%% file: main_arxiv2.tex
\documentclass[10pt,twocolumn,letterpaper]{article}

\usepackage{cvpr}              %

\usepackage{times}
\usepackage{graphicx}
\usepackage{amsmath}
\usepackage{amssymb}
\usepackage{pifont}
\usepackage{array}
\usepackage{comment} 
\usepackage{capt-of}
\usepackage{booktabs}
\usepackage{tabularx}
\usepackage{soul}
\usepackage{multirow}
\usepackage{multicol}
\usepackage[dvipsnames]{xcolor}
\usepackage{xhfill}

\usepackage[toc,page,header]{appendix}
\usepackage{minitoc}

\usepackage[pagebackref,breaklinks,colorlinks]{hyperref}

\newcommand{\sketch}{\mathcal{S}}
\newcommand{\image}{\mathcal{I}}
\newcommand{\renderer}{\mathcal{R}}

\newcommand{\cliploss}[2]{CLIP_{\ell_{#1}}(\mathcal{#2})}

\usepackage[capitalize]{cleveref}
\crefname{section}{Sec.}{Secs.}
\Crefname{section}{Section}{Sections}
\Crefname{table}{Table}{Tables}
\crefname{table}{Tab.}{Tabs.}

\makeatletter
\DeclareRobustCommand\onedot{\futurelet\@let@token\@onedot}
\def\@onedot{\ifx\@let@token.\else.\null\fi\xspace}
\def\eg{\emph{e.g}\onedot}

\def\ie{\emph{i.e}\onedot}

\def\etal{\emph{et al}\onedot}

\def\Bezier{B\'{e}zier\xspace}
\def\etal{\emph{et al}\onedot}
\makeatother
\let\shortcite\cite

\begin{document}

\title{CLIPascene: Scene Sketching with Different Types and Levels of Abstraction}

\author{
Yael Vinker\\
Tel Aviv University
\and
Yuval Alaluf\\
Tel Aviv University
\and
Daniel Cohen-Or\\
Tel Aviv University
\and
Ariel Shamir\\
Reichman University
\and\small\url{https://clipascene.github.io/CLIPascene/}
}

\doparttoc %
\faketableofcontents %

\twocolumn[{%
\vspace{-1em}
\maketitle
\renewcommand\twocolumn[1][]{#1}%
\begin{center}
    \centering
        \includegraphics[width=0.98\textwidth]{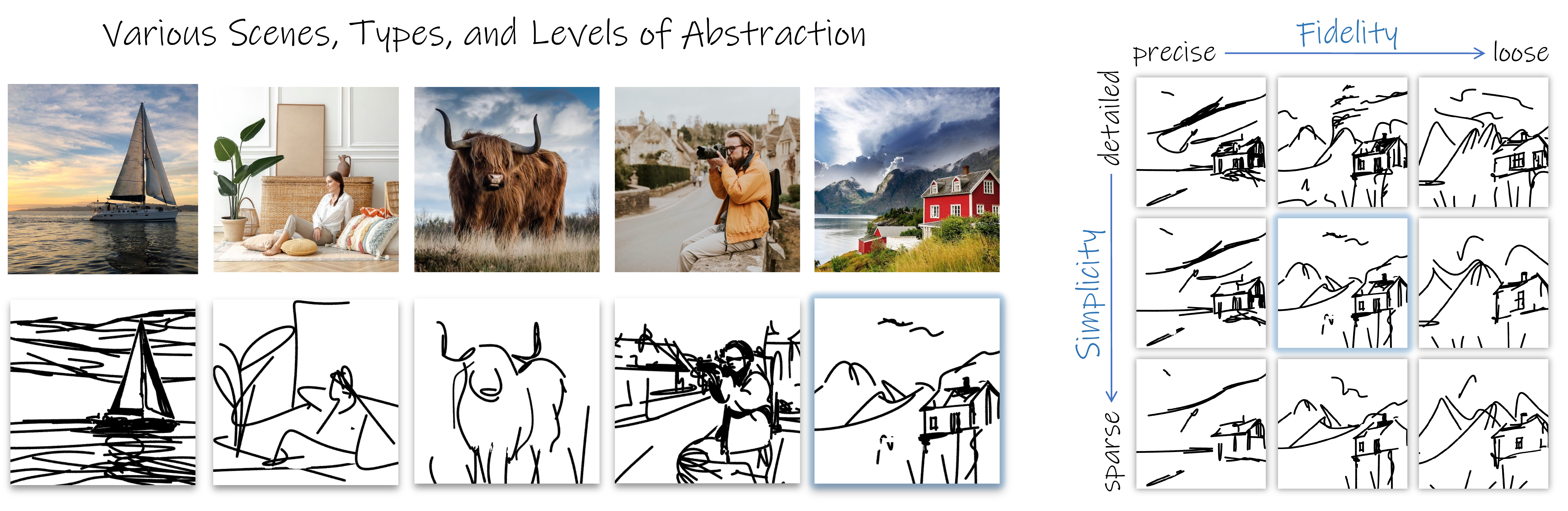}
        \captionof{figure}{Our method converts a \textit{scene} image into a sketch  with different types and levels of abstraction by disentangling abstraction into two axes of control: \textit{fidelity} and \textit{simplicity}. 
        The sketches on the left were selected from a complete \textit{matrix} generated by our method (an example is shown on the right), encompassing a broad range of possible sketch abstractions for a given image. Our sketches are generated in vector form, which can be easily used by designers for further editing.}
    \label{fig:teaser}
\end{center}%
}]

\begin{abstract}
\vspace{-0.5cm}
\input{files/abstract}
\end{abstract}

\input{files/shortIntro}

\input{files/prev_work}

\input{files/method_new}

\input{files/results}

\input{files/limitations}

{\small
\bibliographystyle{ieee_fullname}
\bibliography{bibliography}
}

\clearpage
\appendix
\appendixpage
\addcontentsline{toc}{section}{} %
\part{} %
\vspace{-3em}
\parttoc %
\input{supplementary_arxiv.tex}

\end{document}

%% file: files/abstract.tex
In this paper, we present a method for converting a given \textit{scene} image into a sketch using different \emph{types} and multiple \emph{levels} of abstraction. We distinguish between two types of abstraction.
The first considers the \textit{fidelity} of the sketch, varying its representation from a more precise portrayal of the input to a looser depiction.
The second is defined by the visual \textit{simplicity} of the sketch, moving from a detailed depiction to a sparse sketch.
Using an explicit disentanglement into two abstraction axes --- and multiple levels for each one --- provides users additional control over selecting the desired sketch based on their personal goals and preferences.
To form a sketch at a given level of fidelity and simplification, we train two MLP networks. The first network learns the desired placement of strokes, while the second network learns to gradually remove strokes from the sketch without harming its recognizability and semantics.
Our approach is able to generate sketches of complex scenes including those with complex backgrounds (\eg natural and urban settings) and subjects (\eg animals and people) while depicting gradual abstractions of the input scene in terms of fidelity and simplicity.

%% file: files/shortIntro.tex
\section{Introduction}
Several studies have demonstrated that abstract, minimal representations are not only visually pleasing but also helpful in conveying an idea more effectively by emphasizing the essence of the subject~\cite{Hertzmann_2020,biederman1988surface}. In this paper, we concentrate on converting photographs of natural scenes to sketches as a prominent minimal representation. 

\begin{figure}
\setlength{\belowcaptionskip}{-12pt}
  \centering
  \includegraphics[width=0.925\linewidth]{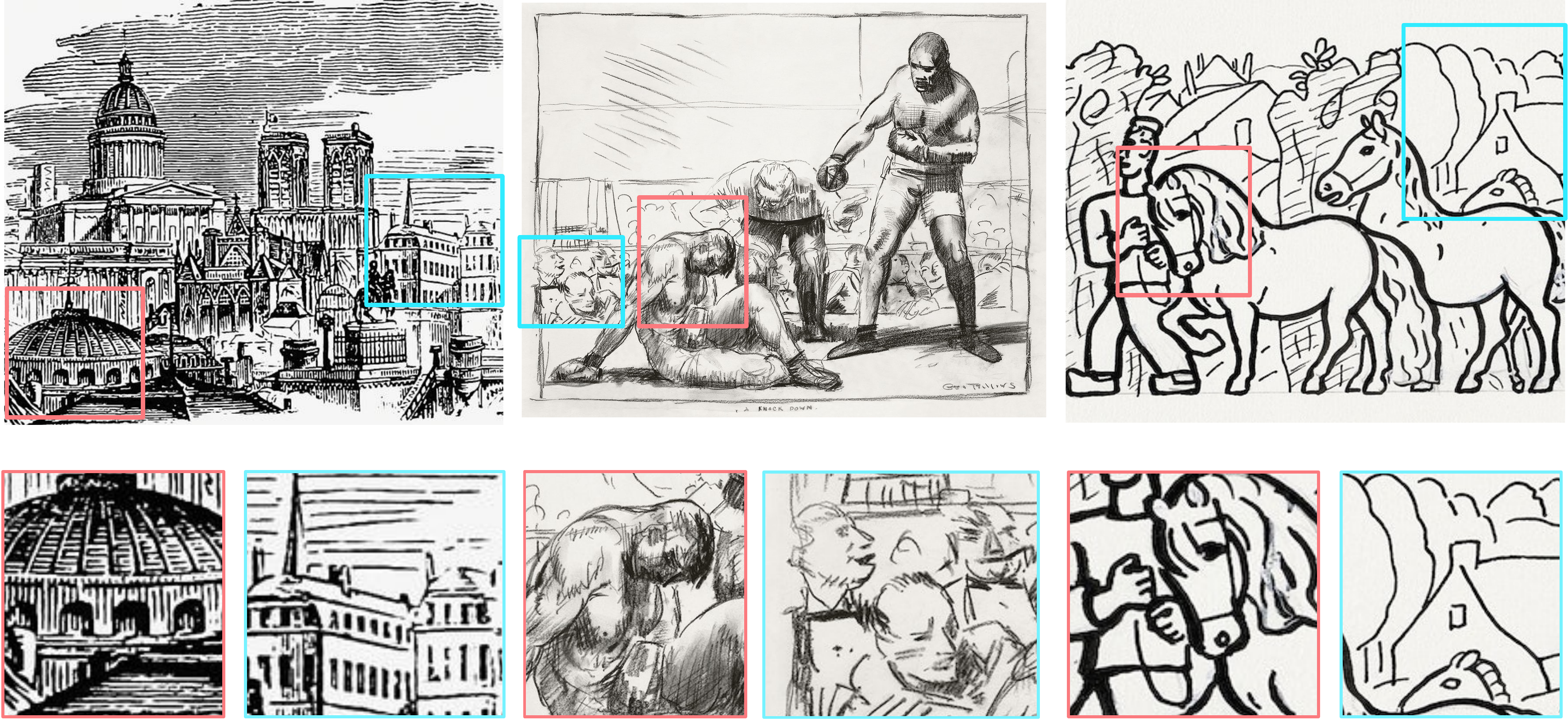}
  \caption{Drawings of different scenes by different artists. Notice the significant differences in style and level of abstraction between the drawings --- moving from more detailed and precise (left) to more abstract (right).
  The second row shows how the level of abstraction not only varies \textit{between} drawings, but also \textit{within} the \textit{same} drawing. Where each drawing contains areas that are relatively more detailed (red) and more abstract (blue).}
 \label{fig:drawing_style_lob}
\end{figure}

Converting a photograph to a sketch involves abstraction, which requires the ability to understand, analyze, and interpret the complexity of the visual scene. 
A scene consists of multiple objects of varying complexity, as well as relationships between the foreground and background (see~\Cref{fig:im_complex}). 
Therefore, when sketching a scene, the artist has many options regarding how to express the various components and the relations between them (see~\Cref{fig:drawing_style_lob}).

In a similar manner, computational sketching methods must deal with scene complexity and consider a variety of abstraction levels. Our work focuses on the challenging task of \emph{scene} sketching while doing so using different \emph{types} and multiple \emph{levels} of abstraction. 
Only a few previous works attempted to produce sketches with multiple levels of abstraction.
However, these works focus specifically on the task of \emph{object} sketching~\cite{vinker2022clipasso,Deep-Sketch-Abstraction} or \emph{portrait} sketching~\cite{Berger2013}, and often simply use the number of strokes to define the level of abstraction. 
We are not aware of any previous work that attempts to separate different \emph{types} of abstractions.
Moreover, existing works for \emph{scene} sketching often focus on producing sketches based on certain styles, without taking into account the abstraction level, which is an essential concept in sketching. Lastly, most existing methods for scene sketching do not produce sketches in vector format. Providing vector-based sketches is a natural choice for sketches as it allows further editing by designers (such as in~\cref{fig:svg_editing}).
\input{files/figures/complexity_level}

\begin{figure}[t]
  \centering
  \setlength{\belowcaptionskip}{-7pt}
  \includegraphics[width=1\linewidth]{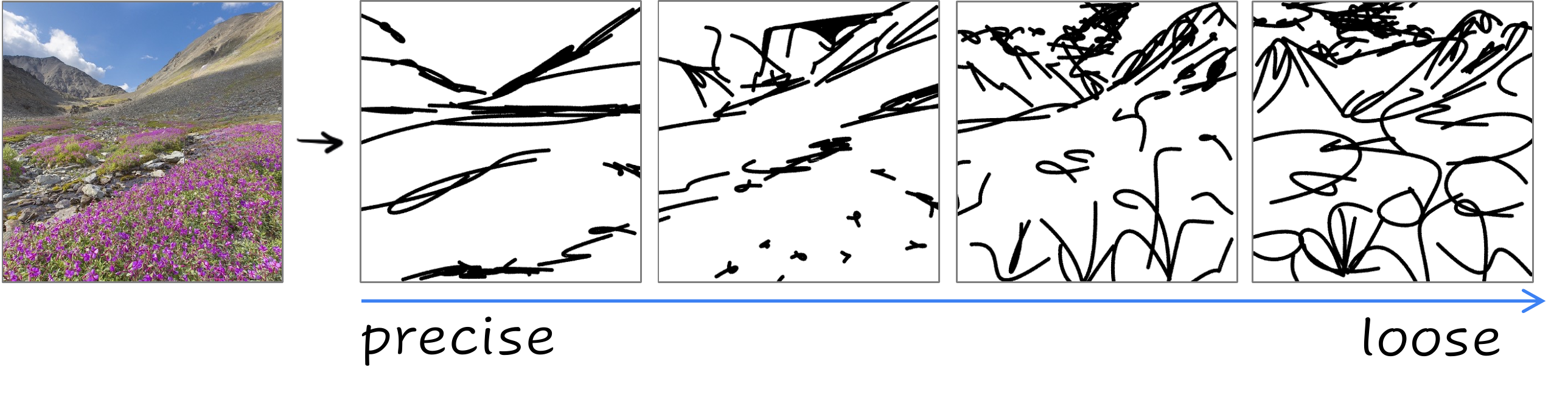}
  \caption{The \emph{fidelity} axis. From left to right, using the same number of strokes the sketches gradually depart from the geometry of the input image, but still convey the semantics of the scene.}
 \label{fig:semantic_axis}
\end{figure}

\begin{figure}[t]
  \centering
  \includegraphics[width=1\linewidth]{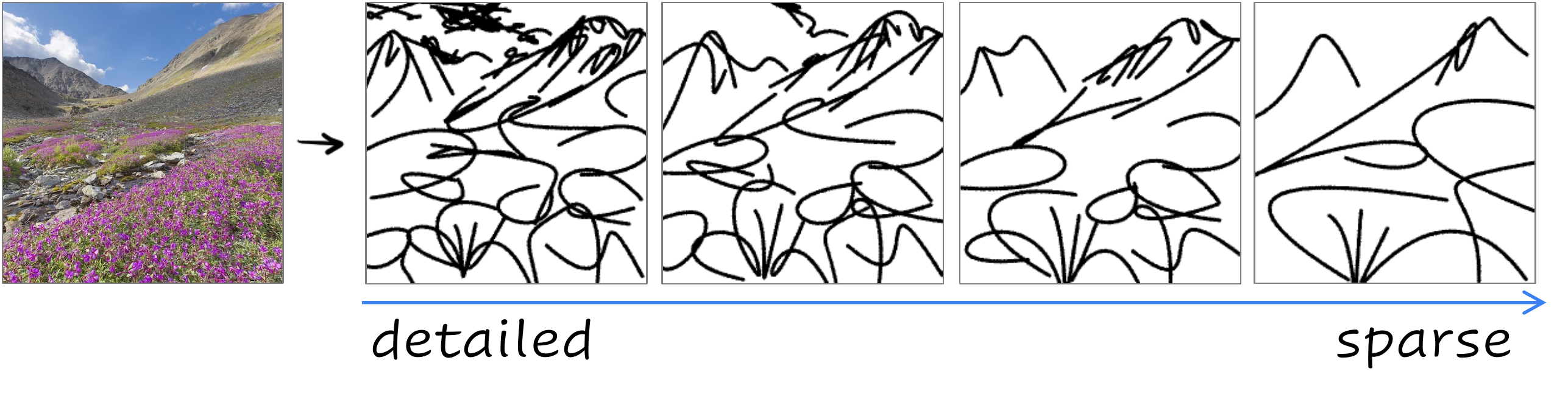}
  \vspace{-0.3cm}
  \caption{The \emph{simplicity} axis. On the left, we start with a more detailed sketch and as we move to the right the sketch is gradually simplified while still remaining consistent with the overall appearance of the initial sketch.}
   \vspace{-0.3cm}
 \label{fig:sparse_axis}
\end{figure}

We define two axes representing two types of abstractions and produce sketches by gradually moving along these axes.
The first axis governs the \emph{fidelity} of the sketch. This axis moves from more precise sketches, where the sketch composition follows the geometry and structure of the photograph to more loose sketches, where the composition relies more on the semantics of the scene. An example is shown in~\Cref{fig:semantic_axis}, where the leftmost sketch follows the contours of the mountains on the horizon, and as we move right, the mountains and the flowers in the front gradually deviate from the edges present in the input, but still convey the correct semantics of the scene.
The second axis governs the level of details of the sketch and moves from detailed to sparse depictions, which appear more abstract. Hence, we refer to this axis as the \textit{simplicity} axis.
An example can be seen in \Cref{fig:sparse_axis}, where the same general characteristics of the scene (\eg the mountains and flowers) are captured in all sketches, but with gradually fewer details.

To deal with scene complexity, we separate the foreground and background elements and sketch each of them separately. This explicit separation and the disentanglement into two abstraction axes provide a more flexible framework for computational sketching, where users can choose the desired sketch from a range of possibilities, according to their goals and personal taste.

We define a sketch as a set of B\'{e}zier\xspace curves, and train a simple multi-layer perceptron (MLP) network to learn the stroke parameters.
Training is performed per image (\eg without an external dataset) and is guided by a pre-trained CLIP-ViT model~\cite{CLIP,dosovitskiy2020image}, leveraging its powerful ability to capture the semantics and global context of the entire scene.

To realize the \emph{fidelity} axis, we utilize different intermediate layers of CLIP-ViT to guide the training process, where shallow layers preserve the geometry of the image and deeper layers encourage the creation of looser sketches that emphasize the scene's semantics. 

To realize the \emph{simplicity} axis, we jointly train an additional MLP network that learns how to best discard strokes gradually and smoothly, without harming the recognizability of the sketch.
As shall be discussed, the use of the networks over a direct optimization-based approach allows us to define the level of details \textit{implicitly} in a learnable fashion, as opposed to explicitly determining the number of strokes.

The resulting sketches demonstrate our ability to cope with various scenes and to capture their core characteristics while providing gradual abstraction along both the fidelity and simplicity axes, as shown in~\Cref{fig:teaser}.
We compare our results with existing methods for scene sketching. 
We additionally evaluate our results quantitatively and demonstrate that the generated sketches, although abstract, successfully preserve the geometry and semantics of the input scene.

%% file: files/figures/complexity_level.tex
\begin{figure}[t]
    \centering
    \setlength{\tabcolsep}{1.5pt}
    {\small
    \begin{tabular}{c c c}
        \includegraphics[width=0.2\linewidth]{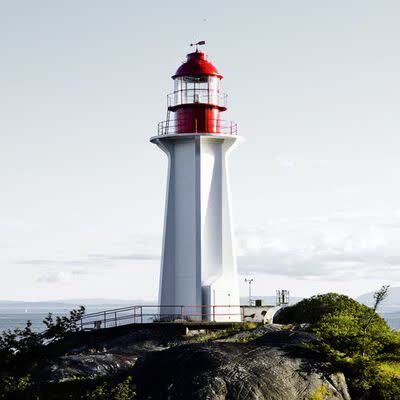} &
        \includegraphics[width=0.2\linewidth]{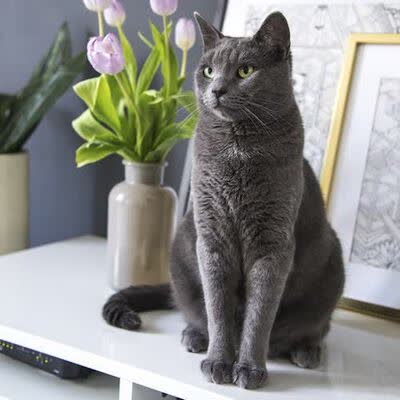} &
        \includegraphics[width=0.2\linewidth]{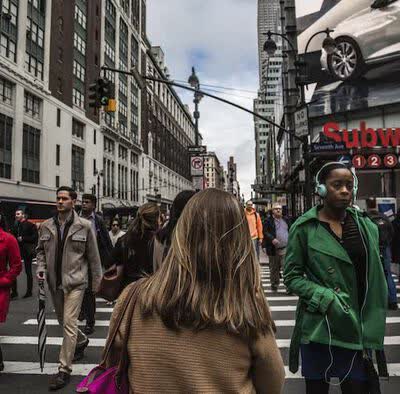} \\
        A & B & C
    \end{tabular}
    }
    \vspace{0.05cm}
    \caption{Scene complexity. (A) contains a single, central object with a simple background, (B) contains multiple objects (the cat and vase) with a slightly more complicated background, and (C) contains both foreground and background that include many details. Our work tackles all types of scenes.}
    \vspace{-0.2cm}
    \label{fig:im_complex}
\end{figure}

%% file: files/prev_work.tex
\begin{figure}
    \centering
    \includegraphics[width=0.9\linewidth]{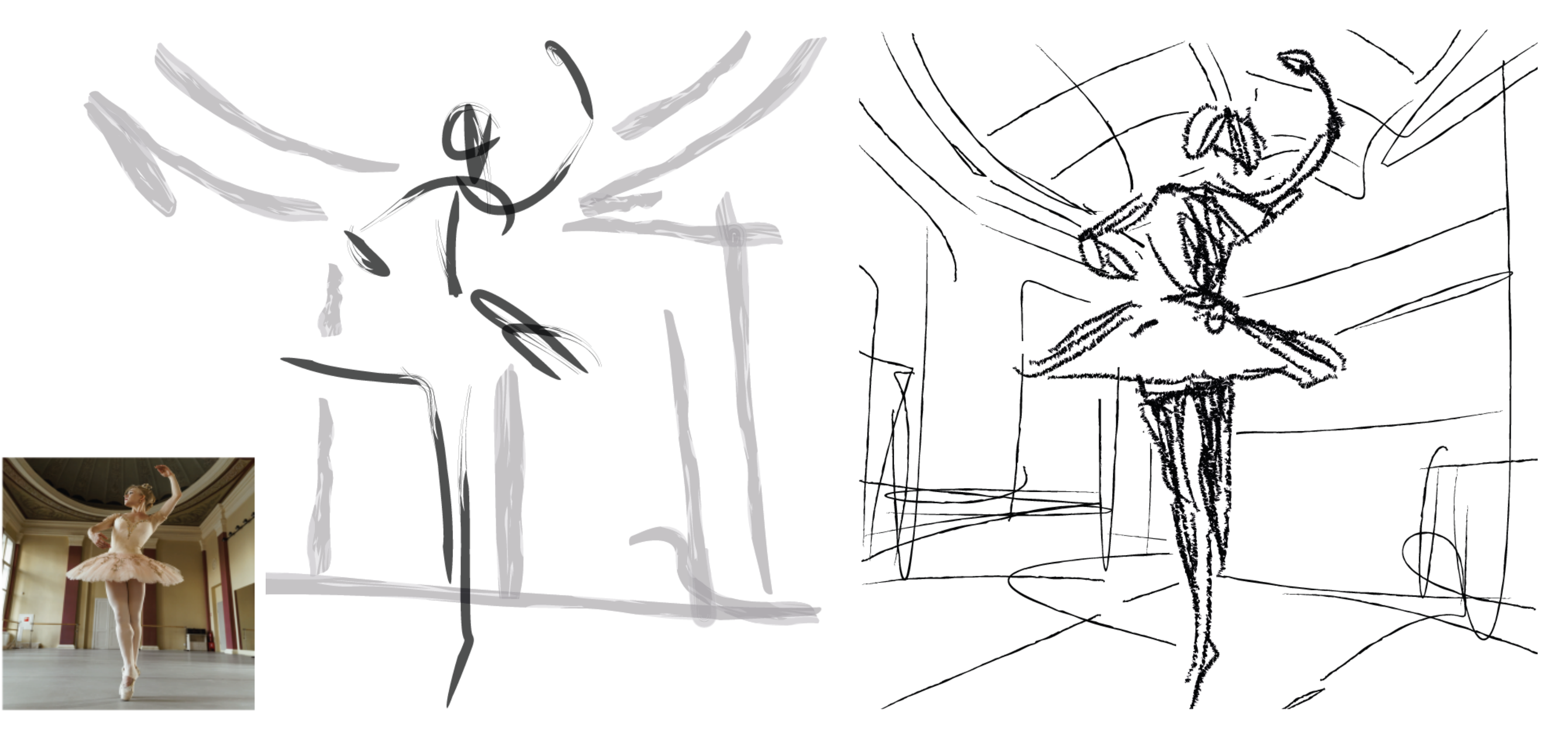}
    \caption{Artistic stylization of the strokes using Adobe Illustrator.}
    \vspace{-0.4cm}
    \label{fig:svg_editing}
\end{figure}

\section{Related Work}
Free-hand sketch generation differs from edge-map extraction \cite{canny1986computational, Winnemller2012XDoGAE} in that it attempts to produce sketches that are representative of the style of human drawings. Yet, there are significant differences in drawing styles among individuals depending on their goals, skill levels, and more (see~\Cref{fig:drawing_style_lob}). As such, computational sketching methods must consider a wide range of sketch representations.

This ranges from methods aiming to produce sketches that are grounded in the edge map of the input image~\cite{li2019photo,xie2015holistically,tong2021sketch,Deformable_Stroke}, to those that aim to produce sketches that are more abstract \cite{bhunia2021doodleformer,CLIPDraw,vinker2022clipasso,ha2017neural,qi2021sketchlattice,ge2021creative,qiu2021emergent,mihai2021learning,Zhou2018LearningTS}. Several works have attempted to develop a unified algorithm that can output sketches with a variety of styles \cite{chan2022learning,yi2020unpaired,liu2021neural}. There are, however, only a few works that attempt to provide various levels of abstraction~\cite{Berger2013,Deep-Sketch-Abstraction,vinker2022clipasso}.
In the following, we focus on scene-sketching approaches, and we refer the reader to \cite{xu2022deep} for a comprehensive survey on computational sketching techniques.

\vspace{-0.35cm}
\paragraph{Photo-Sketch Synthesis}
Various works formulate this task as an image-to-image translation task using paired data of corresponding images and sketches~\cite{li2019im2pencil,yi2019apdrawinggan,li2019photo,artline}. 
Others approach the translation task via unpaired data, often relying on a cycle consistency constraint~\cite{yi2020unpaired,song2018learning,chan2022learning}.
Li~\etal~\shortcite{li2019photo} introduce a GAN-based contour generation algorithm and utilize multiple ground truth sketches to guide the training process.
Yi~\etal~\shortcite{yi2020unpaired} generate portrait drawings with unpaired data by employing a cycle-consistency objective and a discriminator trained to learn a specific style.

Recently, Chan~\etal~\shortcite{chan2022learning} propose an unpaired GAN-based approach. They train a generator to map a given image into a sketch with multiple styles defined explicitly from four existing sketch datasets with a dedicated model trained for each desired style. 
They utilize a CLIP-based loss to achieve semantically-aware sketches.
As these works rely on curated datasets, they require training a new model for each desired style while supporting a single level of sketch abstraction.
In contrast, our approach does not rely on any explicit dataset and is not limited to a pre-defined set of styles. Instead, we leverage the powerful semantics captured by a pre-trained CLIP model~\cite{CLIP}.
Additionally, our work is the only one among the alternative scene sketching approaches that provides sketches with multiple levels of abstraction and in vector form, which allows for a wider range of editing and manipulation.

\vspace{-0.3cm}
\paragraph{Sketch Abstraction}
While abstractions are fundamental to sketches, only a few works have attempted to create sketches at multiple levels of abstraction, while no previous works have done so over an entire scene.
Berger~\etal~\shortcite{Berger2013} collected portrait sketches at different levels of abstraction from seven artists to learn a mapping from a face photograph to a portrait sketch. Their method is limited to faces only and requires a new dataset for each desired level of abstraction. Muhammad~\etal~\shortcite{Deep-Sketch-Abstraction} train a reinforcement learning agent to remove strokes from a given sketch without harming the sketch's recognizability. The recognition signal is given by a sketch classifier trained on nine classes from the QuickDraw dataset~\cite{ha2017neural}. Their method is therefore limited only to objects from the classes seen during training and requires extensive training.

\vspace{-0.3cm}
\paragraph{CLIPasso}
Most similar to our work is CLIPasso~\cite{vinker2022clipasso} which was designed for \textit{object} sketching at multiple levels of abstraction.
They define a sketch as a set of \Bezier curves and optimize the stroke parameters with respect to a CLIP-based \cite{CLIP} similarity loss between the input image and generated sketch. Multiple levels of abstraction are realized by reducing the number of strokes used to compose the sketch.
In contrast to CLIPasso, our method is not restricted to objects and can handle the challenging task of \textit{scene} sketching. 
Additionally, while Vinker~\etal examine only a single form of abstraction, we disentangle abstraction into two distinct axes controlling both the simplicity and the fidelity of the sketch.
Moreover, in CLIPasso, the user is required to \textit{explicitly} define the number of strokes needed to obtain the desired abstraction. However, different images require a different number of strokes, which is difficult to determine in advance. In contrast, we \textit{implicitly} learn the desired number of strokes by training two MLP networks to achieve a desired trade-off between simplicity and fidelity with respect to the input image.

%% file: files/method_new.tex
\setlength{\abovedisplayskip}{6pt}
\setlength{\belowdisplayskip}{6pt}

\section{Method}~\label{sec:method}
Given an input image $\mathcal{I}$ of a scene, our goal is to produce a set of corresponding sketches at $n$ levels of \textit{fidelity} and $m$ levels of \textit{simplicity},  forming a sketch abstraction matrix of size $m \times n$.
We begin by producing a set of sketches along the \textit{fidelity} axis (\Cref{sec:training_scheme,sec:sem_abs}) with no simplification, thus forming the top row in the abstraction matrix. Next, for each sketch at a given level of fidelity, we perform an iterative visual \textit{simplification} by learning how to best remove select strokes and adjust the locations of the remaining strokes (\Cref{sec:sketch_simp}).
For clarity, in the following we describe our method taking into account the entire scene as a whole. 
However, to allow for greater control over the appearance of the output sketches, and to tackle the high complexity presented in a whole scene, our final scheme splits the image into two regions -- the salient foreground object(s), and the background. We apply our 2-axes abstraction method to each region separately, and then combine them to form the matrix of sketches (details in \Cref{sec:split_for_back}).

\vspace{-0.1cm}
\subsection{Training Scheme}~\label{sec:training_scheme}
We define a sketch as a set of $n$ strokes placed over a white background, where each stroke is a two-dimensional \Bezier curve with four control points.
We mark the $i$-th stroke by its set of control points $z_i = \{(x_i,y_i)^j\}_{j=1}^4$, and denote the set of the $n$ strokes by $Z=\{z_i\}_{i=1}^n$. Our goal is to find the set of stroke parameters that produces a sketch adequately depicting the input scene image.

An overview of our training scheme used to produce a single sketch image is presented in the gray area of~\Cref{fig:pipeline_mlp}.
We train an MLP network, denoted by $MLP_{loc}$, that receives an initial set of control points $Z_{init} \in R^{n \times 4 \times 2}$ (marked in blue) and returns a vector of offsets $MLP_{loc}(Z_{init}) = \Delta Z \in \mathbb{R}^{n \times 4 \times 2}$ with respect to the initial stroke locations. The final set of control points are then given by $Z = Z_{init} + \Delta Z$, which are then passed to a differentiable rasterizer $\mathcal{R}$~\cite{diffvg} that outputs the rasterized sketch,
\begin{equation}
    \mathcal{S} = \renderer(Z_{init} + \Delta Z).
\end{equation}

For initializing the locations of the $n$ strokes, we follow the saliency-based initialization introduced in Vinker~\etal~\shortcite{vinker2022clipasso}, in which, strokes are initialized in salient regions based on a relevancy map extracted automatically~\cite{Chefer_2021_ICCV}.

To guide the training process, we leverage a pre-trained CLIP model due to its capabilities of encoding shared information from both sketches and natural images.
As opposed to Vinker~\etal~\shortcite{vinker2022clipasso} that use the ResNet-based~\cite{he2016deep} CLIP model for the sketching process (and struggles with depicting a scene image),
we find that the ViT-based~\cite{dosovitskiy2020image} CLIP model is able to capture the global context required for generating a coherent sketch of a whole scene, including both foreground and background. This also follows the observation of Raghu~\etal~\shortcite{raghu2021vision} that ViT models better capture more global information at lower layers compared to ResNet-based models. We further analyze this design choice in the supplementary material.

The loss function is then defined as the L2 distance between the activations of CLIP on the image $\mathcal{I}$ and sketch $\mathcal{S}$ at a layer $\ell_k$:
\begin{equation}~\label{eq:clip_loss}
    \mathcal{L}_{CLIP}(\sketch,\image,\ell_k) = \big{\|}\ \cliploss{k}{\sketch} - \cliploss{k}{\image} \big{\|}_2^2. 
\end{equation}
At each step during training, we back-propagate the loss through the CLIP model and the differentiable rasterizer $\renderer$ whose weights are frozen, and only update the weights of $MLP_{loc}$. This process is repeated iteratively until convergence. Observe that no external dataset is needed for guiding the training process, as we rely solely on the expressiveness and semantics captured by the pre-trained CLIP model.
This training scheme produces a \textit{single} sketch image at a \textit{single} level of fidelity and simplicity. Below, we describe how to control these two axes of abstraction.

\input{files/figures/mlp}

\subsection{Fidelity Axis}
\label{sec:sem_abs}
To achieve different levels of fidelity, as illustrated by a single row in our abstraction matrix, we select different activation layers of the CLIP-ViT model for computing the loss defined in~\Cref{eq:clip_loss}. 
Optimizing via deeper layers leads to sketches that are more semantic in nature and do not necessarily confine to the precise geometry of the input. 
Specifically, in all our examples we train a separate $MLP_{loc}$ using layers $\{\ell_2,\ell_7,\ell_8,\ell_{11}\}$ of CLIP-ViT and set the number of strokes to $n=64$.
Note that it is possible to use the remaining layers to achieve additional fidelity levels (see the supplementary material).

\subsection{Simplicity Axis}
\label{sec:sketch_simp}
Given a sketch $\sketch_k$ at fidelity level $k$, our goal is to find a set of sketches $\{\sketch_k^1, ..., \sketch_k^m\}$ that are visually and conceptually similar to $\sketch_k$ but have a gradually simplified appearance. In practice, we would like to learn how to best remove select strokes from a given sketch and refine the locations of the remaining strokes without harming the overall recognizability of the sketch. 

We illustrate our sketch simplification scheme for generating a single simplified sketch $\sketch_k^j$ in the bottom left region of ~\Cref{fig:pipeline_mlp}. We train an additional network, denoted as $MLP_{simp}$ (marked in orange), that receives a random-valued vector and is tasked with learning an $n$-dimensional vector $P = \{p_i\}_{i=1}^n$, where $p_i\in[0,1]$ represents the probability of the $i$-th stroke appearing in the rendered sketch.
$P$ is passed as an additional input to $\renderer$ which outputs the simplified sketch $\sketch_k^j$ in accordance.

To implement the probabilistic-based removal or addition of strokes (which are discrete operations) into our learning framework, we multiply the width of each stroke $z_i$ by $p_i$. When rendering the sketch, strokes with a very low probability will be ``hidden'' due to their small width.

Similar to Mo~\etal~\cite{mo2021virtualsketching}, to encourage a sparse representation of the sketch (\ie one with fewer strokes) we minimize the normalized L1 norm of $P$:
\begin{equation}~\label{eq:struct_loss}
    \mathcal{L}_{sparse}(P) = \frac{\|P\|_1}{n}.
\end{equation}

To ensure that the resulting sketch still resembles the original input image, we additionally minimize the $\mathcal{L}_{CLIP}$ loss presented in~\Cref{eq:clip_loss}, and continue to fine-tune $MLP_{loc}$ during the training of $MLP_{simp}$. 
Formally, we minimize the sum:
\begin{equation}
    \mathcal{L}_{CLIP}(\sketch_k^j,\image,\ell_k) + \mathcal{L}_{sparse}(P).
\end{equation}
We back-propagate the gradients from $\mathcal{L}_{CLIP}$ to both $MLP_{loc}$ and $MLP_{simp}$ while $\mathcal{L}_{sparse}$ is used only for training $MLP_{simp}$ (as indicated by the red and purple dashed arrows in~\Cref{fig:pipeline_mlp}).

\input{files/figures/ratio_same_rs}

Note that using the MLP network rather than performing a direct optimization over the stroke parameters (as is done in Vinker~\etal) is crucial as it allows the optimization to restore strokes that may have been previously removed. If we were to use direct optimization, the gradients of deleted strokes would remain removed since they were multiplied by a probability of $0$.

Here, both MLP networks are simple $3$-layer networks with SeLU~\cite{klambauer2017self} activations. For $MLP_{simp}$ we append a Sigmoid activation to convert the outputs to probabilities.

\vspace{-0.4cm}
\paragraph{\textbf{Balancing the Losses.}}
Naturally, there is a trade-off between $\mathcal{L}_{CLIP}$ and $\mathcal{L}_{sparse}$, which affects the appearance of the simplified sketch (see ~\Cref{fig:loss_tradeoff}).
We utilize this trade-off to gradually alter the level of simplicity. 

Finding a balance between $\mathcal{L}_{sparse}$ and $\mathcal{L}_{CLIP}$ is essential for achieving recognizable sketches with varying degrees of abstraction.
Thus, we define the following loss:
\begin{equation}~\label{eq:clip_loss_simp}
    \mathcal{L}_{ratio} = \big{\|}\ \frac{\mathcal{L}_{sparse}}{\mathcal{L}_{CLIP}} - r 
 \big{\|}_2^2,
\end{equation}
where the scalar factor $r$ (denoting the \textit{ratio} of the two losses) controls the strength of simplification.
As we decrease $r$, we encourage the network to output a sparser sketch and vice-versa.
The final objective for generating a single simplified sketch $\sketch_k^j$, is then given by:
\begin{equation}~\label{eq:clip_loss_simp_sum}
   \mathcal{L}_{simp} = \mathcal{L}_{CLIP} + \mathcal{L}_{sparse} + \mathcal{L}_{ratio}.
\end{equation}

To achieve the set of gradually simplified sketches $\{\sketch_k^1, ..., \sketch_k^m\}$, we define a set of corresponding factors $\{r_k^1, ..., r_k^m\}$ to be applied in~\Cref{eq:clip_loss_simp}. 
The first factor $r_k^1$, is derived directly from~\Cref{eq:clip_loss_simp}, aiming to reproduce the strength of simplification present in $\sketch_k$:
\begin{equation}
    r_k^1 = \frac{1}{\mathcal{L}_{CLIP}(\sketch_k,\image,\ell_k)},
\end{equation}
where $\mathcal{L}_{sparse}$ equal to $1$ means that no simplification is performed.
The derivation of the remaining factors $r_k^j$ is described next. 

\vspace{-0.4cm}
\paragraph{\textbf{Perceptually Smooth Simplification}}
\input{files/figures/exponent_fig}

As introduced above, the set of factors determines the strength of the visual simplification. When defining the set of factors $r_k^j$, we aim to achieve a \textit{smooth} simplification. By \textit{smooth} we mean that there is no large change perceptually between two consecutive steps.
This is illustrated in~\Cref{fig:exponent}, where the first row provides an example of \emph{smooth} transitions, and the second row demonstrates a non-smooth transition, where there is a large perceptual ``jump'' in the abstraction level between the second and third sketches, and almost no perceptual change in the following levels.

We find that the simplification appears more smooth when $\mathcal{L}_{sparse}$ is exponential with respect to $\mathcal{L}_{CLIP}$.
The two graphs at the bottom of~\Cref{fig:exponent} describe this observation quantitatively, illustrating the trade-off between $\mathcal{L}_{sparse}$ and $\mathcal{L}_{CLIP}$ for each sketch. 
The smooth transition in the first row forms an exponential relation between $\mathcal{L}_{sparse}$ and $\mathcal{L}_{CLIP}$, while the large ``jump'' in the second row is clearly shown in the right graph.

Given this, we define an exponential function recursively by $f(j) = f(j-1)/2$. The initial value of the function is defined differently for each fidelity level $k$ as $f_k(1) = r_k^1$. To define the following set of factors $\{r_k^2, .. r_k^m\}$ we sample the function $f_k$, where for each $k$, the sampling step size is set proportional to the strength of the $\mathcal{L}_{CLIP}$ loss at level $k$. 
Hence, layers that incur a large $\mathcal{L}_{CLIP}$ value are sampled with a larger step size.
We found this procedure achieves simplifications that are perceptually smooth. 
This observation aligns well with the Weber-Fechner law~\cite{Weber1834-ms,fechnerlaw} which states that human perception is linear with respect to an exponentially-changing signal. 
An analysis of the factors and additional details regarding our design choices are provided in the supplementary material.

\input{files/figures/iterative_simplification}

\vspace{-0.4cm}
\paragraph{\textbf{Generating the Simplified Sketches}}
To generate the set of simplified sketches $\{\sketch_k^1 ... \sketch_k^m\}$, we apply the training procedure iteratively, as illustrated in~\Cref{fig:iterative_simplification}.
We begin with generating $\sketch_k^1$ w.r.t $r_k^1$, by fine-tuning $MLP_{loc}$ and training $MLP_{simp}$ from scratch.
After generating $\sketch_k^1$, we sequentially generate each $\sketch_k^j$ for $2\leq j \leq m$ by continuing training both networks for 500 steps and applying $\mathcal{L}_{ratio}$ with the corresponding factor $r_k^j$.

\input{files/figures/additional_control}

\subsection{Decomposing the Scene}~\label{sec:split_for_back}
The process described above takes the entire scene as a whole. However, in practice, we separate the scene's foreground subject from the background and sketch each of them independently. We use a pretrained U$^2$-Net~\cite{qin2020u2} to extract the salient object(s), and then apply a pretrained LaMa~\cite{suvorov2022resolution} inpainting model to recover the missing regions (see~\Cref{fig:additional_control}, top right).

We find that this separation helps in producing more visually pleasing and stable results. When performing object sketching, we additionally compute $\mathcal{L}_{CLIP}$ over layer $l_4$. This helps in preserving the object's geometry and finer details. On the left part of~\Cref{fig:additional_control} we demonstrate the artifacts that may occur when scene separation is not applied. For example, over-exaggeration of features of the subject at a low fidelity level, such as the panda's face, or, the object might ``blend'' into the background.
Additionally, this explicit separation provides users with more control over the appearance of the final sketches (\Cref{fig:additional_control}, bottom right). For example, users can easily edit the vector file by modifying the brush's style or combine the foreground and background sketches at different levels of abstraction.

%% file: files/figures/mlp.tex
\begin{figure}
    \centering
    \includegraphics[width=1\linewidth]{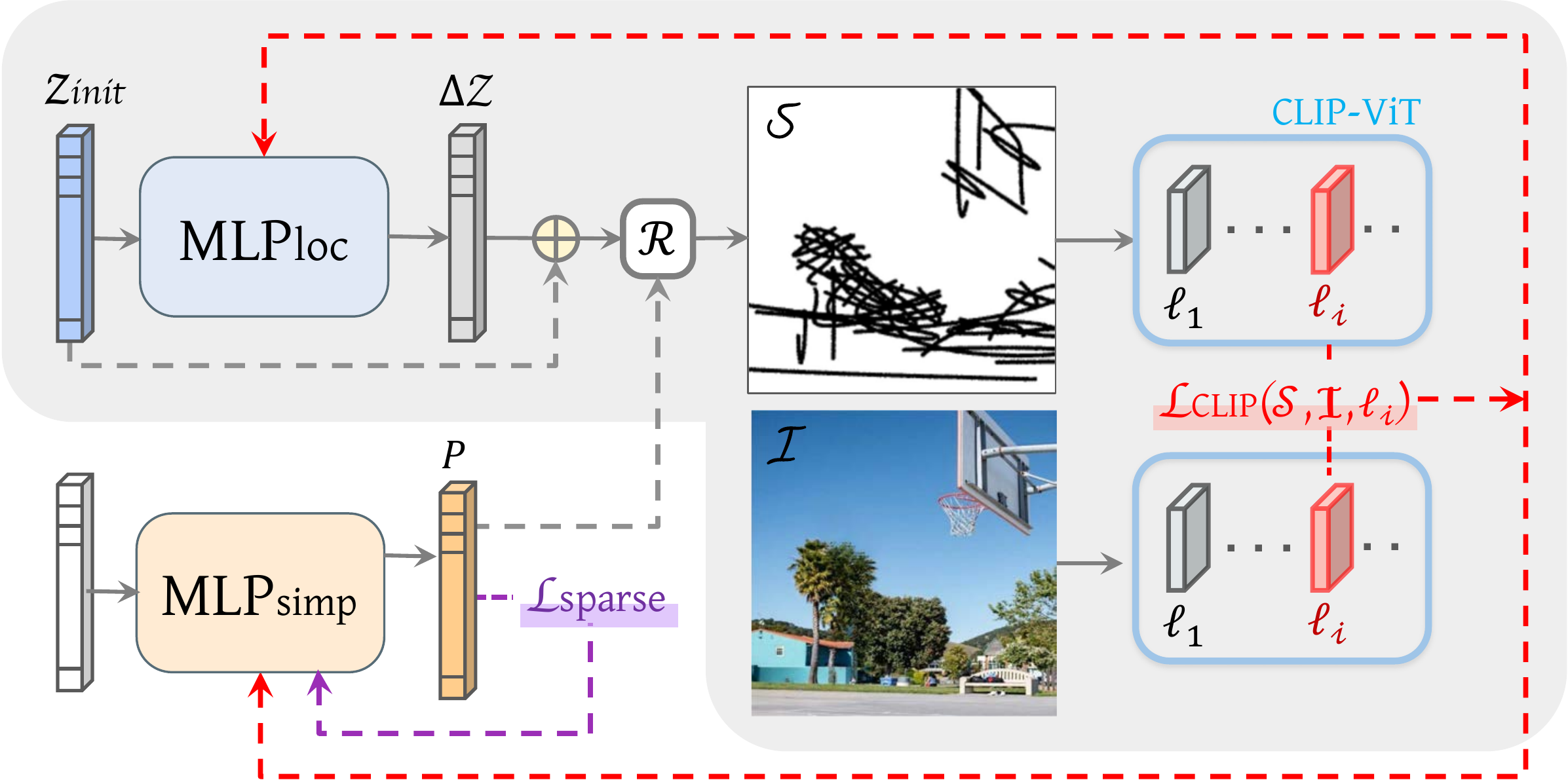}
    \caption{Single sketch generation scheme. 
    In gray, we show our training scheme for producing a single sketch image at a single level of fidelity. 
    In the bottom left we show the additional components used to generate a single sketch at a single level of simplicity.}
    \vspace{-0.4cm}
    \label{fig:pipeline_mlp}
\end{figure}

%% file: files/figures/ratio_same_rs.tex
\begin{figure}
\centering
        \includegraphics[width=0.925\linewidth]{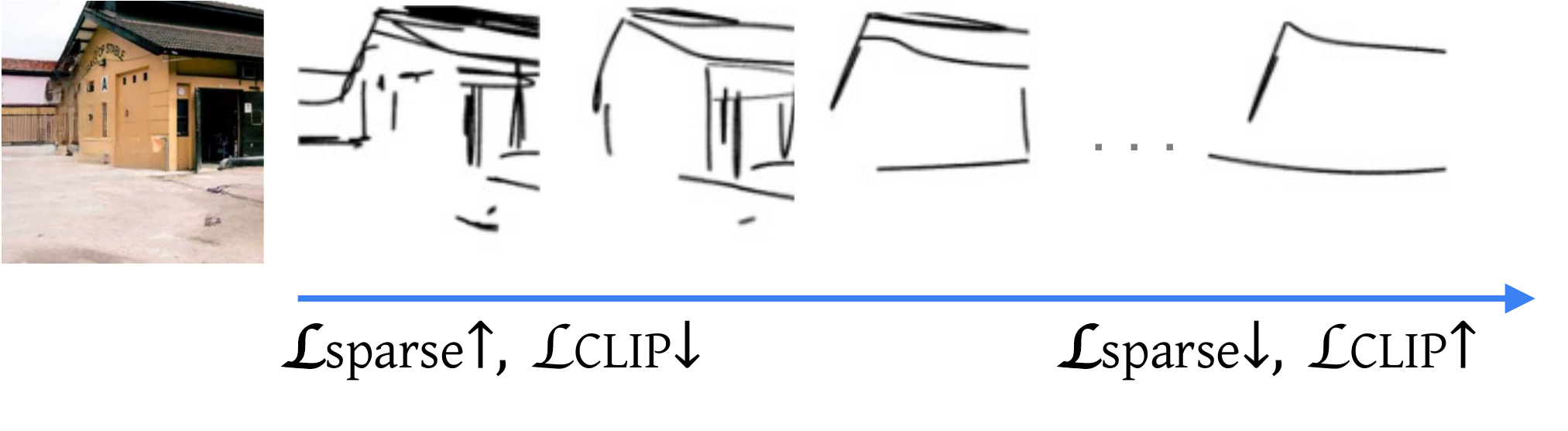}
 \caption{Trade-off between $\mathcal{L}_{sparse}$ and $\mathcal{L}_{CLIP}$. As the sketch becomes sparser, $\mathcal{L}_{sparse}$ obtains lower score. However, the sketch also becomes less recognizable with respect to the input image, resulting in a higher penalty for $\mathcal{L}_{CLIP}$.}
  \vspace{-0.4cm}
         \label{fig:loss_tradeoff}
\end{figure}

%% file: files/figures/exponent_fig.tex
\begin{figure}
    \setlength{\belowcaptionskip}{-4pt}
    \centering
    \includegraphics[width=\linewidth]{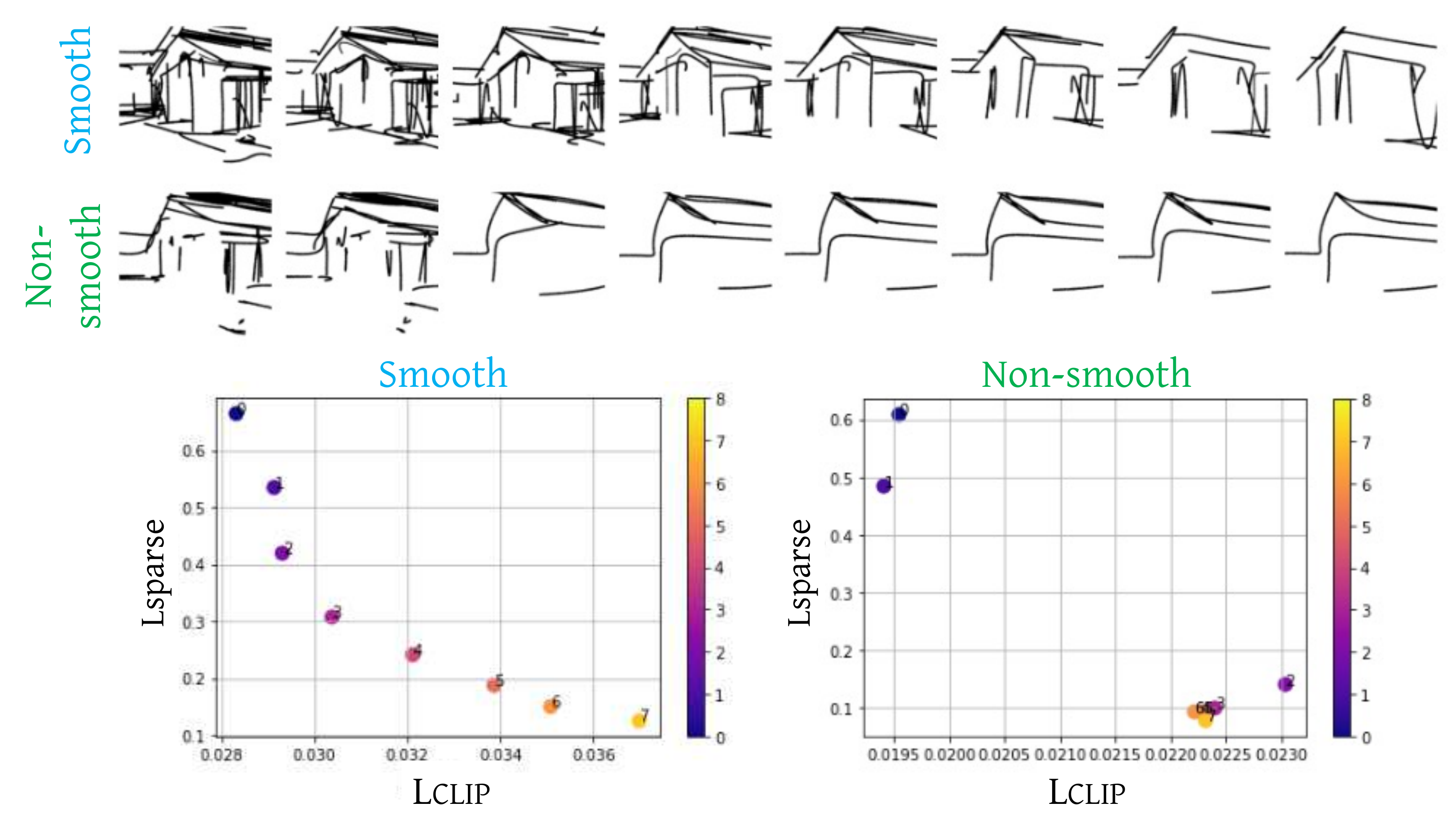}
    \caption{Smooth v.s. non-smooth simplification.
    In the first row, the simplification appears perceptually smooth, where a consistent change in the degree of abstraction is performed. 
    The second row demonstrates a non-smooth simplification, as there is a visible ``jump'' between the second and third sketches. These visual patterns are illustrated quantitatively in the corresponding graphs, where each dot in the graph represents a single sketch.}
    \vspace{-0.4cm}
    \label{fig:exponent}
\end{figure}

%% file: files/figures/iterative_simplification.tex
\begin{figure}
    \centering
    \setlength{\belowcaptionskip}{-5pt}
    \includegraphics[width=0.85\linewidth]{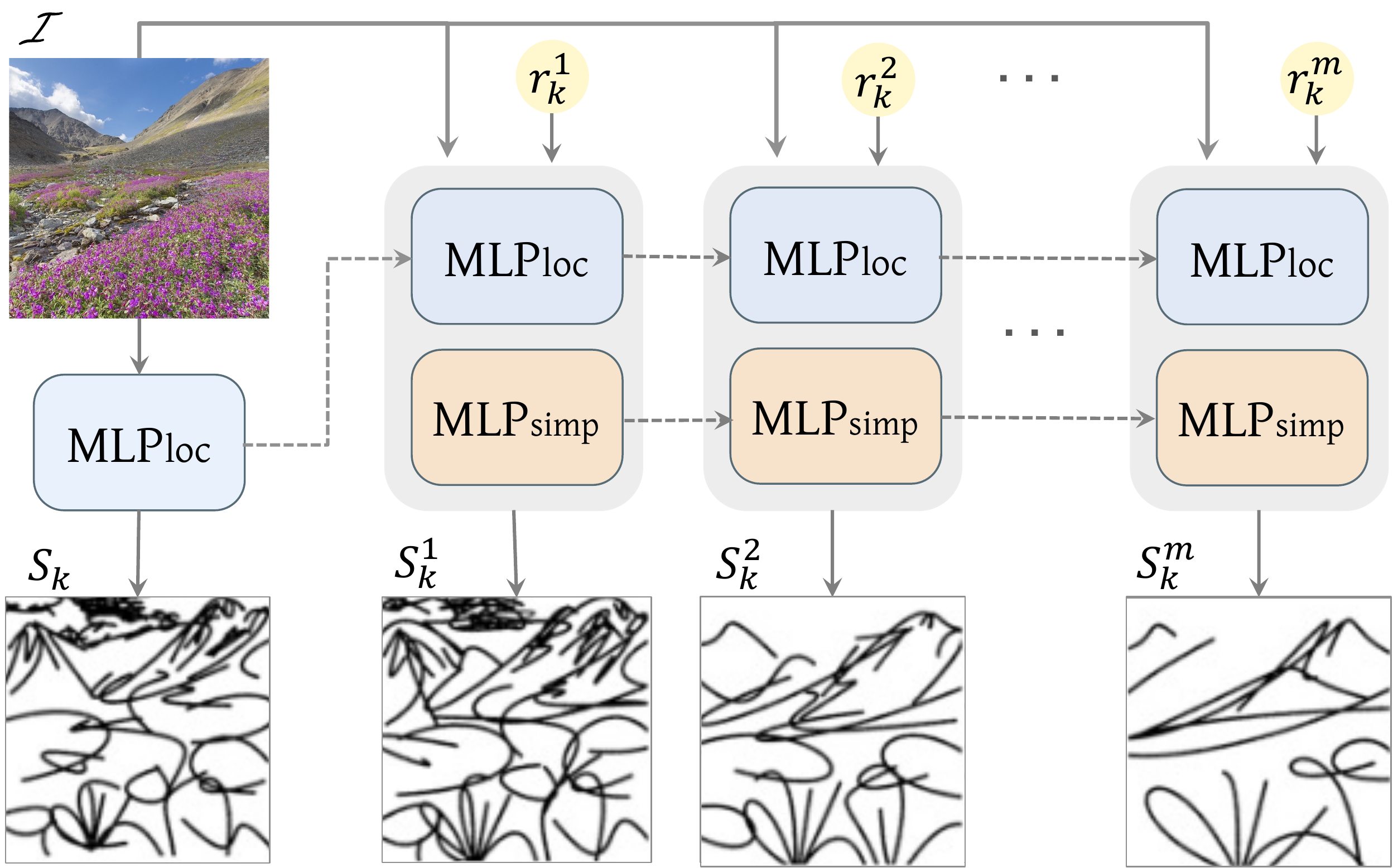}
    \caption{Iterative simplification of the sketch $\sketch_{k}$. 
    To produce a simplified sketch $\sketch_k^j$ we iteratively fine-tune $MLP_{loc}$ (blue) and $MLP_{simp}$ (orange) w.r.t $\mathcal{L}_{ratio}$ loss defined by each $r_k^j$.}
     \vspace{-0.35cm}
    \label{fig:iterative_simplification}
\end{figure}

%% file: files/figures/additional_control.tex
\begin{figure}[t]
    \centering
    \includegraphics[width=0.89\linewidth]{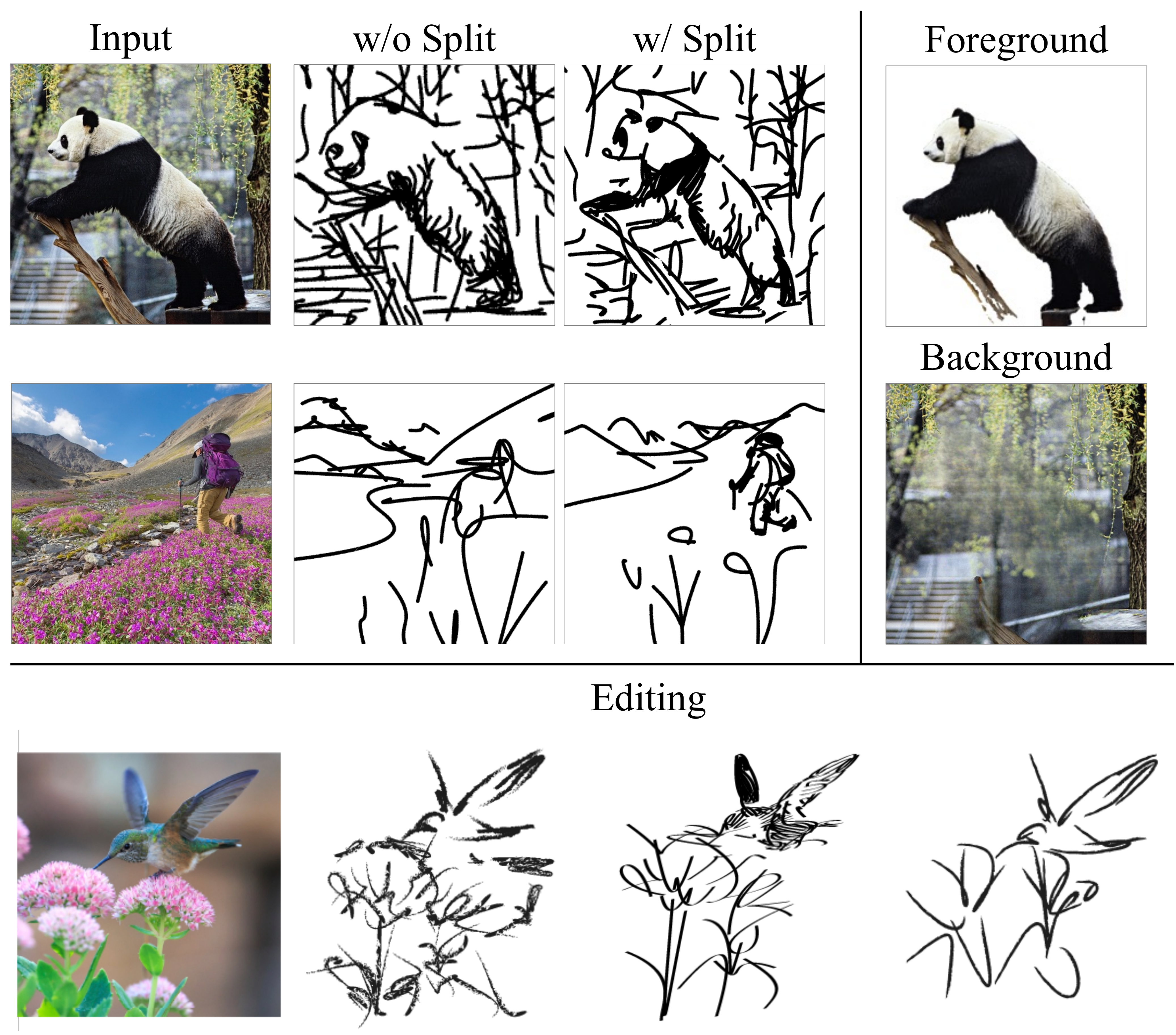}
    \caption{Scene decomposition. Top right -- an example of the separation technique. Left -- scene sketching results obtained with and without decomposing the scene. 
    Bottom -- examples of sketch  editing by modifying the style of strokes.}
     \vspace{-0.4cm}
    \label{fig:additional_control}
\end{figure}

%% file: files/results.tex
\vspace{-0.1cm}
\section{Results}~\label{sec:results}
In the following, we demonstrate the performances of our scene sketching technique qualitatively and quantitatively, and provide comparisons to state-of-the-art sketching methods.
Further analysis, results, and a user study are provided in the supplementary material.

\vspace{-0.1cm}
\subsection{Qualitative Evaluation}
In~\Cref{fig:qualitative_pairs,fig:teaser} we show sketches at different levels of abstraction on various scenes generated by our method. Notice how it is easy to recognize that the sketches are depicting the same scene even though they vary significantly in their abstraction level.

\begin{figure}
    \centering
    \includegraphics[width=0.95\linewidth]{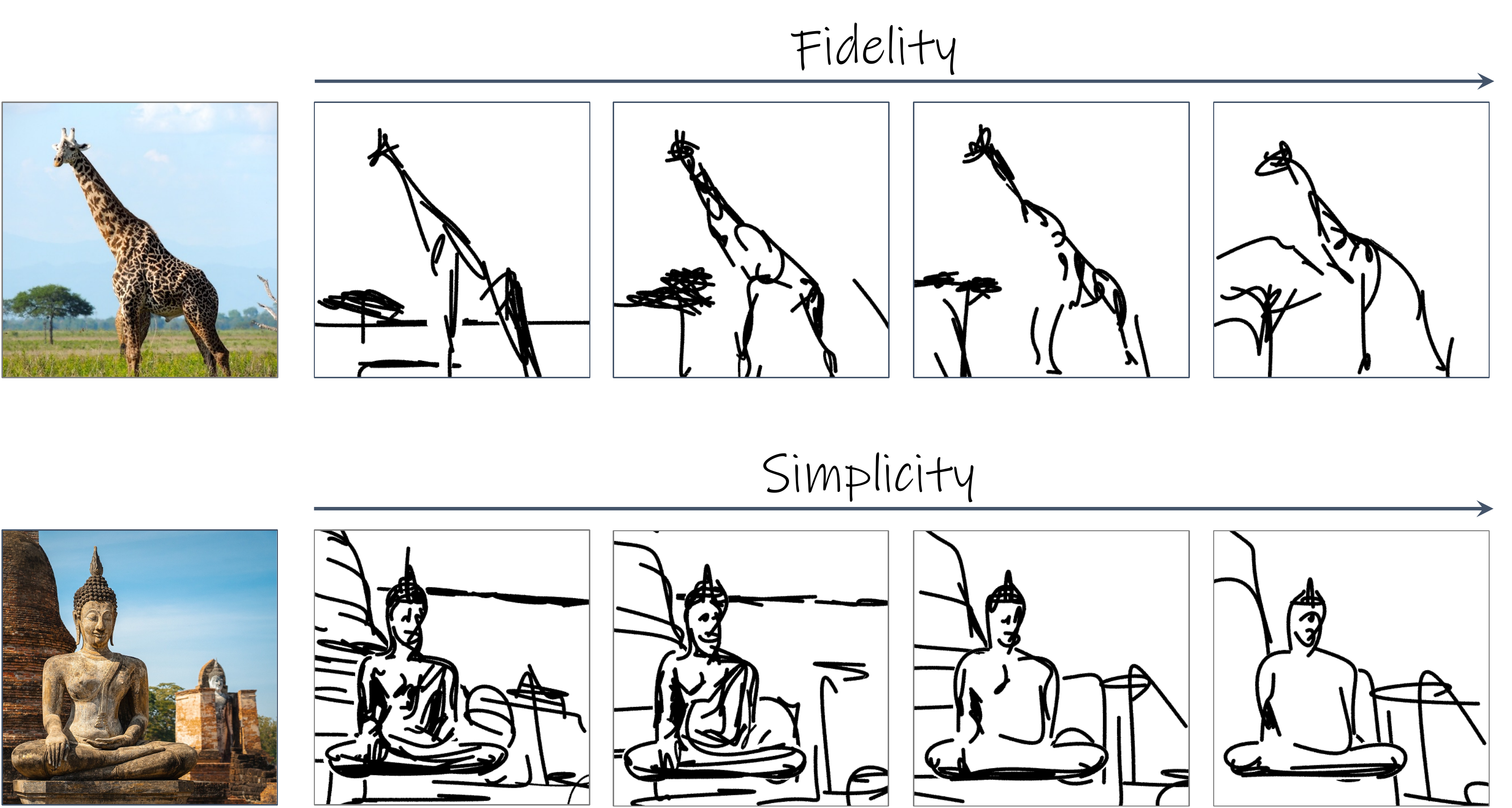}
    \vspace{-0.05cm}
    \caption{Sketches along the two abstraction axes.}
    \vspace{-0.4cm}
    \label{fig:two_axes}
\end{figure}

In~\Cref{fig:teaser,fig:semantic_axis,fig:two_axes} (top) we show sketch abstractions along the \textit{fidelity} axis, where the sketches become less precise as we move from left to right, while still conveying the semantics of the images (for example the mountains in the background in~\Cref{fig:teaser} and the tree and giraffe's body in~\Cref{fig:two_axes}). 
In~\Cref{fig:teaser,fig:sparse_axis,fig:two_axes} (bottom) we show sketch abstractions along the \textit{simplicity} axis. Our method successfully simplifies the sketches in a smooth manner, while still capturing the core characteristics of the scene. For example, notice how the shape of the Buddha sculpture is preserved across all levels.
Observe that these simplifications are achieved \textit{implicitly} using our iterative sketch simplification technique. 
Please refer to the supplemental file for many more results.

\vspace{-0.01cm}
\subsection{Comparison with Existing Methods}
\label{sec:compareMethods}
\begin{figure}
    \centering
    \includegraphics[width=0.95\linewidth]{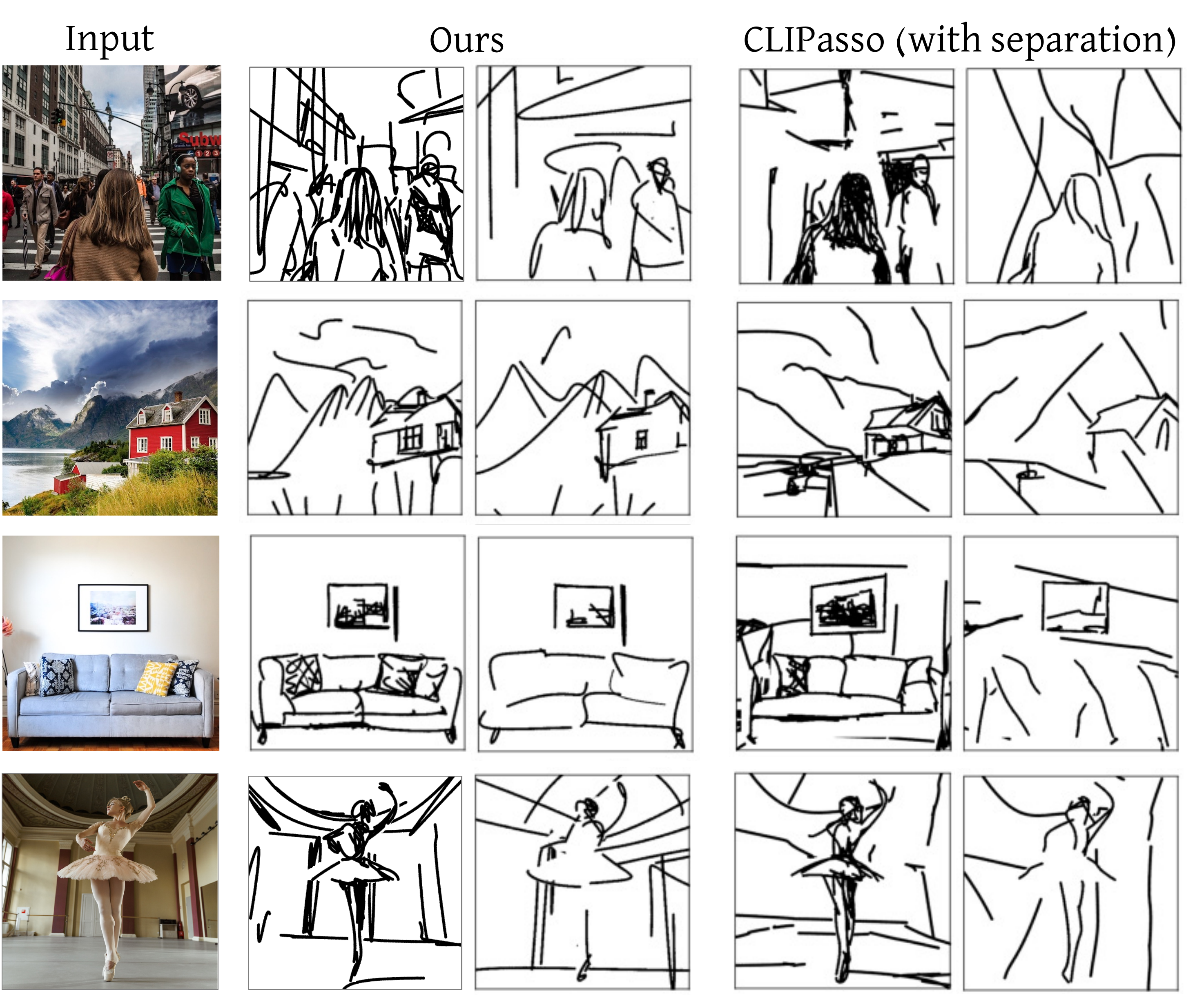}
    \caption{Comparison to CLIPasso~\cite{vinker2022clipasso}. Note how CLIPasso fails to capture the background in most cases, especially at higher abstraction levels, despite having the same stroke budget.}
    \vspace{-0.5cm}
    \label{fig:qualitative_pairs}
\end{figure}

In~\Cref{fig:qualitative_pairs} we present a comparison to CLIPasso~\cite{vinker2022clipasso}. For a fair comparison, we use our scene decomposition technique to separate the input images into foreground and background and use CLIPasso to sketch each part separately before combining them. Also, since CLIPasso requires a predefined number of strokes as input, we set the number of strokes in CLIPasso to be the same as that learned implicitly by our method.
For each image we show two sketches with two different levels of abstraction. 
As expected, CLIPasso is able to portray objects accurately, as it was designed for this purpose. However, in some cases, such as the sofa, CLIPasso fails to depict the object at a higher abstraction level. This drawback may result from the abstraction being learned from scratch per a given number of strokes, rather than gradually. Additionally, in most cases CLIPasso completely fails to capture the background, even when using many strokes (\eg in the first and fourth rows). Our method captures both the foreground and the background in a visually pleasing manner, thanks to our \textit{learned} simplification approach. For example, our method is able to convey the notion of the buildings in the first row or mountains in the second row with only a small number of simple scribbles. Similarly, our approach successfully depicts the subjects across all scenes.

In~\Cref{fig:scene_sketching_comparisons} we present a comparison with three state-of-the-art methods for scene sketching~\cite{yi2020unpaired, li2019photo, chan2022learning}.
On the left, as a simple baseline, we present the edge maps of the input images obtained using XDoG~\cite{Winnemller2012XDoGAE}. 
On the right, we present three sketches produced by our method depicting three representative levels of abstraction.

The sketches produced by UPDG~\cite{yi2020unpaired} and Chan~\etal~\shortcite{chan2022learning} are detailed, closely following the edge maps of the input images (such as the buildings in row $2$). 
These sketches are most similar to the sketches shown in the leftmost column of our set of results, which also align well with the input scene structure.
The sketches produced by Photo-Sketching~\cite{li2019photo} are less detailed and may lack the semantic meaning of the input scene. For example, in the first row, it is difficult to identify the sketch as being that of a person.
Importantly, none of the alternative \textit{scene} sketching approaches can produce sketches with varying abstraction levels, nor can they produce sketches in vector format.
We note that in contrast to the methods considered in~\Cref{fig:scene_sketching_comparisons}, our method operates per image and requires no training data. However, this comes with the disadvantage of longer running time, taking 6 minutes to produce one sketch on a commercial GPU.

\input{files/figures/scene_sketching_comparison}

\subsection{Quantitative Evaluation}
In this section, we provide a quantitative evaluation of our method's ability to produce sketch abstractions along both the simplicity and fidelity axes.
To this end, we collected a variety of images spanning five classes of scene imagery: people, urban, nature, indoor, and animals, with seven images for each class. For each image, we created the $4\times 4$ sketch abstraction matrix -- resulting in a total of $560$ sketches, and 
created sketches using the different methods presented in Section~\ref{sec:compareMethods}.
To make a fair comparison with CLIPasso we generated sketches with four levels of abstraction, using the average number of strokes obtained by our method at the four simplicity levels. For UPDG and Chan~\etal, we obtained sketches with three different styles, and averaged the quantitative scores across the three styles, as they represent the same abstraction level. For Photo-Sketching only one level of abstraction and one style is supported.

\vspace{-0.2cm}
\paragraph{\textbf{Fidelity Changes}}
To measure the fidelity level of the generated sketches, we compute the MS-SSIM~\cite{wang2003multiscale} score between the edge map of each input image (extracted using XDoG) and the corresponding sketch.
In~\Cref{tb:geometry_metrics_by_fidelity} we show the average resulting scores among all categories, where a higher score indicates a higher similarity.
Examining the results matrix of our method, as we move right along the fidelity axis, the scores gradually decrease.  This indicates that the sketches become ``looser'' with respect to the input geometry, as desired.
The sketches by UPDG and Chan~\etal obtained high scores, which is consistent with our observation that their method produces sketches that follow the edges of the input image.
The scores for CLIPasso show that the fidelity level of their sketches does not change much across simplification levels and is similar to the fidelity of sketches of our method at the last two levels (the two rightmost columns). This suggests that CLIPasso is not capable of producing large variations of fidelity abstractions.

\input{files/figures/geometry_preservation}

\vspace{-0.35cm}
\paragraph{Sketch Recognizability} 
A key requirement for successful abstraction is that the input scene will remain recognizable in the sketches across different levels of abstraction. 
To evaluate this, we devise the following recognition experiment on the set of images described above. Using a pre-trained ViT-B/16 CLIP model (different than the one used for training), we performed zero-shot image classification over each input image and the corresponding resulting sketches from the different methods. We use a set of $200$ class names taken from commonly used image classification and object detection datasets~\cite{lin2014microsoft,krizhevsky2009learning} and compute the percent of sketches where at least $2$ of the top $5$ classes predicted for the input image were also present in the sketch's top $5$ classes.
We consider the top $5$ predicted classes since a scene image naturally contains multiple objects.

~\Cref{tb:recognizability_metrics} shows the average recognition rates across all images for each of the described methods.
The recognizability of the sketches produced by our method remains relatively consistent across different simplicity and fidelity levels, with a naturally slight decrease as we increase the simplicity level. 
We do observe a large decrease in the recognition score in the first column. This discrepancy can be attributed to the first fidelity level following the image structure closely, which makes it more difficult to depict the scene with fewer strokes.
CLIPasso's fidelity level is most similar to our two rightmost columns (as shown in \Cref{tb:geometry_metrics_by_fidelity}). When comparing our recognition rates along these columns to the results of CLIPasso, one can observe that at higher simplicity levels, their method looses the scene's semantics.

\input{files/figures/recognizability_metrics}

%% file: files/figures/scene_sketching_comparison.tex
\begin{figure*}[ht]
    \centering
    \setlength{\tabcolsep}{1.5pt}
    {\small
    \begin{tabular}{c @{\hspace{0.2cm}} | c c c c @{\hspace{0.2cm}} | c c c}

        \includegraphics[width=0.09\textwidth]{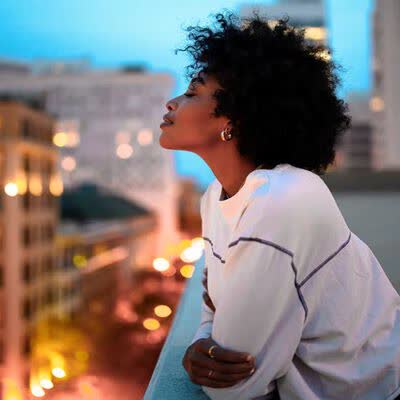} &
        \hspace{0.1cm}
        \includegraphics[width=0.09\textwidth]{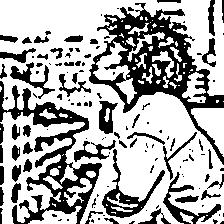} &
        \includegraphics[width=0.09\textwidth]{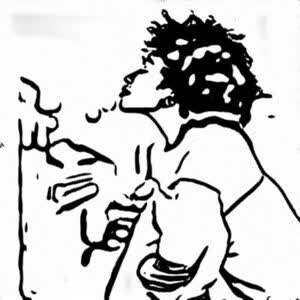} &
        \includegraphics[width=0.09\textwidth]{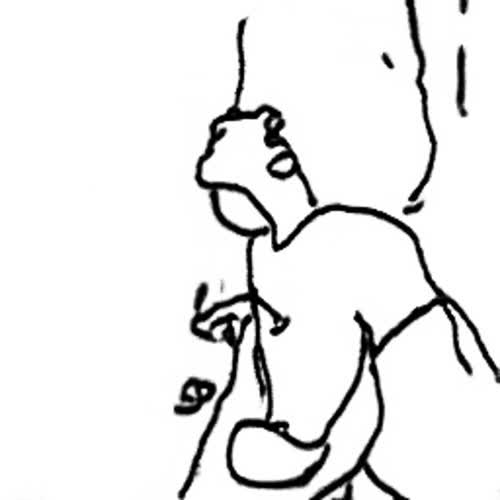} &
        \includegraphics[width=0.09\textwidth]{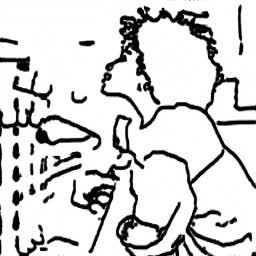} &
        \hspace{0.1cm}
        \includegraphics[width=0.09\textwidth]{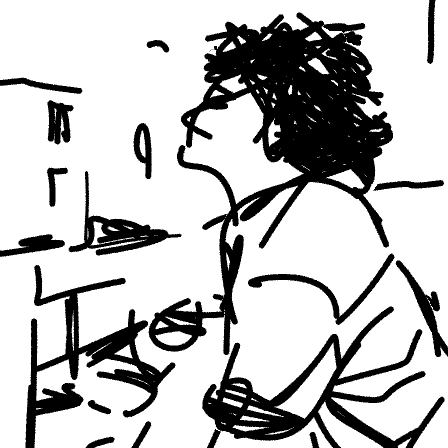} &
        \includegraphics[width=0.09\textwidth]{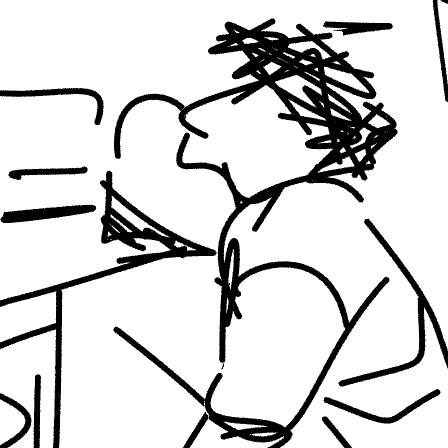} &
        \includegraphics[width=0.09\textwidth]{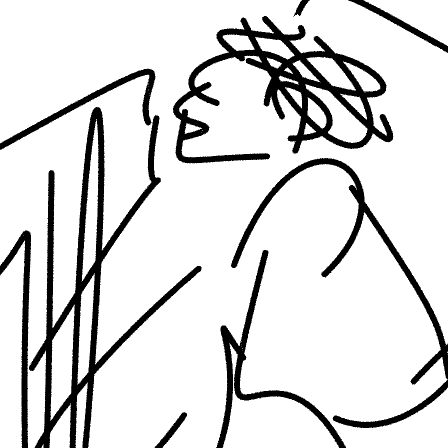}\\

        \includegraphics[width=0.09\textwidth]{figs/inputs/woman_city.jpg} &
        \hspace{0.1cm}
        \includegraphics[width=0.09\textwidth]{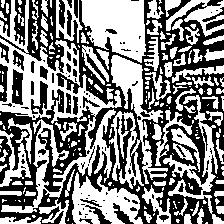} &
        \includegraphics[width=0.09\textwidth]{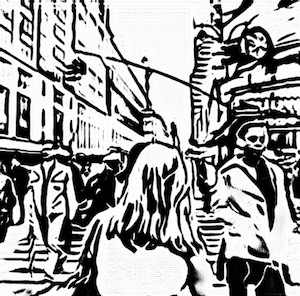} &
        \includegraphics[width=0.09\textwidth]{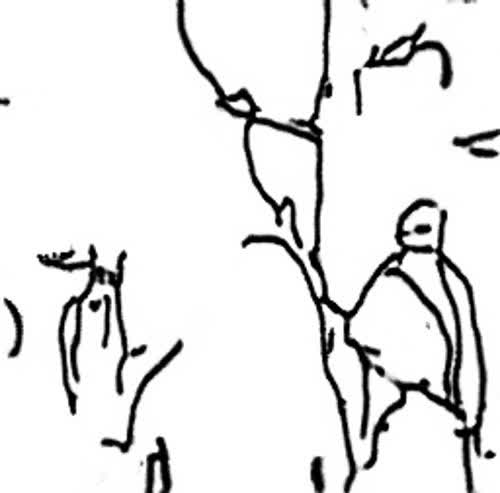} &
        \includegraphics[width=0.09\textwidth]{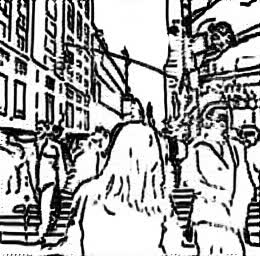} &
        \hspace{0.1cm}
        \includegraphics[width=0.09\textwidth]{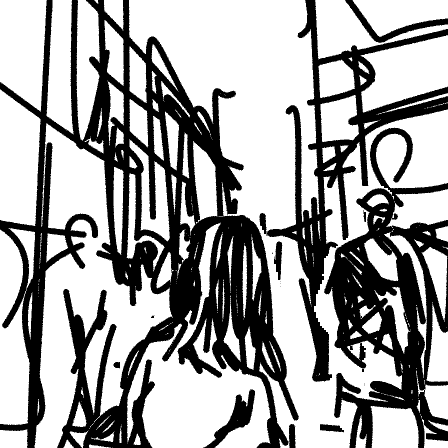} &
        \includegraphics[width=0.09\textwidth]{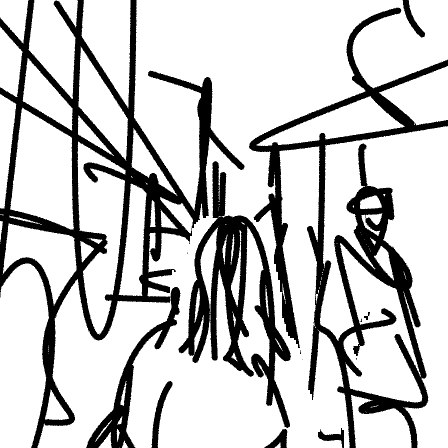} &
        \includegraphics[width=0.09\textwidth]{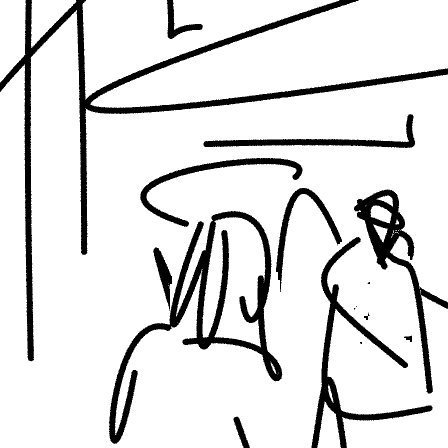} \\
        
        \includegraphics[width=0.09\textwidth]{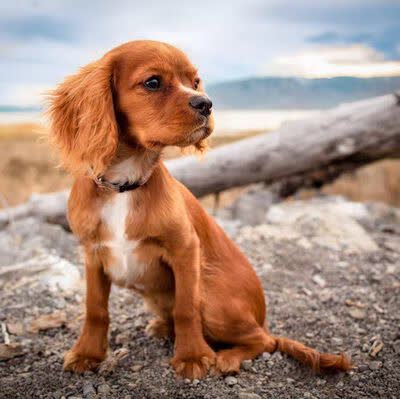} &
        \hspace{0.1cm}
        \includegraphics[width=0.09\textwidth]{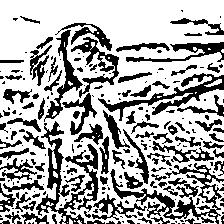} &
        \includegraphics[width=0.09\textwidth]{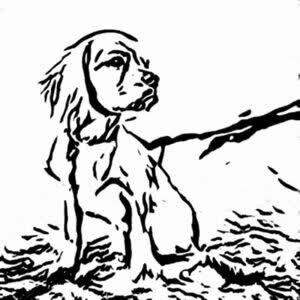} &
        \includegraphics[width=0.09\textwidth]{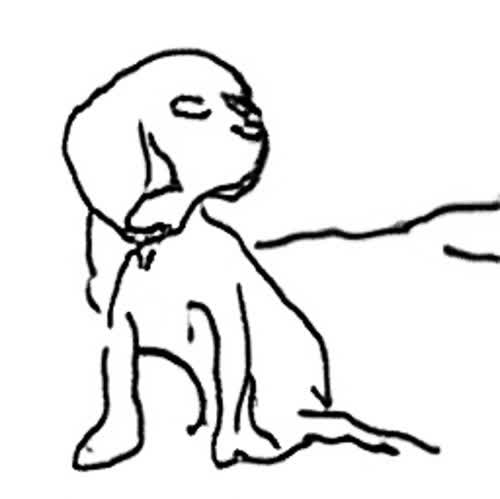} &
        \includegraphics[width=0.09\textwidth]{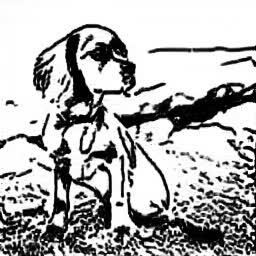} &
        \hspace{0.1cm}
        \includegraphics[width=0.09\textwidth]{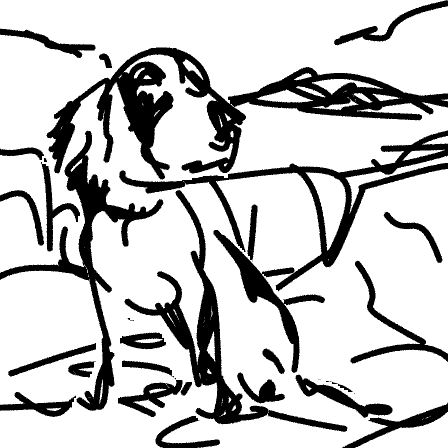} &
        \includegraphics[width=0.09\textwidth]{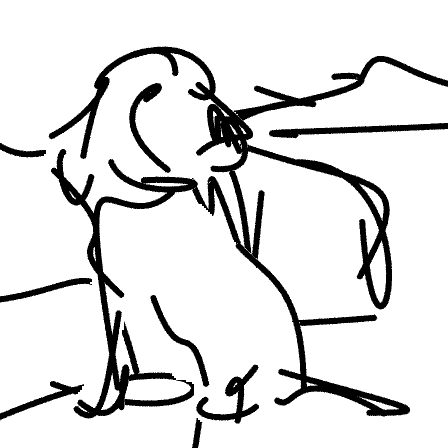} &
        \includegraphics[width=0.09\textwidth]{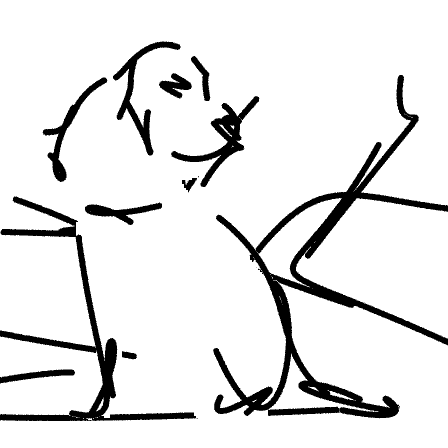} \\

        Input & XDoG & UPDG & Photo-Sketching & Chan~\etal &
        \multicolumn{3}{c}{\hspace{0.1cm} \xrfill[0.5ex]{0.5pt}\quad Ours \quad \xrfill[0.5ex]{0.5pt}}

    \end{tabular}
    
    }
    \vspace{0.1cm}
    \caption{Scene sketching results and comparisons. From left to right are the sketches obtained using XDoG~\cite{Winnemller2012XDoGAE}, UPDG \cite{yi2020unpaired}, Photo-Sketching \cite{li2019photo}, and Chan~\etal~\shortcite{chan2022learning}. On the right, are three representative sketches produced by our method depicting three levels of abstraction. Note that UPDG and Chan~\etal can produce sketches with three different styles, however all the sketches represent a similar level of abstraction. We choose one representative style but provide more style comparisons in the supplementary material.}
    \vspace{-0.2cm}
    \label{fig:scene_sketching_comparisons}
\end{figure*}

%% file: files/figures/geometry_preservation.tex
\begin{table}
\small
\centering
\setlength{\tabcolsep}{4pt}
\caption{Comparison of the average MS-SSIM score, computed  between the edge map of the input images and generated sketches.}
\vspace{0.05cm}
\begin{tabular}{c@{\hskip6pt}c@{\hskip6pt}c@{\hskip6pt}c@{\hskip6pt}ccccc} 
\toprule
& \multicolumn{4}{c}{Ours} & \multirow{2}{*}{CLIPasso} & \multirow{2}{*}{UPDG} & \multirow{2}{*}{\begin{tabular}[c]{@{}c@{}}Chan\\ et al. \end{tabular}} & \multirow{2}{*}{\begin{tabular}[c]{@{}c@{}}Photo-\\Sketch\end{tabular}}  \\ 
\cmidrule(l{2pt}r{2pt}){2-5}
& \multicolumn{4}{c}{Fidelity} & & & & \\ 
\cmidrule(l{2pt}r{2pt}){6-6}\cmidrule(l{2pt}r{2pt}){7-7}\cmidrule(l{2pt}r{2pt}){8-8}\cmidrule(l{2pt}r{2pt}){9-9}
\multirow{4}{*}{\rotatebox{90}{Simplicity}}
& 0.39 & 0.23 & 0.22 & 0.17 & 0.21 & \multirow{4}{*}{0.57} & \multirow{4}{*}{0.55} & \multirow{4}{*}{0.27} \\ 
& 0.37 & 0.23 & 0.21 & 0.19 & 0.21 &   &   & \\
& 0.36 & 0.22 & 0.20 & 0.18 & 0.15 &   &   & \\
& 0.34 & 0.22 & 0.18 & 0.14 & 0.13 &   &   & \\  
\bottomrule
\end{tabular}
\vspace{-0.4cm}
\label{tb:geometry_metrics_by_fidelity}
\end{table}

%% file: files/figures/recognizability_metrics.tex
\begin{table}
\small
\centering
\setlength{\tabcolsep}{4pt}
\caption{Recognizability scores, using a CLIP ViT-B/16 model for zero-shot classification on the input image and generated sketches.}
\begin{tabular}{c@{\hskip6pt}c@{\hskip6pt}c@{\hskip6pt}c@{\hskip6pt}ccccc} 
\toprule
& \multicolumn{4}{c}{Ours} & \multirow{2}{*}{CLIPasso} & \multirow{2}{*}{UPDG} & \multirow{2}{*}{\begin{tabular}[c]{@{}c@{}}Chan\\et al.\end{tabular}} & \multirow{2}{*}{\begin{tabular}[c]{@{}c@{}}Photo-\\Sketch\end{tabular}}  \\ 
\cmidrule(l{2pt}r{2pt}){2-5}
& \multicolumn{4}{c}{Fidelity} & & & & \\ 
\cmidrule(l{2pt}r{2pt}){6-6}\cmidrule(l{2pt}r{2pt}){7-7}\cmidrule(l{2pt}r{2pt}){8-8}\cmidrule(l{2pt}r{2pt}){9-9}
\multirow{4}{*}{\rotatebox{90}{Simplicity}} 

& 0.92 & 1.00 & 0.95 & 0.97 & 0.92 & \multirow{4}{*}{0.87} & \multirow{4}{*}{0.91} & \multirow{4}{*}{0.62} \\
& 0.54 & 0.97 & 1.00 & 0.91 & 0.83 &   &  & \\
& 0.54 & 0.93 & 0.94 & 0.89 & 0.70 &   &  & \\
& 0.44 & 0.79 & 0.91 & 0.85 & 0.43 &   &  & \\
\bottomrule
\end{tabular}
\label{tb:recognizability_metrics}
\vspace{-0.4cm}
\end{table}

%% file: files/limitations.tex
\section{Conclusions}
\label{sec:discussion} 
We presented a method for performing scene sketching with different types and multiple levels of abstraction.
We disentangled the concept of sketch abstraction into two axes: \textit{fidelity} and \textit{simplicity}. 
We demonstrated the ability to cover a wide range of abstractions across various challenging scene images and the advantage of using vector representation and scene decomposition to allow for greater artistic control.
It is our hope that our work will open the door for further research in the emerging area of computational generation of visual abstractions. Future research could focus on further extending these axes and formulate innovative ideas for controlling visual abstractions.

%% file: supplementary_arxiv.tex
\null
\newpage
\section{Implementation Details}~\label{sec:implementations}
In this section, we provide specific details about the implementation of our method. We will further release all code and image sets used for evaluations to facilitate further research and comparisons.

\begin{figure}[h]
    \centering
    \includegraphics[width=1\linewidth]{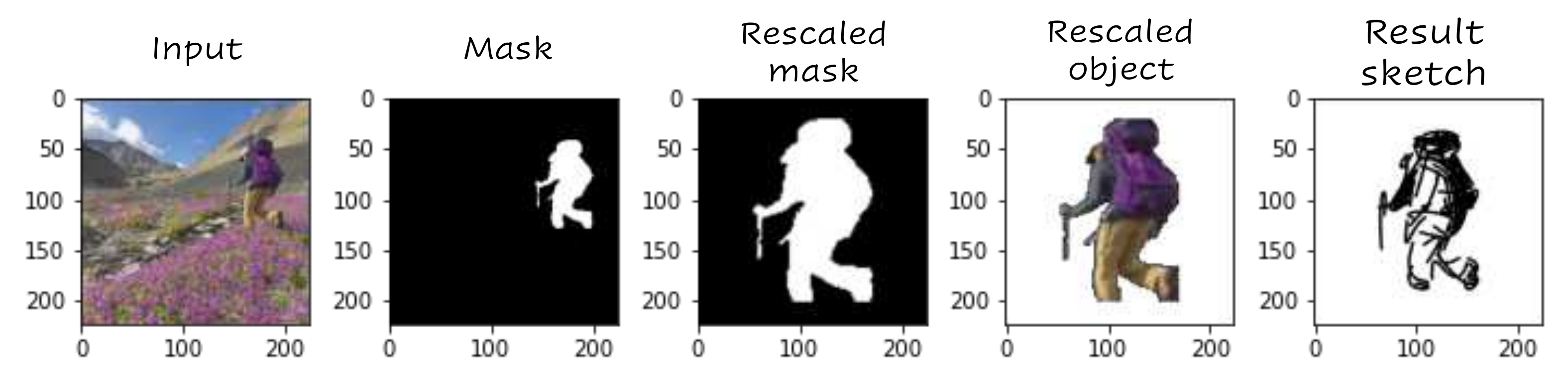}
    \vspace{-0.3cm}
    \caption{Object re-scaling procedure.}
    \vspace{-0.275cm}
    \label{fig:rescaling}
\end{figure}

\subsection{Image Preprocessing}~\label{sec:image_preprocessing}
As stated in the paper, we use a pre-trained U$^2$-Net salient object detector~\cite{qin2020u2} to extract the scene's salient object(s). To receive a binary map, we threshold the resulting map from U$^2$-Net such that pixels with a value smaller than $0.5$ are classified as background, and the remaining pixels are classified as salient objects.
We then use this mask as an input to a pre-trained LaMa~\cite{suvorov2022resolution} inpainting model to recover the missing regions in the background image. 

\paragraph*{\textbf{Object Scaling}} 
In the case where the saliency detection process detected only one object, and this object fills less than $70\%$ of the image size, we perform an additional pre-processing step to increase the size of the object before sketching. 
This assists in sketching key features of the object when using a large number of strokes. 
Specifically, we first take the masked object and compute its bounding box. We then shift the masked object to the center of the image and resize the object such that it covers $\approx 70\%$ of the image.
We apply the sketching procedure on the scaled object and then resize and shift the resulting sketch back to the original location in the input image. Note that since our sketches are given in vector representation, it is possible to re-scale and shift them without changing their resolution.
This process is illustrated in~\Cref{fig:rescaling}.

\subsection{MLP Training}

\paragraph*{\textbf{Hyper-parameters}}
In all experiments, we set the number of strokes to $n = 64$ in the first phase of sketching and train $MLP_{loc}$ for $2,000$ iterations. 
For generating the series of simplified sketches (Section 3.4 in the main paper), we perform $8$ iterative steps. As discussed in Section 3.4, for each fidelity level $k$, we define a separate function $f_k$ for defining the set of ratios used in $\mathcal{L}_{ratio}$. Along with this function, we define a separate step size for sampling the function $f_k$. For simplifying the background sketches, we set this step size to be $\{0.35, 0.45, 0.5, 0.9\}$ for layers $\{2, 7, 8, 11\}$, respectively. For simplifying the object sketches, we set the step sizes to be $\{0.45, 0.4, 0.5, 0.9\}$. Each simplification step is obtained by training $MLP_{simp}$ and $MLP_{loc}$ for $500$ iterations. 
We employ the Adam optimizer with a constant learning rate of $1\mathrm{e}{-4}$ for training both MLP networks.
The input to $MLP_{simp}$ is set to be a random-valued vector of dimension $n$. 

\paragraph*{\textbf{Augmentations}}
As also done in Vinker~\etal~\shortcite{vinker2022clipasso}, we apply random affine augmentations (\ie random perspective and random cropping transformations) to both the input image and generated sketch before passing them as inputs to the CLIP model for computing the loss. 

\paragraph*{\textbf{GradNorm}}
We train $MLP_{simp}$ and $MLP_{loc}$ with three different losses simultaneously in order to achieve our visual simplifications. As these losses compete with each other, training has the potential to be highly unstable. For example, when training with multiple losses, the gradients of one loss may be stronger than the other, resulting in the need to weigh the losses accordingly.
To help achieve a more stable training process and to ensure that each loss contributes equally to the optimization process, we use GradNorm~\cite{GradNorm}, which automatically balances the training process by dynamically adjusting the gradient magnitudes. This balancing is achieved by weighing the losses inversely proportional to their contribution to the overall gradient.

\subsection{Matrix Composition}
As stated in the main paper, we separate the scene into two regions (based on their saliency map) and apply the sketching scheme to both independently, we then combine the resulting sketches to form the final matrix.
To combine the foreground and background we simply aggregate the corresponding strokes at a given level of fidelity and simplicity.
Note that we also export the mask used to separate them, if the user wish to locate it behind the object to avoid the collision of strokes.
We also export the separate matrices, to allow users to combine sketches from different levels of abstraction as a post process.

\begin{figure}
    \centering
    \includegraphics[width=0.8\linewidth]{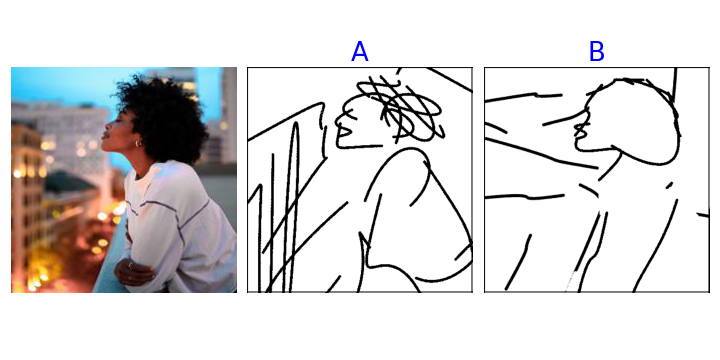}
    \vspace{-0.5cm}
    \caption{Example of how images were presented to participants in the user study.}
    \label{fig:user_study_example}
\end{figure}

\input{files/figures/supplementary/recognizability_example_images}
\input{files/figures/quantitative_num_strokes}
\section{User Study}
In Section 4.3 of the main paper, we presented a quantitative evaluation measuring the fidelity and recognizability of our sketches based on automatic metrics.
As opposed to the fidelity measure, which can be determined by measuring the distance from the edge map of the input scene, validating the recognizability is more challenging.
To this end, we also conducted a user study to further validate the findings presented by the CLIP zero-shot classification approach.

The user study examines how well the sketches depict the input scene, considering both the foreground and background.
Using $30$ images from the set described in Section 4.3, we compared our sketches with three alternative methods: CLIPasso~\cite{vinker2022clipasso}, Chan~\etal~\cite{chan2022learning}, and Photo-Sketching~\cite{li2019photo}. 
The participants were presented with the input image along with two sketches, one produced by our method and the other by the alternative method (with the sketches presented in random order). An example is provided in~\Cref{fig:user_study_example}.
The following question was posed to participants:

\begin{table}
\setlength{\tabcolsep}{3pt} 
 \renewcommand{\arraystretch}{1.2} 
\begin{center}
\begin{tabular}{c|c||c|c||c|c}
    \hline
    \multicolumn{2}{c||}{\begin{tabular}[c]{@{}c@{}}Ours v.s. \\ CLIPasso\end{tabular}} & \multicolumn{2}{c||}{\begin{tabular}[c]{@{}c@{}}Ours v.s. \\ Chan~\etal \end{tabular}} & \multicolumn{2}{c}{\begin{tabular}[c]{@{}c@{}}Ours v.s. \\ Photo-Sketching\end{tabular}} \\
    \toprule
    Ours & 84.8\% & Ours & 52.5\% & Ours & 75.6\% \\
    \midrule
     CLIPasso & 7.0\% & Chan~\etal & 27.3\% & \begin{tabular}[c]{@{}c@{}}Photo-\\ Sketching\end{tabular} & 15.8\% \\
     \midrule
     Equal & 8.2\% & Equal & 20.2\% & Equal & 8.6\%\\
    \bottomrule
\end{tabular}
\end{center}
\vspace{-0.1cm}
\caption{Results of our user study. We compare $30$ sketches produced by our method to three alternative methods: CLIPasso~\cite{vinker2022clipasso}, Chan~\etal~\cite{chan2022learning}, and Photo-Sketching~\cite{li2019photo}. For each method, we specify the percent of responses that preferred our sketch, the sketch of the alternative method, or found the sketches to be similar in their ability to capture the scene semantics.}
\label{tab:user_study}
\end{table}

\textit{Which sketch better depicts the image content?}

\textit{In your answer, please relate to:}

\textit{(1) Preservation of both foreground and background.}

\textit{(2) Semantic preservation - i.e., reflecting the meaning of the elements.}

Participants could choose between three options: ``A", ``B", and ``A and B at a similar level``.

It is important to note that in order to make a fair comparison, we compared the methods which produce abstract sketches (CLIPasso~\cite{vinker2022clipasso} and Photo-Sketching~\cite{li2019photo}) with our abstract sketches (in row 4 of the matrix, i.e., at the highest abstraction level).
Conversely, we compared the sketches of Chan~\etal~\cite{chan2022learning}, which are more detailed and have greater fidelity, to our sketches in their most detailed form (top left corner of the matrix).
We applied CLIPasso using our scene decomposition technique, and the same number of strokes as learned by our method, and for Chan~\etal we used the contour style sketches.
We collected responses from $25$ participants for the survey, which contained a total of $90$ questions (i.e., $2,250$ responses were collected in total).

The average resulting scores among all participants and images are shown in~\Cref{tab:user_study}.
In the first, second, and third columns, we show the scores obtained when comparing to CLIPasso~\cite{vinker2022clipasso}, Chan~\etal~\cite{chan2022learning}, and Photo-Sketching~\cite{li2019photo} respectively.
Compared to CLIPasso and Photo-Sketching, our method achieved significantly higher rates ($84.8\%$ and $75.6\%$ of responses favored our method over the respective alternatives). Conversely, only $7\%$ and $15.8\%$ of responses preferred the results of the alternative method, respectively.
Although sketches produced by Chan~\etal~\cite{chan2022learning} are highly detailed, $52.5\%$ of the responses preferred our sketches, while only $20\%$ considered our sketches and Chan \etal's sketches to be similar.

The results of the user study support the findings presented in the main paper, demonstrating that sketches produced by our method faithfully capture both the foreground and background elements in the scene among varying abstraction levels.

\section{Additional Quantitative Analysis}
In this section, we provide additional details, examples, and results regarding the quantitative evaluations presented in the paper. First, in~\Cref{fig:recognizability_example_images} we present example inputs and representative generated sketches for each of our five scene categories used for evaluations.

\subsection{Sketch Recognizability}
To compute our recognizability metrics, we perform zero-shot classification using a pre-trained ViT-B/16~\cite{dosovitskiy2020image} CLIP model. Observe this model is different than the ViT-B/32 model used to generate sketches, ensuring a more fair evaluation of our sketches. When performing the zero-shot classification, we follow the evaluation setup used in CLIP~\cite{CLIP} and apply $80$ prompt templates when defining our $200$ classes to CLIP's text encoder. This includes prompts of the form: ``a rendering of a \{\}'', ``a drawing of a \{\}`, and ``a sketch of a \{\}''. We then compute the cosine similarity between all text embeddings and the embedding corresponding to either our input image or generated sketches.

In~\Cref{fig:recognizability_example_predictions}, we present example zero-shot classification results obtained on various input images and sketches across our five scene categories.

\subsection{Number of Strokes by Simplicity Level}
We examine our method's ability to generate sketches at varying levels of simplicity. For that purpose, we measure the final number of strokes used to generate the sketches.
We extract this information from the generated sketch SVGs across all $560$ sketches.
We present the results in~\Cref{fig:quantitative_num_strokes}, split between the different fidelity levels (indicated by different colors) and simplicity levels (shown along the x-axis). As can be seen, the number of strokes decreases as we move along the simplicity axis, across all fidelity levels.

In~\Cref{fig:quantitative_num_strokes_by_class}, we present the same results but split between the different scene categories and split between composing the foreground and background sketches. As can be seen, the resulting functions for the different fidelity levels follow an exponential relation as we strengthen the simplification level. 

\input{files/figures/supplementary/recognizability_sample_results}

\input{files/figures/supplementary/quantitative_num_strokes_by_class}
\hfill

\input{files/figures/supplementary/ablation_resnet}
\null\newpage
\null\newpage
\null\newpage
\null\newpage

\section{Ablation Study: General Design Choices}
\subsection{ViT vs. ResNet}
In~\Cref{fig:ablation_full_scene_resnet}, we demonstrate the scene sketching results obtained with a ResNet101-based CLIP compared to those obtained with the ViT-based CLIP model employed in this work. When computing $\mathcal{L}_{CLIP}$ using a single layer of ResNet (\ie layer $2$, $3$, or $4$), we are unable to capture the input scene, indicating that a combination of the layers must be used for capturing the more global details of a complete scene. However, even when computing $\mathcal{L}_{CLIP}$ using multiple ResNet layers (as was done in CLIPasso), the network still struggles in capturing the details of the scene. For example, in row $3$, although we are able to roughly capture the outline of the bull's head and horns, the network is unable to capture the bull's body and scene background. In contrast, when replacing the ResNet model with the more powerful ViT model, we are able to capture both the scene foreground and background, even when using a single layer for computing the loss. This naturally allows us to control the level of fidelity of the generated sketch by simply altering the single ViT layer that is used for computing $\mathcal{L}_{CLIP}$.

\subsection{Foreground-Background Separation}
In Section 3.4 of the main paper, we introduced our scene decomposition technique where the foreground and background components of the input are sketched separately and then merged. In~\Cref{fig:ablation_foreground_background_supp} we provide additional sketching results obtained with and without the scene decomposition for both abstraction axes. 
At the top, we present the resulting sketches along the fidelity axis.
Observe how the house in the leftmost sketch in the first row appears to disappear within the entire scene. Furthermore, note the artifacts that appear in the face of the dog as the abstraction increases.
In contrast, by explicitly separating the foreground and background, we can apply additional constraints over the foreground sketches to help mitigate unwanted artifacts. As a result, we are able to better maintain the correct structure of both the house and dog in the provided examples. 

At the bottom of~\Cref{fig:ablation_foreground_background_supp} we show the resulting sketches along the simplification axis. Note how the house in the first row almost disappears completely, and that there are not enough strokes to depict the mountains in the background. By considering the entire scene as a whole, the model has no explicit control over how to balance the level of details placed between the object and the background. As a result, more strokes are typically used to sketch the background, which consumes a larger portion of the entire image (and therefore leads to a larger reduction in $\mathcal{L}_{CLIP}$).

\input{files/figures/supplementary/ablation_foreground_background}

\null\newpage
\null\newpage
\input{files/figures/supplementary/ablation_explicit_define_num_strokes}
\input{files/figures/supplementary/ablation_direct_lratio}

\null\newpage

\section{Ablation Study: Simplicity Axis}
In this section, we analyze the design choices for the simplification training scheme and corresponding loss objectives.

\subsection{Explicitly Defining the Number of Strokes}~\label{sec:explicit_num_strokes}
We begin by analyzing our simplification scheme as a whole. That is, are we able to achieve a smooth simplification of an input scene by simply varying the number of strokes used to sketch the scene? In~\Cref{fig:ablation_explicit_define_num_strokes} we provide results comparing our implicit simplification scheme (on the right) with results obtained by sketching the scene using a varying number of strokes defined in advance (on the left). For the latter results, we use $64, 32, 16$, and $8$ strokes for sketching. For each input, we present the simplifications achieved at the last level of fidelity (\ie using layer $11$ of CLIP-ViT for training). 

Knowing in advance the number of strokes needed to achieve a specific level of abstraction is often challenging and varies between different inputs. 
For example, consider the image in the first row, containing a simpler scene of a house.  
When using only $16$ strokes for sketching this image, we are able to capture the components of the scene such as the existence of the house in the center, its roof, and its door. However, when sketching the more complex urban scene in the third row, using $16$ strokes may struggle to capture the general structure of the buildings or may not converge at all. By \textit{learning} how to simplify each sketch, our simplification scheme is able to adjust the number of strokes needed to more faithfully sketch a given image at various levels of simplification, while adapting to the complexity of the input scene in the learning process. 

Moreover, we observe that when defining the number of strokes explicitly, we may fail to get a smooth simplification of the initial sketch since each sketch is generated independently and may converge to a different local minimum.
For example, in the second row on the left, the sketches do not appear to be simplified versions of each previous step (\eg between the second and third steps), but rather new sketches of the input scene with an increased level of visual simplification. 
In contrast, each of our simplification results is initialized with the previous result, resulting in a smoother transition between each image.

\subsection{Replace the Ratio Loss With a Target Number of Strokes}
We note in the main paper that achieving gradual visual simplification requires balancing between $\mathcal{L}_{sparse}$ and $\mathcal{L}_{CLIP}$.
In our method, we do so by defining a set of factors used to define a balance between the relative strengths of the losses.
This section examines another possible approach: encouraging the training process to achieve a certain number of strokes during training and reducing this number at each level.
As opposed to~\Cref{sec:explicit_num_strokes} where we restrict the number of strokes completely, here we include the target number of strokes as another objective in the training process, thus allowing deviance from this number. 

To implement this approach we simply redefine $L_{ratio}$ as follows:
\begin{equation}
    \mathcal{L}_{ratio} = ||\mathcal{L}_{sparse} - n_{target}||,
\end{equation}
where $n_{target}$ is the desired number of strokes. Specifically, we define four levels of abstraction using $64$, $32$, $16$, and $8$ strokes. We then normalize the number of strokes to be between $0$ and $1$ and define $n_{target}$ as $\{1, 0.5, 0.25, 0.125\}$ for each level, respectively.

In~\Cref{fig:direct_lratio} we show the result of this experiment. As can be seen on the left, such an approach does not achieve the desired simplification.
As can be seen, on the left, the levels of abstraction are less gradual than on the right and do not reach full abstraction. This approach also relies on an arbitrary fixed number of strokes per abstraction level for all images, as opposed to allowing the ratio itself to implicitly define it in a content-dependent manner.

\subsection{Fine-tuning MLP-loc During Simplification}
As described in Section 3.3 of the main paper, when performing the simplification of a given sketch by training $MLP_{simp}$, we continue fine-tuning $MLP_{loc}$. 
Doing so is important since by training $MLP_{loc}$, we allow to slightly adjust the locations of strokes in the canvas, which helps encourage the simplified sketch to resemble the original input image. 
In~\Cref{fig:ablation_finetune_mlp_loc}, we show sketch simplifications across various inputs obtained with and without the fine-tuning of $MLP_{loc}$.

When $MLP_{loc}$ is held fixed, the simplification process is equivalent to simply selecting a subset of strokes to remove at each step. This approach will result in the appearance of visual simplification, but may not be sufficient to maintain the semantics of the scene. For example, in the last simplification step of the first example, the mountains in the background have disappeared, as have the buildings in the third example. In addition, the house in the second image can no longer be identified.
On the other hand, our results, obtained with the fine-tuning of $MLP_{loc}$, produce the desired visual simplification, while still preserving the same semantics of the input scenes. 

\input{files/figures/supplementary/ablation_finetune_mlp_loc}

\input{files/figures/supplementary/ablation_linear_fk}

\subsection{Defining the Function f-k as an Exponential}
In Section 3.3 of the main paper, we describe the process of selecting the set of factors used to achieve a gradual simplification of a given sketch. To do so, we defined a function $f_k$ for each fidelity level $k$ defining the balance between $\mathcal{L}_{CLIP}$ and $\mathcal{L}_{sparse}$. We find that an $f_k$ that models in an exponential relationship between $L_{CLIP}$ and $L_{sparse}$ achieves a simplification that is perceived smooth.

In this section, we demonstrate this effect by visually demonstrating the gradual simplification we achieve when choosing an $f_k$ that gives a linear relation between $\mathcal{L}_{CLIP}$ and $\mathcal{L}_{sparse}$.
To do so, we define the linear $f_k$ such that the sampled set of factors $\{r_k^1, .. r_k^m\}$ represent a constant step size by encouraging the removal of $8$ strokes in each step.

In~\Cref{fig:ablation_linear_fk} we present the results of this alternative setup.
For each set of generated sketches, we additionally present two graphs: (1) the resulting $L_{sparse}$ as a function of $L_{CLIP}$ (left), and (2) the final number of strokes as a function of the simplification step (right). 
Each point in the graphs corresponds to a single sketch with the color of the points indicating the location of the corresponding sketch along the simplification axis. That is, $0$ (or dark blue) indicates the leftmost, non-simplified sketch while $7$ (or yellow) indicates the rightmost sketch with the highest level of simplification.
Recall, as discussed in the main paper, the left graph should ideally depict an exponential relation between the two loss objectives in order for the simplification to appear smooth.

The results presented on the left side of~\Cref{fig:ablation_linear_fk} show the sketches and corresponding graphs produced when using a linear $f_k$ as defined above. The results on the right-hand side of the Figure show sketches obtained with our method when using the exponential $f_k$ as described in the main paper. As can be seen, the sketches in the linear alternative (left) remain too detailed at the initial abstraction levels and do not convey the smooth and gradual change perceptually as is present with the exponential function (right).

\input{files/figures/supplementary/same_ratios}

\input{files/figures/supplementary/same_step_size_fk}
\null\newpage

\subsection{Defining a Different Set of Factors for Each Layer}
In Section 3.3 of the main paper, we describe the process of selecting the set of factors $\{r_k^1, ..., r_k^m\}$ used to achieve a gradual simplification of a given sketch. 
In this section, we validate the use of different sets of factors for each fidelity level $k$.
Note that, as stated in the main paper, the set of factors $r_k^j$ determine the balance between $\mathcal{L}_{CLIP}$ and $\mathcal{L}_{sparse}$, which directly determines the level of visual simplification.

We show on the left-hand side of~\Cref{fig:same_ratios} the simplified sketches obtained for different levels $k$ of fidelity when using the same set of factors. Specifically, we apply the set of factors used for layer $\ell_8$ to the remaining ViT layers.
As can be seen, the perceived level of visual simplification is not uniform between different layers: the simplification achieved for layer $\ell_{11}$ is too weak with very little perceived change realized across all steps. In contrast, for layer $\ell_2$ the simplification is too strong with the background quickly ``disappearing'' as we move to the right. 
On the right-hand side of~\Cref{fig:same_ratios}, we present the results obtained with our method, where we fit a dedicated set of factors for each fidelity level $k$. As can be seen, the perceived level of abstraction among the different fidelity levels is more uniform and smooth.

\subsection{Defining a Different Sampling Step for Each f-k}
After defining the function $f_k$ for each fidelity level $k$, we use a different step size for sampling this function for each $k$.
We wish to achieve a similar appearance of simplification across different fidelity levels, and we find that in order to do so, using a different step size for sampling $f_k$ is crucial.

In~\Cref{fig:same_step_size}, on the left, we demonstrate results obtained when using the same step size for our four fidelity levels (\ie ViT layers $\ell_2,\ell_7,\ell_8,\ell_{11}$). 
As can be seen, for layers $\ell_2$ and $\ell_7$ the gradual simplification is not smooth and a noticeable jump in the strength of the abstraction can be seen between the second and third sketches.
Furthermore, for layer $\ell_7$, the change in step size caused the networks to converge to a noisy solution. Observe how the second sketch does not resemble a simplification of the previous one.

In contrast, the results on the right are obtained using the proposed approach of selecting different step sizes for each fidelity level $k$. As shown, the sketches do not suffer from the perceived artifacts present on the left. Moreover, the simplification results are also smooth and uniform in appearance between the different layers.

\input{files/figures/supplementary/ablation_all_vit_layers}

\section{Ablation Study: Fidelity Axis}
Finally, in this section, we perform various ablation studies to validate the design choices made with respect to our fidelity abstraction axis.

\subsection{Using l-4 for Object Sketching}
When decomposing the scene and sketching for the foreground image, we additionally compute $L_{CLIP}$ over layer $\ell_4$ of ViT. We found that doing so may help in preserving the geometry of more complex subjects, as illustrated in~\Cref{fig:ablation_l4}. This is most noticeable in finer details such as in the facial details of the old man and the dog or the body shape of the panda, for example.

\input{files/figures/supplementary/ablation_with_l4}

\subsection{Using Other ViT Layers for Training}
In order to obtain different levels of fidelity, we train $MLP_{loc}$ guided by different layers of the CLIP-ViT model for computing $\mathcal{L}_{CLIP}$.
Our model is based on the ViT-B/32 architecture that includes $11$ intermediate layers.
Our main paper presents the results of applying our training scheme to a subset of four layers: $2$, $7$, $8$, and $11$.
This subset of layers represents a range of possible fidelity levels that can be achieved by our method. 
While we focus on presenting results using only these four layers, our method can naturally generate additional levels of fidelity by using the remaining  intermediate layers.
We present the results of using additional layers in~\Cref{fig:ablation_all_vit_layers}.

\newpage
\section{Additional Results}
\begin{figure*}
    \centering
    \includegraphics[width=0.7\linewidth]{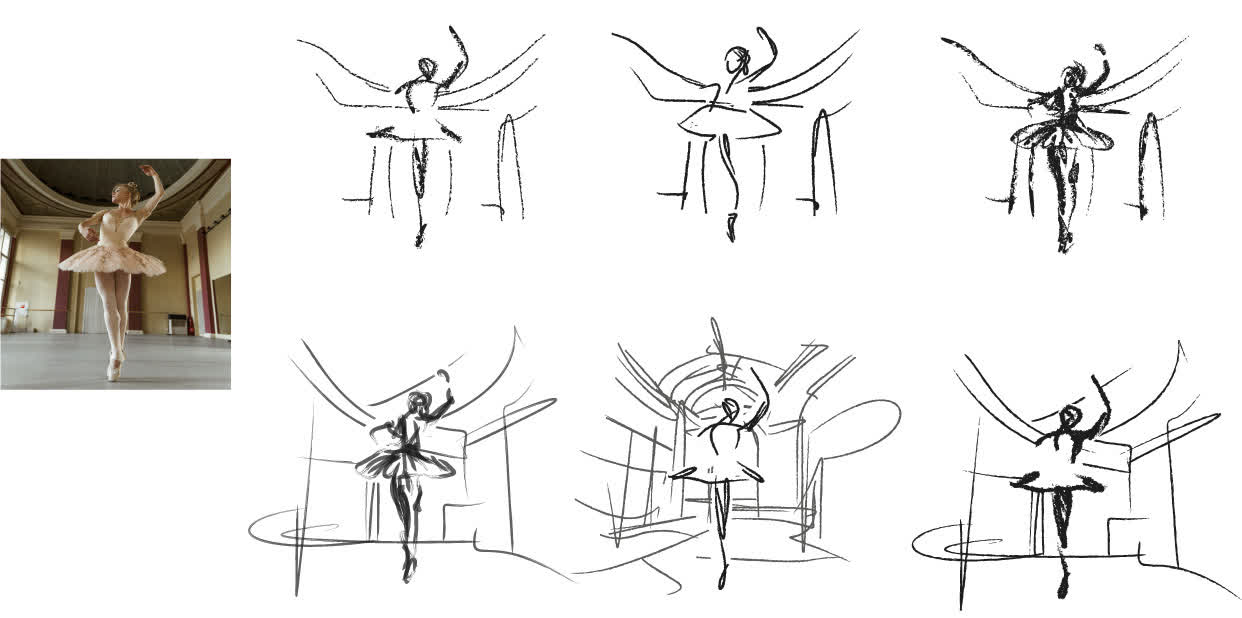}
    \rule[0.5ex]{0.7\linewidth}{1pt}
    \includegraphics[width=0.7\linewidth]{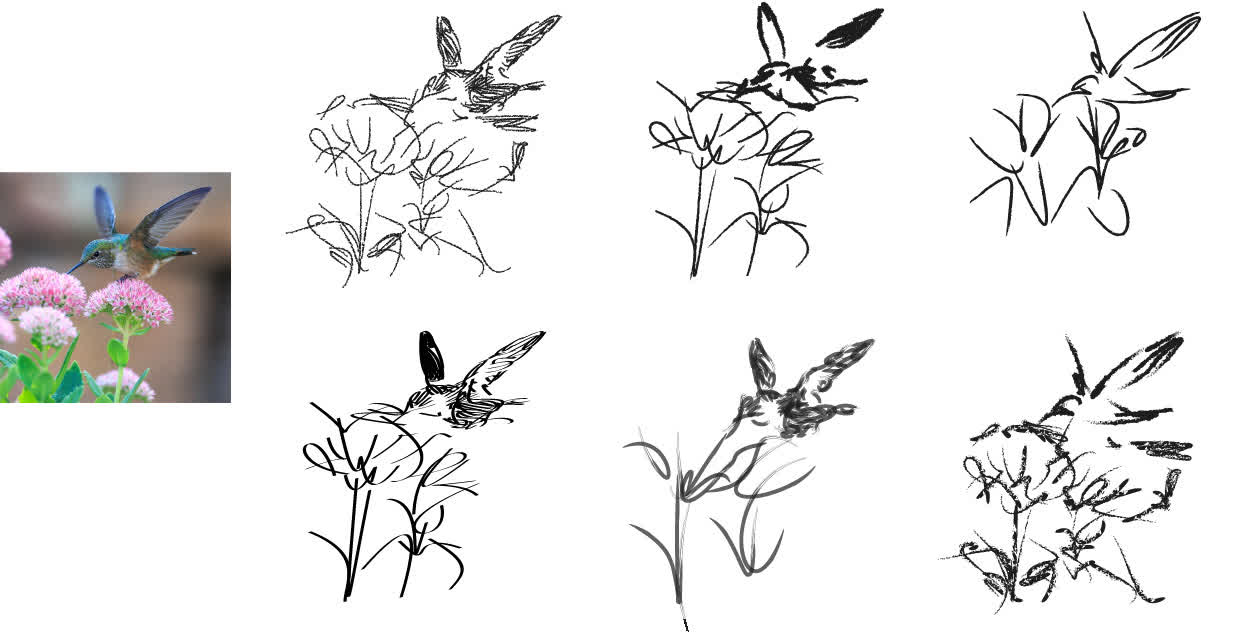}
    \rule[0.5ex]{0.7\linewidth}{1pt}
    \includegraphics[width=0.7\linewidth]{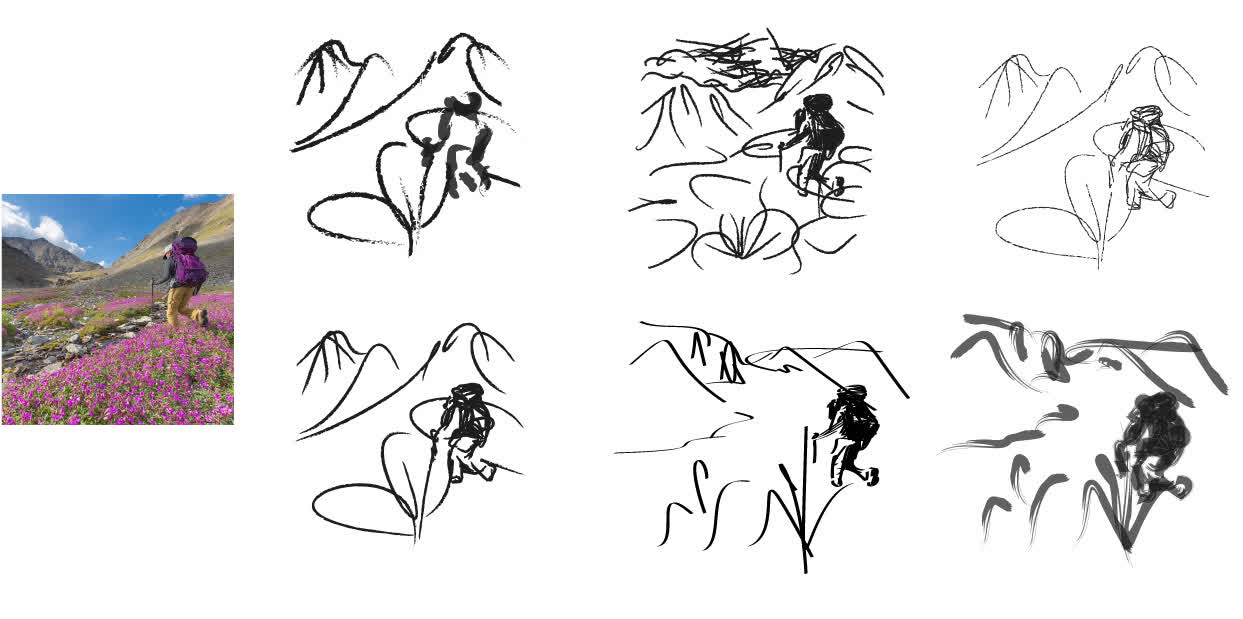}
    \caption{Additional control. For each image, we combine foreground and background sketches from different levels of abstraction, and edit the style of strokes using Adobe Illustrator. This illustrates the power of our method in providing various options for the user to edit the resulted sketches.}
    \label{fig:additional_control_supp}
\end{figure*}

We begin with additional results generated by our method. In~\Cref{fig:matrix1,fig:matrix2,fig:matrix3,fig:matrix4,fig:matrix5} we provide $4\times4$ abstraction matrices for various scene images. In addition, we present additional examples of the added control provided by the separation technique in~\Cref{fig:additional_control_supp}. This includes: (1) editing the style of strokes using Adobe Illustrator and (2) combining the foreground and background sketches and varying levels of abstractions to achieve various artistic effects.

\input{files/figures/supplementary/our_matrices/our_matrices_1}
\input{files/figures/supplementary/our_matrices/our_matrices_2}
\input{files/figures/supplementary/our_matrices/our_matrices_3}
\input{files/figures/supplementary/our_matrices/our_matrices_4}
\input{files/figures/supplementary/our_matrices/our_matrices_5}

\null\newpage
\null\newpage
\null\newpage
\null\newpage
\null\newpage
\null\newpage
\null\newpage
\null\newpage
\null\newpage
\null\newpage
\null\newpage
\null\newpage

\section{Additional Comparisons}

\input{files/figures/supplementary/diffusion_comparison}

\subsection{Diffusion Models}
\label{subsec:diffusion}
Recent advancements in diffusion models~\cite{ho2020denoising} have demonstrated an unprecedented ability to generate amazing imagery guided by a target text prompt or image~\cite{rombach2021highresolution,ramesh2021zero,ramesh2022hierarchical,nichol2021glide,yu2022scaling,saharia2022photorealistic}. In this section, we explore whether such models can be leveraged to generate abstract sketches of a given scene. We begin by exploring the recent Stable Diffusion model~\cite{rombach2021highresolution}. Given an input image, we perform a text-guided image-to-image translation of the input using text prompts such as: ``A black and white sketch image'' and ``A black and white single line abstract sketch.'' Results are illustrated in~\Cref{fig:diffusion_comparison}. As can be seen in the top row, Stable Diffusion struggles in capturing the sketch style even when guiding the denoising process with keywords such as ``A black and white sketch''. We do note that better results may be achieved with heavy prompt engineering or tricks such as prompt re-weighting
methods~\cite{automaticWebUi2022}. However, doing so would require heavy manual overhead for each input image. 

Another approach to assist in better capturing the desired sketch style would be to fine-tune the entire diffusion model on a collection of sketch images. However, this would require collecting a few hundred or thousands of images with matching captions and training a separate model for each desired style and level of abstraction.

As another diffusion-based approach, we consider the recent Textual Inversion (TI) technique from Gal~\etal~\shortcite{gal2022image}. Given a few images (\eg 5) of the desired style (\eg sketch), TI can be used to learn a new ``word'' representing the style. Users can then use a pre-trained text-to-image model such as the recent Latent Diffusion Model~\cite{rombach2021highresolution} to generate images of the learned style. For example, users can generate a sketch image of a house using the prompt ``A photo of a house in the style of $S_*$'' where $S_*$ represents our learned sketch style. 

To evaluate TI's ability to generate sketch images supported by our method, we collect $10$ sketches generated by our method --- $5$ detailed and $5$ abstract --- and learn a new token representing each of the sketch styles. In a similar fashion, we can learn a new word representing a unique object of interest (\eg the headless statue shown in~\Cref{fig:diffusion_comparison}). We can then generate images of the learned object in our learned style using prompts of the form ``A drawing of a $S_{statue}$ in the style of $S_{detailed}$'' or ``A drawing of a $S_{statue}$ in the style of $S_{abstract}$''. Example results are presented in the bottom half of~\Cref{fig:diffusion_comparison}. As can be seen, TI struggles in composing both the learned style and subject in a single image. Specifically, TI either struggles in capturing the unique shape of the statue (\eg its missing head) or struggles in capturing the learned sketch style (\eg TI may generate images in color). In contrast, our method is able to generate a range of possible sketch abstractions that successfully capture the input subject.

\input{files/figures/supplementary/clipasso_comparison_main}

\subsection{Scene Sketching Approaches}
In~\Cref{fig:clipasso_comp_main}, we provide a comparison to CLIPasso. We applied CLIPasso on the masked object and obtained the abstraction by explicitly specifying the number of strokes (\ie $64,32,16,$ and $8$ strokes). In the second and third rows we show the simplification results obtained by our method, at two different fidelity levels.
Since CLIPasso only offers a single axis of abstraction (mostly governed by simplification), the fidelity level of the sketch can not be explicitly controlled.
Additionally, unlike CLIPasso, where the user must manually determine the number of strokes required to achieve different levels of abstraction, our approach \textit{learns} the desired number of strokes. 

Lastly, observe that since each sketch of CLIPasso is generated independently, the resulting sketches may not portray a gradual, smooth simplification of the sketch since each optimization process may converge to a different local minimum. By training an MLP network to \textit{learn} this gradual simplification, our resulting sketches depict a smoother simplification, where each sketch is a simplified version of the previous one.

\begin{figure*}
    \setlength{\tabcolsep}{14pt}
    \renewcommand{\arraystretch}{1}
    \centering
    
    \begin{tabular}{c c}

    \includegraphics[width=0.4\linewidth]{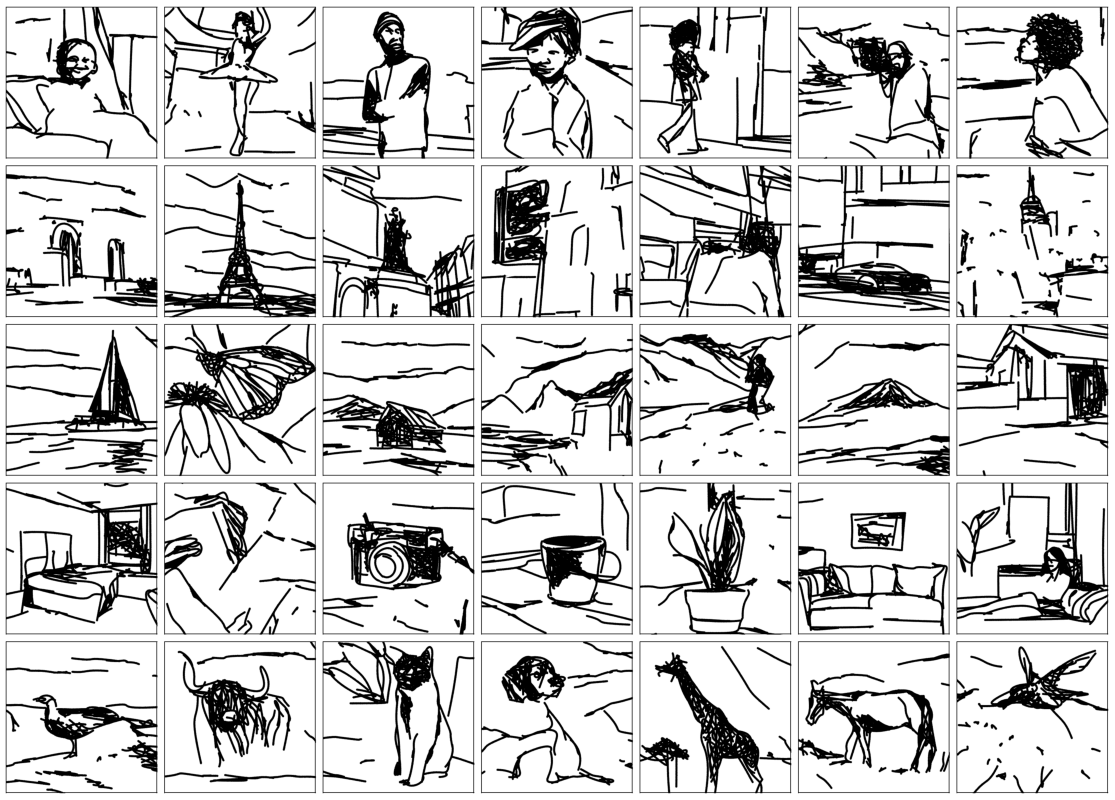} & \includegraphics[width=0.4\linewidth]{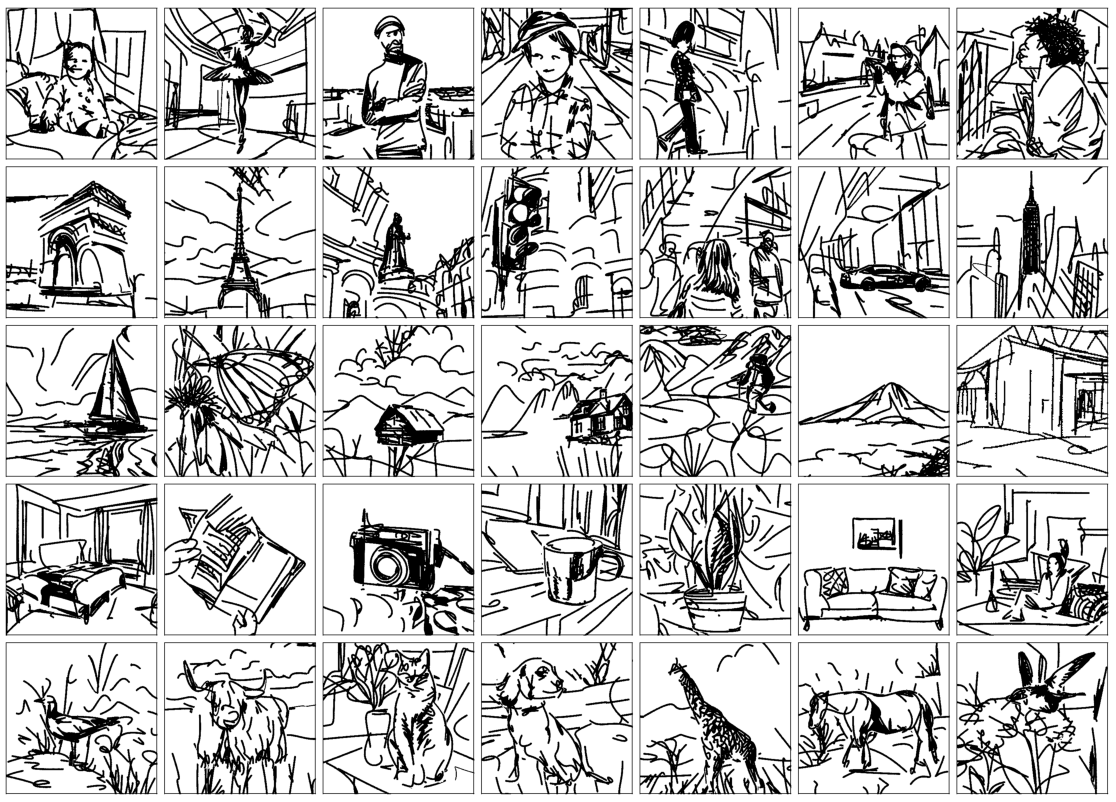} \\

    \includegraphics[width=0.4\linewidth]{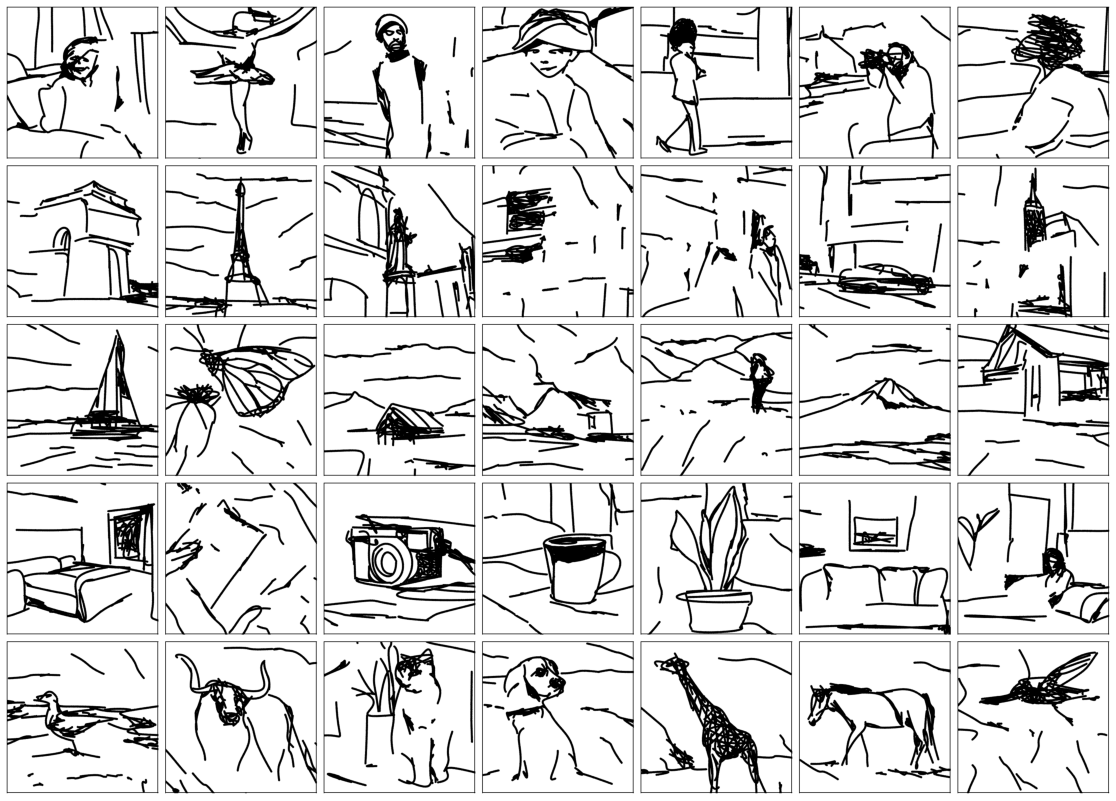} & \includegraphics[width=0.4\linewidth]{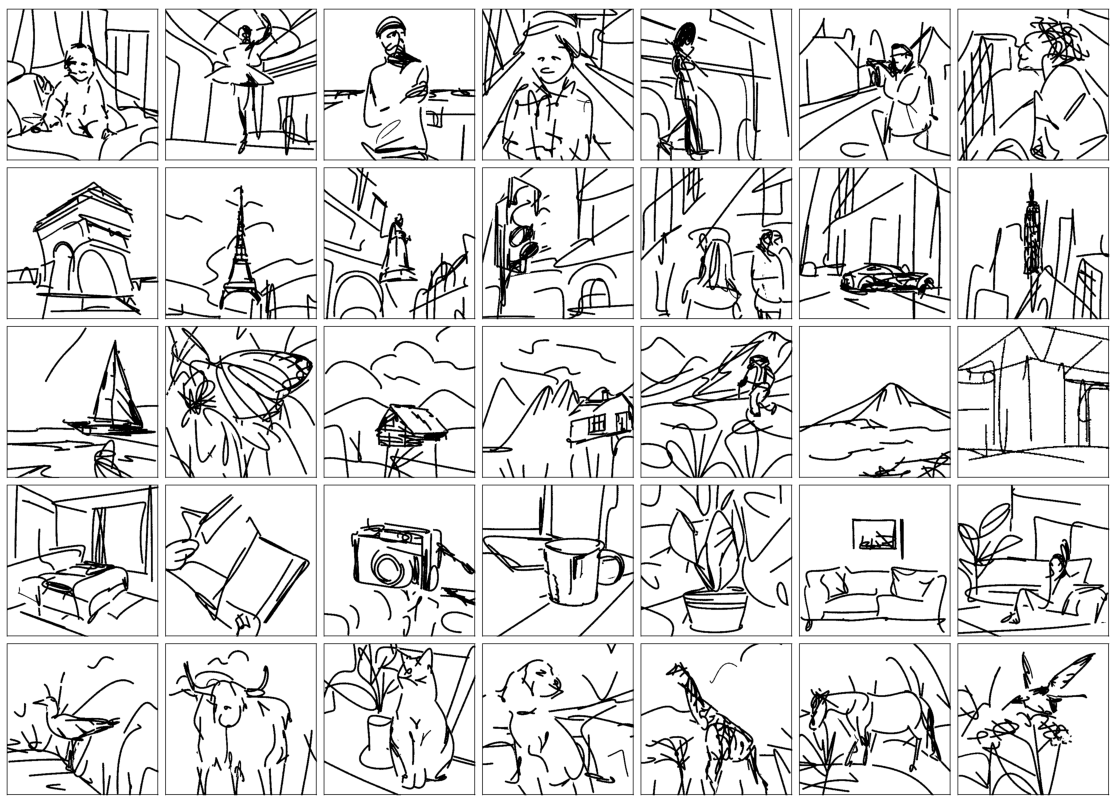} \\

    \includegraphics[width=0.4\linewidth]{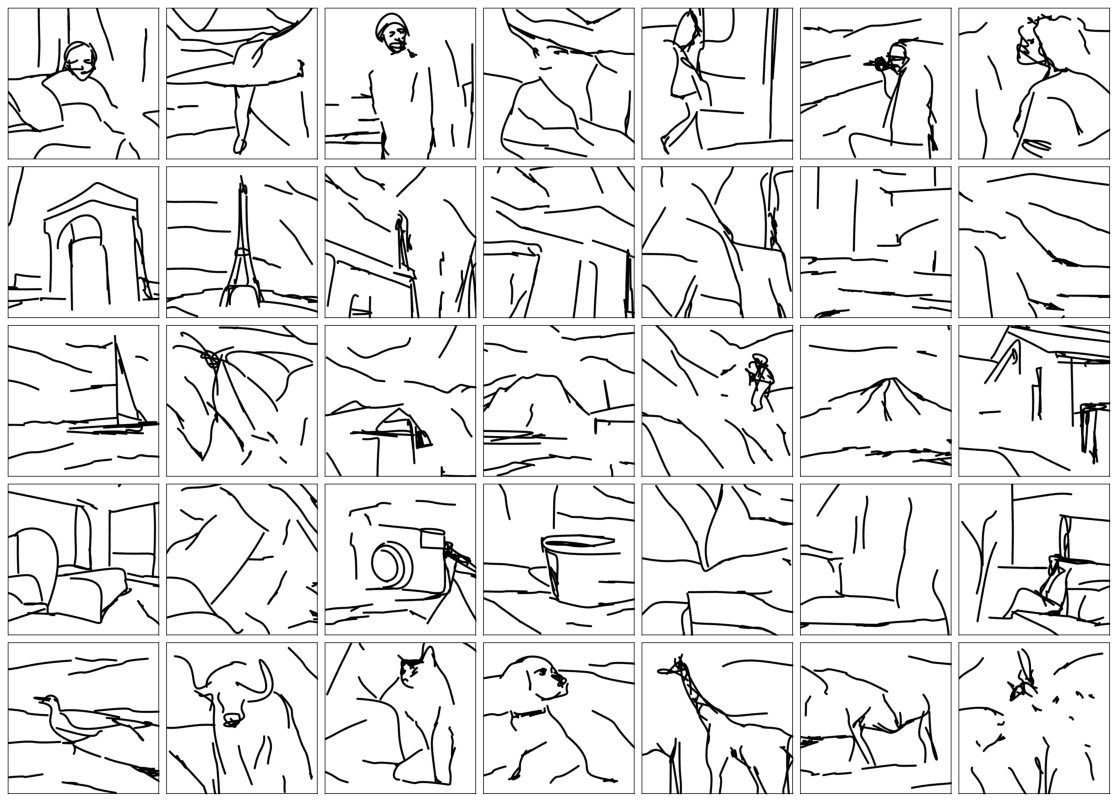} & \includegraphics[width=0.4\linewidth]{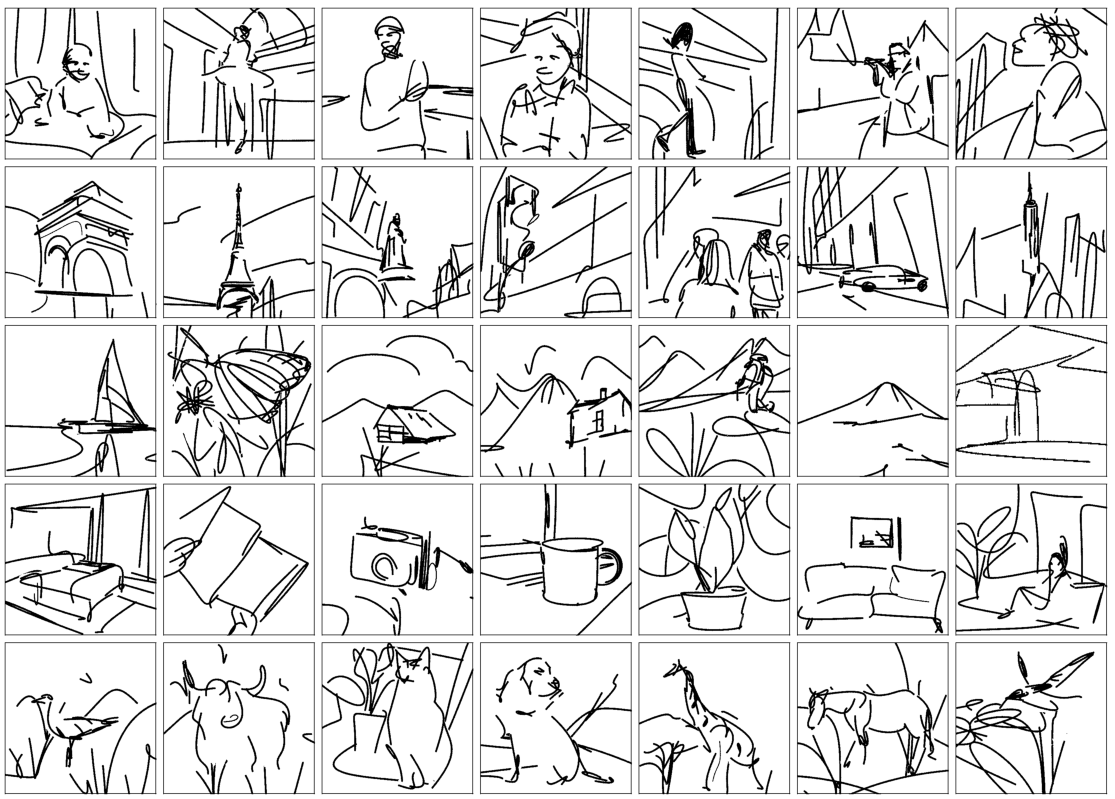} \\

    \includegraphics[width=0.4\linewidth]{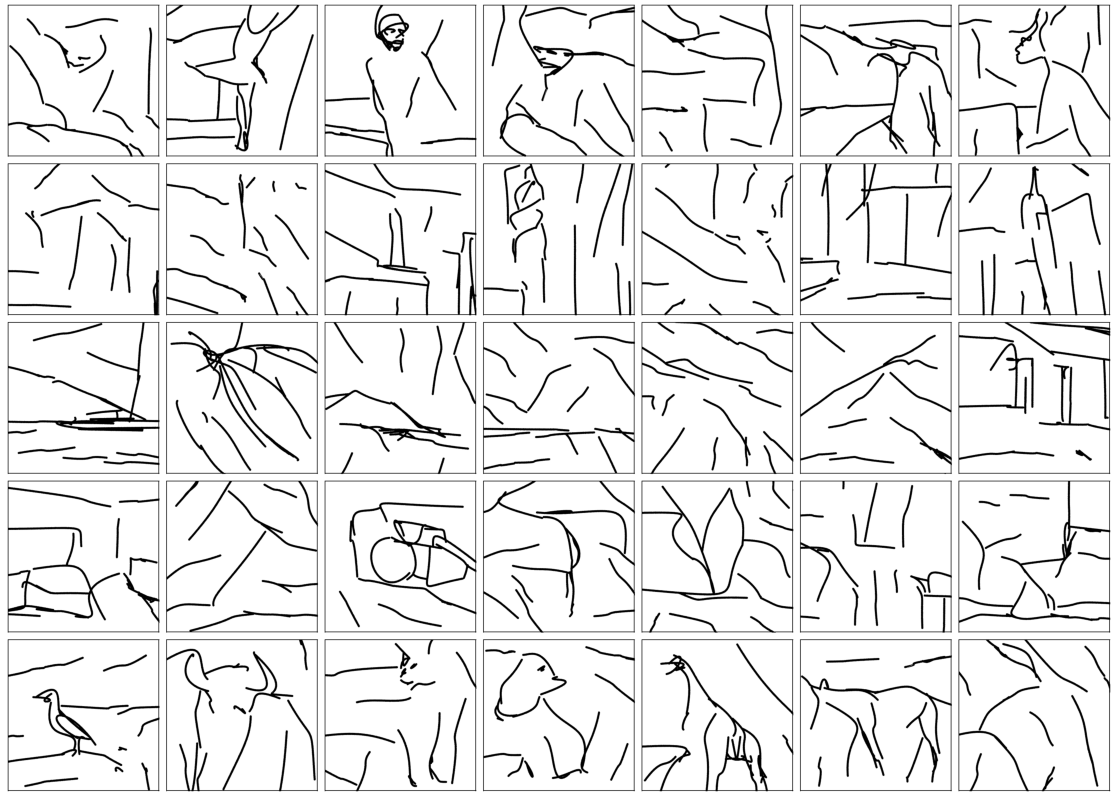} & \includegraphics[width=0.4\linewidth]{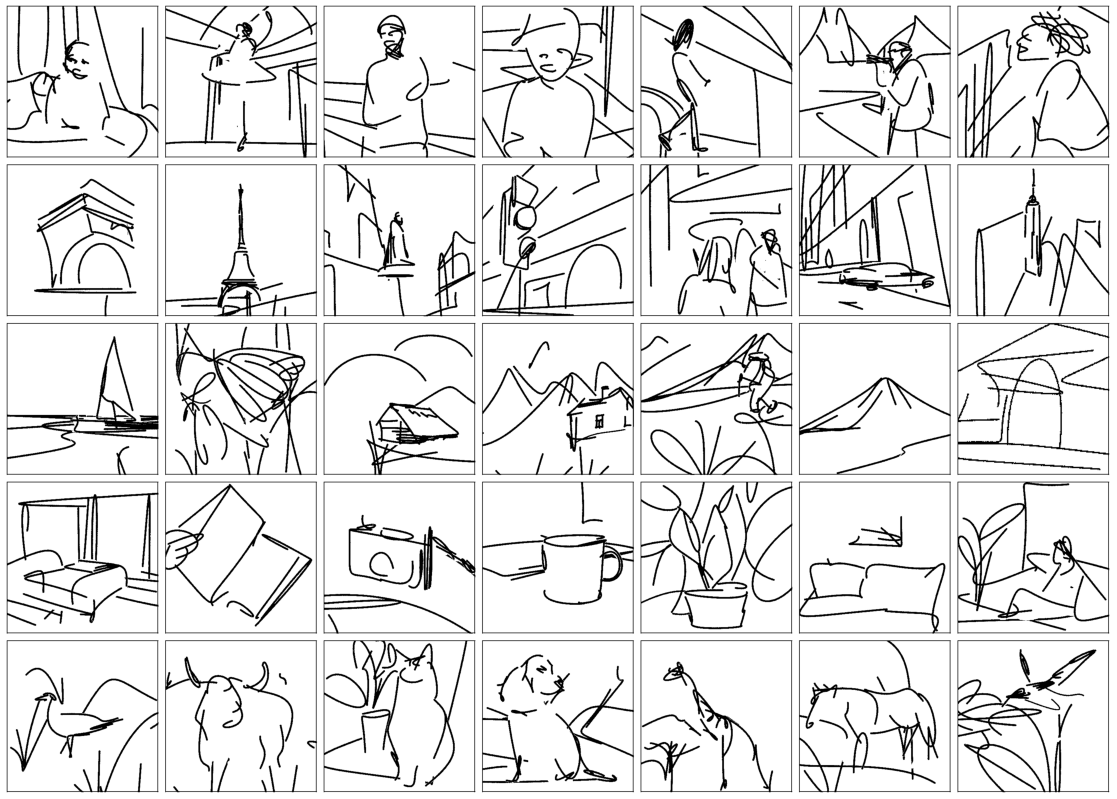} \\

    \end{tabular}
    \caption{The $35$ sketches produced for the quantitative experiment. On the left are the results of CLIPasso with four levels of abstraction, and on the right are our results with four levels of abstraction obtained using layer 11 of CLIP-ViT.}
    
    \label{fig:quant_res_clipasso}
\end{figure*}

\begin{figure*}
    \setlength{\tabcolsep}{14pt}
    \renewcommand{\arraystretch}{1}
    \centering
    
    \begin{tabular}{c c}

    \includegraphics[width=0.4\linewidth]{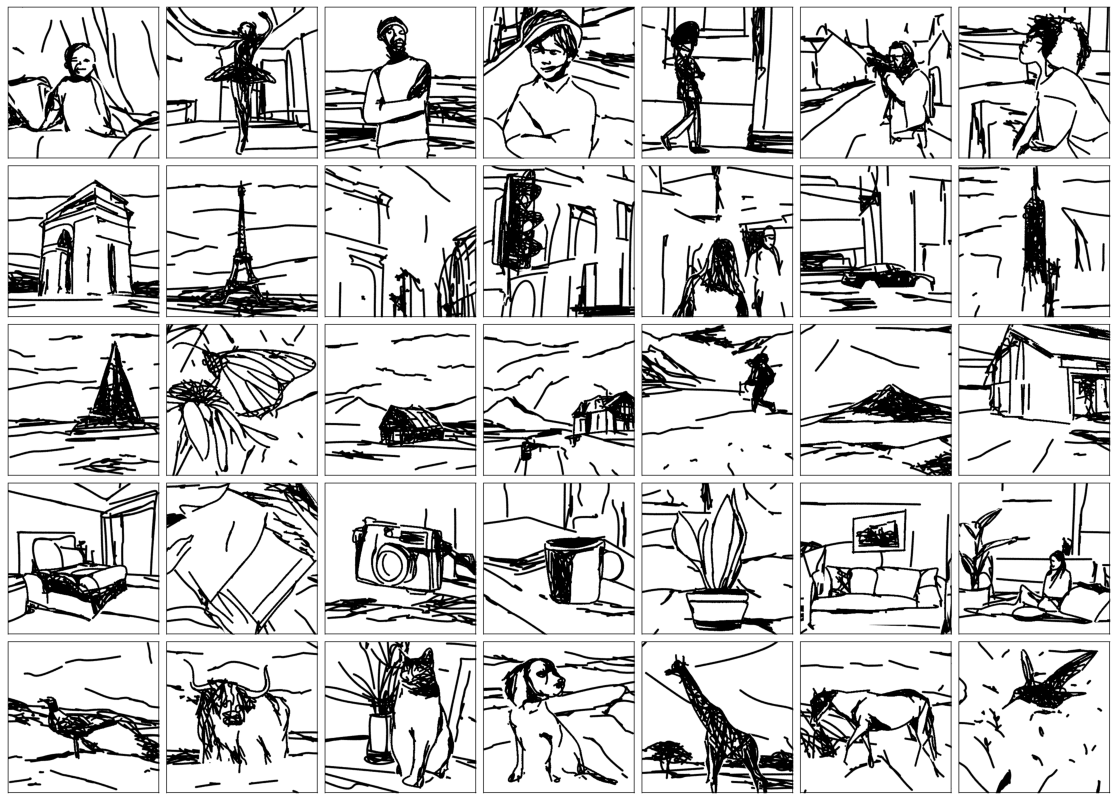} & \includegraphics[width=0.4\linewidth]{figs/quant_all/ours_12.png} \\

    \includegraphics[width=0.4\linewidth]{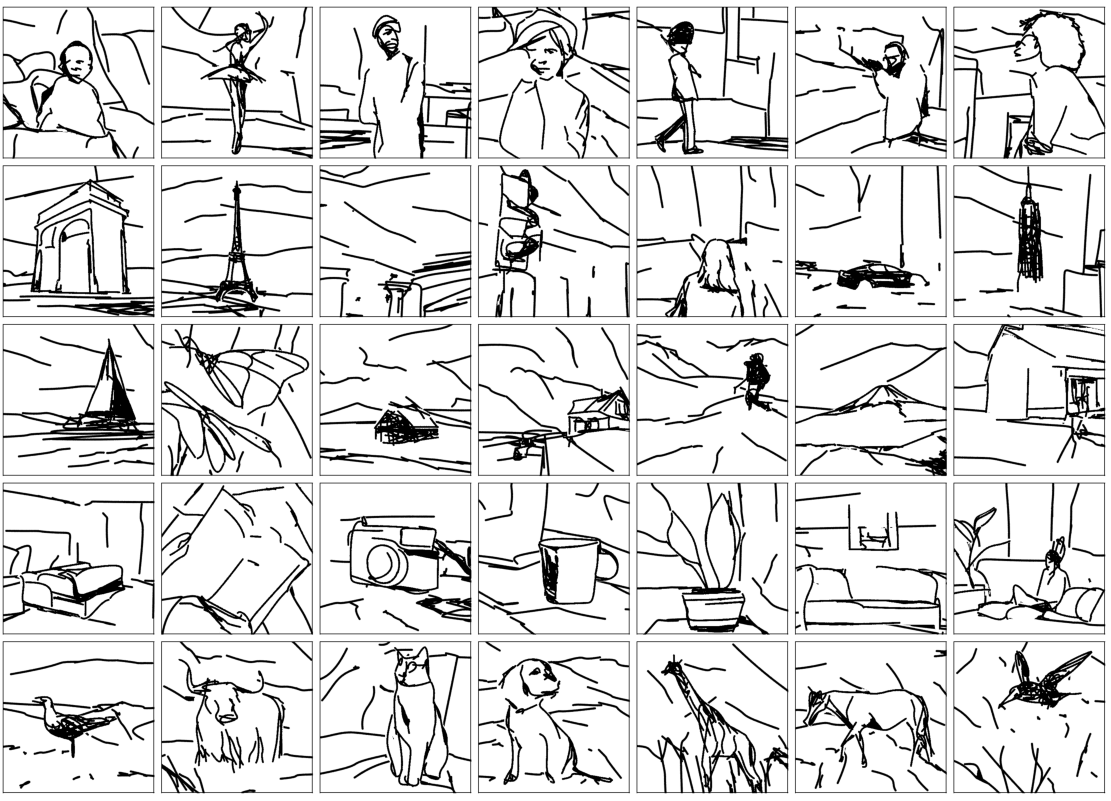} & \includegraphics[width=0.4\linewidth]{figs/quant_all/ours_13.png} \\

    \includegraphics[width=0.4\linewidth]{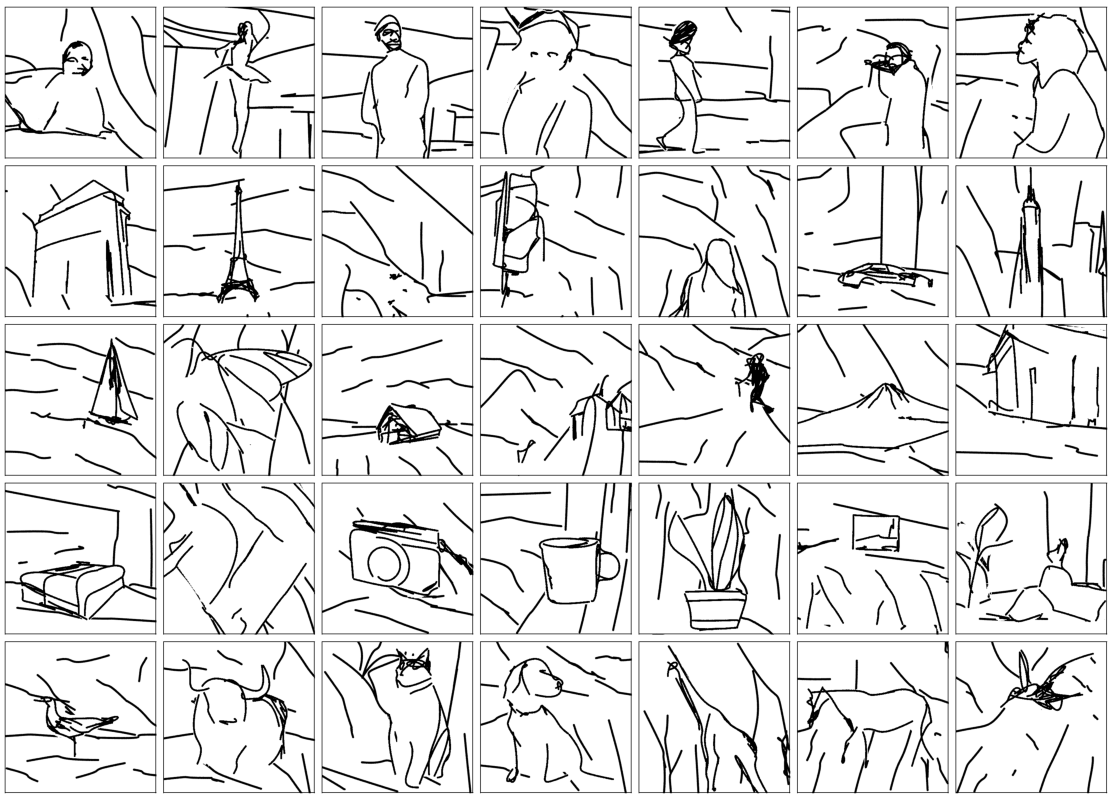} & \includegraphics[width=0.4\linewidth]{figs/quant_all/ours_14.png} \\

    \includegraphics[width=0.4\linewidth]{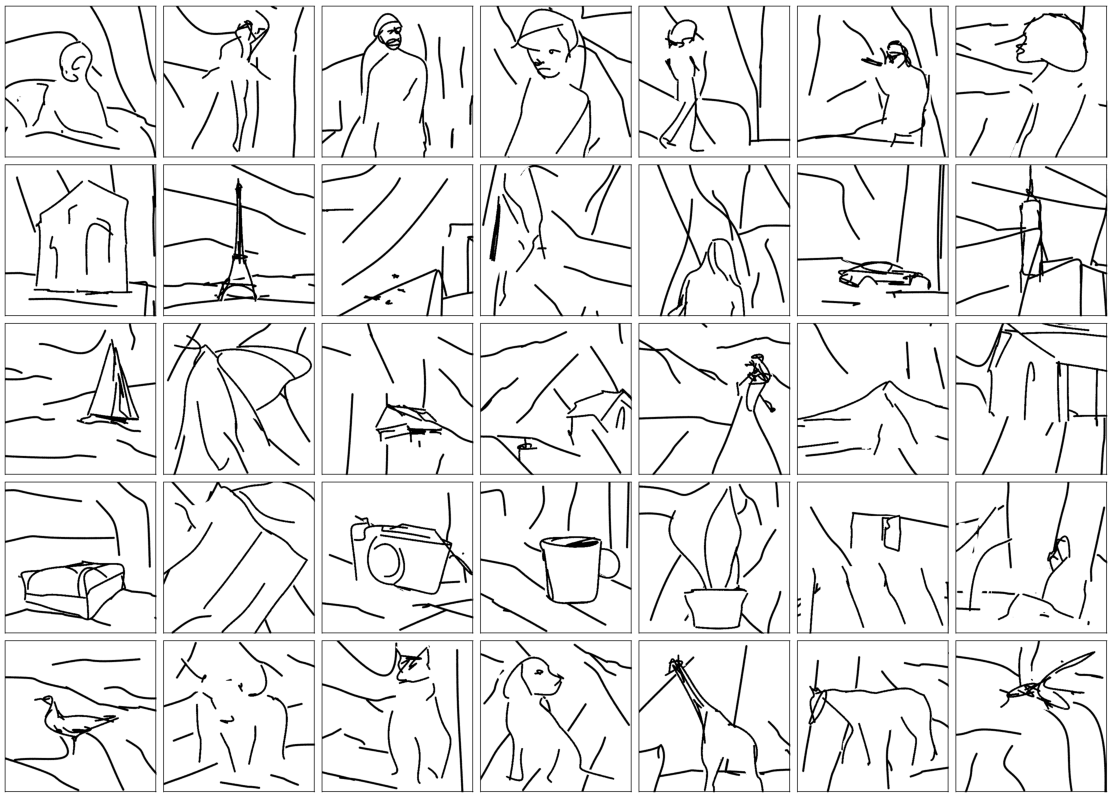} & \includegraphics[width=0.4\linewidth]{figs/quant_all/ours_15.png} \\

    \end{tabular}
    \caption{A comparison to CLIPasso with the scene separation technique on the $35$ images used for the quantitative experiment. On the left are the results of CLIPasso with four levels of abstraction, when we separate the image into foreground and background, and sketch each of the separately. On the right are our results with four levels of abstraction obtained using layer 11 of CLIP-ViT.}
    
    \label{fig:quant_res_clipasso_combined}
\end{figure*}

\begin{figure*}
    \setlength{\tabcolsep}{14pt}
    \renewcommand{\arraystretch}{1}
    \centering
    
    \begin{tabular}{c c}

    \includegraphics[width=0.4\linewidth]{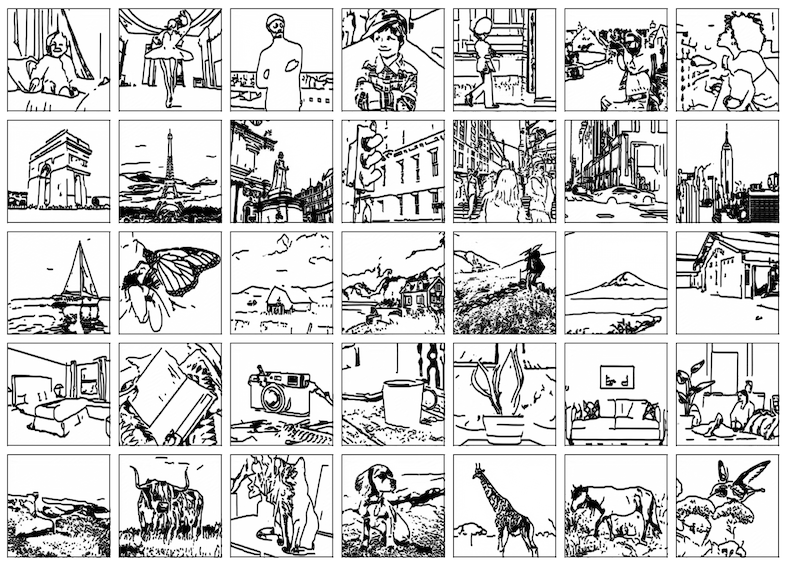} & \includegraphics[width=0.4\linewidth]{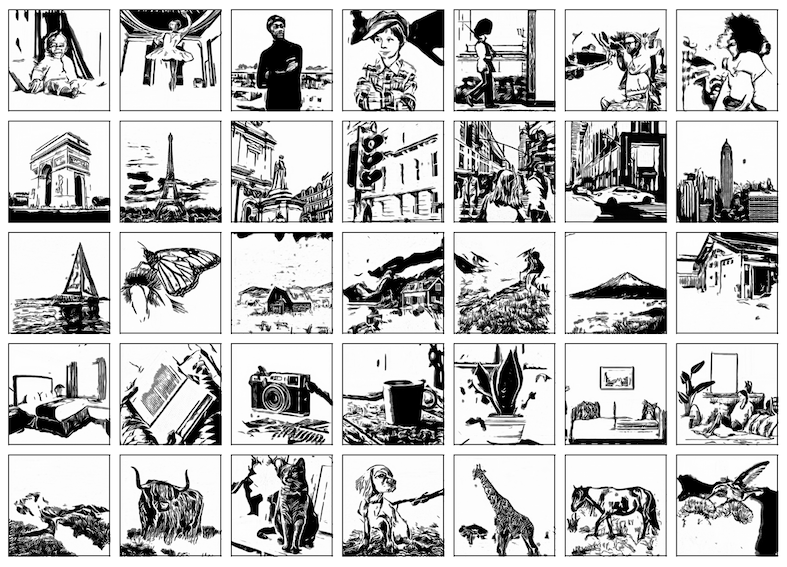} \\

    \includegraphics[width=0.4\linewidth]{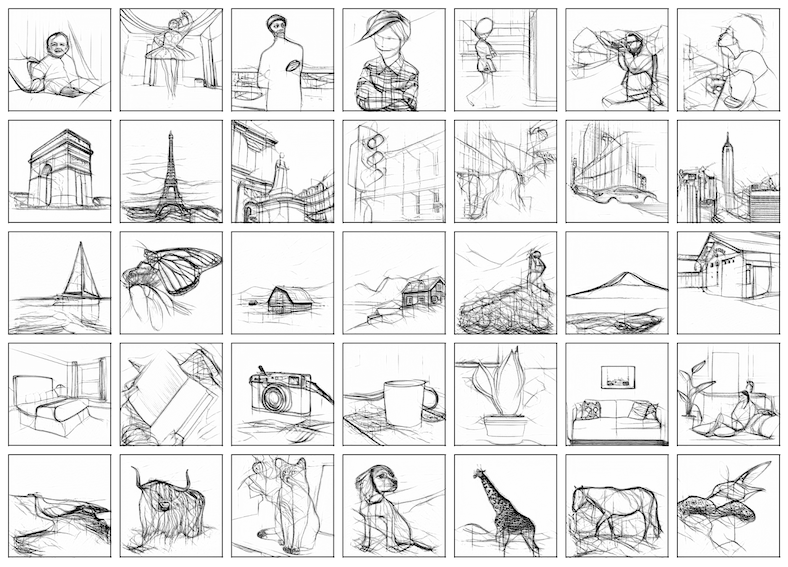} & \includegraphics[width=0.4\linewidth]{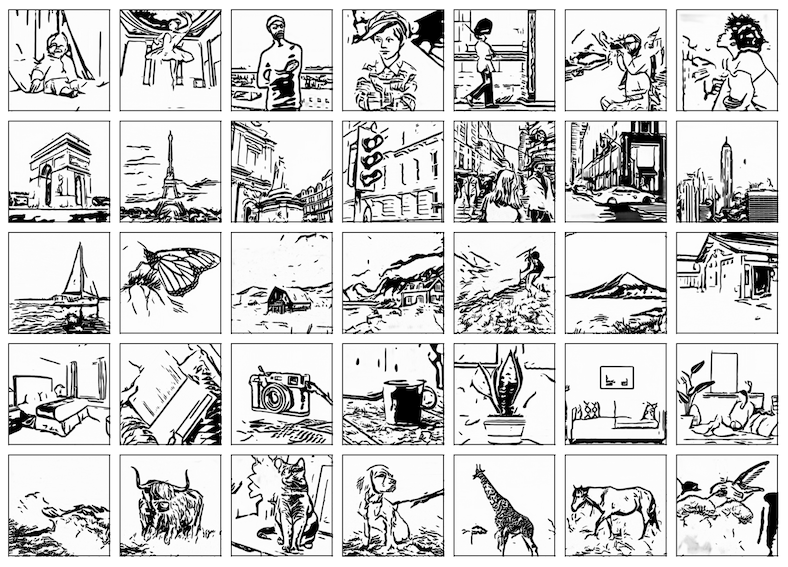} \\

    \includegraphics[width=0.4\linewidth]{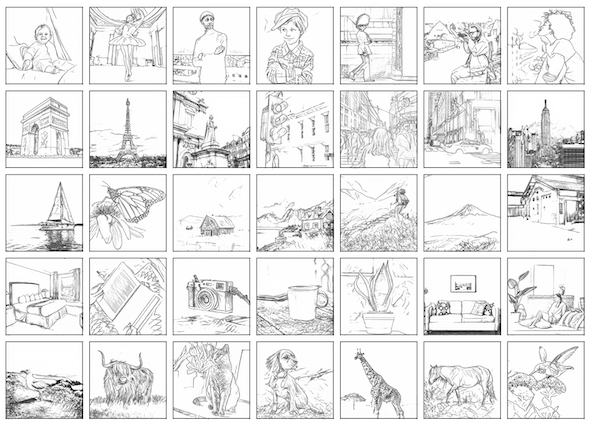} & \includegraphics[width=0.4\linewidth]{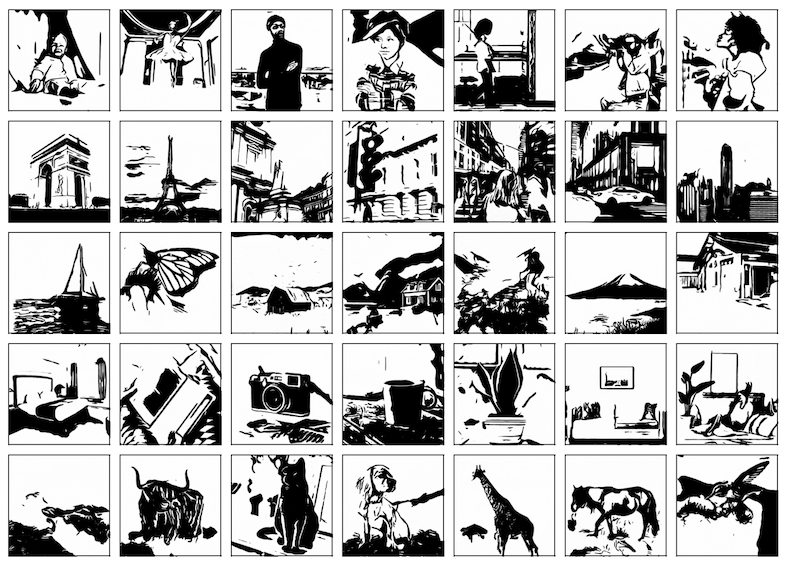} \\

    \includegraphics[width=0.4\linewidth]{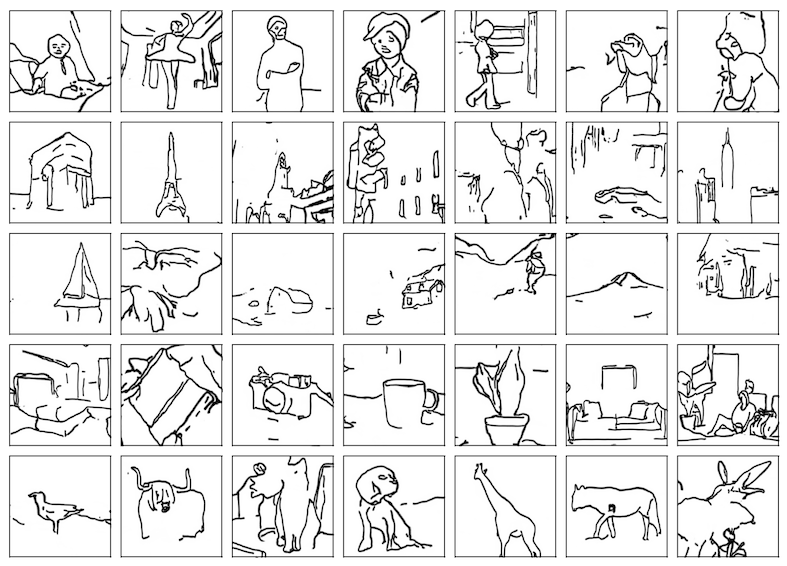} &

    \end{tabular}
    \caption{The $35$ sketches produced for the quantitative experiment. On the left are the results by Chan~\etal~\shortcite{chan2022learning} with the three provided styles, the last row on the left is by Photo-Sketch~\cite{li2019photo}. On the right are the results by UPDG~\cite{yi2020unpaired} with the three provided styles.}
    \label{fig:quant_res_chan}
\end{figure*}

In~\Cref{fig:scene_sketching_comparisons_supp} we provide additional scene sketching comparisons to alternative scene sketching methods. In~\Cref{fig:scene_sketching_comparisons_chan_supp} we provide additional sketch comparisons to all styles supported by UPDG~\cite{yi2020unpaired} and Chan~\etal~\shortcite{chan2022learning}. In~\Cref{fig:clipasso_comparisons_supp} we provide additional comparisons to CLIPasso. 
In~\Cref{fig:quant_res_clipasso,fig:quant_res_clipasso_combined,fig:quant_res_chan} we show the $35$ sketches produced by the different sketch approaches used for the quantitative experiment.
Note that in~\Cref{fig:quant_res_clipasso_combined} we show the results obtained by CLIPasso when using our scene decomposition technique, specifically, we separate the input images into foreground and background and use CLIPasso to sketch each image separately, and then combine the results.
\input{files/figures/supplementary/scene_sketch_comparison}
\input{files/figures/supplementary/chan_comparison}
\input{files/figures/supplementary/clipasso_comparison}

\null\newpage
\null\newpage
\null\newpage
\null\newpage
\null\newpage
\null\newpage

%% file: files/figures/supplementary/recognizability_example_images.tex
\begin{figure*}[hbt!]
    \centering
    \setlength{\belowcaptionskip}{-6pt}
    \setlength{\tabcolsep}{1.5pt}
    {\small
    \begin{tabular}{c c@{\hspace{0.2cm}} | c c@{\hspace{0.2cm}} | c c@{\hspace{0.2cm}} | c c@{\hspace{0.2cm}} | c c@{\hspace{0.2cm}} }

        \multicolumn{2}{c}{\textbf{People}} &
        \multicolumn{2}{c}{\textbf{Urban}} &
        \multicolumn{2}{c}{\textbf{Nature}} &
        \multicolumn{2}{c}{\textbf{Indoor}} &
        \multicolumn{2}{c}{\textbf{Animals}} \\

        \includegraphics[width=0.085\textwidth]{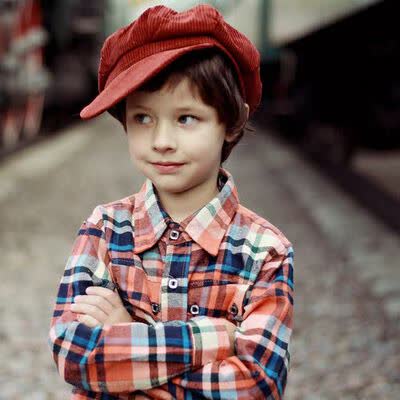} &
        \includegraphics[width=0.085\textwidth]{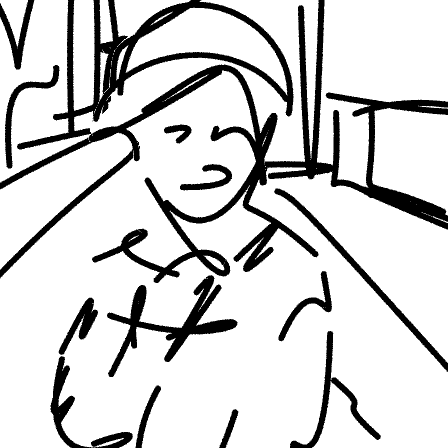} &
        \hspace{0.1cm}
        \includegraphics[width=0.085\textwidth]{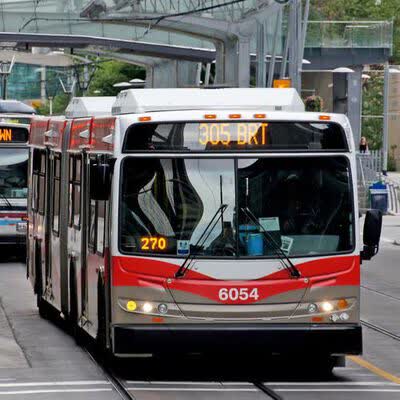} &
        \includegraphics[width=0.085\textwidth]{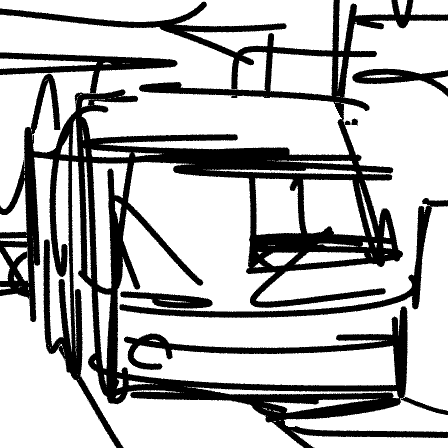} &
        \hspace{0.1cm}
        \includegraphics[height=0.0875\textwidth,width=0.085\textwidth]{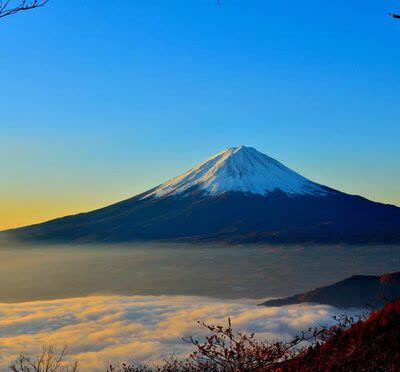} &
        \includegraphics[width=0.085\textwidth]{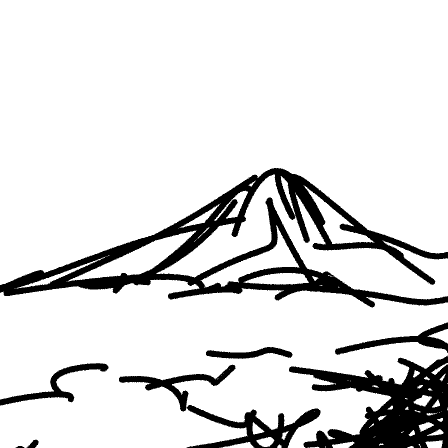} &
        \hspace{0.1cm}
        \includegraphics[width=0.085\textwidth]{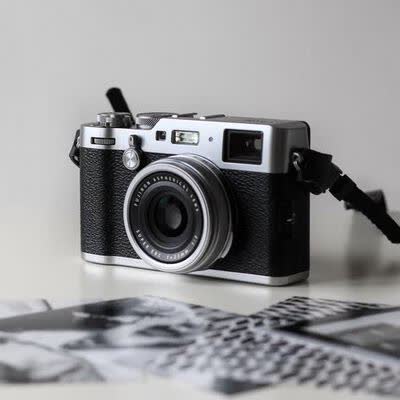} &
        \includegraphics[width=0.085\textwidth]{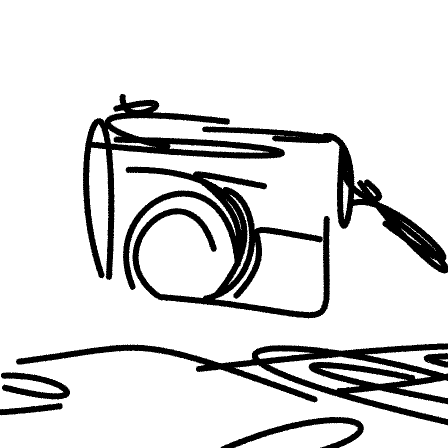} &
        \hspace{0.1cm}
        \includegraphics[width=0.085\textwidth]{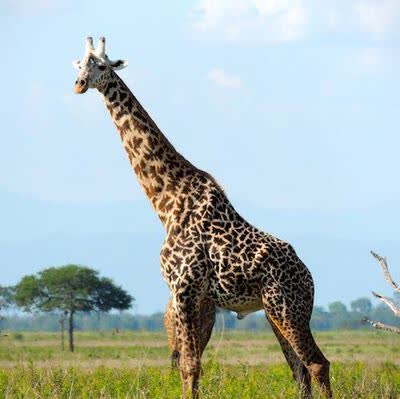} &
        \includegraphics[width=0.085\textwidth]{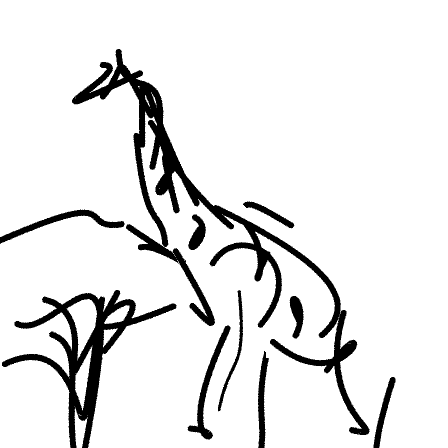} \\

        \includegraphics[width=0.085\textwidth]{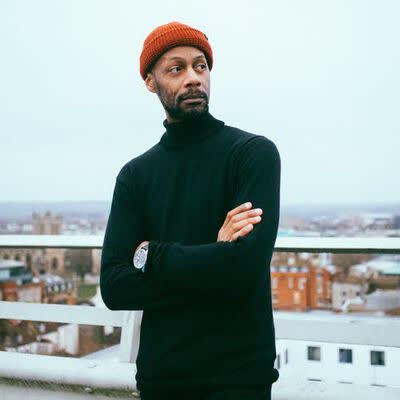} &
        \includegraphics[width=0.085\textwidth]{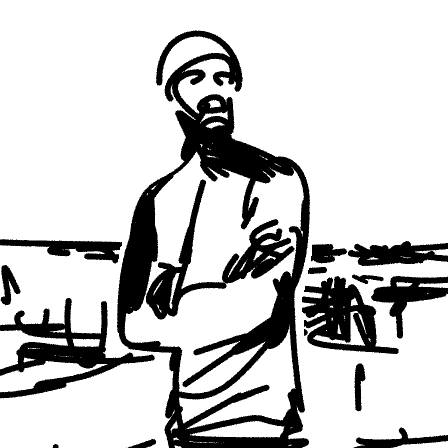} &
        \hspace{0.1cm}
        \includegraphics[width=0.085\textwidth]{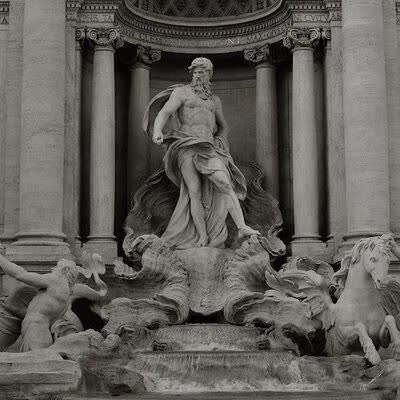} &
        \includegraphics[width=0.085\textwidth]{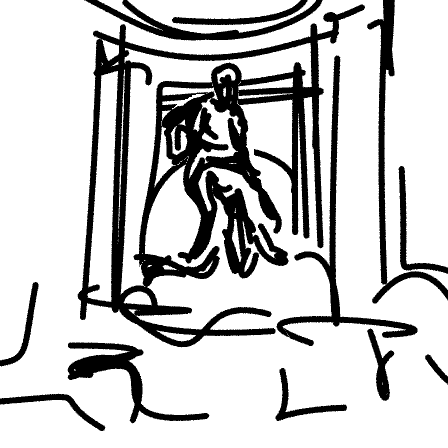} &
        \hspace{0.1cm}
        \includegraphics[width=0.085\textwidth]{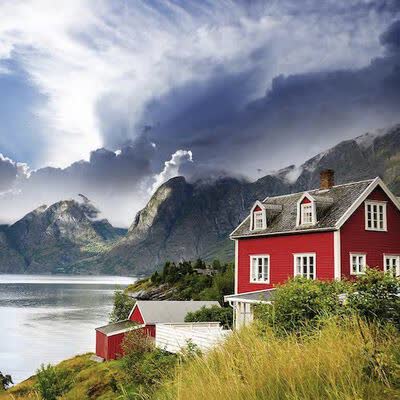} &
        \includegraphics[width=0.085\textwidth]{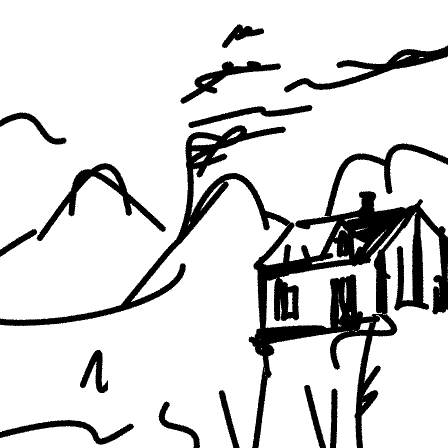} &
        \hspace{0.1cm}
        \includegraphics[width=0.085\textwidth]{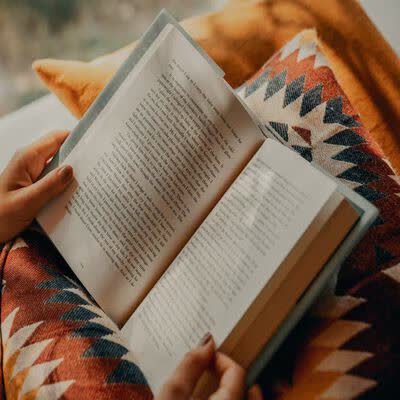} &
        \includegraphics[width=0.085\textwidth]{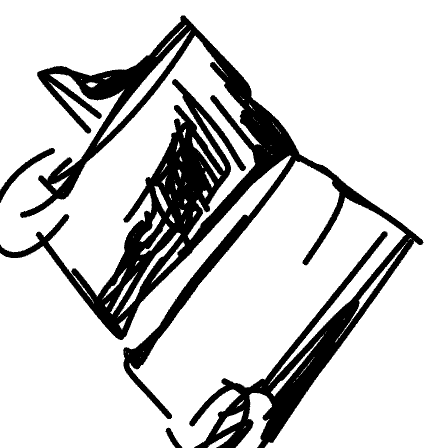} &
        \hspace{0.1cm}
        \includegraphics[width=0.085\textwidth]{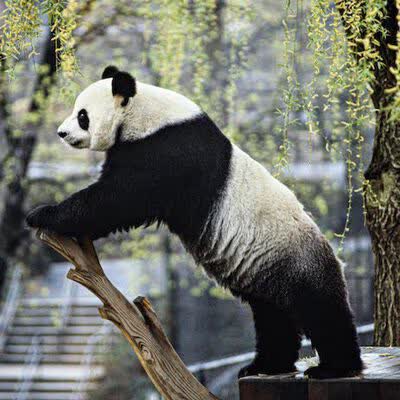} &
        \includegraphics[width=0.085\textwidth]{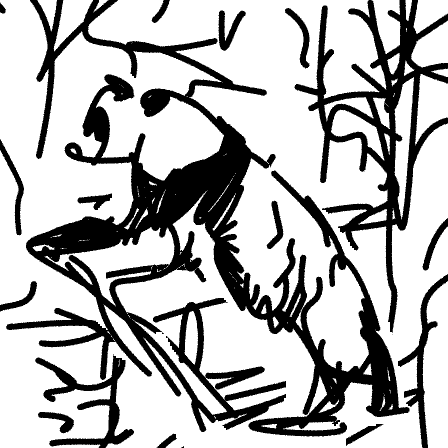} \\

        \includegraphics[width=0.085\textwidth]{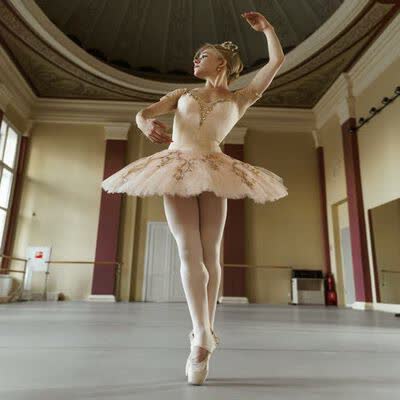} &
        \includegraphics[width=0.085\textwidth]{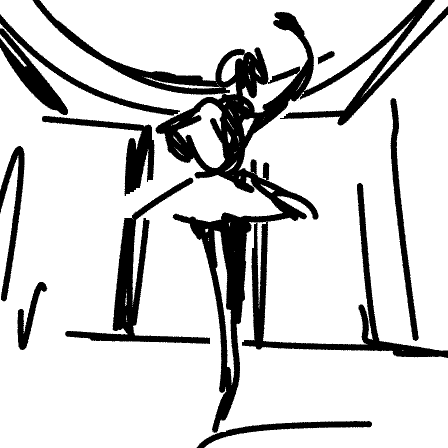} &
        \hspace{0.1cm}
        \includegraphics[height=0.0875\textwidth,width=0.085\textwidth]{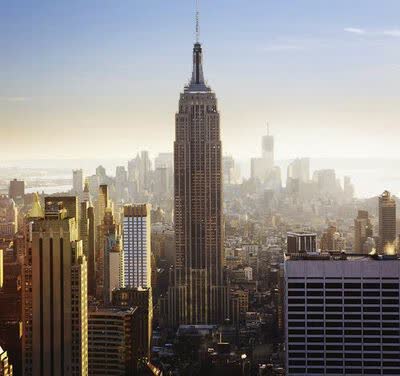} &
        \includegraphics[width=0.085\textwidth]{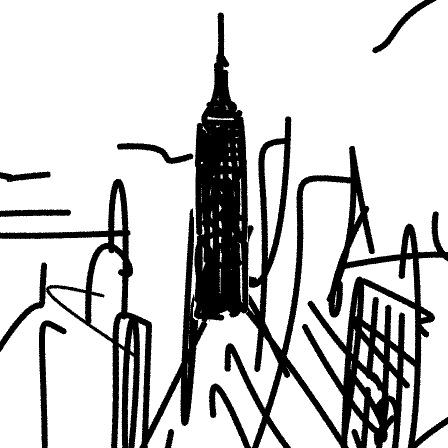} &
        \hspace{0.1cm}
        \includegraphics[width=0.085\textwidth]{figs/inputs/lighthouse-2.jpg} &
        \includegraphics[width=0.085\textwidth]{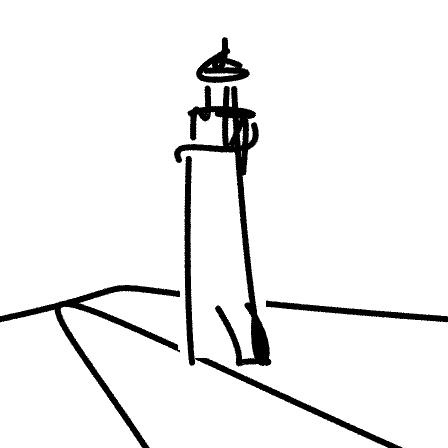} &
        \hspace{0.1cm}
        \includegraphics[width=0.085\textwidth]{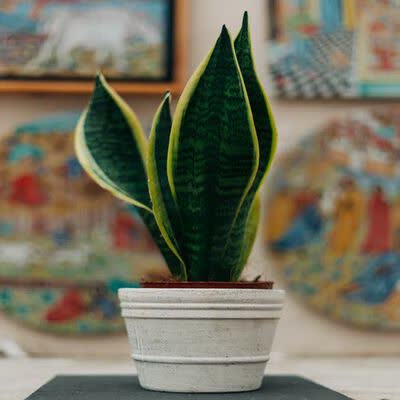} &
        \includegraphics[width=0.085\textwidth]{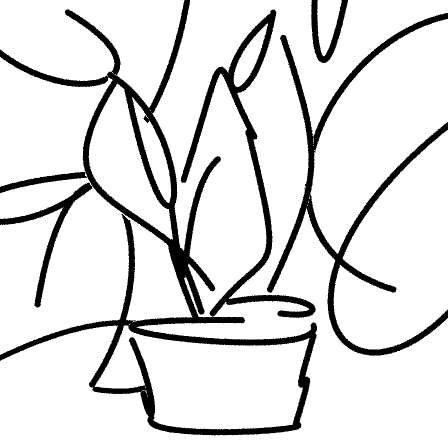} &
        \hspace{0.1cm}
        \includegraphics[width=0.085\textwidth]{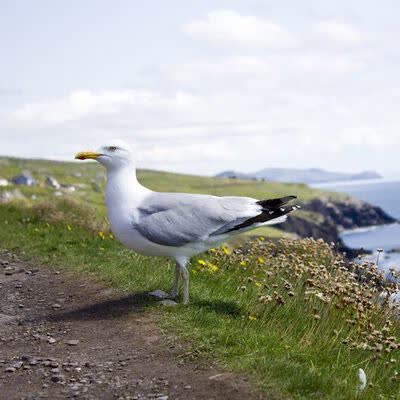} &
        \includegraphics[width=0.085\textwidth]{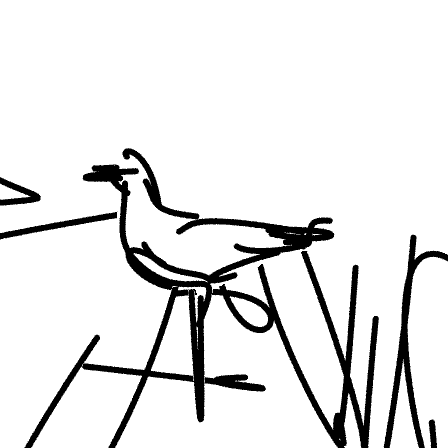} \\

        \includegraphics[width=0.085\textwidth]{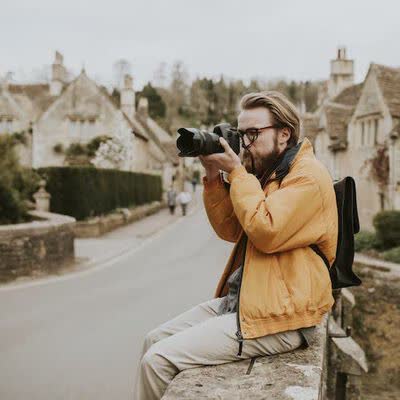} &
        \includegraphics[width=0.085\textwidth]{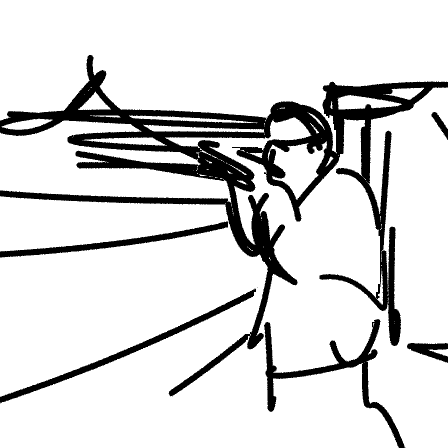} &
        \hspace{0.1cm}
        \includegraphics[width=0.085\textwidth]{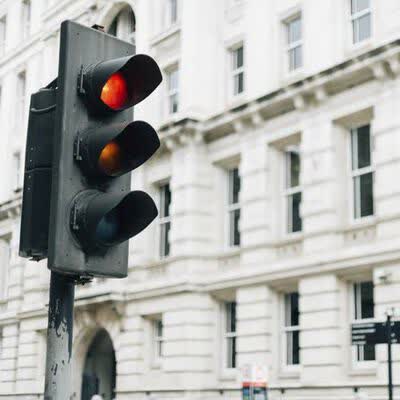} &
        \includegraphics[width=0.085\textwidth]{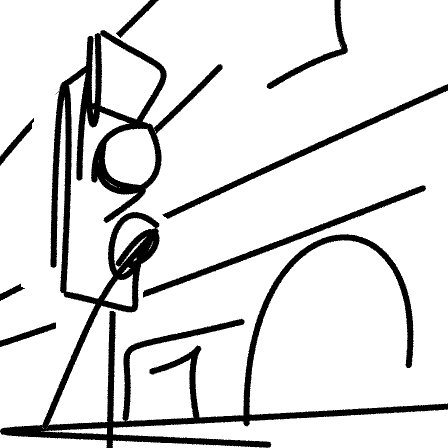} &
        \hspace{0.1cm}
        \includegraphics[width=0.085\textwidth]{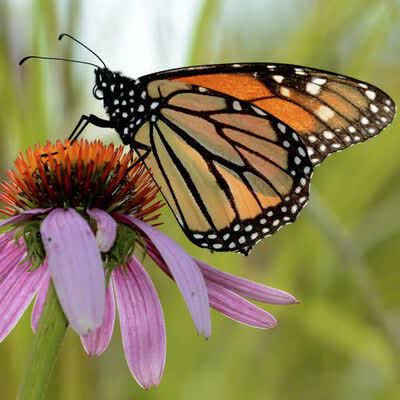} &
        \includegraphics[width=0.085\textwidth]{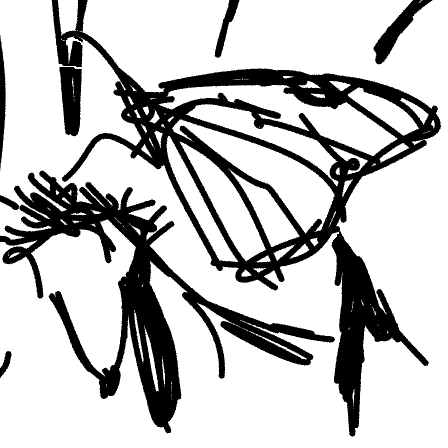} &
        \hspace{0.1cm}
        \includegraphics[width=0.085\textwidth]{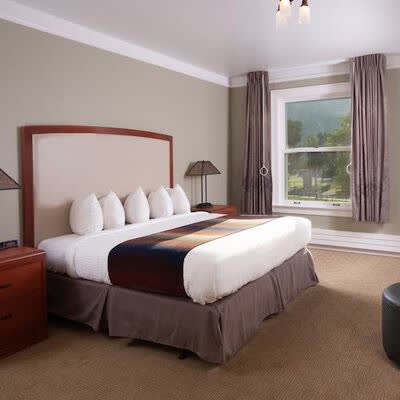} &
        \includegraphics[width=0.085\textwidth]{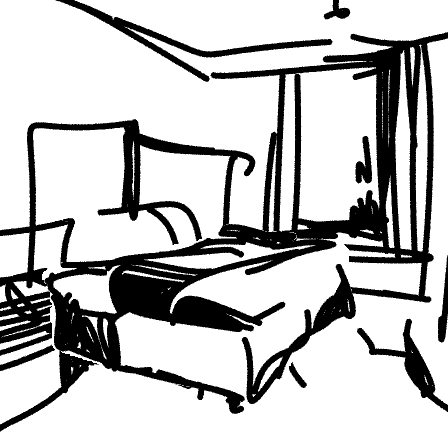} &
        \hspace{0.1cm}
        \includegraphics[width=0.085\textwidth]{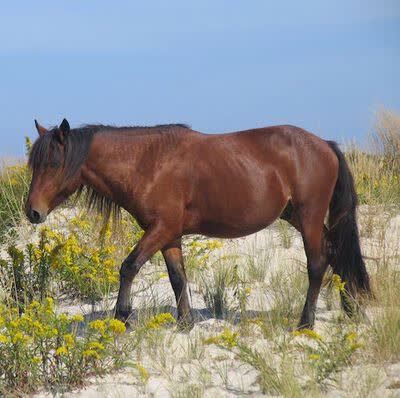} &
        \includegraphics[width=0.085\textwidth]{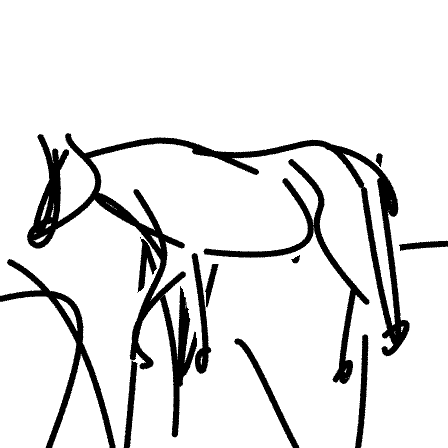} \\

    \end{tabular}
    
    }
    \caption{Example images and representative sketches used for our quantitative evaluations.}
    \label{fig:recognizability_example_images}
\end{figure*}

%% file: files/figures/quantitative_num_strokes.tex
\begin{figure}[h]
    \centering
    \setlength{\tabcolsep}{0.55pt}
    \renewcommand{\arraystretch}{1.0}
    
    \begin{tabular}{c}

    \includegraphics[width=1\linewidth]{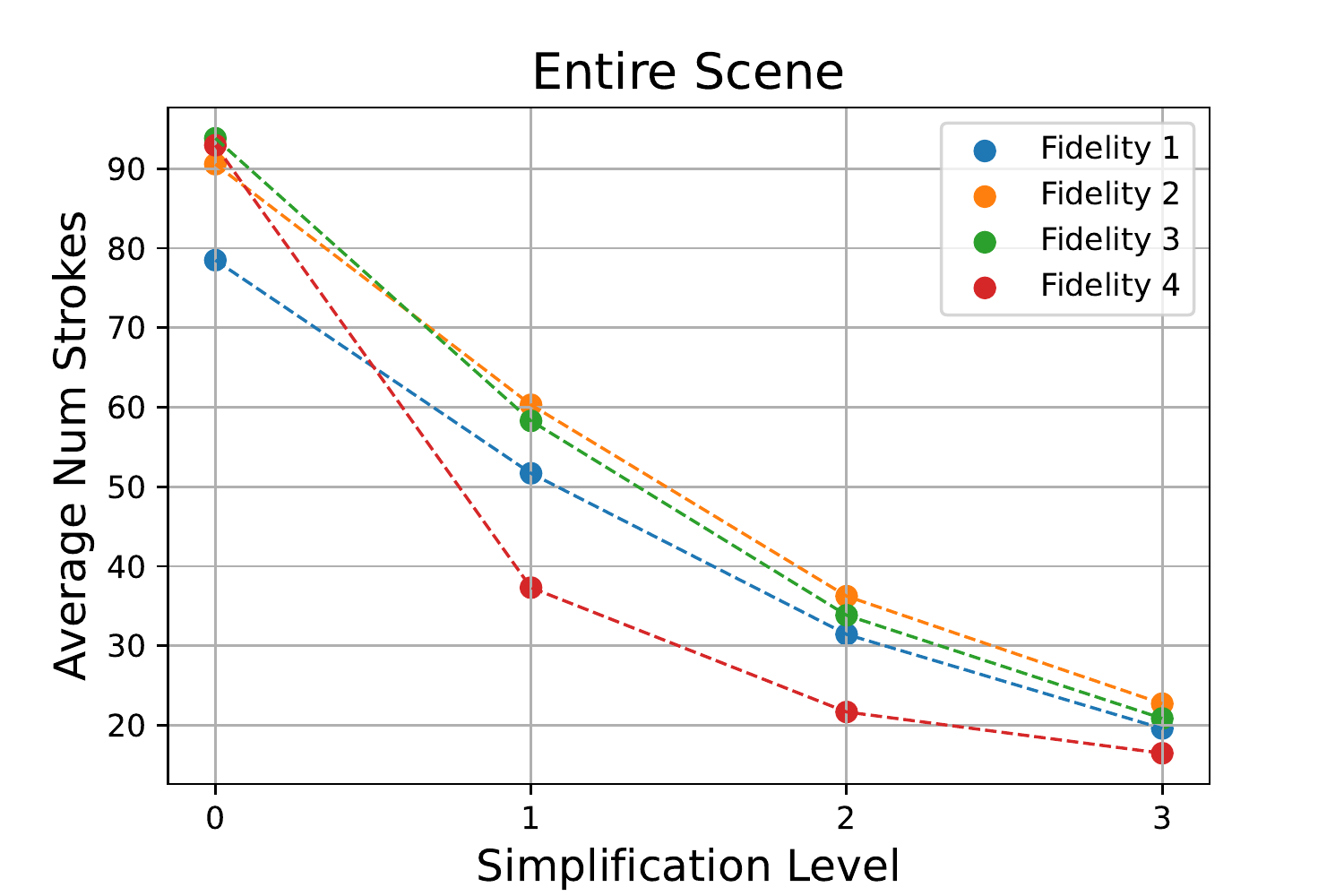}     
    \end{tabular}
    \caption{Examining the number of strokes used to compose the sketch across each fidelity and simplification level. 
    Results are averaged across all images across all scene categories. In the supplementary materials, we additionally illustrate the number of strokes split between foreground and background and between the five scene categories.}
    \label{fig:quantitative_num_strokes}
    
\end{figure}

%% file: files/figures/supplementary/recognizability_sample_results.tex
\begin{figure}[t]
    \centering
    \setlength{\tabcolsep}{2pt}
    {\small
    \begin{tabular}{c c c c}

        \\ \\ \\
        \\ \\ \\
        \\

        \raisebox{0.175in}{\rotatebox{90}{People}} &
        \includegraphics[width=0.095\textwidth]{figs/inputs/black_woman.jpg} &
        \includegraphics[width=0.095\textwidth]{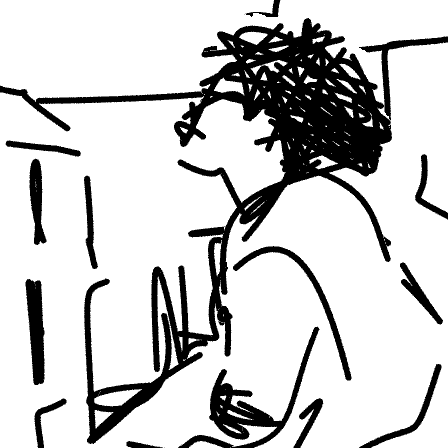} &
        \includegraphics[width=0.095\textwidth]{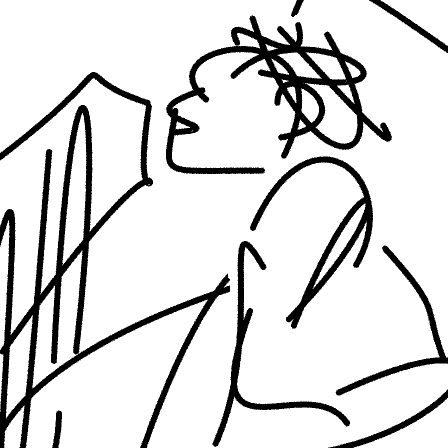} \\

        & \multicolumn{3}{l}{\textbf{Image Top 5:} person, plain, skyscraper, bridge, tower} \\
        & \multicolumn{3}{l}{\textbf{Sketch 1 Top 5:} person, laptop, skyscraper, pencil, glasses} \\        
        & \multicolumn{3}{l}{\textbf{Sketch 2 Top 5:} skyscraper, person, tower, statue, hair drier} \\

        \raisebox{0.175in}{\rotatebox{90}{Urban}} &
        \includegraphics[width=0.095\textwidth]{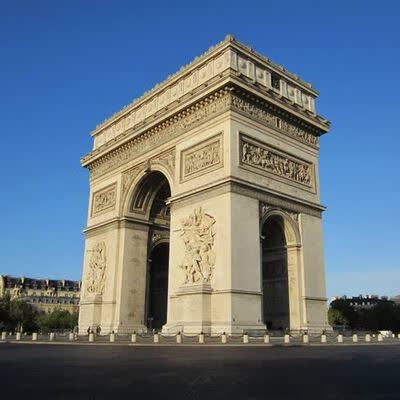} &
        \includegraphics[width=0.095\textwidth]{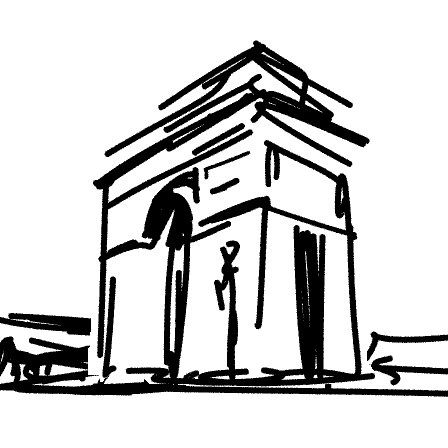} &
        \includegraphics[width=0.095\textwidth]{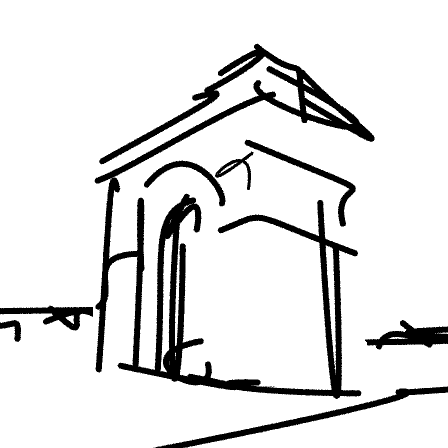} \\
        
        & \multicolumn{3}{l}{\textbf{Image Top 5:} tower, statue, plain, castle, toilet} \\
        & \multicolumn{3}{l}{\textbf{Sketch 1 Top 5:} tower, house, statue, toilet, oven} \\        
        & \multicolumn{3}{l}{\textbf{Sketch 2 Top 5:} house, tower, castle, bridge, lighthouse} \\

        \raisebox{0.175in}{\rotatebox{90}{Nature}} &
        \includegraphics[width=0.095\textwidth]{figs/inputs/house4.jpg} &
        \includegraphics[width=0.095\textwidth]{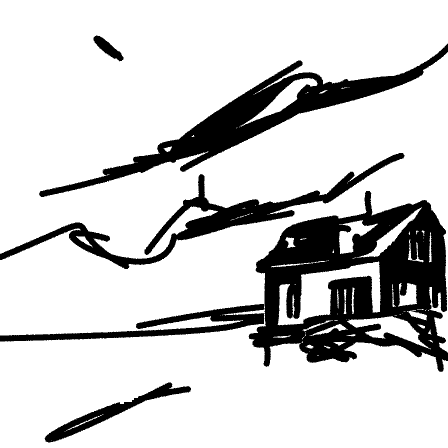} &
        \includegraphics[width=0.095\textwidth]{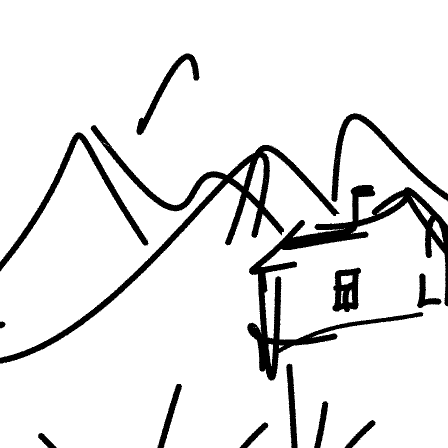} \\

        & \multicolumn{3}{l}{\textbf{Image Top 5:} house, lighthouse, mountain, sea, plain} \\
        & \multicolumn{3}{l}{\textbf{Sketch 1 Top 5:} house, lighthouse, mountain, plain, kite} \\        
        & \multicolumn{3}{l}{\textbf{Sketch 2 Top 5:} mountain, house, plain, snowboard, castle} \\

        \raisebox{0.175in}{\rotatebox{90}{Indoor}} &
        \includegraphics[width=0.095\textwidth]{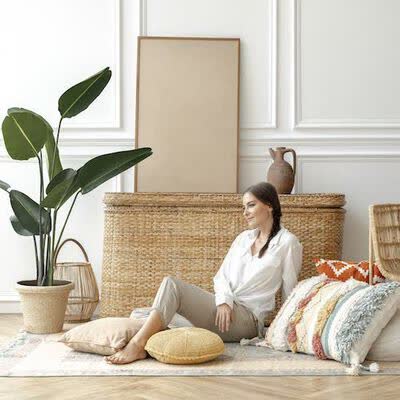} &
        \includegraphics[width=0.095\textwidth]{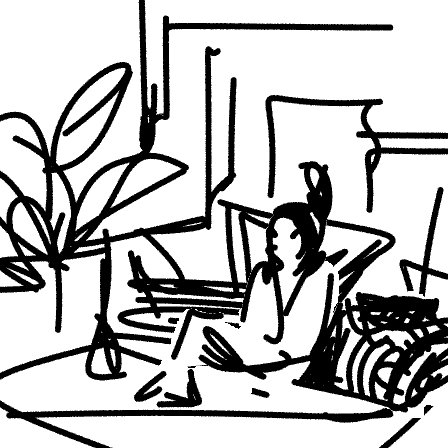} &
        \includegraphics[width=0.095\textwidth]{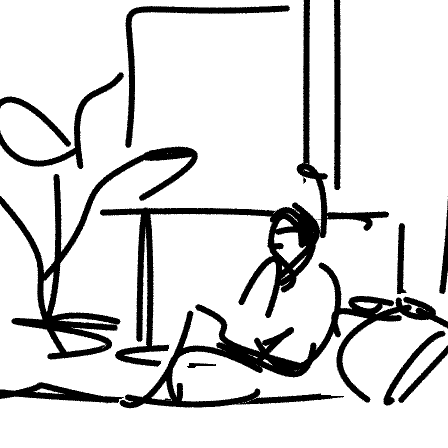} \\

        & \multicolumn{3}{l}{\textbf{Image Top 5:} bed, couch, table, vase, chair} \\
        & \multicolumn{3}{l}{\textbf{Sketch 1 Top 5:} laptop, couch, table, remote, chair} \\        
        & \multicolumn{3}{l}{\textbf{Sketch 2 Top 5:} plant, couch, vase, kettle, bowl} \\

        \raisebox{0.175in}{\rotatebox{90}{Animals}} &
        \includegraphics[width=0.095\textwidth]{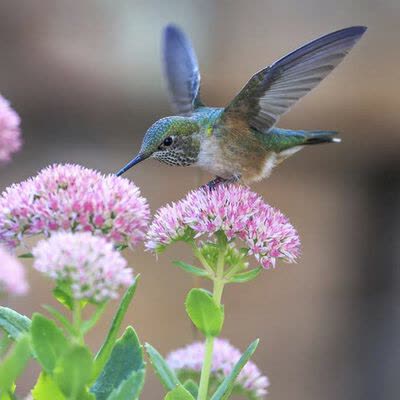} &
        \includegraphics[width=0.095\textwidth]{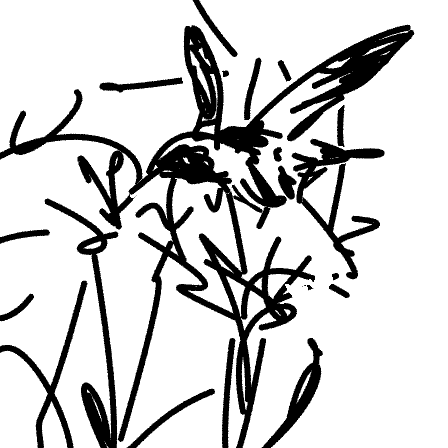} &
        \includegraphics[width=0.095\textwidth]{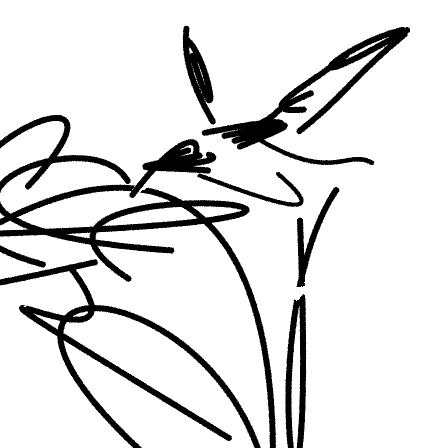} \\

        & \multicolumn{3}{l}{\textbf{Image Top 5:} bird, bee, flower, butterfly, camera} \\
        & \multicolumn{3}{l}{\textbf{Sketch 1 Top 5:} bird, bee, mouse, rabbit, butterfly} \\        
        & \multicolumn{3}{l}{\textbf{Sketch 2 Top 5:} bee, butterfly, flower, bird, plant} \\

    \end{tabular}
    
    }
    \vspace{0.2cm}
    \caption{Examples of CLIP zero-shot class predictions on various input images and representative sketches of varying abstractions. These predictions are then used to compute a recognizability metric for each scene category across different levels of abstractions (see Section 4.3 in the main paper).}
    \label{fig:recognizability_example_predictions}
\end{figure}

%% file: files/figures/supplementary/quantitative_num_strokes_by_class.tex
\begin{figure*}
    \centering
    \setlength{\tabcolsep}{0.55pt}
    \renewcommand{\arraystretch}{1.0}
    
    \begin{tabular}{c c c}

    \\ \\ \\ \\ \\
    
    \textbf{All Categories} &
    \textbf{People} &
    \textbf{Urban} \\

    \includegraphics[width=0.3\linewidth]{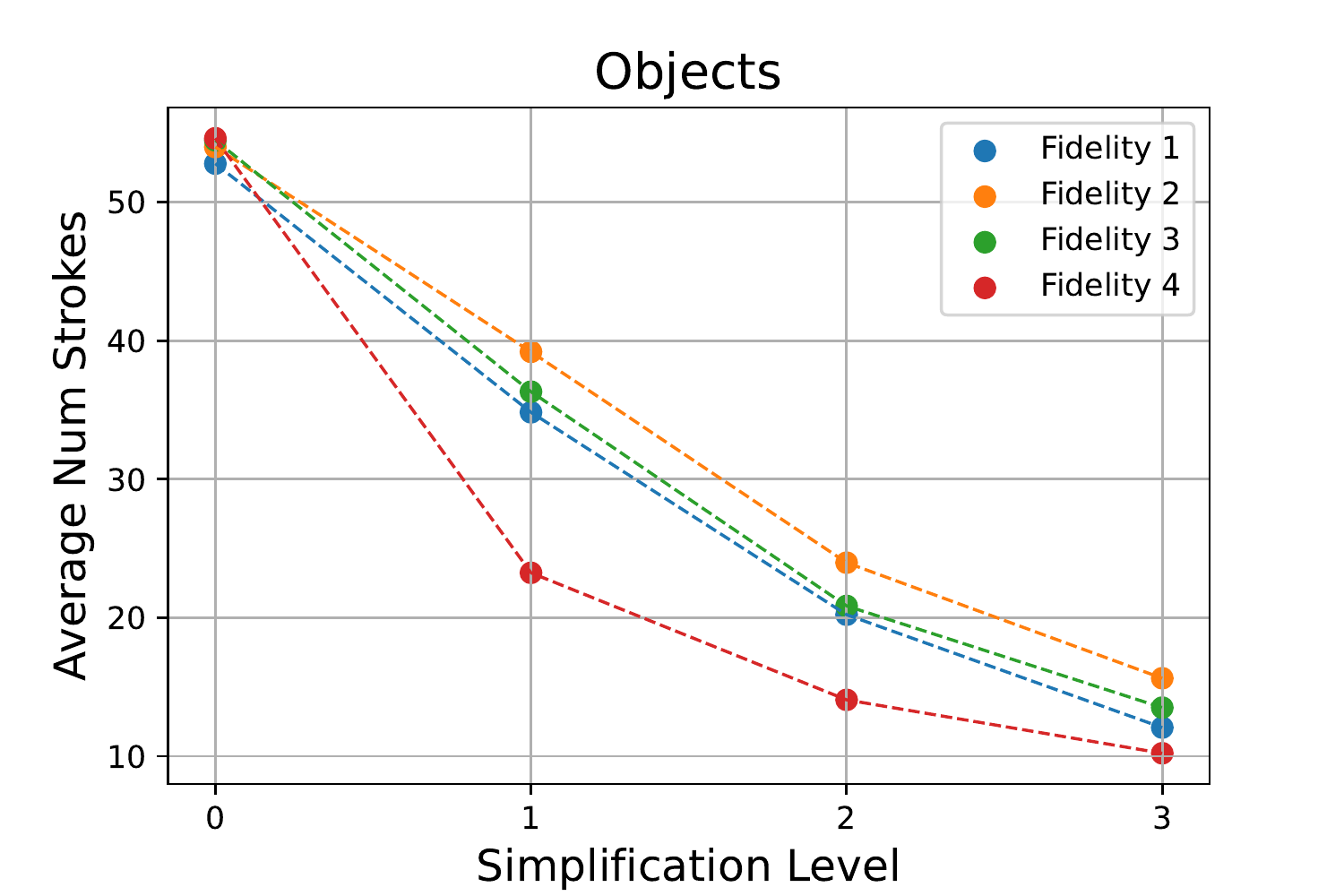} &
    \includegraphics[width=0.3\linewidth]{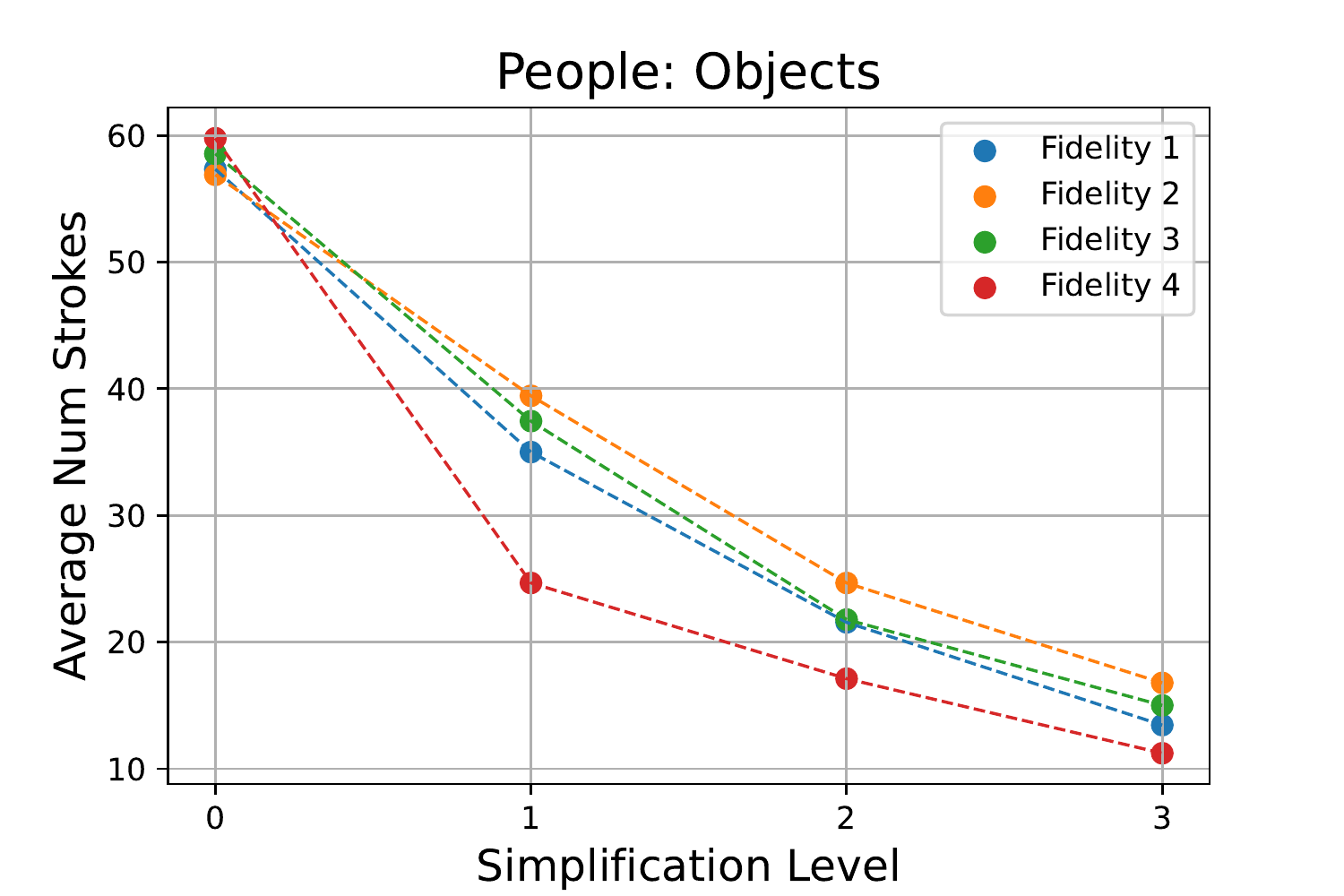} &
    \includegraphics[width=0.3\linewidth]{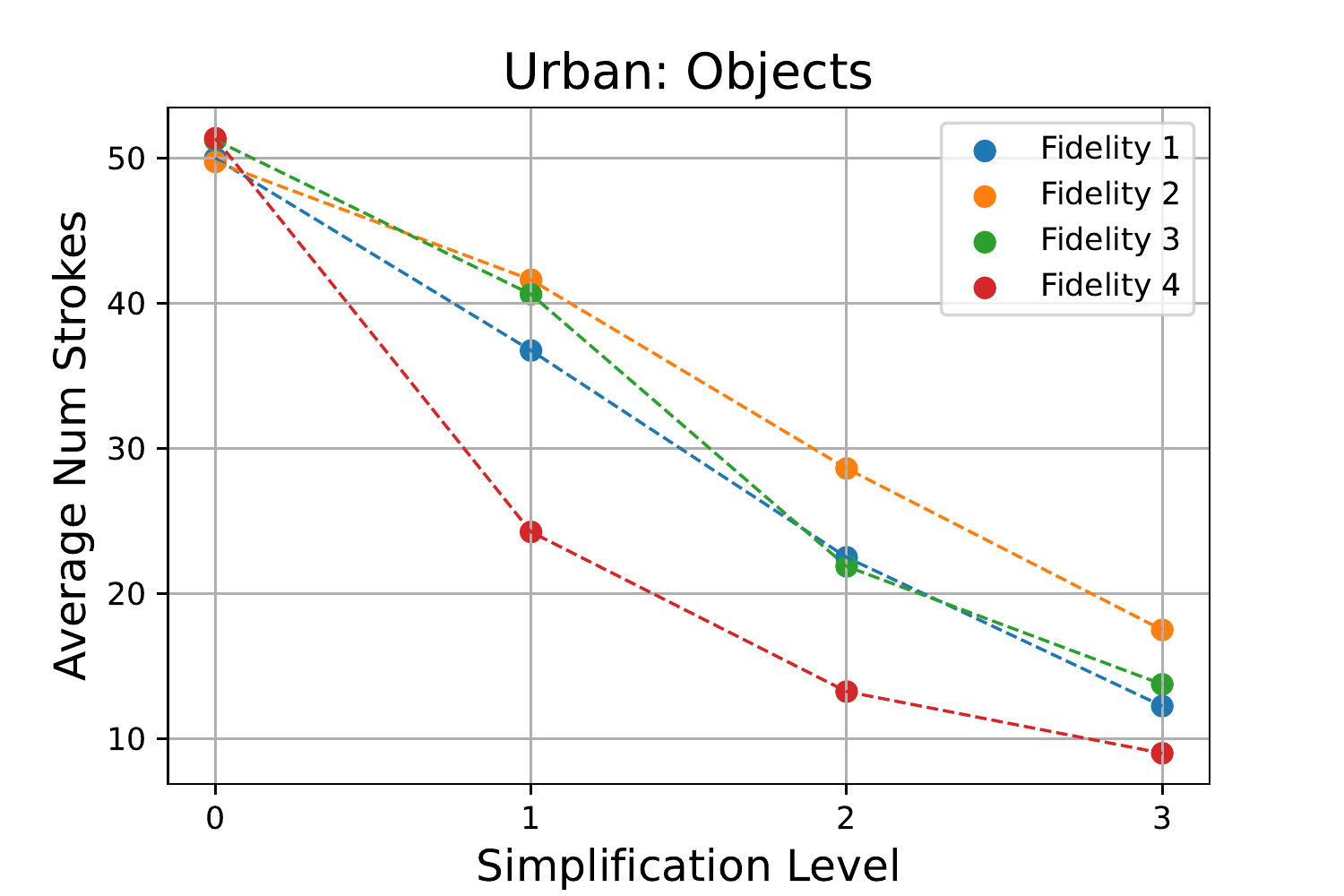} \\

    \includegraphics[width=0.3\linewidth]{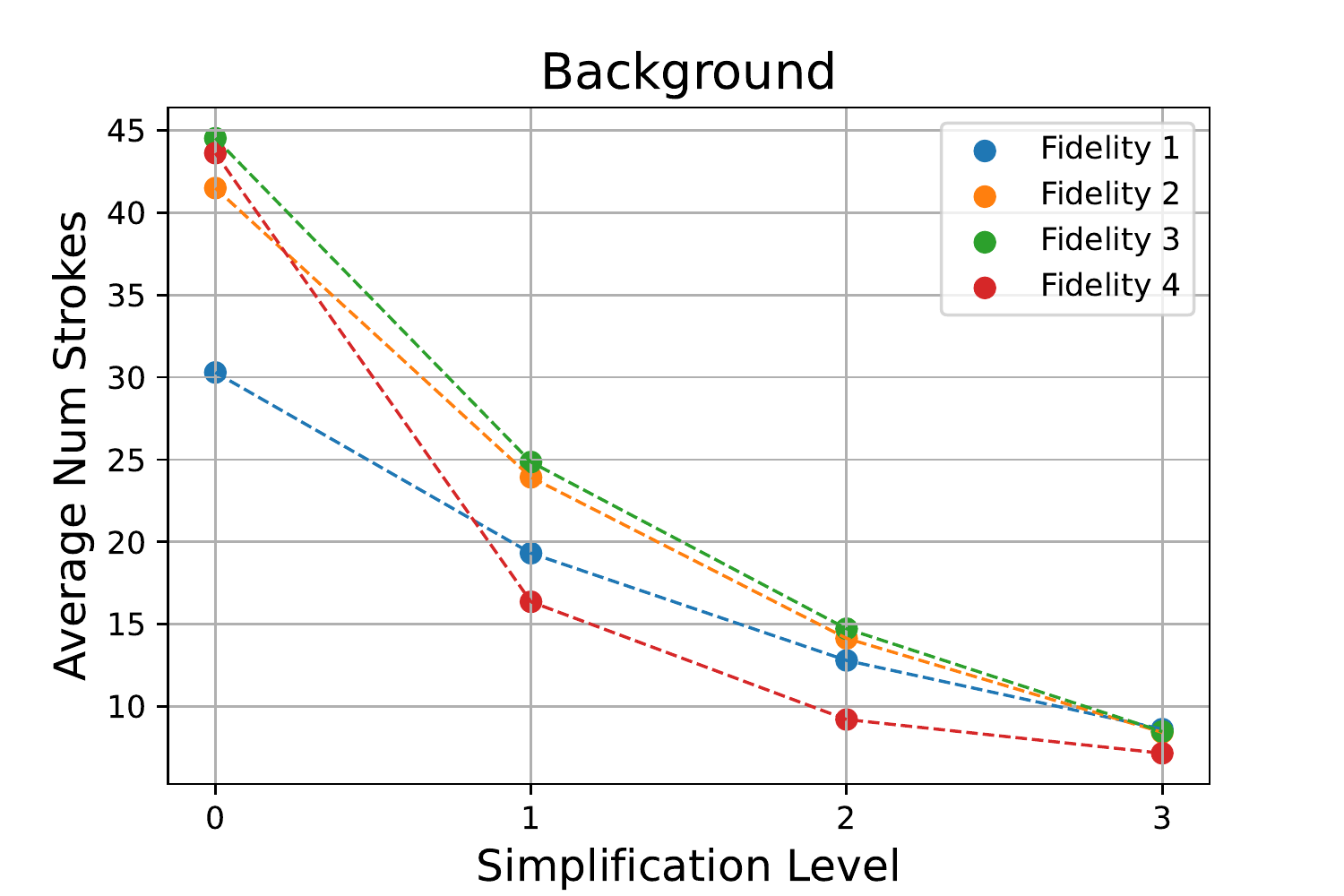} &
    \includegraphics[width=0.3\linewidth]{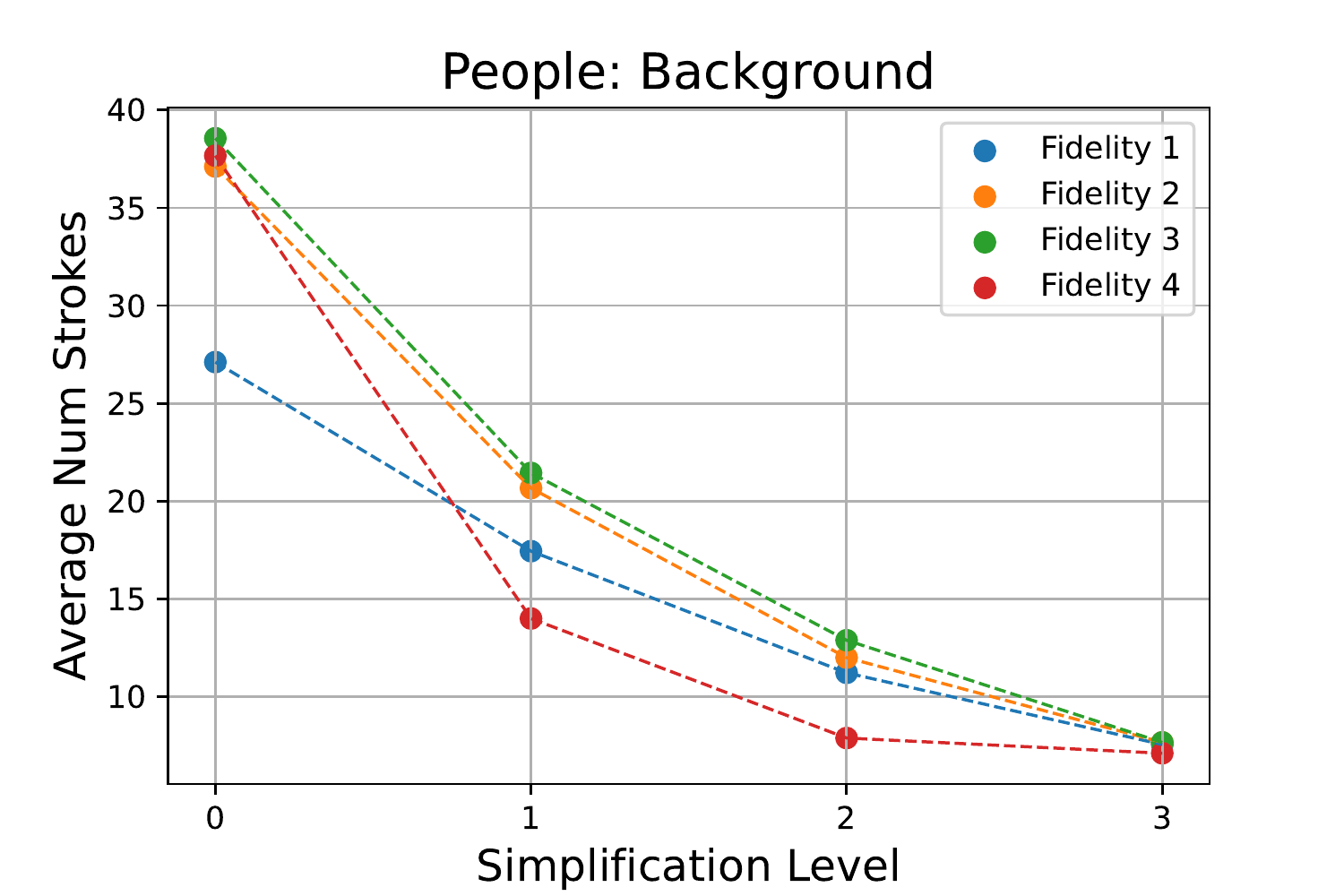} &
    \includegraphics[width=0.3\linewidth]{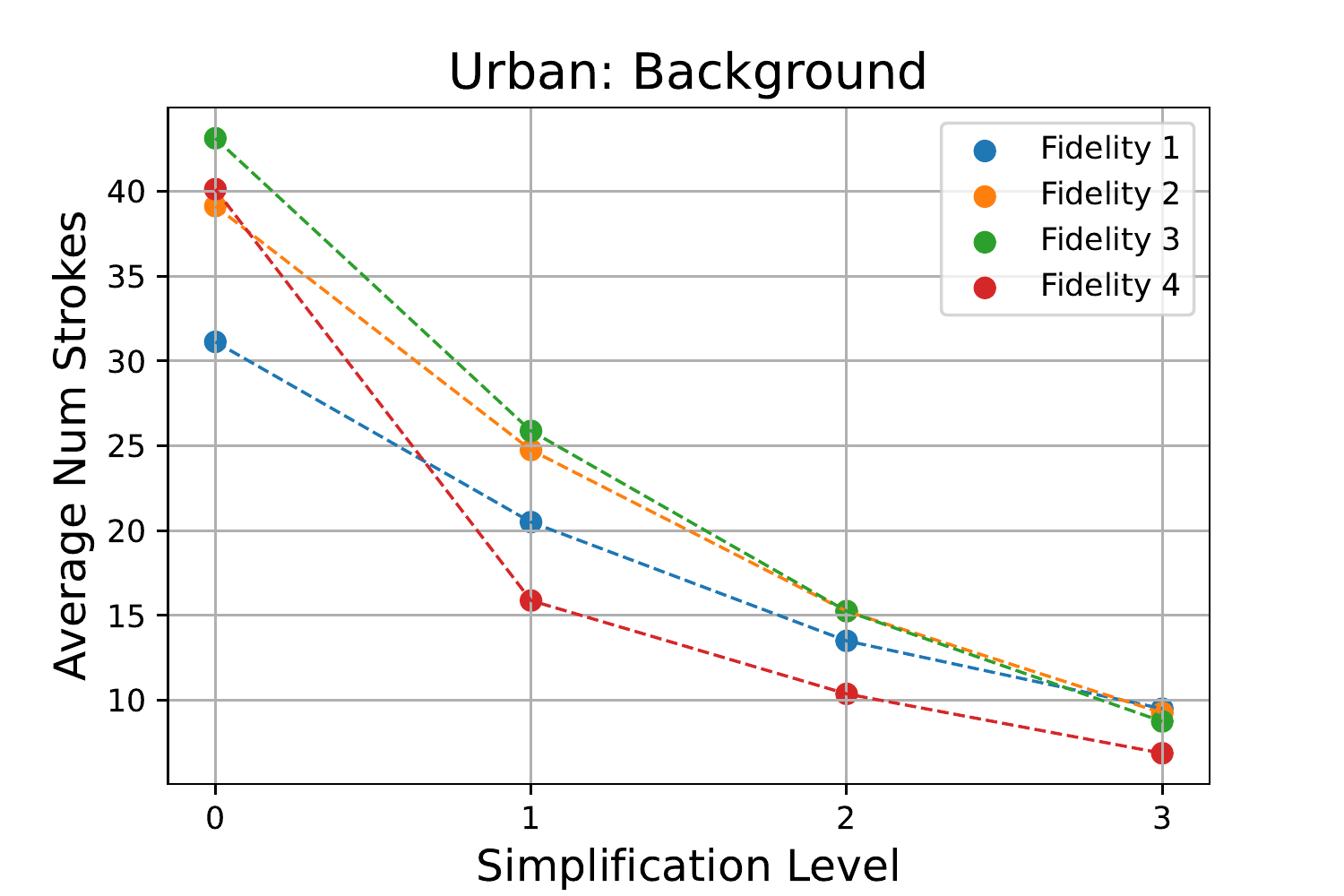} \\

    \\
    \end{tabular}

    \begin{tabular}{c c c}

    \textbf{Nature} &
    \textbf{Indoor} &
    \textbf{Animals} \\

    \includegraphics[width=0.3\linewidth]{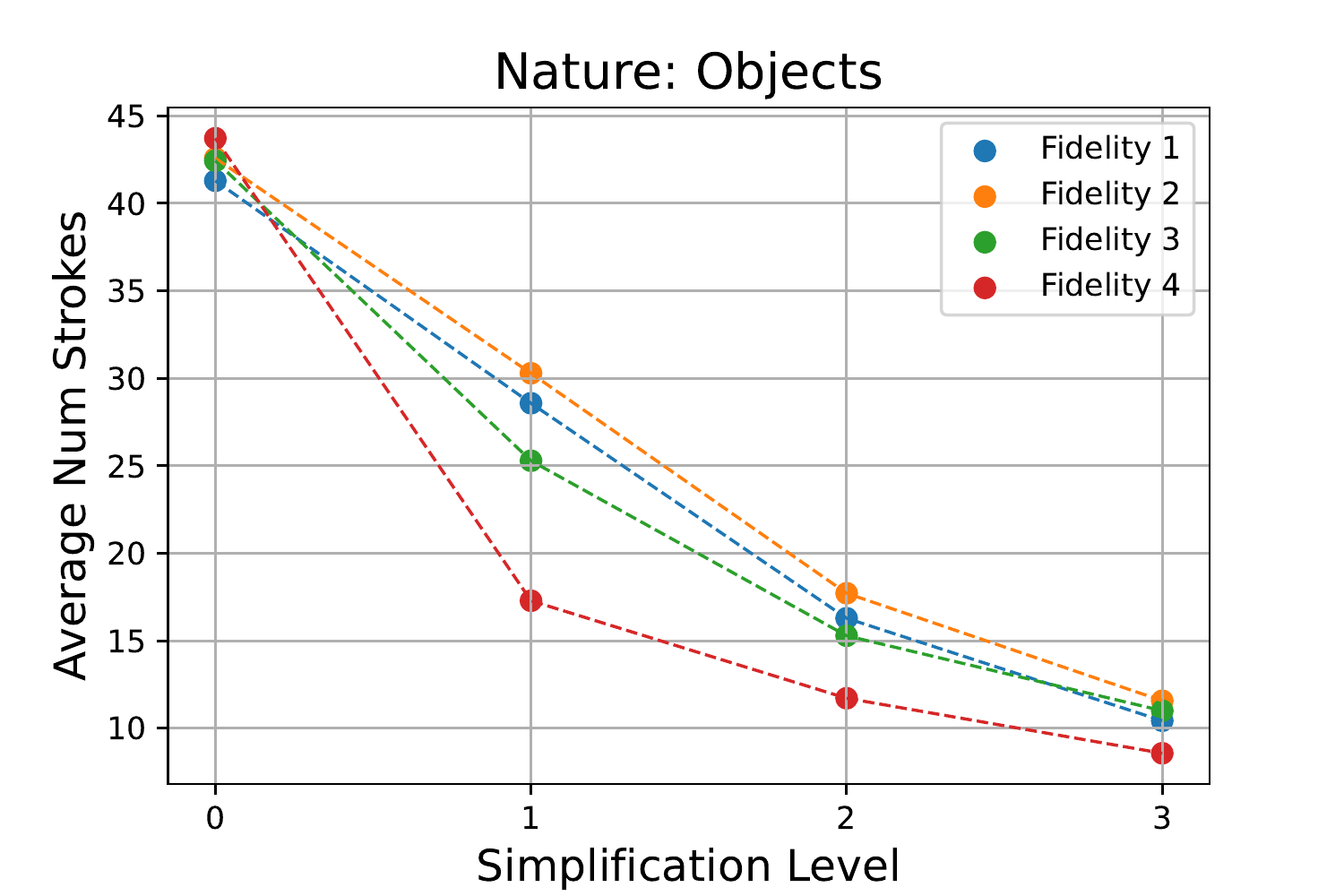} &
    \includegraphics[width=0.3\linewidth]{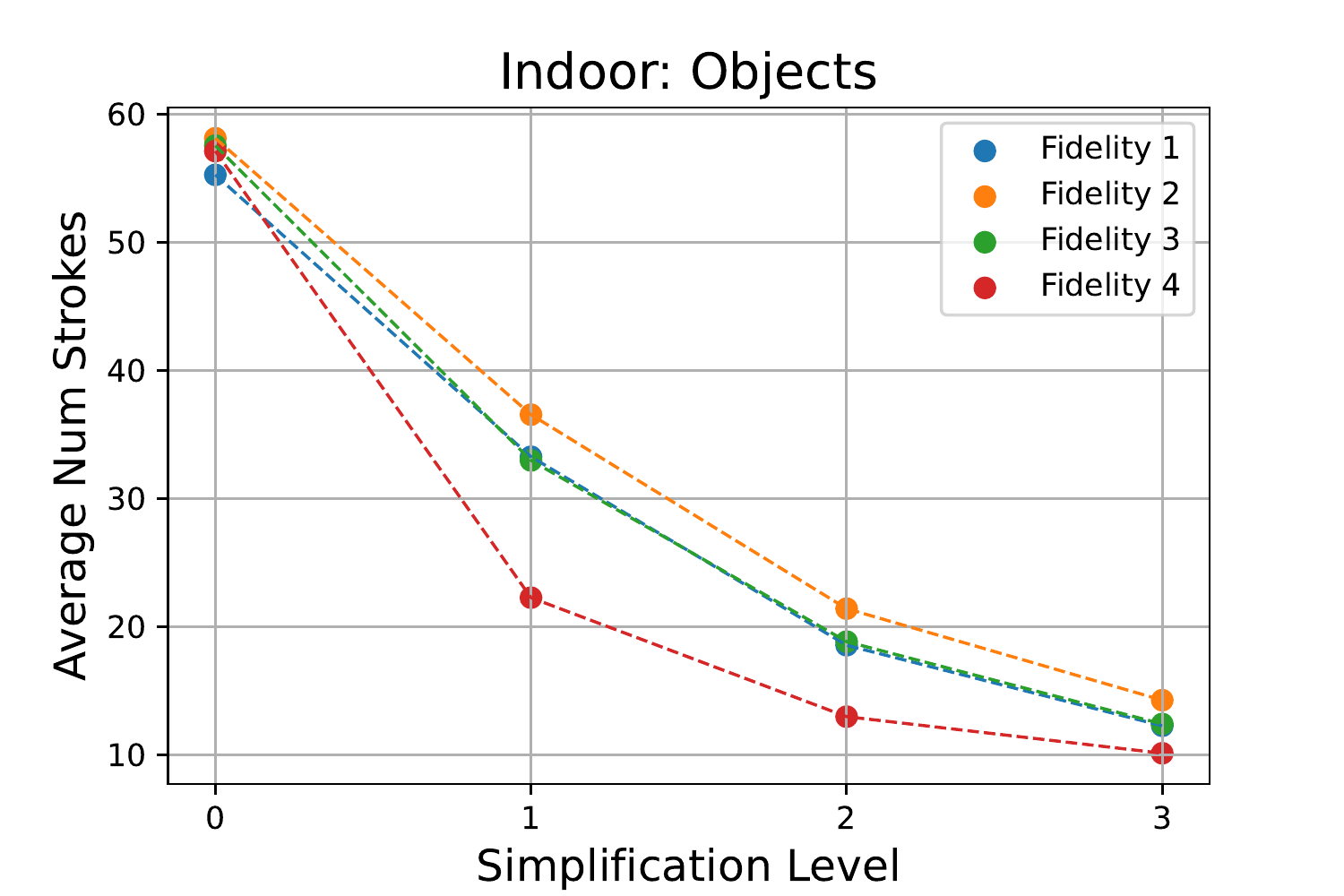} &
    \includegraphics[width=0.3\linewidth]{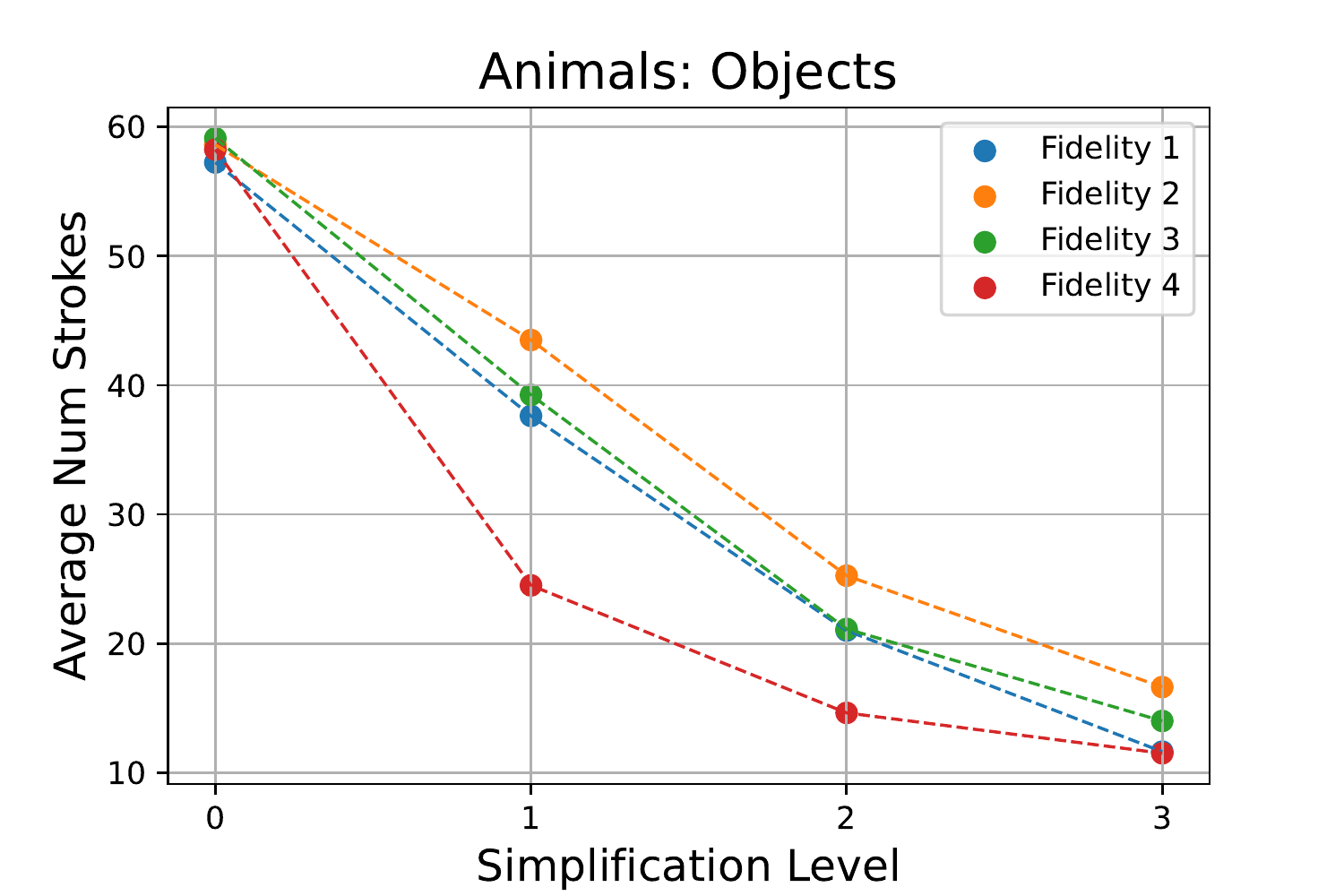} \\

    \includegraphics[width=0.3\linewidth]{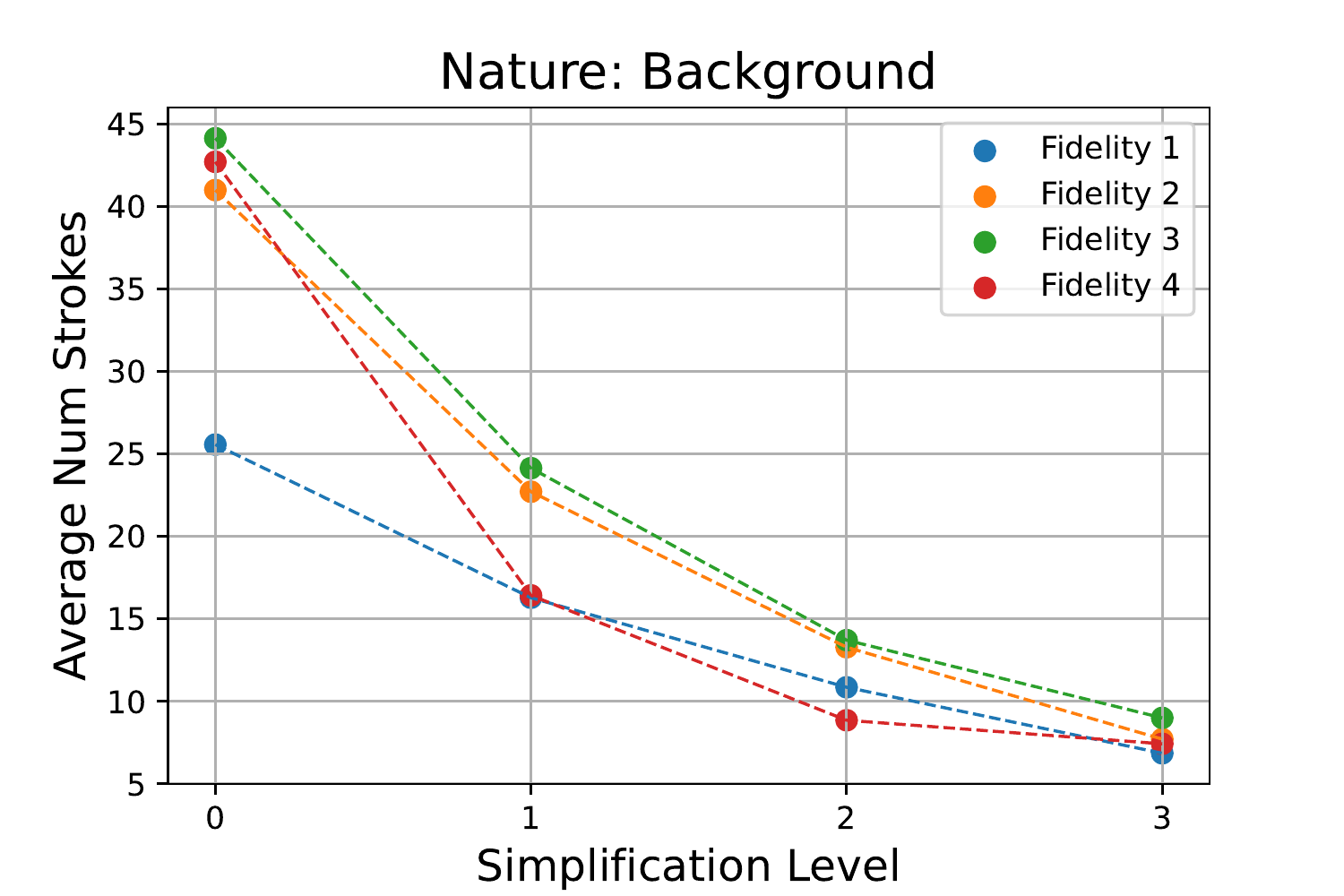} &
    \includegraphics[width=0.3\linewidth]{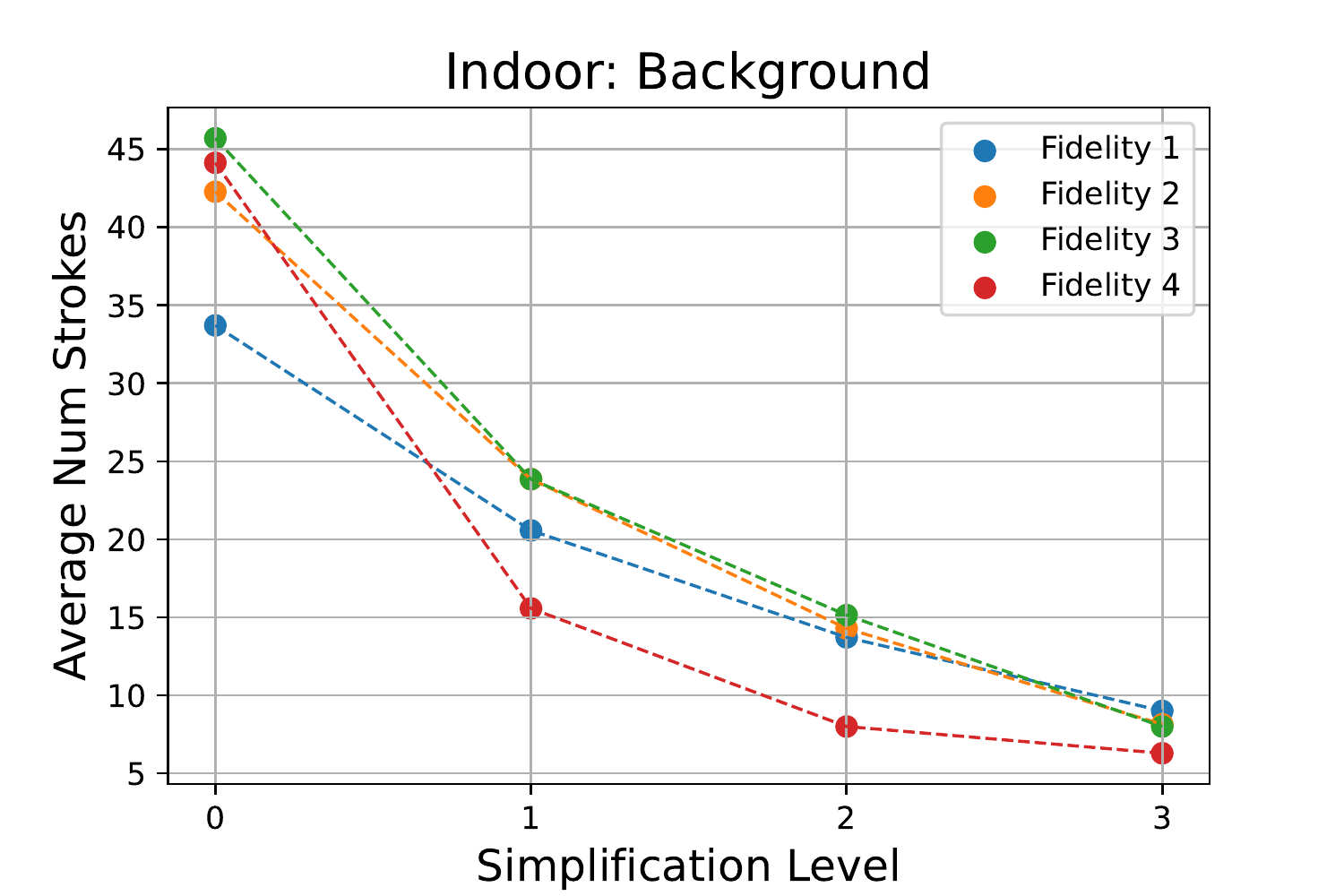} &
    \includegraphics[width=0.3\linewidth]{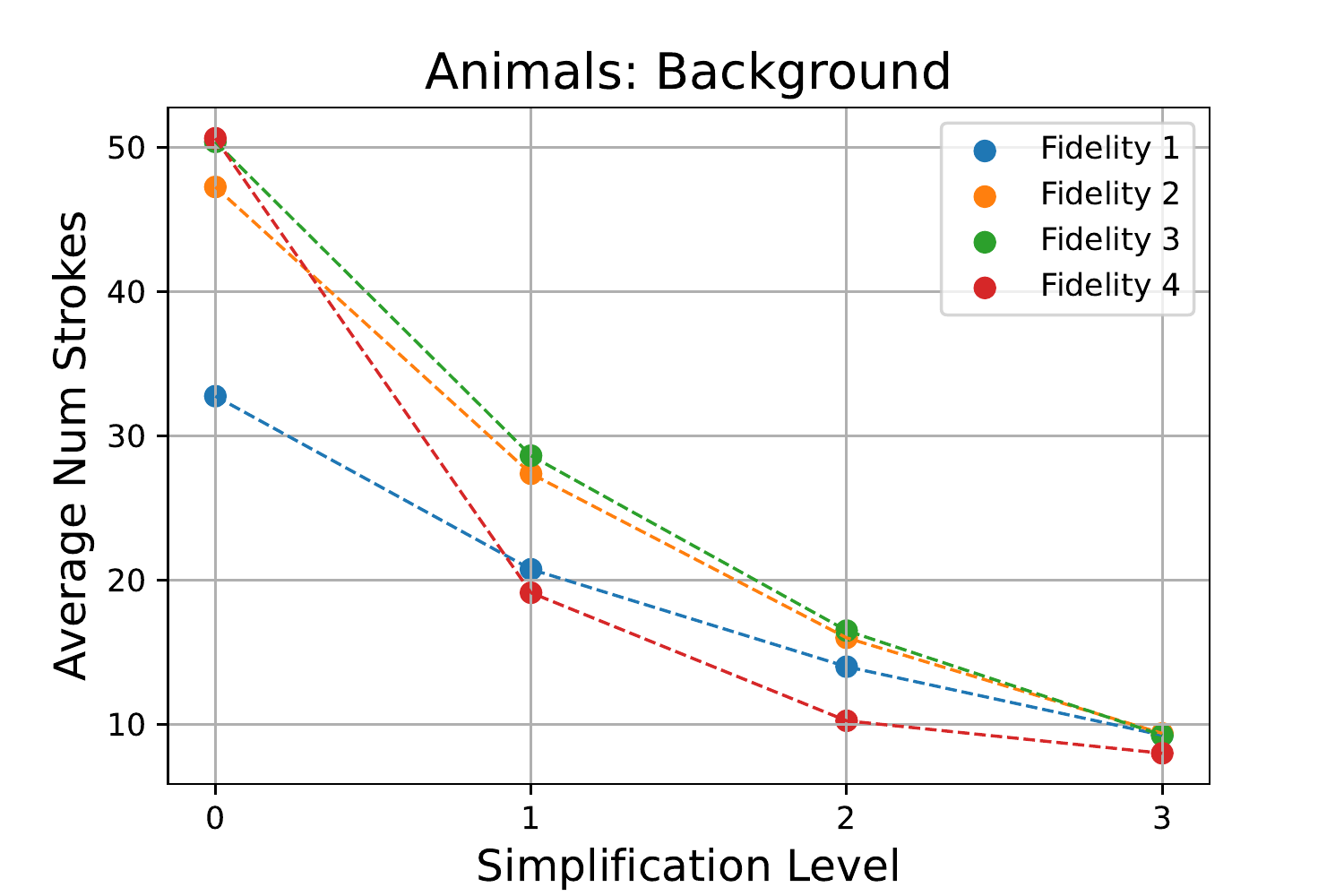} \\

    \end{tabular}
    
    \caption{
    Examining the number of strokes used to compose the sketch across each fidelity and simplification level, split between the different scene categories.
    }
    \label{fig:quantitative_num_strokes_by_class}
    
\end{figure*}

%% file: files/figures/supplementary/ablation_resnet.tex
\begin{figure*}[t!]

    \centering
    \setlength{\tabcolsep}{1.5pt}
    {\small
    \begin{tabular}{c c c c c@{\hspace{0.3cm}} | c c c}

        \includegraphics[width=0.1105\textwidth]{figs/inputs/woman_city.jpg} &
        \hspace{0.1cm}
        \includegraphics[width=0.1105\textwidth]{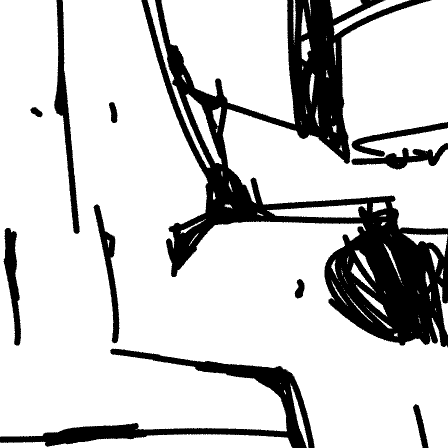} &
        \includegraphics[width=0.1105\textwidth]{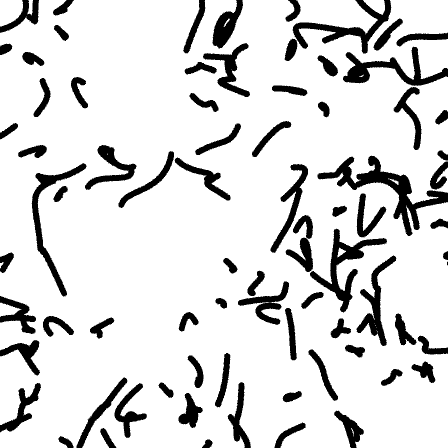} &
        \includegraphics[width=0.1105\textwidth]{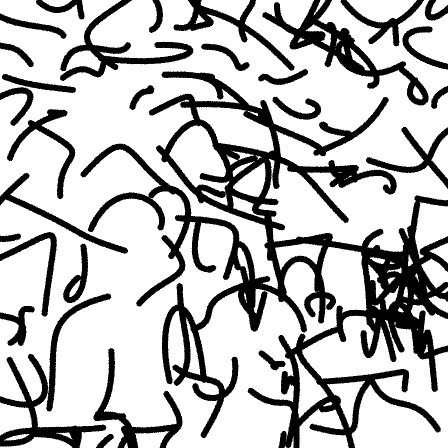} &
        \includegraphics[width=0.1105\textwidth]{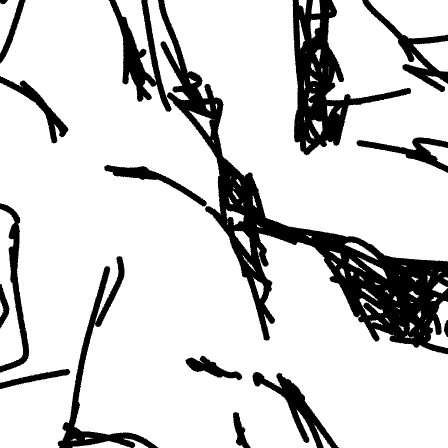} &    
        \hspace{0.1cm}
        \includegraphics[width=0.1105\textwidth]{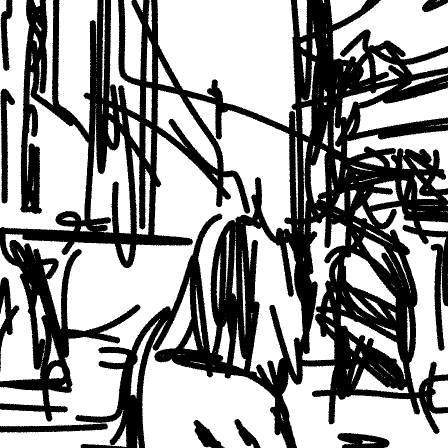} &
        \includegraphics[width=0.1105\textwidth]{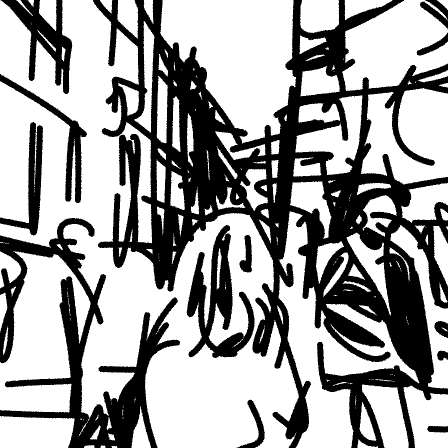} &
        \includegraphics[width=0.1105\textwidth]{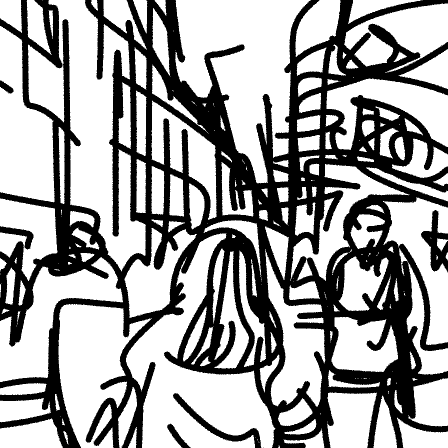} \\

        \includegraphics[width=0.1105\textwidth]{figs/inputs/man_camera.jpg} &
        \hspace{0.1cm}
        \includegraphics[width=0.1105\textwidth]{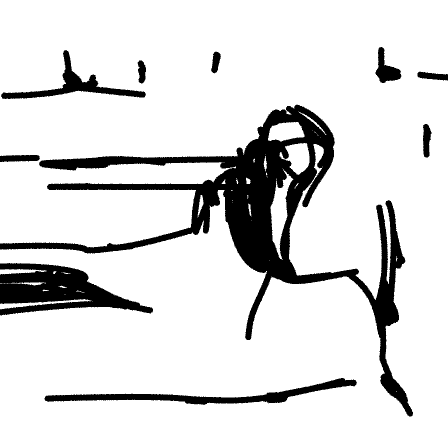} &
        \includegraphics[width=0.1105\textwidth]{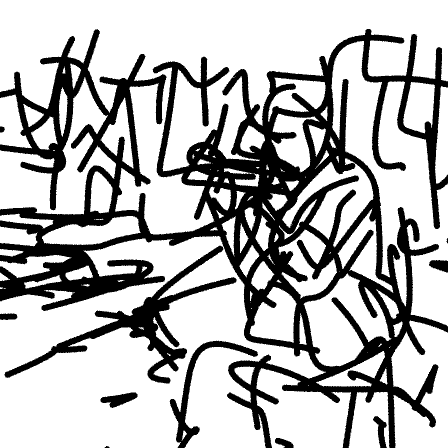} &
        \includegraphics[width=0.1105\textwidth]{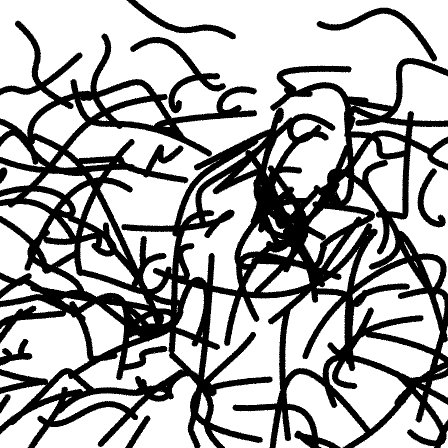} &
        \includegraphics[width=0.1105\textwidth]{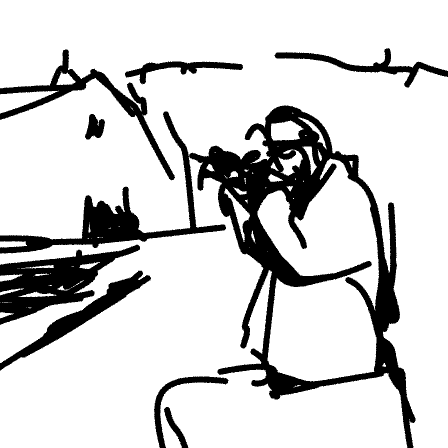} &
        \hspace{0.1cm}
        \includegraphics[width=0.1105\textwidth]{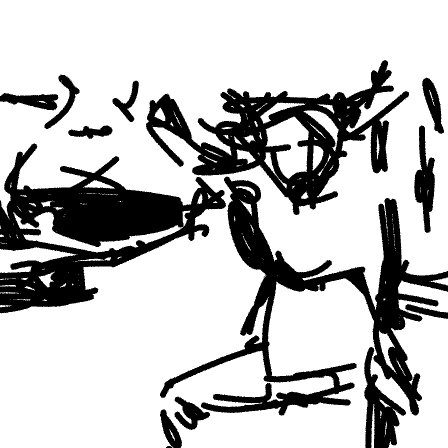} &
        \includegraphics[width=0.1105\textwidth]{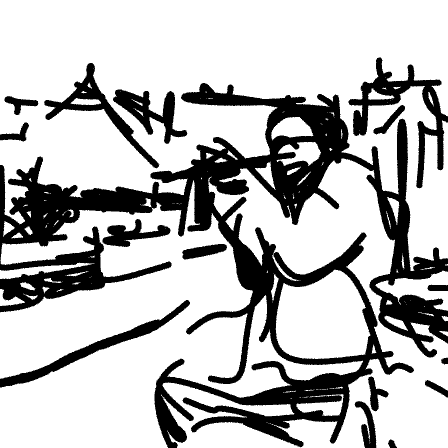} &
        \includegraphics[width=0.1105\textwidth]{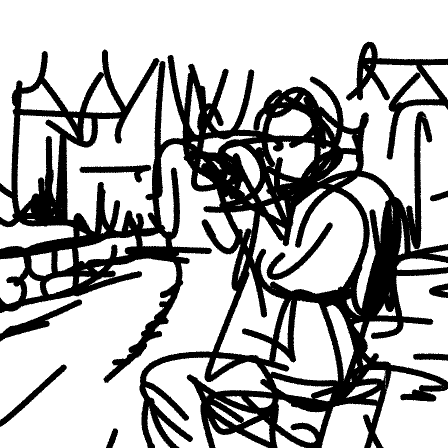} \\

        \includegraphics[width=0.1105\textwidth]{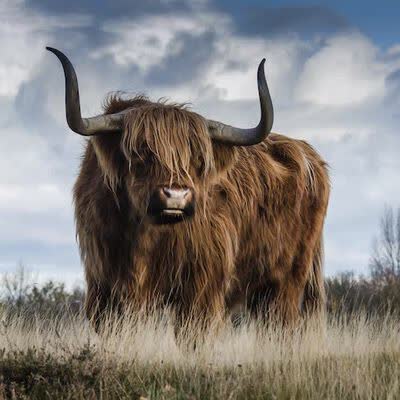} &
        \hspace{0.1cm}
        \includegraphics[width=0.1105\textwidth]{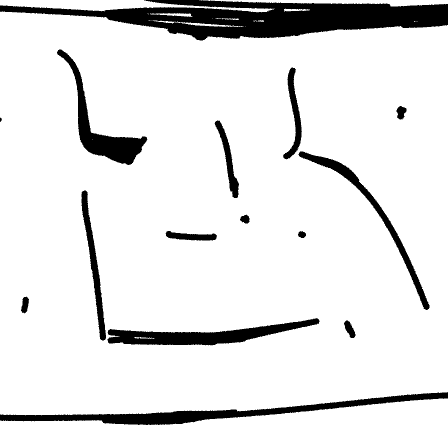} &
        \includegraphics[width=0.1105\textwidth]{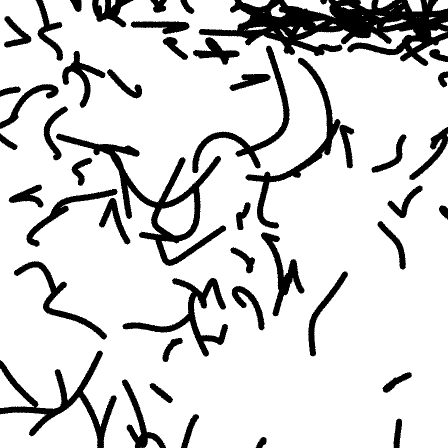} &
        \includegraphics[width=0.1105\textwidth]{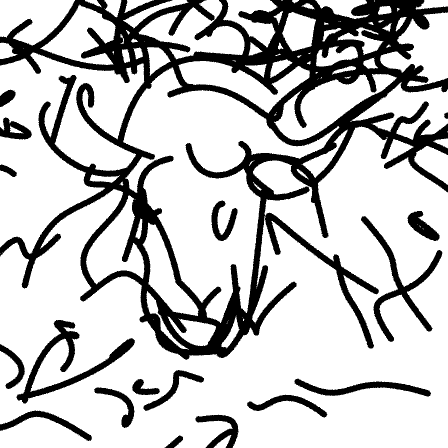} &
        \includegraphics[width=0.1105\textwidth]{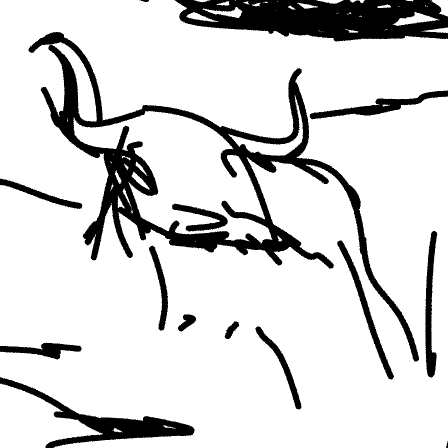} &
        \hspace{0.1cm}
        \includegraphics[width=0.1105\textwidth]{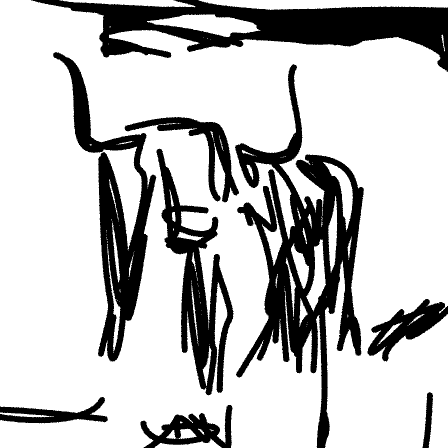} &
        \includegraphics[width=0.1105\textwidth]{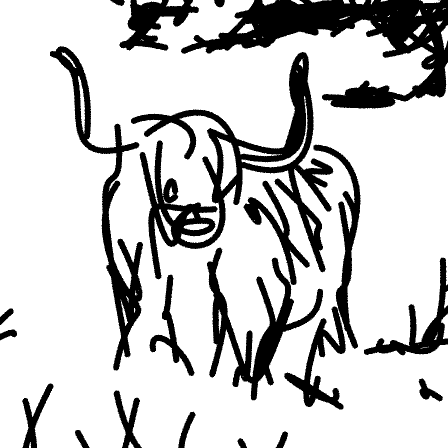} &
        \includegraphics[width=0.1105\textwidth]{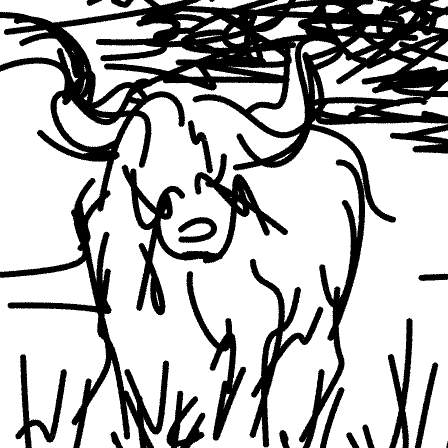} \\

        \includegraphics[width=0.1105\textwidth]{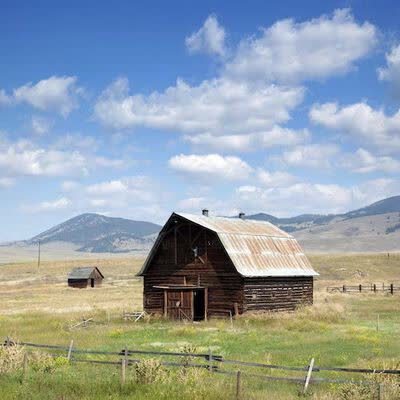} &
        \hspace{0.1cm}
        \includegraphics[width=0.1105\textwidth]{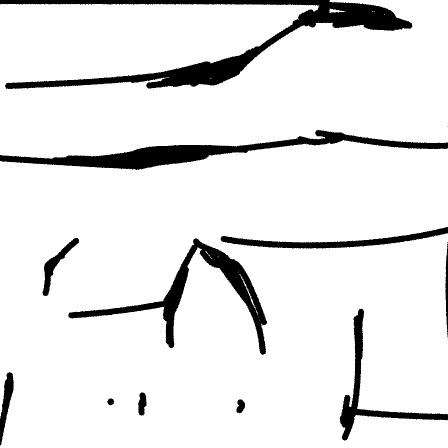} &
        \includegraphics[width=0.1105\textwidth]{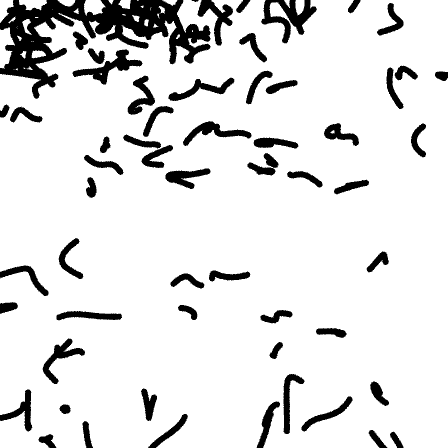} &
        \includegraphics[width=0.1105\textwidth]{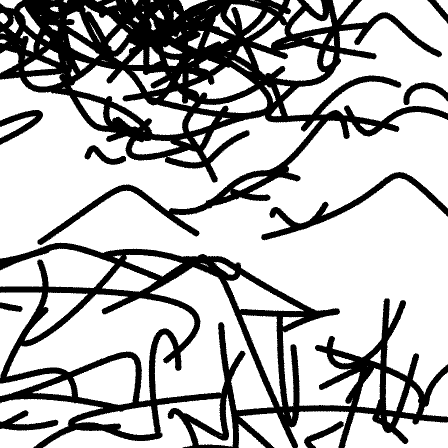} &
        \includegraphics[width=0.1105\textwidth]{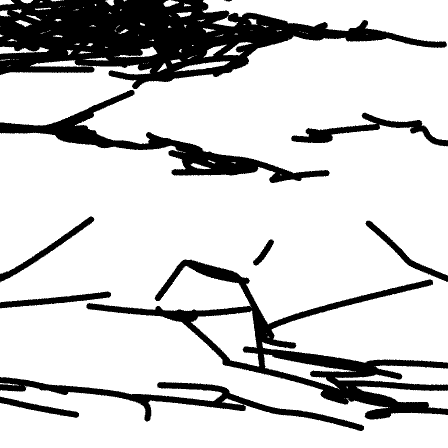} &
        \hspace{0.1cm}
        \includegraphics[width=0.1105\textwidth]{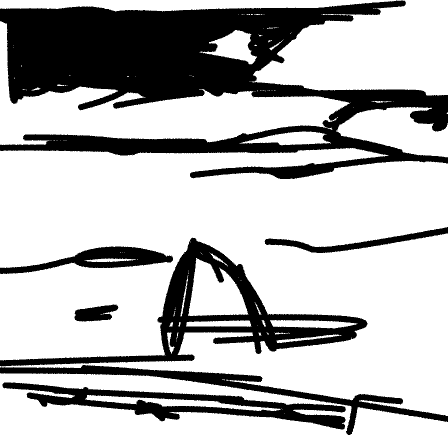} &
        \includegraphics[width=0.1105\textwidth]{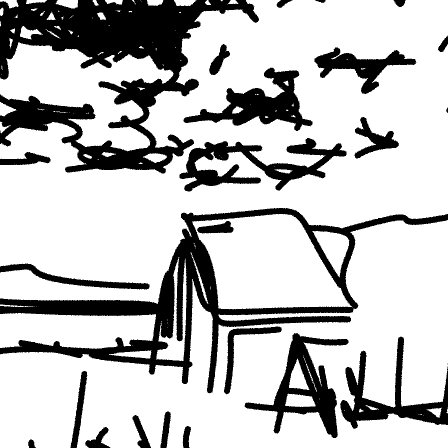} &
        \includegraphics[width=0.1105\textwidth]{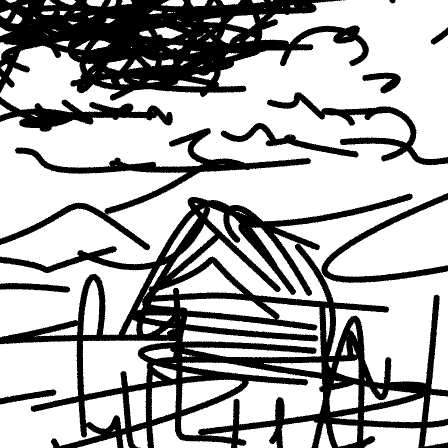} \\
        
        Input & 
        \begin{tabular}{c} ResNet \\ (Layer 2) \end{tabular} &
        \begin{tabular}{c} ResNet \\ (Layer 3) \end{tabular} &
        \begin{tabular}{c} ResNet \\ (Layer 4) \end{tabular} &
        \begin{tabular}{c} ResNet \\ Multi-Layer \end{tabular} &
        \begin{tabular}{c} ViT \\ (Layer 2) \end{tabular} &
        \begin{tabular}{c} ViT \\ (Layer 7) \end{tabular} &
        \begin{tabular}{c} ViT \\ (Layer 11) \end{tabular}
                
    \end{tabular}
    
    }
    \vspace{0.15cm}
    \caption{Ablation study on using ResNet-CLIP and ViT-CLIP for guiding the training process. 
    For both variants, we generate the sketches for the entire scene together (\ie no scene decomposition is performed) and set the number of strokes to $128$.
    In addition to evaluating ResNet using a single layer for computing $\mathcal{L}_{CLIP}$, we additionally show results obtained when multiple ResNet layers are used to compute $\mathcal{L}_{CLIP}$ (marked as ``ResNet Multi-Layer''). For this variant, we follow CLIPasso~\cite{vinker2022clipasso} as set the layer weights to $0,0,1,1,0$ for layers $\ell_1$ to $\ell_5$, respectively. In addition, we use the output of the fully connected layer and set its weight to $0.1$ in the loss computation.}
    \vspace{-0.2cm}
    \label{fig:ablation_full_scene_resnet}
\end{figure*}

%% file: files/figures/supplementary/ablation_foreground_background.tex
\begin{figure*}[ht]
    \centering
    \setlength{\belowcaptionskip}{-6pt}
    \setlength{\tabcolsep}{1.5pt}
    {\small
    
    \begin{tabular}{c c c c c c}

        \\ \\ \\ \\

        \includegraphics[width=0.1\linewidth]{figs/inputs/house3.jpg} &
        \raisebox{0.05in}{\rotatebox{90}{\textcolor{cyan}{w/o Split}}} &
        \includegraphics[width=0.1\linewidth]{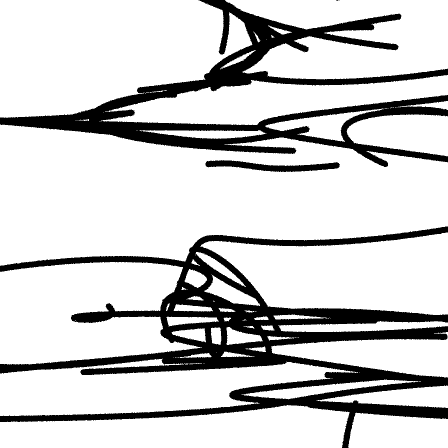} &
        \includegraphics[width=0.1\linewidth]{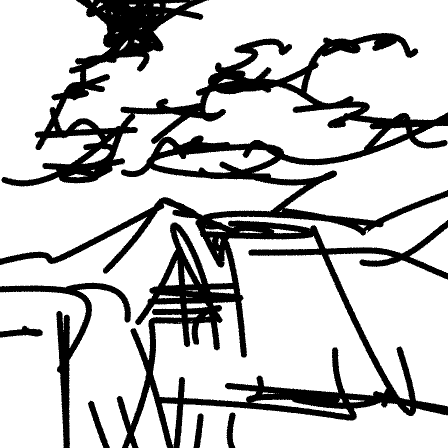} &
        \includegraphics[width=0.1\linewidth]{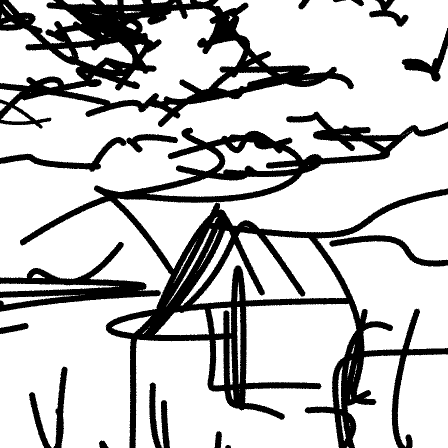} &
        \includegraphics[width=0.1\linewidth]{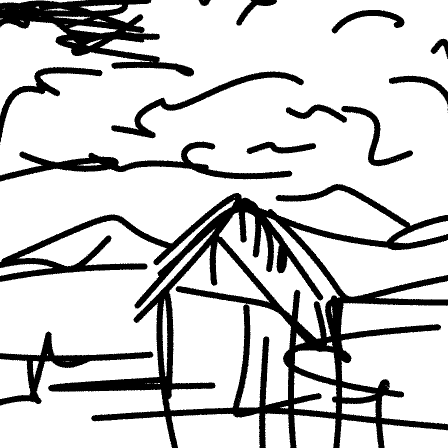} \\
        
        &
        \raisebox{0.05in}{\rotatebox{90}{\textcolor{orange}{w/ Split}}} &
        \includegraphics[width=0.1\linewidth]{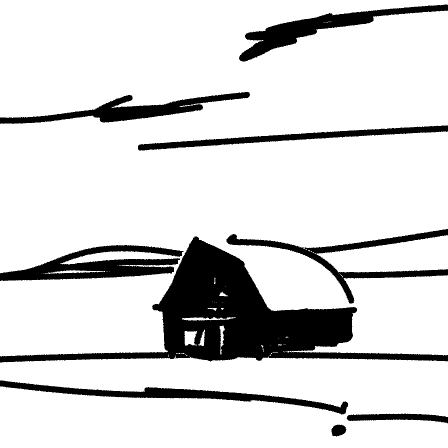} & 
        \includegraphics[width=0.1\linewidth]{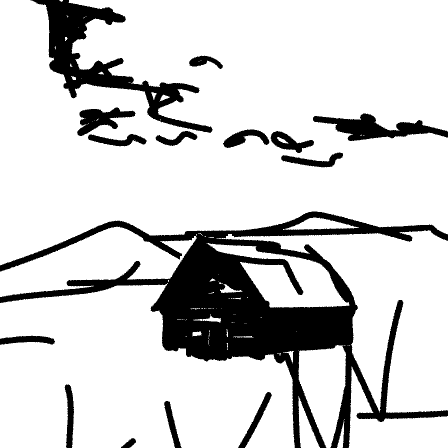} &
        \includegraphics[width=0.1\linewidth]{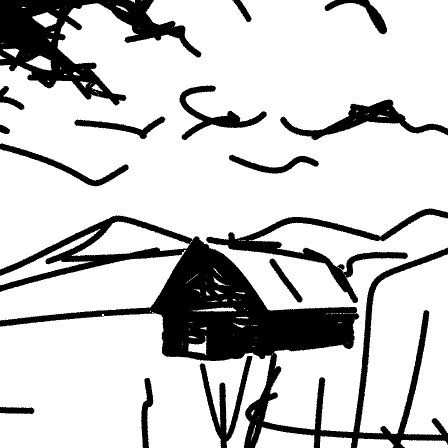} &
        \includegraphics[width=0.1\linewidth]{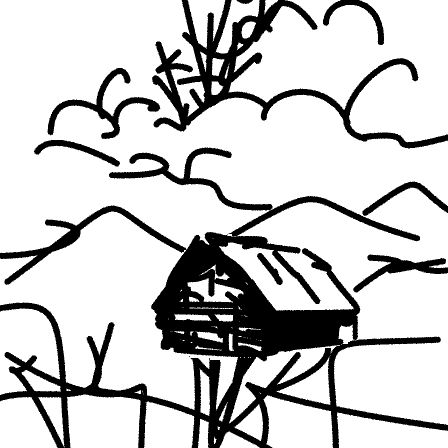} \\

        \includegraphics[width=0.1\linewidth]{figs/inputs/dog.jpg} &
        \raisebox{0.05in}{\rotatebox{90}{\textcolor{cyan}{w/o Split}}} &
        \includegraphics[width=0.1\linewidth]{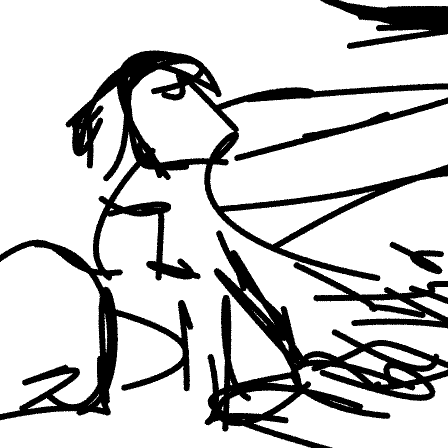} &
        \includegraphics[width=0.1\linewidth]{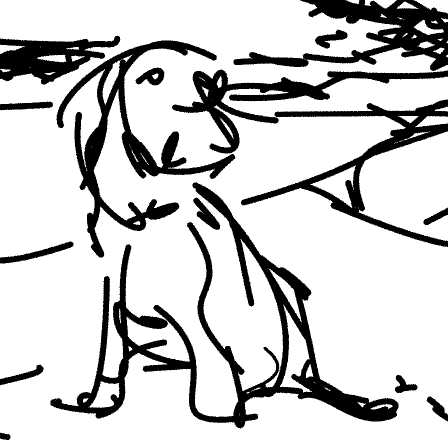} &
        \includegraphics[width=0.1\linewidth]{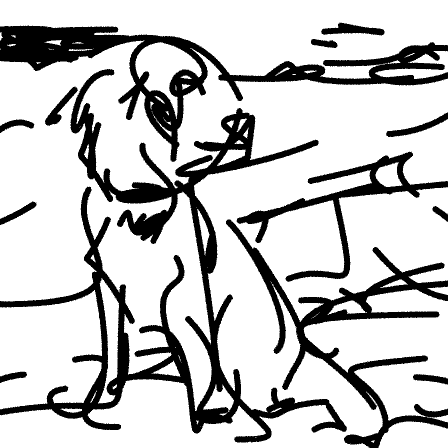} &
        \includegraphics[width=0.1\linewidth]{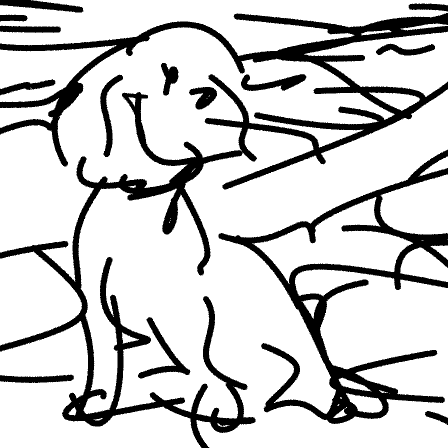} \\
        
        &
        \raisebox{0.05in}{\rotatebox{90}{\textcolor{orange}{w/ Split}}} &
        \includegraphics[width=0.1\linewidth]{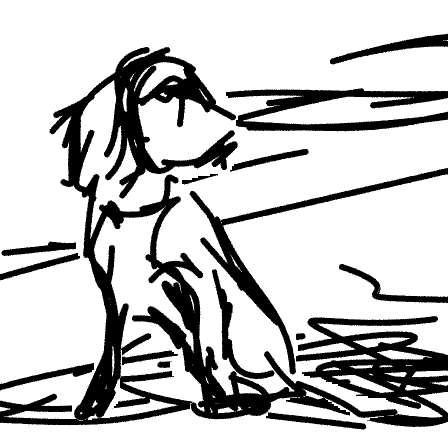} & 
        \includegraphics[width=0.1\linewidth]{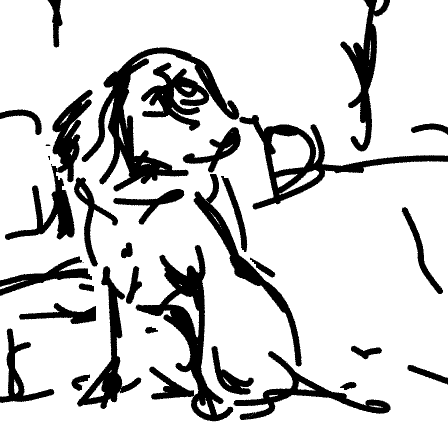} &
        \includegraphics[width=0.1\linewidth]{figs/matrices_black/dog_row0col2_black.png} &
        \includegraphics[width=0.1\linewidth]{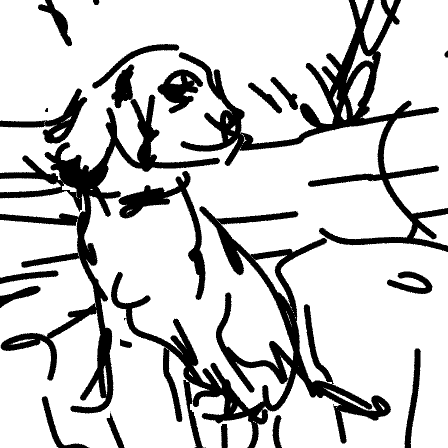} \\

        & \multicolumn{4}{l}{\hspace{0.35cm} Fidelity Axis $\longrightarrow$} \\

        \\

        \includegraphics[width=0.1\linewidth]{figs/inputs/house3.jpg} &
        \raisebox{0.05in}{\rotatebox{90}{\textcolor{cyan}{w/o Split}}} &
        \includegraphics[width=0.1\linewidth]{figs/ablations/full_scene/house3_row0col2_full.png} &
        \includegraphics[width=0.1\linewidth]{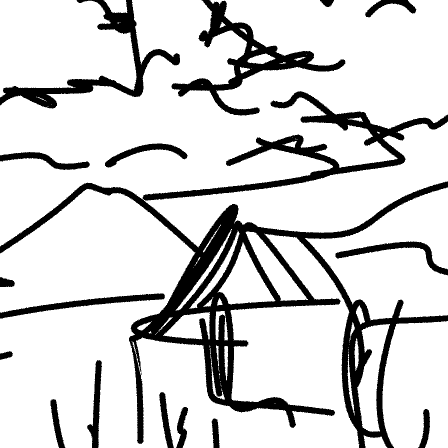} &
        \includegraphics[width=0.1\linewidth]{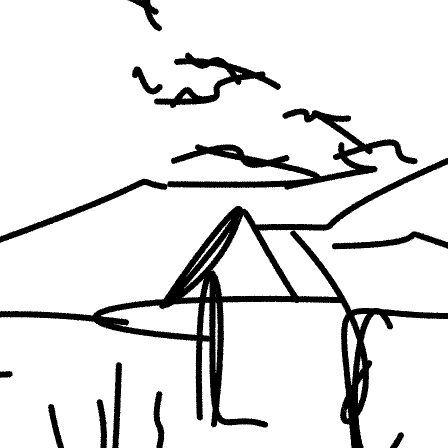} &
        \includegraphics[width=0.1\linewidth]{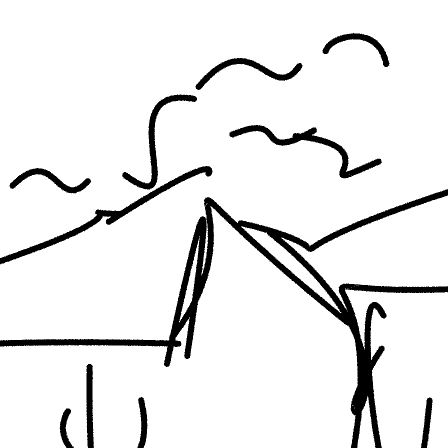} \\

        &
        \raisebox{0.05in}{\rotatebox{90}{\textcolor{orange}{w/ Split}}} &
        \includegraphics[width=0.1\linewidth]{figs/matrices_black/house3_row0col2_black.png} &
        \includegraphics[width=0.1\linewidth]{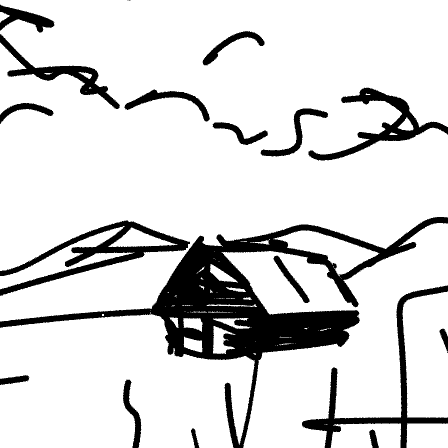} &
        \includegraphics[width=0.1\linewidth]{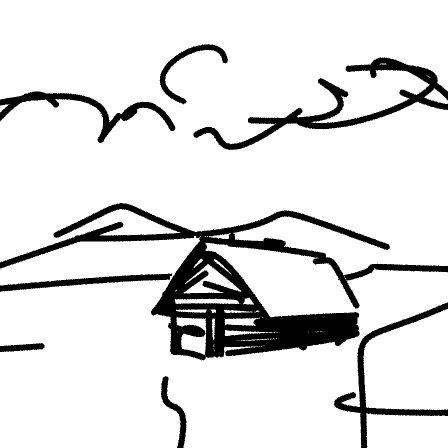} &
        \includegraphics[width=0.1\linewidth]{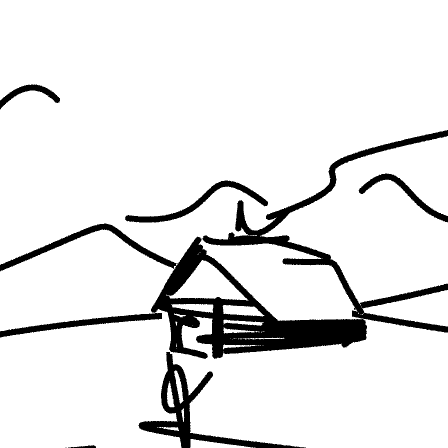} \\

        \includegraphics[width=0.1\linewidth]{figs/inputs/dog.jpg} &
        \raisebox{0.05in}{\rotatebox{90}{\textcolor{cyan}{w/o Split}}} &
        \includegraphics[width=0.1\linewidth]{figs/ablations/full_scene/dog_row0col2_full.png} &
        \includegraphics[width=0.1\linewidth]{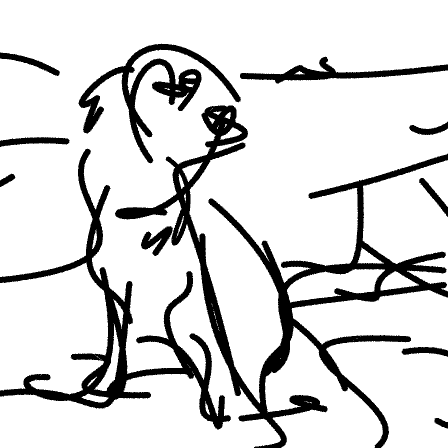} &
        \includegraphics[width=0.1\linewidth]{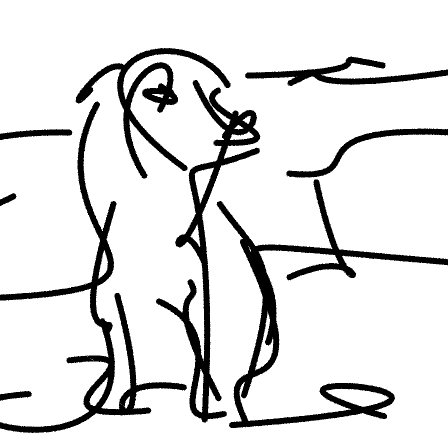} &
        \includegraphics[width=0.1\linewidth]{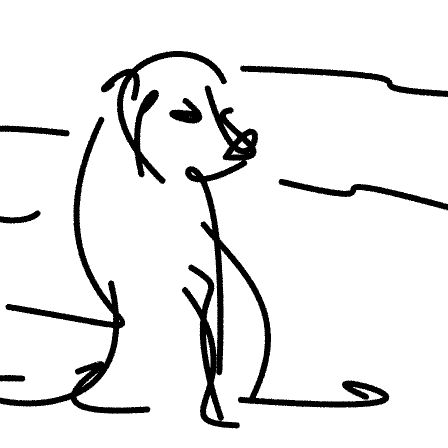} \\

        &
        \raisebox{0.05in}{\rotatebox{90}{\textcolor{orange}{w/ Split}}} &
        \includegraphics[width=0.1\linewidth]{figs/matrices_black/dog_row0col2_black.png} &
        \includegraphics[width=0.1\linewidth]{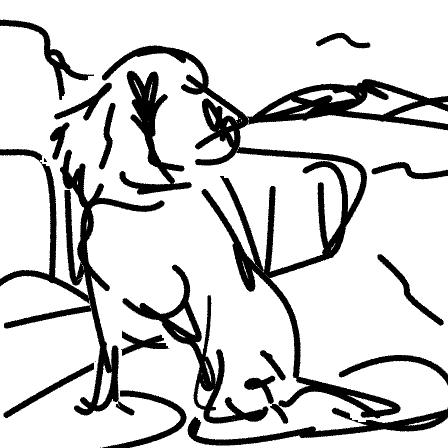} &
        \includegraphics[width=0.1\linewidth]{figs/matrices_black/dog_row2col2_black.png} &
        \includegraphics[width=0.1\linewidth]{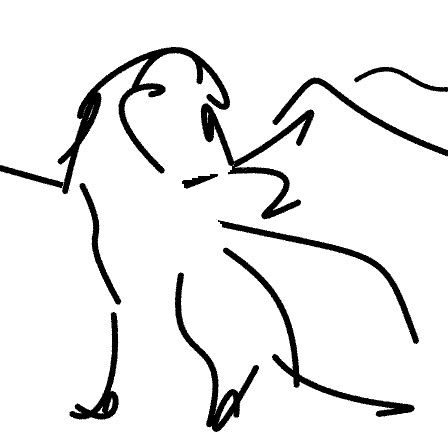} \\

        & & \multicolumn{4}{l}{\hspace{0.05cm} Simplicity Axis $\longrightarrow$} \\

    \end{tabular}
    
    }
    \vspace{0.2cm}
    \caption{Scene sketching results obtained with and without decomposing the scene. We show sketches generated across both the fidelity axis (first four rows) and the simplicity axis (bottom four rows).}
    \label{fig:ablation_foreground_background_supp}
\end{figure*}

%% file: files/figures/supplementary/ablation_explicit_define_num_strokes.tex
\begin{figure*}
    \centering
    \setlength{\tabcolsep}{1.5pt}
    
    \begin{tabular}{c c c c c@{\hspace{0.2cm}} | c  c c c c}

    \includegraphics[width=0.0875\linewidth]{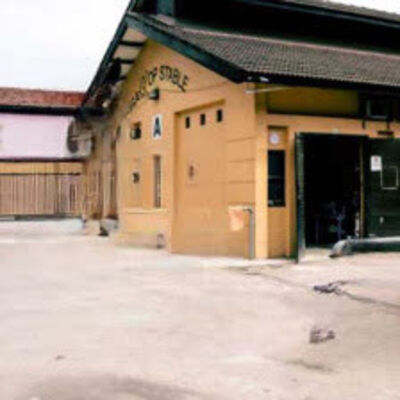} &
    \hspace{0.1cm}
    \includegraphics[width=0.0875\linewidth]{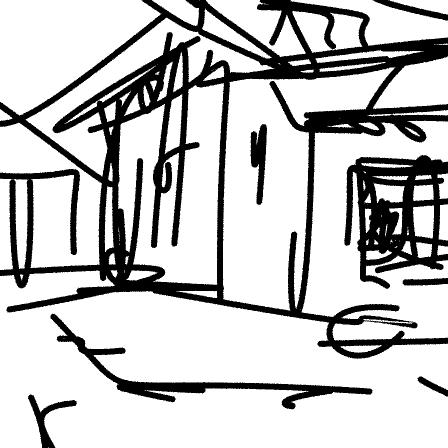} &
    \includegraphics[width=0.0875\linewidth]{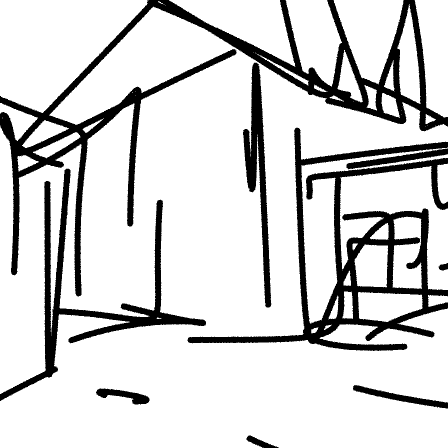} &
    \includegraphics[width=0.0875\linewidth]{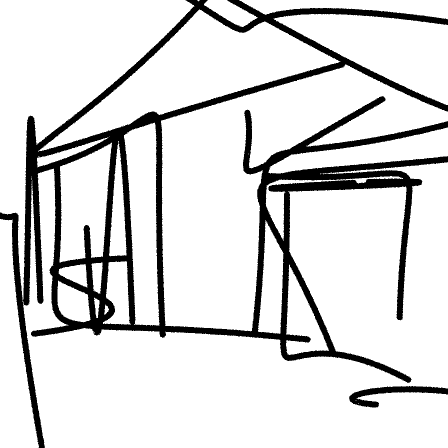} &
    \includegraphics[width=0.0875\linewidth]{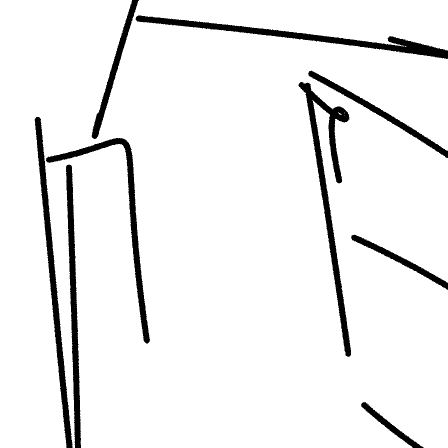} &  
    \hspace{0.1cm}
    \includegraphics[width=0.0875\linewidth]{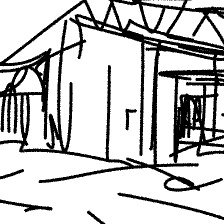} &
    \includegraphics[width=0.0875\linewidth]{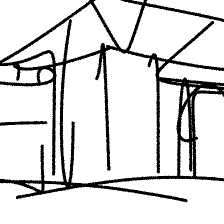} &
    \includegraphics[width=0.0875\linewidth]{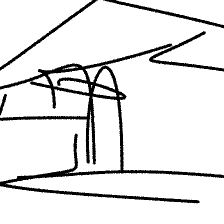} &
    \includegraphics[width=0.0875\linewidth]{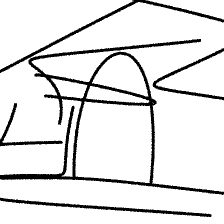} \\

    \includegraphics[width=0.0875\linewidth]{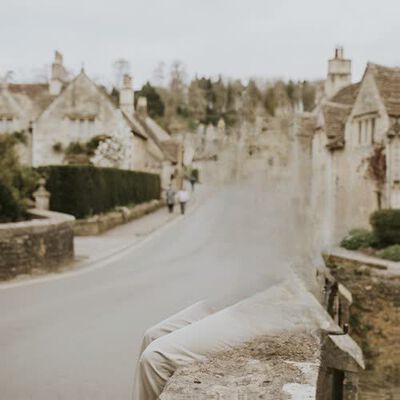} &
    \hspace{0.1cm}
    \includegraphics[width=0.0875\linewidth]{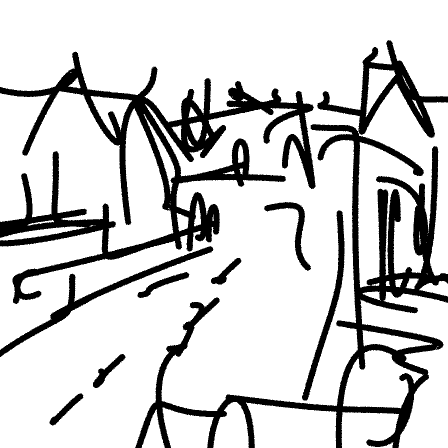} &
    \includegraphics[width=0.0875\linewidth]{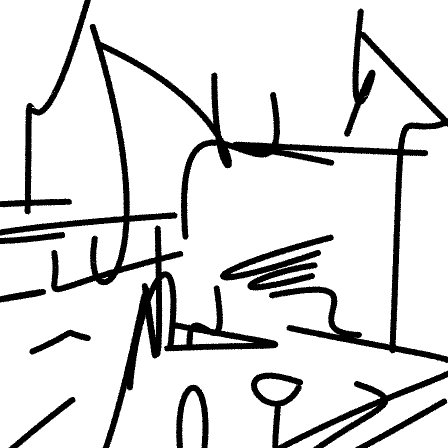} &
    \includegraphics[width=0.0875\linewidth]{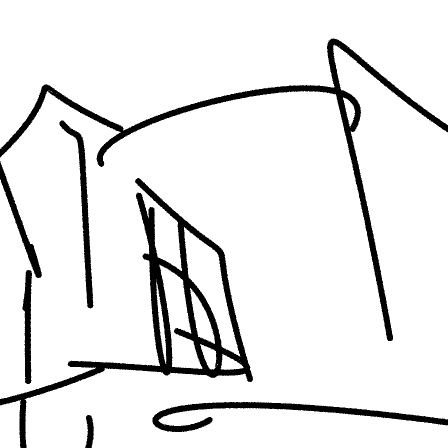} &
    \includegraphics[width=0.0875\linewidth]{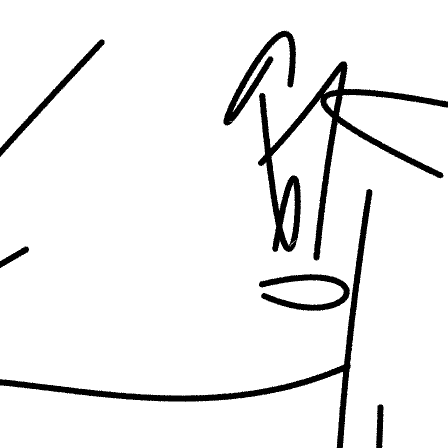} &
    \hspace{0.1cm}
    \includegraphics[width=0.0875\linewidth]{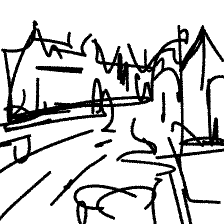} &
    \includegraphics[width=0.0875\linewidth]{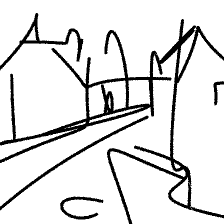} &
    \includegraphics[width=0.0875\linewidth]{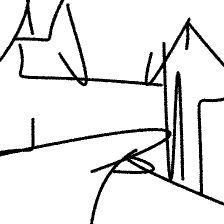} &
    \includegraphics[width=0.0875\linewidth]{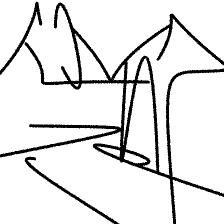} \\

    \includegraphics[width=0.0875\linewidth]{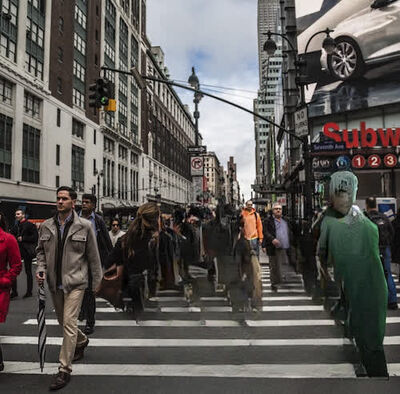} &
    \hspace{0.1cm}
    \includegraphics[width=0.0875\linewidth]{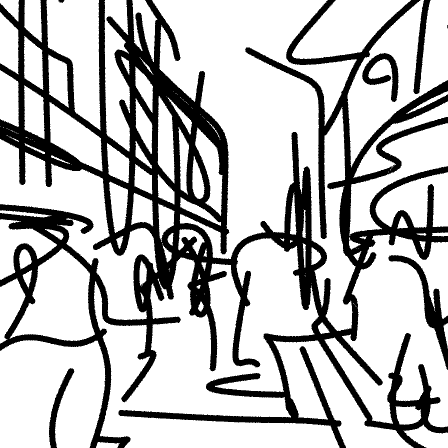} &
    \includegraphics[width=0.0875\linewidth]{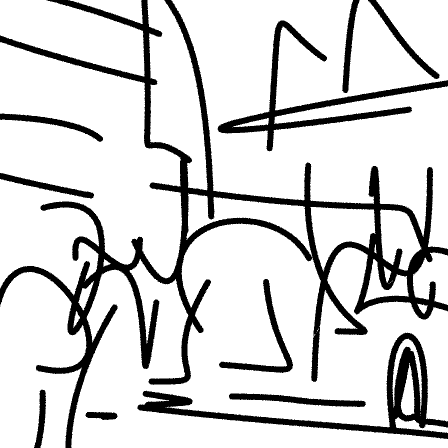} &
    \includegraphics[width=0.0875\linewidth]{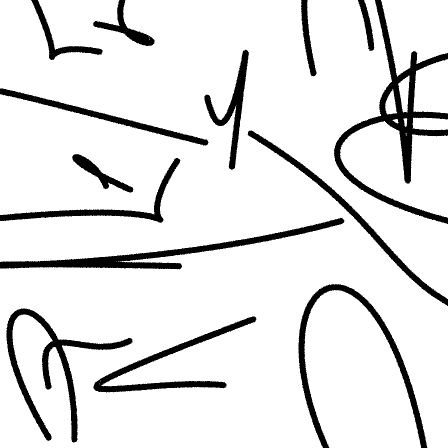} &
    \includegraphics[width=0.0875\linewidth]{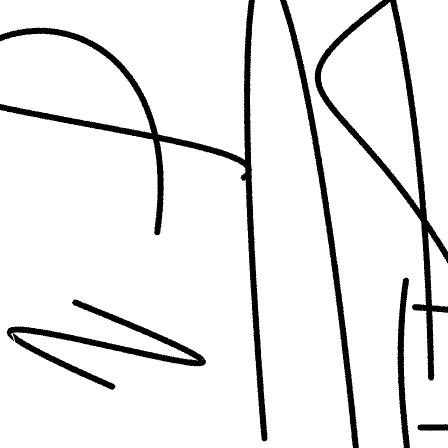} &
    \hspace{0.1cm}
    \includegraphics[width=0.0875\linewidth]{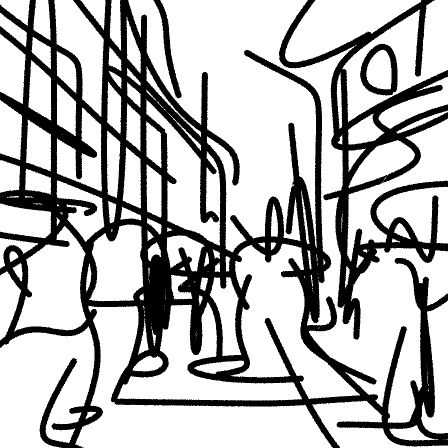} &
    \includegraphics[width=0.0875\linewidth]{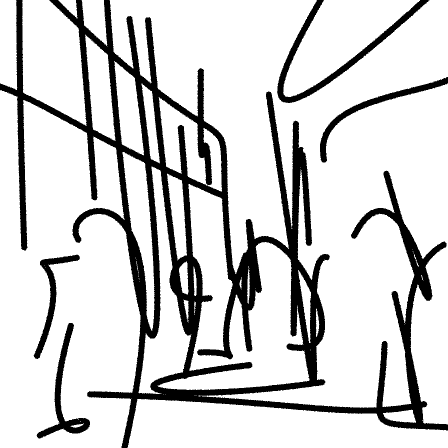} &
    \includegraphics[width=0.0875\linewidth]{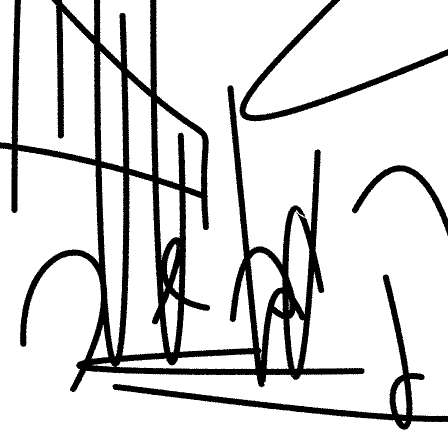} &
    \includegraphics[width=0.0875\linewidth]{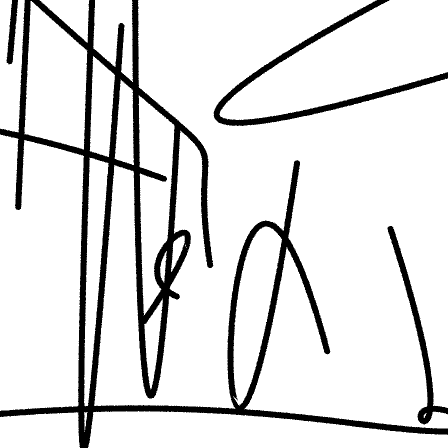} \\

    & 64 & 32 & 16 & 8 &  &  &  & \\ 
    
    Input &
    \multicolumn{4}{c}{\hspace{0.1cm} \xrfill[0.5ex]{0.5pt}\; Explicit Number of Strokes \; \xrfill[0.5ex]{0.5pt} \hspace{0.3cm}} &
    \multicolumn{4}{c}{\hspace{0.1cm} \xrfill[0.5ex]{0.5pt}\; Ours \; \xrfill[0.5ex]{0.5pt}}

    \end{tabular}

    \vspace{0.2cm}
    \caption{Ablation results of explicitly defining the number of strokes (left) compared to using our implicit simplification scheme (right). The presented sketches of all three input images are of the last fidelity level (obtained using layer $11$ of CLIP-ViT).}
    \label{fig:ablation_explicit_define_num_strokes}
    
\end{figure*}

%% file: files/figures/supplementary/ablation_direct_lratio.tex
\begin{figure*}[h]
    \centering
    \setlength{\tabcolsep}{1.5pt}
    
    \begin{tabular}{c}
    
    \begin{tabular}{c c c c c@{\hspace{0.2cm}} | c  c c c c}

    \includegraphics[width=0.0875\linewidth]{figs/inputs_mask/semi-complex_mask.jpg} &
    \hspace{0.1cm}
    \includegraphics[width=0.0875\linewidth]{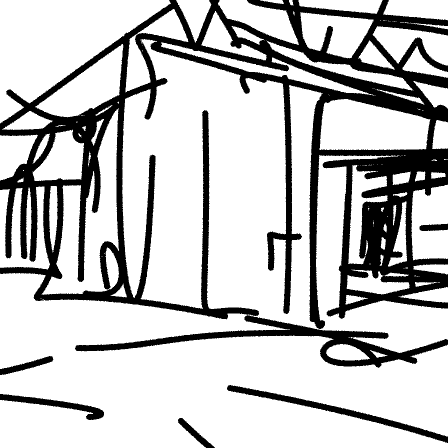} &
    \includegraphics[width=0.0875\linewidth]{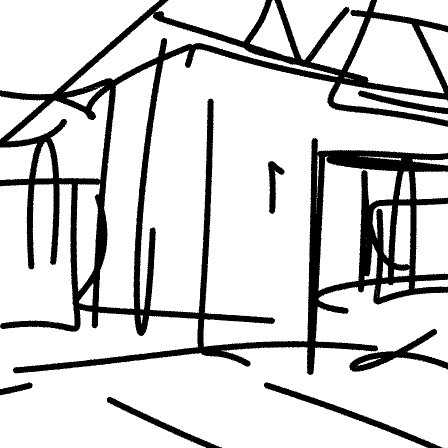} &
    \includegraphics[width=0.0875\linewidth]{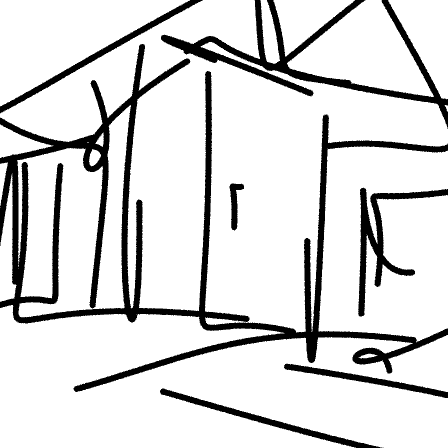} &
    \includegraphics[width=0.0875\linewidth]{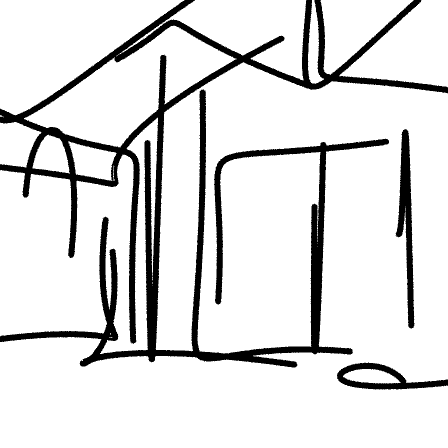} & 
    \hspace{0.1cm}
    \includegraphics[width=0.0875\linewidth]{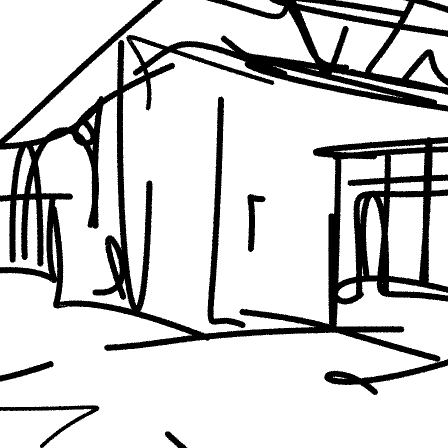} &
    \includegraphics[width=0.0875\linewidth]{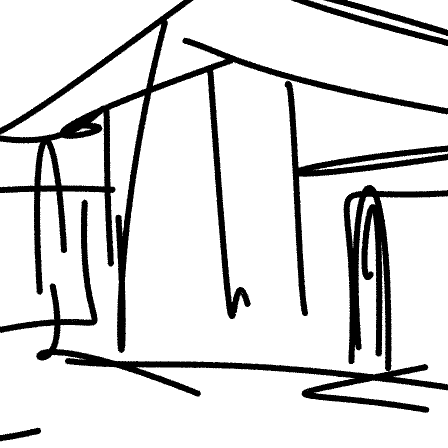} &
    \includegraphics[width=0.0875\linewidth]{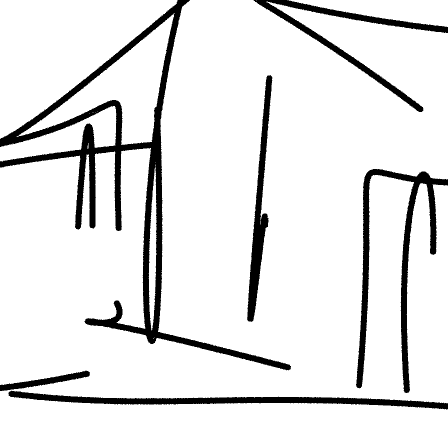} &
    \includegraphics[width=0.0875\linewidth]{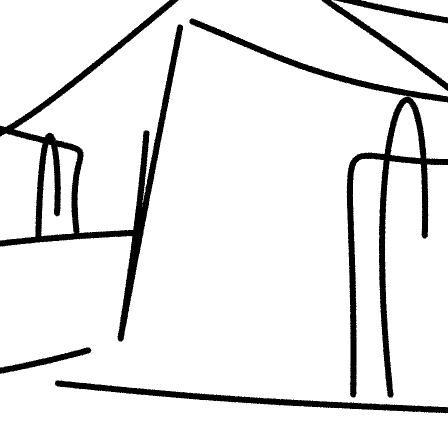} \\

    \includegraphics[width=0.0875\linewidth]{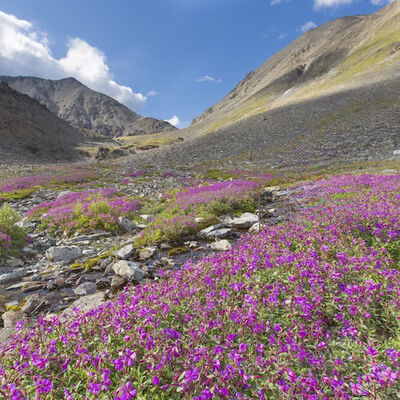} &
    \hspace{0.1cm}
    \includegraphics[width=0.0875\linewidth]{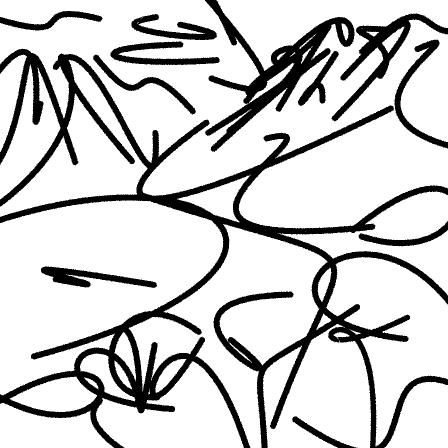} &
    \includegraphics[width=0.0875\linewidth]{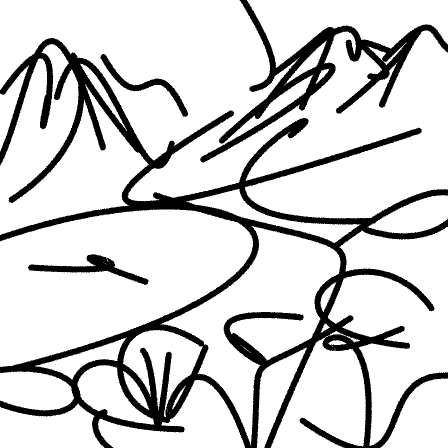} &
    \includegraphics[width=0.0875\linewidth]{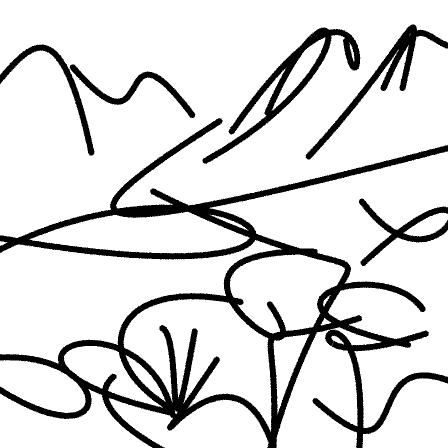} &
    \includegraphics[width=0.0875\linewidth]{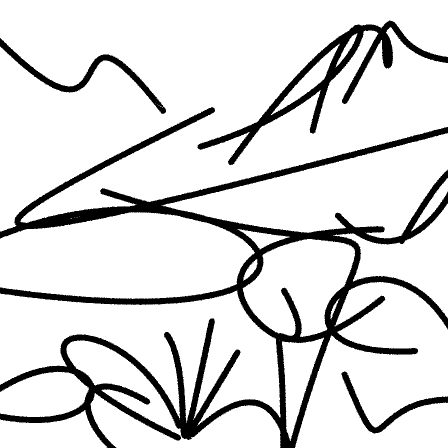} &
    \hspace{0.1cm}
    \includegraphics[width=0.0875\linewidth]{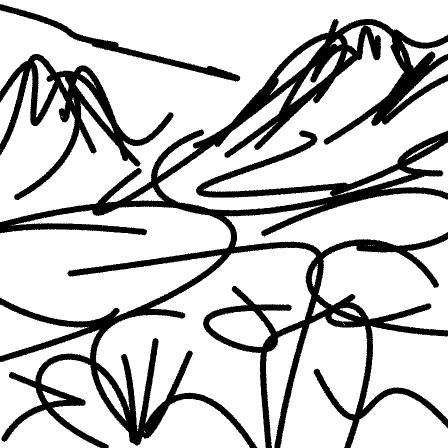} &
    \includegraphics[width=0.0875\linewidth]{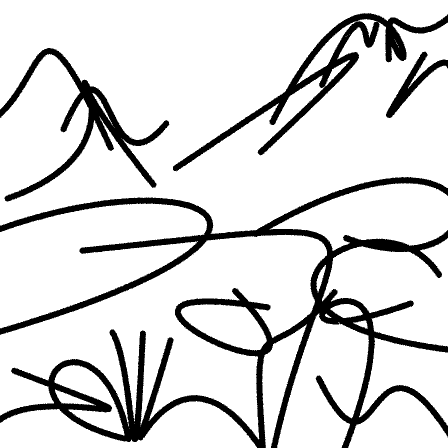} &
    \includegraphics[width=0.0875\linewidth]{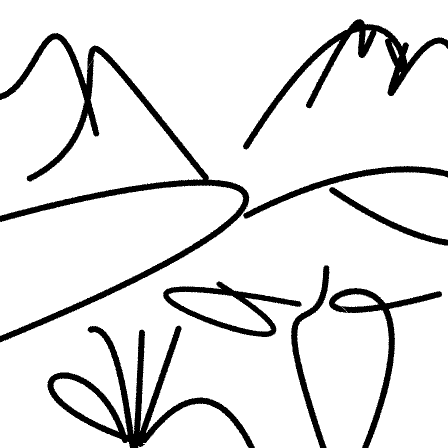} &
    \includegraphics[width=0.0875\linewidth]{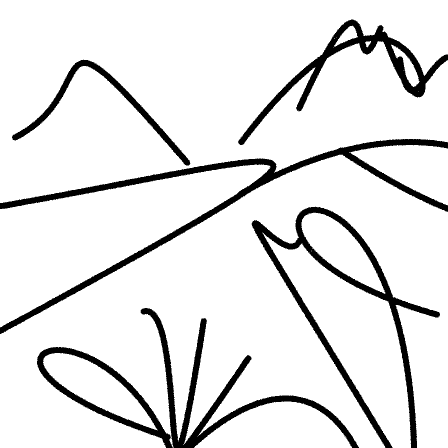} \\

    Input &
    \multicolumn{4}{c}{\hspace{0.1cm} \xrfill[0.5ex]{0.5pt}\; Without $L_{ratio}$ \; \xrfill[0.5ex]{0.5pt} \hspace{0.3cm}} &
    \multicolumn{4}{c}{\hspace{0.1cm} \xrfill[0.5ex]{0.5pt}\; Ours \; \xrfill[0.5ex]{0.5pt}}
    
    \end{tabular}
    
    \end{tabular}
    \vspace{0.2cm}
    \caption{Ablation results of replacing our $\mathcal{L}_{ratio}$ loss with an alternative loss $||\mathcal{L}_{sparse} - n_{target}||$  guided by a desired number of strokes $n_{target}$.}
    \label{fig:direct_lratio}
    
\end{figure*}

%% file: files/figures/supplementary/ablation_finetune_mlp_loc.tex
\begin{figure}[h]
    \centering
    \setlength{\tabcolsep}{1.5pt}
    {\small
    \begin{tabular}{c c c c c c}
        
        \includegraphics[width=0.1625\linewidth]{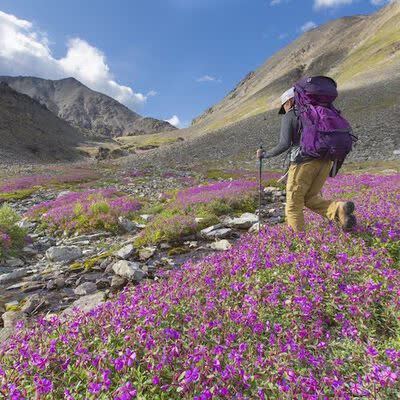} &
        \raisebox{0.05in}{\rotatebox{90}{\textcolor{cyan}{\begin{tabular}{c} w/o \\ Fine-tune \end{tabular}}}} &
        \includegraphics[width=0.1625\linewidth]{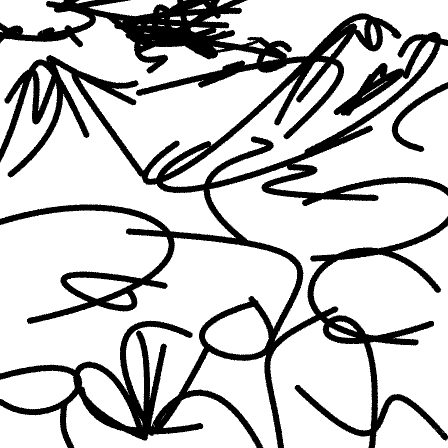} &
        \includegraphics[width=0.1625\linewidth]{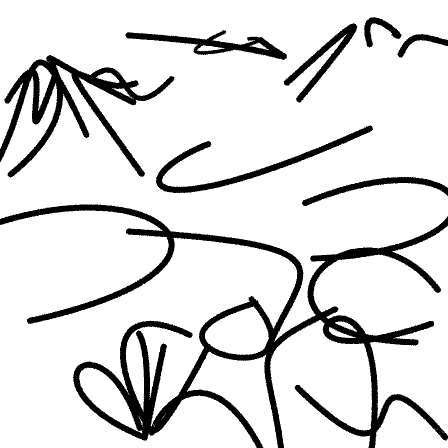} &
        \includegraphics[width=0.1625\linewidth]{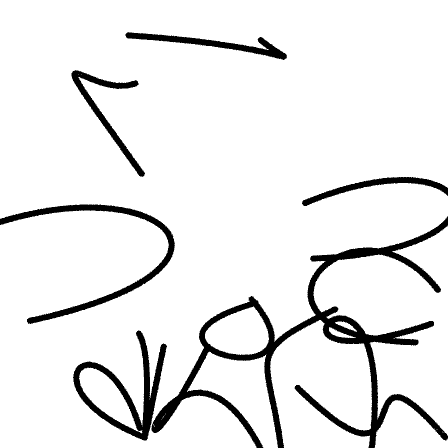} &
        \includegraphics[width=0.1625\linewidth]{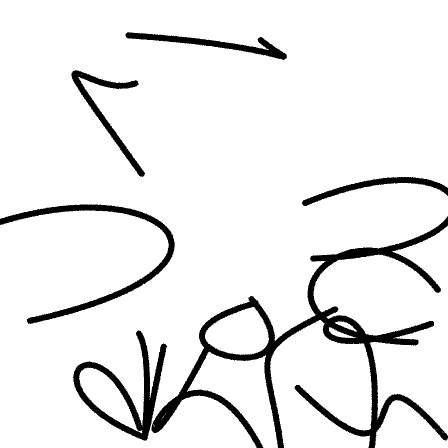} \\

        &
        \raisebox{0.05in}{\rotatebox{90}{\textcolor{orange}{\begin{tabular}{c} w/ \\ Fine-tune \end{tabular}}}} &
        \includegraphics[width=0.1625\linewidth]{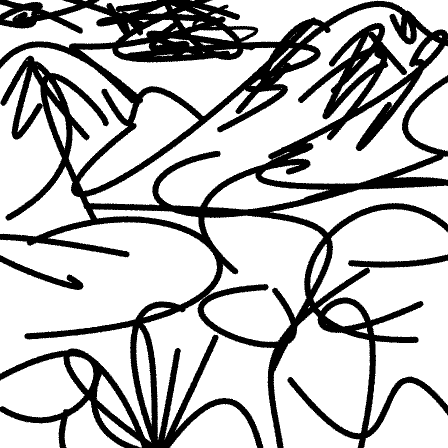} &
        \includegraphics[width=0.1625\linewidth]{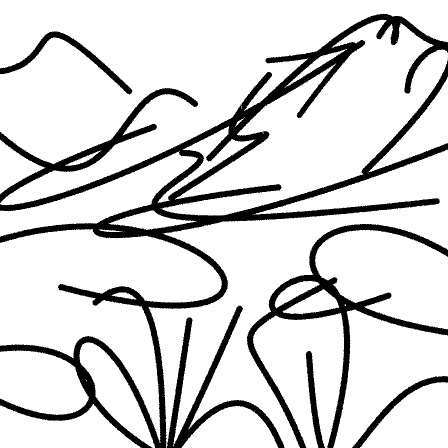} &
        \includegraphics[width=0.1625\linewidth]{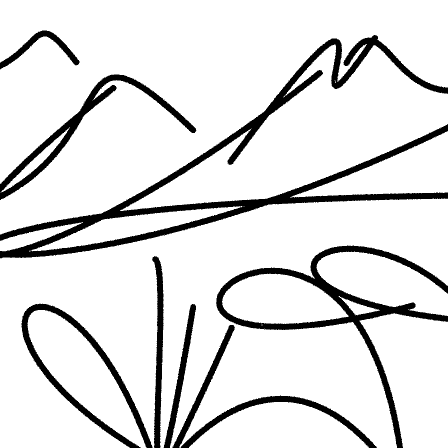} &
        \includegraphics[width=0.1625\linewidth]{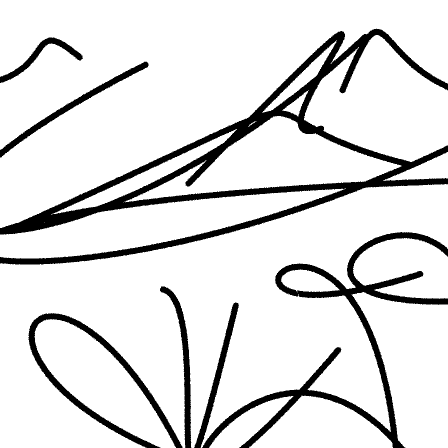} \\
        
        \cline{2-6} \\

        \includegraphics[width=0.1625\linewidth]{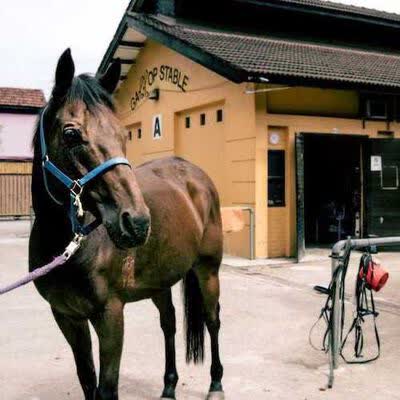} &
        \raisebox{0.05in}{\rotatebox{90}{\textcolor{cyan}{\begin{tabular}{c} w/o \\ Fine-tune \end{tabular}}}} &
        \includegraphics[width=0.1625\linewidth]{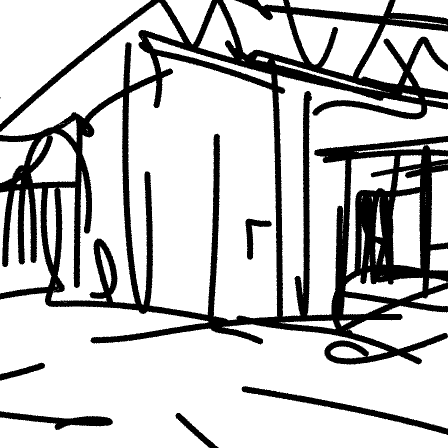} &
        \includegraphics[width=0.1625\linewidth]{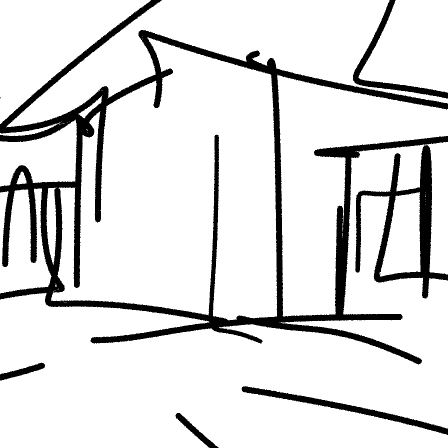} &
        \includegraphics[width=0.1625\linewidth]{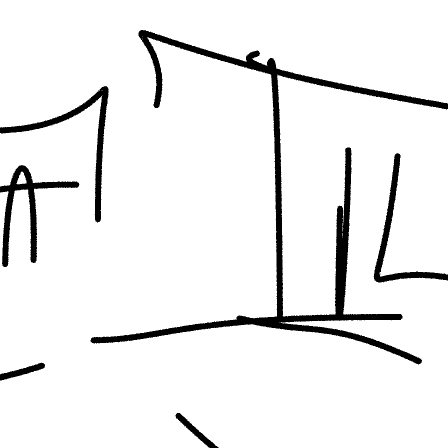} &
        \includegraphics[width=0.1625\linewidth]{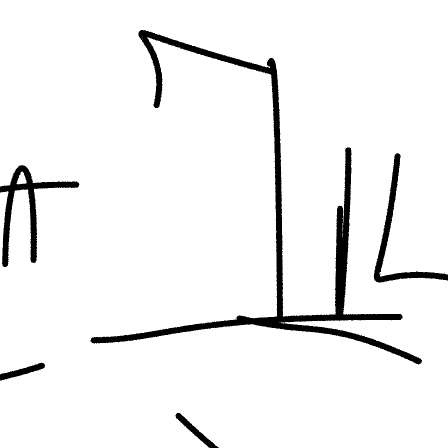} \\

        &
        \raisebox{0.05in}{\rotatebox{90}{\textcolor{orange}{\begin{tabular}{c} w/ \\ Fine-tune \end{tabular}}}} &
        \includegraphics[width=0.1625\linewidth]{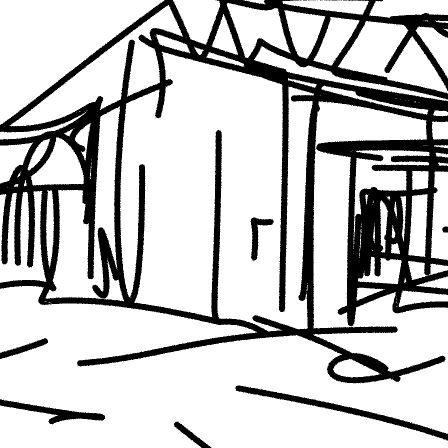} &
        \includegraphics[width=0.1625\linewidth]{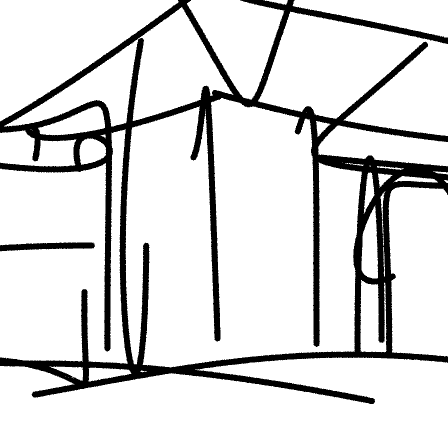} &
        \includegraphics[width=0.1625\linewidth]{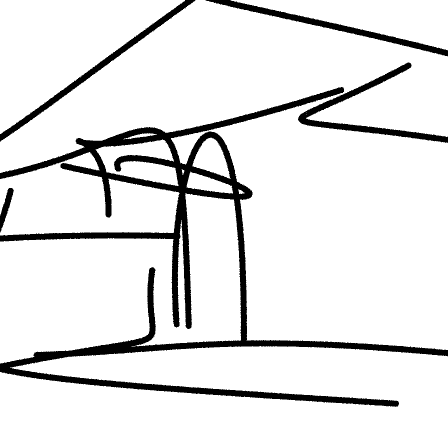} &
        \includegraphics[width=0.1625\linewidth]{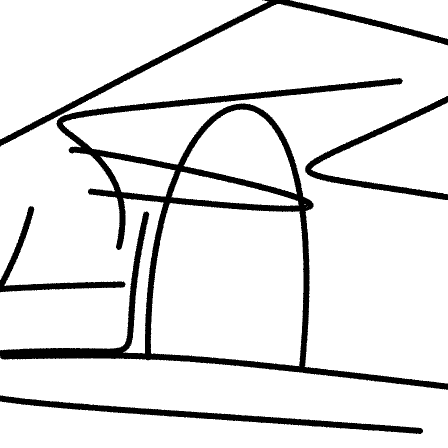} \\
        
        \cline{2-6} \\

        \includegraphics[width=0.1625\linewidth]{figs/inputs/woman_city.jpg} &
        \raisebox{0.05in}{\rotatebox{90}{\textcolor{cyan}{\begin{tabular}{c} w/o \\ Fine-tune \end{tabular}}}} &
        \includegraphics[width=0.1625\linewidth]{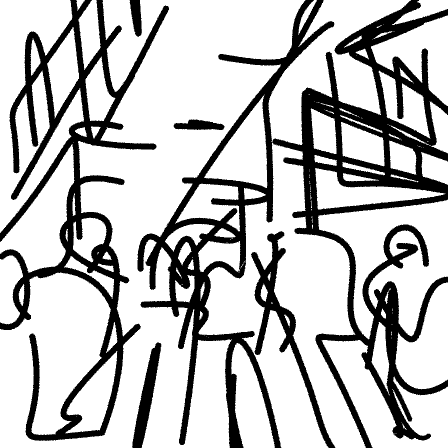} &
        \includegraphics[width=0.1625\linewidth]{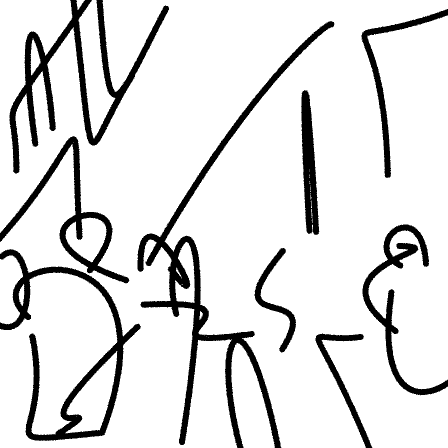} &
        \includegraphics[width=0.1625\linewidth]{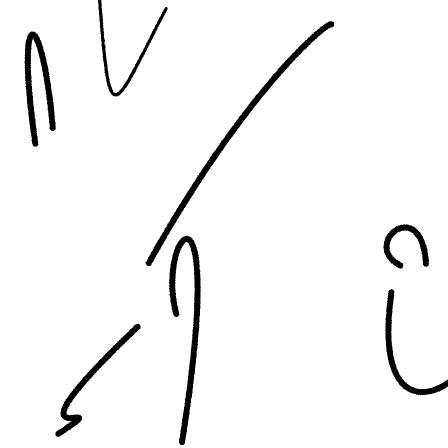} &
        \includegraphics[width=0.1625\linewidth]{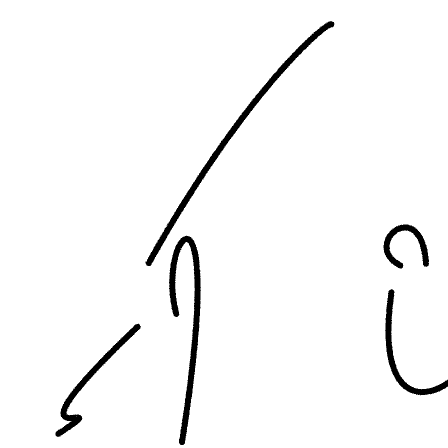} \\

        &
        \raisebox{0.05in}{\rotatebox{90}{\textcolor{orange}{\begin{tabular}{c} w/ \\ Fine-tune \end{tabular}}}} &
        \includegraphics[width=0.1625\linewidth]{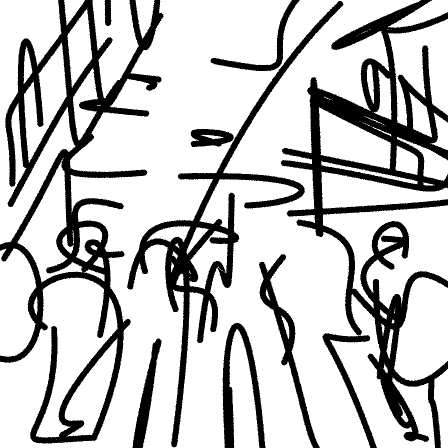} &
        \includegraphics[width=0.1625\linewidth]{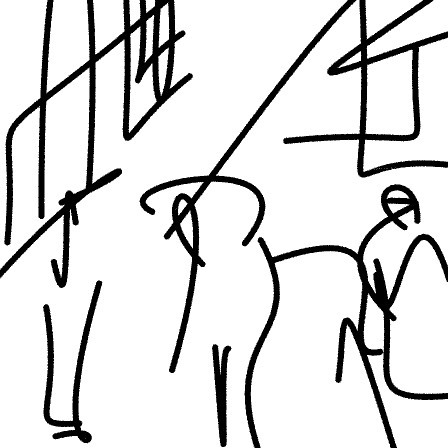} &
        \includegraphics[width=0.1625\linewidth]{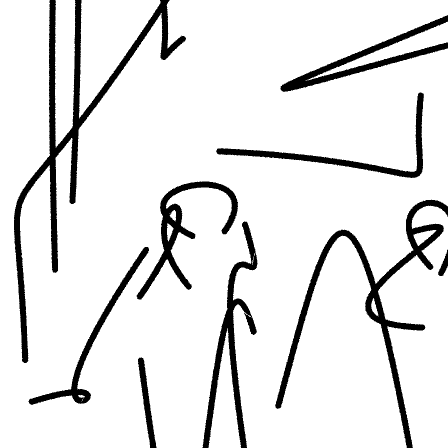} &
        \includegraphics[width=0.1625\linewidth]{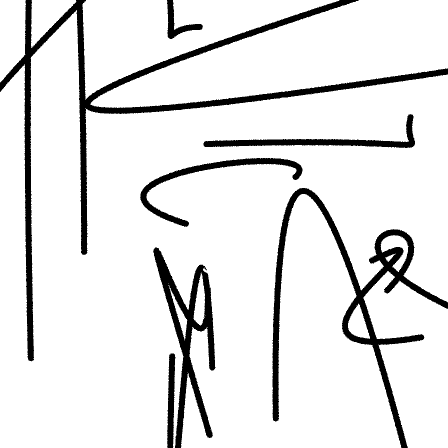} \\
        & & \multicolumn{4}{l}{\hspace{0.05cm} Simplicity Axis $\longrightarrow$} \\

    \end{tabular}
    
    }
    \vspace{0.2cm}
    \caption{Ablation of fine-tuning $MLP_{loc}$ when training $MLP_{simp}$ during the simplification process. The sketches presented here are obtained using layer $11$ of the CLIP-ViT model.}
    \label{fig:ablation_finetune_mlp_loc}
\end{figure}

%% file: files/figures/supplementary/ablation_linear_fk.tex
\begin{figure*}[h]
    \centering
    \setlength{\belowcaptionskip}{-6pt}
    \setlength{\tabcolsep}{1.5pt}
    {\small
    \begin{tabular}{c c c c c c c c @{\hspace{0.2cm}} | c c c c c c c c}
        
        \multicolumn{8}{c}{\xrfill[0.5ex]{0.5pt}\; Linear \; \xrfill[0.5ex]{0.5pt}} &
        \multicolumn{8}{c}{\hspace{0.1cm} \xrfill[0.5ex]{0.5pt}\; Ours (Exponential) \; \xrfill[0.5ex]{0.5pt} \hspace{0.2cm}} \\
        
        \cmidrule{1-8} 
        \cmidrule{9-16}
        
        \includegraphics[width=0.055\linewidth]{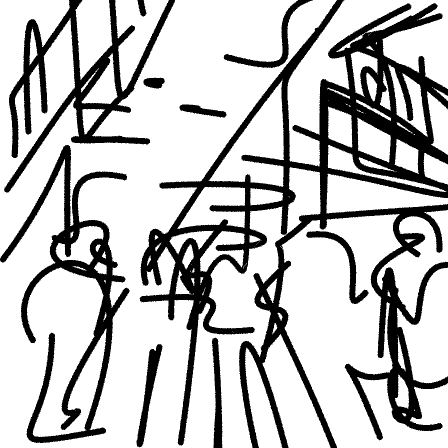} &
        \includegraphics[width=0.055\linewidth]{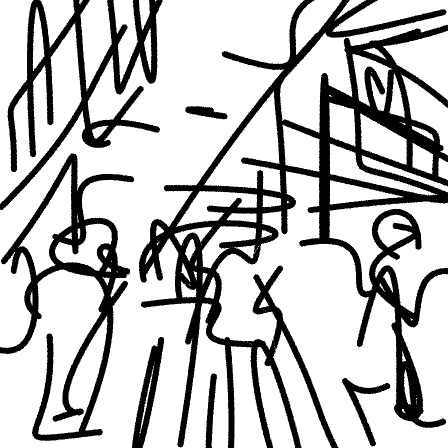} &
        \includegraphics[width=0.055\linewidth]{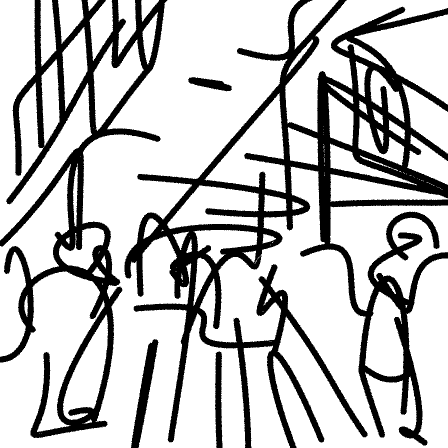} &
        \includegraphics[width=0.055\linewidth]{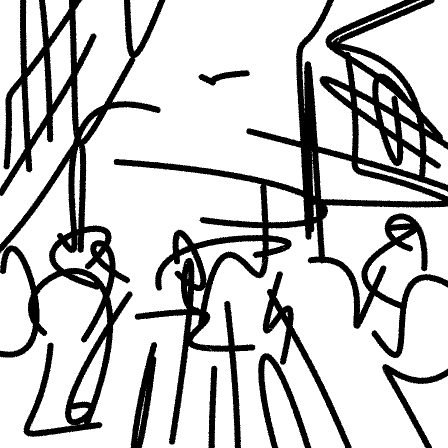} &
        \includegraphics[width=0.055\linewidth]{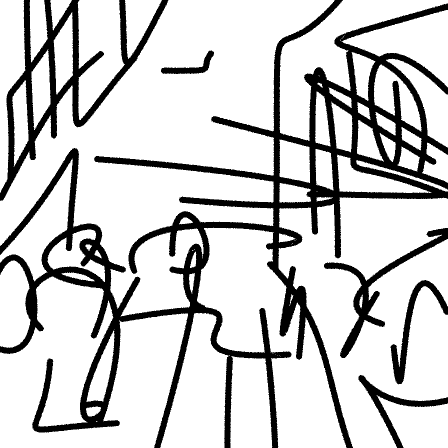} &
        \includegraphics[width=0.055\linewidth]{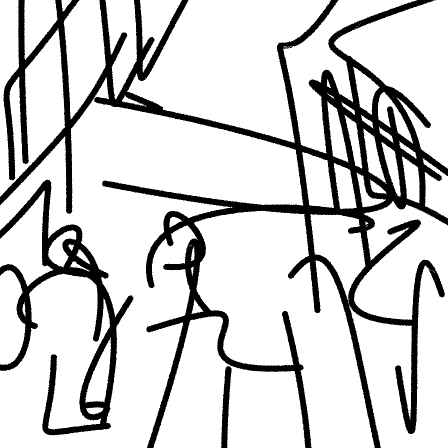} &
        \includegraphics[width=0.055\linewidth]{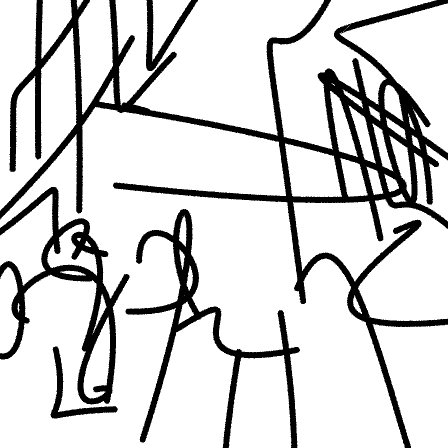} &
        \includegraphics[width=0.055\linewidth]{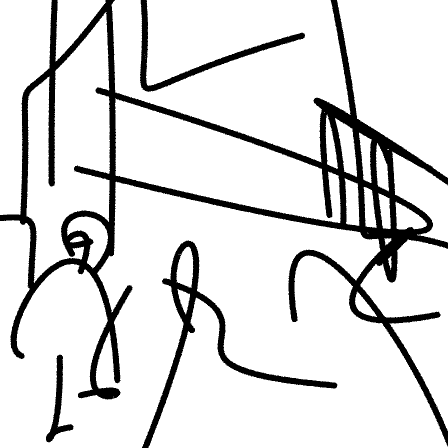} &
        \hspace{0.1cm}
        \includegraphics[width=0.055\linewidth]{figs/ablations/linear_fk/woman_city_exp_abs1.png} &
        \includegraphics[width=0.055\linewidth]{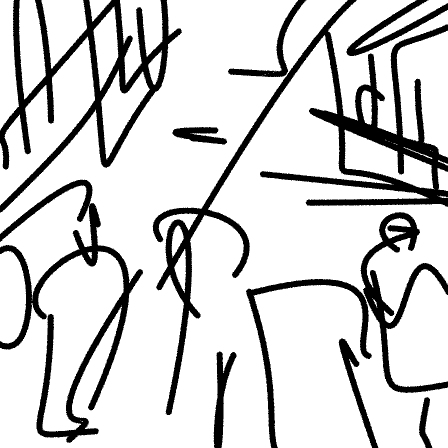} &
        \includegraphics[width=0.055\linewidth]{figs/ablations/linear_fk/woman_city_exp_abs3.png} &
        \includegraphics[width=0.055\linewidth]{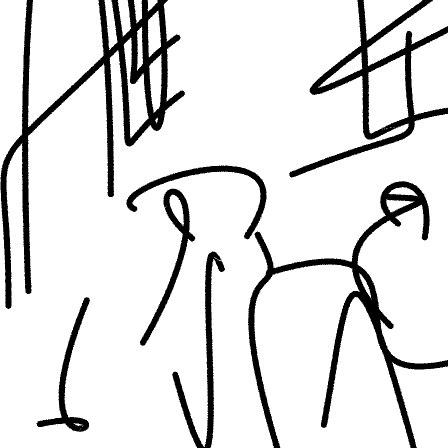} &
        \includegraphics[width=0.055\linewidth]{figs/ablations/linear_fk/woman_city_exp_abs5.png} &
        \includegraphics[width=0.055\linewidth]{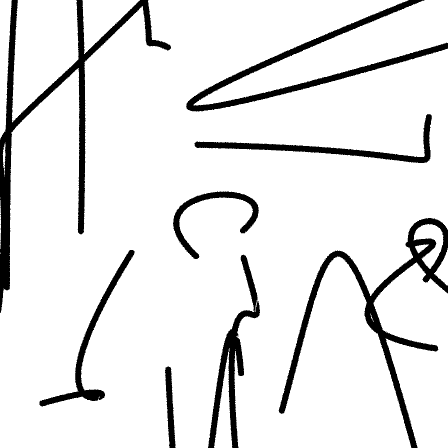} &
        \includegraphics[width=0.055\linewidth]{figs/ablations/linear_fk/woman_city_exp_abs7.png} &
        \includegraphics[width=0.055\linewidth]{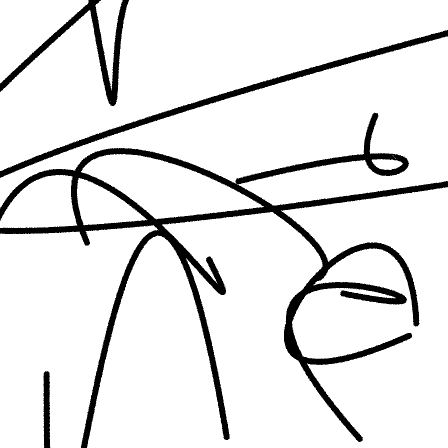} \\

        \multicolumn{4}{c}{\includegraphics[width=0.18\linewidth]{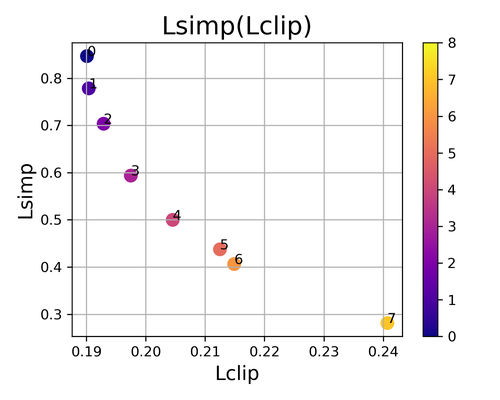}} & 
        \multicolumn{4}{c|}{\includegraphics[width=0.18\linewidth]{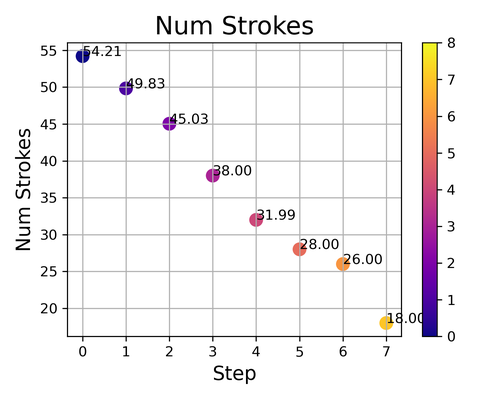}} &
        \multicolumn{4}{c}{\includegraphics[width=0.18\linewidth]{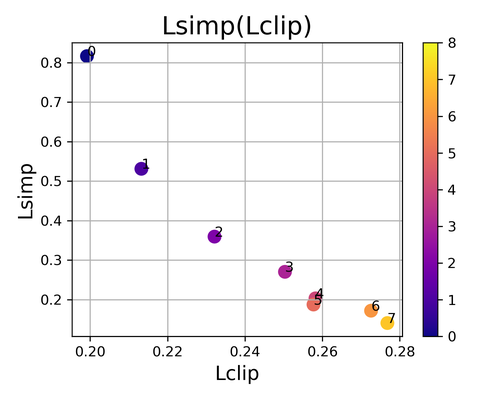}} &
        \multicolumn{4}{c}{\includegraphics[width=0.18\linewidth]{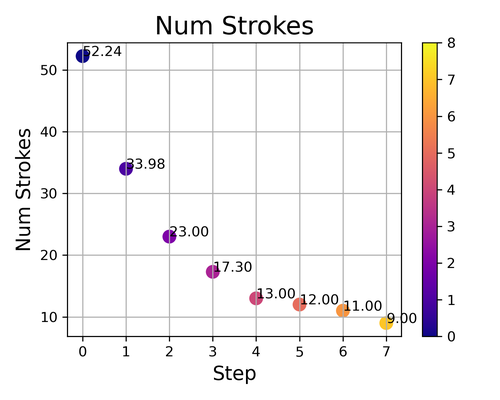}} \\

        \cmidrule{1-8} 
        \cmidrule{9-16}

        \includegraphics[width=0.055\linewidth]{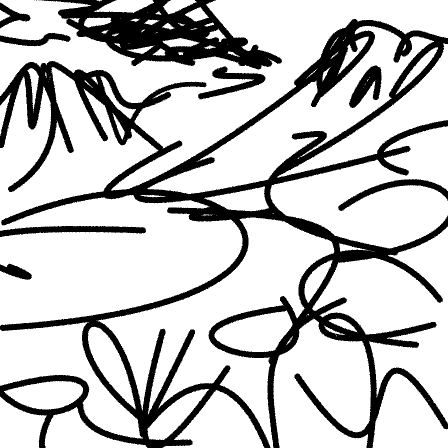} &
        \includegraphics[width=0.055\linewidth]{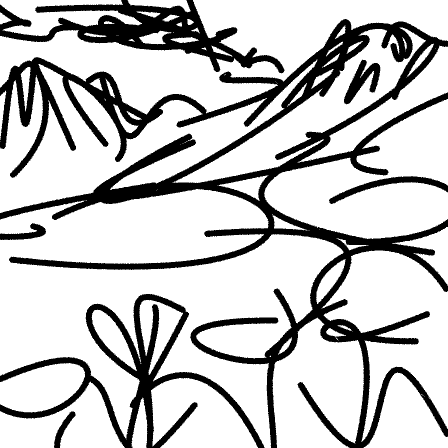} &
        \includegraphics[width=0.055\linewidth]{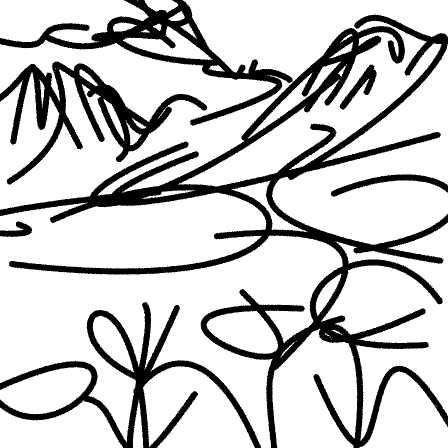} &
        \includegraphics[width=0.055\linewidth]{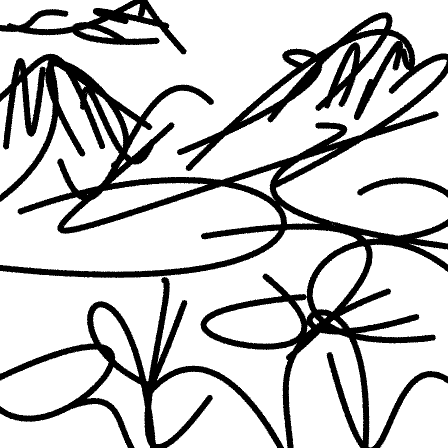} &
        \includegraphics[width=0.055\linewidth]{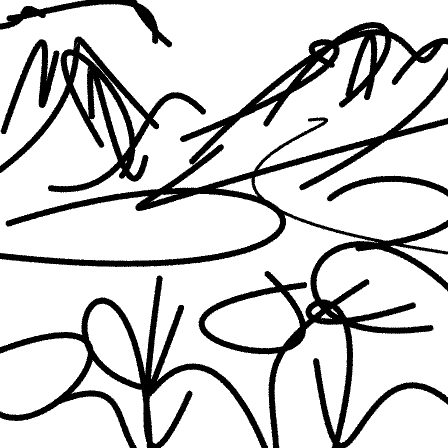} &
        \includegraphics[width=0.055\linewidth]{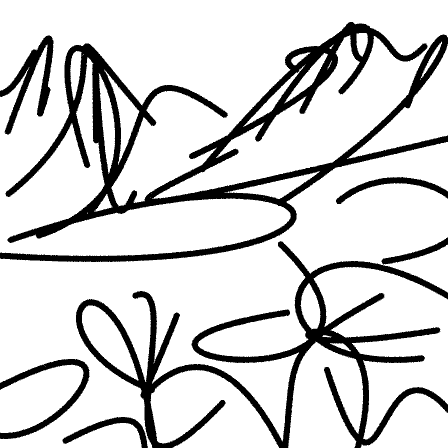} &
        \includegraphics[width=0.055\linewidth]{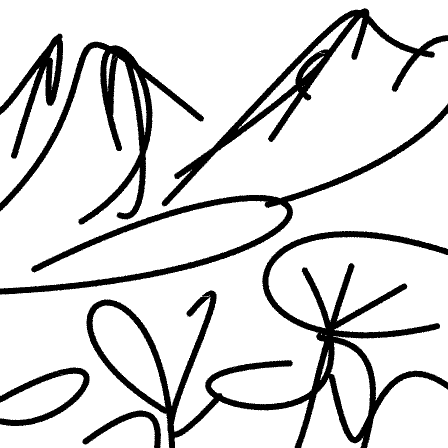} &
        \includegraphics[width=0.055\linewidth]{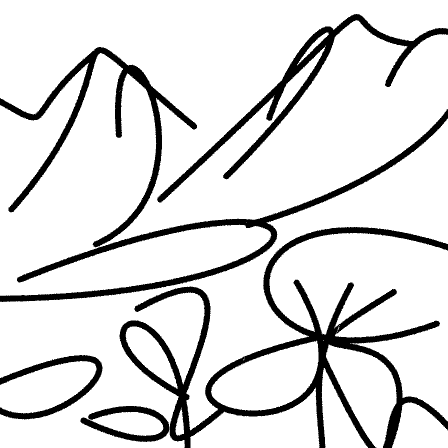} &
        \hspace{0.1cm}
        \includegraphics[width=0.055\linewidth]{figs/ablations/linear_fk/l11_man_flowers_exp_abs1.png} &
        \includegraphics[width=0.055\linewidth]{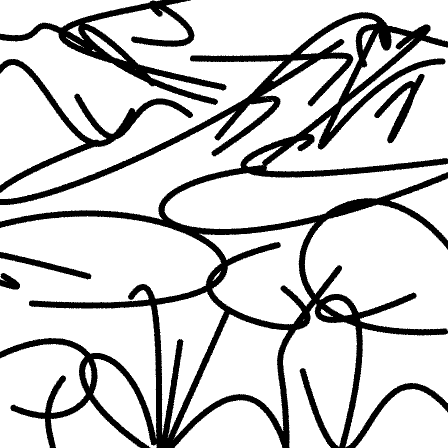} &
        \includegraphics[width=0.055\linewidth]{figs/ablations/linear_fk/l11_man_flowers_exp_abs3.png} &
        \includegraphics[width=0.055\linewidth]{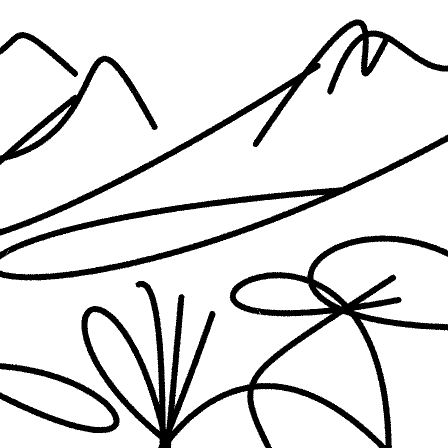} &
        \includegraphics[width=0.055\linewidth]{figs/ablations/linear_fk/l11_man_flowers_exp_abs5.png} &
        \includegraphics[width=0.055\linewidth]{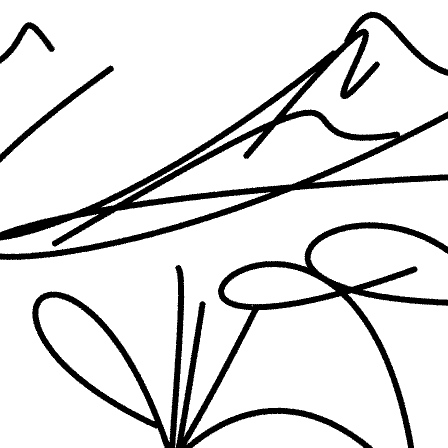} &
        \includegraphics[width=0.055\linewidth]{figs/ablations/linear_fk/l11_man_flowers_exp_abs7.png} &
        \includegraphics[width=0.055\linewidth]{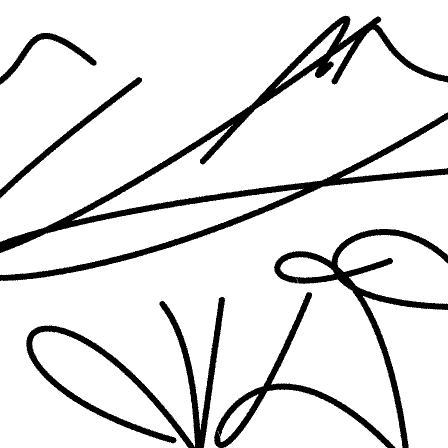} \\

        \multicolumn{4}{c}{\includegraphics[width=0.18\linewidth]{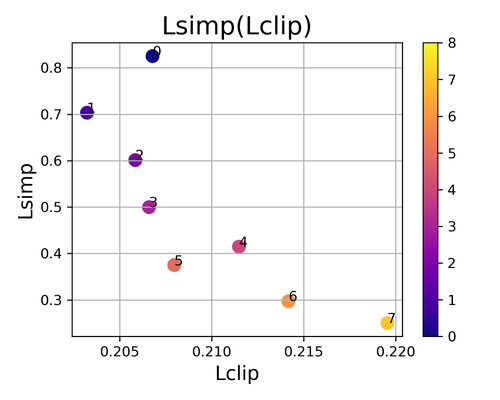}} & 
        \multicolumn{4}{c|}{\includegraphics[width=0.18\linewidth]{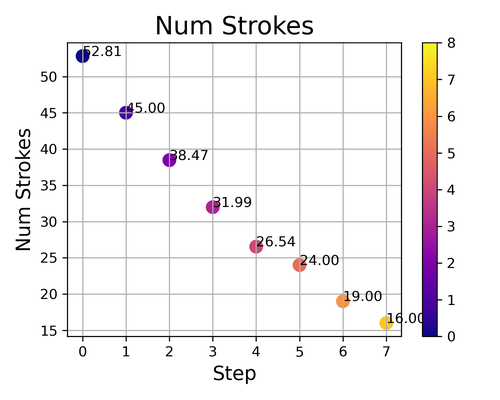}} &
        \multicolumn{4}{c}{\includegraphics[width=0.18\linewidth]{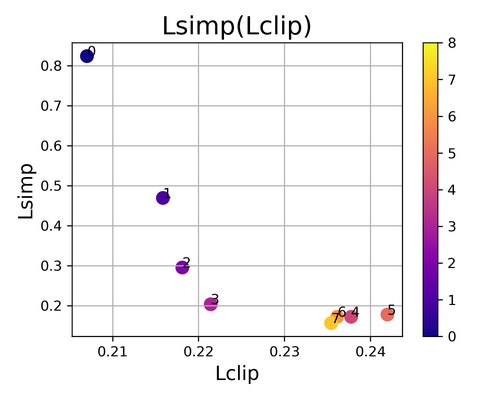}} &
        \multicolumn{4}{c}{\includegraphics[width=0.18\linewidth]{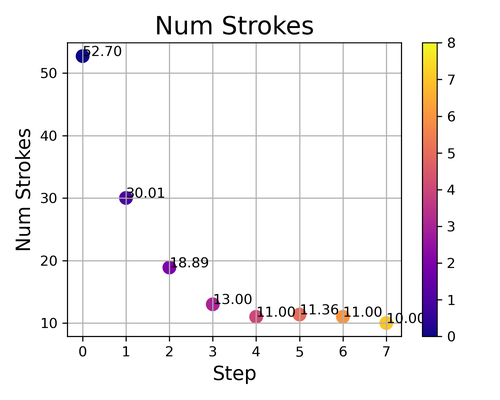}} \\

        \cmidrule{1-8} 
        \cmidrule{9-16}
    
        \includegraphics[width=0.055\linewidth]{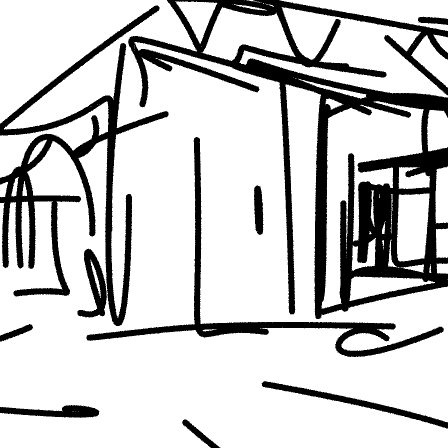} &
        \includegraphics[width=0.055\linewidth]{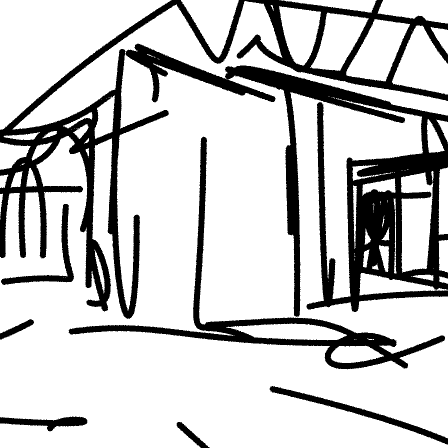} &
        \includegraphics[width=0.055\linewidth]{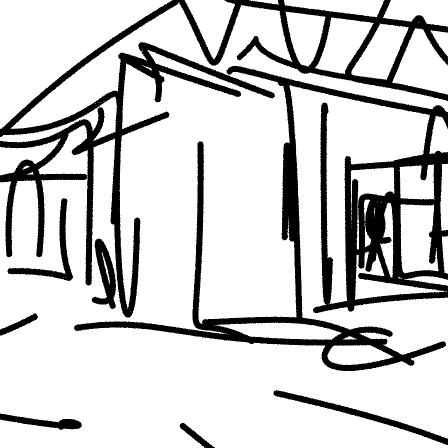} &
        \includegraphics[width=0.055\linewidth]{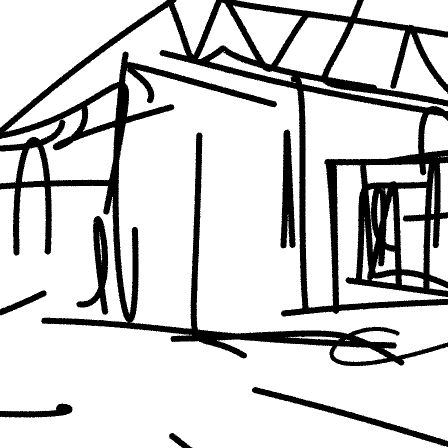} &
        \includegraphics[width=0.055\linewidth]{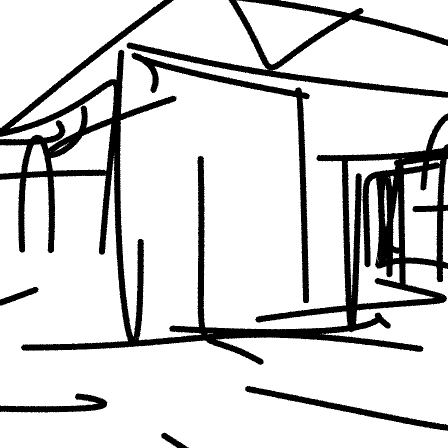} &
        \includegraphics[width=0.055\linewidth]{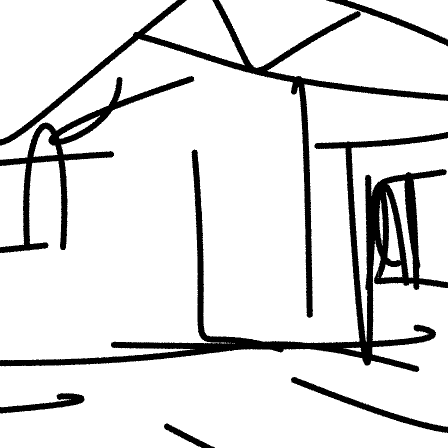} &
        \includegraphics[width=0.055\linewidth]{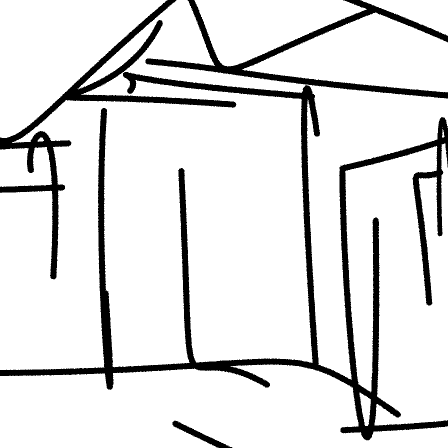} &
        \includegraphics[width=0.055\linewidth]{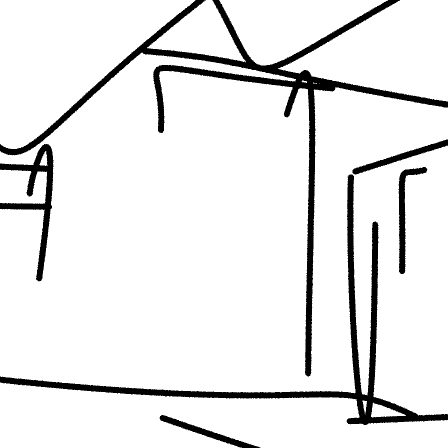} &
        \hspace{0.1cm}
        \includegraphics[width=0.055\linewidth]{figs/ablations/linear_fk/l11_semi-complex_exp_abs1.png} &
        \includegraphics[width=0.055\linewidth]{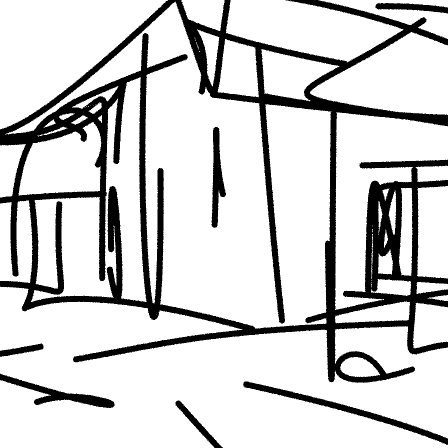} &
        \includegraphics[width=0.055\linewidth]{figs/ablations/linear_fk/l11_semi-complex_exp_abs3.png} &
        \includegraphics[width=0.055\linewidth]{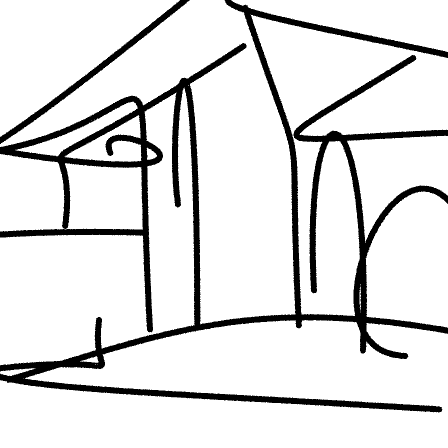} &
        \includegraphics[width=0.055\linewidth]{figs/ablations/linear_fk/l11_semi-complex_exp_abs5.png} &
        \includegraphics[width=0.055\linewidth]{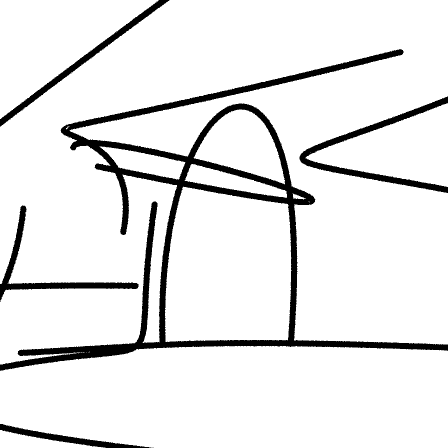} &
        \includegraphics[width=0.055\linewidth]{figs/ablations/linear_fk/l11_semi-complex_exp_abs7.png} &
        \includegraphics[width=0.055\linewidth]{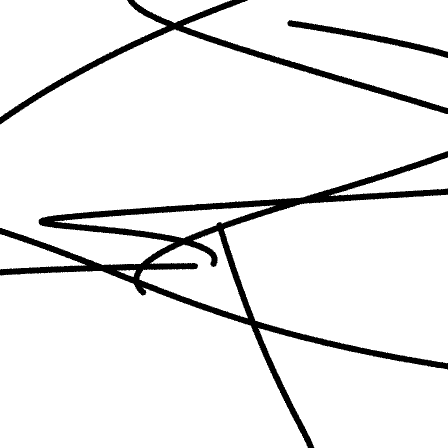} \\

        \multicolumn{4}{c}{\includegraphics[width=0.18\linewidth]{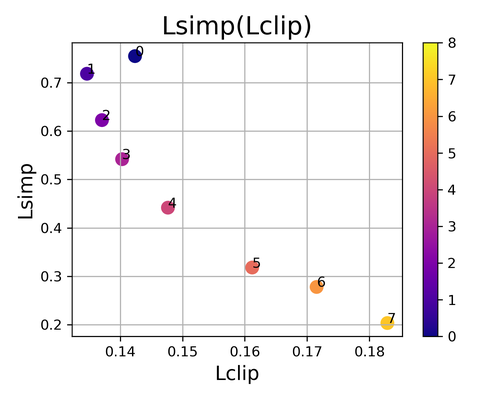}} & 
        \multicolumn{4}{c|}{\includegraphics[width=0.18\linewidth]{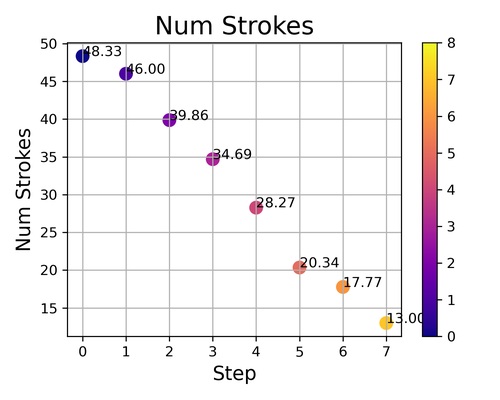}} &
        \multicolumn{4}{c}{\includegraphics[width=0.18\linewidth]{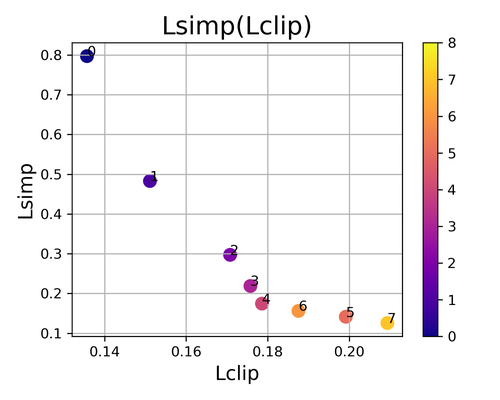}} &
        \multicolumn{4}{c}{\includegraphics[width=0.18\linewidth]{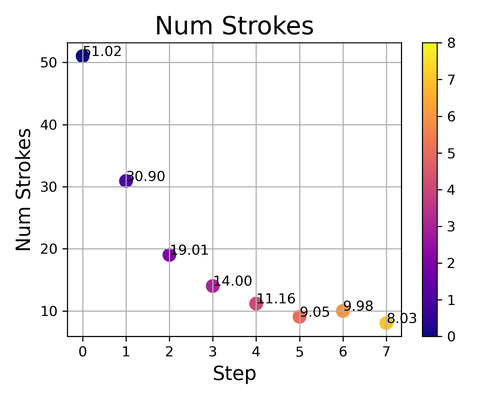}} \\

        \cmidrule{1-8} 
        \cmidrule{9-16}

        \multicolumn{8}{c}{\xrfill[0.5ex]{0.5pt}\; Linear \; \xrfill[0.5ex]{0.5pt}} &
        \multicolumn{8}{c}{\hspace{0.1cm} \xrfill[0.5ex]{0.5pt}\; Ours (Exponential) \; \xrfill[0.5ex]{0.5pt} \hspace{0.2cm}} \\
        
    \end{tabular}
    
    }
    \vspace{0.2cm}
    \caption{Ablation of the defining an exponential $f_k$ when performing an iterative simplification of the sketch. On the right-hand side, we define a linear relation between the two loss objectives. On the left side, we show sketch results obtained when defining an exponential relation between $\mathcal{L}_{CLIP}$ and $\mathcal{L}_{sparse}$. For each set of simplified sketches, we additionally present two graphs depicting (1) the relation between the two loss objectives at each simplification step (left graph) and (2) the number of strokes used to compose the sketch (right graph).}
    \label{fig:ablation_linear_fk}
\end{figure*}

%% file: files/figures/supplementary/same_ratios.tex
\begin{figure*}[!ht]
    \centering
    \setlength{\belowcaptionskip}{-6pt}
    \setlength{\tabcolsep}{1.5pt}
    {\small
        
    \begin{tabular}{c c c c c c c c c @{\hspace{0.2cm}} | c c c c c c c c}

        \cmidrule{2-9} 
        \cmidrule{10-17}

        {\footnotesize\raisebox{0.05in}{\rotatebox{90}{Layer 2}}} &
        \includegraphics[width=0.055\linewidth]{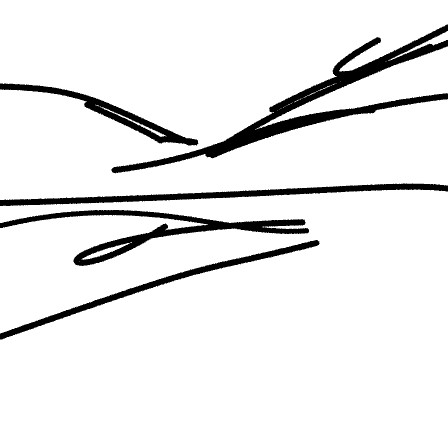} &
        \includegraphics[width=0.055\linewidth]{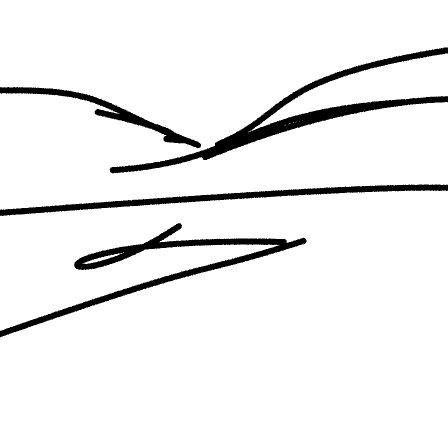} &
        \includegraphics[width=0.055\linewidth]{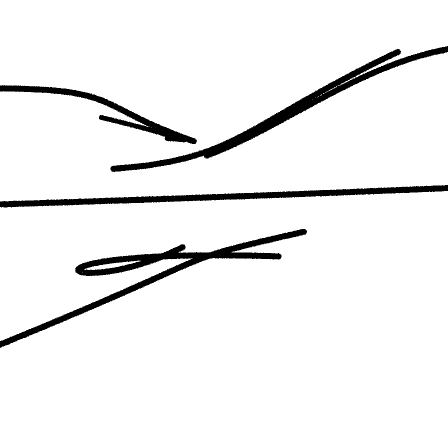} &
        \includegraphics[width=0.055\linewidth]{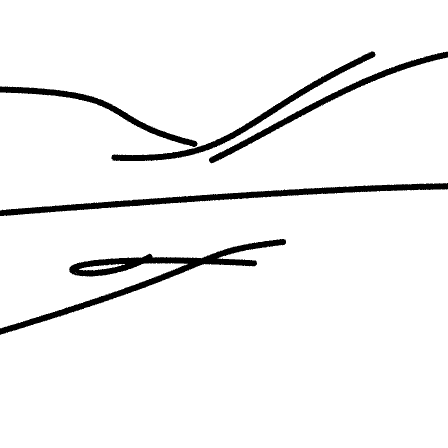} &
        \includegraphics[width=0.055\linewidth]{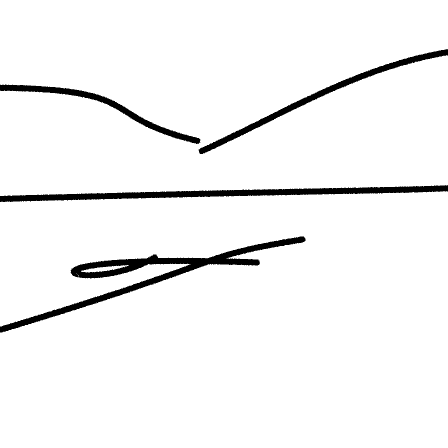} &
        \includegraphics[width=0.055\linewidth]{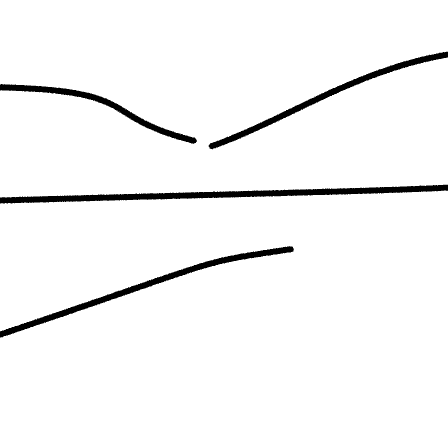} &
        \includegraphics[width=0.055\linewidth]{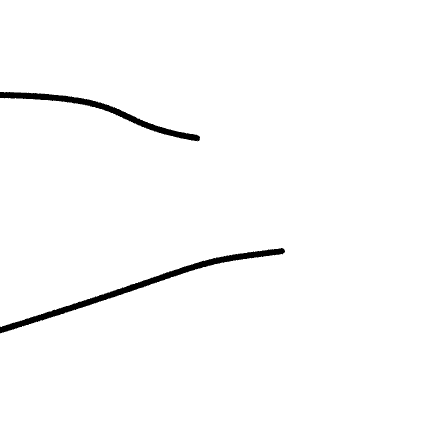} &
        \includegraphics[width=0.055\linewidth]{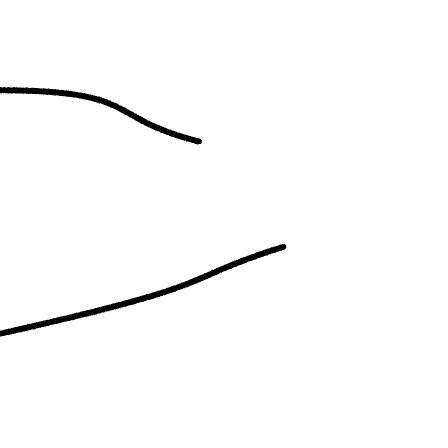} &
        \hspace{0.1cm}
        \includegraphics[width=0.055\linewidth]{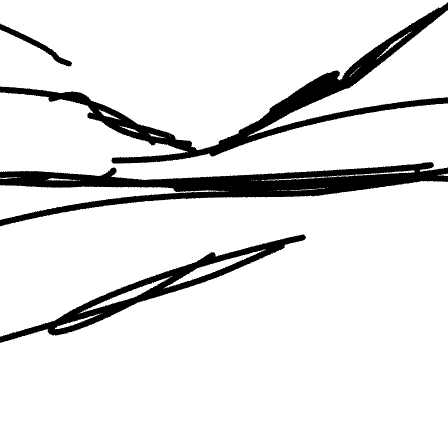} &
        \includegraphics[width=0.055\linewidth]{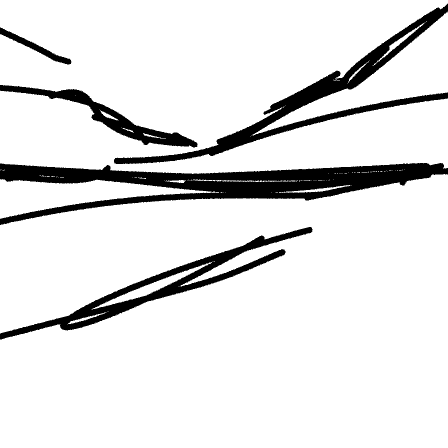} &
        \includegraphics[width=0.055\linewidth]{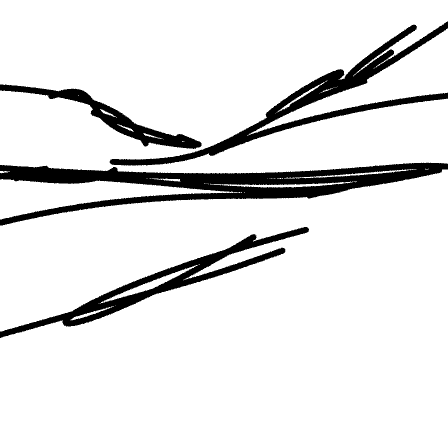} &
        \includegraphics[width=0.055\linewidth]{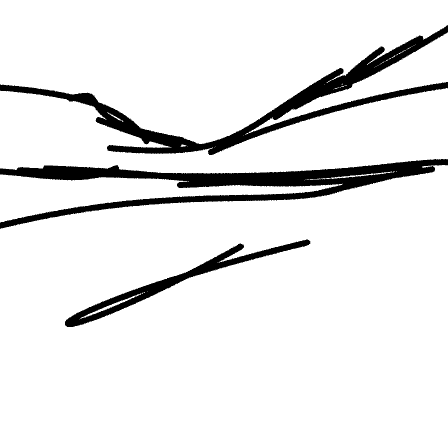} &
        \includegraphics[width=0.055\linewidth]{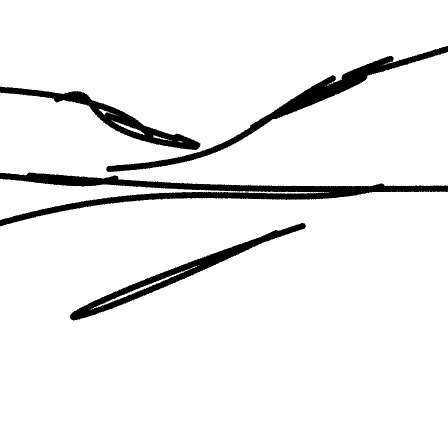} &
        \includegraphics[width=0.055\linewidth]{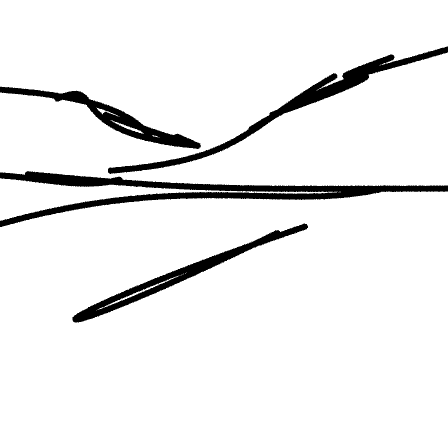} &
        \includegraphics[width=0.055\linewidth]{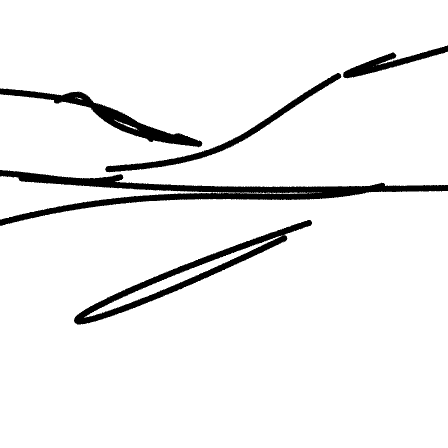} &
        \includegraphics[width=0.055\linewidth]{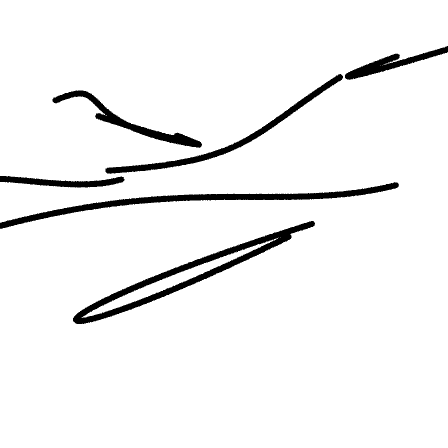} \\

        {\footnotesize\raisebox{0.05in}{\rotatebox{90}{Layer 7}}} &
        \includegraphics[width=0.055\linewidth]{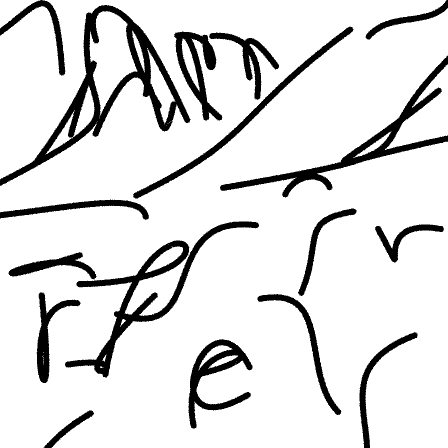} &
        \includegraphics[width=0.055\linewidth]{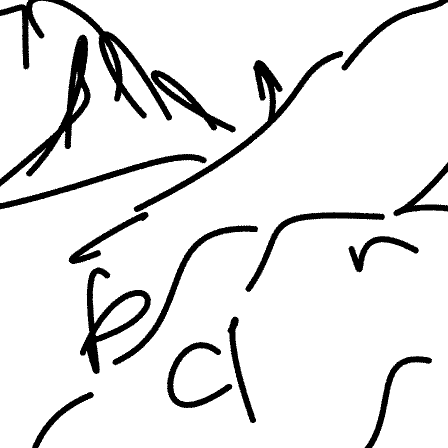} &
        \includegraphics[width=0.055\linewidth]{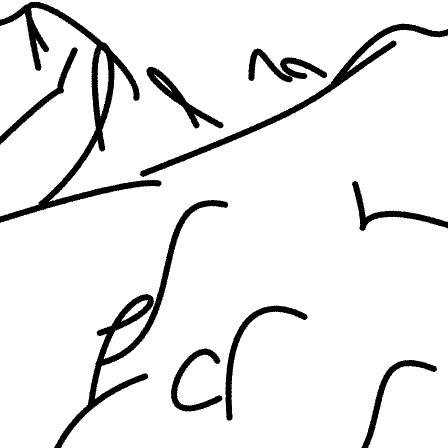} &
        \includegraphics[width=0.055\linewidth]{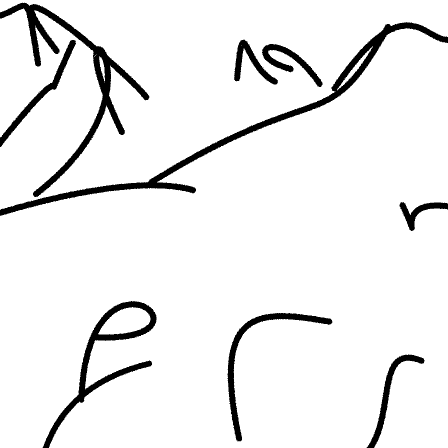} &
        \includegraphics[width=0.055\linewidth]{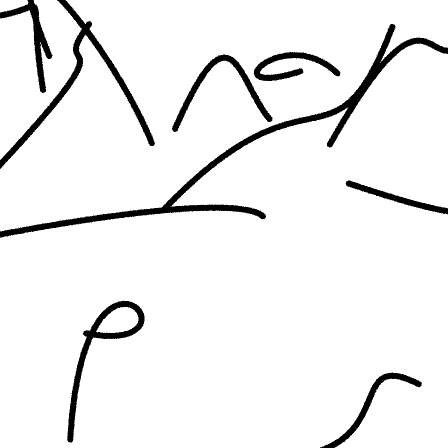} &
        \includegraphics[width=0.055\linewidth]{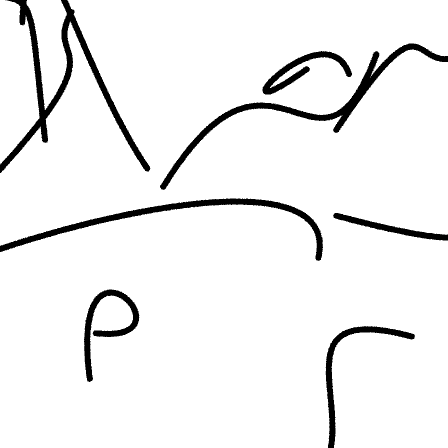} &
        \includegraphics[width=0.055\linewidth]{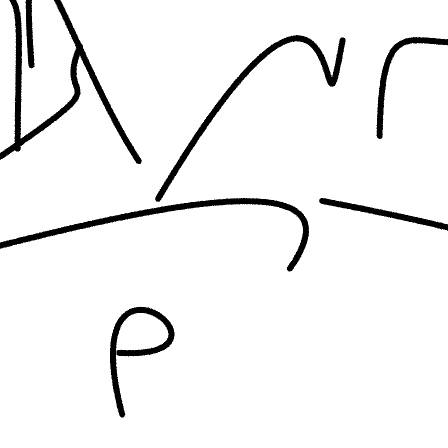} &
        \includegraphics[width=0.055\linewidth]{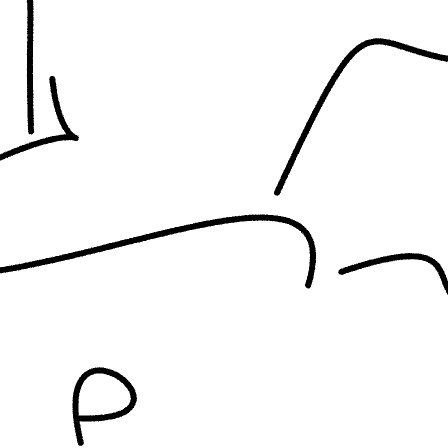} &
        \hspace{0.1cm}
        \includegraphics[width=0.055\linewidth]{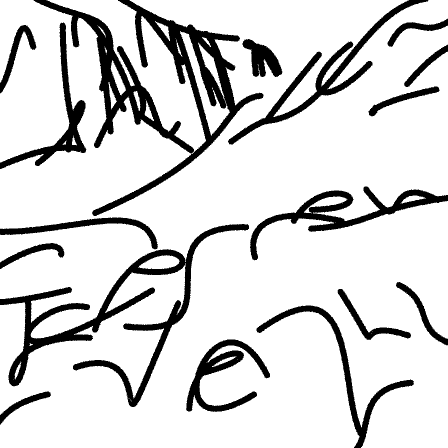} &
        \includegraphics[width=0.055\linewidth]{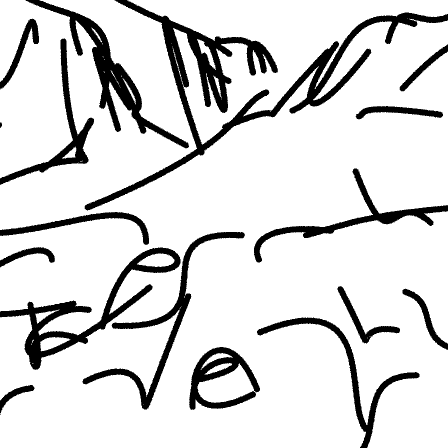} &
        \includegraphics[width=0.055\linewidth]{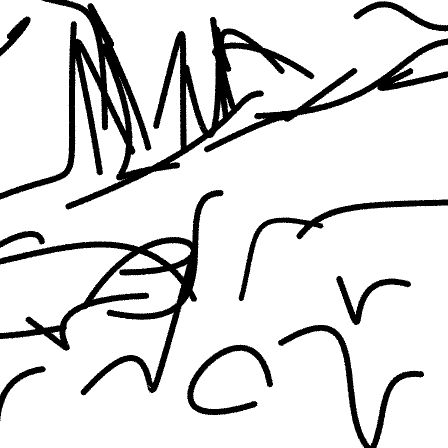} &
        \includegraphics[width=0.055\linewidth]{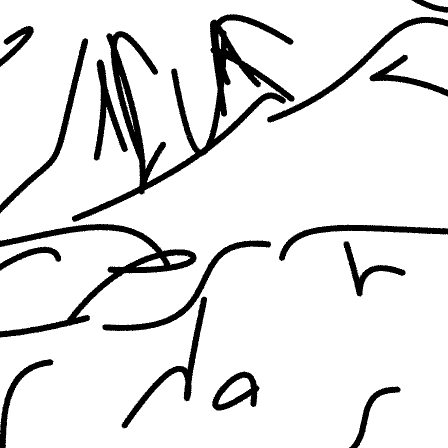} &
        \includegraphics[width=0.055\linewidth]{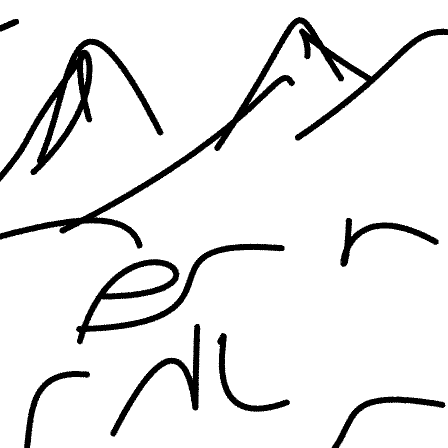} &
        \includegraphics[width=0.055\linewidth]{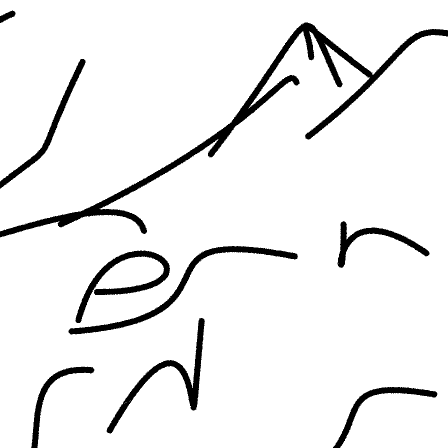} &
        \includegraphics[width=0.055\linewidth]{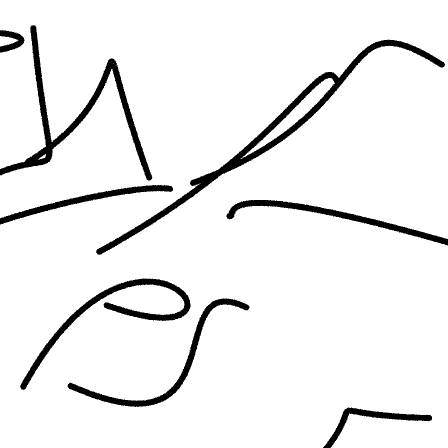} &
        \includegraphics[width=0.055\linewidth]{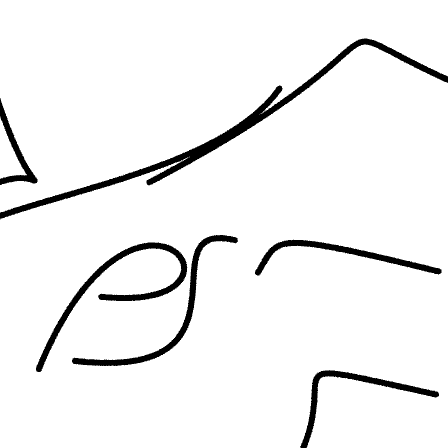} \\

        {\footnotesize\raisebox{0.05in}{\rotatebox{90}{Layer 8}}} &
        \includegraphics[width=0.055\linewidth]{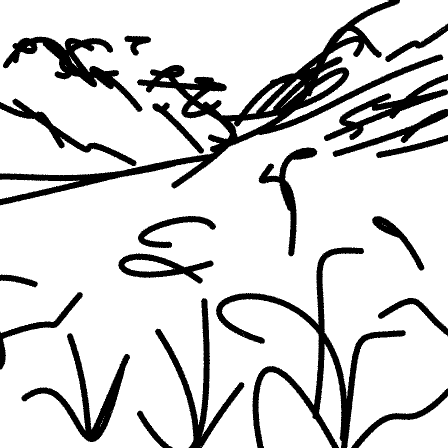} &
        \includegraphics[width=0.055\linewidth]{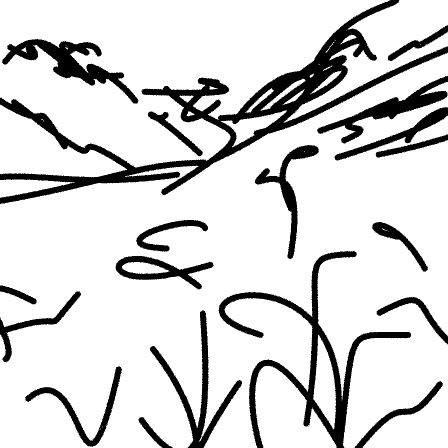} &
        \includegraphics[width=0.055\linewidth]{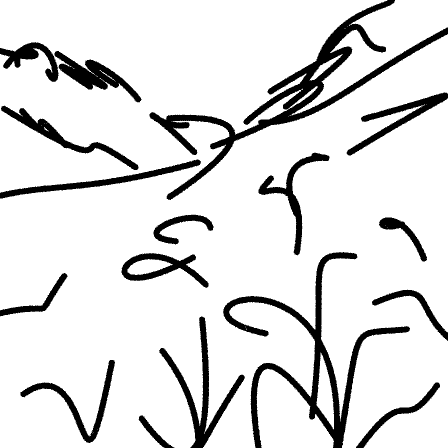} &
        \includegraphics[width=0.055\linewidth]{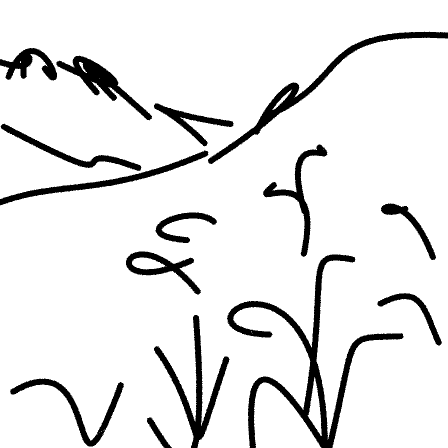} &
        \includegraphics[width=0.055\linewidth]{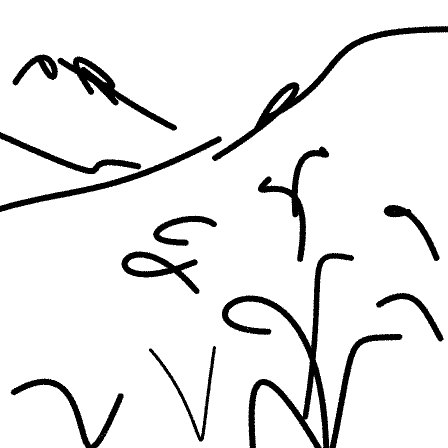} &
        \includegraphics[width=0.055\linewidth]{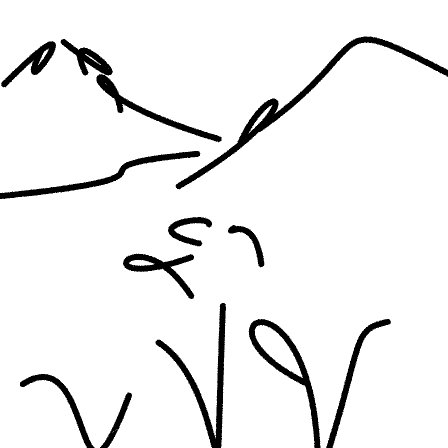} &
        \includegraphics[width=0.055\linewidth]{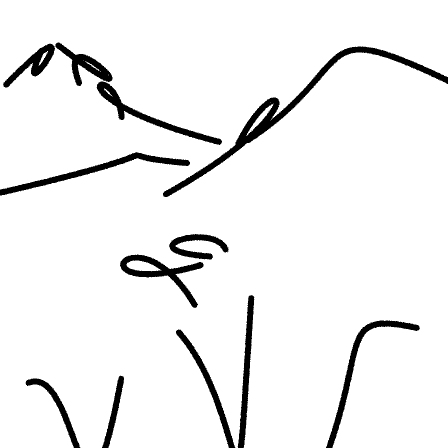} &
        \includegraphics[width=0.055\linewidth]{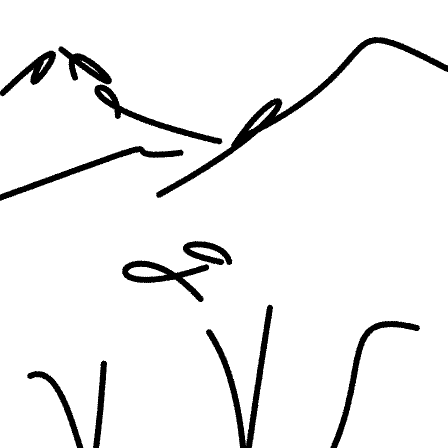} &
        \hspace{0.1cm}
        \includegraphics[width=0.055\linewidth]{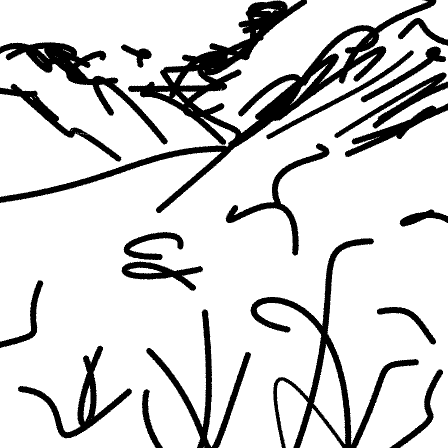} &
        \includegraphics[width=0.055\linewidth]{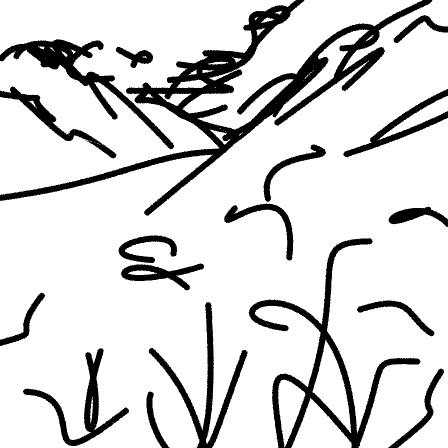} &
        \includegraphics[width=0.055\linewidth]{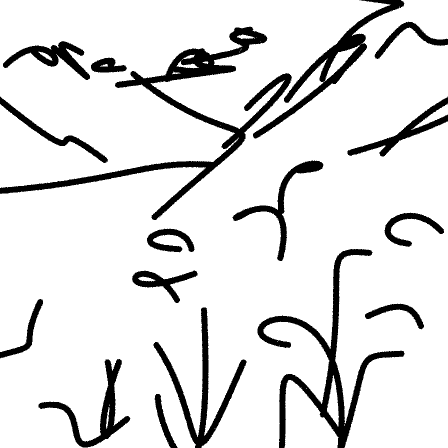} &
        \includegraphics[width=0.055\linewidth]{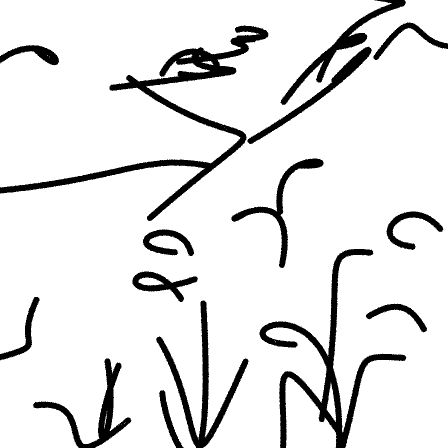} &
        \includegraphics[width=0.055\linewidth]{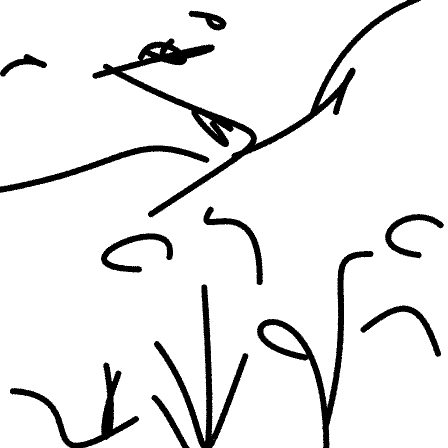} &
        \includegraphics[width=0.055\linewidth]{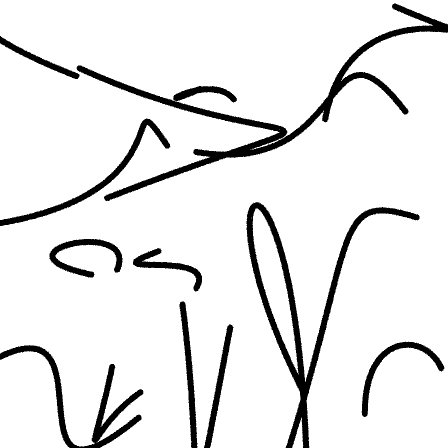} &
        \includegraphics[width=0.055\linewidth]{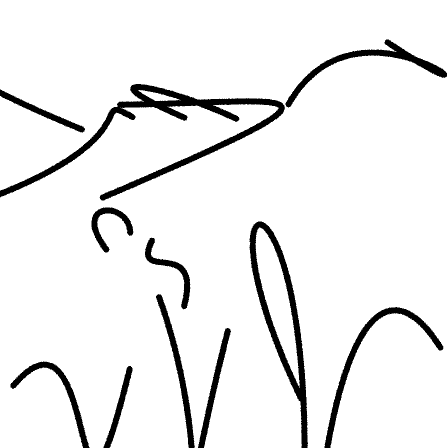} &
        \includegraphics[width=0.055\linewidth]{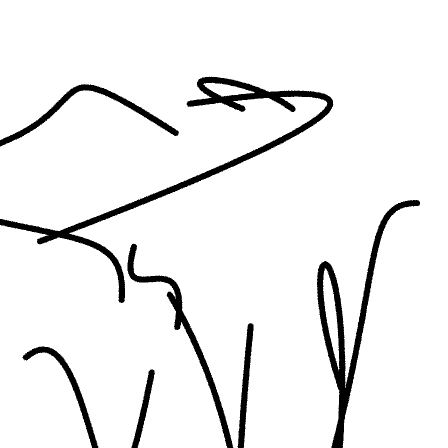} \\

        {\footnotesize\raisebox{0.05in}{\rotatebox{90}{Layer 11}}} &
        \includegraphics[width=0.055\linewidth]{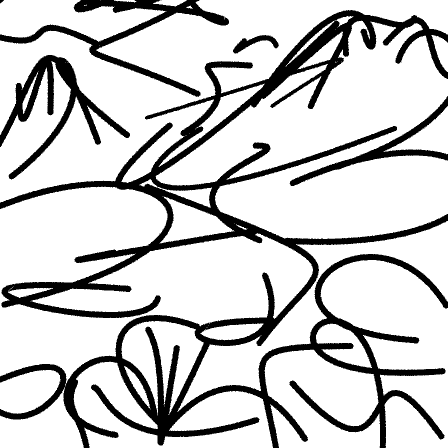} &
        \includegraphics[width=0.055\linewidth]{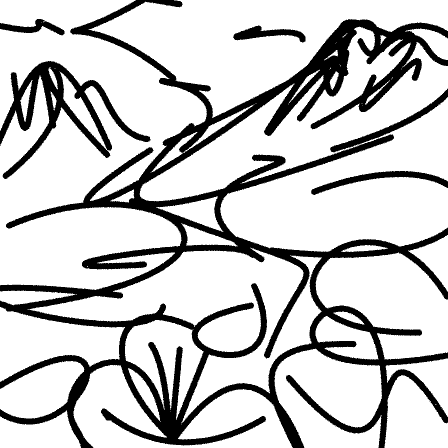} &
        \includegraphics[width=0.055\linewidth]{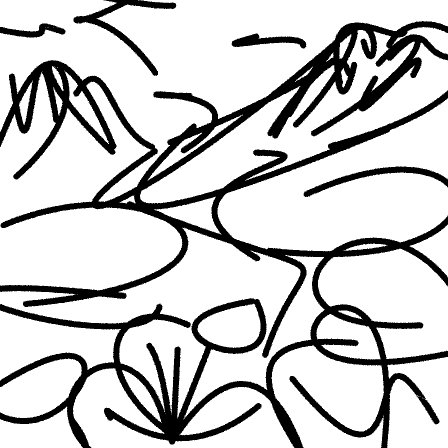} &
        \includegraphics[width=0.055\linewidth]{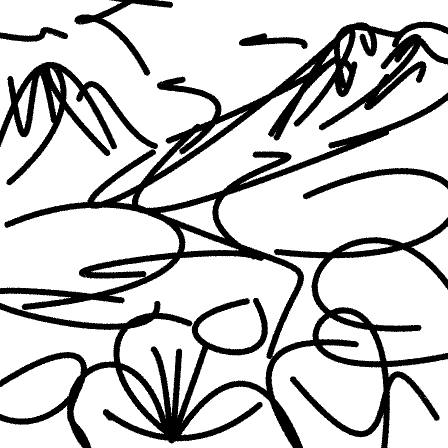} &
        \includegraphics[width=0.055\linewidth]{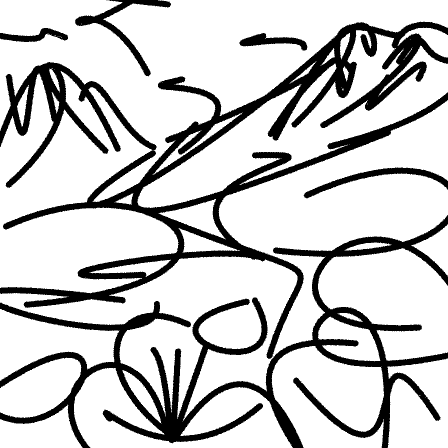} &
        \includegraphics[width=0.055\linewidth]{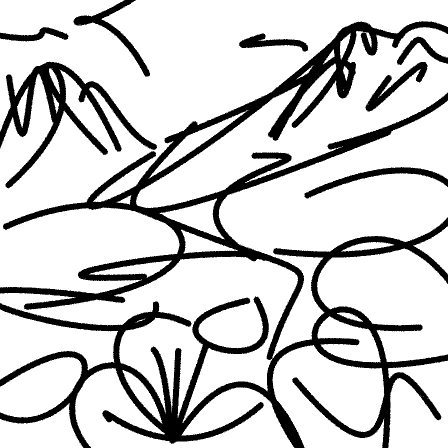} &
        \includegraphics[width=0.055\linewidth]{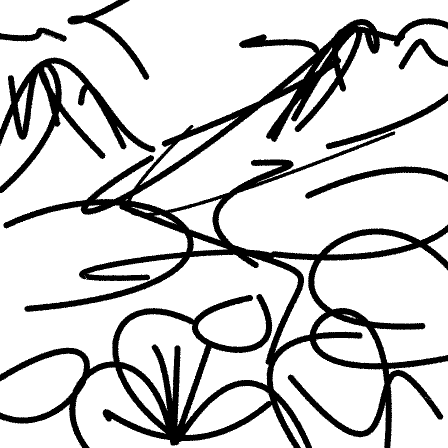} &
        \includegraphics[width=0.055\linewidth]{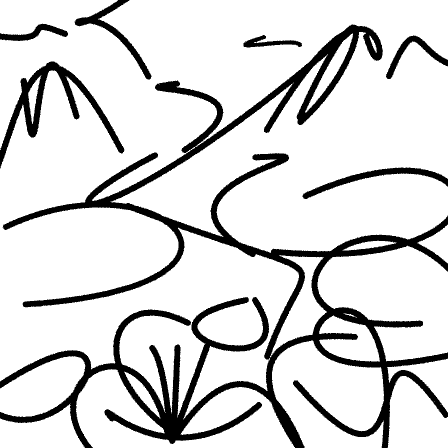} &
        \hspace{0.1cm}
        \includegraphics[width=0.055\linewidth]{figs/ablations/our_results_new/l11_man_flowers_original_abs1.png} &
        \includegraphics[width=0.055\linewidth]{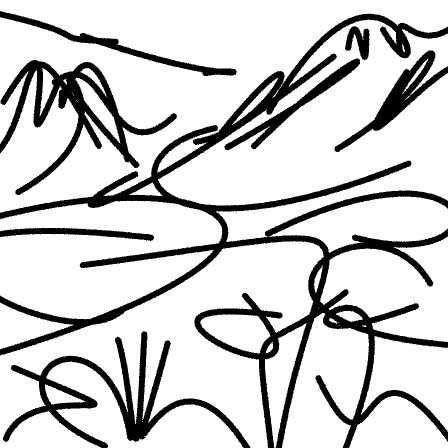} &
        \includegraphics[width=0.055\linewidth]{figs/ablations/our_results_new/l11_man_flowers_original_abs3.png} &
        \includegraphics[width=0.055\linewidth]{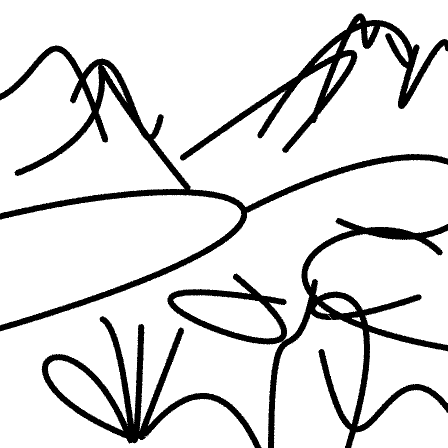} &
        \includegraphics[width=0.055\linewidth]{figs/ablations/our_results_new/l11_man_flowers_original_abs5.png} &
        \includegraphics[width=0.055\linewidth]{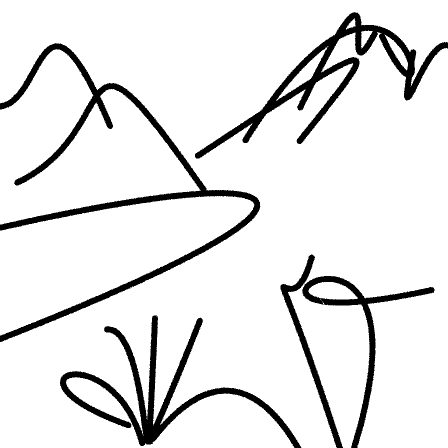} &
        \includegraphics[width=0.055\linewidth]{figs/ablations/our_results_new/l11_man_flowers_original_abs7.png} &
        \includegraphics[width=0.055\linewidth]{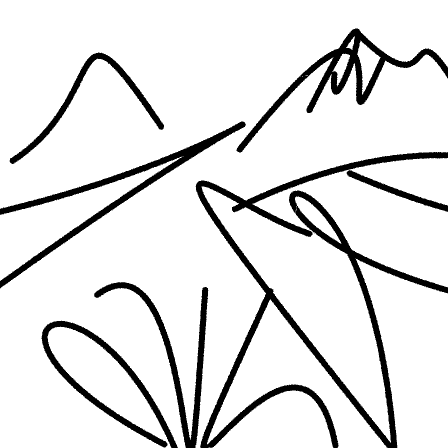} \\

        & \multicolumn{8}{l}{\hspace{0.05cm} Simplicity Axis $\longrightarrow$} & \multicolumn{8}{l}{\hspace{0.05cm} Simplicity Axis $\longrightarrow$}\\

        \cmidrule{2-9} 
        \cmidrule{10-17}

        &
        \multicolumn{8}{c}{\xrfill[0.5ex]{0.5pt}\; Same Factors for All Layers \; \xrfill[0.5ex]{0.5pt} \hspace{0.7cm}} &
        \multicolumn{8}{c}{\xrfill[0.5ex]{0.5pt}\; Different Factors for All Layers \; \xrfill[0.5ex]{0.5pt} \hspace{0.2cm}} \\

    \end{tabular}
    
    }
    \vspace{0.3cm}
    \caption{Ablation study on using a different set of factors for each fidelity level $k$ when defining our $\mathcal{L}_{ratio}$ loss. On the left, we should simplification results obtained when applying the same set of factors across all levels. On the right side, we should our simplification results obtained by adjusting the factors for each fidelity level.}
    \label{fig:same_ratios}
\end{figure*}

%% file: files/figures/supplementary/same_step_size_fk.tex
\begin{figure*}
    \centering
    \setlength{\belowcaptionskip}{-6pt}
    \setlength{\tabcolsep}{1.5pt}
    {\small
    \begin{tabular}{c c c c c c c c c @{\hspace{0.1cm}} | c c c c c c c c}
        
        \cmidrule{2-9} 
        \cmidrule{10-17}
        
        {\footnotesize\raisebox{0.05in}{\rotatebox{90}{Layer 2}}} &
        \includegraphics[width=0.055\linewidth]{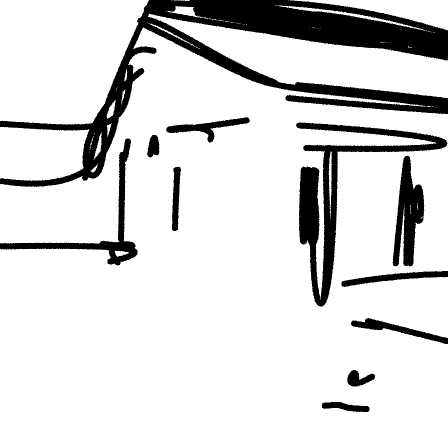} &
        \includegraphics[width=0.055\linewidth]{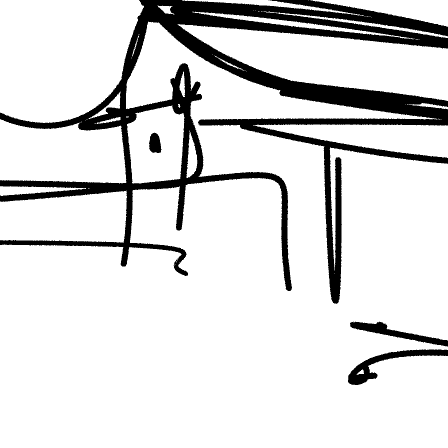} &
        \includegraphics[width=0.055\linewidth]{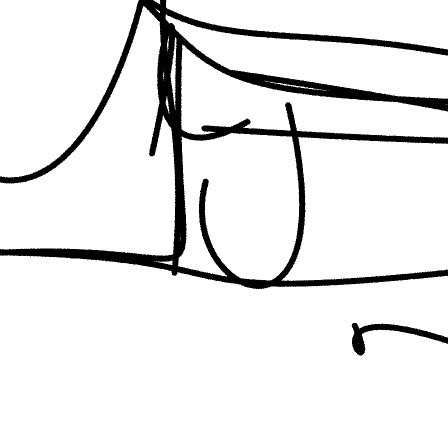} &
        \includegraphics[width=0.055\linewidth]{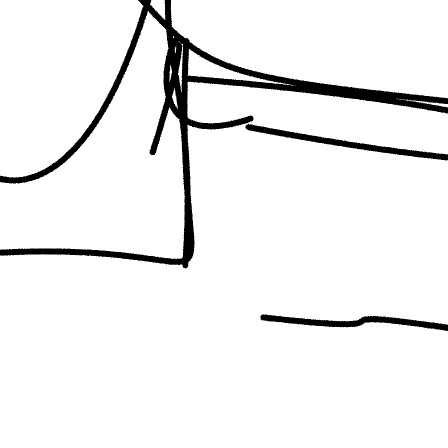} &
        \includegraphics[width=0.055\linewidth]{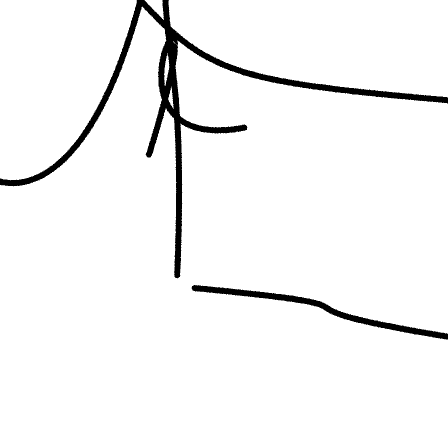} &
        \includegraphics[width=0.055\linewidth]{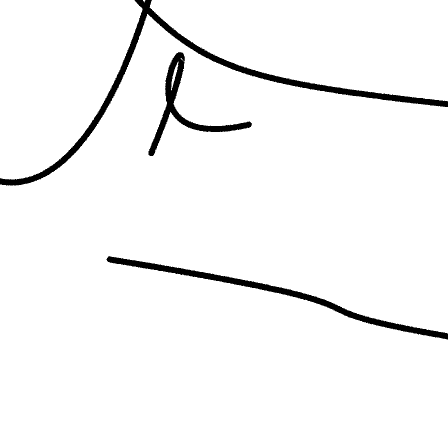} &
        \includegraphics[width=0.055\linewidth]{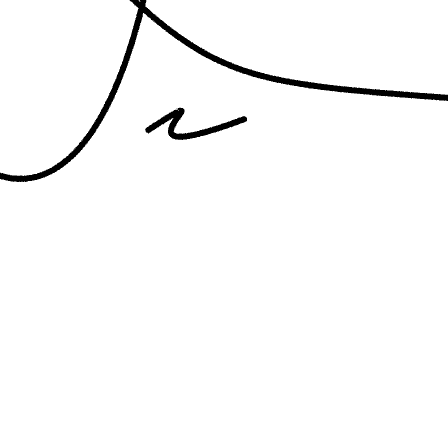} &
        \includegraphics[width=0.055\linewidth]{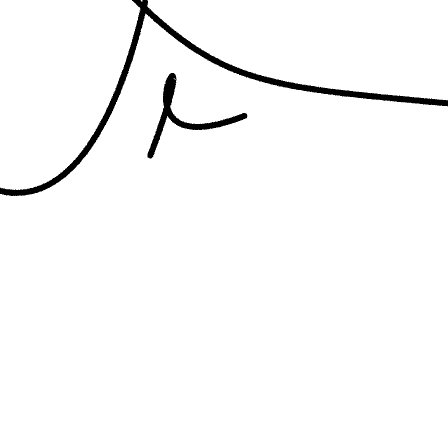} &
        \hspace{0.1cm}
        \includegraphics[width=0.055\linewidth]{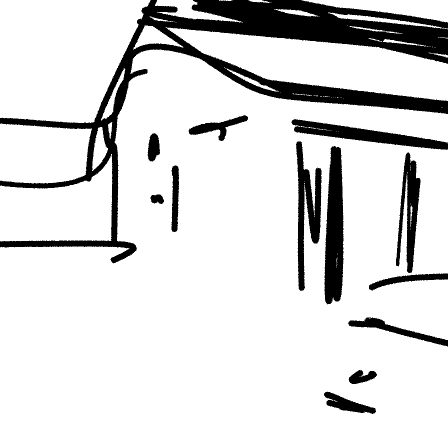} &
        \includegraphics[width=0.055\linewidth]{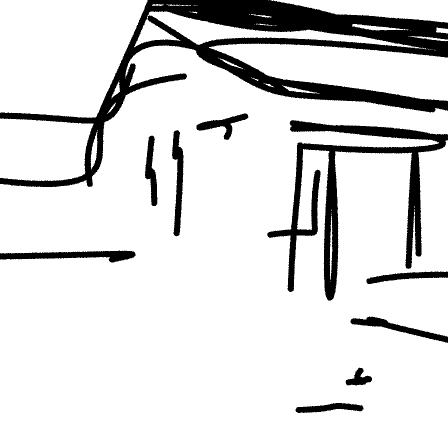} &
        \includegraphics[width=0.055\linewidth]{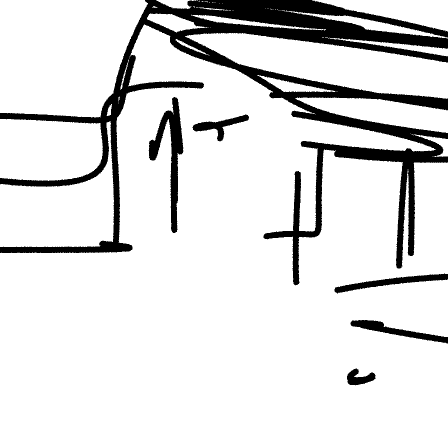} &
        \includegraphics[width=0.055\linewidth]{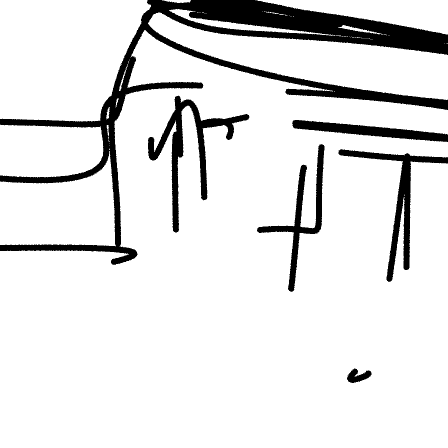} &
        \includegraphics[width=0.055\linewidth]{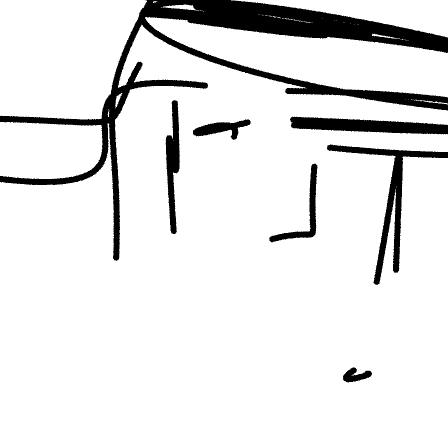} &
        \includegraphics[width=0.055\linewidth]{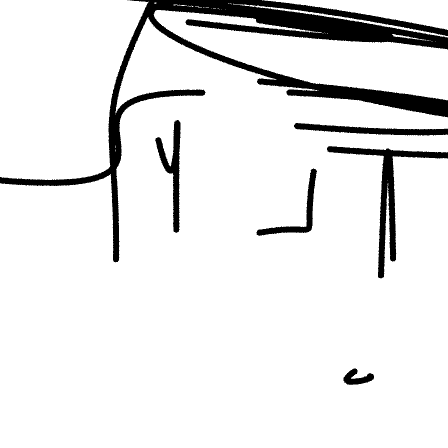} &
        \includegraphics[width=0.055\linewidth]{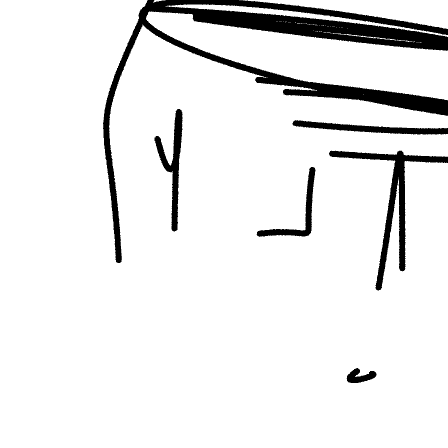} &
        \includegraphics[width=0.055\linewidth]{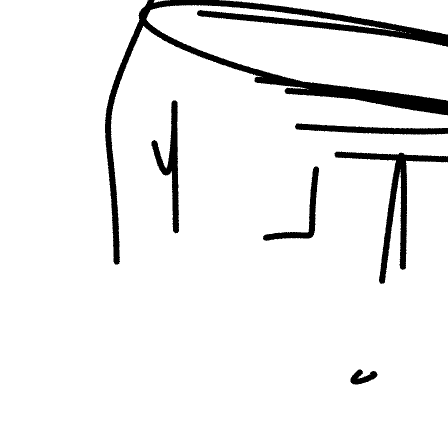} \\

        {\footnotesize\raisebox{0.05in}{\rotatebox{90}{Layer 7}}} &
        \includegraphics[width=0.055\linewidth]{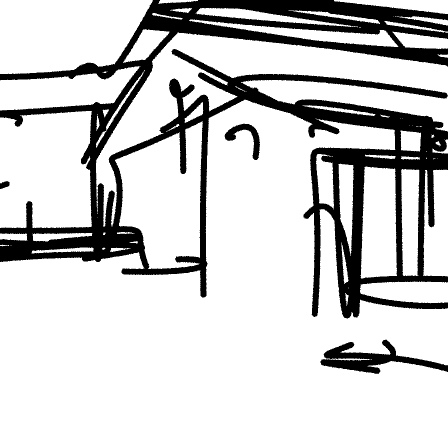} &
        \includegraphics[width=0.055\linewidth]{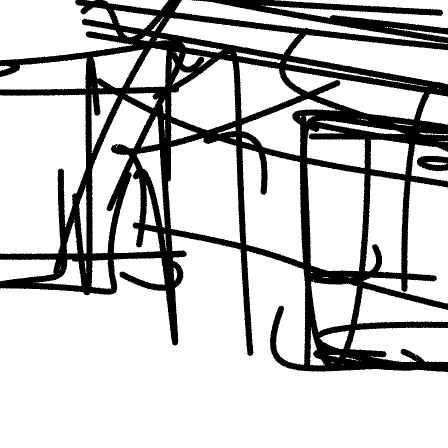} &
        \includegraphics[width=0.055\linewidth]{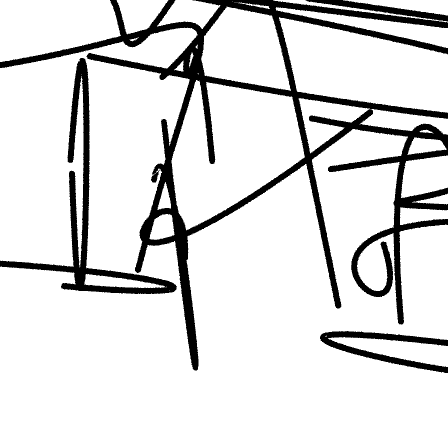} &
        \includegraphics[width=0.055\linewidth]{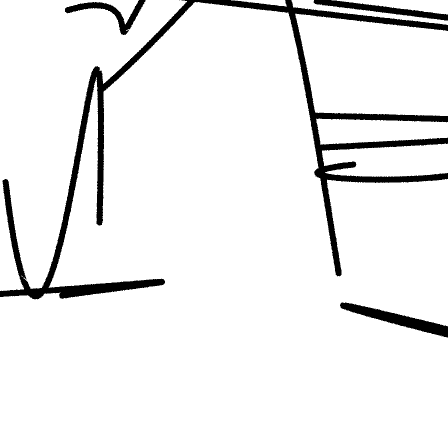} &
        \includegraphics[width=0.055\linewidth]{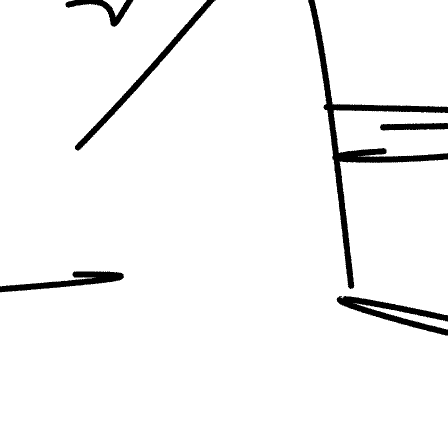} &
        \includegraphics[width=0.055\linewidth]{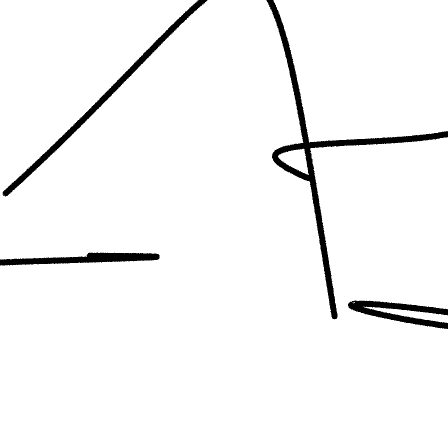} &
        \includegraphics[width=0.055\linewidth]{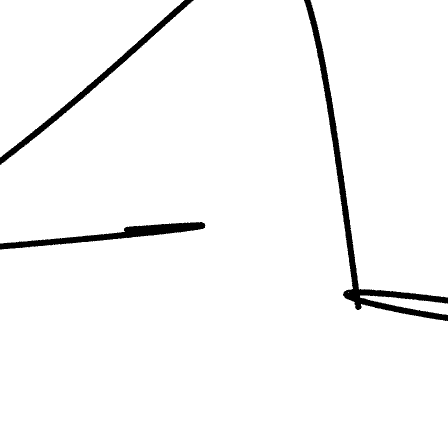} &
        \includegraphics[width=0.055\linewidth]{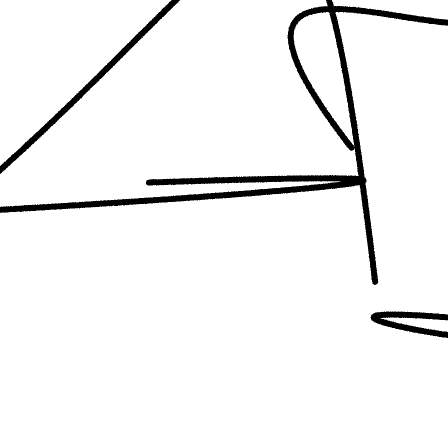} &
        \hspace{0.1cm}
        \includegraphics[width=0.055\linewidth]{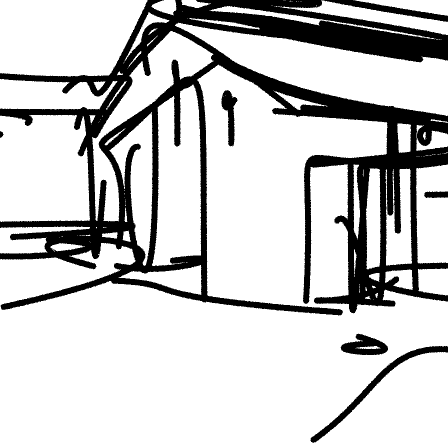} &
        \includegraphics[width=0.055\linewidth]{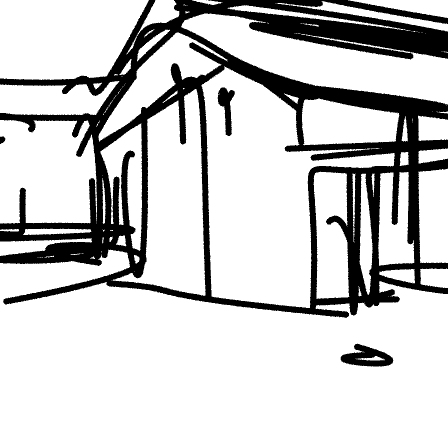} &
        \includegraphics[width=0.055\linewidth]{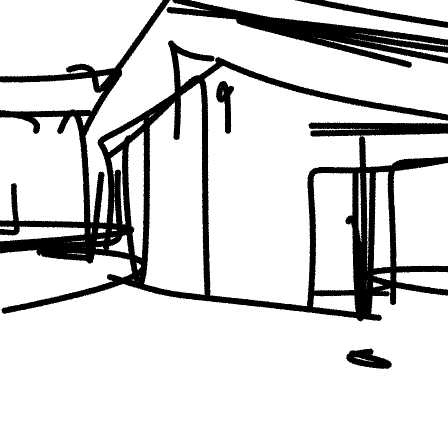} &
        \includegraphics[width=0.055\linewidth]{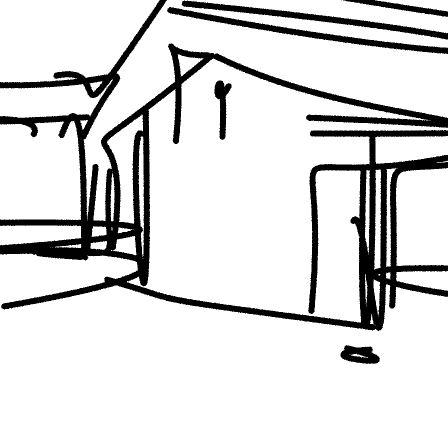} &
        \includegraphics[width=0.055\linewidth]{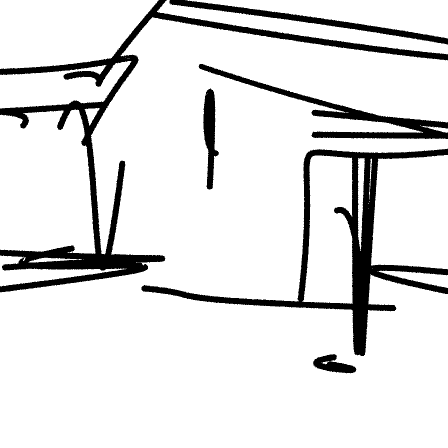} &
        \includegraphics[width=0.055\linewidth]{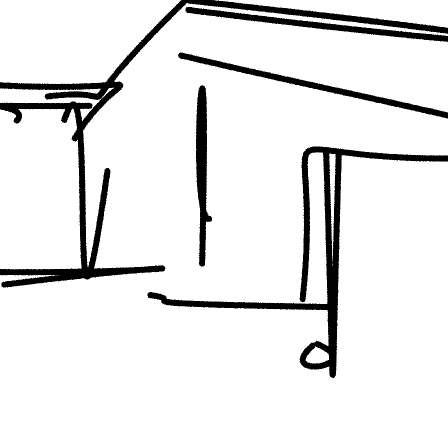} &
        \includegraphics[width=0.055\linewidth]{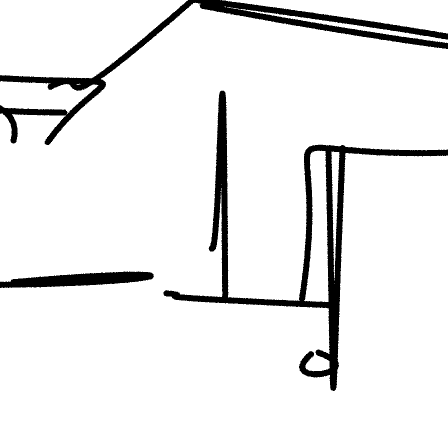} &
        \includegraphics[width=0.055\linewidth]{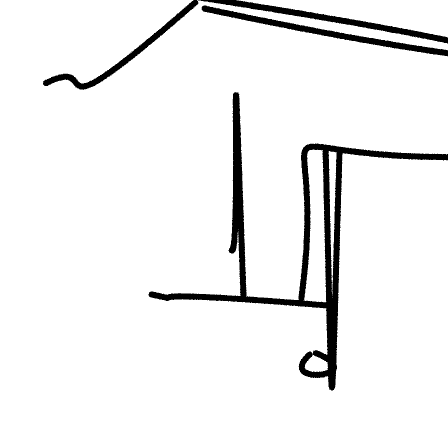} \\

       {\footnotesize\raisebox{0.05in}{\rotatebox{90}{Layer 8}}} &
        \includegraphics[width=0.055\linewidth]{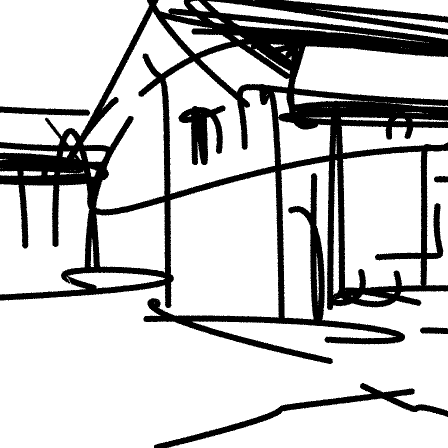} &
        \includegraphics[width=0.055\linewidth]{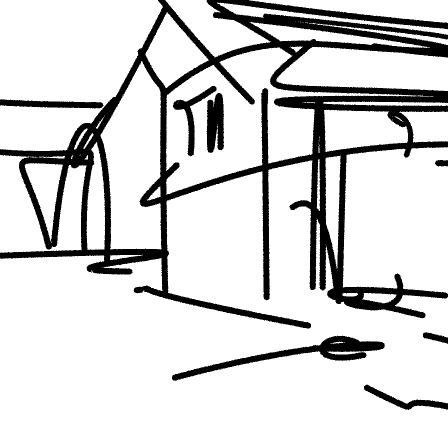} &
        \includegraphics[width=0.055\linewidth]{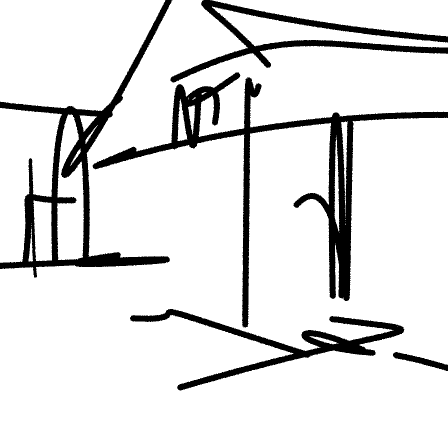} &
        \includegraphics[width=0.055\linewidth]{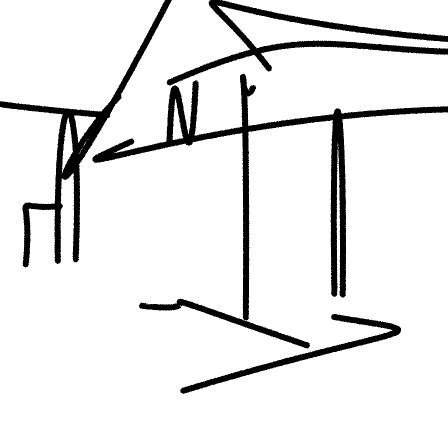} &
        \includegraphics[width=0.055\linewidth]{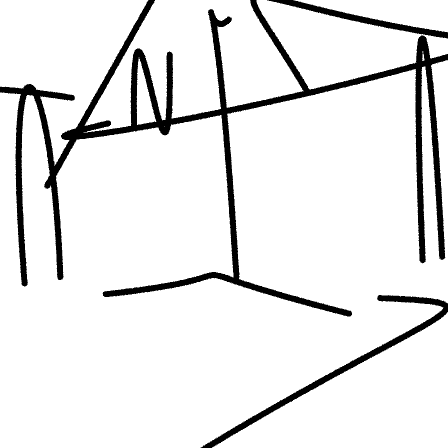} &
        \includegraphics[width=0.055\linewidth]{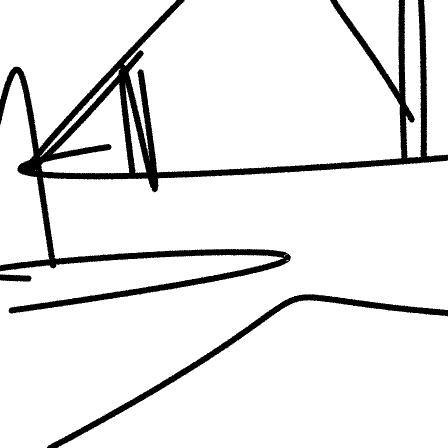} &
        \includegraphics[width=0.055\linewidth]{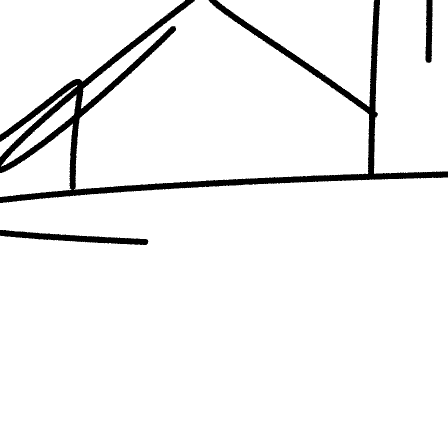} &
        \includegraphics[width=0.055\linewidth]{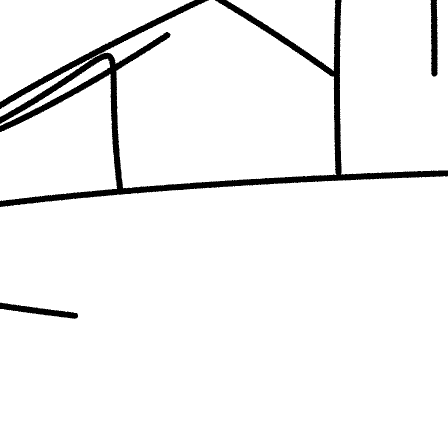} &
        \hspace{0.1cm}
        \includegraphics[width=0.055\linewidth]{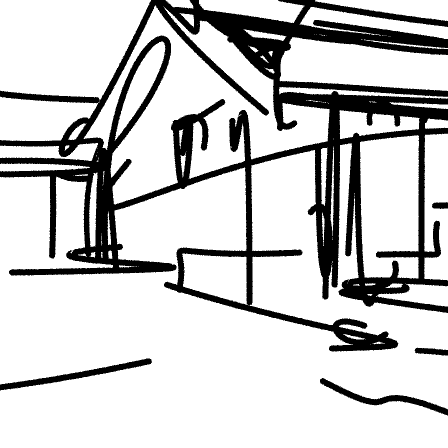} &
        \includegraphics[width=0.055\linewidth]{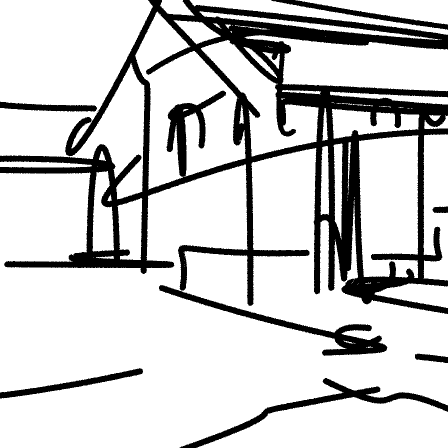} &
        \includegraphics[width=0.055\linewidth]{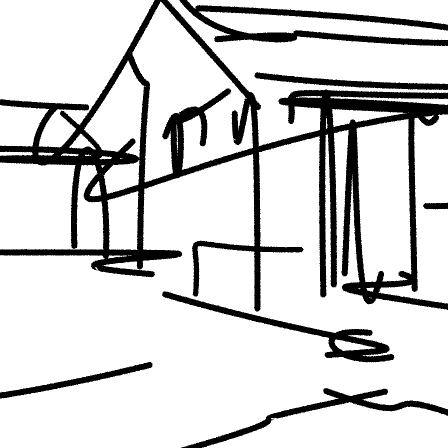} &
        \includegraphics[width=0.055\linewidth]{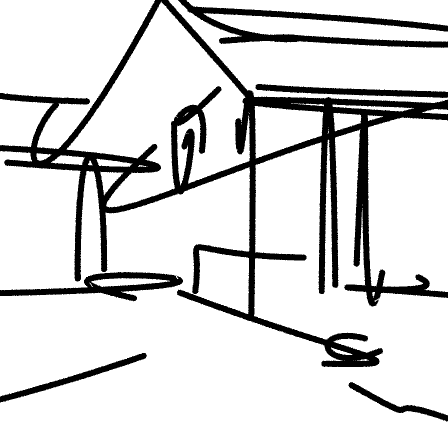} &
        \includegraphics[width=0.055\linewidth]{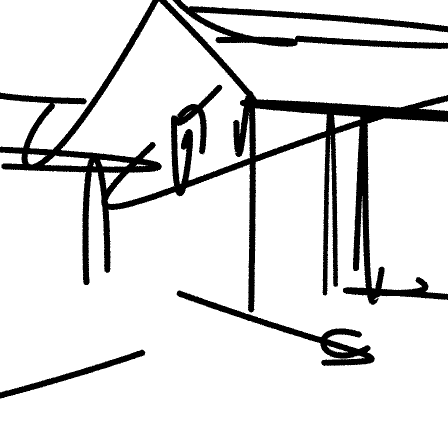} &
        \includegraphics[width=0.055\linewidth]{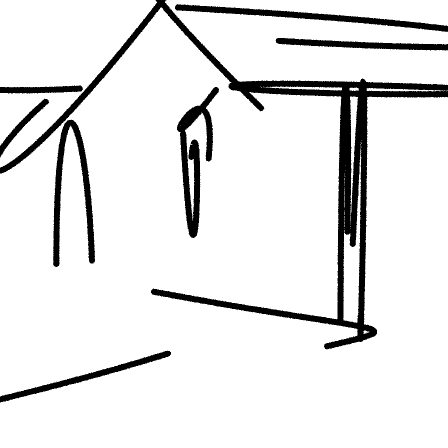} &
        \includegraphics[width=0.055\linewidth]{figs/ablations/our_results_new/l8_semi-complex_abs6.png} &
        \includegraphics[width=0.055\linewidth]{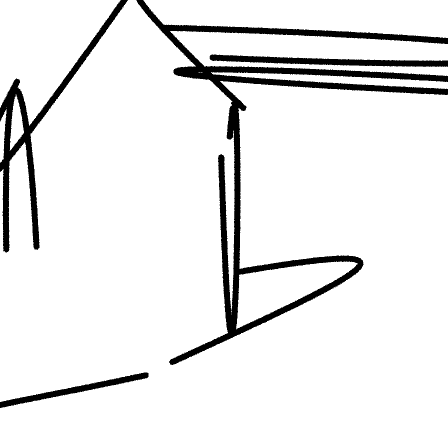} \\

        {\footnotesize\raisebox{0.05in}{\rotatebox{90}{Layer 11}}} &
        \includegraphics[width=0.055\linewidth]{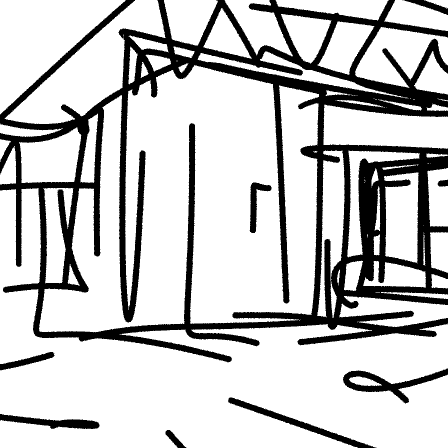} &
        \includegraphics[width=0.055\linewidth]{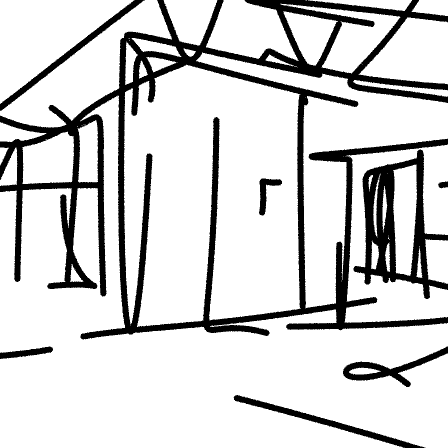} &
        \includegraphics[width=0.055\linewidth]{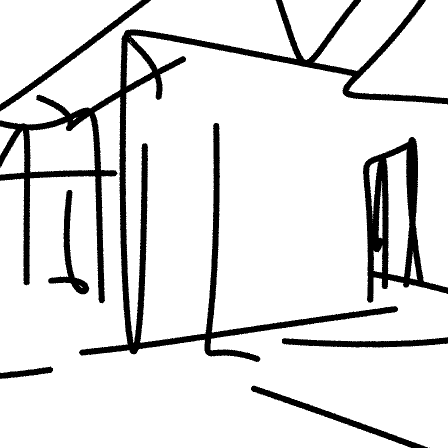} &
        \includegraphics[width=0.055\linewidth]{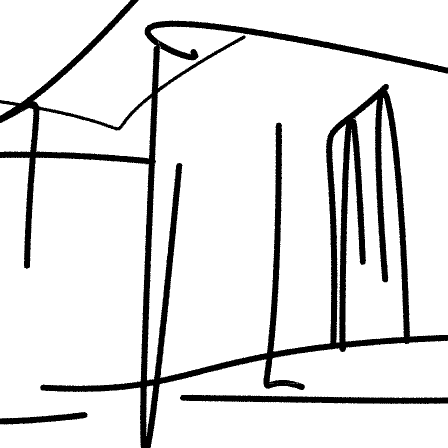} &
        \includegraphics[width=0.055\linewidth]{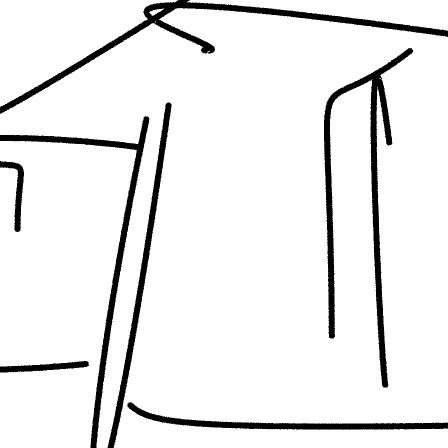} &
        \includegraphics[width=0.055\linewidth]{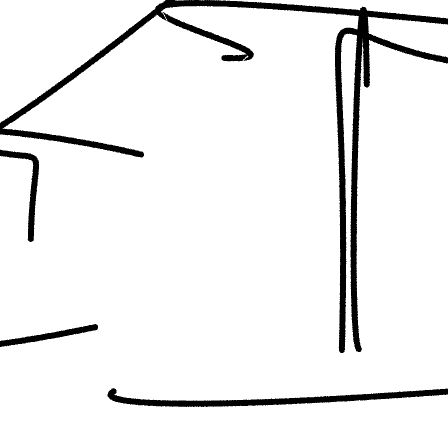} &
        \includegraphics[width=0.055\linewidth]{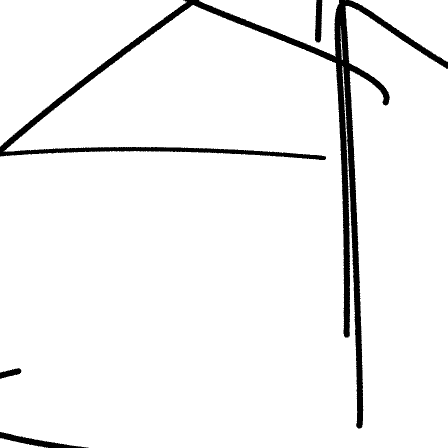} &
        \includegraphics[width=0.055\linewidth]{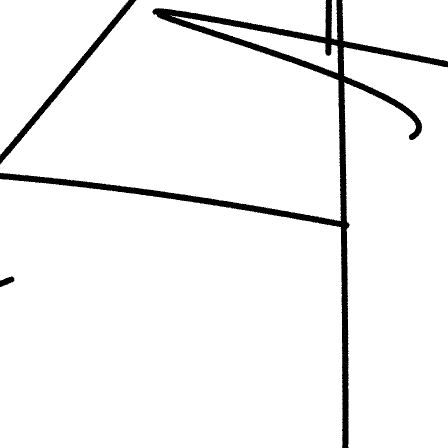} &
        \hspace{0.1cm}
        \includegraphics[width=0.055\linewidth]{figs/ablations/our_results_new/l11_semi-complex_original_abs1.png} &
        \includegraphics[width=0.055\linewidth]{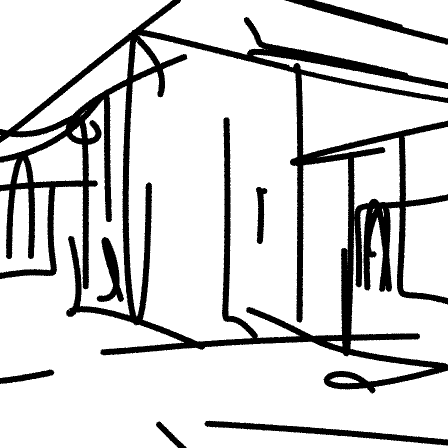} &
        \includegraphics[width=0.055\linewidth]{figs/ablations/our_results_new/l11_semi-complex_original_abs3.png} &
        \includegraphics[width=0.055\linewidth]{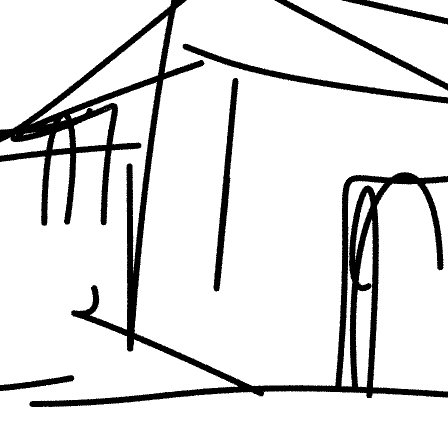} &
        \includegraphics[width=0.055\linewidth]{figs/ablations/our_results_new/l11_semi-complex_original_abs5.png} &
        \includegraphics[width=0.055\linewidth]{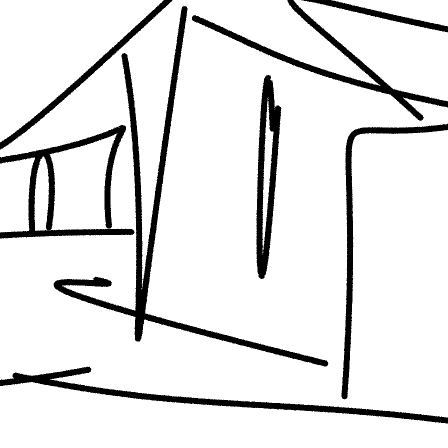} &
        \includegraphics[width=0.055\linewidth]{figs/ablations/our_results_new/l11_semi-complex_original_abs7.png} &
        \includegraphics[width=0.055\linewidth]{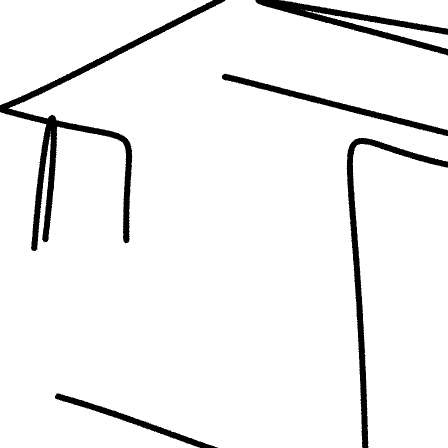} \\
        
        & \multicolumn{8}{l}{\hspace{0.05cm} Simplicity Axis $\longrightarrow$} & \multicolumn{8}{l}{\hspace{0.05cm} Simplicity Axis $\longrightarrow$}\\

        \cmidrule{2-9} 
        \cmidrule{10-17}

        &
        \multicolumn{8}{c}{\xrfill[0.5ex]{0.5pt}\; Same Step Size for All Layers \; \xrfill[0.5ex]{0.5pt}} &
        \multicolumn{8}{c}{\hspace{0.1cm} \xrfill[0.5ex]{0.5pt}\;  Different Step Size Per Layer  \; \xrfill[0.5ex]{0.5pt} \hspace{0.1cm}} \\

    \end{tabular}
    
    }
    \vspace{0.3cm}
    \caption{Ablation study on applying the same step size for sampling each function $f_k$ during simplification. On the left, we apply the same step size across all fidelity levels, resulting in non-smooth and non-uniform simplifications between the different fidelity levels. On the right, we show our simplification results obtained by adjusting the step size for each level. As shown, we achieve smoother simplifications that are more consistent between the four different levels.}
    \label{fig:same_step_size}

\end{figure*}

%% file: files/figures/supplementary/ablation_all_vit_layers.tex
\begin{figure*}[h]

    \centering
    \setlength{\tabcolsep}{1.5pt}
    {\small
    \begin{tabular}{c c c c c c c c c c c c}

        \includegraphics[width=0.0785\textwidth]{figs/inputs/man_camera.jpg} &
        \includegraphics[width=0.0785\textwidth]{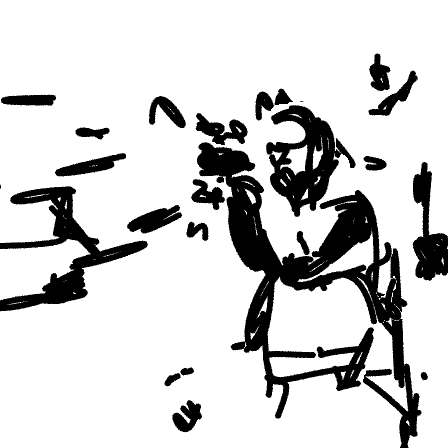} &
        \includegraphics[width=0.0785\textwidth]{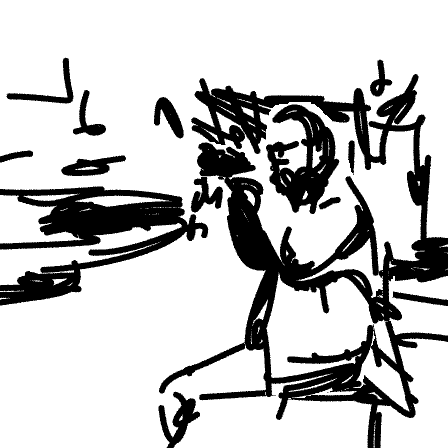} &
        \includegraphics[width=0.0785\textwidth]{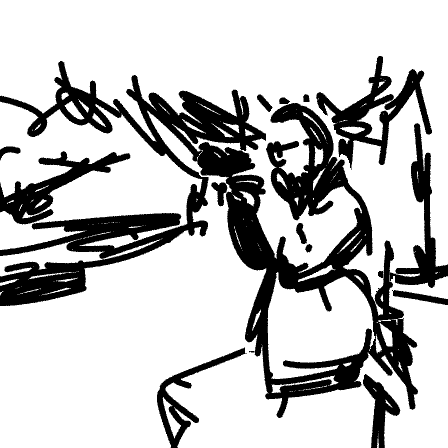} &
        \includegraphics[width=0.0785\textwidth]{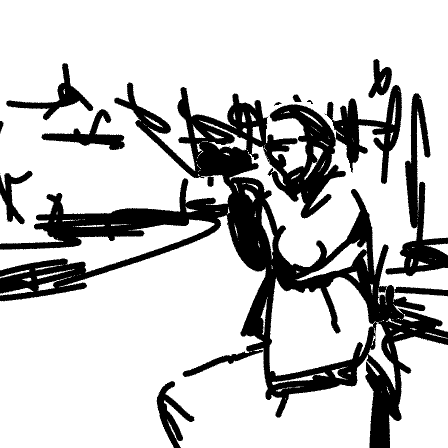} &
        \includegraphics[width=0.0785\textwidth]{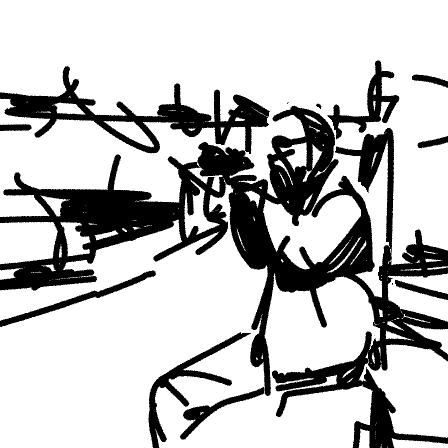} &
        \includegraphics[width=0.0785\textwidth]{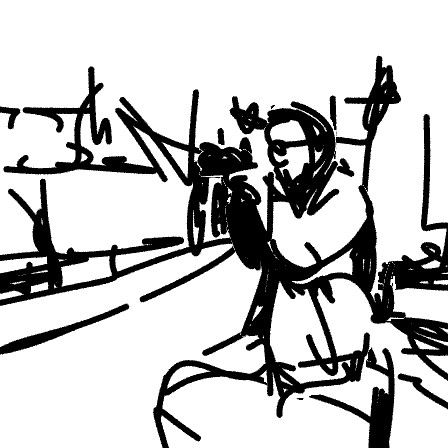} &
        \includegraphics[width=0.0785\textwidth]{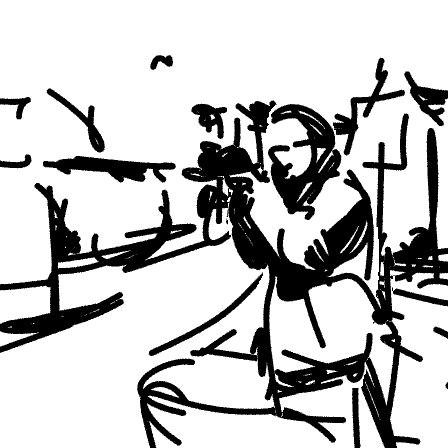} &
        \includegraphics[width=0.0785\textwidth]{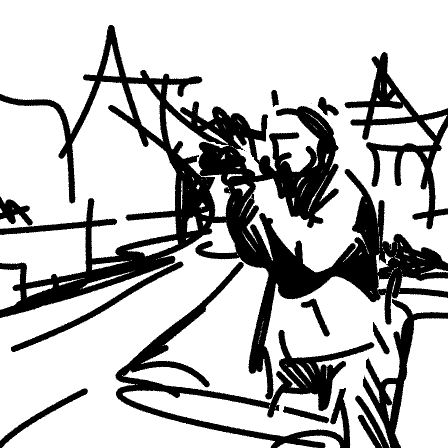} &
        \includegraphics[width=0.0785\textwidth]{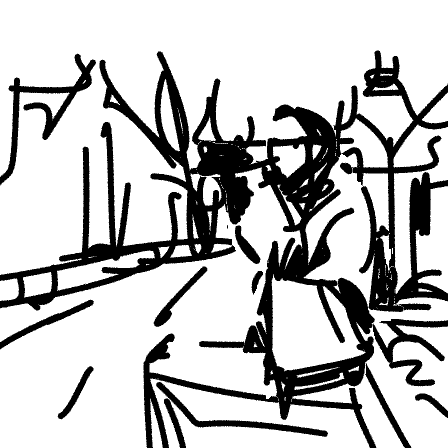} &
        \includegraphics[width=0.0785\textwidth]{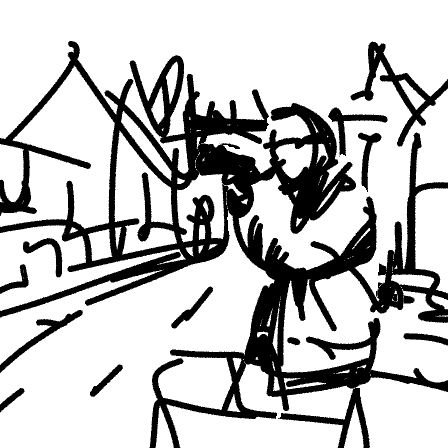} &
        \includegraphics[width=0.0785\textwidth]{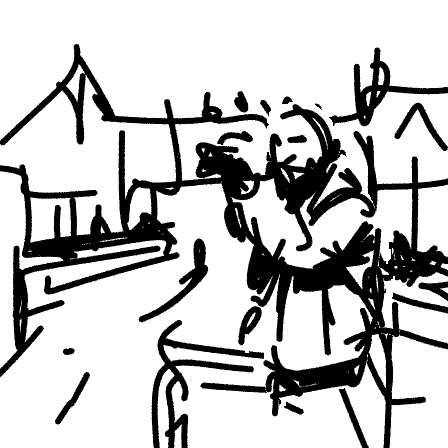} \\

        \includegraphics[width=0.0785\textwidth]{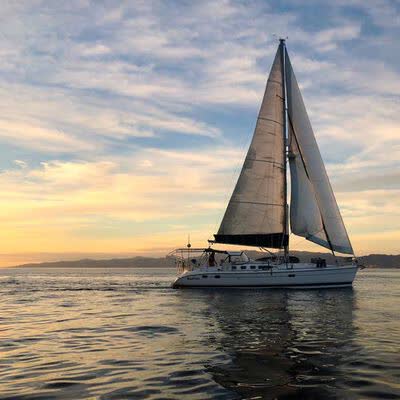} &
        \includegraphics[width=0.0785\textwidth]{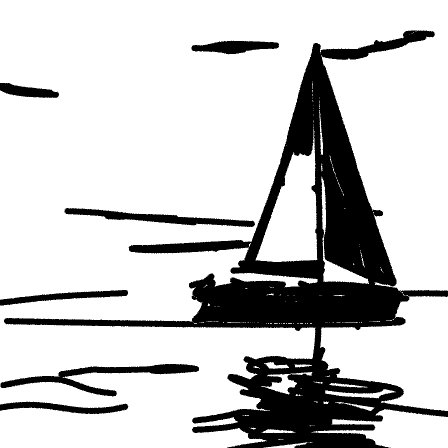} &
        \includegraphics[width=0.0785\textwidth]{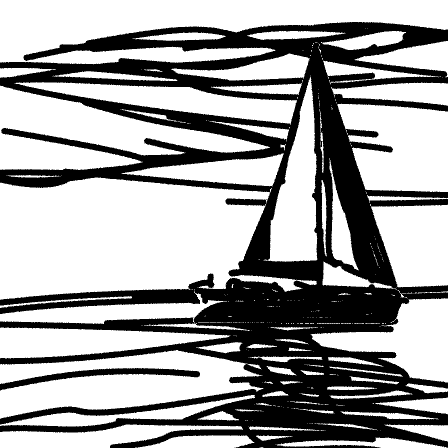} &
        \includegraphics[width=0.0785\textwidth]{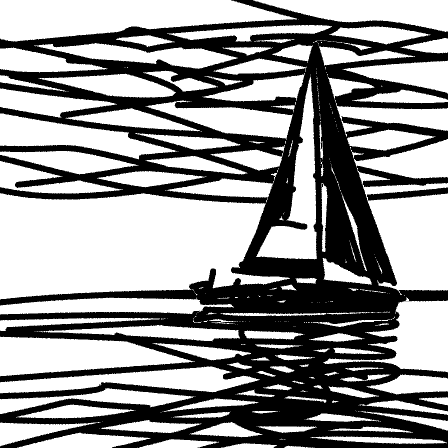} &
        \includegraphics[width=0.0785\textwidth]{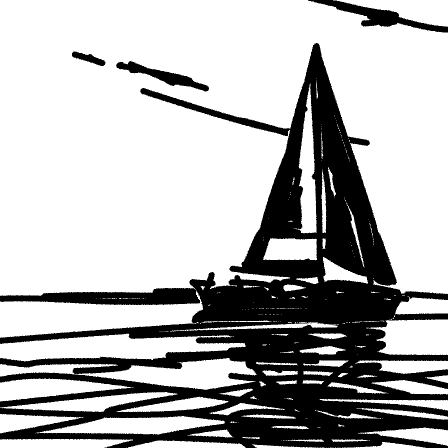} &
        \includegraphics[width=0.0785\textwidth]{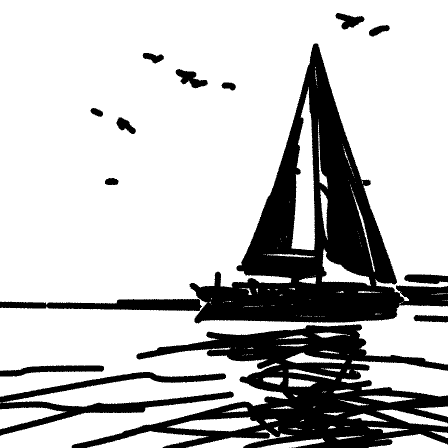} &
        \includegraphics[width=0.0785\textwidth]{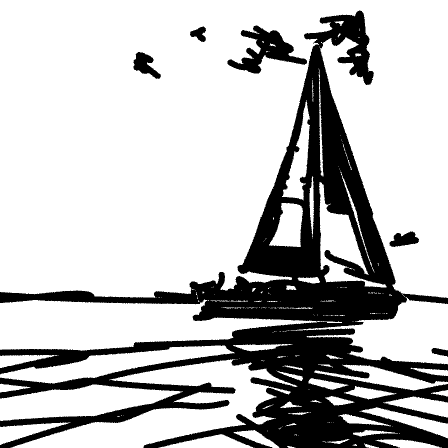} &
        \includegraphics[width=0.0785\textwidth]{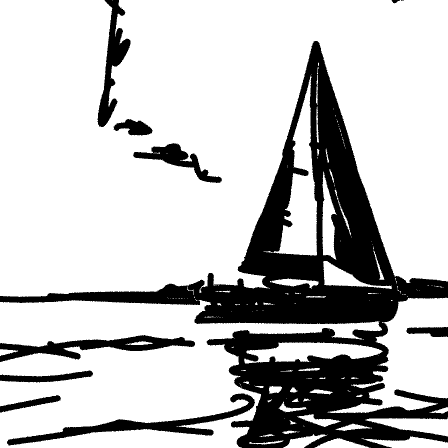} &
        \includegraphics[width=0.0785\textwidth]{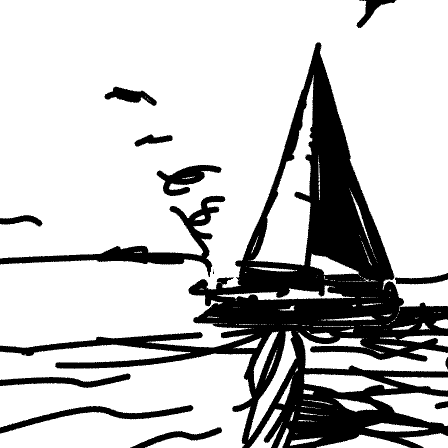} &
        \includegraphics[width=0.0785\textwidth]{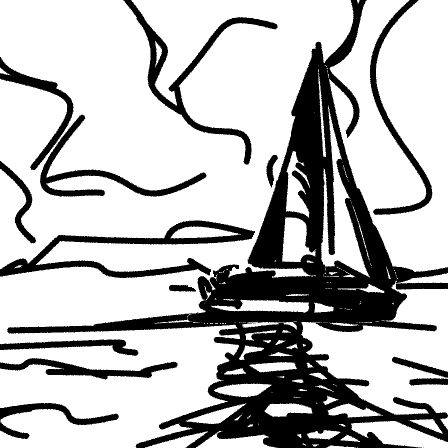} &
        \includegraphics[width=0.0785\textwidth]{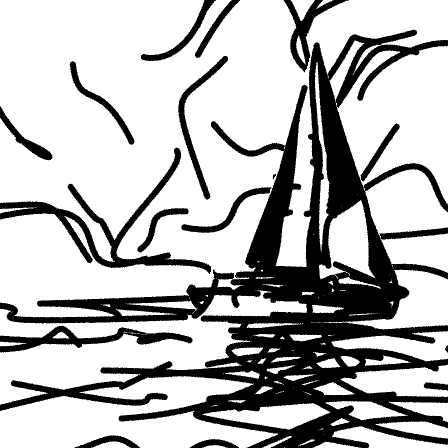} &
        \includegraphics[width=0.0785\textwidth]{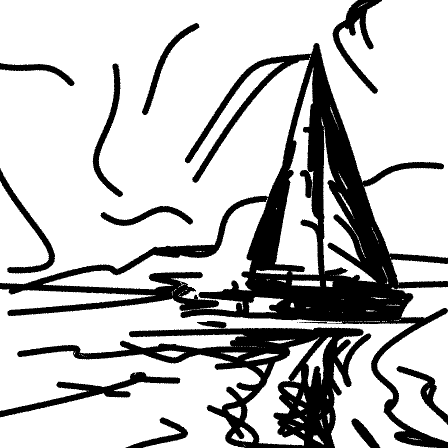} \\

        \includegraphics[width=0.0785\textwidth]{figs/inputs/panda.jpg} &
        \includegraphics[width=0.0785\textwidth]{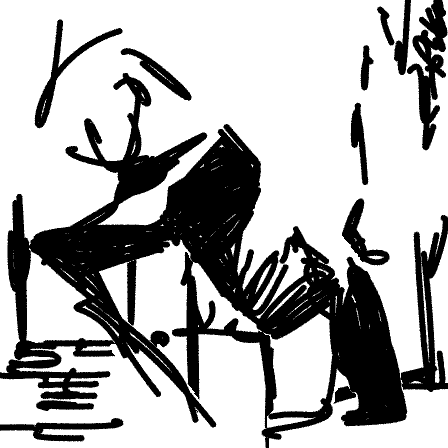} &
        \includegraphics[width=0.0785\textwidth]{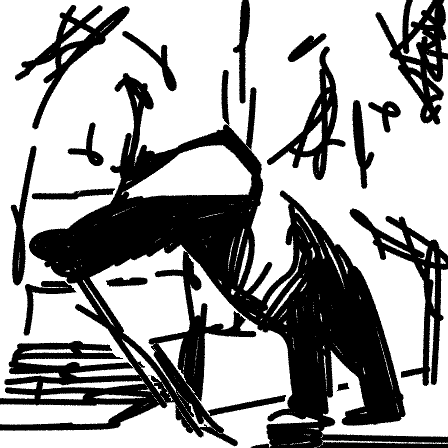} &
        \includegraphics[width=0.0785\textwidth]{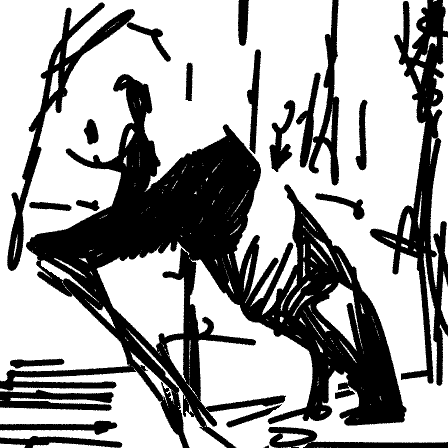} &
        \includegraphics[width=0.0785\textwidth]{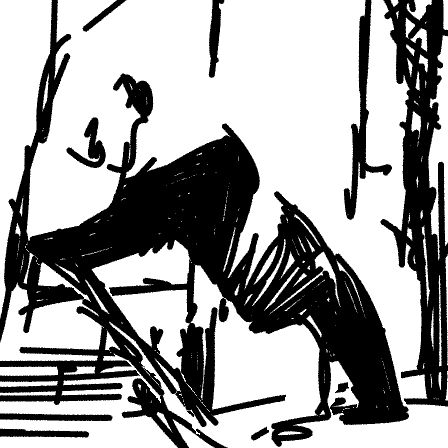} &
        \includegraphics[width=0.0785\textwidth]{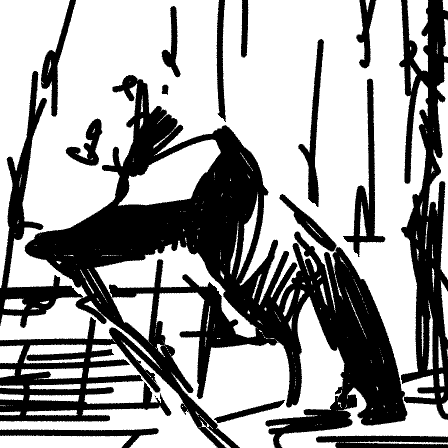} &
        \includegraphics[width=0.0785\textwidth]{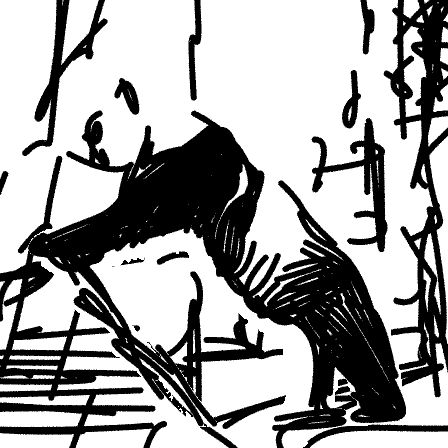} &
        \includegraphics[width=0.0785\textwidth]{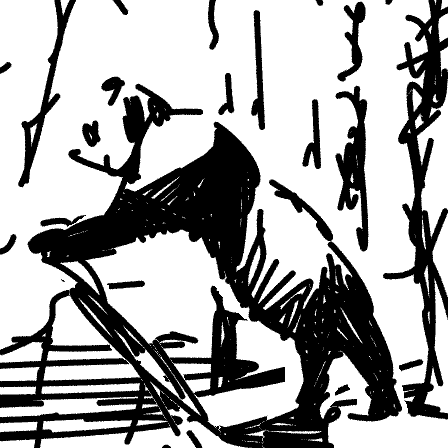} &
        \includegraphics[width=0.0785\textwidth]{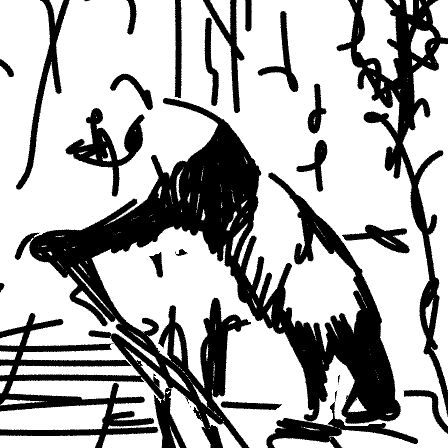} &
        \includegraphics[width=0.0785\textwidth]{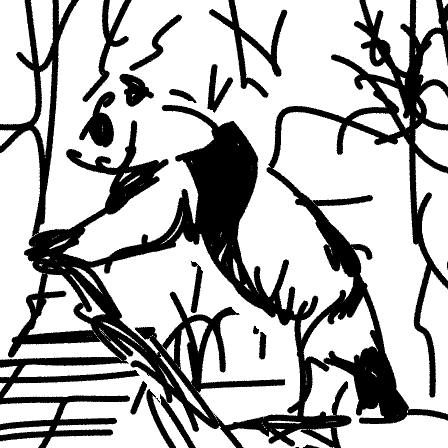} &
        \includegraphics[width=0.0785\textwidth]{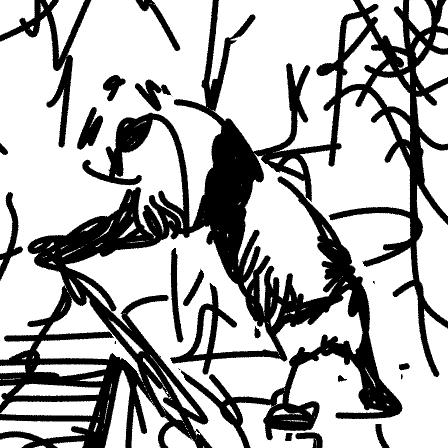} &
        \includegraphics[width=0.0785\textwidth]{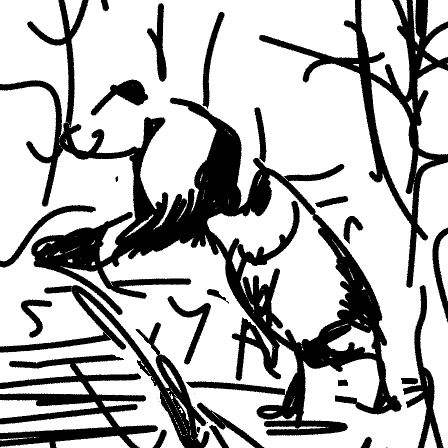} \\

        \includegraphics[width=0.0785\textwidth]{figs/inputs/woman_city.jpg} &
        \includegraphics[width=0.0785\textwidth]{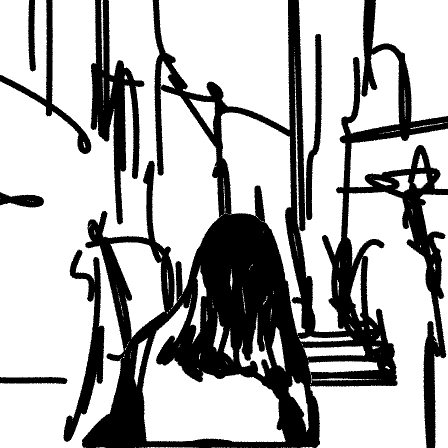} &
        \includegraphics[width=0.0785\textwidth]{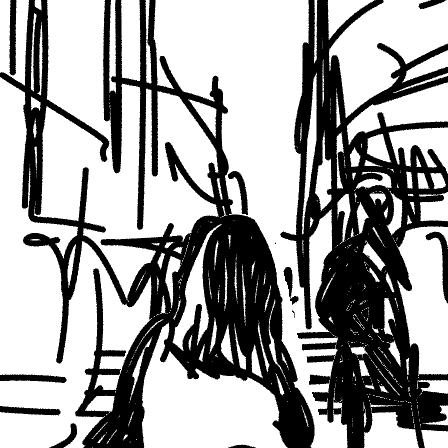} &
        \includegraphics[width=0.0785\textwidth]{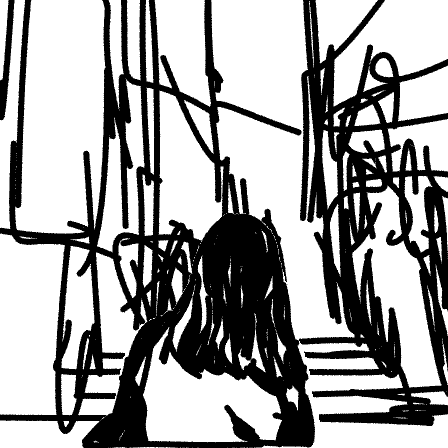} &
        \includegraphics[width=0.0785\textwidth]{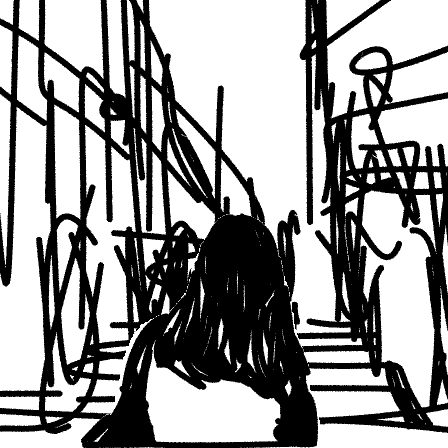} &
        \includegraphics[width=0.0785\textwidth]{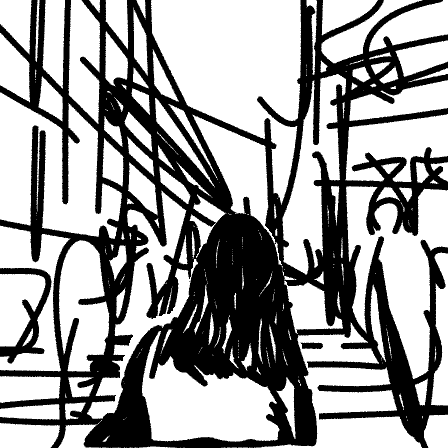} &
        \includegraphics[width=0.0785\textwidth]{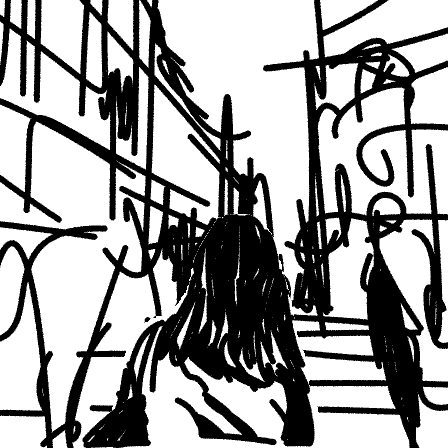} &
        \includegraphics[width=0.0785\textwidth]{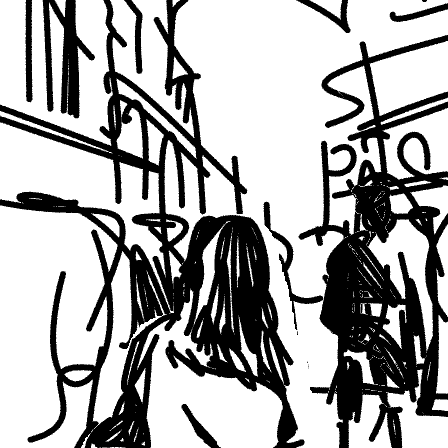} &
        \includegraphics[width=0.0785\textwidth]{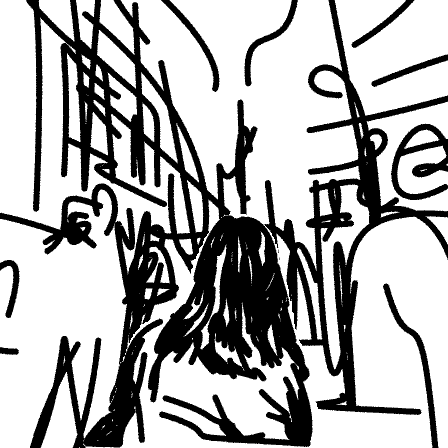} &
        \includegraphics[width=0.0785\textwidth]{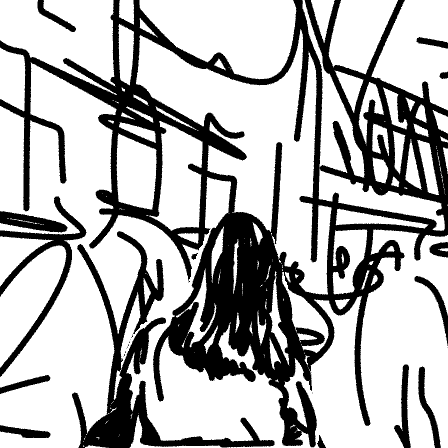} &
        \includegraphics[width=0.0785\textwidth]{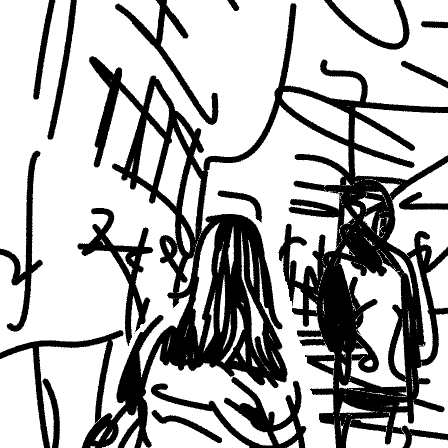} &
        \includegraphics[width=0.0785\textwidth]{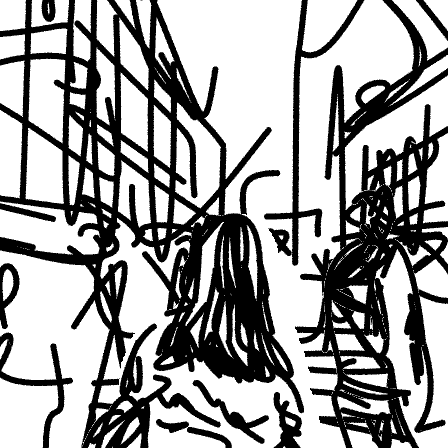} \\

        \includegraphics[width=0.0785\textwidth]{figs/inputs/womanhome.jpg} &
        \includegraphics[width=0.0785\textwidth]{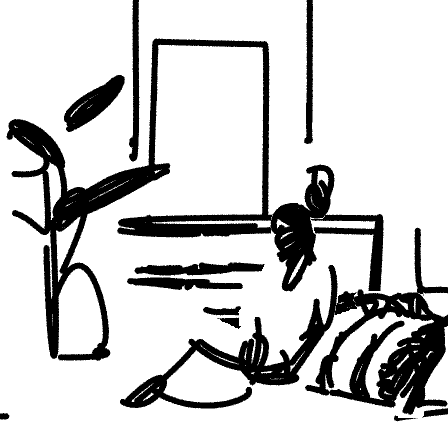} &
        \includegraphics[width=0.0785\textwidth]{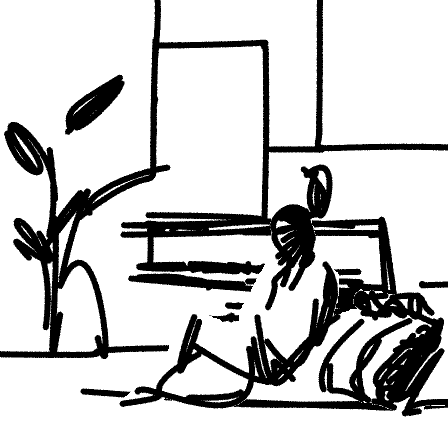} &
        \includegraphics[width=0.0785\textwidth]{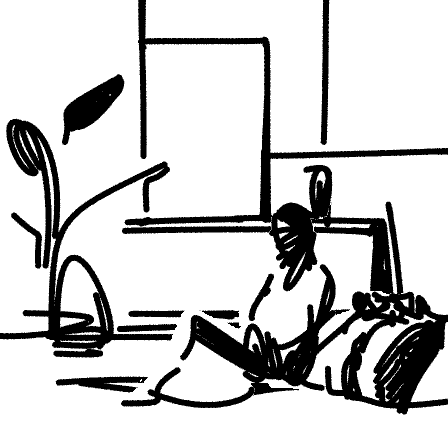} &
        \includegraphics[width=0.0785\textwidth]{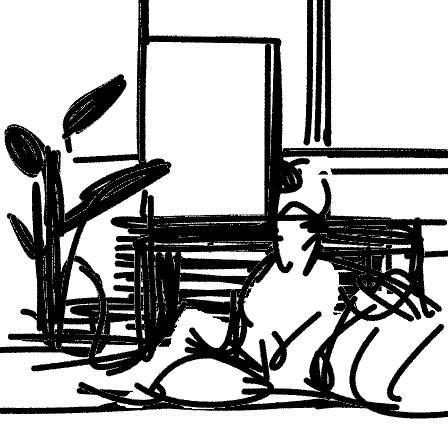} &
        \includegraphics[width=0.0785\textwidth]{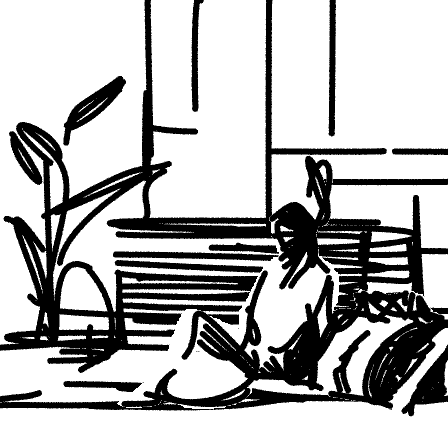} &
        \includegraphics[width=0.0785\textwidth]{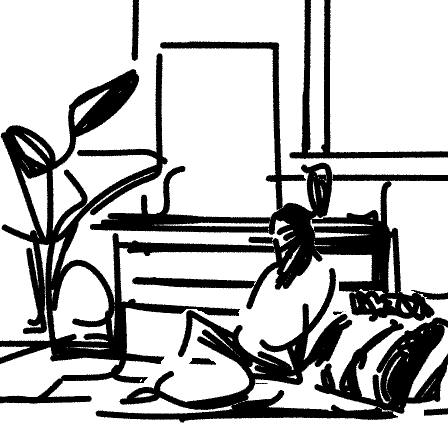} &
        \includegraphics[width=0.0785\textwidth]{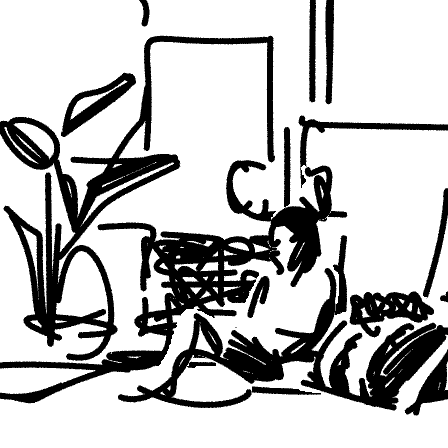} &
        \includegraphics[width=0.0785\textwidth]{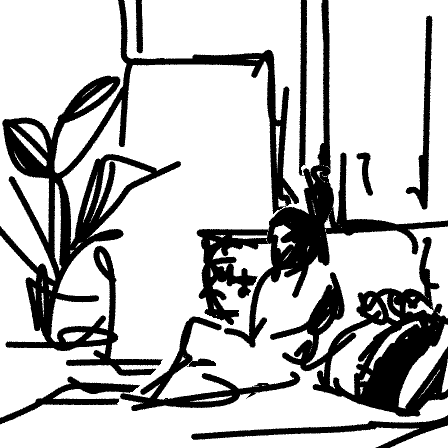} &
        \includegraphics[width=0.0785\textwidth]{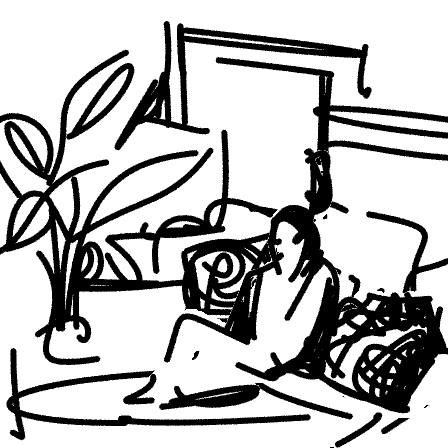} &
        \includegraphics[width=0.0785\textwidth]{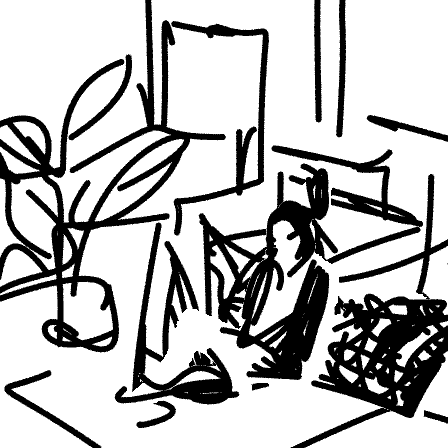} &
        \includegraphics[width=0.0785\textwidth]{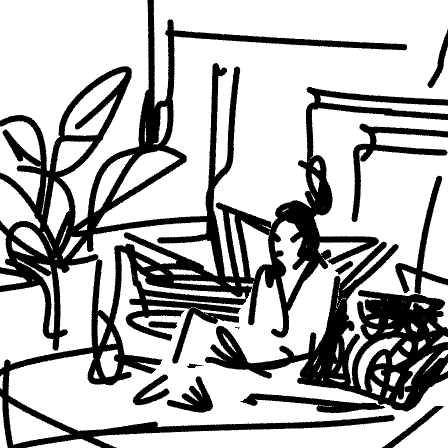} \\

        \includegraphics[width=0.0785\textwidth]{figs/inputs/house3.jpg} &
        \includegraphics[width=0.0785\textwidth]{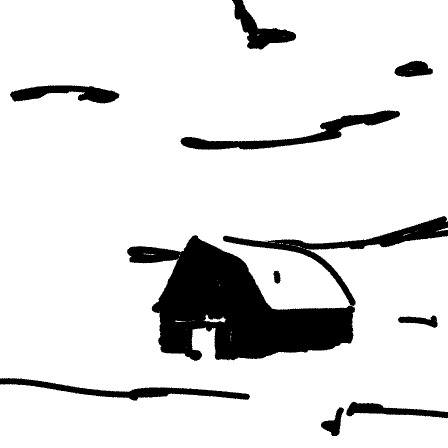} &
        \includegraphics[width=0.0785\textwidth]{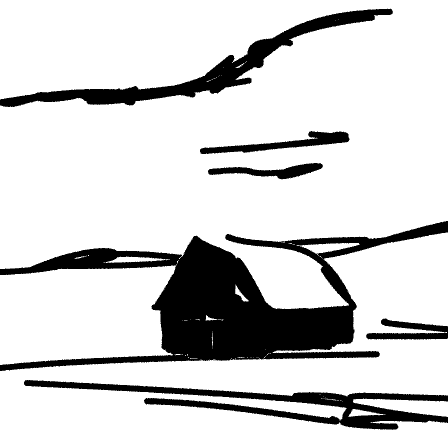} &
        \includegraphics[width=0.0785\textwidth]{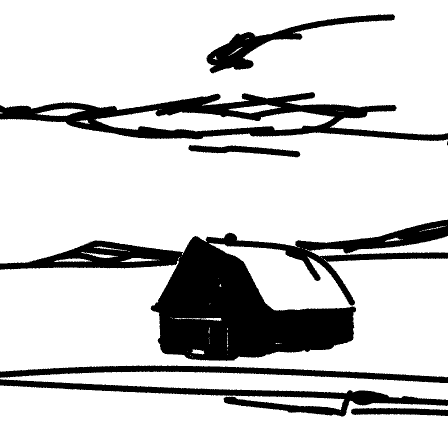} &
        \includegraphics[width=0.0785\textwidth]{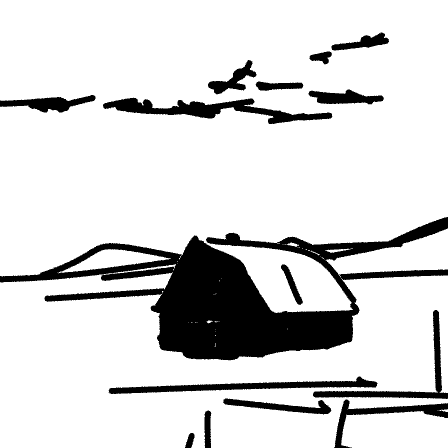} &
        \includegraphics[width=0.0785\textwidth]{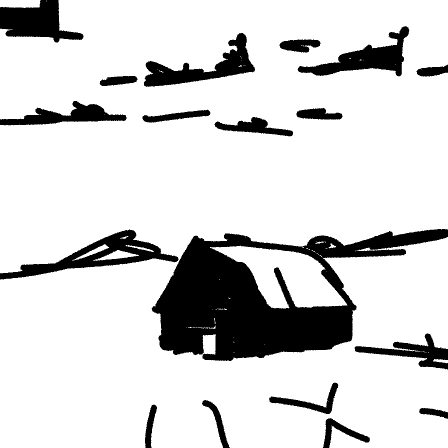} &
        \includegraphics[width=0.0785\textwidth]{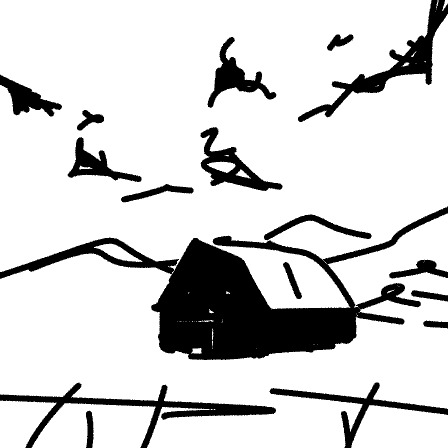} &
        \includegraphics[width=0.0785\textwidth]{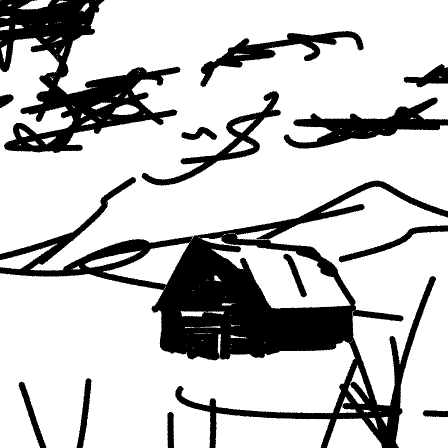} &
        \includegraphics[width=0.0785\textwidth]{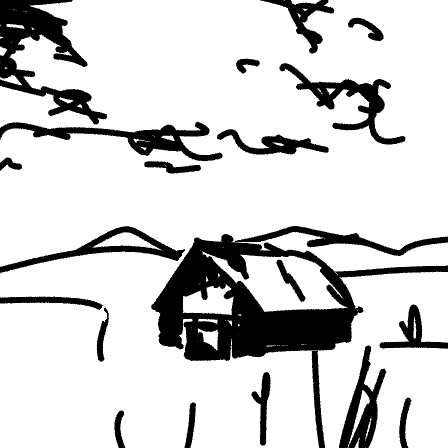} &
        \includegraphics[width=0.0785\textwidth]{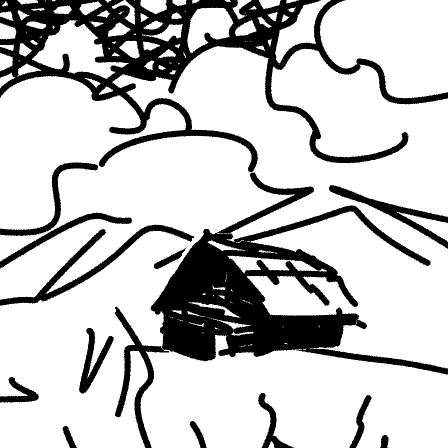} &
        \includegraphics[width=0.0785\textwidth]{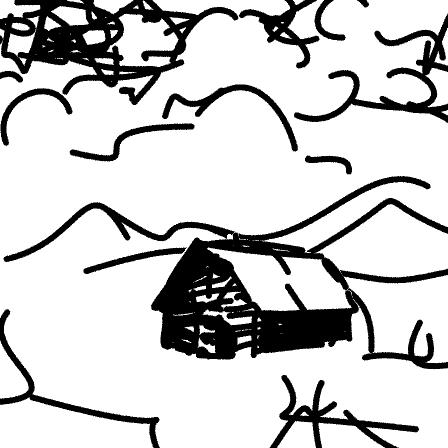} &
        \includegraphics[width=0.0785\textwidth]{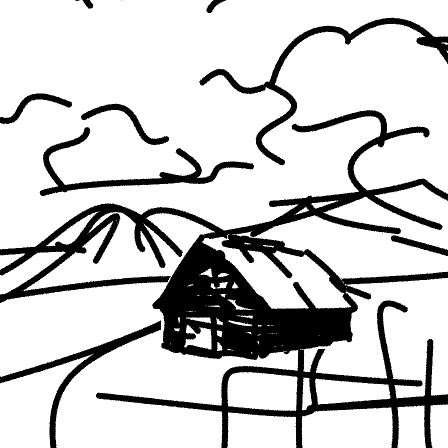} \\

        Input & 
        Layer 1 &
        Layer 2 &
        Layer 3 &
        Layer 4 &
        Layer 5 &
        Layer 6 &
        Layer 7 &
        Layer 8 &
        Layer 9 &
        Layer 10 &
        Layer 11
                
    \end{tabular}
    
    }
    \vspace{0.3cm}
    \caption{Ablation study on using different ViT layers for computing $\mathcal{L}_{CLIP}$ for generating sketches at different levels of fidelity.}
    \label{fig:ablation_all_vit_layers}
\end{figure*}

%% file: files/figures/supplementary/ablation_with_l4.tex
\begin{figure}[h]

    \centering
    \setlength{\tabcolsep}{1.5pt}
    {\small
    \begin{tabular}{c c c c}

        \raisebox{0.225in}{\rotatebox{90}{Input}} &
        \includegraphics[width=0.11\textwidth]{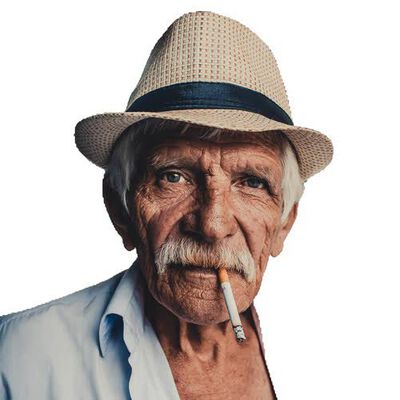} &
        \includegraphics[width=0.11\textwidth]{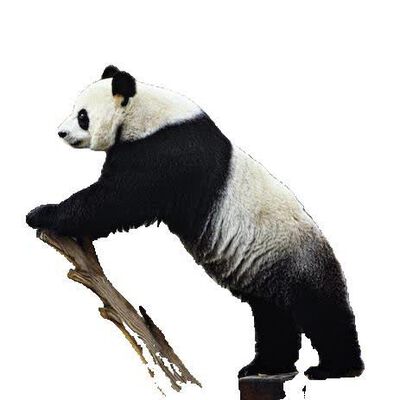} & 
        \includegraphics[width=0.11\textwidth]{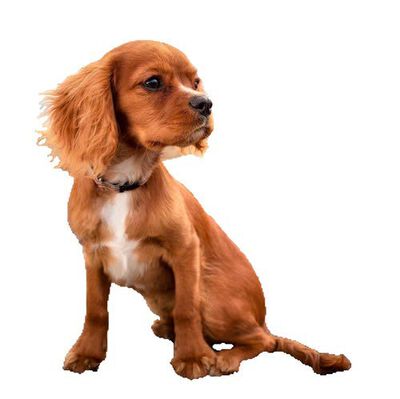} \\

        \raisebox{0.225in}{\rotatebox{90}{{w/o $\ell_4$}}} &
        \includegraphics[width=0.11\textwidth]{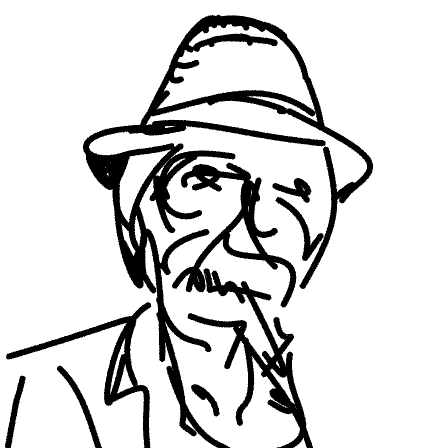} &
        \includegraphics[width=0.11\textwidth]{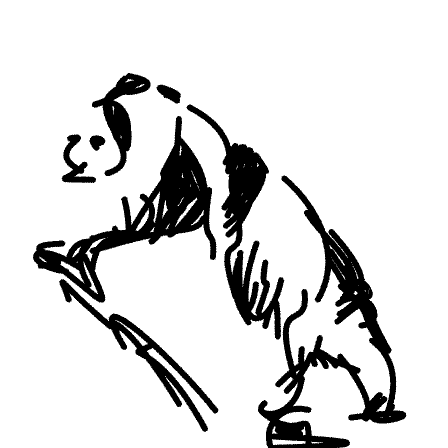} &
        \includegraphics[width=0.11\textwidth]{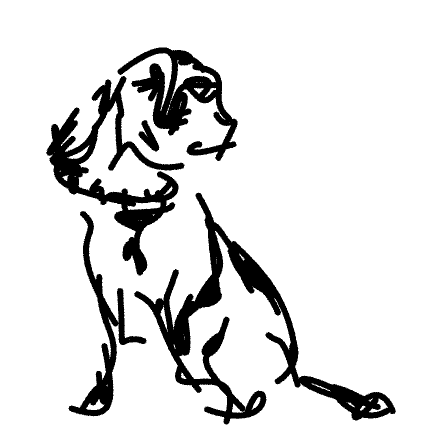} \\

        \raisebox{0.225in}{\rotatebox{90}{w/ $\ell_4$}} &
        \includegraphics[width=0.11\textwidth]{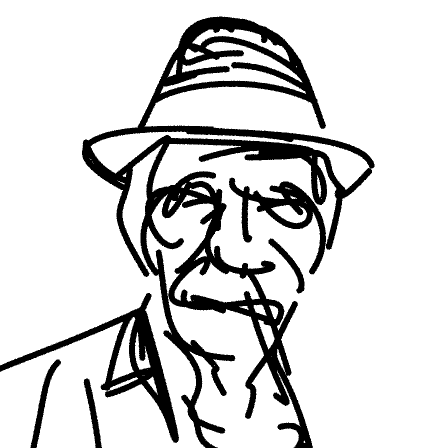} &
        \includegraphics[width=0.11\textwidth]{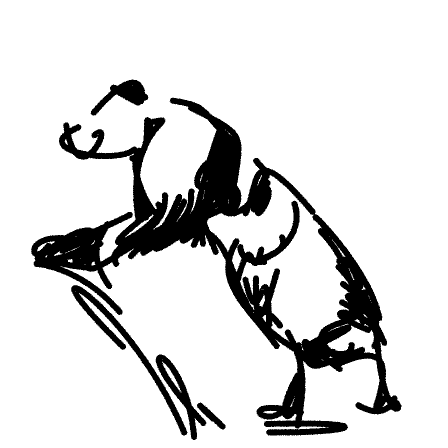} &
        \includegraphics[width=0.11\textwidth]{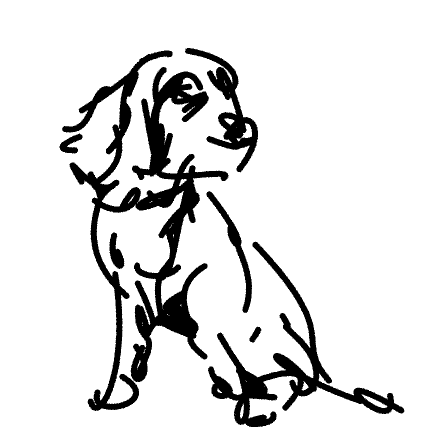} \\
        
    \end{tabular}
    
    }
    \vspace{0.3cm}
    \caption{Ablation results for object sketching when additionally using layer $\ell_4$ when computing $\mathcal{L}_{CLIP}$ for object sketching.}
    \label{fig:ablation_l4}
\end{figure}

%% file: files/figures/supplementary/our_matrices/our_matrices_1.tex
\begin{figure*}
    \centering
    
    \begin{tabular}{c c c c c c c c}

    \includegraphics[width=0.098\textwidth,height=0.098\textwidth]{figs/inputs/arc_de_triomphe.jpg} & & & &
    \hspace{0.5cm}
    \includegraphics[width=0.098\textwidth,height=0.098\textwidth]{figs/inputs/man_camera.jpg} & & & \\
    \frame{\includegraphics[width=0.098\textwidth,height=0.098\textwidth]{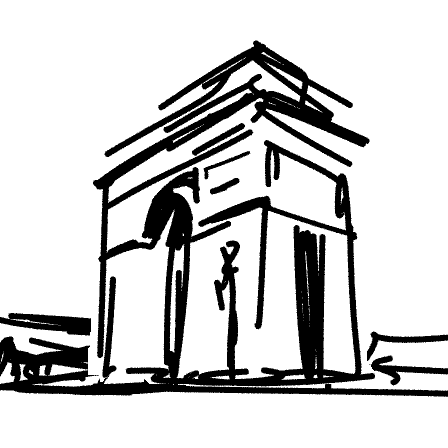}} &
    \frame{\includegraphics[width=0.098\textwidth,height=0.098\textwidth]{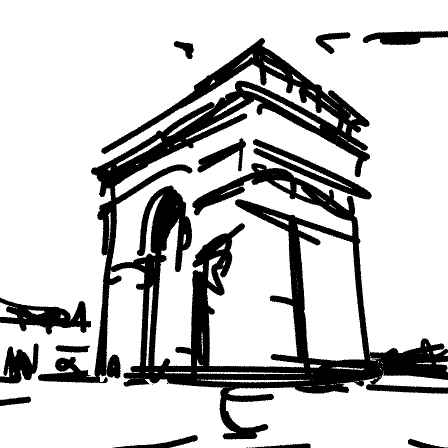}} &
    \frame{\includegraphics[width=0.098\textwidth,height=0.098\textwidth]{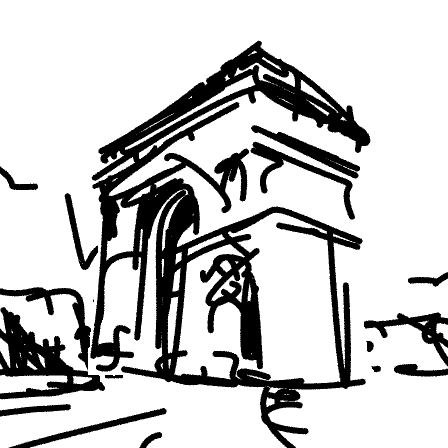}} &
    \frame{\includegraphics[width=0.098\textwidth,height=0.098\textwidth]{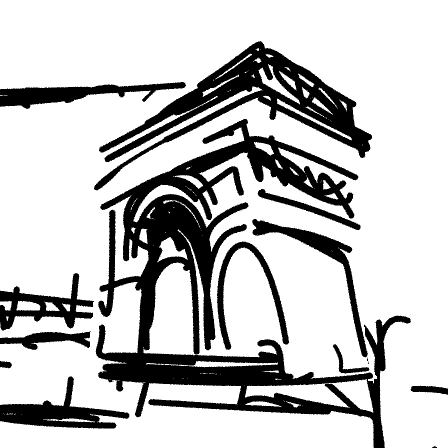}} &
    \hspace{0.5cm}
    \frame{\includegraphics[width=0.098\textwidth,height=0.098\textwidth]{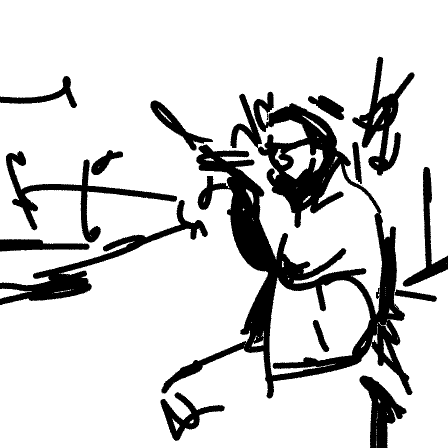}} &
    \frame{\includegraphics[width=0.098\textwidth,height=0.098\textwidth]{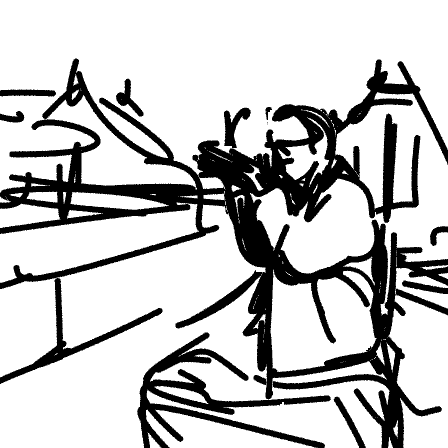}} &
    \frame{\includegraphics[width=0.098\textwidth,height=0.098\textwidth]{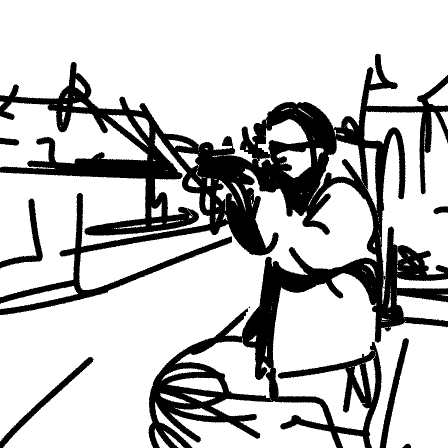}} &
    \frame{\includegraphics[width=0.098\textwidth,height=0.098\textwidth]{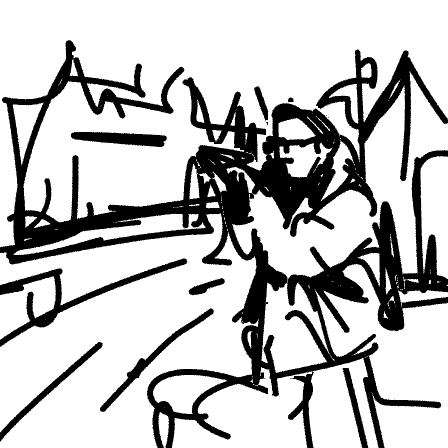}} \\
    
    \frame{\includegraphics[width=0.098\textwidth,height=0.098\textwidth]{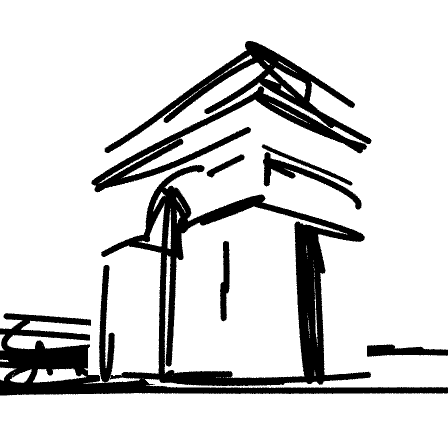}} &
    \frame{\includegraphics[width=0.098\textwidth,height=0.098\textwidth]{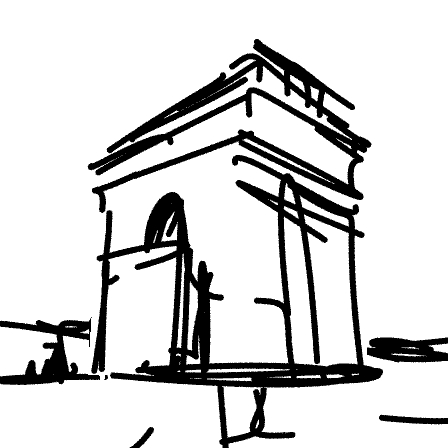}} &
    \frame{\includegraphics[width=0.098\textwidth,height=0.098\textwidth]{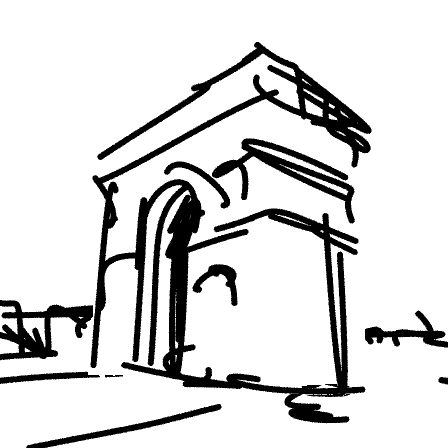}} &
    \frame{\includegraphics[width=0.098\textwidth,height=0.098\textwidth]{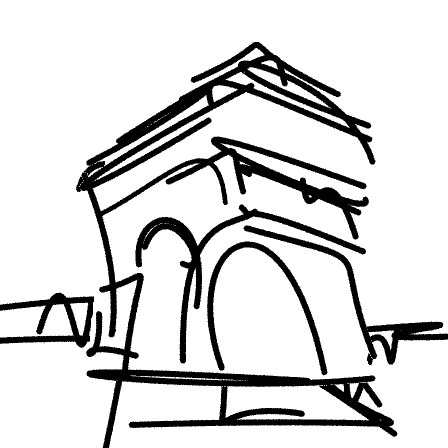}} &
    \hspace{0.5cm}
    \frame{\includegraphics[width=0.098\textwidth,height=0.098\textwidth]{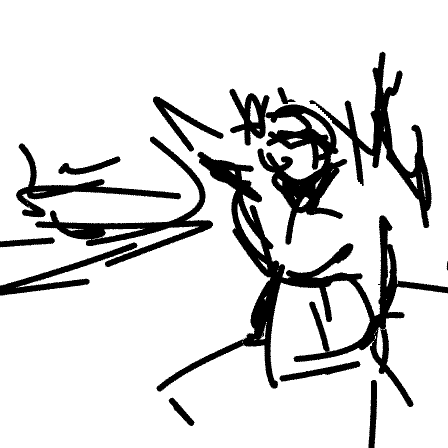}} &
    \frame{\includegraphics[width=0.098\textwidth,height=0.098\textwidth]{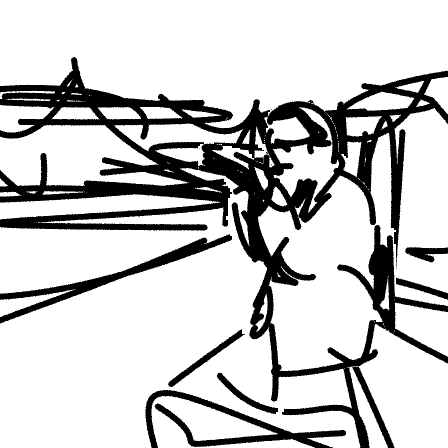}} &
    \frame{\includegraphics[width=0.098\textwidth,height=0.098\textwidth]{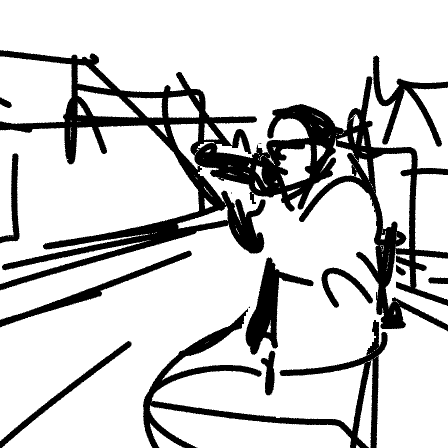}} &
    \frame{\includegraphics[width=0.098\textwidth,height=0.098\textwidth]{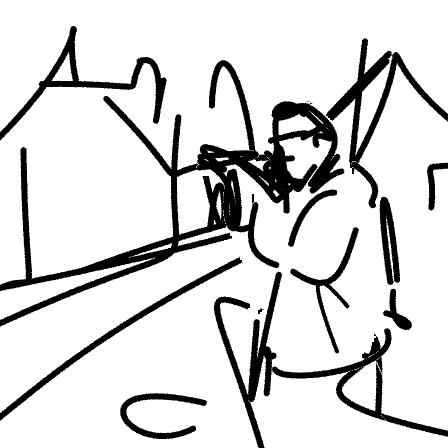}} \\
    
    \frame{\includegraphics[width=0.098\textwidth,height=0.098\textwidth]{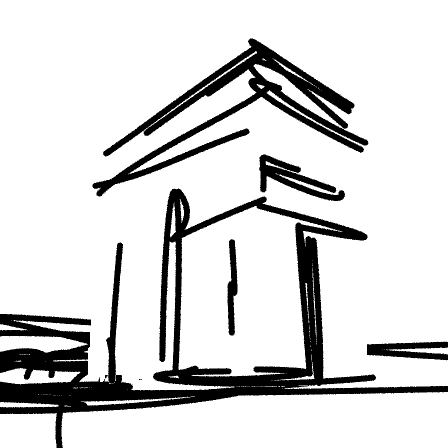}} &
    \frame{\includegraphics[width=0.098\textwidth,height=0.098\textwidth]{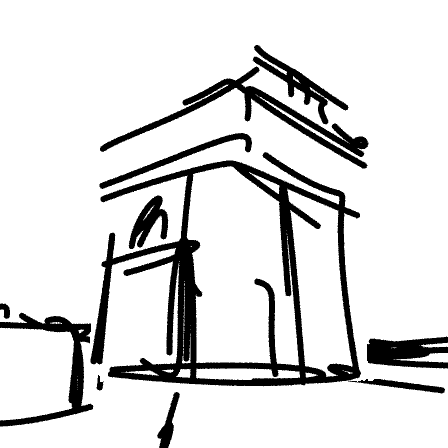}} &
    \frame{\includegraphics[width=0.098\textwidth,height=0.098\textwidth]{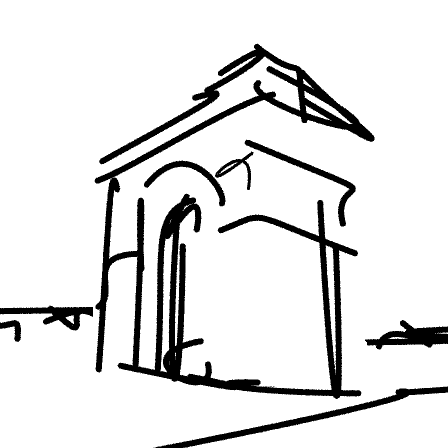}} &
    \frame{\includegraphics[width=0.098\textwidth,height=0.098\textwidth]{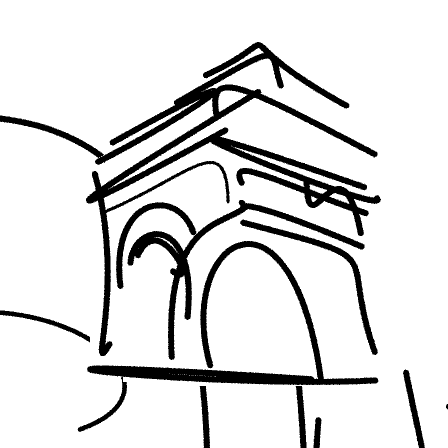}} &
    \hspace{0.5cm}
    \frame{\includegraphics[width=0.098\textwidth,height=0.098\textwidth]{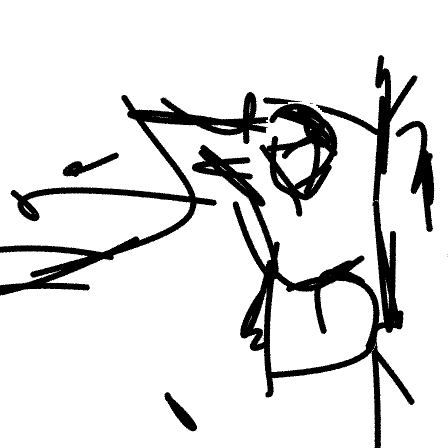}} &
    \frame{\includegraphics[width=0.098\textwidth,height=0.098\textwidth]{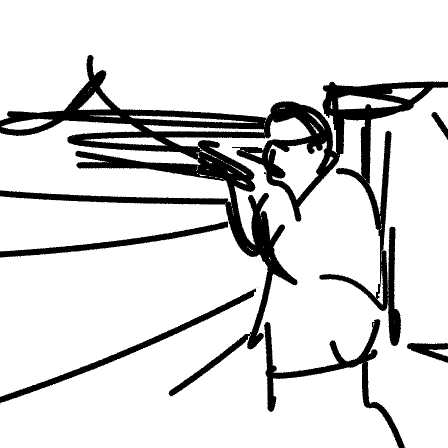}} &
    \frame{\includegraphics[width=0.098\textwidth,height=0.098\textwidth]{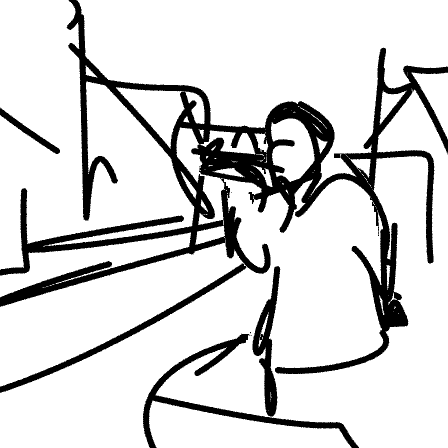}} &
    \frame{\includegraphics[width=0.098\textwidth,height=0.098\textwidth]{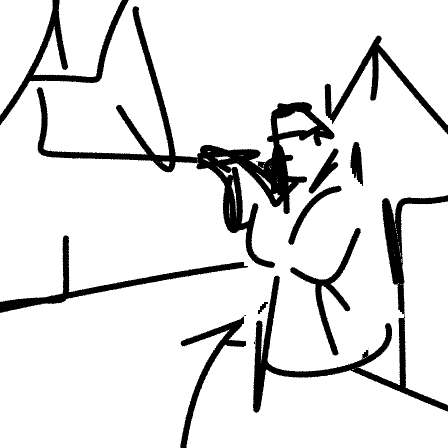}} \\
    
    \frame{\includegraphics[width=0.098\textwidth,height=0.098\textwidth]{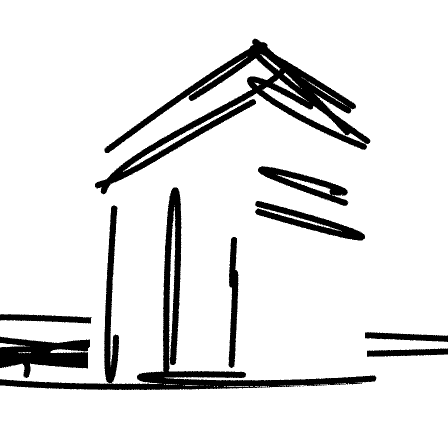}} &
    \frame{\includegraphics[width=0.098\textwidth,height=0.098\textwidth]{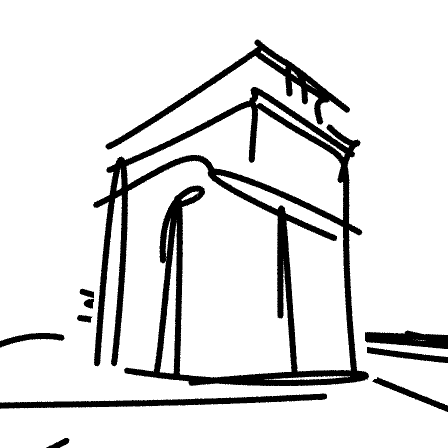}} &
    \frame{\includegraphics[width=0.098\textwidth,height=0.098\textwidth]{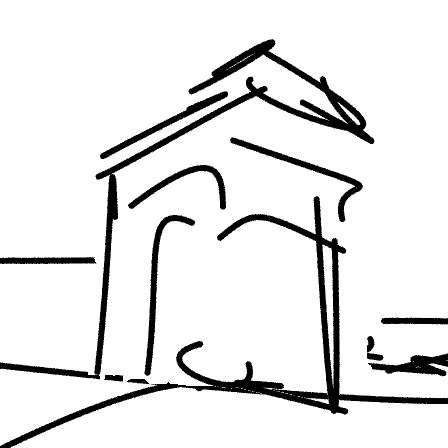}} &
    \frame{\includegraphics[width=0.098\textwidth,height=0.098\textwidth]{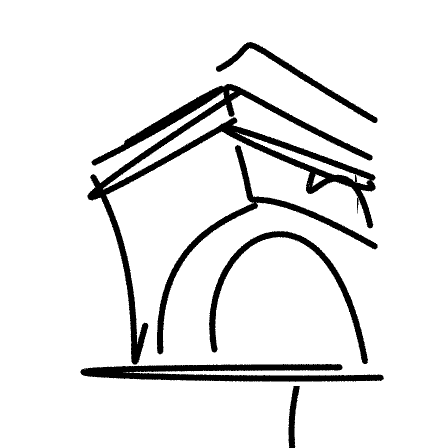}} &
    \hspace{0.5cm}
    \frame{\includegraphics[width=0.098\textwidth,height=0.098\textwidth]{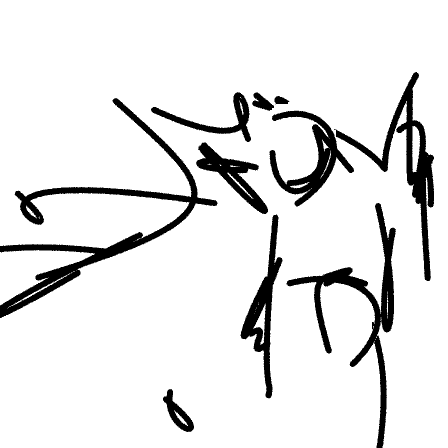}} &
    \frame{\includegraphics[width=0.098\textwidth,height=0.098\textwidth]{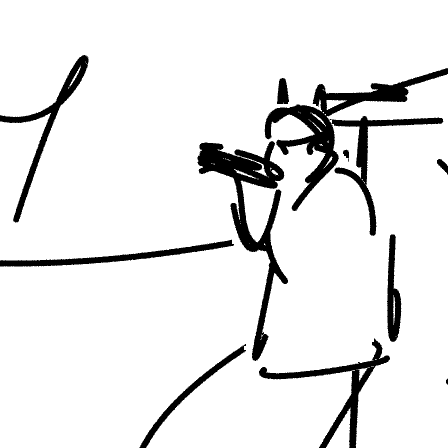}} &
    \frame{\includegraphics[width=0.098\textwidth,height=0.098\textwidth]{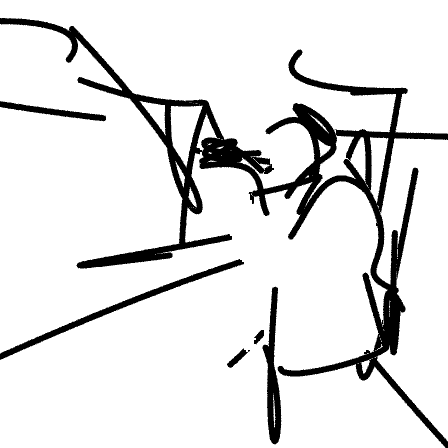}} &
    \frame{\includegraphics[width=0.098\textwidth,height=0.098\textwidth]{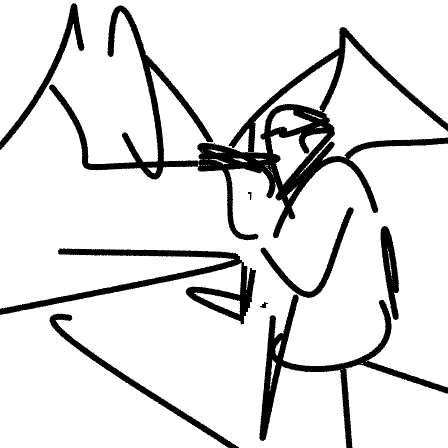}} \\

    \\
    \\
    
    \includegraphics[width=0.098\textwidth,height=0.098\textwidth]{figs/inputs/house4.jpg} & & & &
    \hspace{0.5cm}
    \includegraphics[width=0.098\textwidth,height=0.098\textwidth]{figs/inputs/ballerina.jpg} & & & \\

    \frame{\includegraphics[width=0.098\textwidth,height=0.098\textwidth]{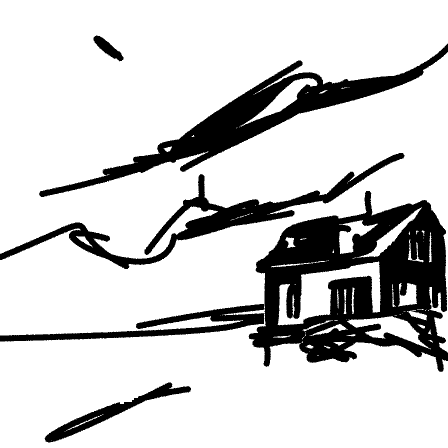}} &
    \frame{\includegraphics[width=0.098\textwidth,height=0.098\textwidth]{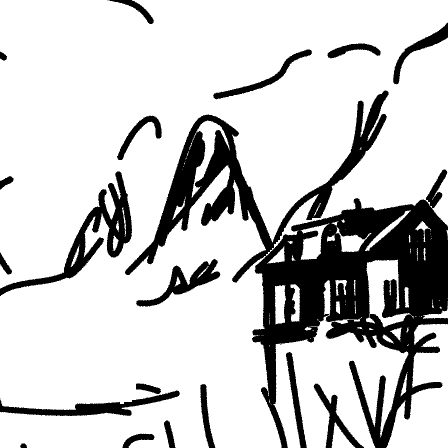}} &
    \frame{\includegraphics[width=0.098\textwidth,height=0.098\textwidth]{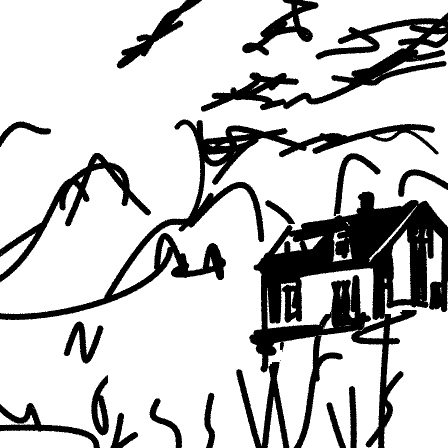}} &
    \frame{\includegraphics[width=0.098\textwidth,height=0.098\textwidth]{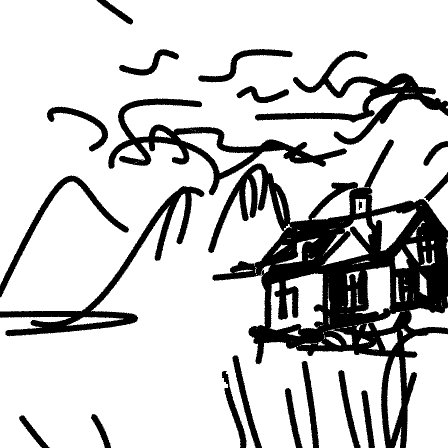}} &
    \hspace{0.5cm}
    \frame{\includegraphics[width=0.098\textwidth,height=0.098\textwidth]{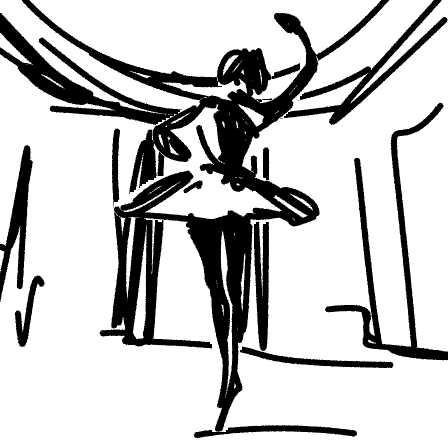}} &
    \frame{\includegraphics[width=0.098\textwidth,height=0.098\textwidth]{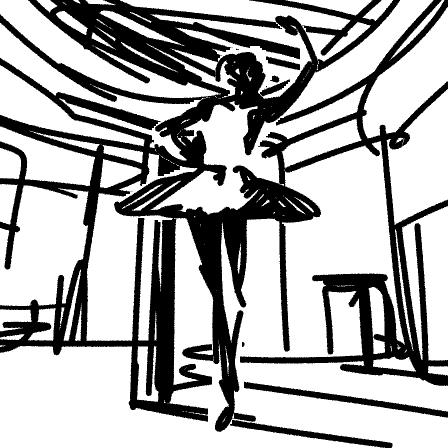}} &
    \frame{\includegraphics[width=0.098\textwidth,height=0.098\textwidth]{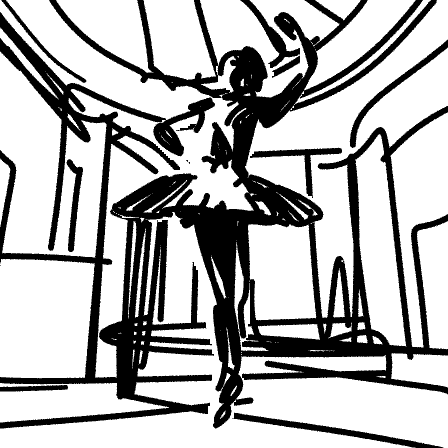}} &
    \frame{\includegraphics[width=0.098\textwidth,height=0.098\textwidth]{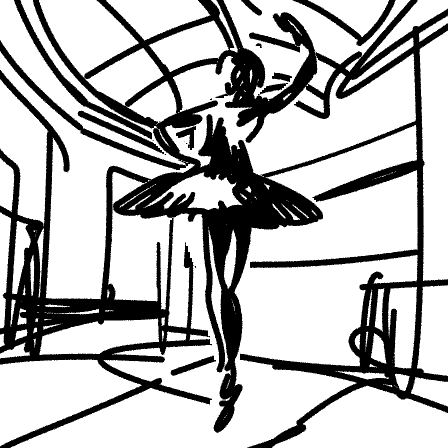}} \\
    
    \frame{\includegraphics[width=0.098\textwidth,height=0.098\textwidth]{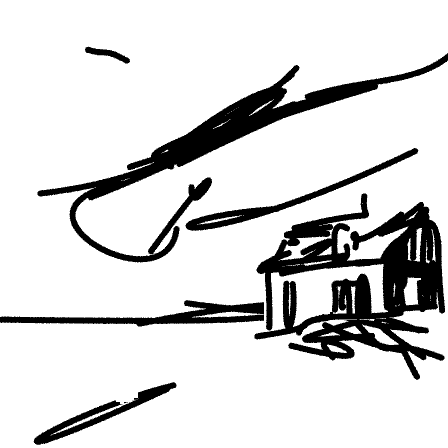}} &
    \frame{\includegraphics[width=0.098\textwidth,height=0.098\textwidth]{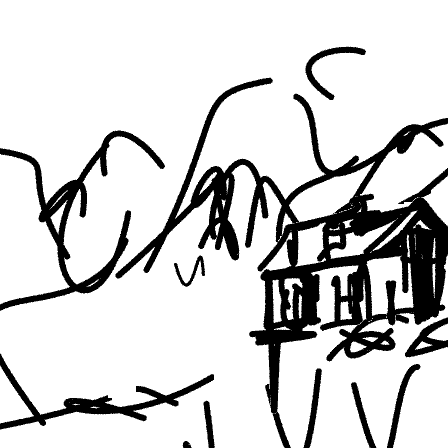}} &
    \frame{\includegraphics[width=0.098\textwidth,height=0.098\textwidth]{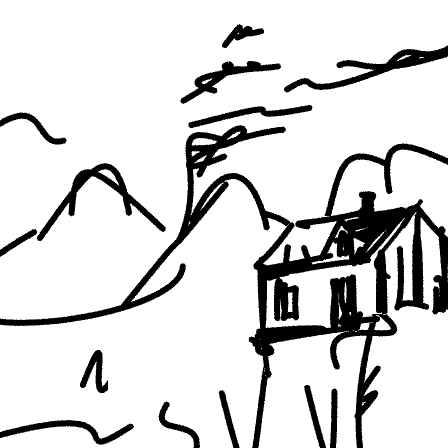}} &
    \frame{\includegraphics[width=0.098\textwidth,height=0.098\textwidth]{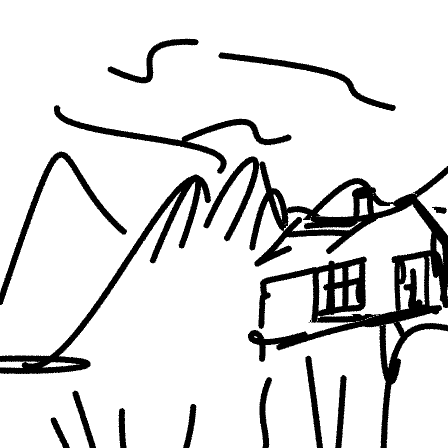}} &
    \hspace{0.5cm}
    \frame{\includegraphics[width=0.098\textwidth,height=0.098\textwidth]{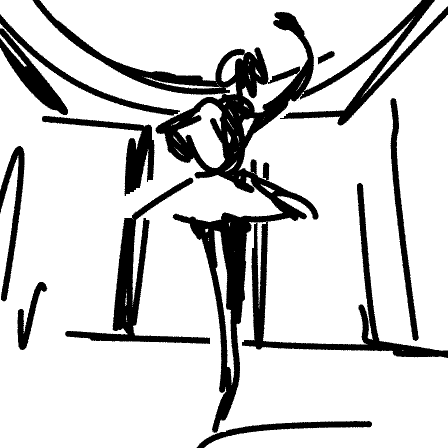}} &
    \frame{\includegraphics[width=0.098\textwidth,height=0.098\textwidth]{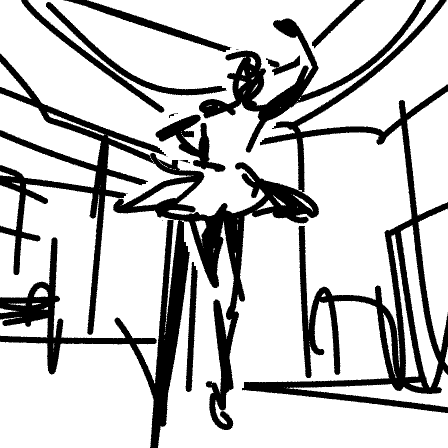}} &
    \frame{\includegraphics[width=0.098\textwidth,height=0.098\textwidth]{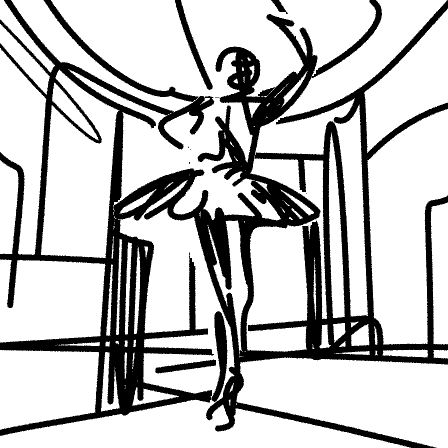}} &
    \frame{\includegraphics[width=0.098\textwidth,height=0.098\textwidth]{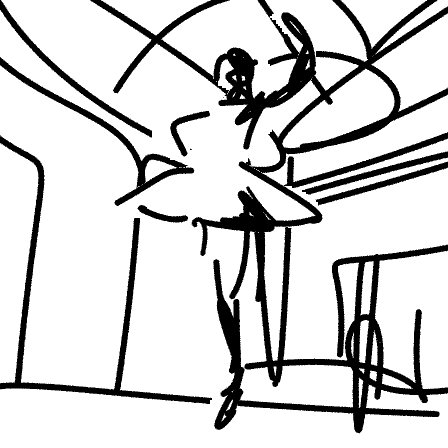}} \\
    
    \frame{\includegraphics[width=0.098\textwidth,height=0.098\textwidth]{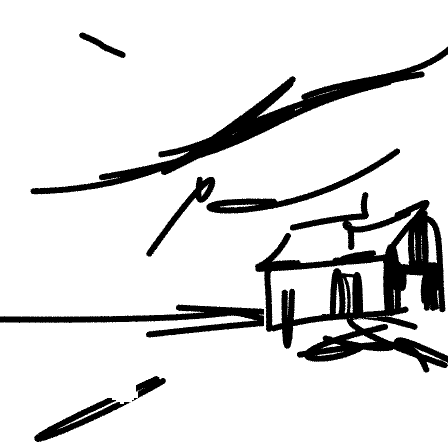}} &
    \frame{\includegraphics[width=0.098\textwidth,height=0.098\textwidth]{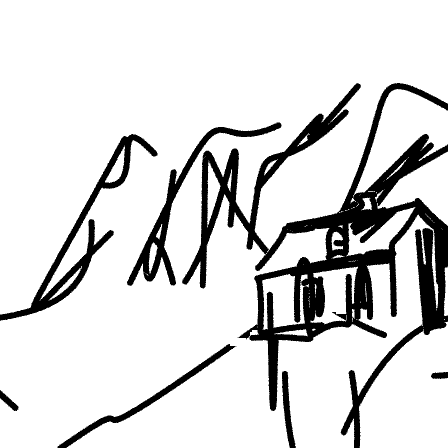}} &
    \frame{\includegraphics[width=0.098\textwidth,height=0.098\textwidth]{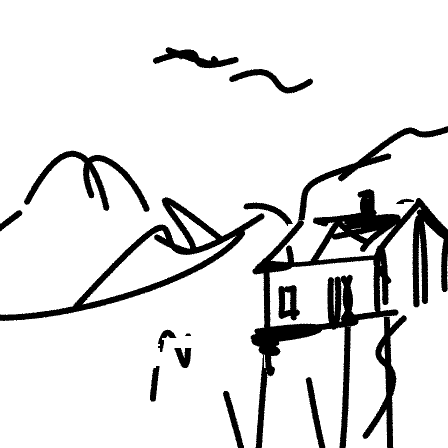}} &
    \frame{\includegraphics[width=0.098\textwidth,height=0.098\textwidth]{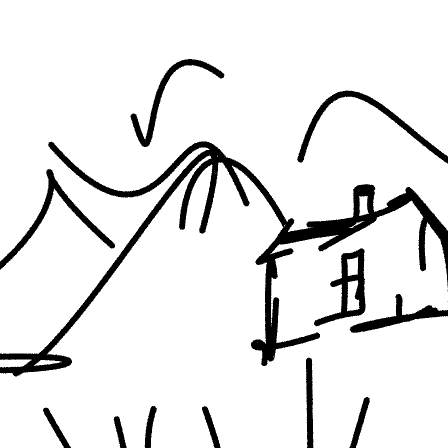}} &
    \hspace{0.5cm}
    \frame{\includegraphics[width=0.098\textwidth,height=0.098\textwidth]{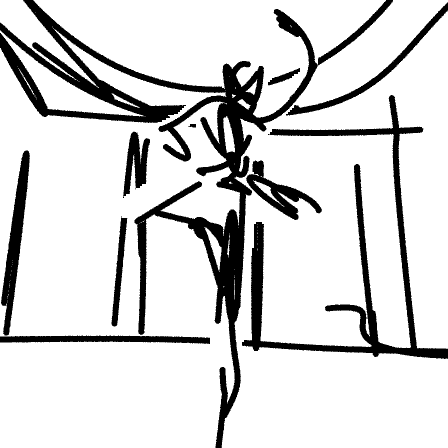}} &
    \frame{\includegraphics[width=0.098\textwidth,height=0.098\textwidth]{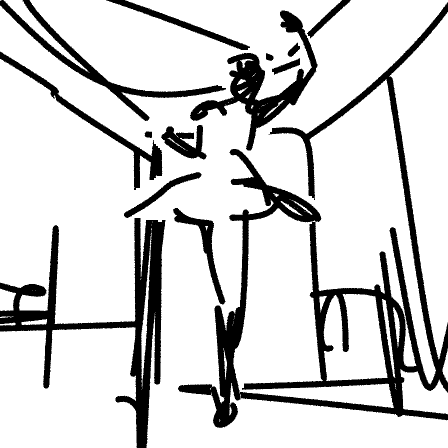}} &
    \frame{\includegraphics[width=0.098\textwidth,height=0.098\textwidth]{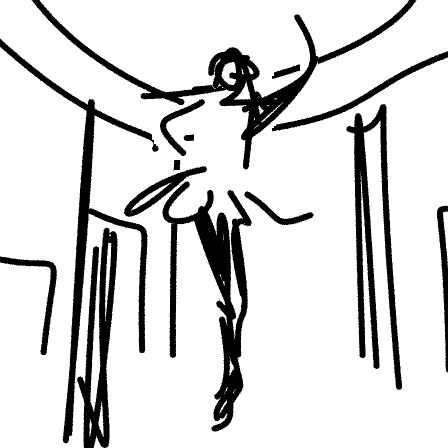}} &
    \frame{\includegraphics[width=0.098\textwidth,height=0.098\textwidth]{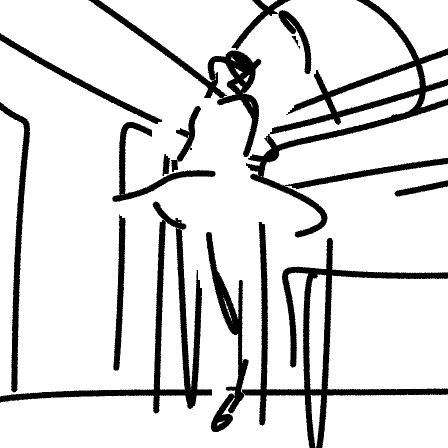}} \\
    
    \frame{\includegraphics[width=0.098\textwidth,height=0.098\textwidth]{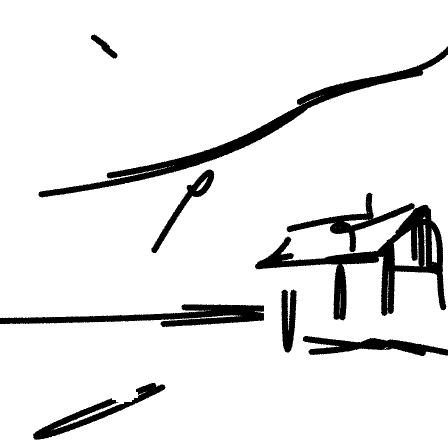}} &
    \frame{\includegraphics[width=0.098\textwidth,height=0.098\textwidth]{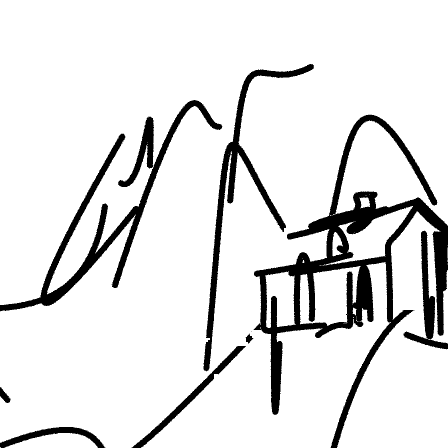}} &
    \frame{\includegraphics[width=0.098\textwidth,height=0.098\textwidth]{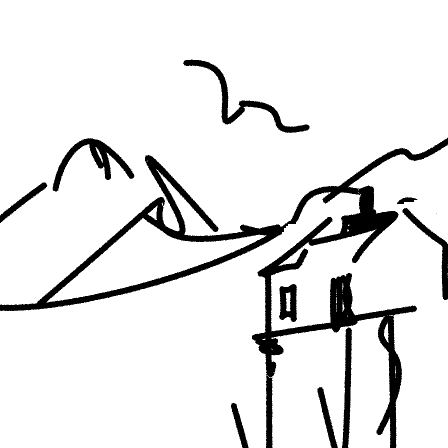}} &
    \frame{\includegraphics[width=0.098\textwidth,height=0.098\textwidth]{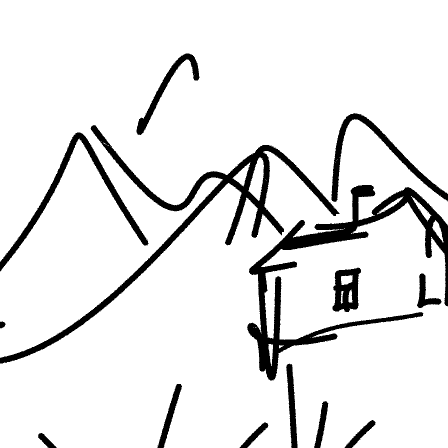}} &
    \hspace{0.5cm}
    \frame{\includegraphics[width=0.098\textwidth,height=0.098\textwidth]{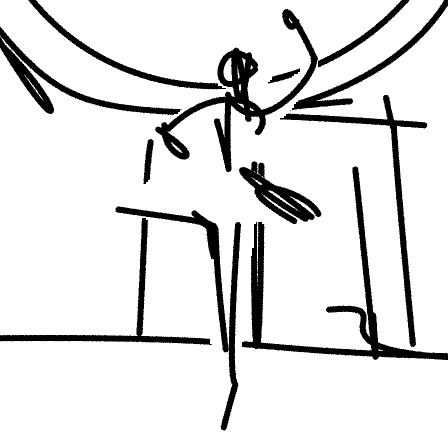}} &
    \frame{\includegraphics[width=0.098\textwidth,height=0.098\textwidth]{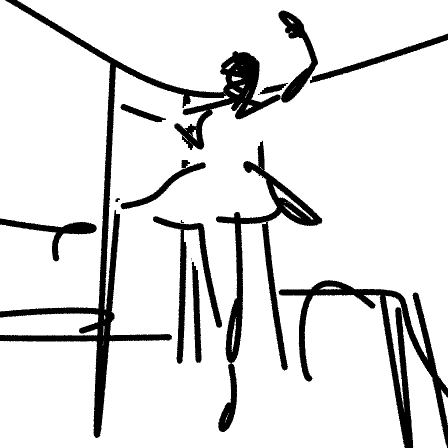}} &
    \frame{\includegraphics[width=0.098\textwidth,height=0.098\textwidth]{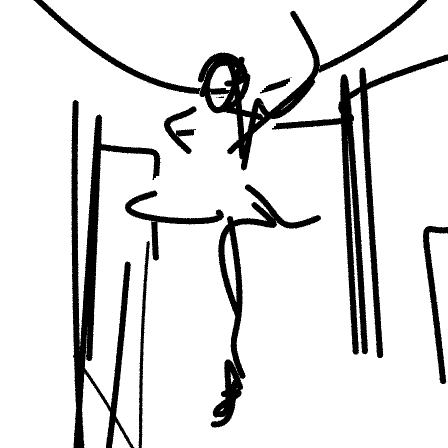}} &
    \frame{\includegraphics[width=0.098\textwidth,height=0.098\textwidth]{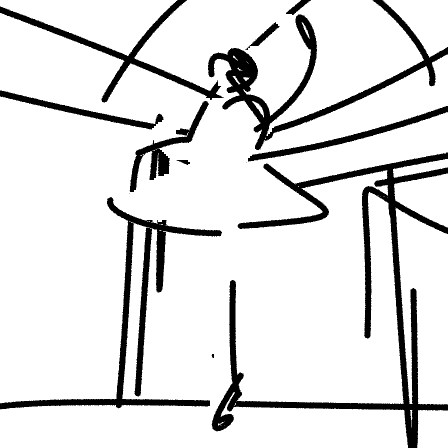}} \\

    \end{tabular}
    \caption{The $4\times4$ matrix of sketches produced by our method. Columns from left to right illustrate the change in fidelity, from precise to loose, and rows from top to bottom illustrate the visual simplification.}
    
    \label{fig:matrix1}
\end{figure*}

%% file: files/figures/supplementary/our_matrices/our_matrices_2.tex
\begin{figure*}
    \centering
    
    \begin{tabular}{c c c c c c c c}

    \includegraphics[width=0.098\textwidth,height=0.098\textwidth]{figs/inputs/man_flowers.jpg} & & & &
    \hspace{0.5cm}
    \includegraphics[width=0.098\textwidth,height=0.098\textwidth]{figs/inputs/dog.jpg} & & & \\
    
    \frame{\includegraphics[width=0.098\textwidth,height=0.098\textwidth]{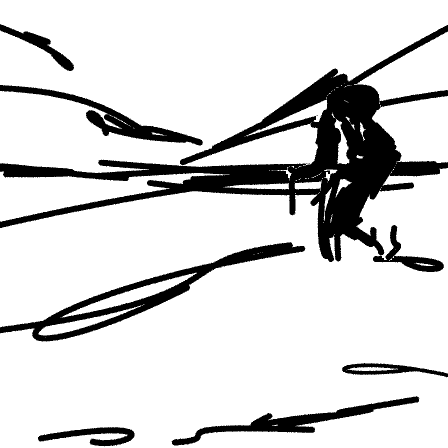}} &
    \frame{\includegraphics[width=0.098\textwidth,height=0.098\textwidth]{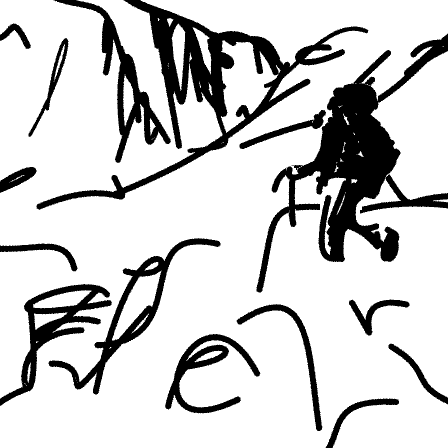}} &
    \frame{\includegraphics[width=0.098\textwidth,height=0.098\textwidth]{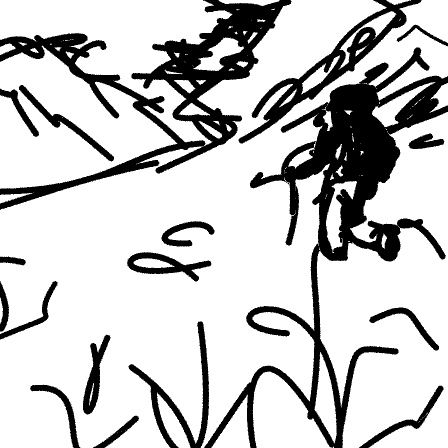}} &
    \frame{\includegraphics[width=0.098\textwidth,height=0.098\textwidth]{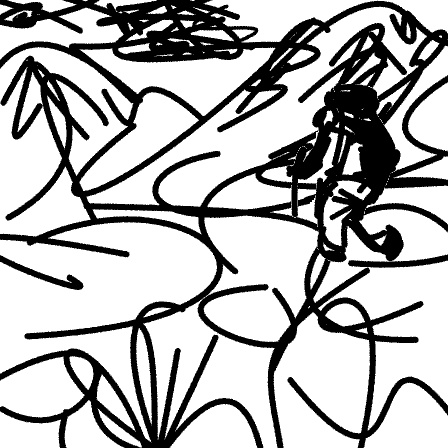}} &
    \hspace{0.5cm}
    \frame{\includegraphics[width=0.098\textwidth,height=0.098\textwidth]{figs/matrices_black/dog_row0col0_black.png}} &
    \frame{\includegraphics[width=0.098\textwidth,height=0.098\textwidth]{figs/matrices_black/dog_row0col1_black.png}} &
    \frame{\includegraphics[width=0.098\textwidth,height=0.098\textwidth]{figs/matrices_black/dog_row0col2_black.png}} &
    \frame{\includegraphics[width=0.098\textwidth,height=0.098\textwidth]{figs/matrices_black/dog_row0col3_black.png}} \\
    
    \frame{\includegraphics[width=0.098\textwidth,height=0.098\textwidth]{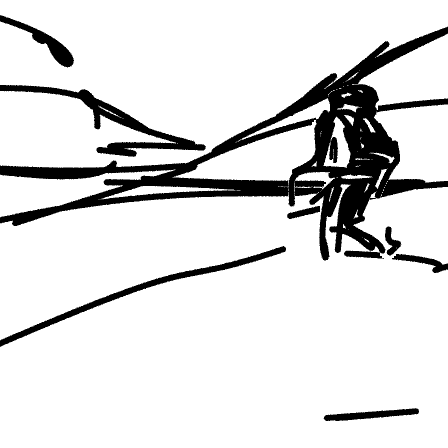}} &
    \frame{\includegraphics[width=0.098\textwidth,height=0.098\textwidth]{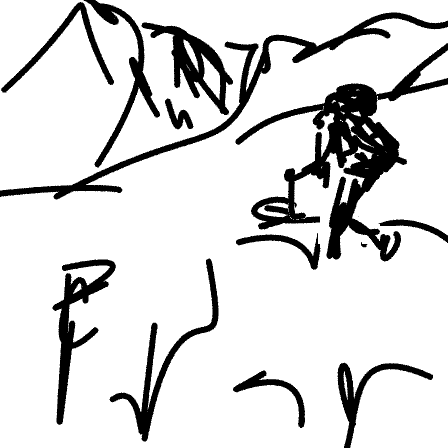}} &
    \frame{\includegraphics[width=0.098\textwidth,height=0.098\textwidth]{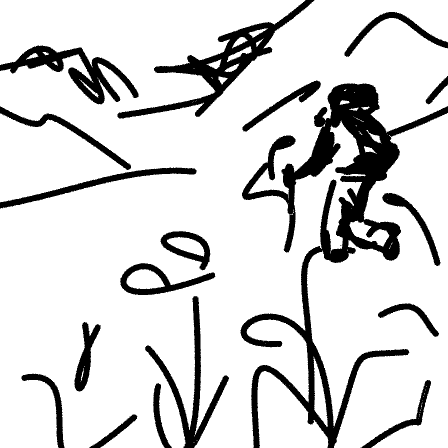}} &
    \frame{\includegraphics[width=0.098\textwidth,height=0.098\textwidth]{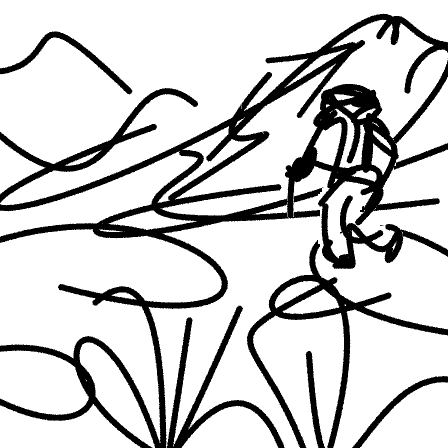}} &
    \hspace{0.5cm}
    \frame{\includegraphics[width=0.098\textwidth,height=0.098\textwidth]{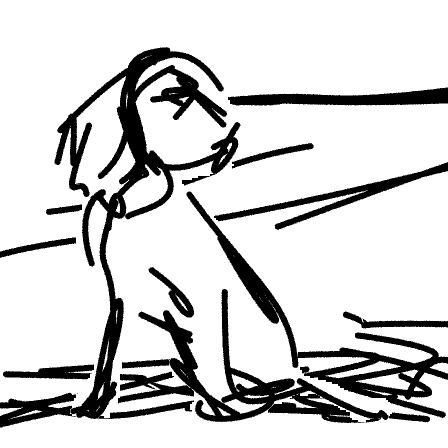}} &
    \frame{\includegraphics[width=0.098\textwidth,height=0.098\textwidth]{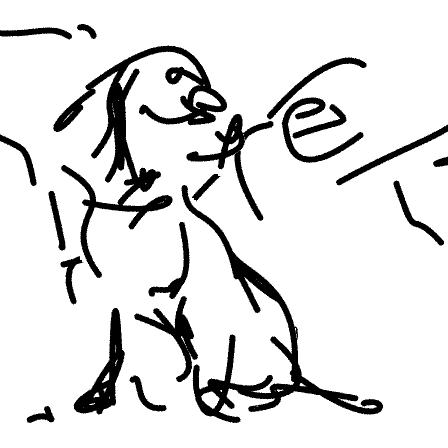}} &
    \frame{\includegraphics[width=0.098\textwidth,height=0.098\textwidth]{figs/matrices_black/dog_row1col2_black.png}} &
    \frame{\includegraphics[width=0.098\textwidth,height=0.098\textwidth]{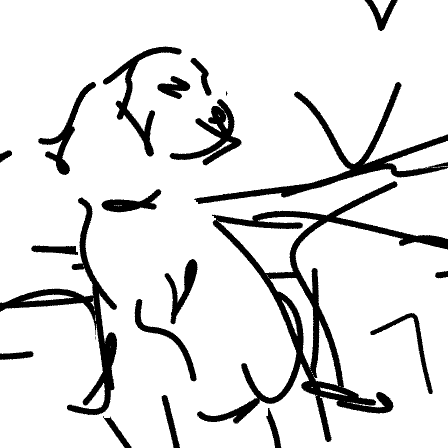}} \\
    
    \frame{\includegraphics[width=0.098\textwidth,height=0.098\textwidth]{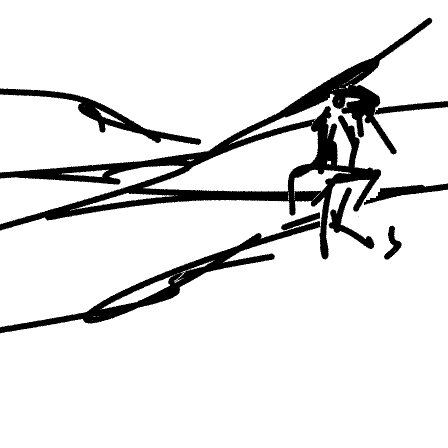}} &
    \frame{\includegraphics[width=0.098\textwidth,height=0.098\textwidth]{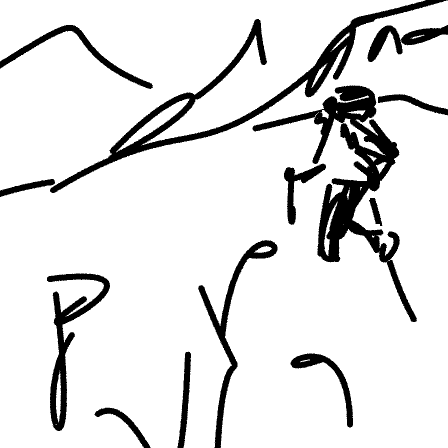}} &
    \frame{\includegraphics[width=0.098\textwidth,height=0.098\textwidth]{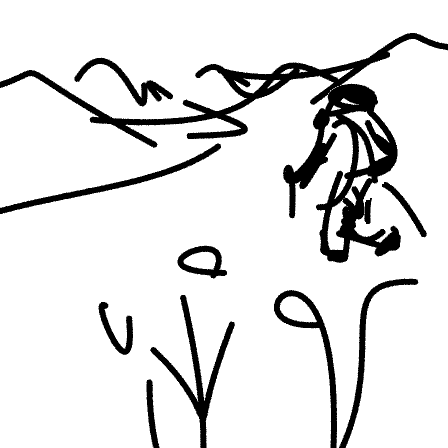}} &
    \frame{\includegraphics[width=0.098\textwidth,height=0.098\textwidth]{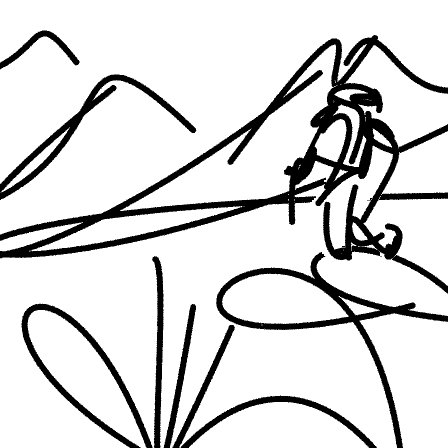}} &
    \hspace{0.5cm}
    \frame{\includegraphics[width=0.098\textwidth,height=0.098\textwidth]{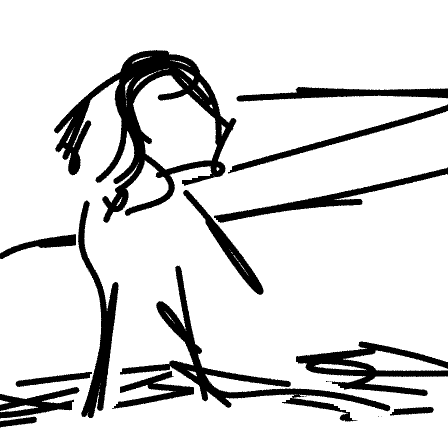}} &
    \frame{\includegraphics[width=0.098\textwidth,height=0.098\textwidth]{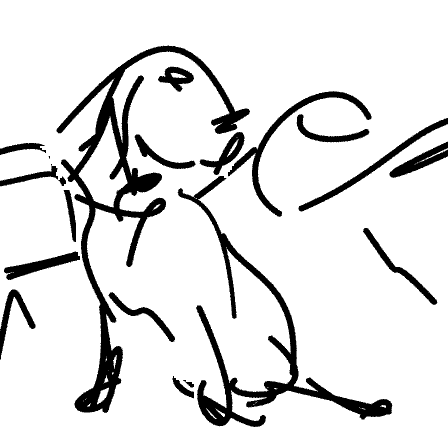}} &
    \frame{\includegraphics[width=0.098\textwidth,height=0.098\textwidth]{figs/matrices_black/dog_row2col2_black.png}} &
    \frame{\includegraphics[width=0.098\textwidth,height=0.098\textwidth]{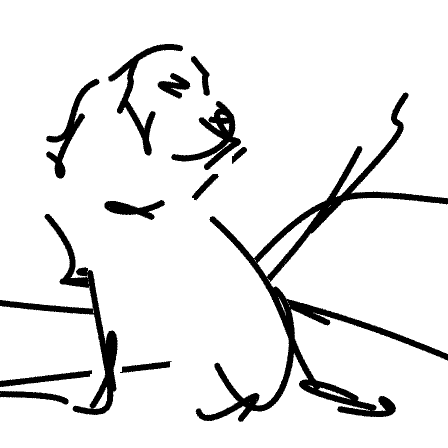}} \\
    
    \frame{\includegraphics[width=0.098\textwidth,height=0.098\textwidth]{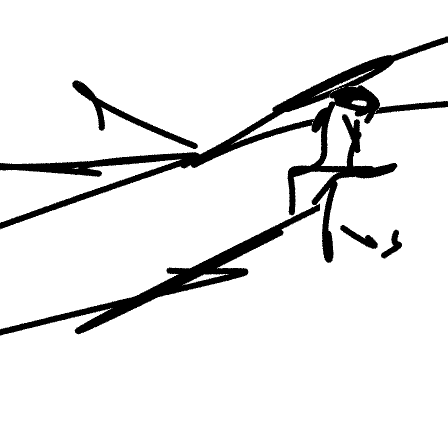}} &
    \frame{\includegraphics[width=0.098\textwidth,height=0.098\textwidth]{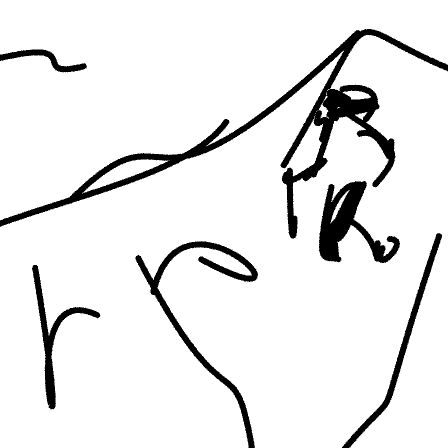}} &
    \frame{\includegraphics[width=0.098\textwidth,height=0.098\textwidth]{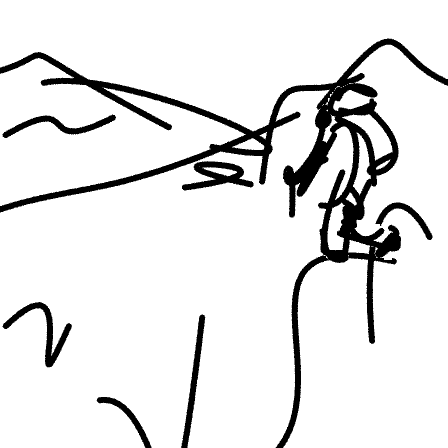}} &
    \frame{\includegraphics[width=0.098\textwidth,height=0.098\textwidth]{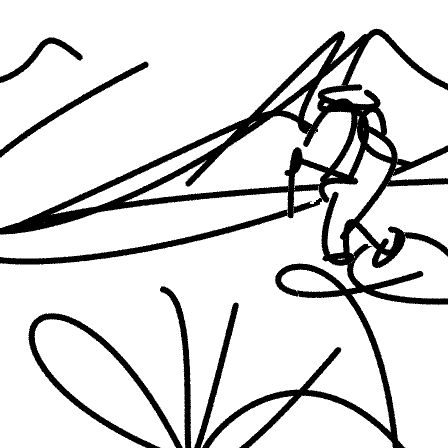}} &
    \hspace{0.5cm}
    \frame{\includegraphics[width=0.098\textwidth,height=0.098\textwidth]{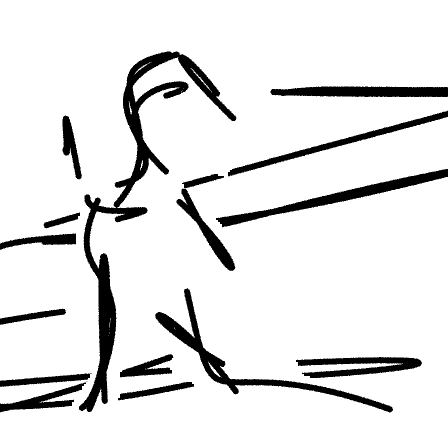}} &
    \frame{\includegraphics[width=0.098\textwidth,height=0.098\textwidth]{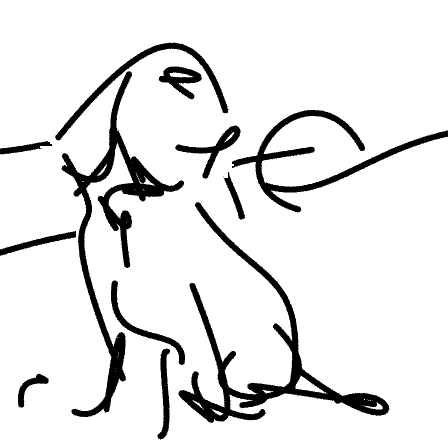}} &
    \frame{\includegraphics[width=0.098\textwidth,height=0.098\textwidth]{figs/matrices_black/dog_row3col2_black.png}} &
    \frame{\includegraphics[width=0.098\textwidth,height=0.098\textwidth]{figs/matrices_black/dog_row3col3_black.png}} \\

    \\
    \\
    
    \includegraphics[width=0.098\textwidth,height=0.098\textwidth]{figs/inputs/bull.jpg} & & & &
    \hspace{0.5cm}
    \includegraphics[width=0.098\textwidth,height=0.098\textwidth]{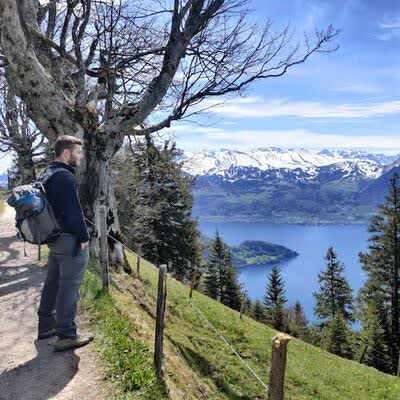} & & & \\

    \frame{\includegraphics[width=0.098\textwidth,height=0.098\textwidth]{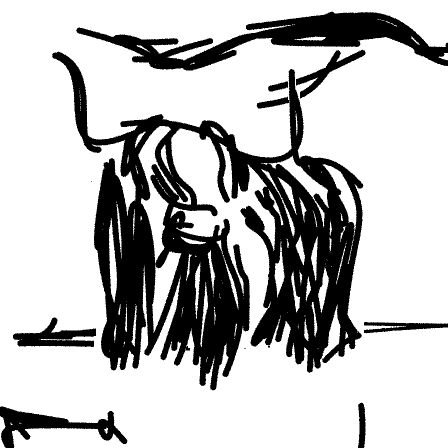}} &
    \frame{\includegraphics[width=0.098\textwidth,height=0.098\textwidth]{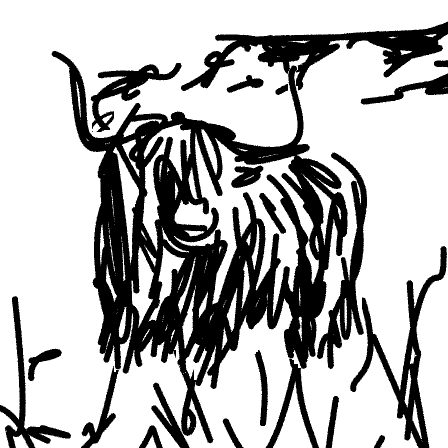}} &
    \frame{\includegraphics[width=0.098\textwidth,height=0.098\textwidth]{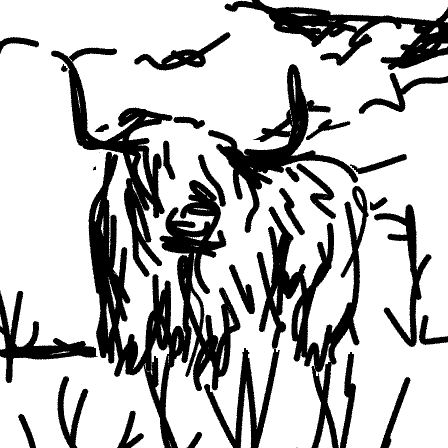}} &
    \frame{\includegraphics[width=0.098\textwidth,height=0.098\textwidth]{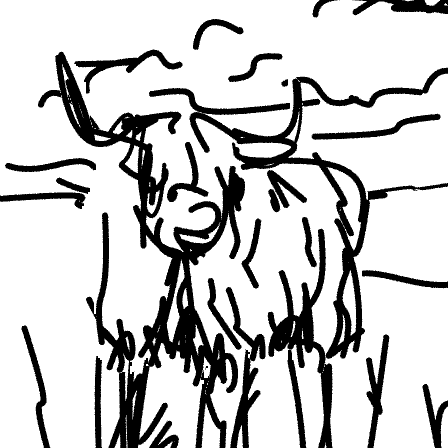}} &
    \hspace{0.5cm}
    \frame{\includegraphics[width=0.098\textwidth,height=0.098\textwidth]{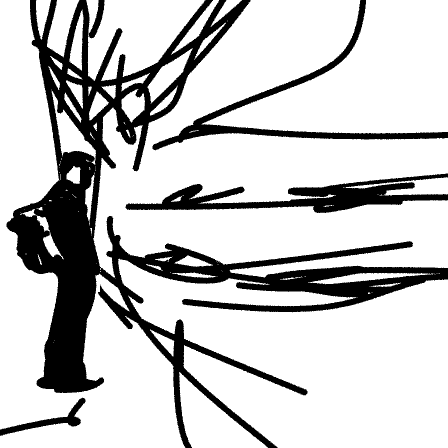}} &
    \frame{\includegraphics[width=0.098\textwidth,height=0.098\textwidth]{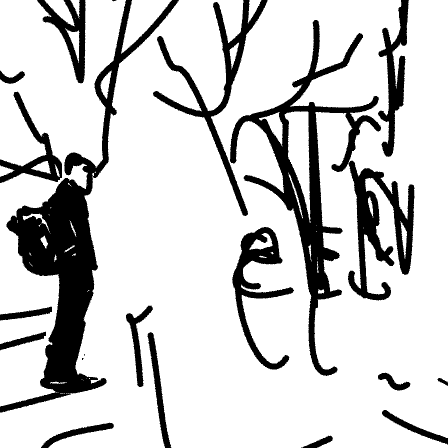}} &
    \frame{\includegraphics[width=0.098\textwidth,height=0.098\textwidth]{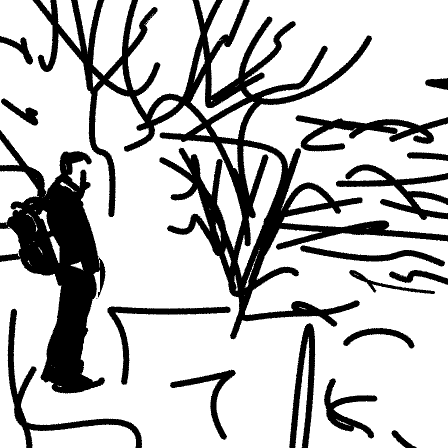}} &
    \frame{\includegraphics[width=0.098\textwidth,height=0.098\textwidth]{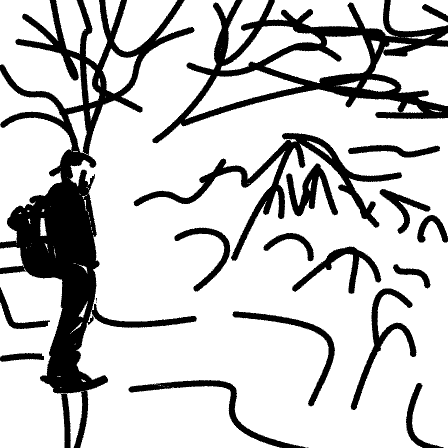}} \\
    
    \frame{\includegraphics[width=0.098\textwidth,height=0.098\textwidth]{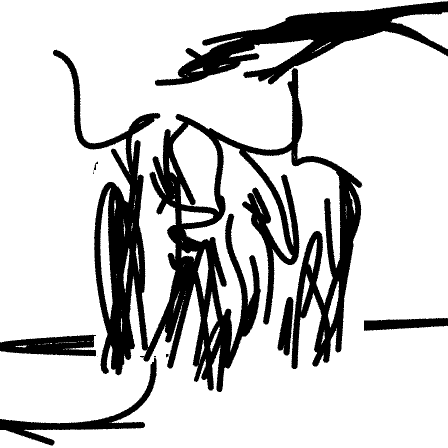}} &
    \frame{\includegraphics[width=0.098\textwidth,height=0.098\textwidth]{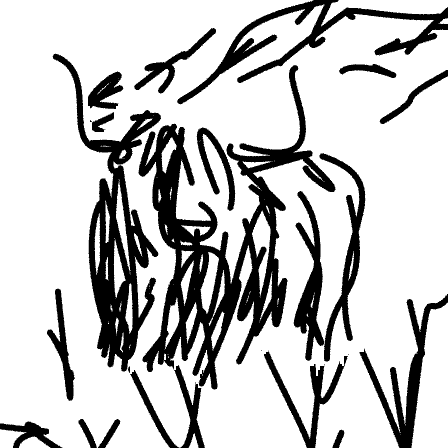}} &
    \frame{\includegraphics[width=0.098\textwidth,height=0.098\textwidth]{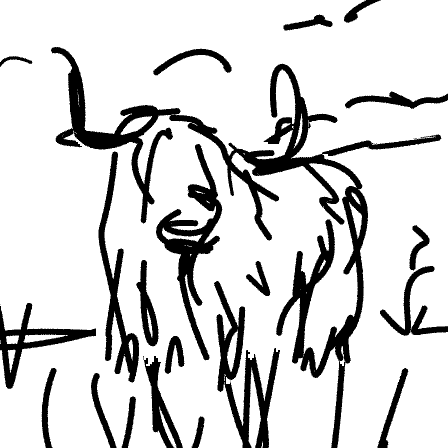}} &
    \frame{\includegraphics[width=0.098\textwidth,height=0.098\textwidth]{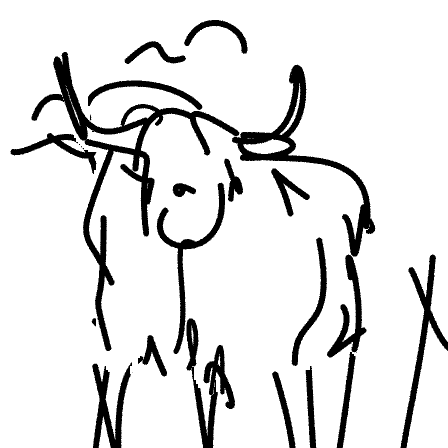}} &
    \hspace{0.5cm}
    \frame{\includegraphics[width=0.098\textwidth,height=0.098\textwidth]{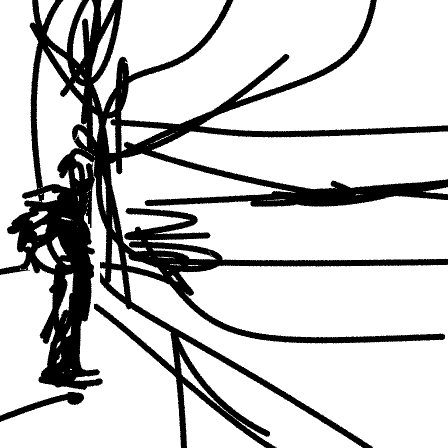}} &
    \frame{\includegraphics[width=0.098\textwidth,height=0.098\textwidth]{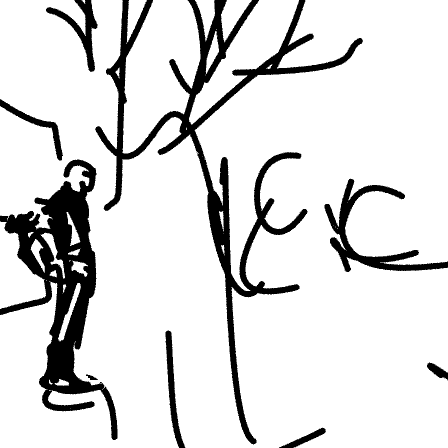}} &
    \frame{\includegraphics[width=0.098\textwidth,height=0.098\textwidth]{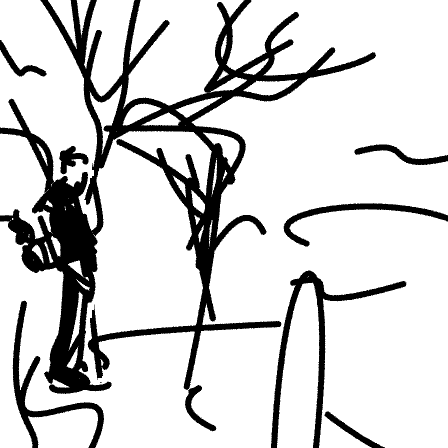}} &
    \frame{\includegraphics[width=0.098\textwidth,height=0.098\textwidth]{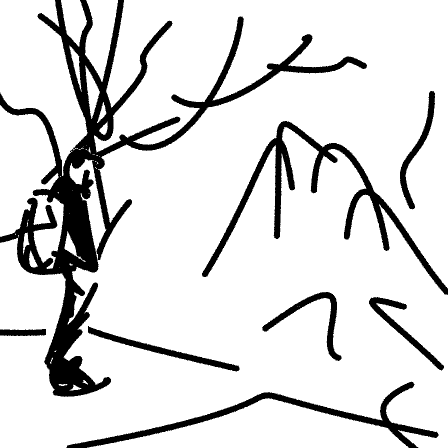}} \\
    
    \frame{\includegraphics[width=0.098\textwidth,height=0.098\textwidth]{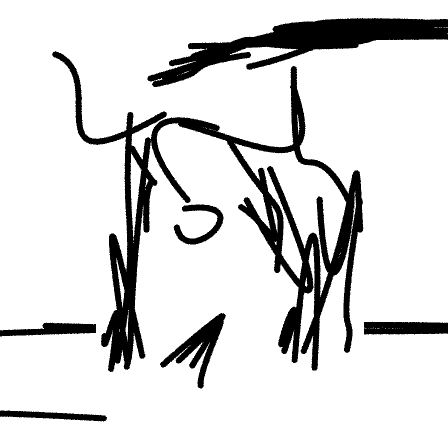}} &
    \frame{\includegraphics[width=0.098\textwidth,height=0.098\textwidth]{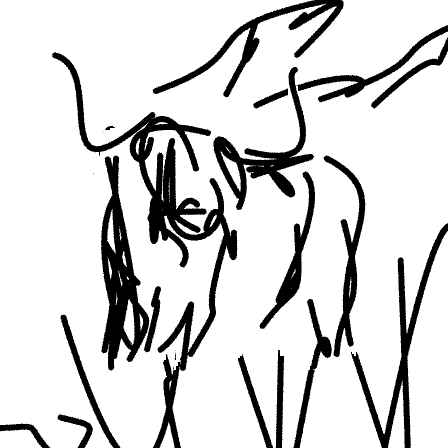}} &
    \frame{\includegraphics[width=0.098\textwidth,height=0.098\textwidth]{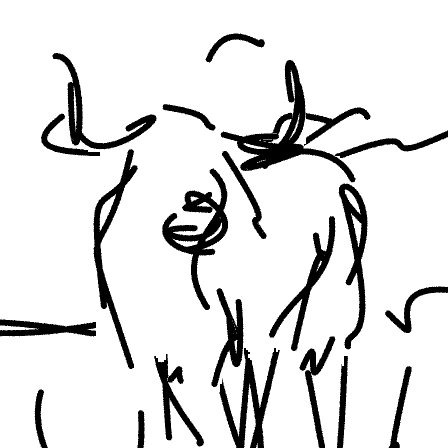}} &
    \frame{\includegraphics[width=0.098\textwidth,height=0.098\textwidth]{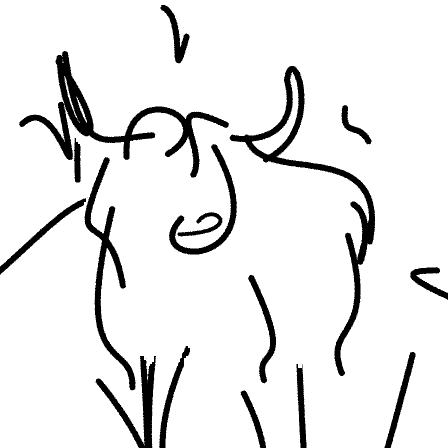}} &
    \hspace{0.5cm}
    \frame{\includegraphics[width=0.098\textwidth,height=0.098\textwidth]{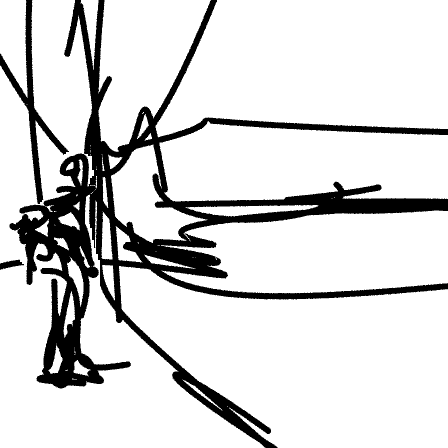}} &
    \frame{\includegraphics[width=0.098\textwidth,height=0.098\textwidth]{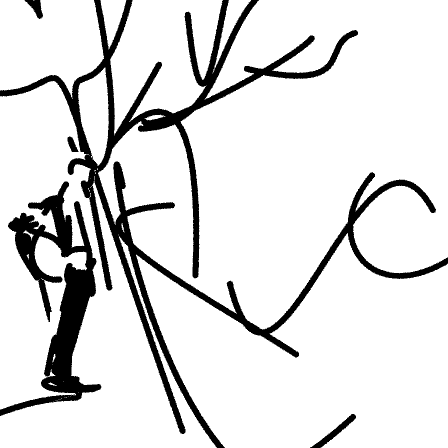}} &
    \frame{\includegraphics[width=0.098\textwidth,height=0.098\textwidth]{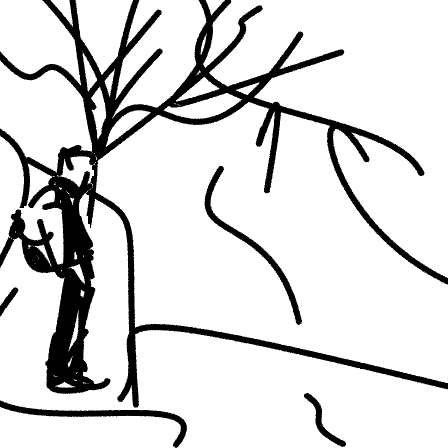}} &
    \frame{\includegraphics[width=0.098\textwidth,height=0.098\textwidth]{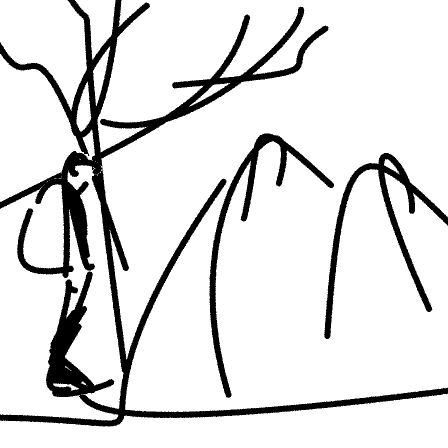}} \\
    
    \frame{\includegraphics[width=0.098\textwidth,height=0.098\textwidth]{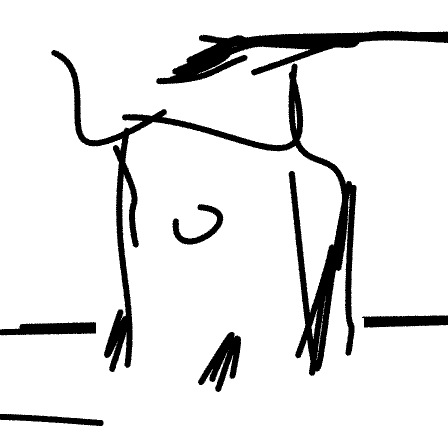}} &
    \frame{\includegraphics[width=0.098\textwidth,height=0.098\textwidth]{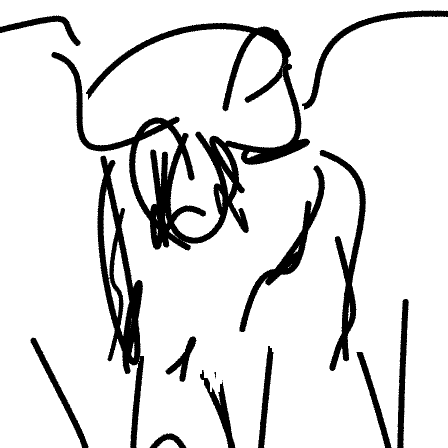}} &
    \frame{\includegraphics[width=0.098\textwidth,height=0.098\textwidth]{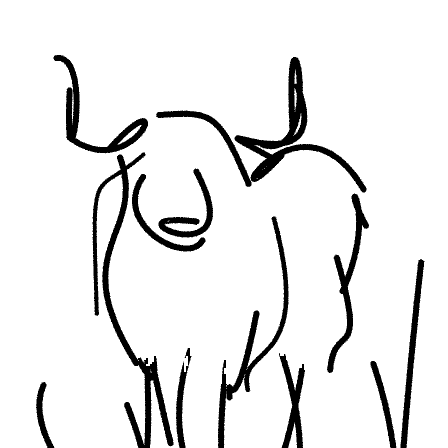}} &
    \frame{\includegraphics[width=0.098\textwidth,height=0.098\textwidth]{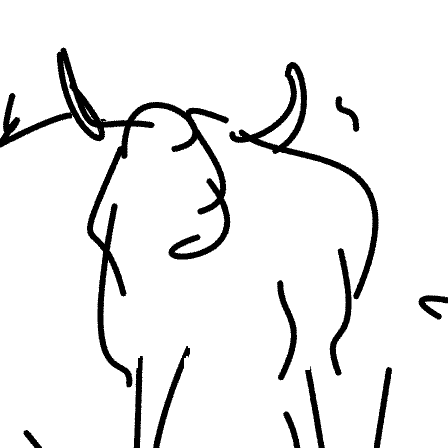}} &
    \hspace{0.5cm}
    \frame{\includegraphics[width=0.098\textwidth,height=0.098\textwidth]{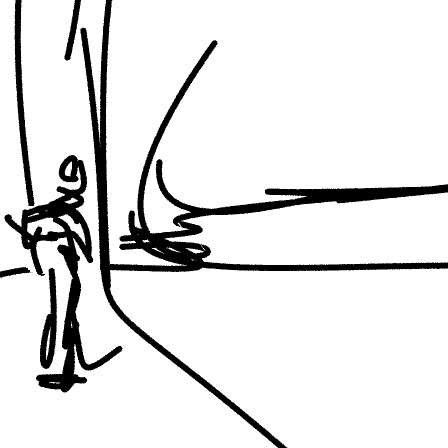}} &
    \frame{\includegraphics[width=0.098\textwidth,height=0.098\textwidth]{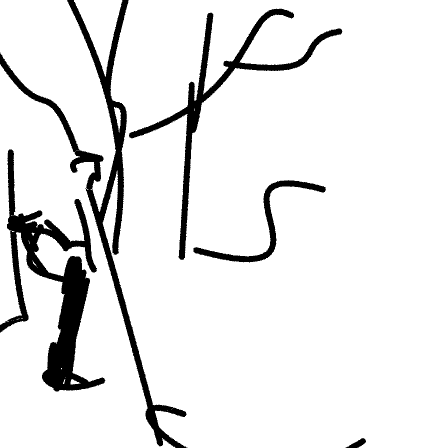}} &
    \frame{\includegraphics[width=0.098\textwidth,height=0.098\textwidth]{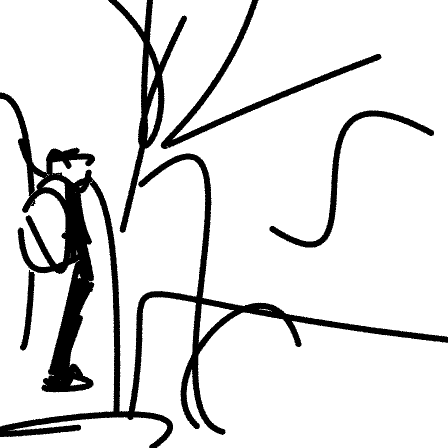}} &
    \frame{\includegraphics[width=0.098\textwidth,height=0.098\textwidth]{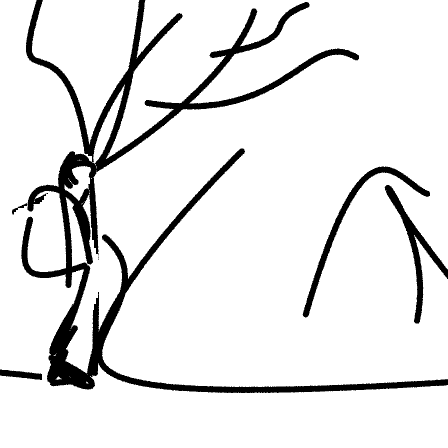}} \\

    \end{tabular}
    \caption{The $4\times4$ matrix of sketches produced by our method. Columns from left to right illustrate the change in fidelity, from precise to loose, and rows from top to bottom illustrate the visual simplification.}
    
    \label{fig:matrix2}
\end{figure*}

%% file: files/figures/supplementary/our_matrices/our_matrices_3.tex
\begin{figure*}
    \centering
    
    \begin{tabular}{c c c c c c c c}

    \includegraphics[width=0.098\textwidth,height=0.098\textwidth]{figs/inputs/woman_city.jpg} & & & &
    \hspace{0.5cm}
    \includegraphics[width=0.098\textwidth,height=0.098\textwidth]{figs/inputs/black_woman.jpg} & & & \\
    
    \frame{\includegraphics[width=0.098\textwidth,height=0.098\textwidth]{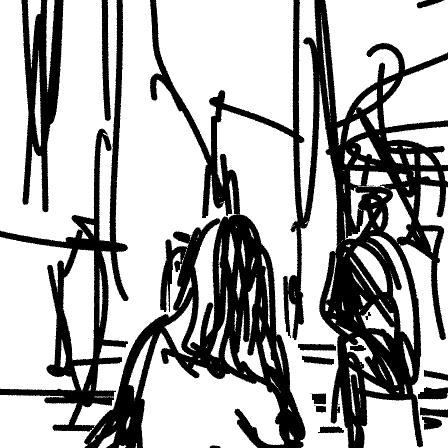}} &
    \frame{\includegraphics[width=0.098\textwidth,height=0.098\textwidth]{figs/matrices_black/woman_city_row0col1_black.png}} &
    \frame{\includegraphics[width=0.098\textwidth,height=0.098\textwidth]{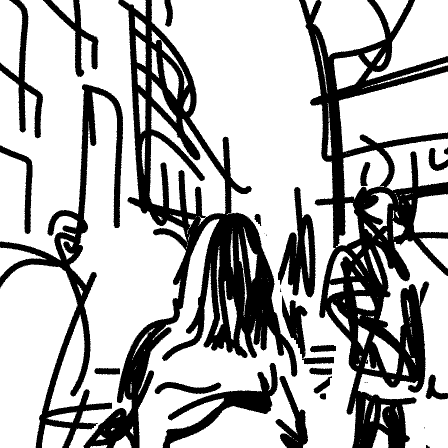}} &
    \frame{\includegraphics[width=0.098\textwidth,height=0.098\textwidth]{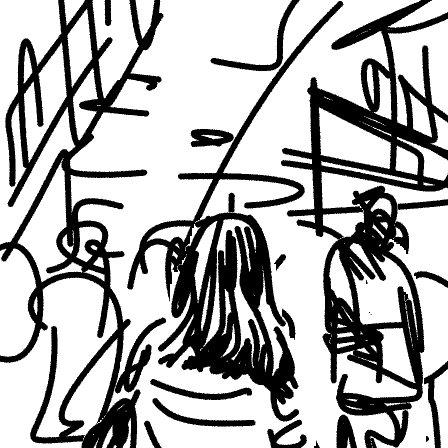}} &
    \hspace{0.5cm}
    \frame{\includegraphics[width=0.098\textwidth,height=0.098\textwidth]{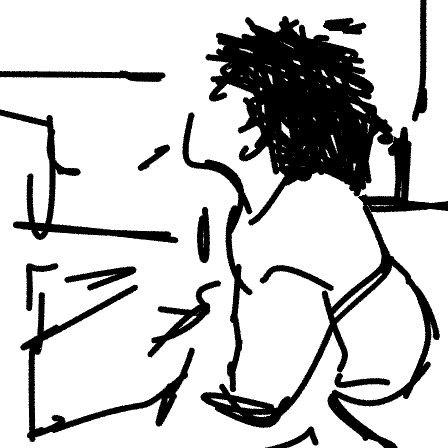}} &
    \frame{\includegraphics[width=0.098\textwidth,height=0.098\textwidth]{figs/matrices_black/black_woman_4.png}} &
    \frame{\includegraphics[width=0.098\textwidth,height=0.098\textwidth]{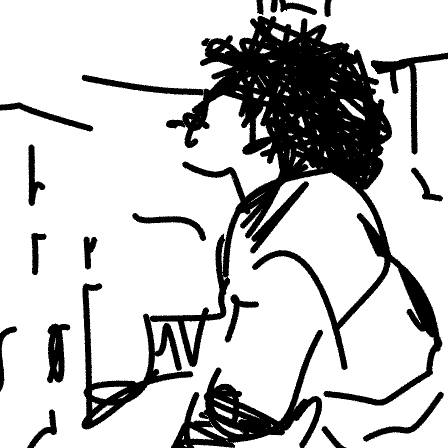}} &
    \frame{\includegraphics[width=0.098\textwidth,height=0.098\textwidth]{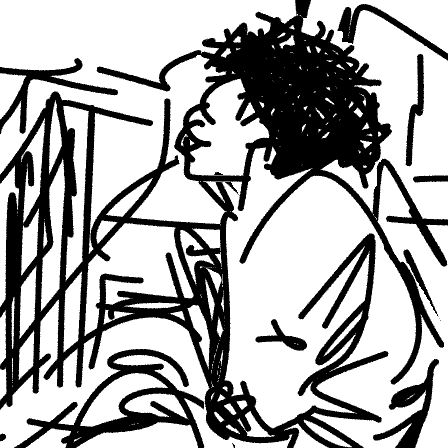}} \\
    
    \frame{\includegraphics[width=0.098\textwidth,height=0.098\textwidth]{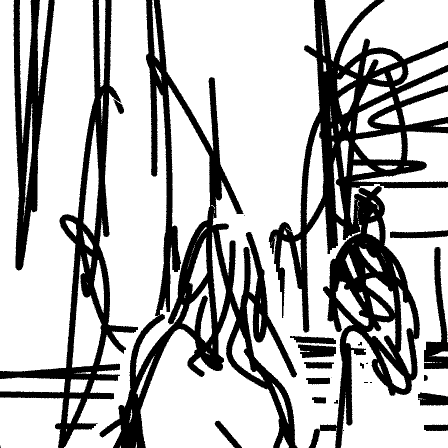}} &
    \frame{\includegraphics[width=0.098\textwidth,height=0.098\textwidth]{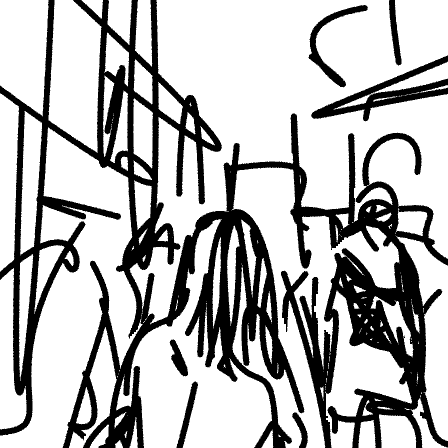}} &
    \frame{\includegraphics[width=0.098\textwidth,height=0.098\textwidth]{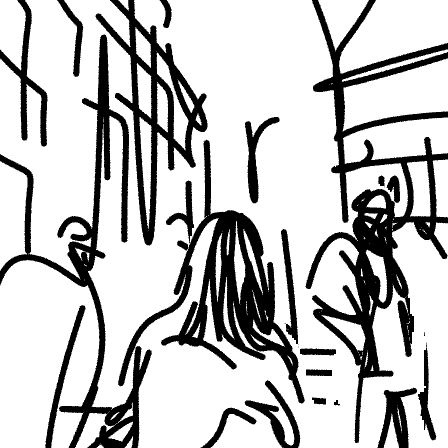}} &
    \frame{\includegraphics[width=0.098\textwidth,height=0.098\textwidth]{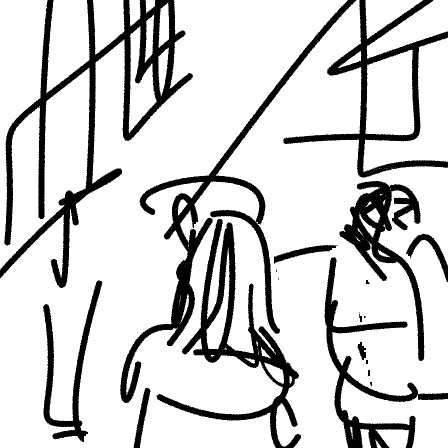}} &
    \hspace{0.5cm}
    \frame{\includegraphics[width=0.098\textwidth,height=0.098\textwidth]{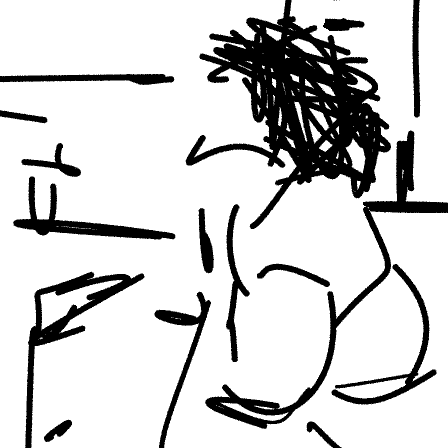}} &
    \frame{\includegraphics[width=0.098\textwidth,height=0.098\textwidth]{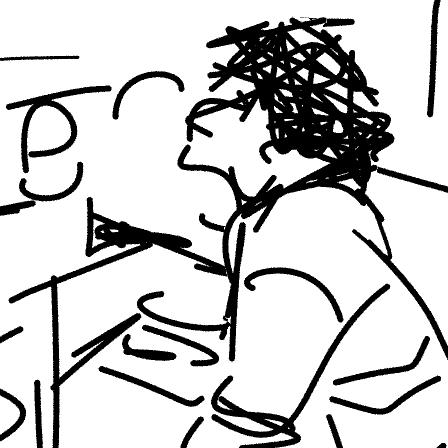}} &
    \frame{\includegraphics[width=0.098\textwidth,height=0.098\textwidth]{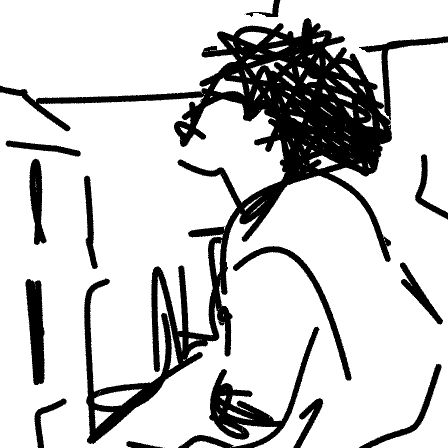}} &
    \frame{\includegraphics[width=0.098\textwidth,height=0.098\textwidth]{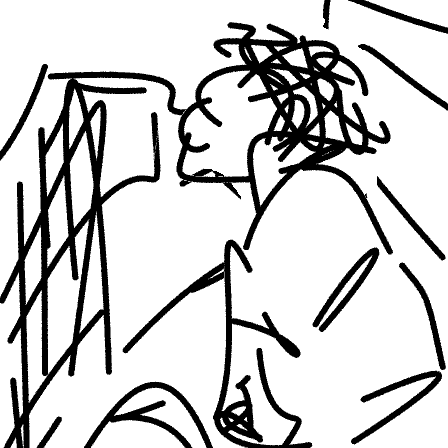}} \\
    
    \frame{\includegraphics[width=0.098\textwidth,height=0.098\textwidth]{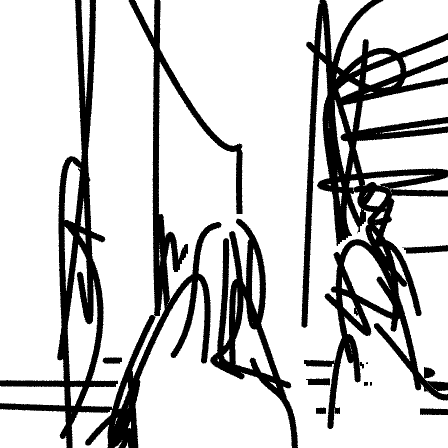}} &
    \frame{\includegraphics[width=0.098\textwidth,height=0.098\textwidth]{figs/matrices_black/woman_city_row2col1_black.png}} &
    \frame{\includegraphics[width=0.098\textwidth,height=0.098\textwidth]{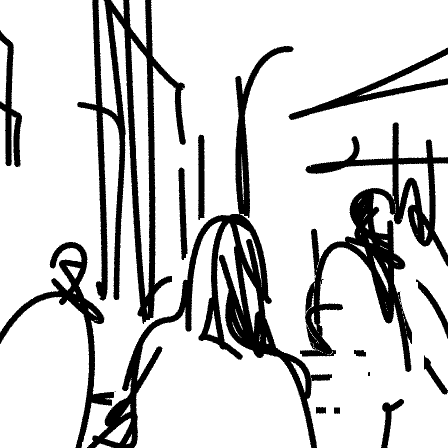}} &
    \frame{\includegraphics[width=0.098\textwidth,height=0.098\textwidth]{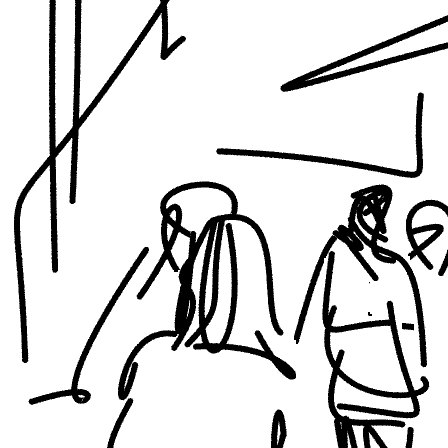}} &
    \hspace{0.5cm}
    \frame{\includegraphics[width=0.098\textwidth,height=0.098\textwidth]{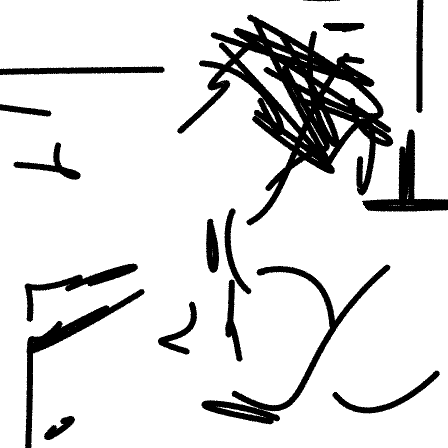}} &
    \frame{\includegraphics[width=0.098\textwidth,height=0.098\textwidth]{figs/matrices_black/black_woman_6.png}} &
    \frame{\includegraphics[width=0.098\textwidth,height=0.098\textwidth]{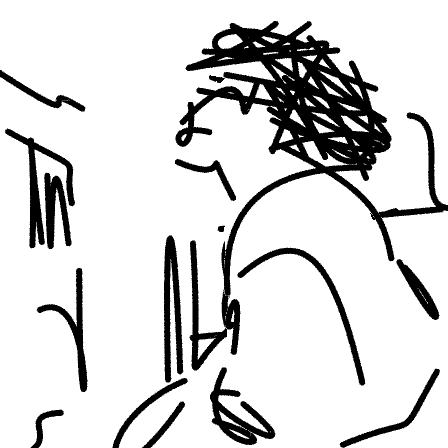}} &
    \frame{\includegraphics[width=0.098\textwidth,height=0.098\textwidth]{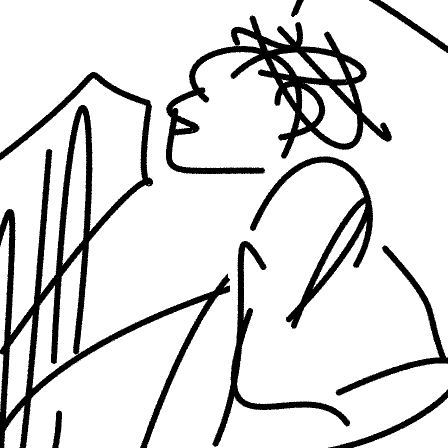}} \\
    
    \frame{\includegraphics[width=0.098\textwidth,height=0.098\textwidth]{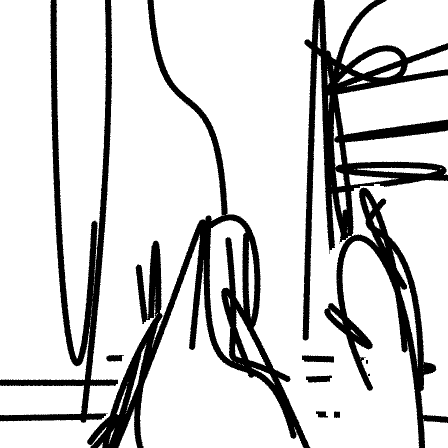}} &
    \frame{\includegraphics[width=0.098\textwidth,height=0.098\textwidth]{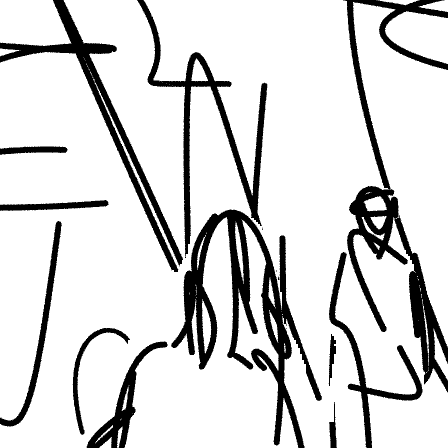}} &
    \frame{\includegraphics[width=0.098\textwidth,height=0.098\textwidth]{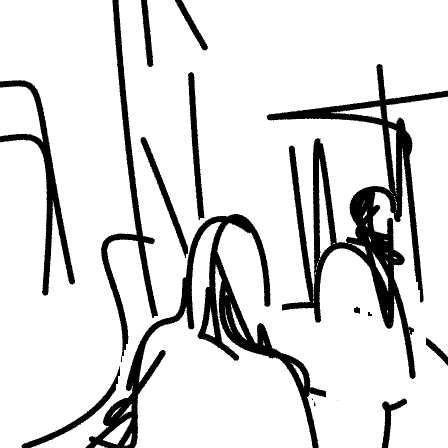}} &
    \frame{\includegraphics[width=0.098\textwidth,height=0.098\textwidth]{figs/matrices_black/woman_city_row3col3_black.png}} &
    \hspace{0.5cm}
    \frame{\includegraphics[width=0.098\textwidth,height=0.098\textwidth]{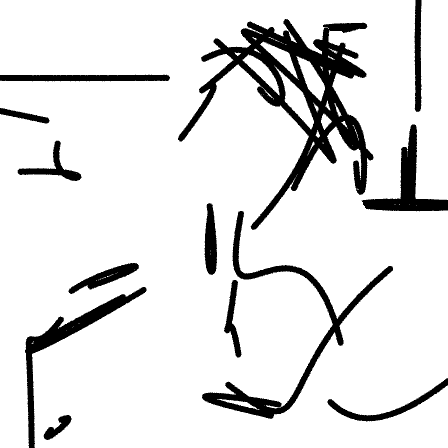}} &
    \frame{\includegraphics[width=0.098\textwidth,height=0.098\textwidth]{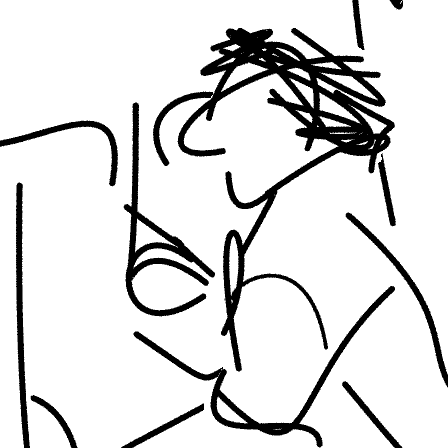}} &
    \frame{\includegraphics[width=0.098\textwidth,height=0.098\textwidth]{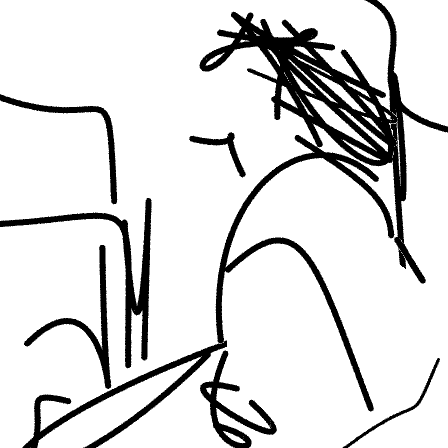}} &
    \frame{\includegraphics[width=0.098\textwidth,height=0.098\textwidth]{figs/matrices_black/black_woman_15.png}} \\

    \\
    \\
    
    \includegraphics[width=0.098\textwidth,height=0.098\textwidth]{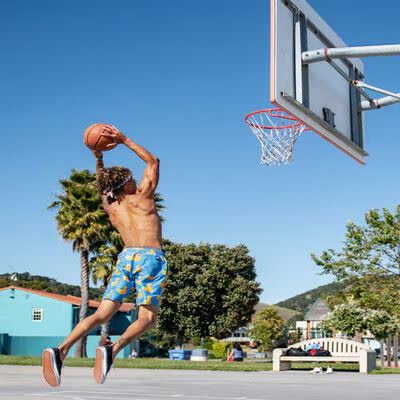} & & & &
    \hspace{0.5cm}
    \includegraphics[width=0.098\textwidth,height=0.098\textwidth]{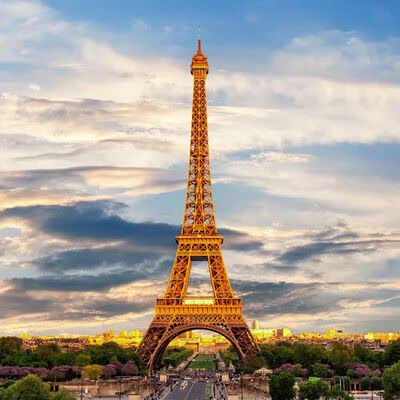} & & & \\

    \frame{\includegraphics[width=0.098\textwidth,height=0.098\textwidth]{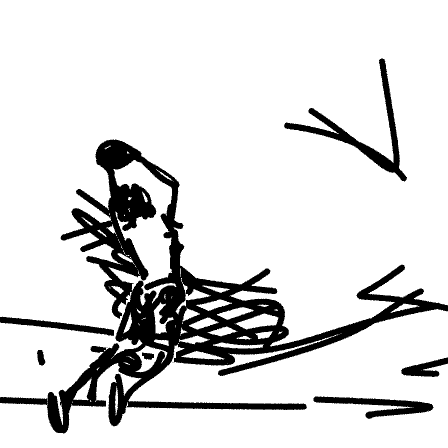}} &
    \frame{\includegraphics[width=0.098\textwidth,height=0.098\textwidth]{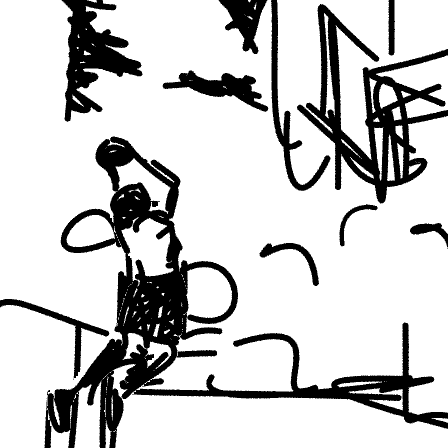}} &
    \frame{\includegraphics[width=0.098\textwidth,height=0.098\textwidth]{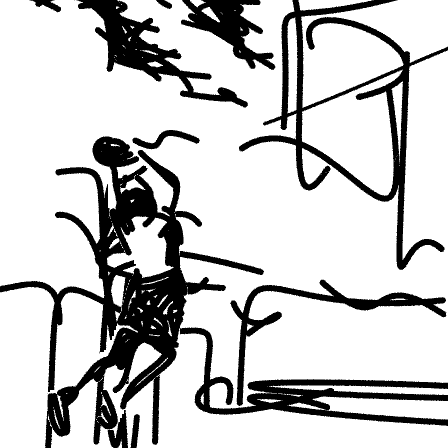}} &
    \frame{\includegraphics[width=0.098\textwidth,height=0.098\textwidth]{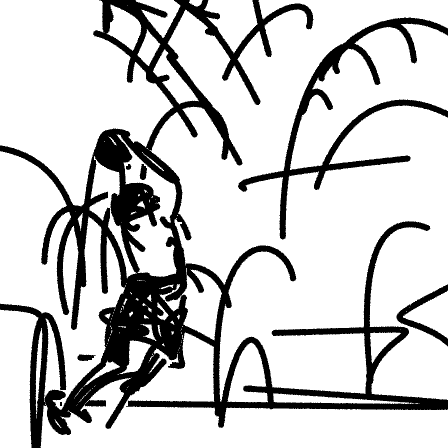}} &
    \hspace{0.5cm}
    \frame{\includegraphics[width=0.098\textwidth,height=0.098\textwidth]{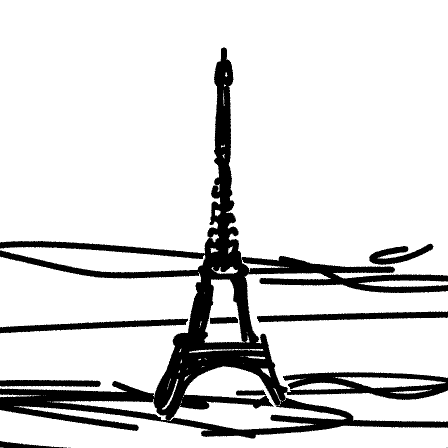}} &
    \frame{\includegraphics[width=0.098\textwidth,height=0.098\textwidth]{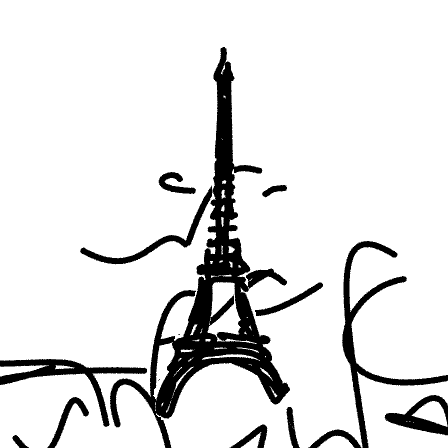}} &
    \frame{\includegraphics[width=0.098\textwidth,height=0.098\textwidth]{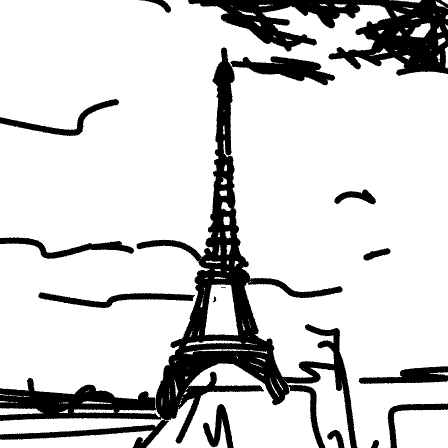}} &
    \frame{\includegraphics[width=0.098\textwidth,height=0.098\textwidth]{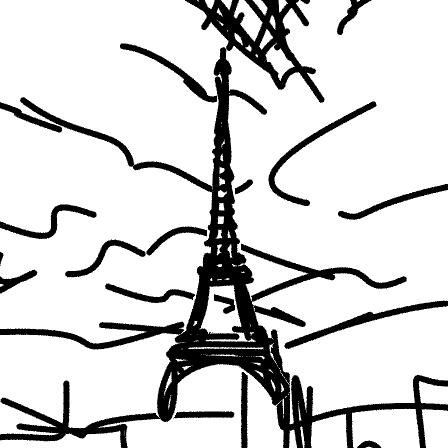}} \\
    
    \frame{\includegraphics[width=0.098\textwidth,height=0.098\textwidth]{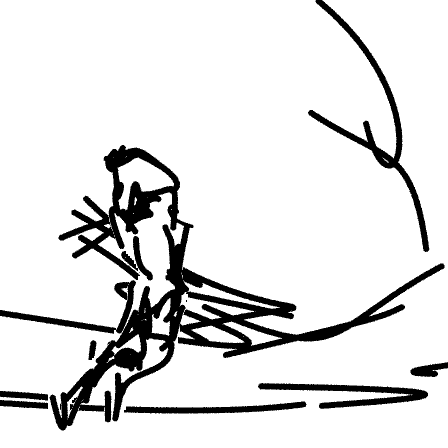}} &
    \frame{\includegraphics[width=0.098\textwidth,height=0.098\textwidth]{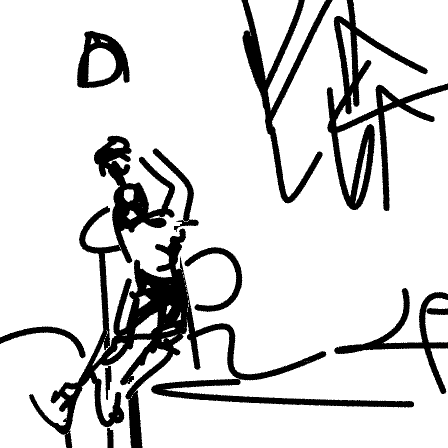}} &
    \frame{\includegraphics[width=0.098\textwidth,height=0.098\textwidth]{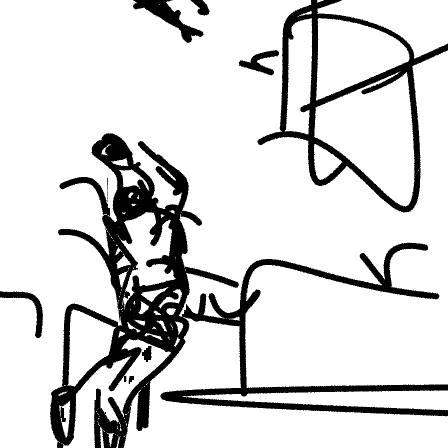}} &
    \frame{\includegraphics[width=0.098\textwidth,height=0.098\textwidth]{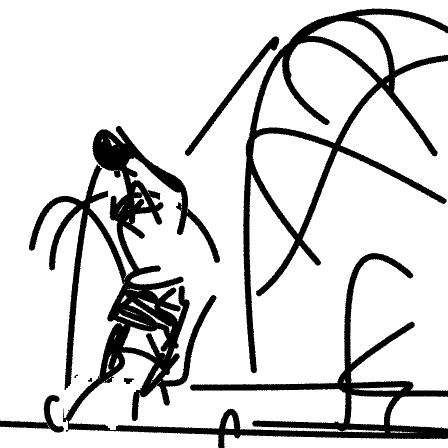}} &
    \hspace{0.5cm}
    \frame{\includegraphics[width=0.098\textwidth,height=0.098\textwidth]{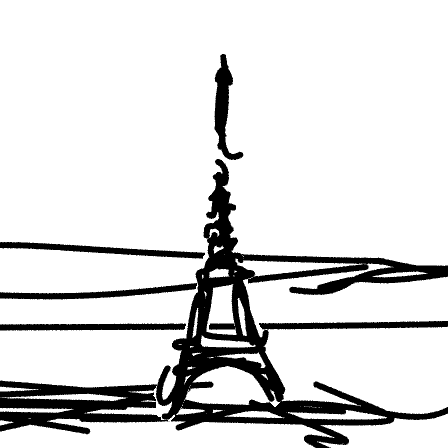}} &
    \frame{\includegraphics[width=0.098\textwidth,height=0.098\textwidth]{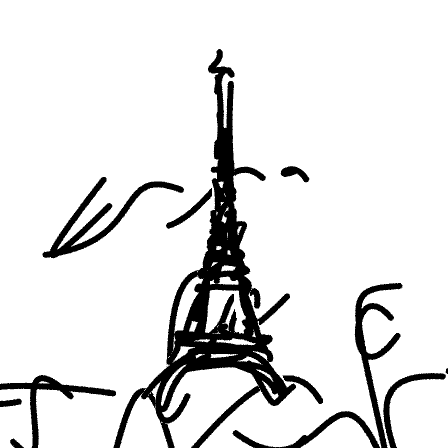}} &
    \frame{\includegraphics[width=0.098\textwidth,height=0.098\textwidth]{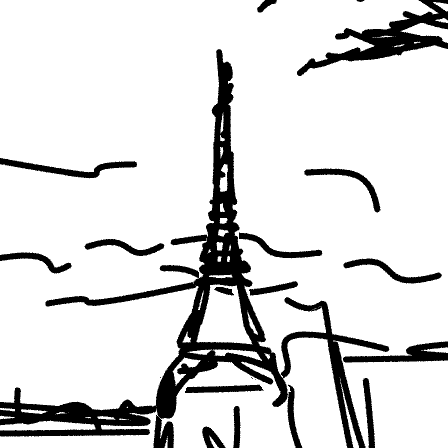}} &
    \frame{\includegraphics[width=0.098\textwidth,height=0.098\textwidth]{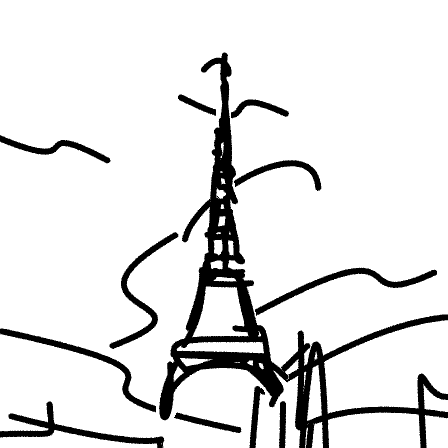}} \\
    
    \frame{\includegraphics[width=0.098\textwidth,height=0.098\textwidth]{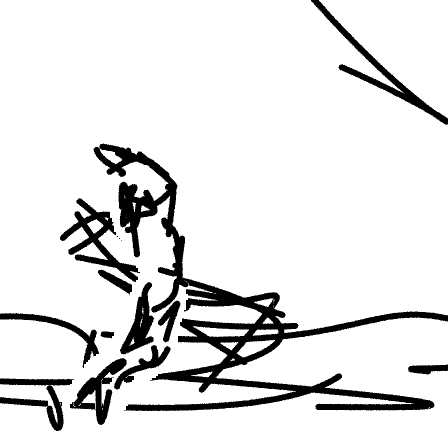}} &
    \frame{\includegraphics[width=0.098\textwidth,height=0.098\textwidth]{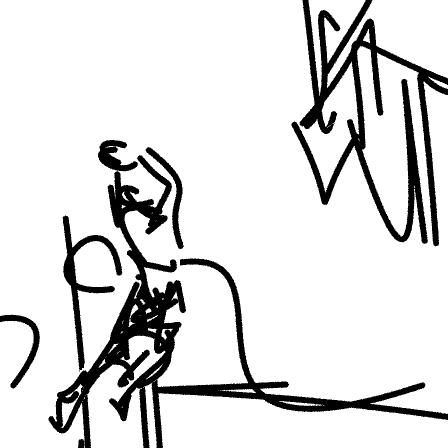}} &
    \frame{\includegraphics[width=0.098\textwidth,height=0.098\textwidth]{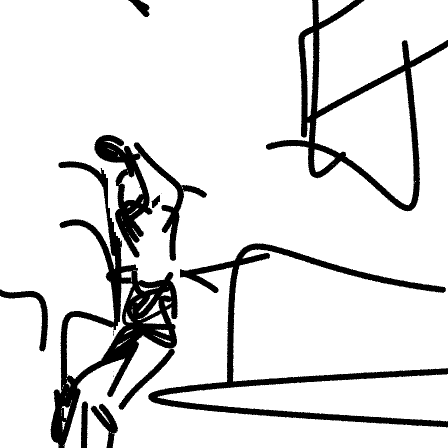}} &
    \frame{\includegraphics[width=0.098\textwidth,height=0.098\textwidth]{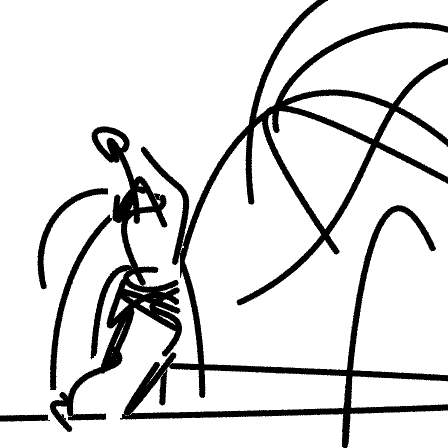}} &
    \hspace{0.5cm}
    \frame{\includegraphics[width=0.098\textwidth,height=0.098\textwidth]{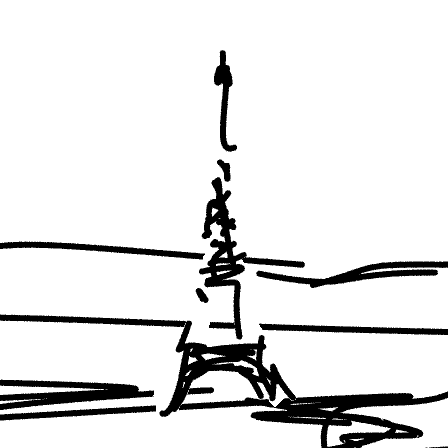}} &
    \frame{\includegraphics[width=0.098\textwidth,height=0.098\textwidth]{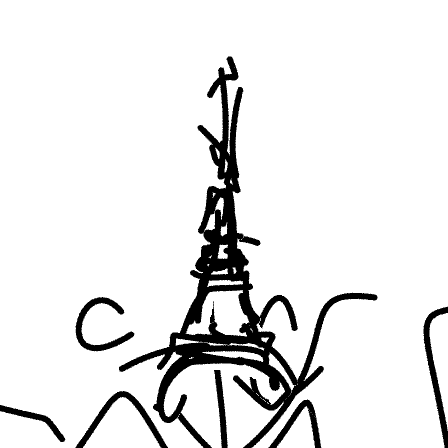}} &
    \frame{\includegraphics[width=0.098\textwidth,height=0.098\textwidth]{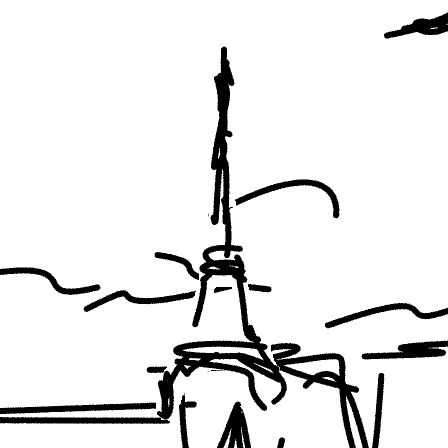}} &
    \frame{\includegraphics[width=0.098\textwidth,height=0.098\textwidth]{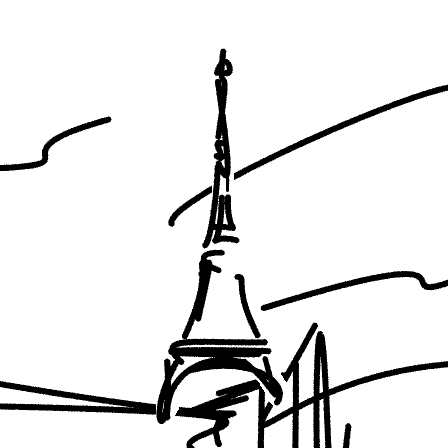}} \\
    
    \frame{\includegraphics[width=0.098\textwidth,height=0.098\textwidth]{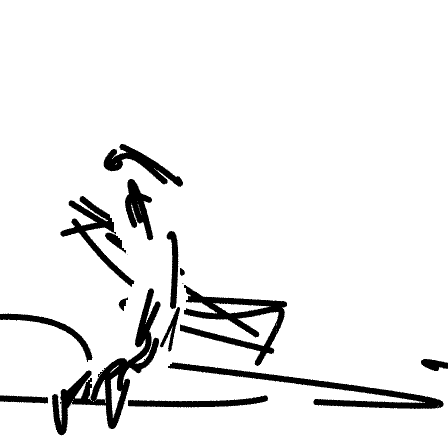}} &
    \frame{\includegraphics[width=0.098\textwidth,height=0.098\textwidth]{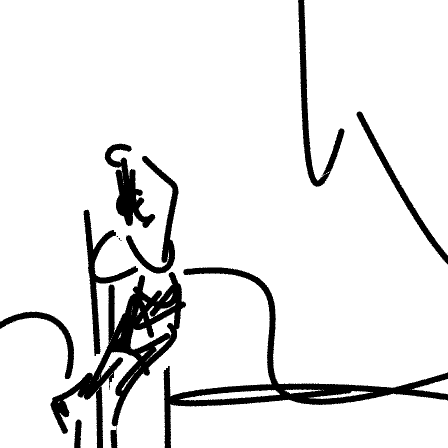}} &
    \frame{\includegraphics[width=0.098\textwidth,height=0.098\textwidth]{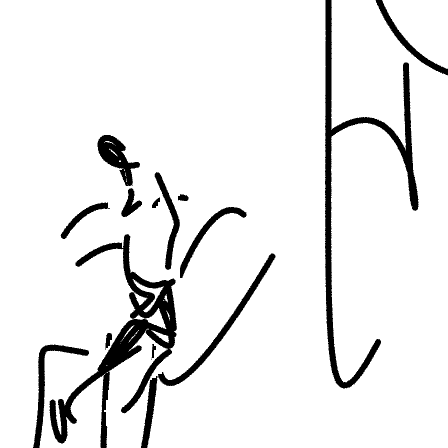}} &
    \frame{\includegraphics[width=0.098\textwidth,height=0.098\textwidth]{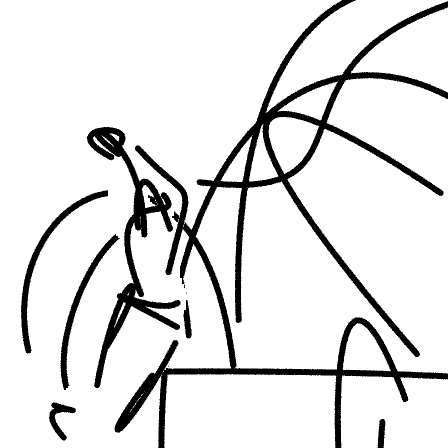}} &
    \hspace{0.5cm}
    \frame{\includegraphics[width=0.098\textwidth,height=0.098\textwidth]{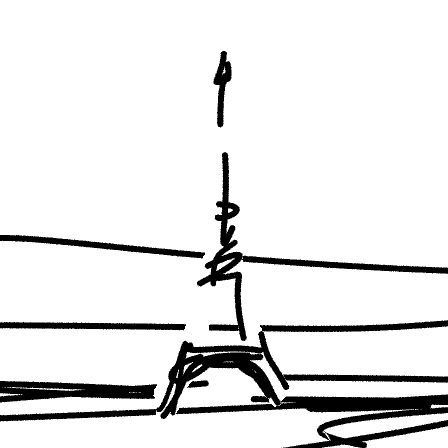}} &
    \frame{\includegraphics[width=0.098\textwidth,height=0.098\textwidth]{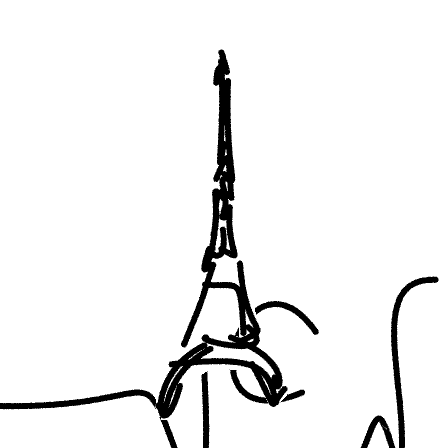}} &
    \frame{\includegraphics[width=0.098\textwidth,height=0.098\textwidth]{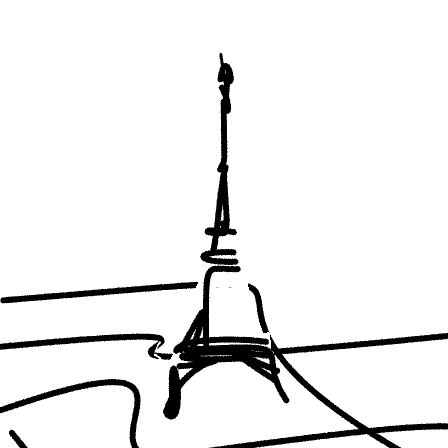}} &
    \frame{\includegraphics[width=0.098\textwidth,height=0.098\textwidth]{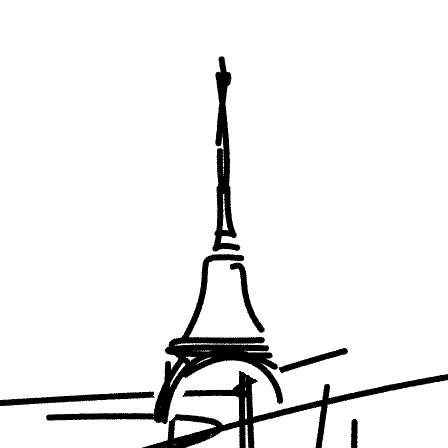}} \\

    \end{tabular}
    \caption{The $4\times4$ matrix of sketches produced by our method. Columns from left to right illustrate the change in fidelity, from precise to loose, and rows from top to bottom illustrate the visual simplification.}
    
    \label{fig:matrix3}
\end{figure*}

%% file: files/figures/supplementary/our_matrices/our_matrices_4.tex
\begin{figure*}
    \centering
    
    \begin{tabular}{c c c c c c c c}

    \includegraphics[width=0.098\textwidth,height=0.098\textwidth]{figs/inputs/camera-2.jpg} & & & &
    \hspace{0.5cm}
    \includegraphics[width=0.098\textwidth,height=0.098\textwidth]{figs/inputs/traffic_light.jpg} & & & \\
    
    \frame{\includegraphics[width=0.098\textwidth,height=0.098\textwidth]{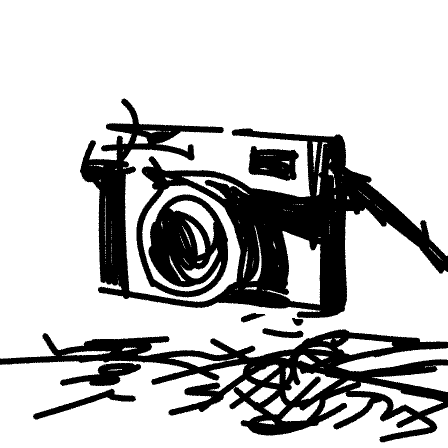}} &
    \frame{\includegraphics[width=0.098\textwidth,height=0.098\textwidth]{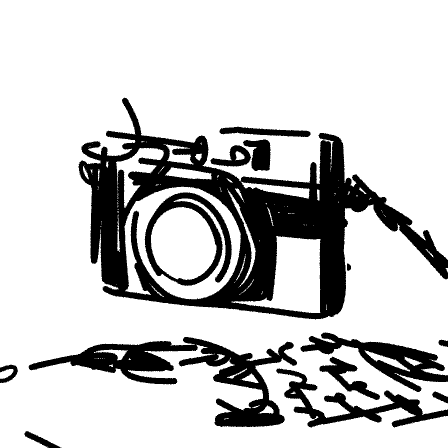}} &
    \frame{\includegraphics[width=0.098\textwidth,height=0.098\textwidth]{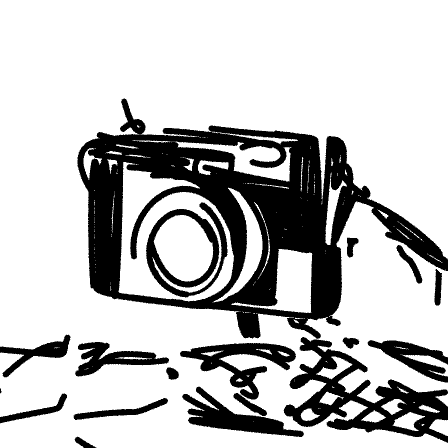}} &
    \frame{\includegraphics[width=0.098\textwidth,height=0.098\textwidth]{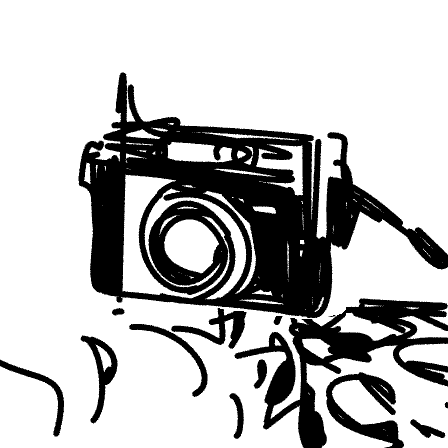}} &
    \hspace{0.5cm}
    \frame{\includegraphics[width=0.098\textwidth,height=0.098\textwidth]{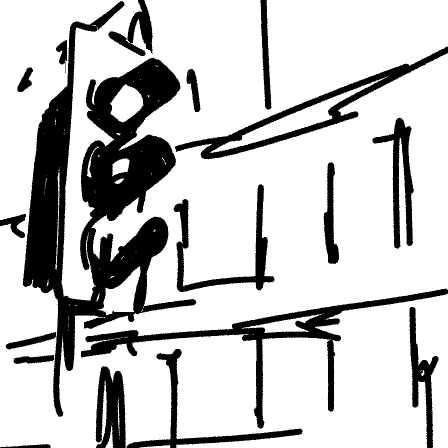}} &
    \frame{\includegraphics[width=0.098\textwidth,height=0.098\textwidth]{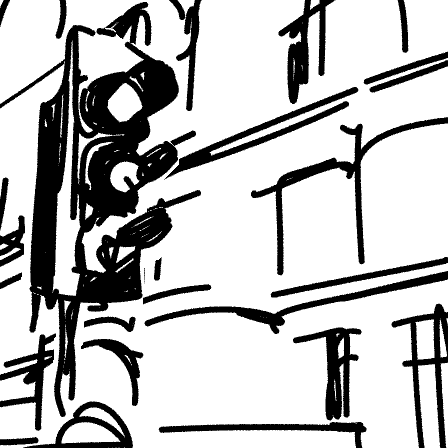}} &
    \frame{\includegraphics[width=0.098\textwidth,height=0.098\textwidth]{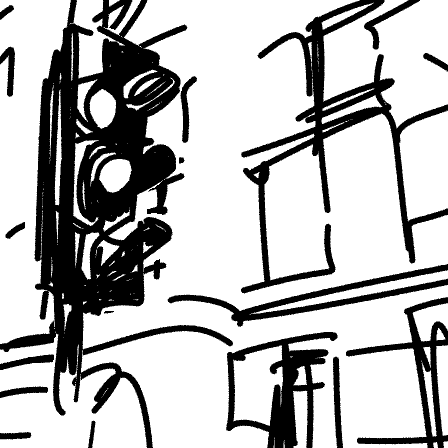}} &
    \frame{\includegraphics[width=0.098\textwidth,height=0.098\textwidth]{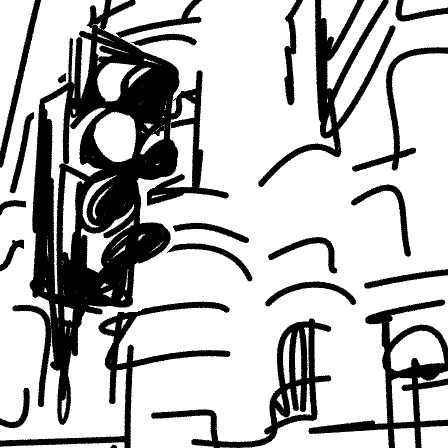}} \\
    
    \frame{\includegraphics[width=0.098\textwidth,height=0.098\textwidth]{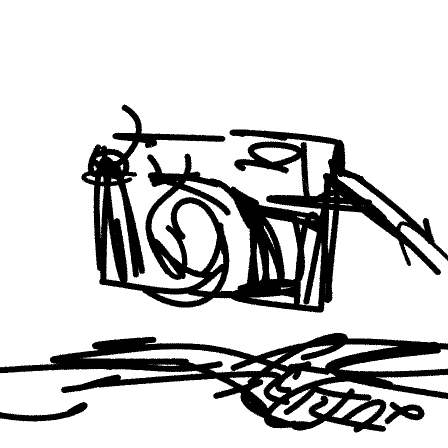}} &
    \frame{\includegraphics[width=0.098\textwidth,height=0.098\textwidth]{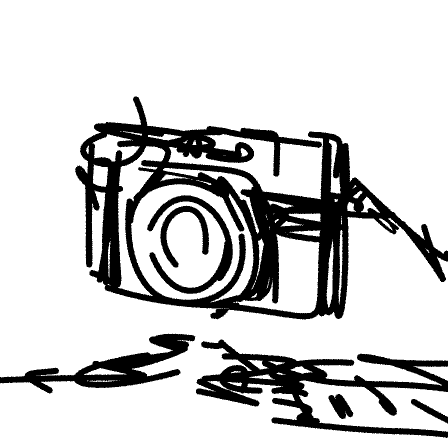}} &
    \frame{\includegraphics[width=0.098\textwidth,height=0.098\textwidth]{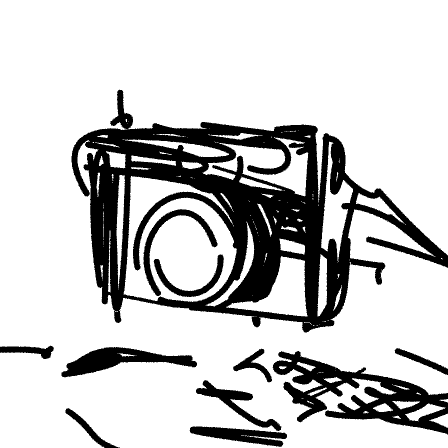}} &
    \frame{\includegraphics[width=0.098\textwidth,height=0.098\textwidth]{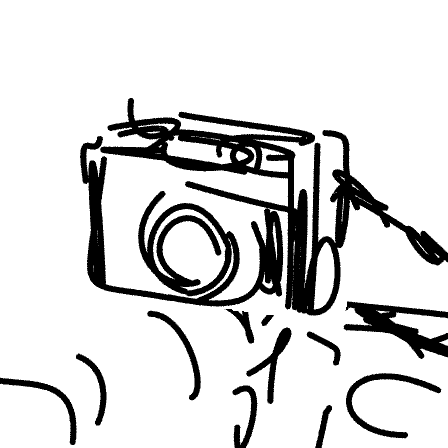}} &
    \hspace{0.5cm}
    \frame{\includegraphics[width=0.098\textwidth,height=0.098\textwidth]{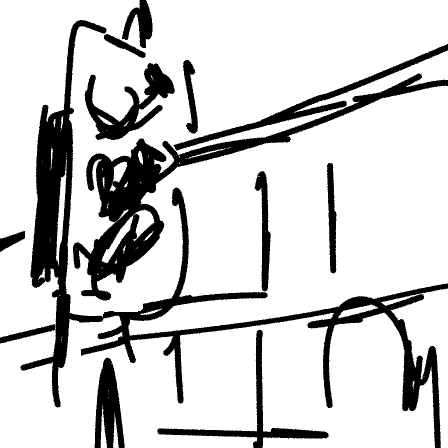}} &
    \frame{\includegraphics[width=0.098\textwidth,height=0.098\textwidth]{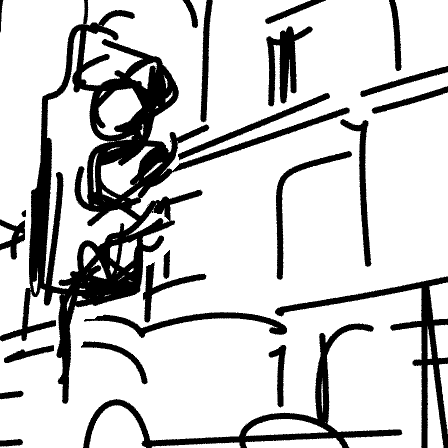}} &
    \frame{\includegraphics[width=0.098\textwidth,height=0.098\textwidth]{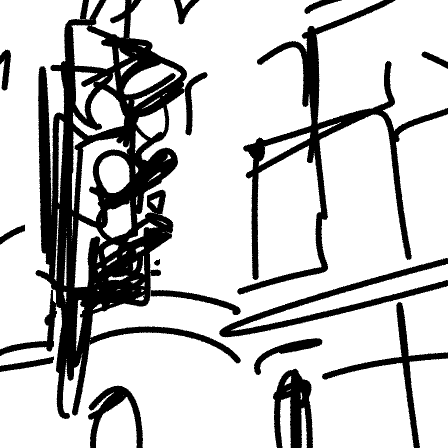}} &
    \frame{\includegraphics[width=0.098\textwidth,height=0.098\textwidth]{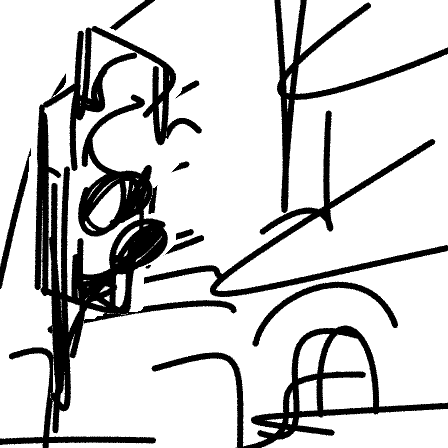}} \\
    
    \frame{\includegraphics[width=0.098\textwidth,height=0.098\textwidth]{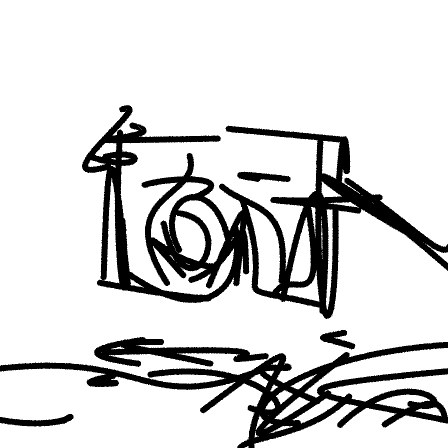}} &
    \frame{\includegraphics[width=0.098\textwidth,height=0.098\textwidth]{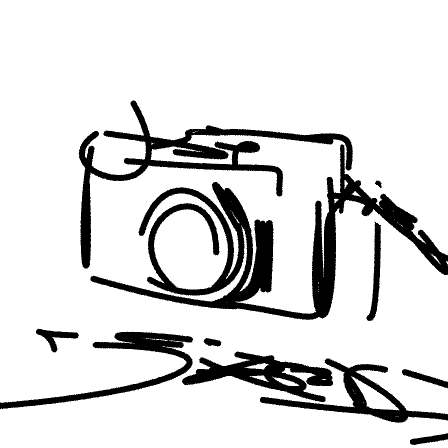}} &
    \frame{\includegraphics[width=0.098\textwidth,height=0.098\textwidth]{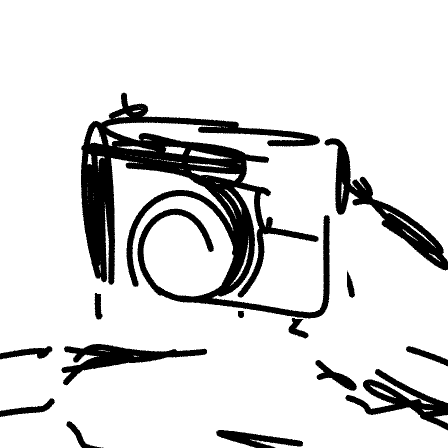}} &
    \frame{\includegraphics[width=0.098\textwidth,height=0.098\textwidth]{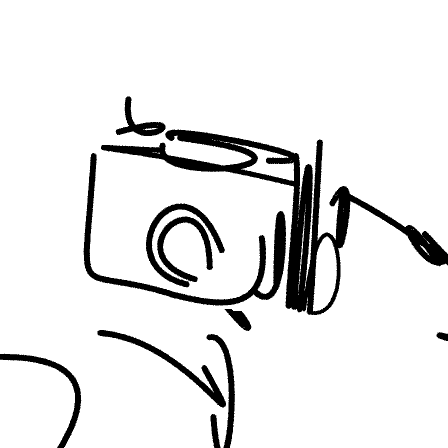}} &
    \hspace{0.5cm}
    \frame{\includegraphics[width=0.098\textwidth,height=0.098\textwidth]{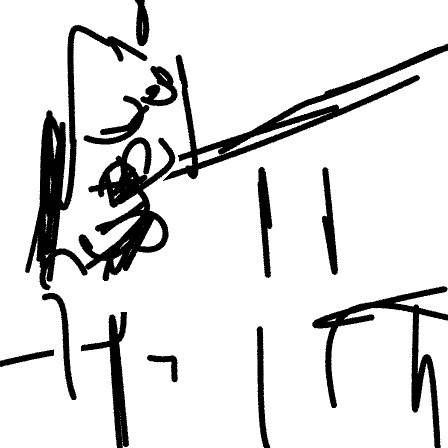}} &
    \frame{\includegraphics[width=0.098\textwidth,height=0.098\textwidth]{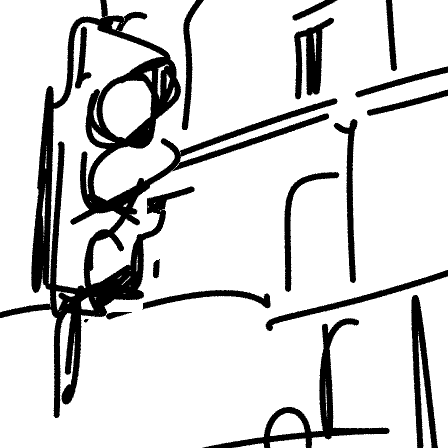}} &
    \frame{\includegraphics[width=0.098\textwidth,height=0.098\textwidth]{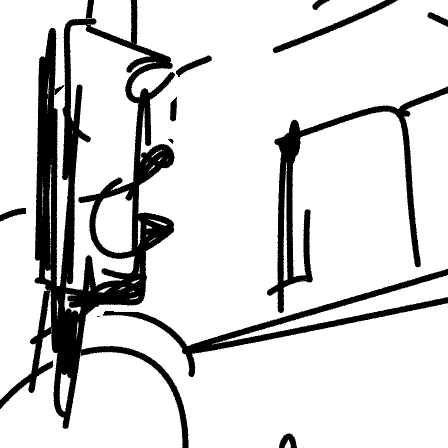}} &
    \frame{\includegraphics[width=0.098\textwidth,height=0.098\textwidth]{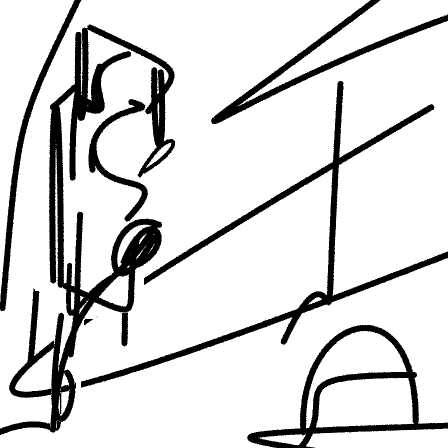}} \\
    
    \frame{\includegraphics[width=0.098\textwidth,height=0.098\textwidth]{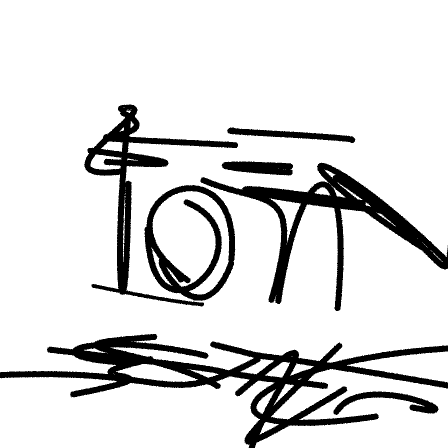}} &
    \frame{\includegraphics[width=0.098\textwidth,height=0.098\textwidth]{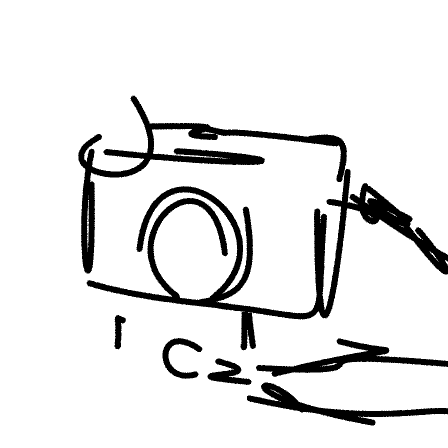}} &
    \frame{\includegraphics[width=0.098\textwidth,height=0.098\textwidth]{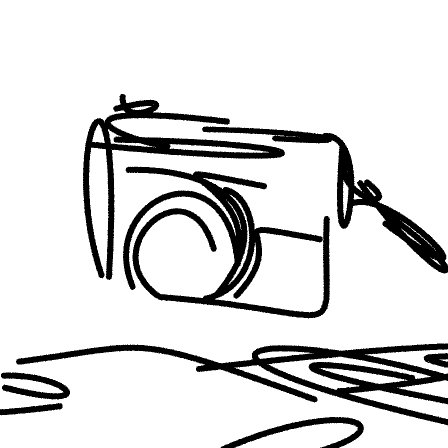}} &
    \frame{\includegraphics[width=0.098\textwidth,height=0.098\textwidth]{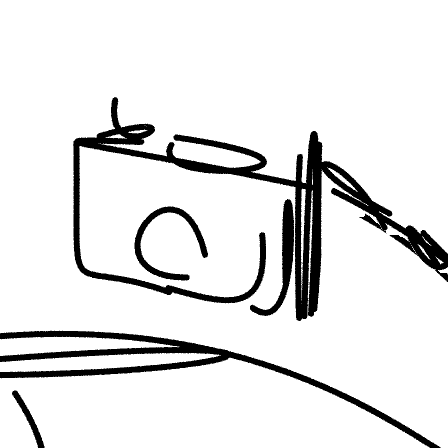}} &
    \hspace{0.5cm}
    \frame{\includegraphics[width=0.098\textwidth,height=0.098\textwidth]{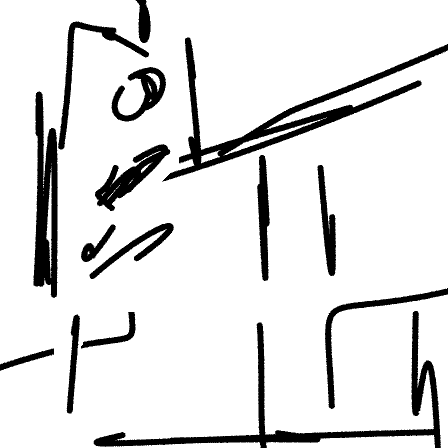}} &
    \frame{\includegraphics[width=0.098\textwidth,height=0.098\textwidth]{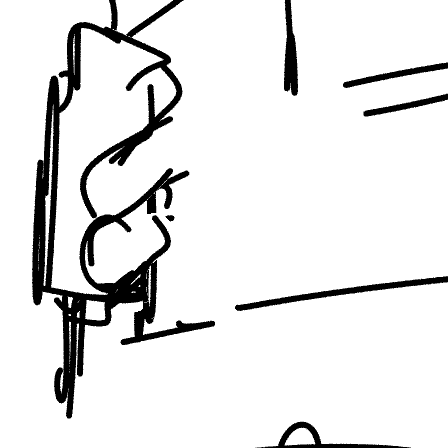}} &
    \frame{\includegraphics[width=0.098\textwidth,height=0.098\textwidth]{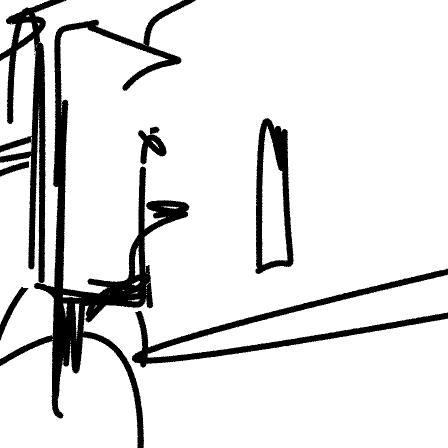}} &
    \frame{\includegraphics[width=0.098\textwidth,height=0.098\textwidth]{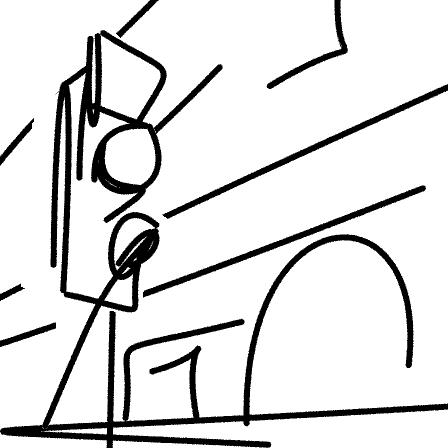}} \\

    \\
    \\
    
    \includegraphics[width=0.098\textwidth,height=0.098\textwidth]{figs/inputs/boat-2.jpg} & & & &
    \hspace{0.5cm}
    \includegraphics[width=0.098\textwidth,height=0.098\textwidth]{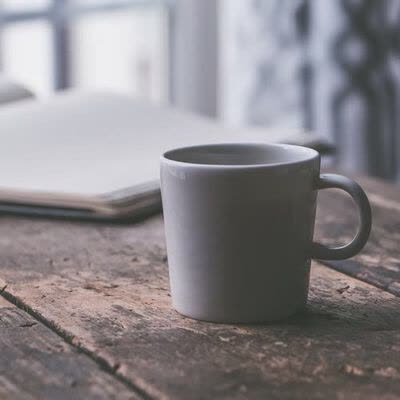} & & & \\

    \frame{\includegraphics[width=0.098\textwidth,height=0.098\textwidth]{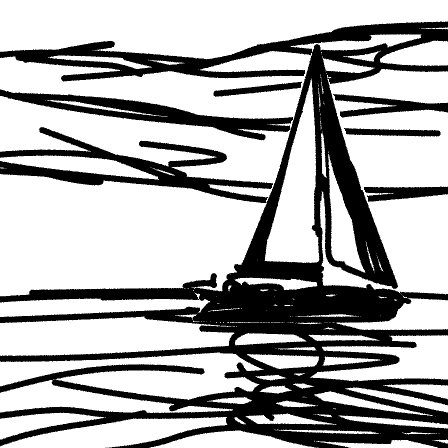}} &
    \frame{\includegraphics[width=0.098\textwidth,height=0.098\textwidth]{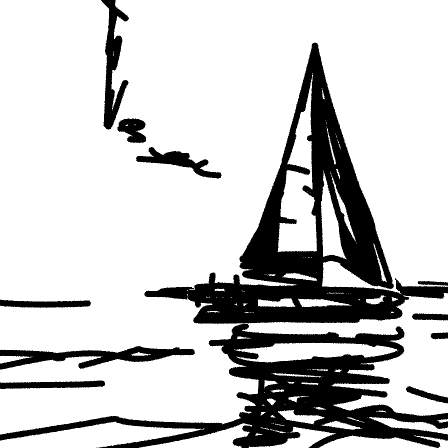}} &
    \frame{\includegraphics[width=0.098\textwidth,height=0.098\textwidth]{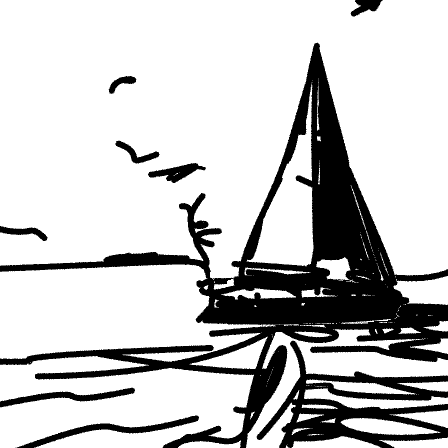}} &
    \frame{\includegraphics[width=0.098\textwidth,height=0.098\textwidth]{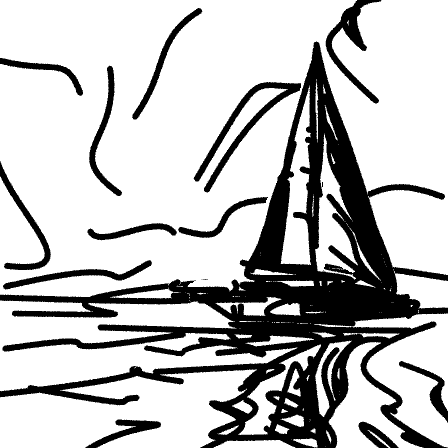}} &
    \hspace{0.5cm}
    \frame{\includegraphics[width=0.098\textwidth,height=0.098\textwidth]{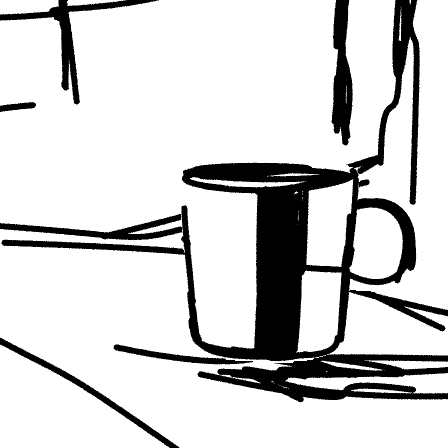}} &
    \frame{\includegraphics[width=0.098\textwidth,height=0.098\textwidth]{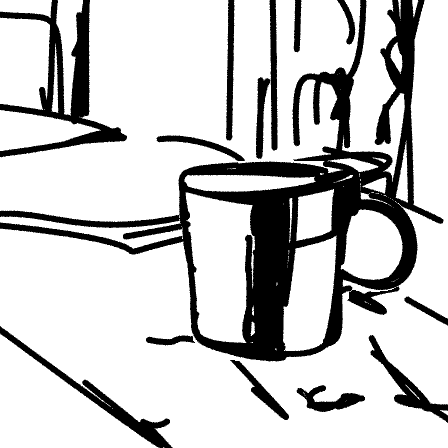}} &
    \frame{\includegraphics[width=0.098\textwidth,height=0.098\textwidth]{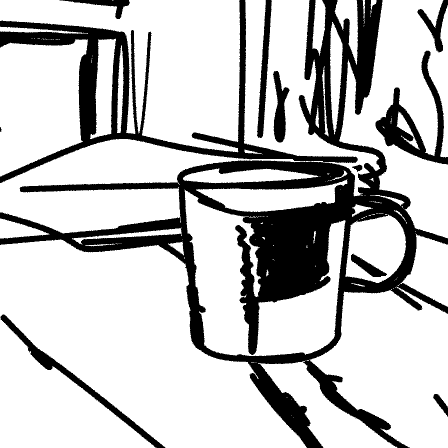}} &
    \frame{\includegraphics[width=0.098\textwidth,height=0.098\textwidth]{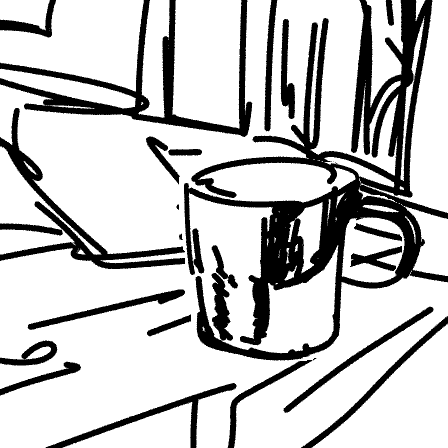}} \\
    
    \frame{\includegraphics[width=0.098\textwidth,height=0.098\textwidth]{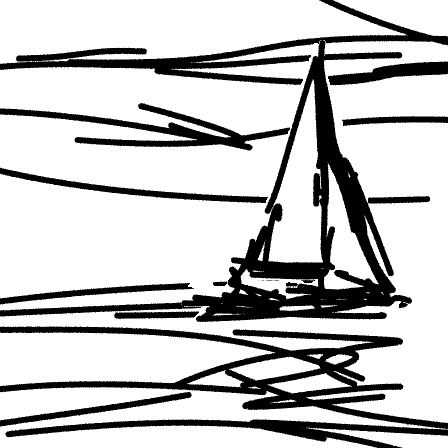}} &
    \frame{\includegraphics[width=0.098\textwidth,height=0.098\textwidth]{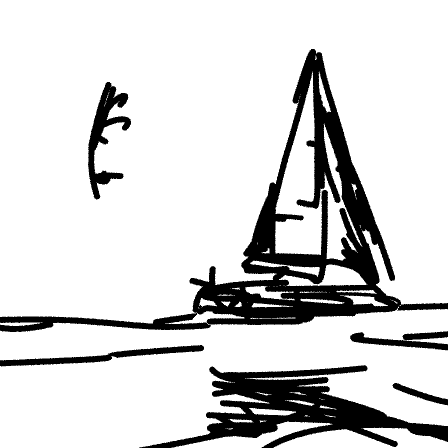}} &
    \frame{\includegraphics[width=0.098\textwidth,height=0.098\textwidth]{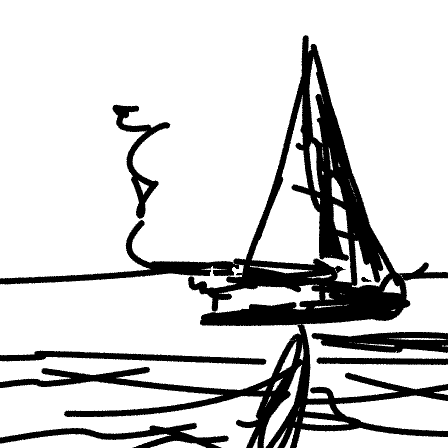}} &
    \frame{\includegraphics[width=0.098\textwidth,height=0.098\textwidth]{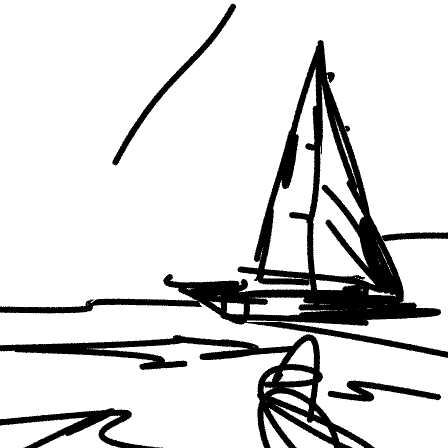}} &
    \hspace{0.5cm}
    \frame{\includegraphics[width=0.098\textwidth,height=0.098\textwidth]{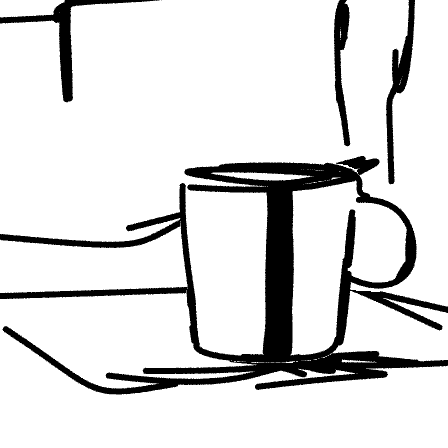}} &
    \frame{\includegraphics[width=0.098\textwidth,height=0.098\textwidth]{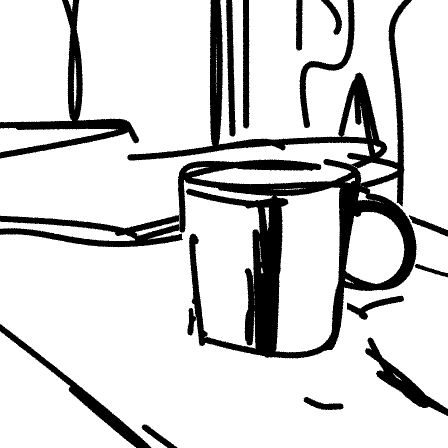}} &
    \frame{\includegraphics[width=0.098\textwidth,height=0.098\textwidth]{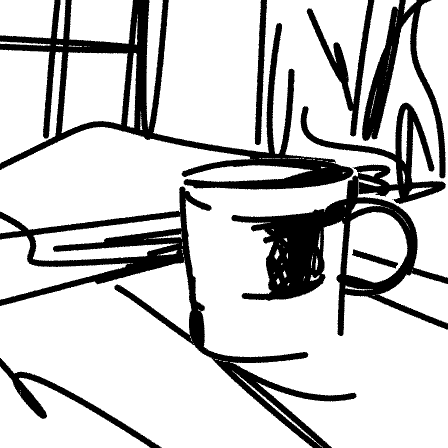}} &
    \frame{\includegraphics[width=0.098\textwidth,height=0.098\textwidth]{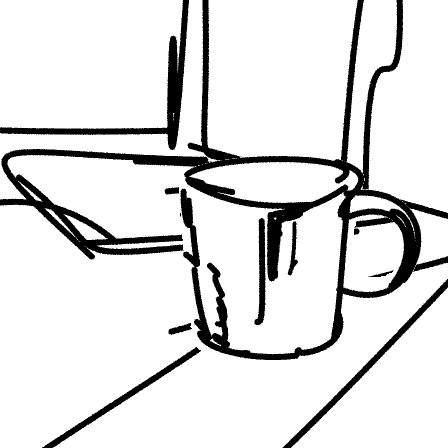}} \\
    
    \frame{\includegraphics[width=0.098\textwidth,height=0.098\textwidth]{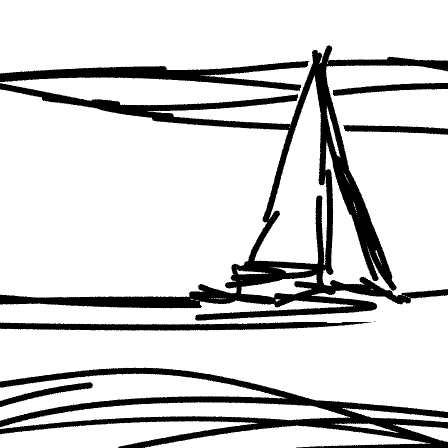}} &
    \frame{\includegraphics[width=0.098\textwidth,height=0.098\textwidth]{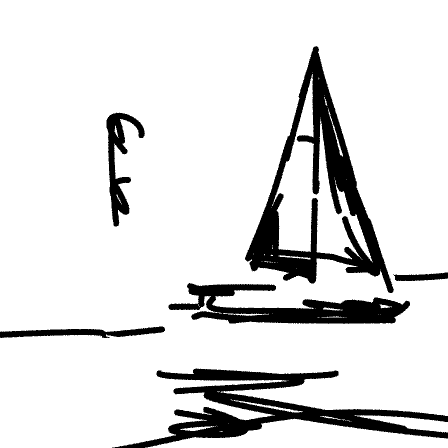}} &
    \frame{\includegraphics[width=0.098\textwidth,height=0.098\textwidth]{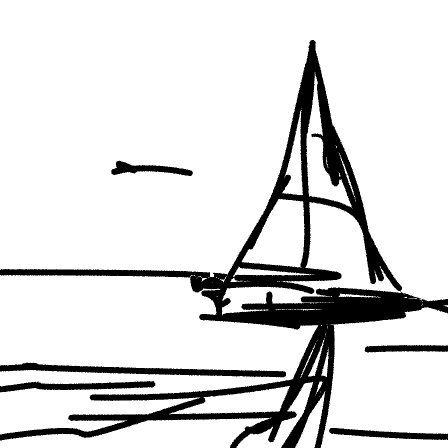}} &
    \frame{\includegraphics[width=0.098\textwidth,height=0.098\textwidth]{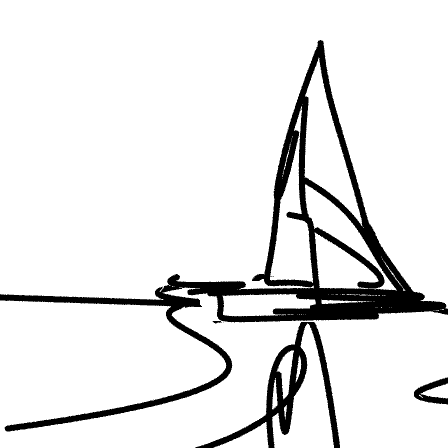}} &
    \hspace{0.5cm}
    \frame{\includegraphics[width=0.098\textwidth,height=0.098\textwidth]{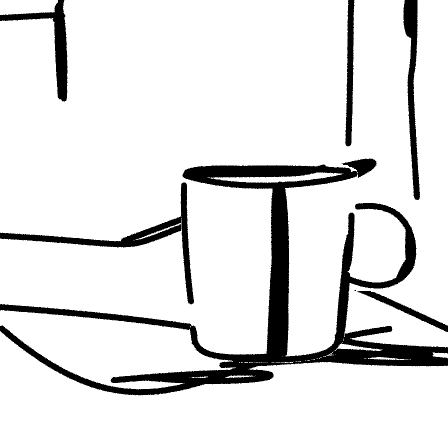}} &
    \frame{\includegraphics[width=0.098\textwidth,height=0.098\textwidth]{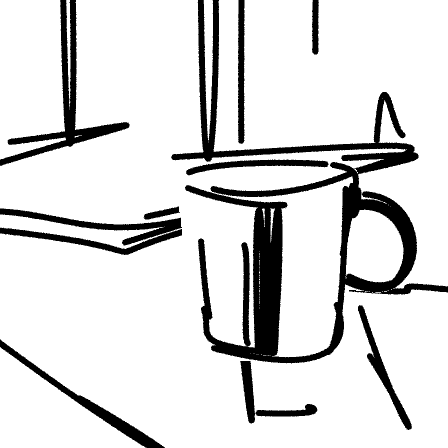}} &
    \frame{\includegraphics[width=0.098\textwidth,height=0.098\textwidth]{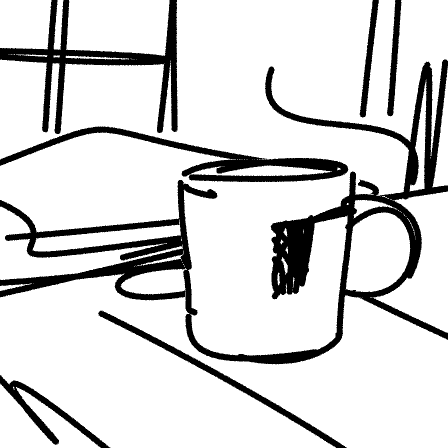}} &
    \frame{\includegraphics[width=0.098\textwidth,height=0.098\textwidth]{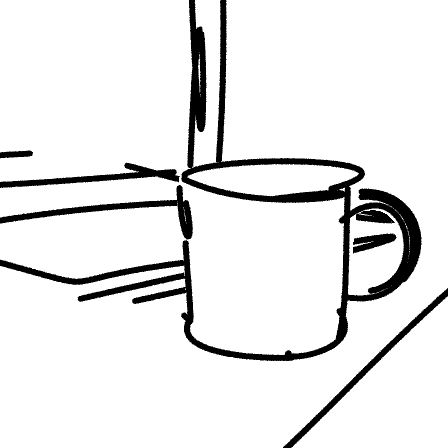}} \\
    
    \frame{\includegraphics[width=0.098\textwidth,height=0.098\textwidth]{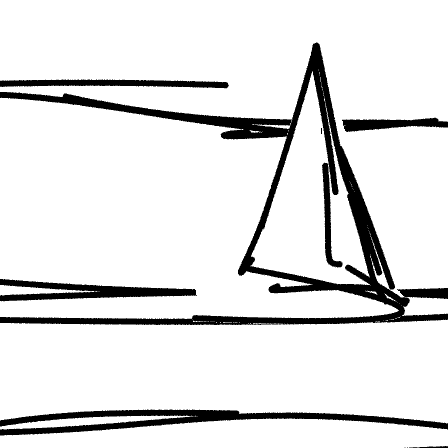}} &
    \frame{\includegraphics[width=0.098\textwidth,height=0.098\textwidth]{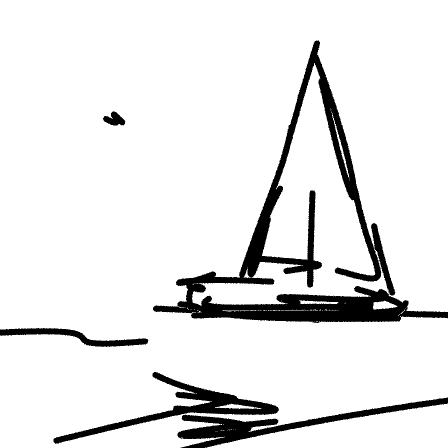}} &
    \frame{\includegraphics[width=0.098\textwidth,height=0.098\textwidth]{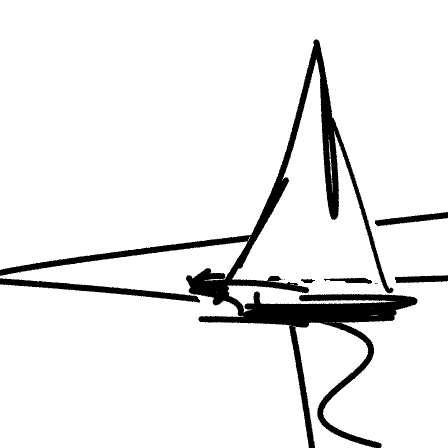}} &
    \frame{\includegraphics[width=0.098\textwidth,height=0.098\textwidth]{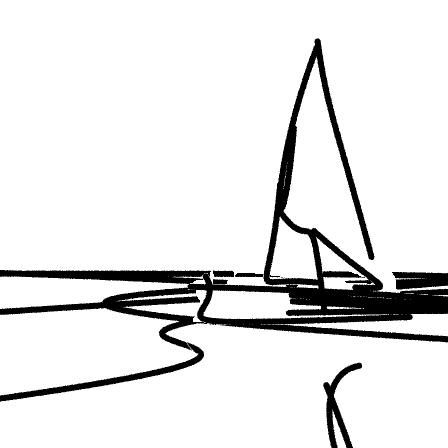}} &
    \hspace{0.5cm}
    \frame{\includegraphics[width=0.098\textwidth,height=0.098\textwidth]{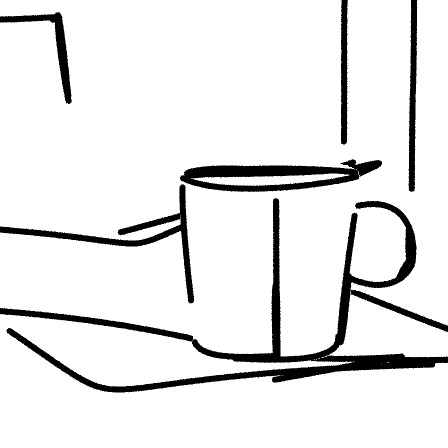}} &
    \frame{\includegraphics[width=0.098\textwidth,height=0.098\textwidth]{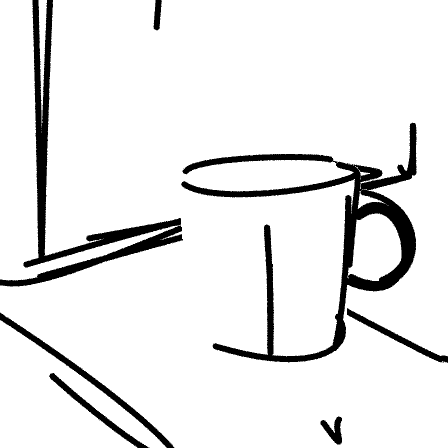}} &
    \frame{\includegraphics[width=0.098\textwidth,height=0.098\textwidth]{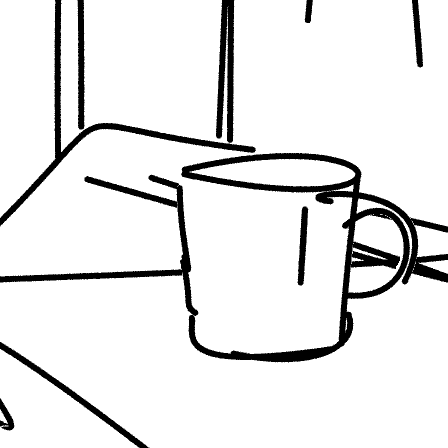}} &
    \frame{\includegraphics[width=0.098\textwidth,height=0.098\textwidth]{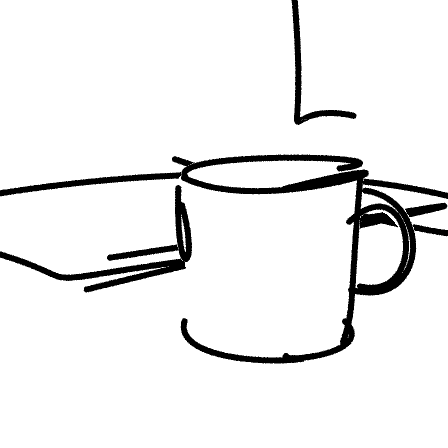}} \\

    \end{tabular}
    \caption{The $4\times4$ matrix of sketches produced by our method. Columns from left to right illustrate the change in fidelity, from precise to loose, and rows from top to bottom illustrate the visual simplification.}
    
    \label{fig:matrix4}
\end{figure*}

%% file: files/figures/supplementary/our_matrices/our_matrices_5.tex
\begin{figure*}
    \centering
    
    \begin{tabular}{c c c c c c c c}

    \includegraphics[width=0.098\textwidth,height=0.098\textwidth]{figs/inputs/giraffe.jpg} & & & &
    \hspace{0.5cm}
    \includegraphics[width=0.098\textwidth,height=0.098\textwidth]{figs/inputs/womanhome.jpg} & & & \\
    
    \frame{\includegraphics[width=0.098\textwidth,height=0.098\textwidth]{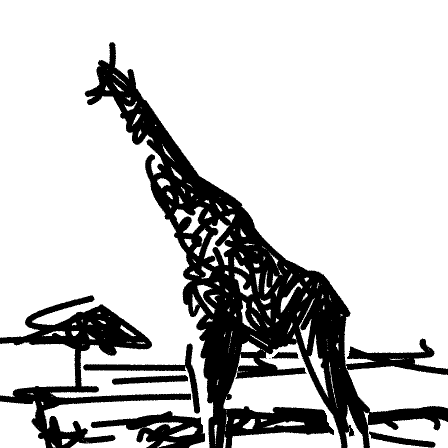}} &
    \frame{\includegraphics[width=0.098\textwidth,height=0.098\textwidth]{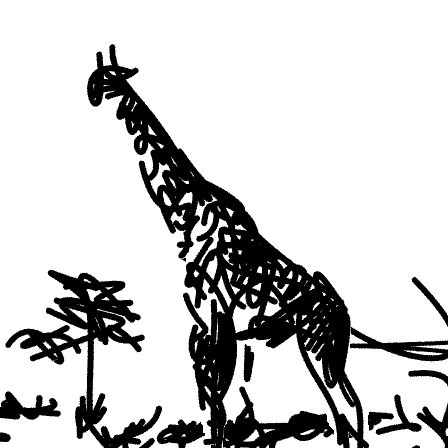}} &
    \frame{\includegraphics[width=0.098\textwidth,height=0.098\textwidth]{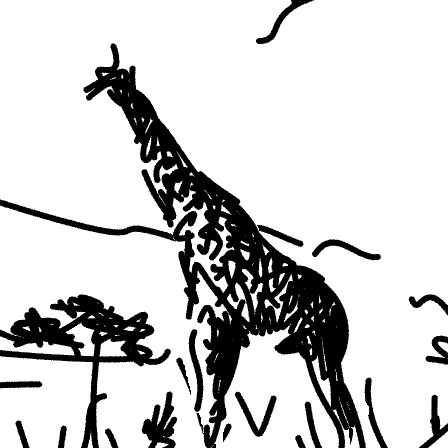}} &
    \frame{\includegraphics[width=0.098\textwidth,height=0.098\textwidth]{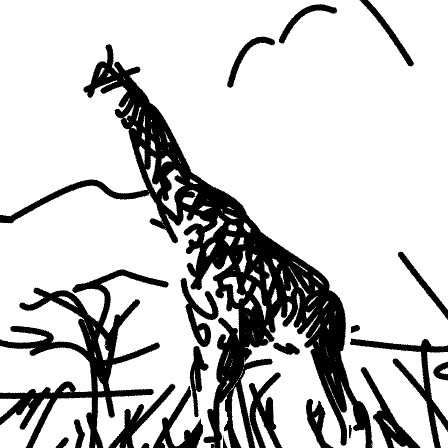}} &
    \hspace{0.5cm}
    \frame{\includegraphics[width=0.098\textwidth,height=0.098\textwidth]{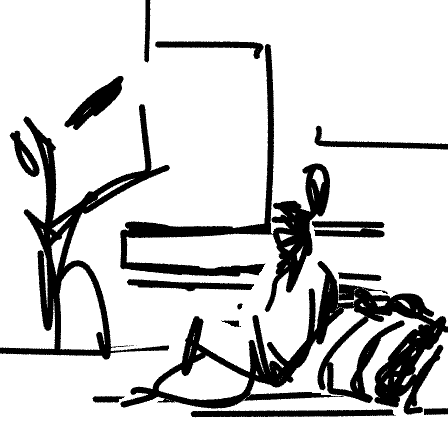}} &
    \frame{\includegraphics[width=0.098\textwidth,height=0.098\textwidth]{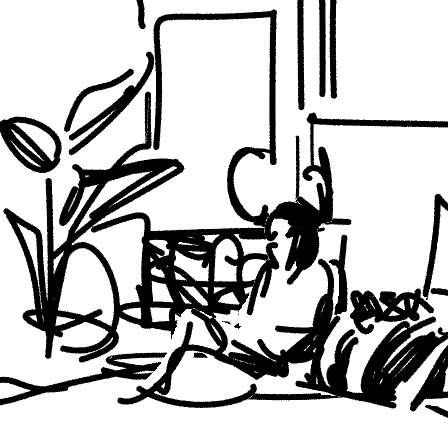}} &
    \frame{\includegraphics[width=0.098\textwidth,height=0.098\textwidth]{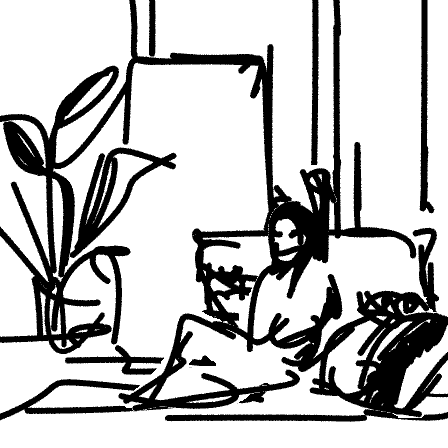}} &
    \frame{\includegraphics[width=0.098\textwidth,height=0.098\textwidth]{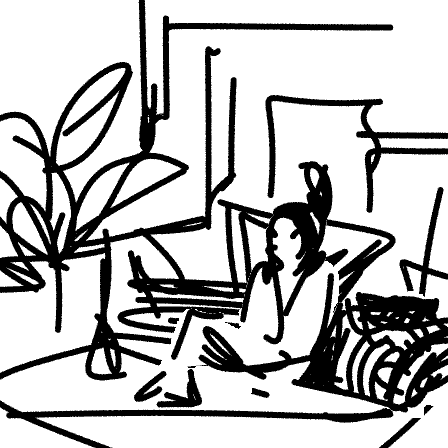}} \\
    
    \frame{\includegraphics[width=0.098\textwidth,height=0.098\textwidth]{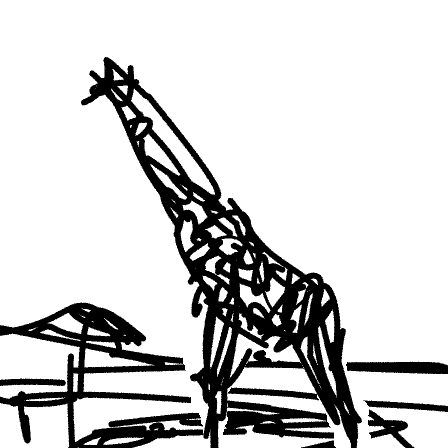}} &
    \frame{\includegraphics[width=0.098\textwidth,height=0.098\textwidth]{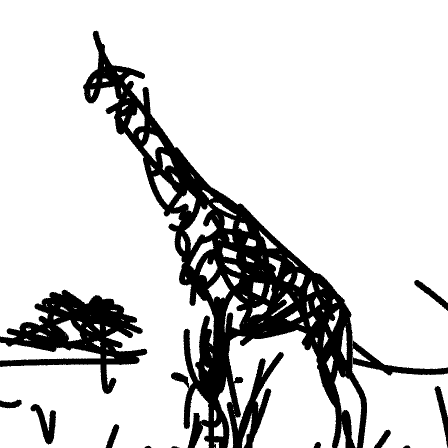}} &
    \frame{\includegraphics[width=0.098\textwidth,height=0.098\textwidth]{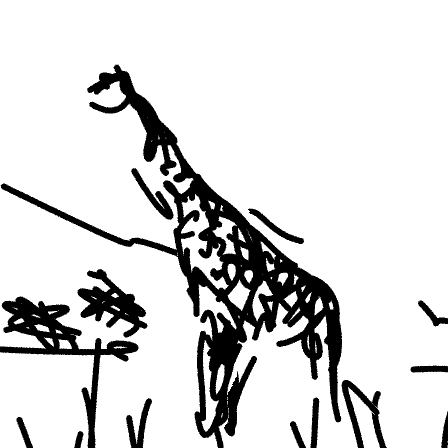}} &
    \frame{\includegraphics[width=0.098\textwidth,height=0.098\textwidth]{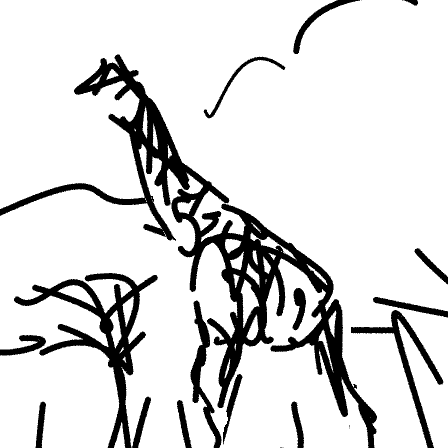}} &
    \hspace{0.5cm}
    \frame{\includegraphics[width=0.098\textwidth,height=0.098\textwidth]{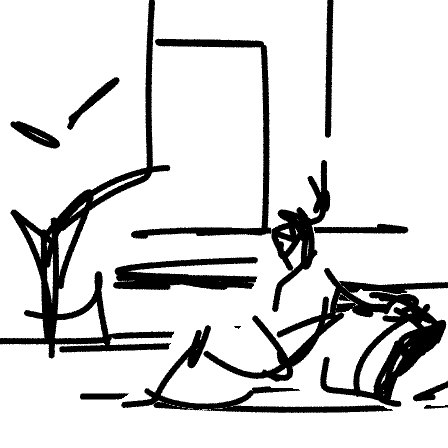}} &
    \frame{\includegraphics[width=0.098\textwidth,height=0.098\textwidth]{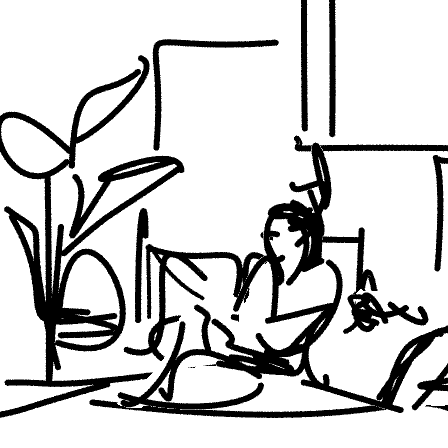}} &
    \frame{\includegraphics[width=0.098\textwidth,height=0.098\textwidth]{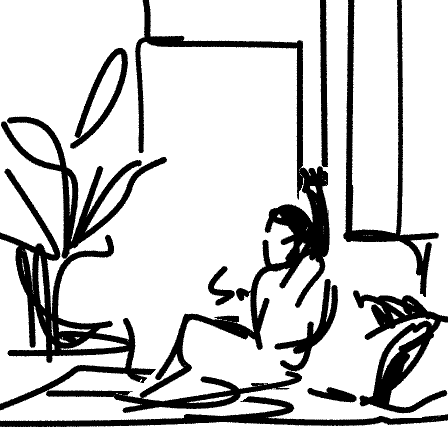}} &
    \frame{\includegraphics[width=0.098\textwidth,height=0.098\textwidth]{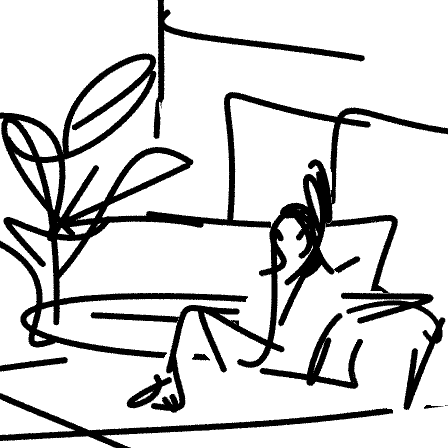}} \\
    
    \frame{\includegraphics[width=0.098\textwidth,height=0.098\textwidth]{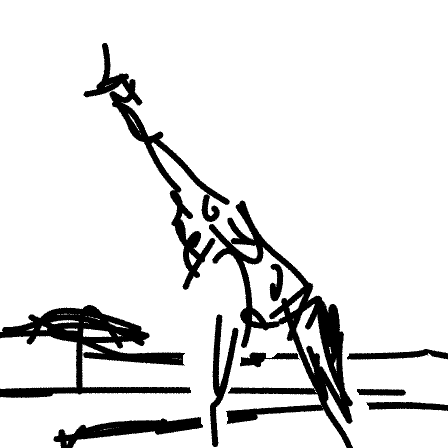}} &
    \frame{\includegraphics[width=0.098\textwidth,height=0.098\textwidth]{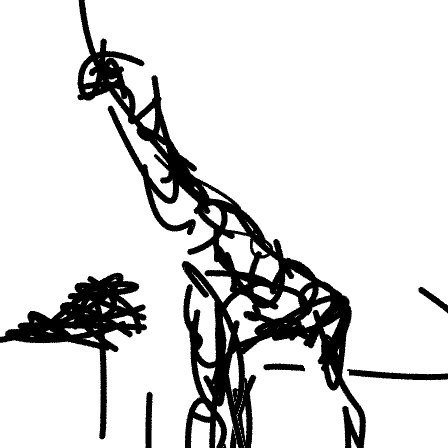}} &
    \frame{\includegraphics[width=0.098\textwidth,height=0.098\textwidth]{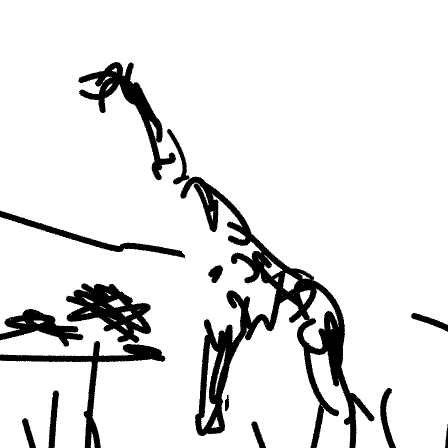}} &
    \frame{\includegraphics[width=0.098\textwidth,height=0.098\textwidth]{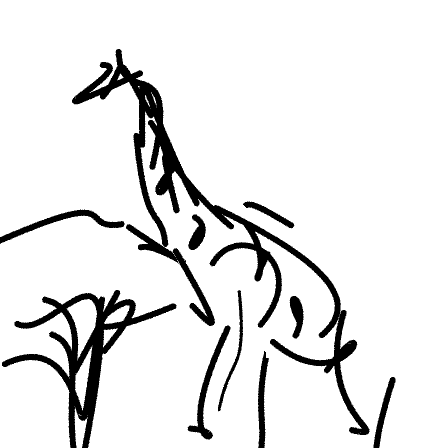}} &
    \hspace{0.5cm}
    \frame{\includegraphics[width=0.098\textwidth,height=0.098\textwidth]{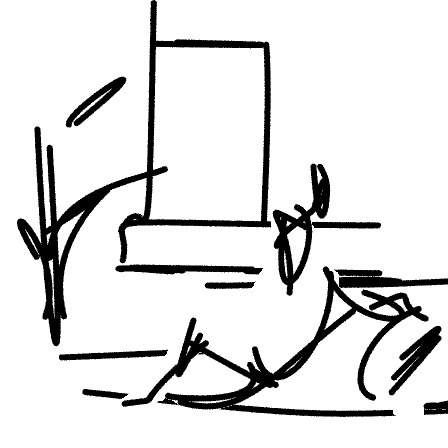}} &
    \frame{\includegraphics[width=0.098\textwidth,height=0.098\textwidth]{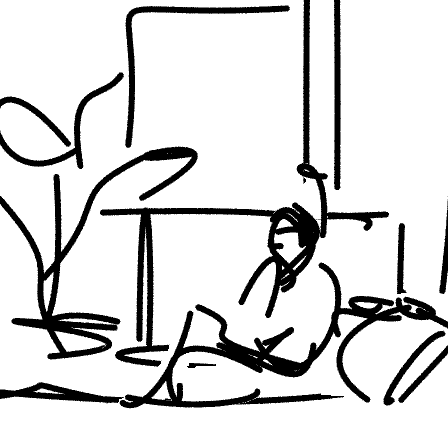}} &
    \frame{\includegraphics[width=0.098\textwidth,height=0.098\textwidth]{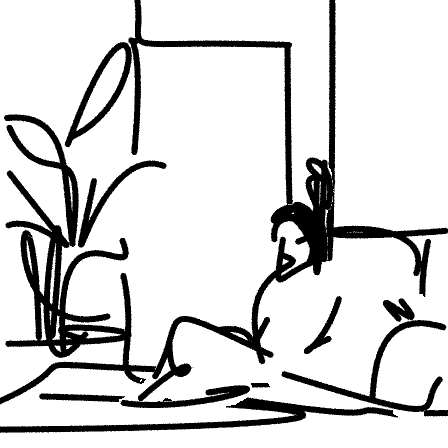}} &
    \frame{\includegraphics[width=0.098\textwidth,height=0.098\textwidth]{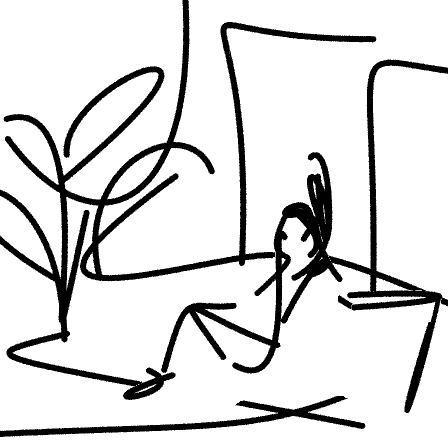}} \\
    
    \frame{\includegraphics[width=0.098\textwidth,height=0.098\textwidth]{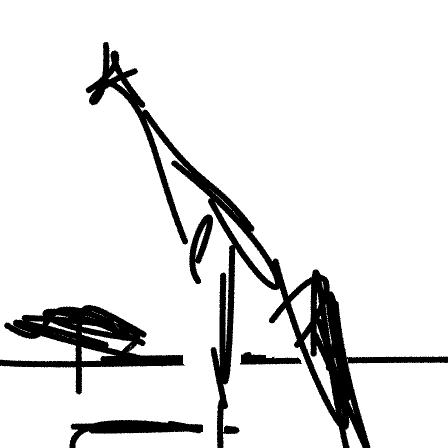}} &
    \frame{\includegraphics[width=0.098\textwidth,height=0.098\textwidth]{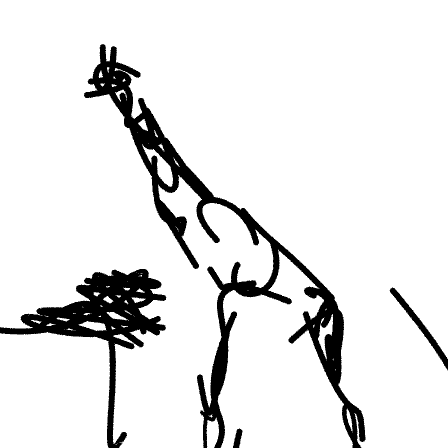}} &
    \frame{\includegraphics[width=0.098\textwidth,height=0.098\textwidth]{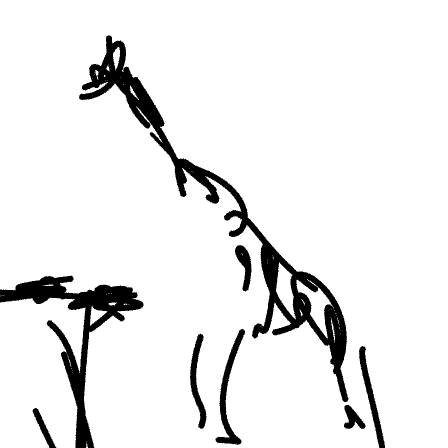}} &
    \frame{\includegraphics[width=0.098\textwidth,height=0.098\textwidth]{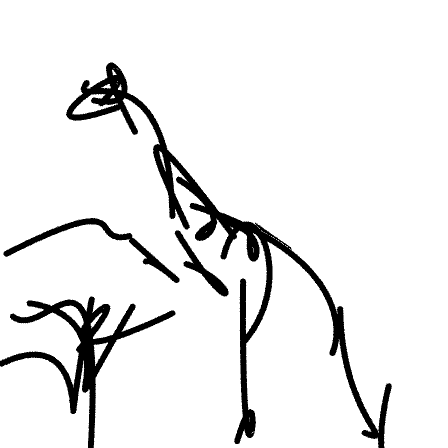}} &
    \hspace{0.5cm}
    \frame{\includegraphics[width=0.098\textwidth,height=0.098\textwidth]{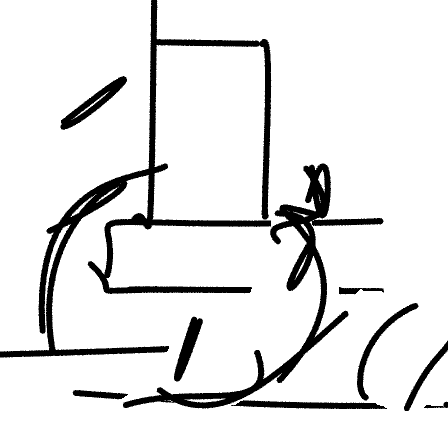}} &
    \frame{\includegraphics[width=0.098\textwidth,height=0.098\textwidth]{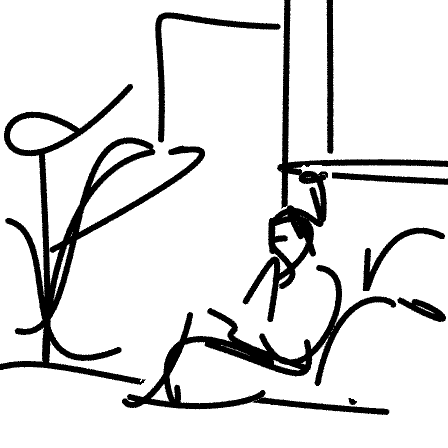}} &
    \frame{\includegraphics[width=0.098\textwidth,height=0.098\textwidth]{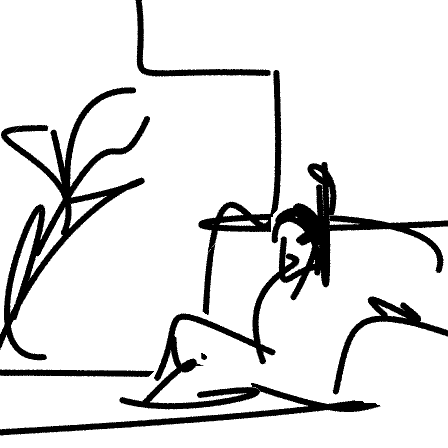}} &
    \frame{\includegraphics[width=0.098\textwidth,height=0.098\textwidth]{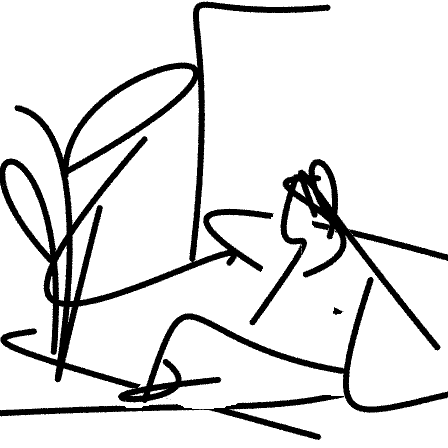}} \\

    \\
    \\
    
    \includegraphics[width=0.098\textwidth,height=0.098\textwidth]{figs/inputs/cat_c.jpg} & & & &
    \hspace{0.5cm}
    \includegraphics[width=0.098\textwidth,height=0.098\textwidth]{figs/inputs/hummingbird.jpg} & & & \\

    \frame{\includegraphics[width=0.098\textwidth,height=0.098\textwidth]{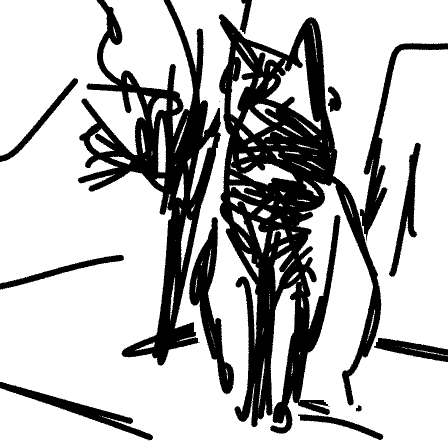}} &
    \frame{\includegraphics[width=0.098\textwidth,height=0.098\textwidth]{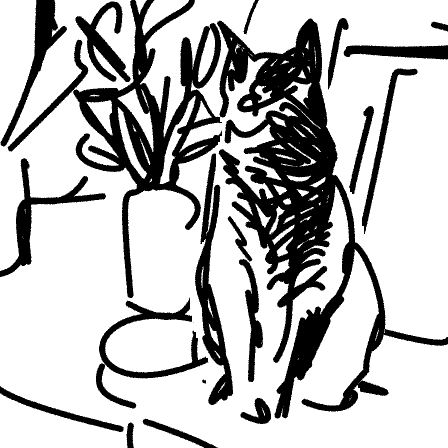}} &
    \frame{\includegraphics[width=0.098\textwidth,height=0.098\textwidth]{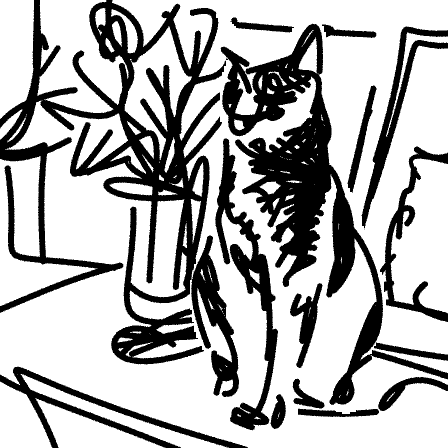}} &
    \frame{\includegraphics[width=0.098\textwidth,height=0.098\textwidth]{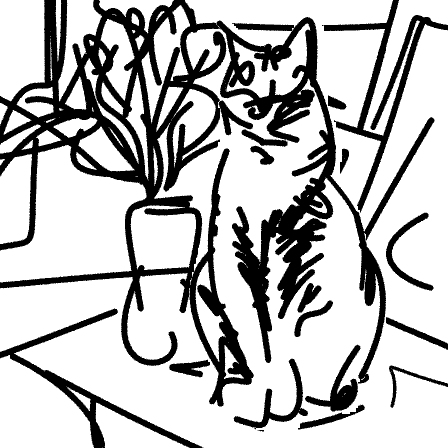}} &
    \hspace{0.5cm}
    \frame{\includegraphics[width=0.098\textwidth,height=0.098\textwidth]{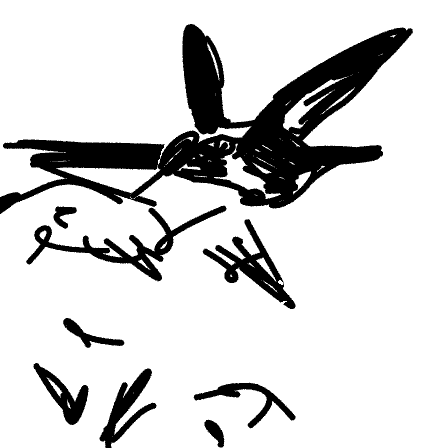}} &
    \frame{\includegraphics[width=0.098\textwidth,height=0.098\textwidth]{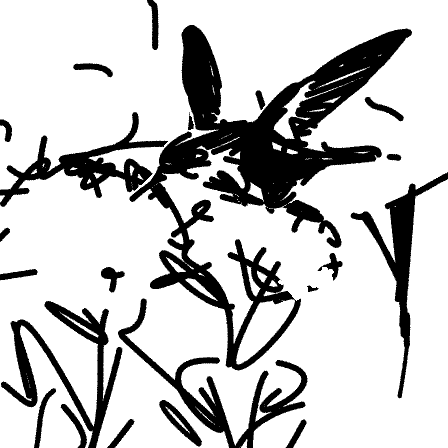}} &
    \frame{\includegraphics[width=0.098\textwidth,height=0.098\textwidth]{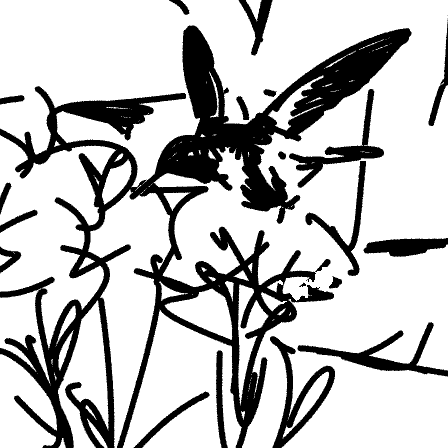}} &
    \frame{\includegraphics[width=0.098\textwidth,height=0.098\textwidth]{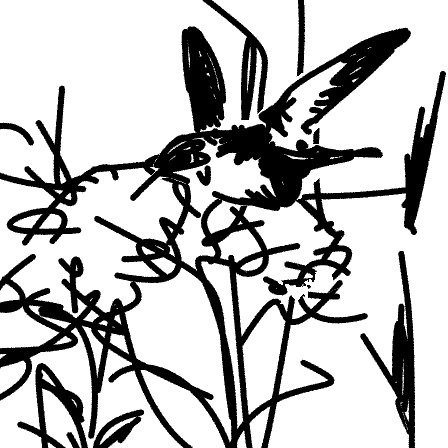}} \\
    
    \frame{\includegraphics[width=0.098\textwidth,height=0.098\textwidth]{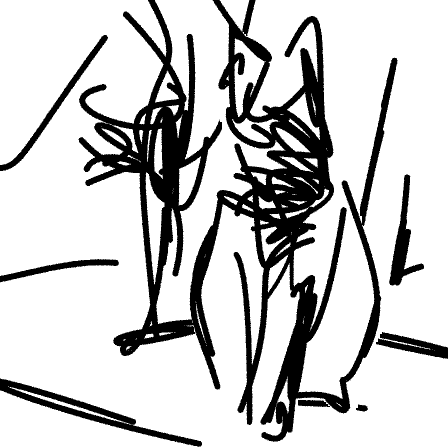}} &
    \frame{\includegraphics[width=0.098\textwidth,height=0.098\textwidth]{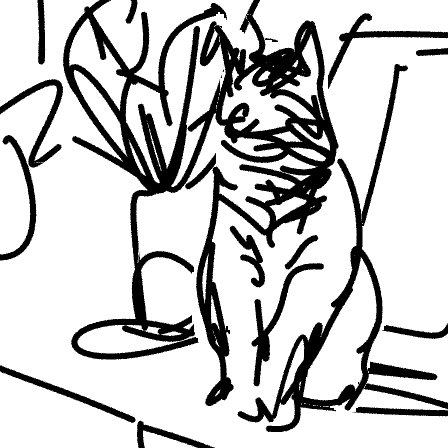}} &
    \frame{\includegraphics[width=0.098\textwidth,height=0.098\textwidth]{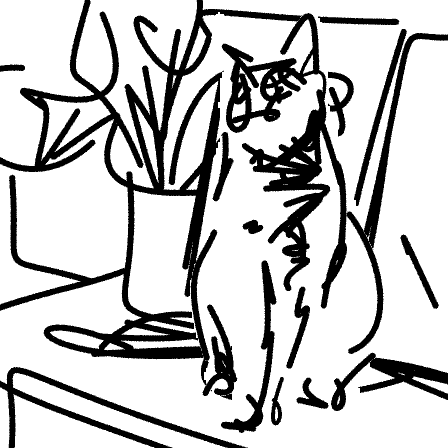}} &
    \frame{\includegraphics[width=0.098\textwidth,height=0.098\textwidth]{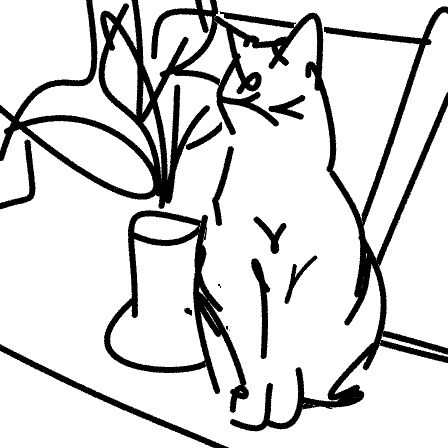}} &
    \hspace{0.5cm}
    \frame{\includegraphics[width=0.098\textwidth,height=0.098\textwidth]{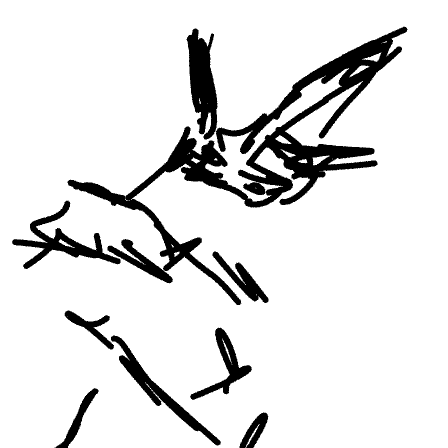}} &
    \frame{\includegraphics[width=0.098\textwidth,height=0.098\textwidth]{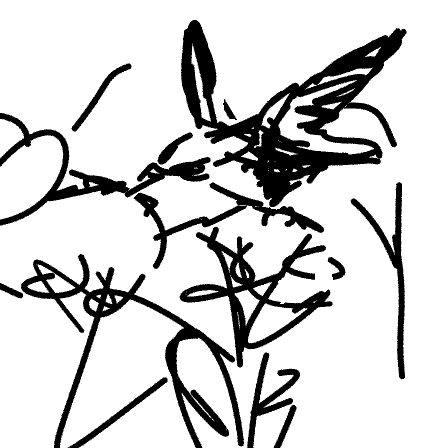}} &
    \frame{\includegraphics[width=0.098\textwidth,height=0.098\textwidth]{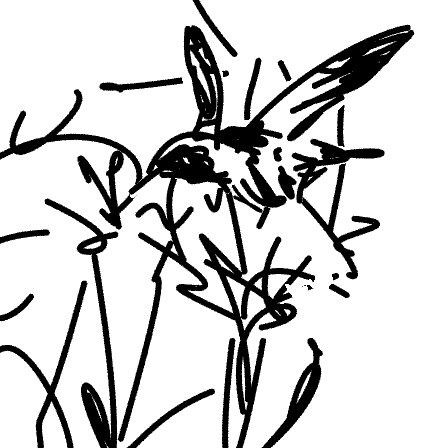}} &
    \frame{\includegraphics[width=0.098\textwidth,height=0.098\textwidth]{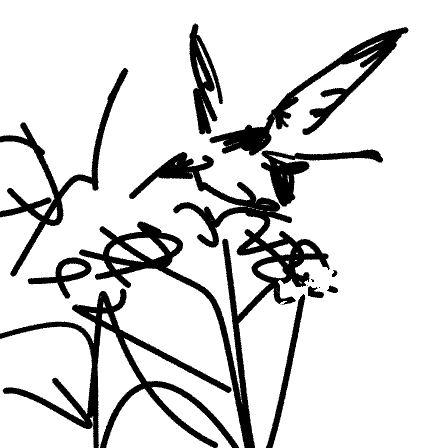}} \\
    
    \frame{\includegraphics[width=0.098\textwidth,height=0.098\textwidth]{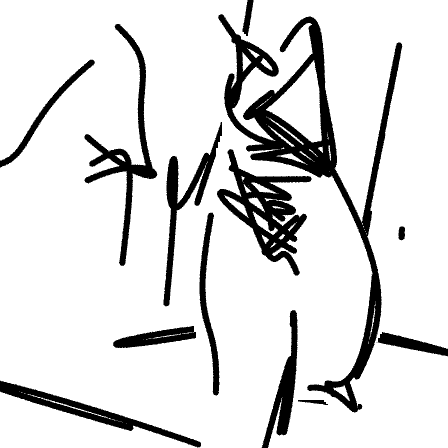}} &
    \frame{\includegraphics[width=0.098\textwidth,height=0.098\textwidth]{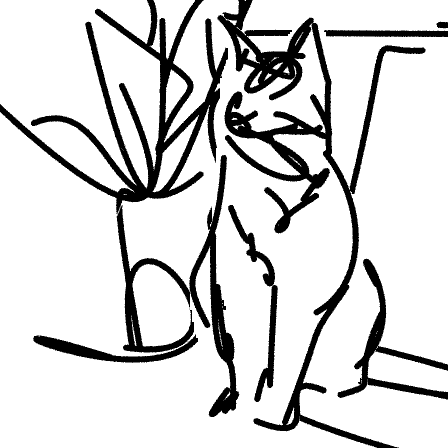}} &
    \frame{\includegraphics[width=0.098\textwidth,height=0.098\textwidth]{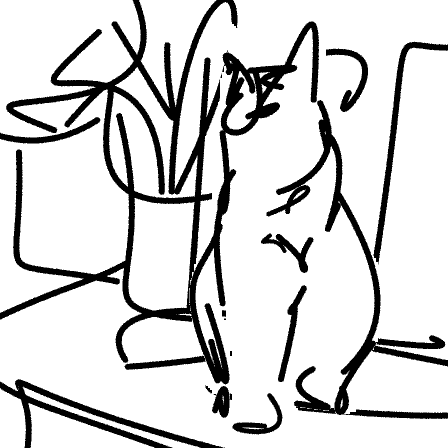}} &
    \frame{\includegraphics[width=0.098\textwidth,height=0.098\textwidth]{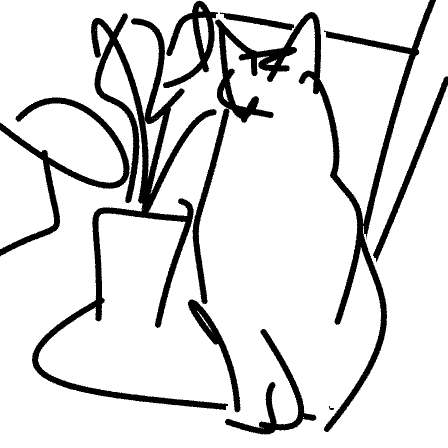}} &
    \hspace{0.5cm}
    \frame{\includegraphics[width=0.098\textwidth,height=0.098\textwidth]{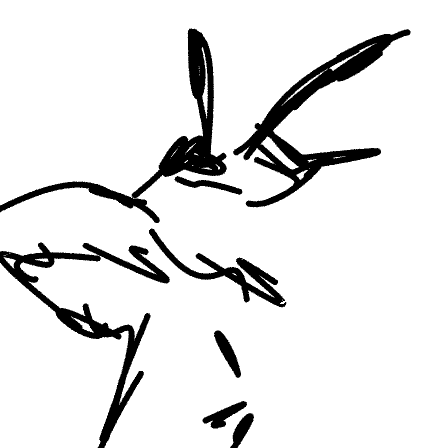}} &
    \frame{\includegraphics[width=0.098\textwidth,height=0.098\textwidth]{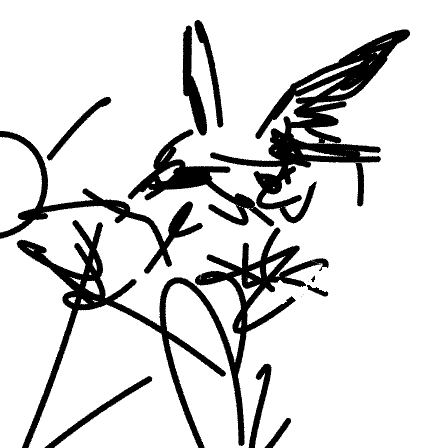}} &
    \frame{\includegraphics[width=0.098\textwidth,height=0.098\textwidth]{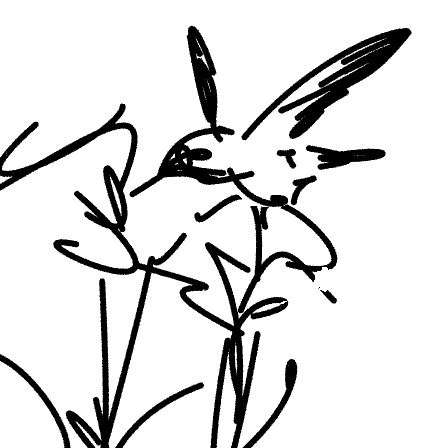}} &
    \frame{\includegraphics[width=0.098\textwidth,height=0.098\textwidth]{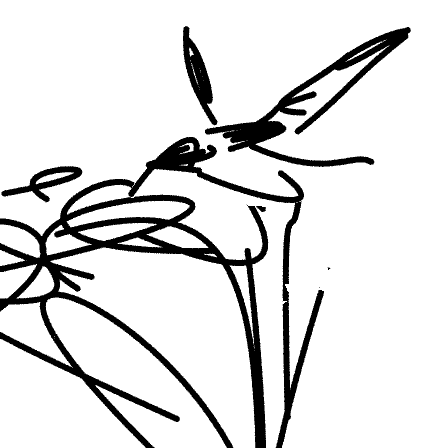}} \\
    
    \frame{\includegraphics[width=0.098\textwidth,height=0.098\textwidth]{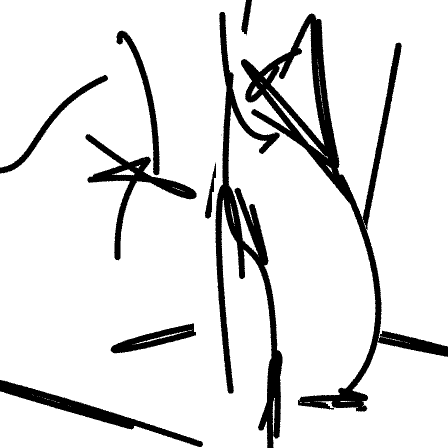}} &
    \frame{\includegraphics[width=0.098\textwidth,height=0.098\textwidth]{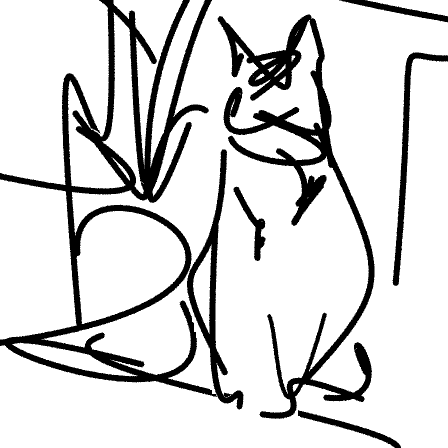}} &
    \frame{\includegraphics[width=0.098\textwidth,height=0.098\textwidth]{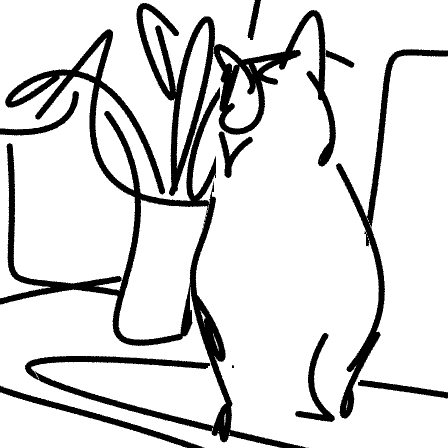}} &
    \frame{\includegraphics[width=0.098\textwidth,height=0.098\textwidth]{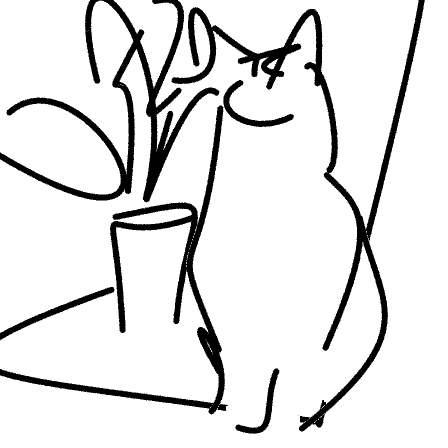}} &
    \hspace{0.5cm}
    \frame{\includegraphics[width=0.098\textwidth,height=0.098\textwidth]{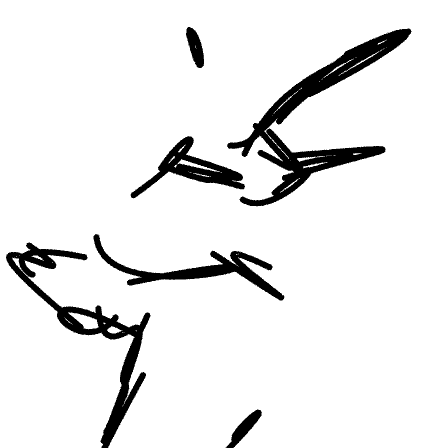}} &
    \frame{\includegraphics[width=0.098\textwidth,height=0.098\textwidth]{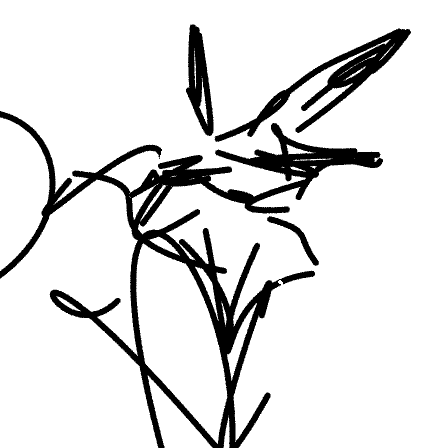}} &
    \frame{\includegraphics[width=0.098\textwidth,height=0.098\textwidth]{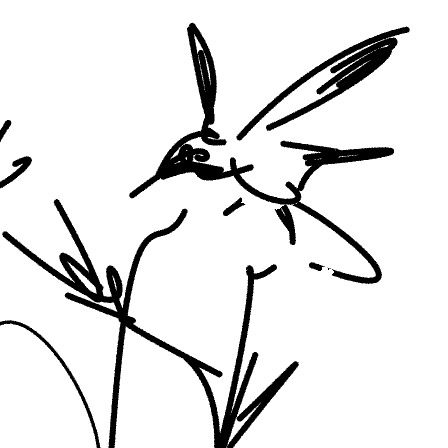}} &
    \frame{\includegraphics[width=0.098\textwidth,height=0.098\textwidth]{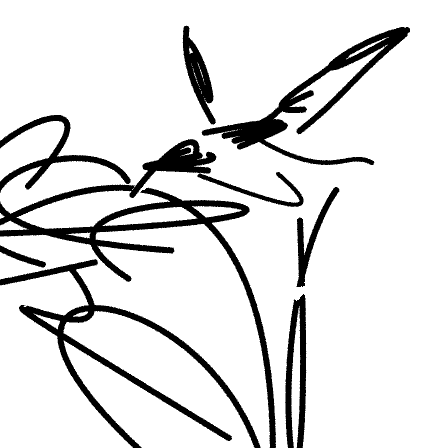}} \\

    \end{tabular}
    \caption{The $4\times4$ matrix of sketches produced by our method. Columns from left to right illustrate the change in fidelity, from precise to loose, and rows from top to bottom illustrate the visual simplification.}
    
    \label{fig:matrix5}
\end{figure*}

%% file: files/figures/supplementary/diffusion_comparison.tex
\begin{figure}
    \centering
    \setlength{\tabcolsep}{1.5pt}
    {\small
    \begin{tabular}{c c c c c c}

        \includegraphics[width=0.085\textwidth]{figs/inputs/house3.jpg} &
        \hspace{0.1cm}
        {\footnotesize
        \raisebox{0.01in}{\rotatebox{90}{Img2Img (SD)}}
        } &
        \includegraphics[width=0.085\textwidth]{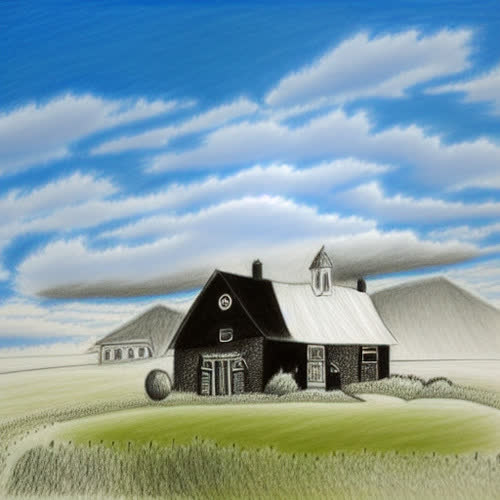} &
        \includegraphics[width=0.085\textwidth]{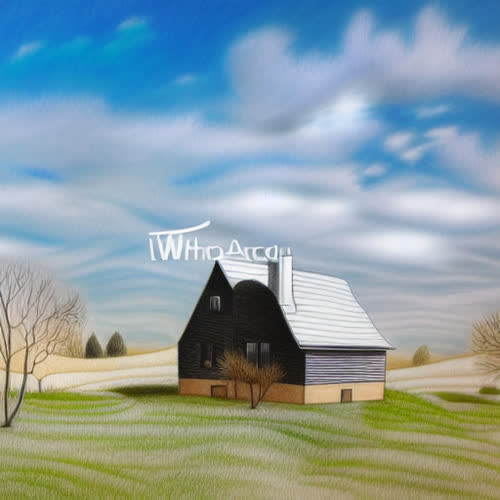} &
        \includegraphics[width=0.085\textwidth]{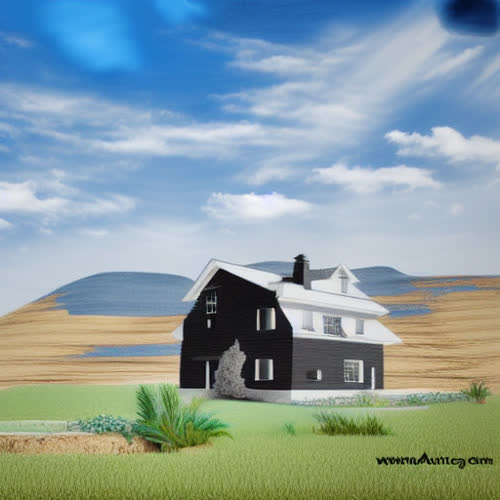} &
        \includegraphics[width=0.085\textwidth]{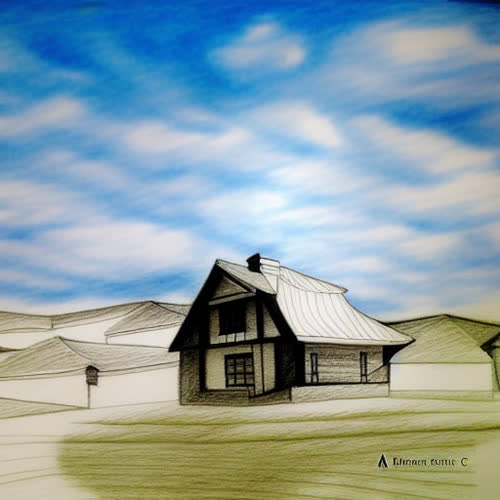} \\
    
        &
        \hspace{0.1cm}
        \raisebox{0.15in}{\rotatebox{90}{Ours}} &
        \includegraphics[width=0.085\textwidth]{figs/matrices_black/house3_row0col0_black.png} &
        \includegraphics[width=0.085\textwidth]{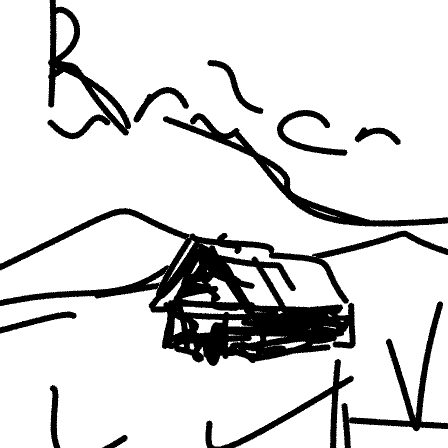} &
        \includegraphics[width=0.085\textwidth]{figs/matrices_black/house3_row2col2_black.png} &
        \includegraphics[width=0.085\textwidth]{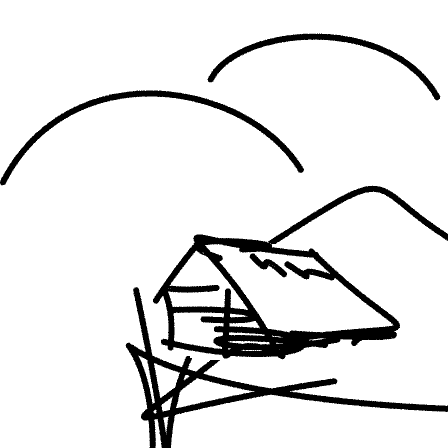} \\

        \cline{2-6} \\
        
        \includegraphics[width=0.085\textwidth]{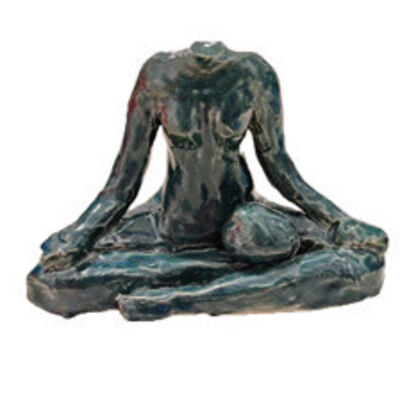} &
        \hspace{0.1cm}
        \raisebox{0.3in}{
        \multirow{2}{*}{\raisebox{0.25in}{\rotatebox{90}{Textual Inversion}}}} &
        \includegraphics[width=0.085\textwidth]{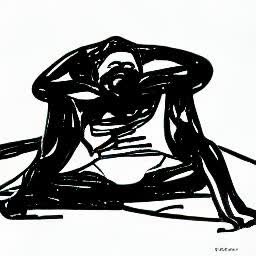} &
        \includegraphics[width=0.085\textwidth]{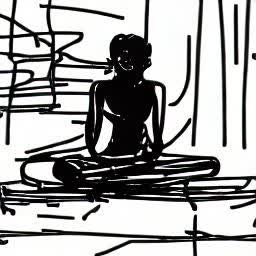} &
        \includegraphics[width=0.085\textwidth]{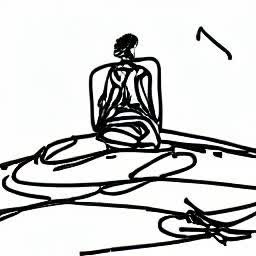} &
        \includegraphics[width=0.085\textwidth]{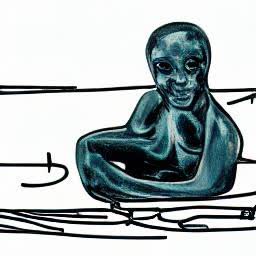} \\
        & & 
        \includegraphics[width=0.085\textwidth]{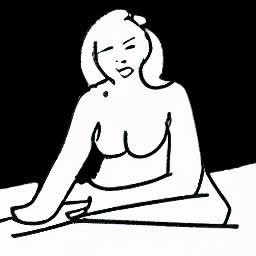} &
        \includegraphics[width=0.085\textwidth]{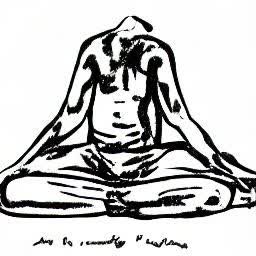} &
        \includegraphics[width=0.085\textwidth]{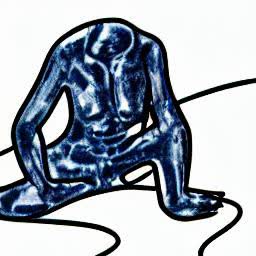} &
        \includegraphics[width=0.085\textwidth]{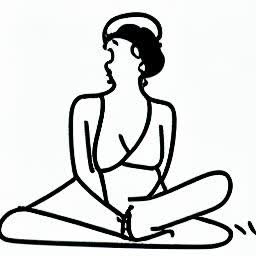} \\

        &
        \hspace{0.1cm}
        \raisebox{0.25in}{\rotatebox{90}{Ours}} &
        \includegraphics[width=0.085\textwidth]{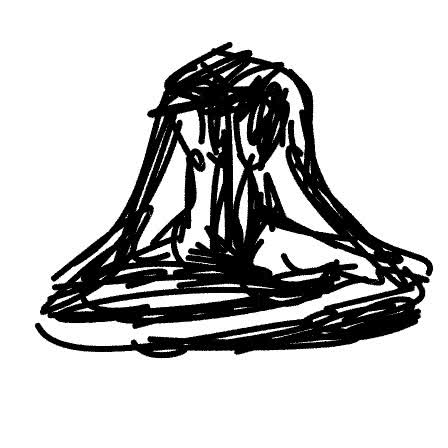} &
        \includegraphics[width=0.085\textwidth]{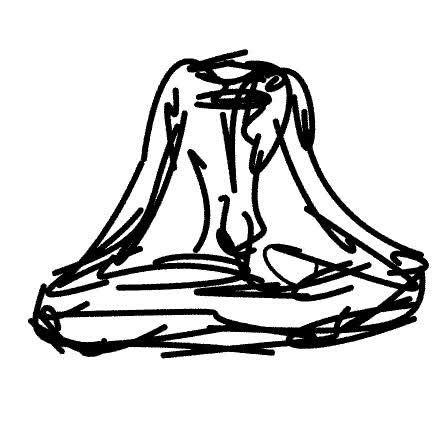} &
        \includegraphics[width=0.085\textwidth]{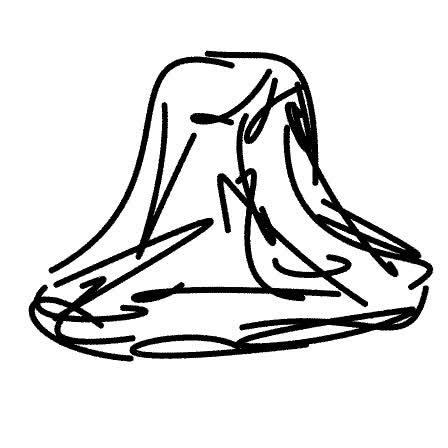} &
        \includegraphics[width=0.085\textwidth]{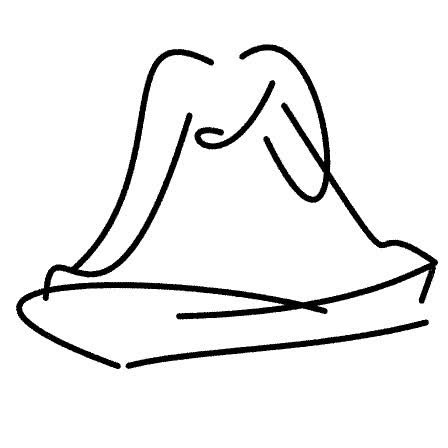} \\

    \end{tabular}
    
    }
    \caption{Comparison to various diffusion model-based techniques. At the top, we compare our sketch results to those obtained using Stable Diffusion~\cite{rombach2021highresolution} image-to-image technique guided by the text prompt ``A black and white sketch image''. At the bottom, we compare with Textual Inversion~\cite{gal2022image} by learning new tokens representing a detailed sketch (top row) and abstract sketch (bottom) row. As shown, both approaches struggle in either capturing the desired sketch style or input subject.}
    \label{fig:diffusion_comparison}
\end{figure}

%% file: files/figures/supplementary/clipasso_comparison_main.tex
\begin{figure}[ht]
    \centering
    \setlength{\tabcolsep}{1pt}
    \begin{tabular}{c c c c c c}

    \includegraphics[width=0.17\linewidth]{figs/inputs/bull.jpg} &
    \raisebox{0.05in}{\rotatebox{90}{CLIPasso}} &
    \includegraphics[width=0.17\linewidth]{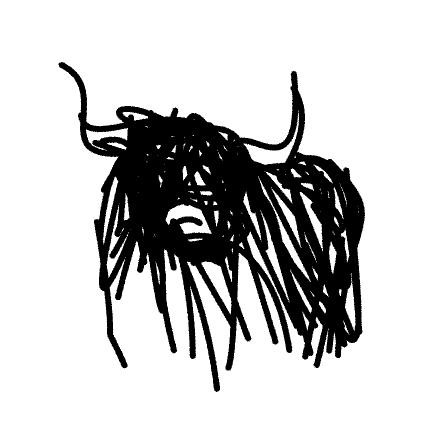} &
    \includegraphics[width=0.17\linewidth]{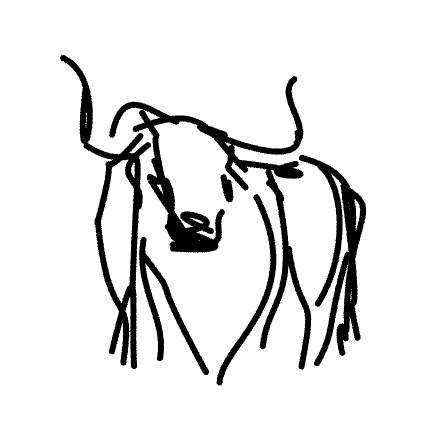} &
    \includegraphics[width=0.17\linewidth]{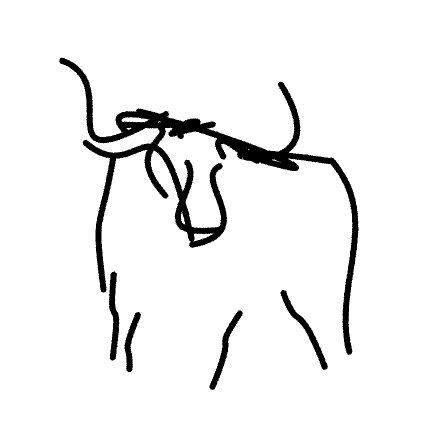} &
    \includegraphics[width=0.17\linewidth]{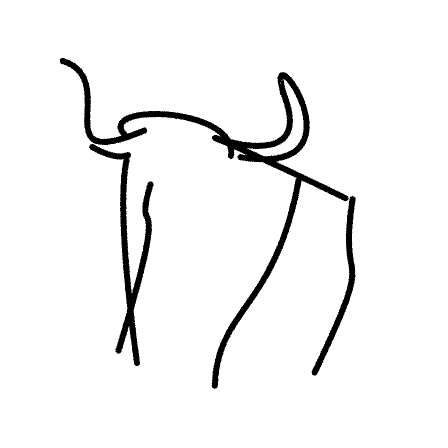} \\

    \cline{2-6}

    & 
    \raisebox{0.35cm}{
        \multirow{2}{*}{\rotatebox{90}{\hspace{0.1cm} \xrfill[0.5ex]{0.5pt}\; Ours \; \xrfill[0.5ex]{0.5pt} }}
    } &
    \includegraphics[width=0.17\linewidth]{figs/matrices_black/bull_row0col0_black.png} &
    \includegraphics[width=0.17\linewidth]{figs/matrices_black/bull_row1col0_black.png} &
    \includegraphics[width=0.17\linewidth]{figs/matrices_black/bull_row2col0_black.png} &
    \includegraphics[width=0.17\linewidth]{figs/matrices_black/bull_row3col0_black.png} \\
    
    &
    &
    \includegraphics[width=0.17\linewidth]{figs/matrices_black/bull_row0col3_black.png} &
    \includegraphics[width=0.17\linewidth]{figs/matrices_black/bull_row1col3_black.png} &
    \includegraphics[width=0.17\linewidth]{figs/matrices_black/bull_row2col3_black.png} &
    \includegraphics[width=0.17\linewidth]{figs/matrices_black/bull_row3col3_black.png} \\
    \end{tabular}
    
    \caption{Comparisons with CLIPasso~\cite{vinker2022clipasso}. We show CLIPasso results obtained when applied over an entire scene image using $128$ and $64$ strokes respectively. We show our sketch results obtained using approximately the same number of strokes.}
    \label{fig:clipasso_comp_main}
\end{figure}

%% file: files/figures/supplementary/scene_sketch_comparison.tex
\begin{figure*}
    \centering
    \setlength{\belowcaptionskip}{-6pt}
    \setlength{\tabcolsep}{1.5pt}
    {\small
    \begin{tabular}{c c c c c c c | c c}

        \includegraphics[width=0.10\textwidth]{figs/inputs/black_man.jpg} &
        \hspace{0.1cm}
        \includegraphics[width=0.10\textwidth]{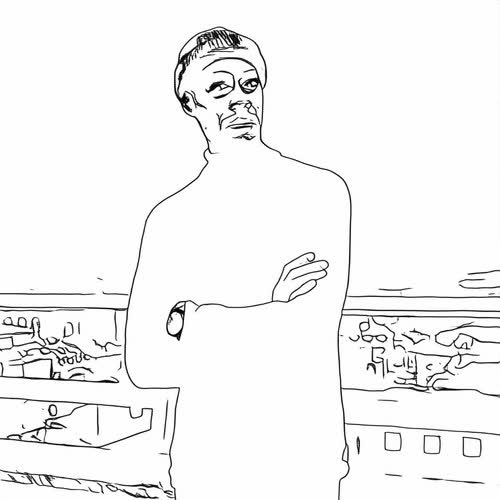} &
        \includegraphics[width=0.10\textwidth]{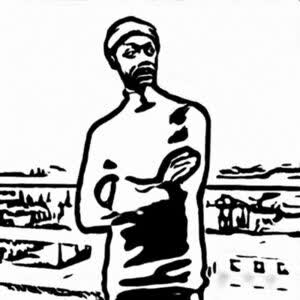} &
        \includegraphics[width=0.10\textwidth]{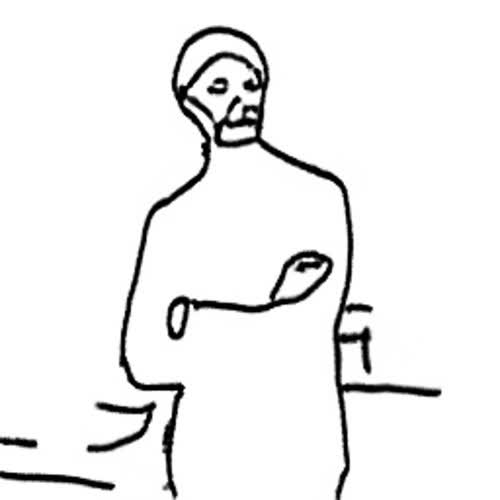} &
        \includegraphics[width=0.10\textwidth]{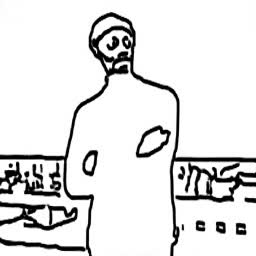} &
        \includegraphics[width=0.10\textwidth]{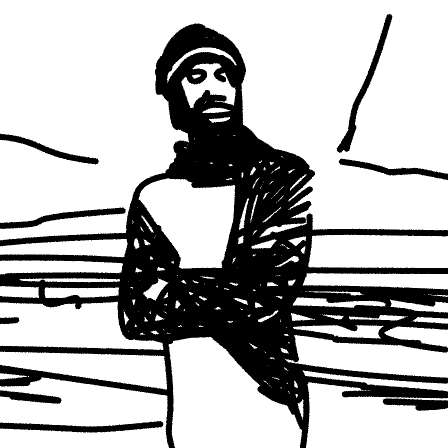} &
        \includegraphics[width=0.10\textwidth]{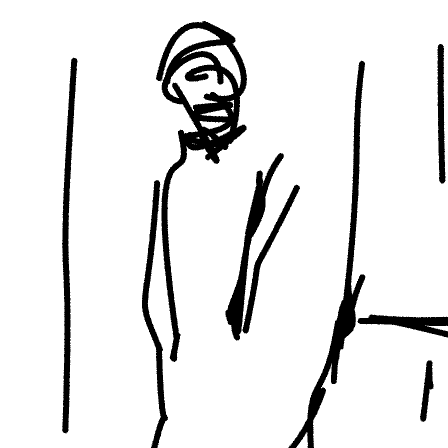} &
        \hspace{0.1cm}
        \includegraphics[width=0.10\textwidth]{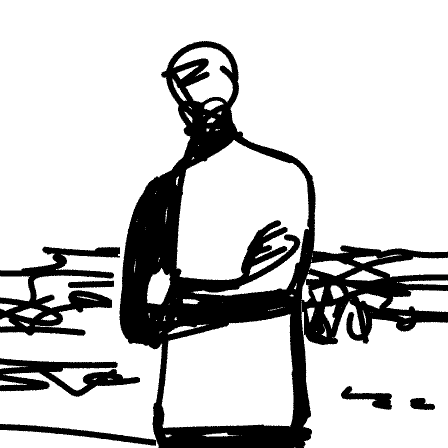} &
        \includegraphics[width=0.10\textwidth]{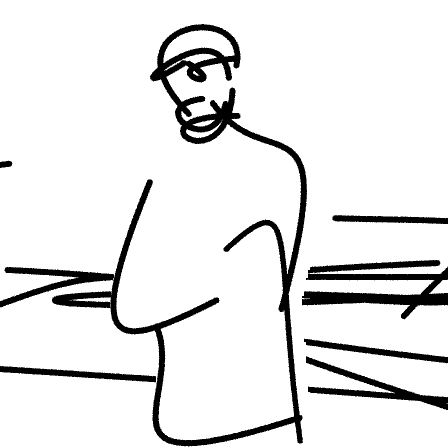} \\

        \includegraphics[width=0.10\textwidth]{figs/inputs/ballerina.jpg} &
        \hspace{0.1cm}
        \includegraphics[width=0.10\textwidth]{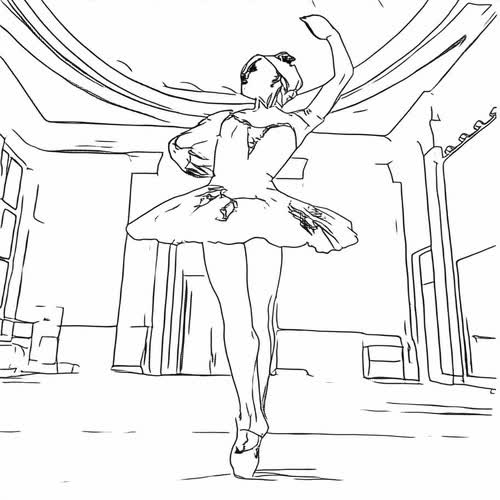} &
        \includegraphics[width=0.10\textwidth]{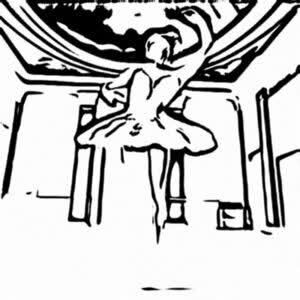} &
        \includegraphics[width=0.10\textwidth]{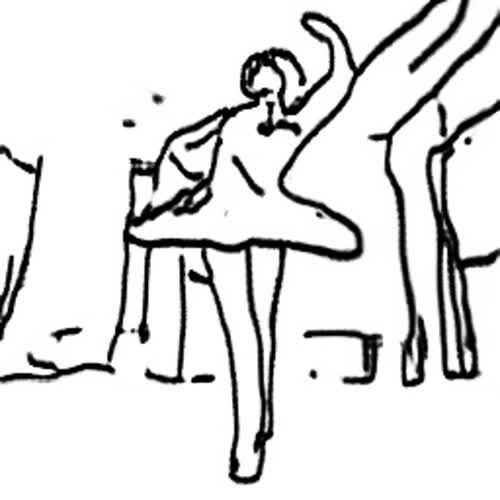} &
        \includegraphics[width=0.10\textwidth]{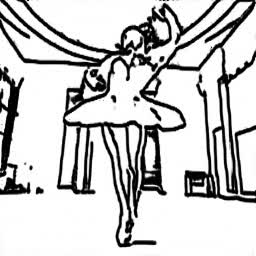} &
        \includegraphics[width=0.10\textwidth]{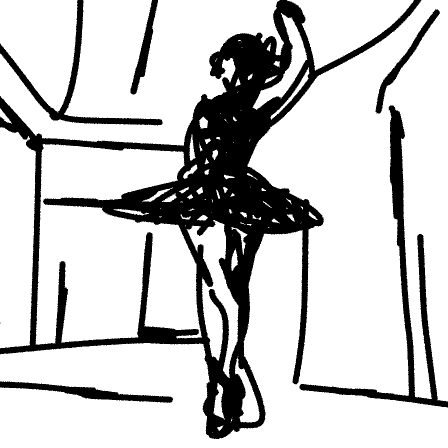} &
        \includegraphics[width=0.10\textwidth]{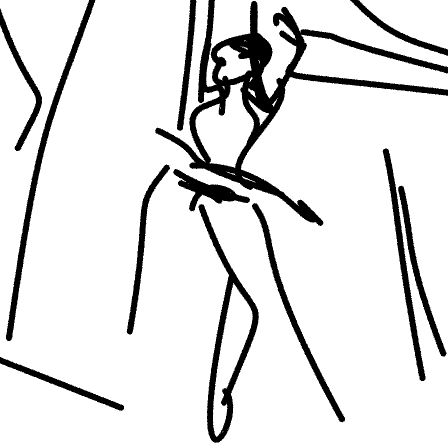} &
        \hspace{0.1cm}
        \includegraphics[width=0.10\textwidth]{figs/matrices_black/ballerina_row0col0_black.png} &
        \includegraphics[width=0.10\textwidth]{figs/matrices_black/ballerina_row3col3_black.png} \\
        
        \includegraphics[width=0.10\textwidth]{figs/inputs/basketball-5.jpg} &
        \hspace{0.1cm}
        \includegraphics[width=0.10\textwidth]{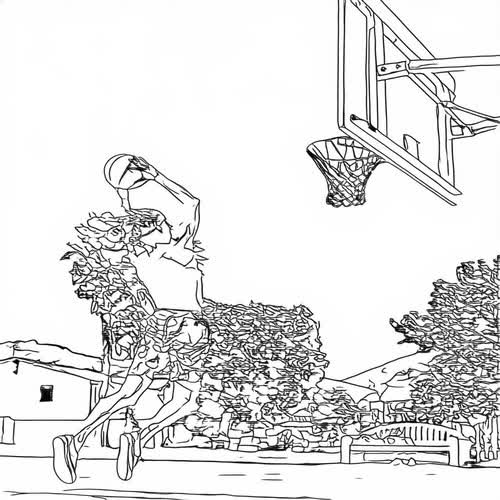} &
        \includegraphics[width=0.10\textwidth]{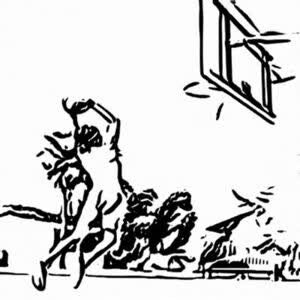} &
        \includegraphics[width=0.10\textwidth]{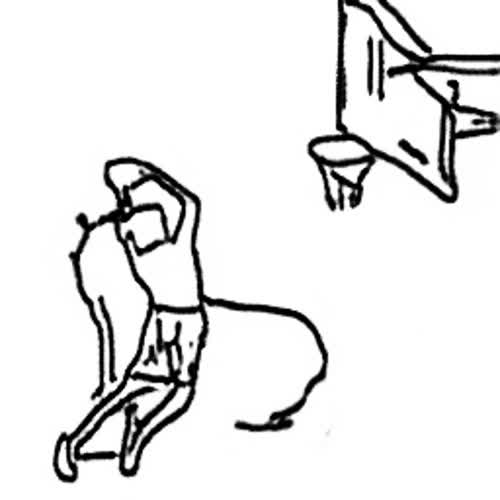} &
        \includegraphics[width=0.10\textwidth]{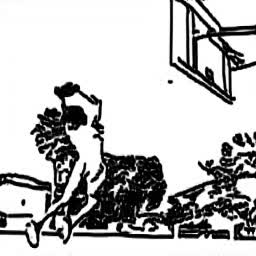} &
        \includegraphics[width=0.10\textwidth]{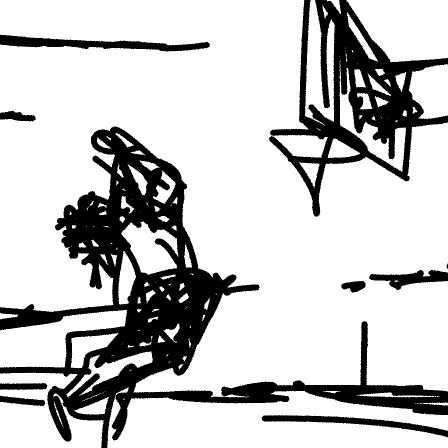} &
        \includegraphics[width=0.10\textwidth]{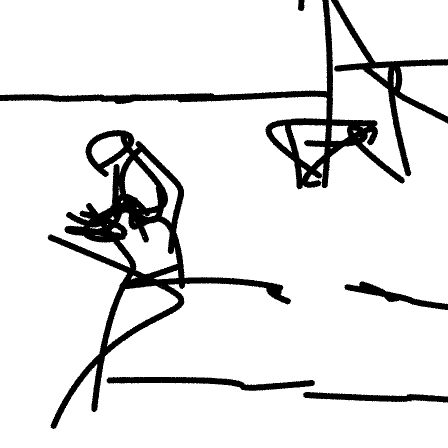} &
        \hspace{0.1cm}
        \includegraphics[width=0.10\textwidth]{figs/matrices_black/basketball-5_row0col0_black.png} & 
        \includegraphics[width=0.10\textwidth]{figs/matrices_black/basketball-5_row3col3_black.png} \\

        \includegraphics[width=0.10\textwidth]{figs/inputs/lighthouse-2.jpg} &
        \hspace{0.1cm}
        \includegraphics[width=0.10\textwidth]{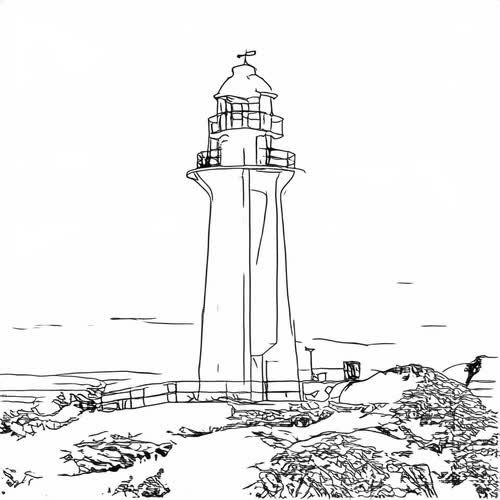} &
        \includegraphics[width=0.10\textwidth]{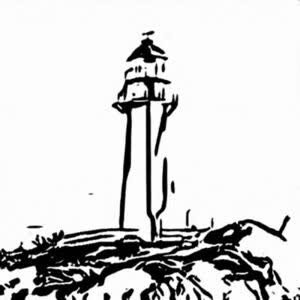} &
        \includegraphics[width=0.10\textwidth]{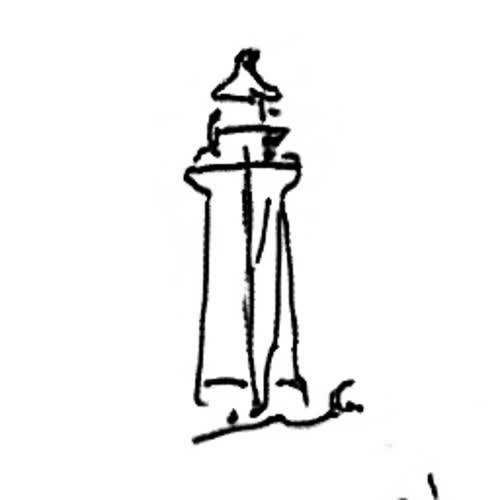} &
        \includegraphics[width=0.10\textwidth]{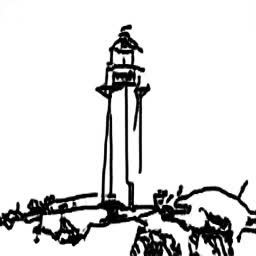} &
        \includegraphics[width=0.10\textwidth]{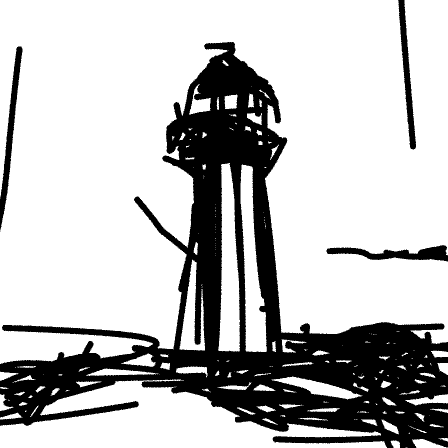} &
        \includegraphics[width=0.10\textwidth]{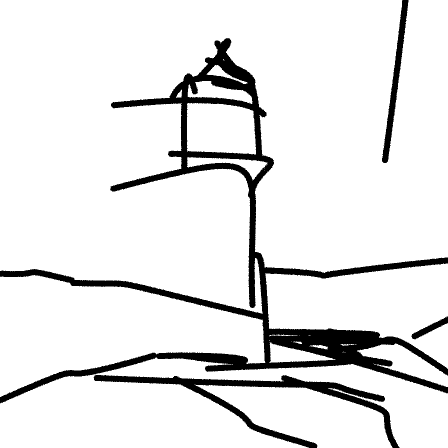} &
        \hspace{0.1cm}
        \includegraphics[width=0.10\textwidth]{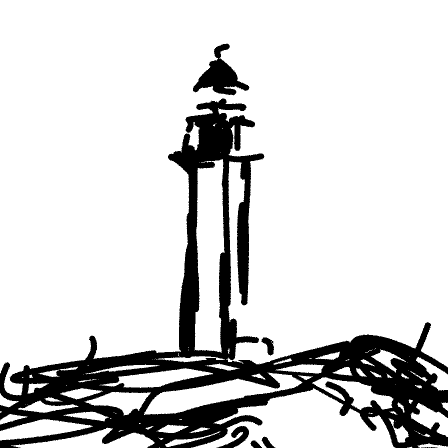} & 
        \includegraphics[width=0.10\textwidth]{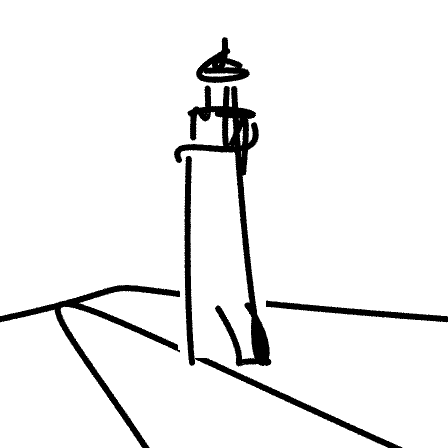} \\

        \includegraphics[width=0.10\textwidth]{figs/inputs/eiffel_tower.jpg} &
        \hspace{0.1cm}
        \includegraphics[width=0.10\textwidth]{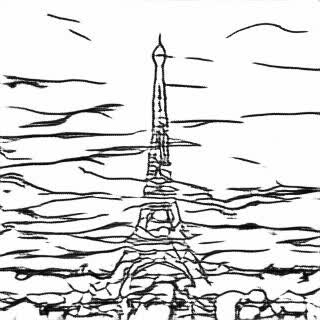} &
        \includegraphics[width=0.10\textwidth]{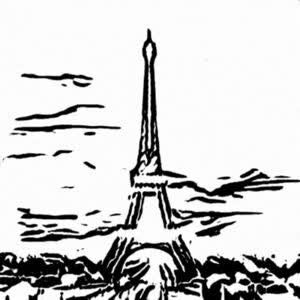} &
        \includegraphics[width=0.10\textwidth]{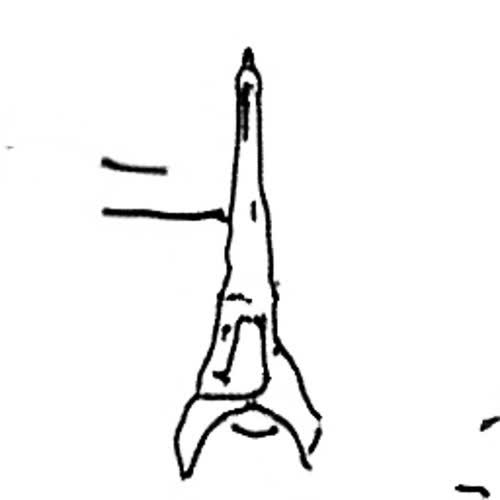} &
        \includegraphics[width=0.10\textwidth]{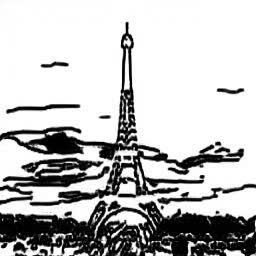} &
        \includegraphics[width=0.10\textwidth]{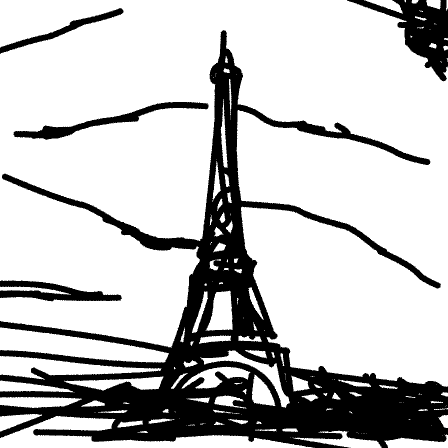} &
        \includegraphics[width=0.10\textwidth]{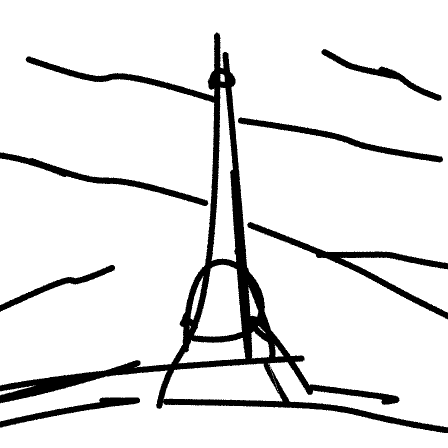} &
        \hspace{0.1cm}
        \includegraphics[width=0.10\textwidth]{figs/matrices_black/eiffel_tower_0.png} &
        \includegraphics[width=0.10\textwidth]{figs/matrices_black/eiffel_tower_15.png} \\

        \includegraphics[width=0.10\textwidth]{figs/inputs/man_flowers.jpg} &
        \hspace{0.1cm}
        \includegraphics[width=0.10\textwidth]{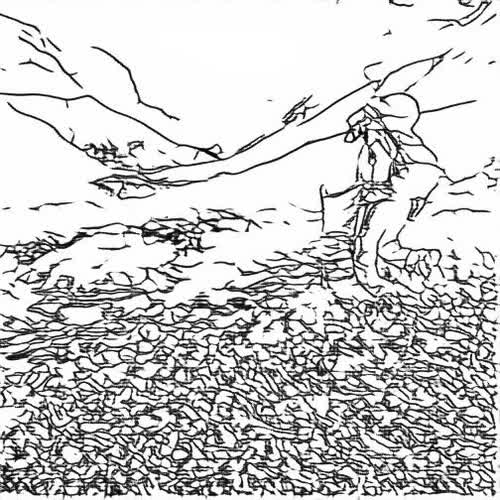} &
        \includegraphics[width=0.10\textwidth]{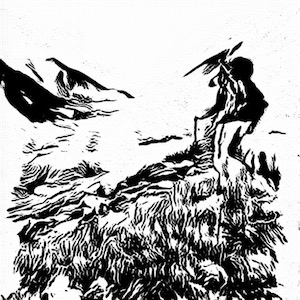} &
        \includegraphics[width=0.10\textwidth]{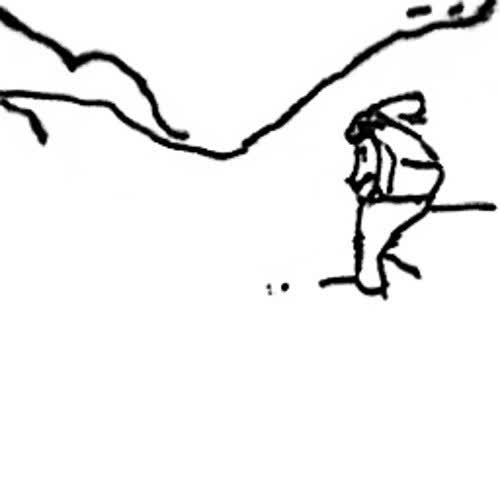} &
        \includegraphics[width=0.10\textwidth]{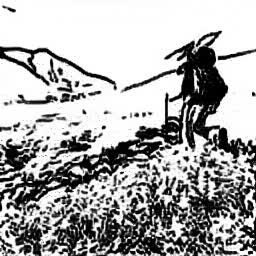} &
        \includegraphics[width=0.10\textwidth]{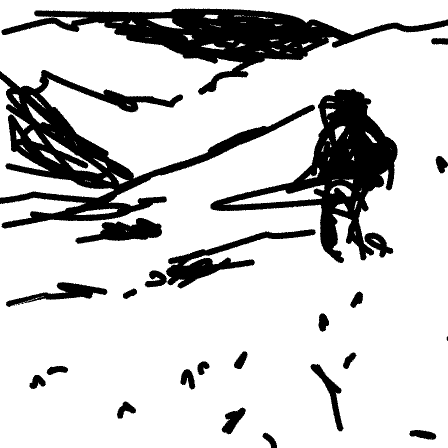} &
        \includegraphics[width=0.10\textwidth]{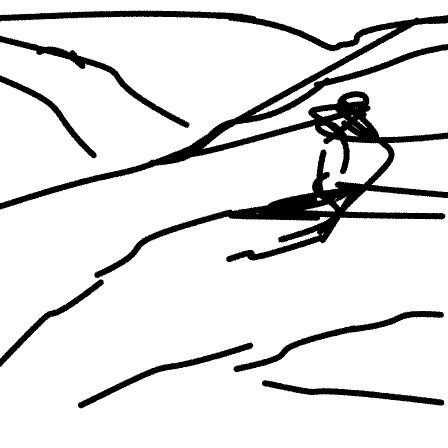} &
        \hspace{0.1cm}
        \includegraphics[width=0.10\textwidth]{figs/matrices_black/man_flowers_row0col0_black.png} & 
        \includegraphics[width=0.10\textwidth]{figs/matrices_black/man_flowers_row3col3_black.png} \\

        \includegraphics[width=0.10\textwidth]{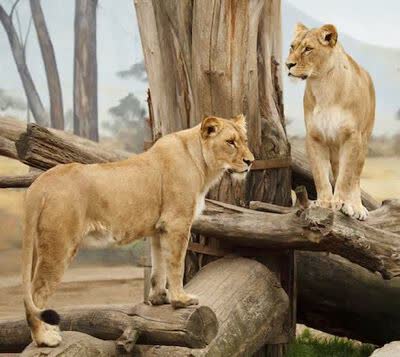} &
        \hspace{0.1cm}
        \includegraphics[width=0.10\textwidth]{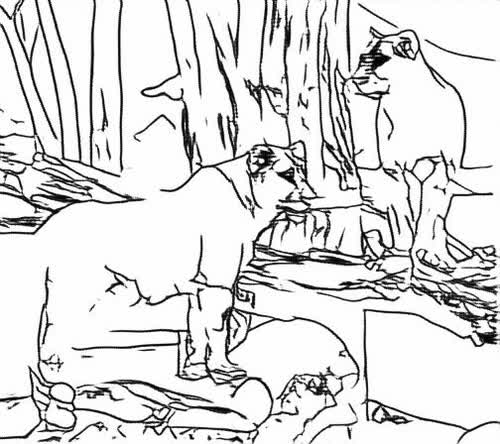} &
        \includegraphics[width=0.10\textwidth]{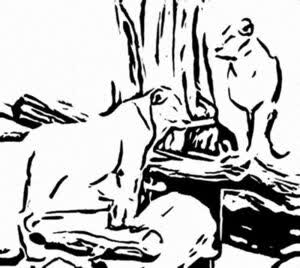} &
        \includegraphics[width=0.10\textwidth]{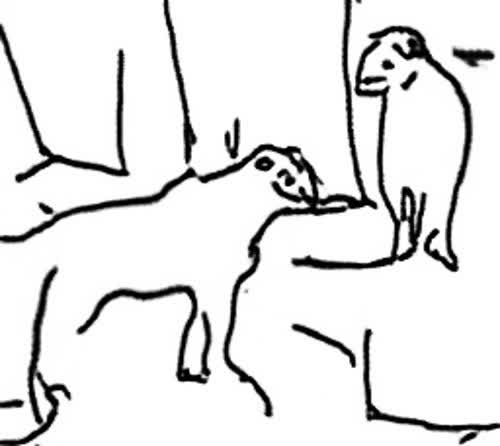} &
        \includegraphics[width=0.10\textwidth]{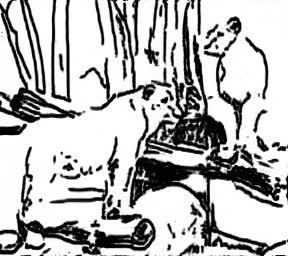} &
        \includegraphics[width=0.10\textwidth]{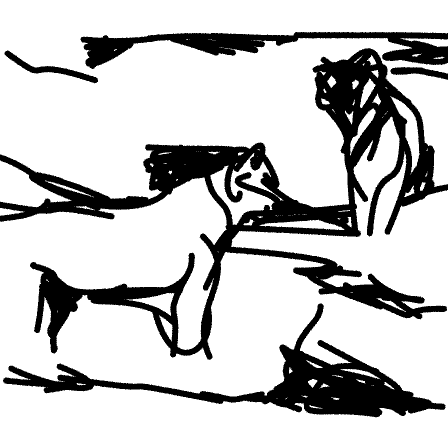} &
        \includegraphics[width=0.10\textwidth]{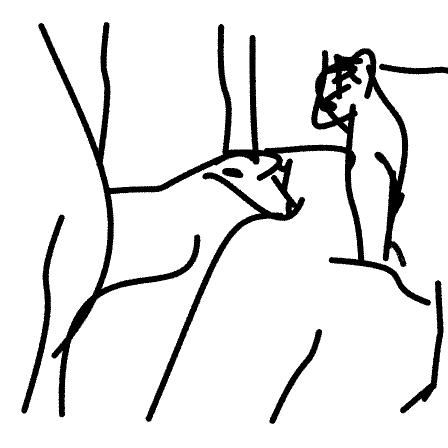} &
        \includegraphics[width=0.10\textwidth]{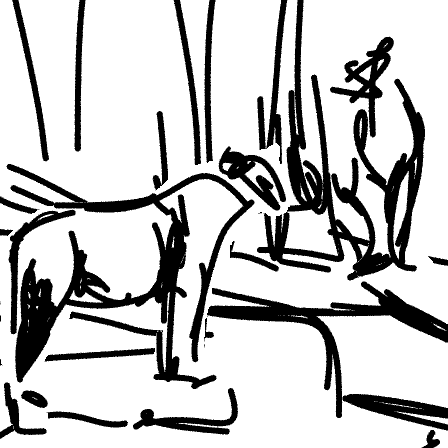} &
        \includegraphics[width=0.10\textwidth]{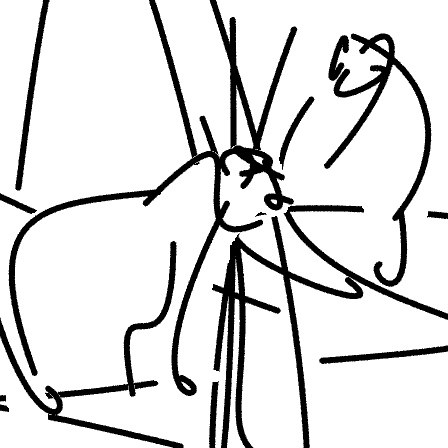} \\
        
        \includegraphics[width=0.10\textwidth]{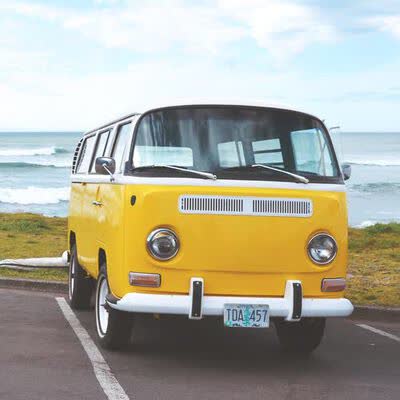} &
        \hspace{0.1cm}
        \includegraphics[width=0.10\textwidth]{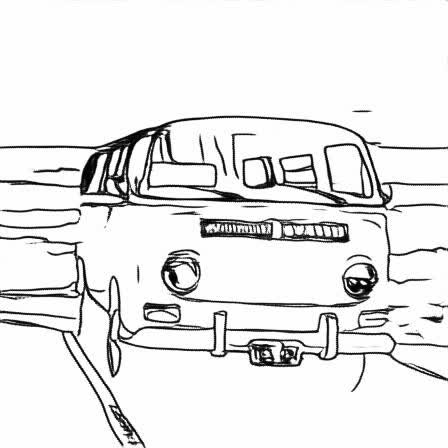} &
        \includegraphics[width=0.10\textwidth]{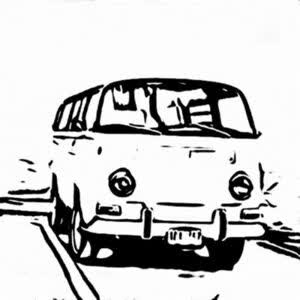} &
        \includegraphics[width=0.10\textwidth]{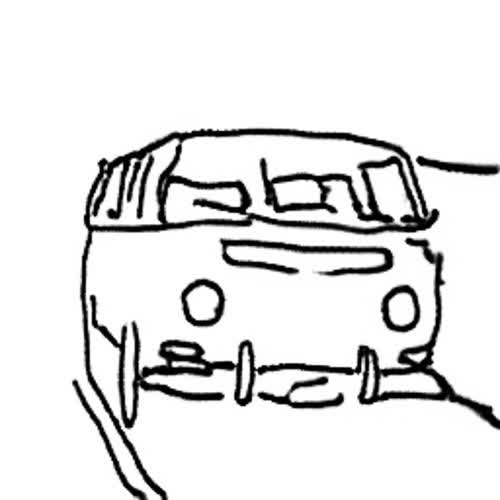} &
        \includegraphics[width=0.10\textwidth]{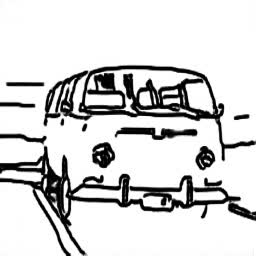} &
        \includegraphics[width=0.10\textwidth]{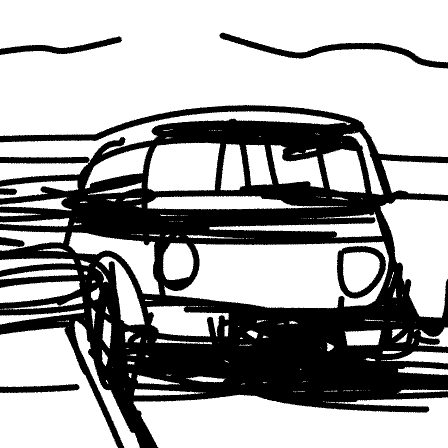} &
        \includegraphics[width=0.10\textwidth]{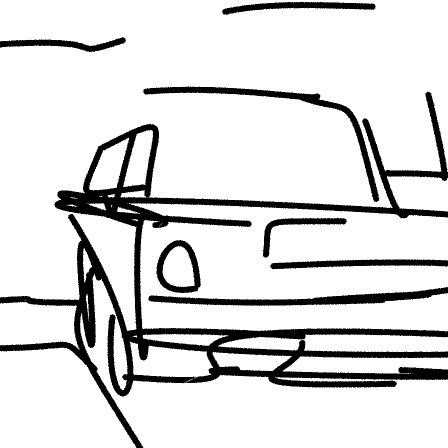} &
        \hspace{0.1cm}
        \includegraphics[width=0.10\textwidth]{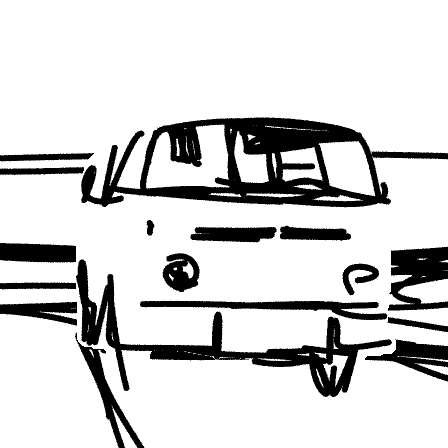} &
        \includegraphics[width=0.10\textwidth]{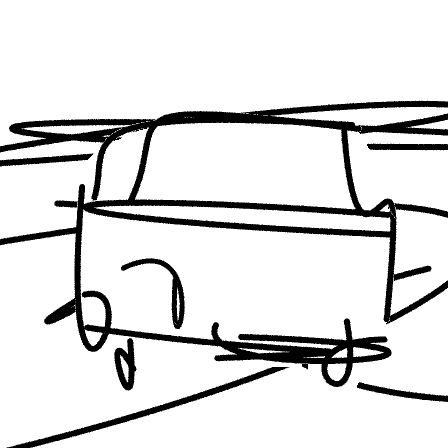} \\

        \includegraphics[width=0.10\textwidth]{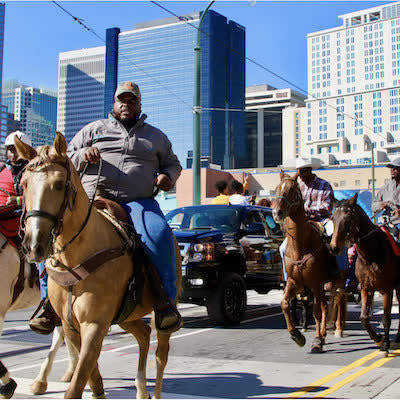} &
        \hspace{0.1cm}
        \includegraphics[width=0.10\textwidth]{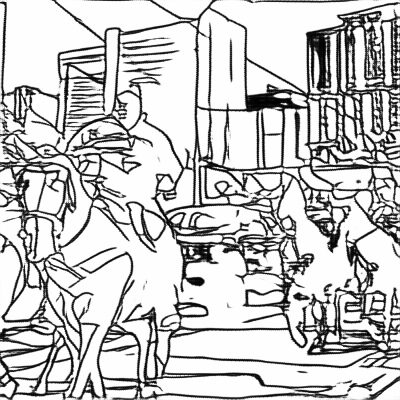} &
        \includegraphics[width=0.10\textwidth]{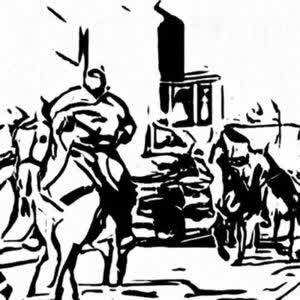} &
        \includegraphics[width=0.10\textwidth]{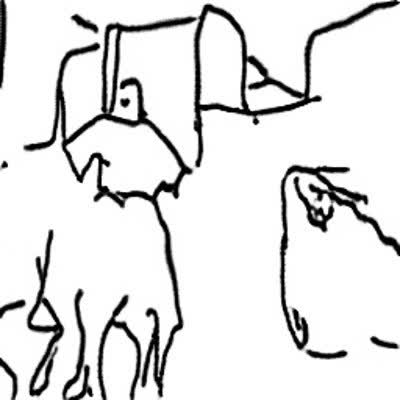} &
        \includegraphics[width=0.10\textwidth]{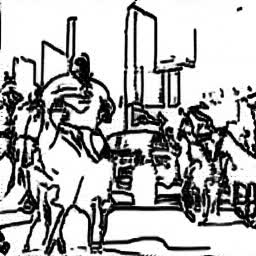} &
        \includegraphics[width=0.10\textwidth]{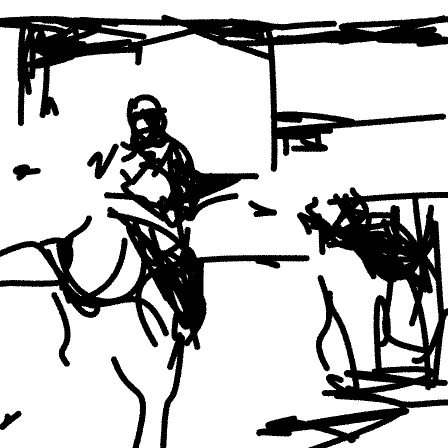} &
        \includegraphics[width=0.10\textwidth]{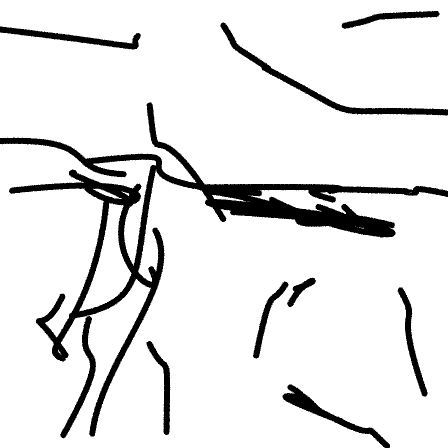} &
        \hspace{0.1cm}
        \includegraphics[width=0.10\textwidth]{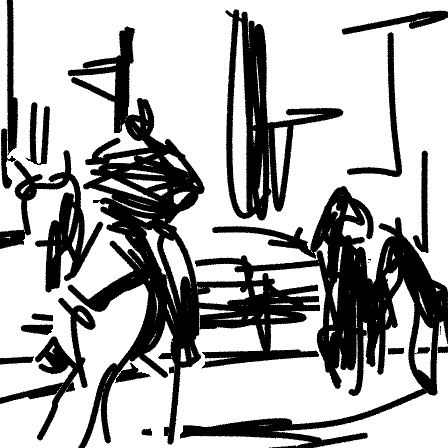} &
        \includegraphics[width=0.10\textwidth]{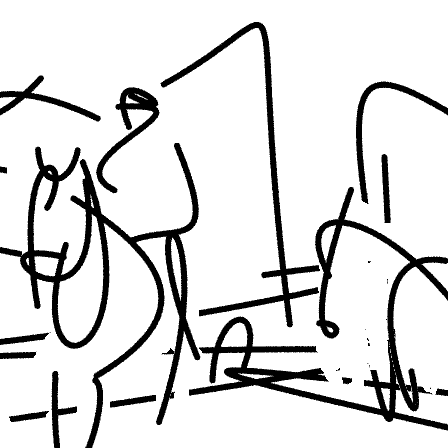} \\

        \includegraphics[width=0.10\textwidth]{figs/inputs/panda.jpg} &
        \hspace{0.1cm}
        \includegraphics[width=0.10\textwidth]{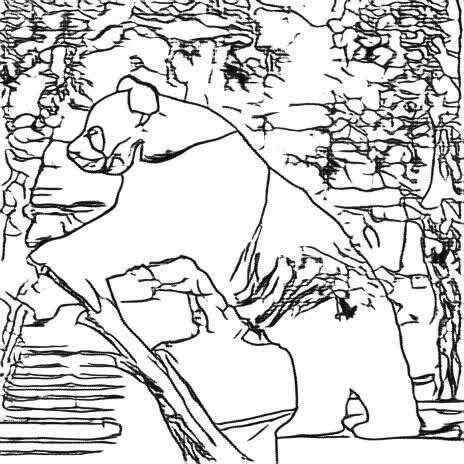} &
        \includegraphics[width=0.10\textwidth]{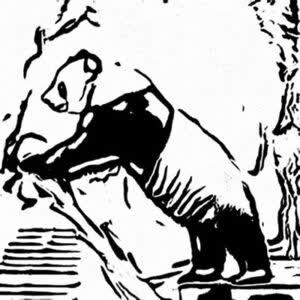} &
        \includegraphics[width=0.10\textwidth]{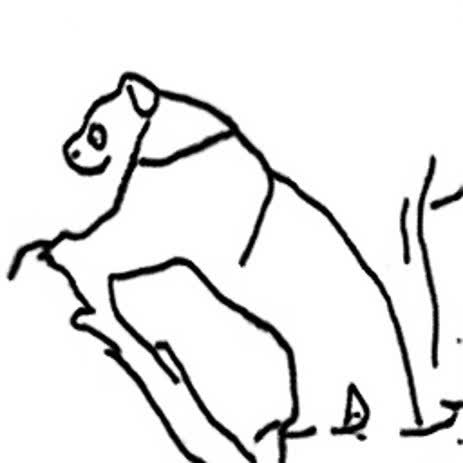} &
        \includegraphics[width=0.10\textwidth]{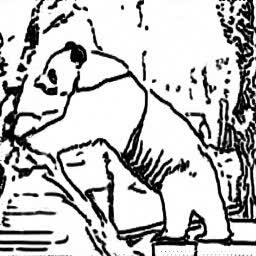} &
        \includegraphics[width=0.10\textwidth]{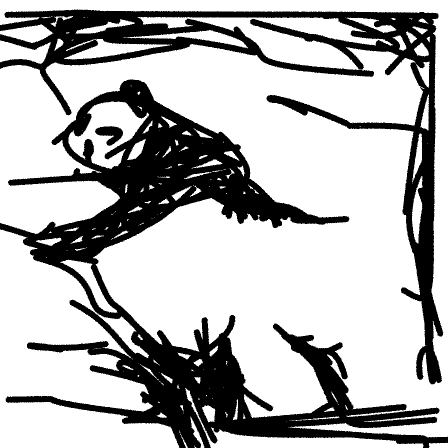} &
        \includegraphics[width=0.10\textwidth]{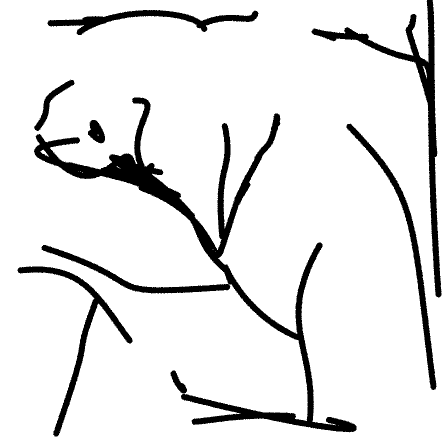} &
        \hspace{0.1cm}
        \includegraphics[width=0.10\textwidth]{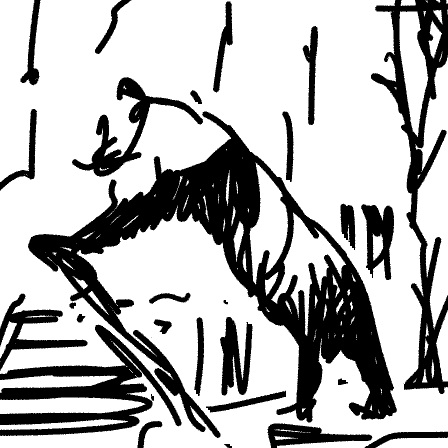} &
        \includegraphics[width=0.10\textwidth]{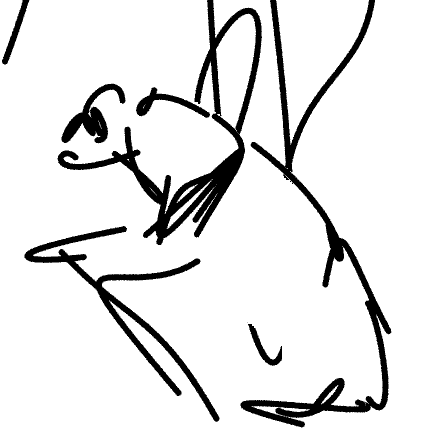} \\

        Input & Edge Extraction & UPDG & Photo-Sketching & Chan~\etal~\shortcite{chan2022learning} & 
        \begin{tabular}{c} CLIPasso \\ (128 Strokes) \end{tabular} &
        \begin{tabular}{c} CLIPasso \\ (32 Strokes) \end{tabular} &
        \begin{tabular}{c} Ours \\ (Layer 2) \end{tabular} &
        \begin{tabular}{c} Ours \\ (Layer 11) \end{tabular} \\
        
    \end{tabular}
    
    }
    \caption{Scene sketching results and comparisons.}
    \label{fig:scene_sketching_comparisons_supp}
\end{figure*}

%% file: files/figures/supplementary/chan_comparison.tex
\begin{figure*}
    \centering
    \setlength{\belowcaptionskip}{-6pt}
    \setlength{\tabcolsep}{1.5pt}
    {\small
    \begin{tabular}{c@{\hspace{0.2cm}} | c c c@{\hspace{0.2cm}} | c c c@{\hspace{0.2cm}} | c c c}

        \includegraphics[width=0.09\textwidth]{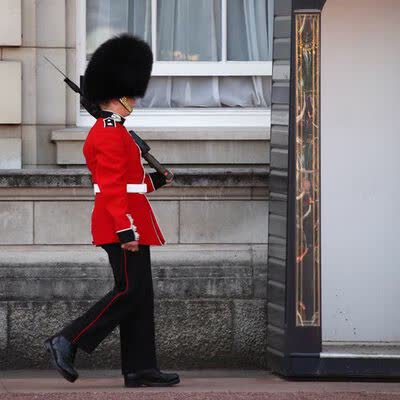} &
        \hspace{0.1cm}
        \includegraphics[width=0.09\textwidth]{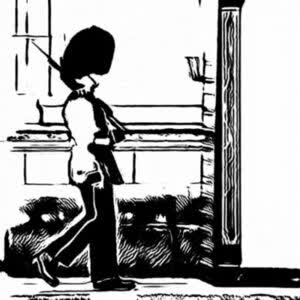} &
        \includegraphics[width=0.09\textwidth]{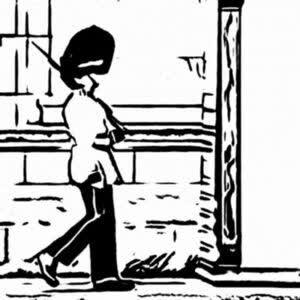} &
        \includegraphics[width=0.09\textwidth]{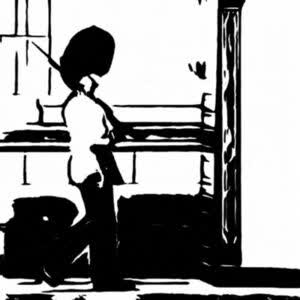} &
        \hspace{0.1cm}
        \includegraphics[width=0.09\textwidth]{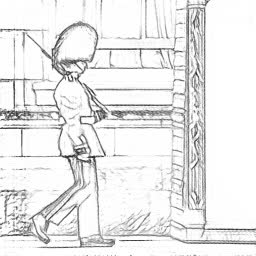} &
        \includegraphics[width=0.09\textwidth]{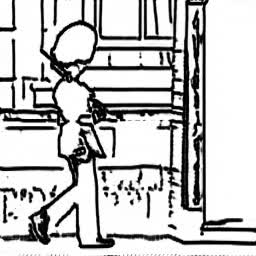} &
        \includegraphics[width=0.09\textwidth]{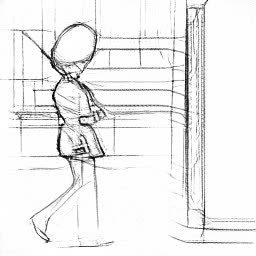} &
        \hspace{0.1cm}
        \includegraphics[width=0.09\textwidth]{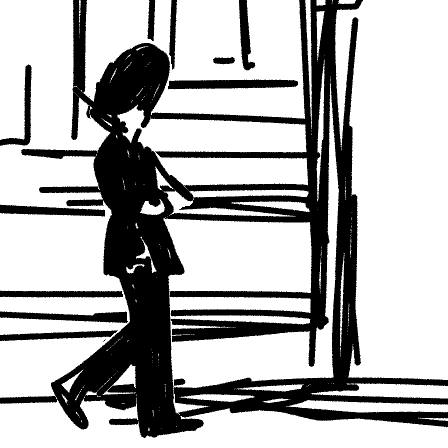} &
        \includegraphics[width=0.09\textwidth]{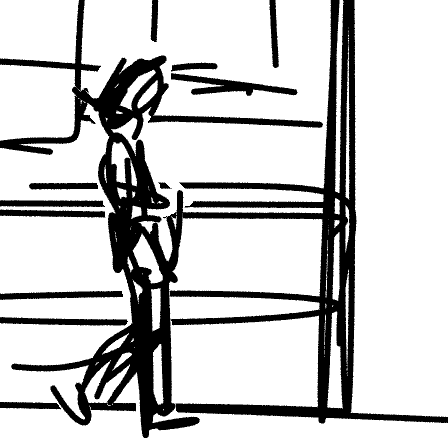} &
        \includegraphics[width=0.09\textwidth]{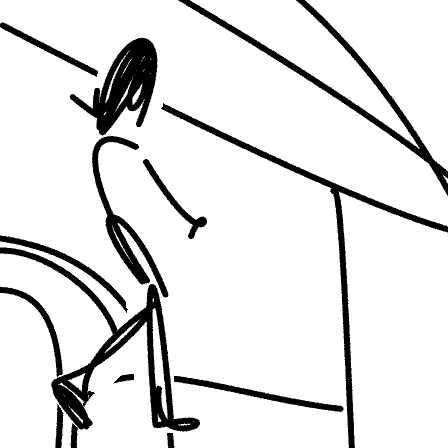} \\

        \includegraphics[width=0.09\textwidth]{figs/inputs/eiffel_tower.jpg} &
        \hspace{0.1cm}
        \includegraphics[width=0.09\textwidth]{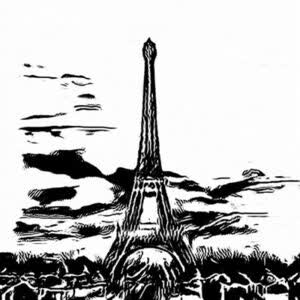} &
        \includegraphics[width=0.09\textwidth]{figs/comparisons/updg/eiffel_tower_fake1.jpg} &
        \includegraphics[width=0.09\textwidth]{figs/comparisons/updg/eiffel_tower_fake1.jpg} &
        \hspace{0.1cm}
        \hspace{0.1cm}
        \includegraphics[width=0.09\textwidth]{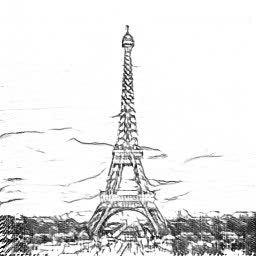} &
        \includegraphics[width=0.09\textwidth]{figs/comparisons/informative_drawings/eiffel_tower_out.jpg} &
        \includegraphics[width=0.09\textwidth]{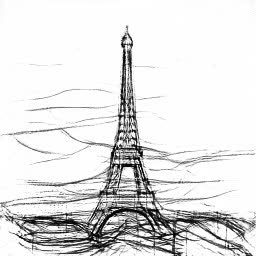} &
        \hspace{0.1cm}
        \includegraphics[width=0.09\textwidth]{figs/matrices_black/eiffel_tower_0.png} &
        \includegraphics[width=0.09\textwidth]{figs/matrices_black/eiffel_tower_9.png} &
        \includegraphics[width=0.09\textwidth]{figs/matrices_black/eiffel_tower_15.png} \\
        
        \includegraphics[width=0.09\textwidth]{figs/inputs/dog.jpg} &
        \hspace{0.1cm}
        \includegraphics[width=0.09\textwidth]{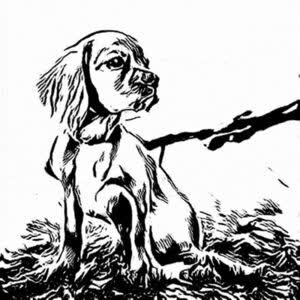} &
        \includegraphics[width=0.09\textwidth]{figs/comparisons/updg/dog_fake2.jpg} &
        \includegraphics[width=0.09\textwidth]{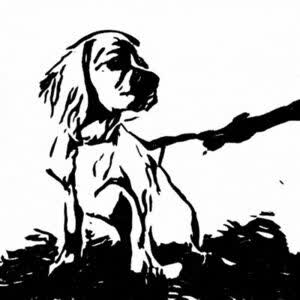} &
        \hspace{0.1cm}
        \includegraphics[width=0.09\textwidth]{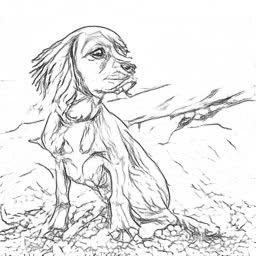} &
        \includegraphics[width=0.09\textwidth]{figs/comparisons/informative_drawings/dog_out.jpg} &
        \includegraphics[width=0.09\textwidth]{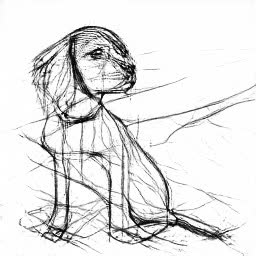} &
        \hspace{0.1cm}
        \includegraphics[width=0.09\textwidth]{figs/matrices_black/dog_row0col0_black.png} &
        \includegraphics[width=0.09\textwidth]{figs/matrices_black/dog_row2col2_black.png} &
        \includegraphics[width=0.09\textwidth]{figs/matrices_black/dog_row3col3_black.png} \\
        
        \includegraphics[width=0.09\textwidth]{figs/inputs/lighthouse-2.jpg} &
        \hspace{0.1cm}
        \includegraphics[width=0.09\textwidth]{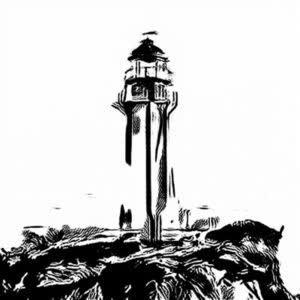} &
        \includegraphics[width=0.09\textwidth]{figs/comparisons/updg/lighthouse-2_fake2.jpg} &
        \includegraphics[width=0.09\textwidth]{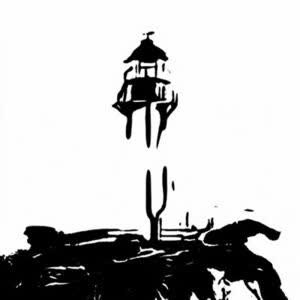} &
        \hspace{0.1cm}
        \includegraphics[width=0.09\textwidth]{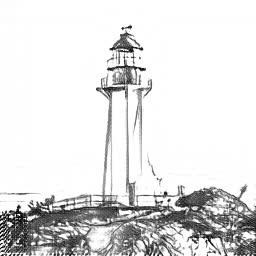} &
        \includegraphics[width=0.09\textwidth]{figs/comparisons/informative_drawings/lighthouse-2_out.jpg} &
        \includegraphics[width=0.09\textwidth]{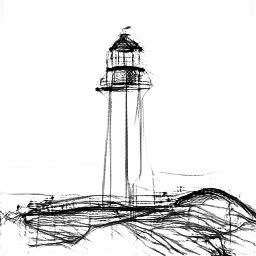} &
        \hspace{0.1cm}
        \includegraphics[width=0.09\textwidth]{figs/matrices_black/lighthouse-2_row0col0_black.png} &
        \includegraphics[width=0.09\textwidth]{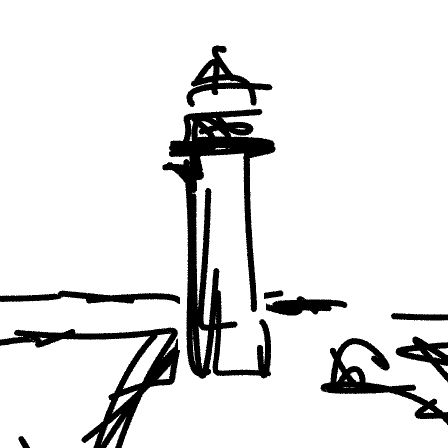} &
        \includegraphics[width=0.09\textwidth]{figs/matrices_black/lighthouse-2_row3col3_black.png} \\

        \includegraphics[width=0.09\textwidth]{figs/inputs/trevi.jpg} &
        \hspace{0.1cm}
        \includegraphics[width=0.09\textwidth]{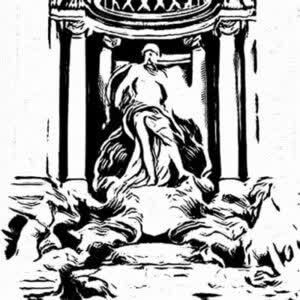} &
        \includegraphics[width=0.09\textwidth]{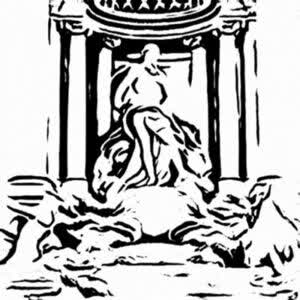} &
        \includegraphics[width=0.09\textwidth]{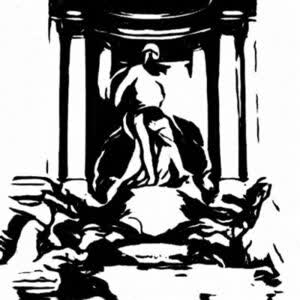} &
        \hspace{0.1cm}
        \includegraphics[width=0.09\textwidth]{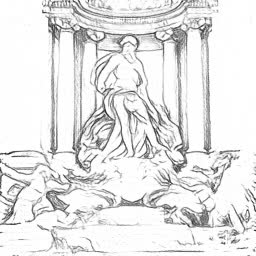} &
        \includegraphics[width=0.09\textwidth]{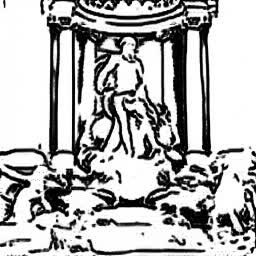} &
        \includegraphics[width=0.09\textwidth]{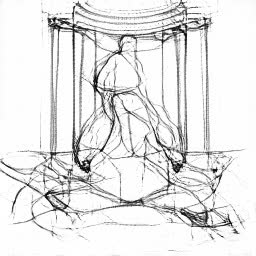} &
        \hspace{0.1cm}
        \includegraphics[width=0.09\textwidth]{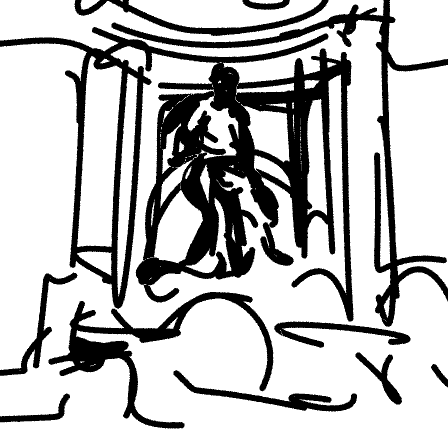} &
        \includegraphics[width=0.09\textwidth]{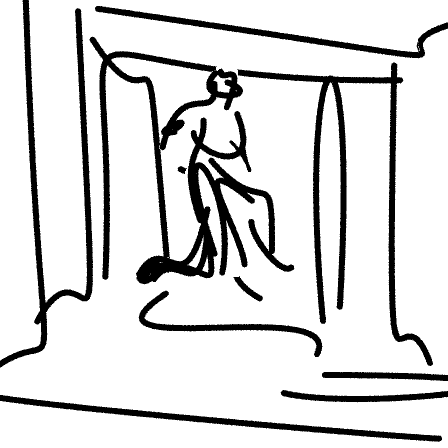} &
        \includegraphics[width=0.09\textwidth]{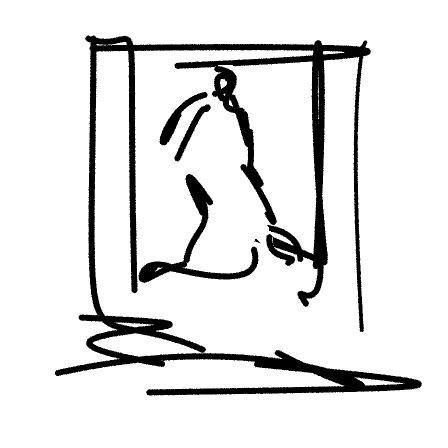} \\

        \includegraphics[width=0.09\textwidth]{figs/inputs/bull.jpg} &
        \hspace{0.1cm}
        \includegraphics[width=0.09\textwidth]{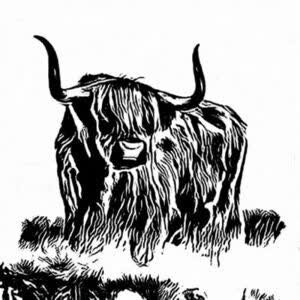} &
        \includegraphics[width=0.09\textwidth]{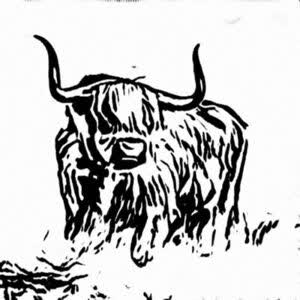} &
        \includegraphics[width=0.09\textwidth]{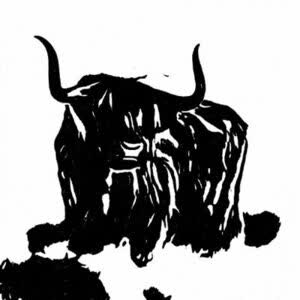} &
        \hspace{0.1cm}
        \includegraphics[width=0.09\textwidth]{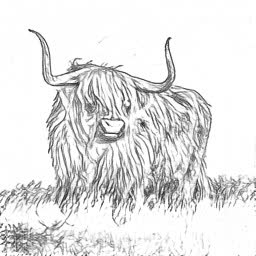} &
        \includegraphics[width=0.09\textwidth]{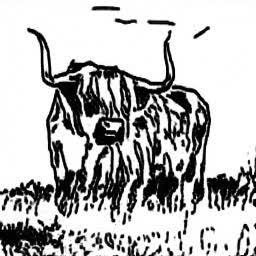} &
        \includegraphics[width=0.09\textwidth]{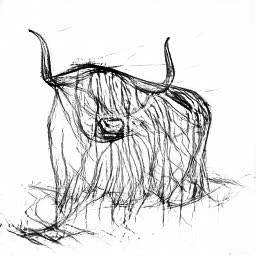} &
        \hspace{0.1cm}
        \includegraphics[width=0.09\textwidth]{figs/matrices_black/bull_row0col0_black.png} &
        \includegraphics[width=0.09\textwidth]{figs/matrices_black/bull_row2col2_black.png} &
        \includegraphics[width=0.09\textwidth]{figs/matrices_black/bull_row3col3_black.png} \\

        Input & 
        \multicolumn{3}{c}{ UPDG~\cite{yi2020unpaired} } &
        \multicolumn{3}{c}{ Chan~\etal~\shortcite{chan2022learning} } &
        \multicolumn{3}{c}{ Ours } 
        
    \end{tabular}
    
    }
    \caption{Scene sketching comparison to Chan~\etal~\cite{chan2022learning} and UPDG~\cite{yi2020unpaired} across three different styles supported by each of their methods. For our results, we show three sketches illustrating the various levels of abstraction that our method is capable of achieving.}
    \label{fig:scene_sketching_comparisons_chan_supp}
\end{figure*}

%% file: files/figures/supplementary/clipasso_comparison.tex
\begin{figure*}
    \centering
    \setlength{\tabcolsep}{2pt}
    
    \begin{tabular}{c}
    
    \begin{tabular}{c c c c c c c c c}

    \includegraphics[width=0.095\linewidth,height=0.095\linewidth]{figs/inputs/lion.jpg} &
    \includegraphics[width=0.095\linewidth,height=0.095\linewidth]{figs/inputs/lion.jpg} &
    \includegraphics[width=0.095\linewidth,height=0.095\linewidth]{figs/inputs/lion.jpg} &
    \hspace{0.01\linewidth}
    \includegraphics[width=0.095\linewidth,height=0.095\linewidth]{figs/inputs/ballerina.jpg} &
    \includegraphics[width=0.095\linewidth,height=0.095\linewidth]{figs/inputs/ballerina.jpg} &
    \includegraphics[width=0.095\linewidth,height=0.095\linewidth]{figs/inputs/ballerina.jpg} &
    \hspace{0.01\linewidth}
    \includegraphics[width=0.095\linewidth,height=0.095\linewidth]{figs/inputs/black_man.jpg} &
    \includegraphics[width=0.095\linewidth,height=0.095\linewidth]{figs/inputs/black_man.jpg} &
    \includegraphics[width=0.095\linewidth,height=0.095\linewidth]{figs/inputs/black_man.jpg} \\

    \includegraphics[width=0.095\linewidth,height=0.095\linewidth]{figs/comparisons/clipasso/lion_128_strokes_seed_1000_best.png} &
    \includegraphics[width=0.095\linewidth,height=0.095\linewidth]{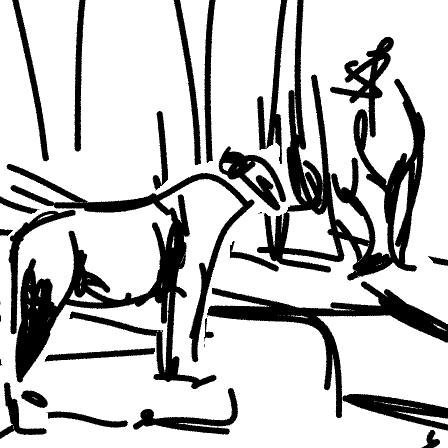} &
    \includegraphics[width=0.095\linewidth,height=0.095\linewidth]{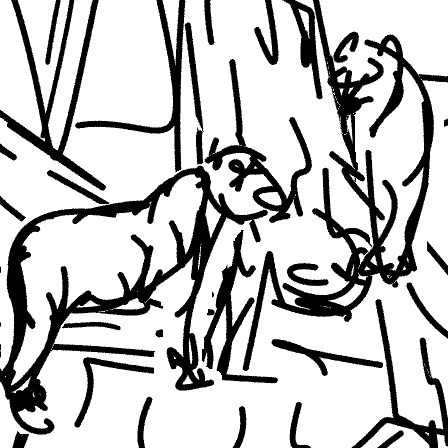} &
    \hspace{0.01\linewidth}
    \includegraphics[width=0.095\linewidth,height=0.095\linewidth]{figs/comparisons/clipasso/ballerina_128_strokes_seed_2000_best.png} &
    \includegraphics[width=0.095\linewidth,height=0.095\linewidth]{figs/matrices_black/ballerina_row0col0_black.png} &
    \includegraphics[width=0.095\linewidth,height=0.095\linewidth]{figs/matrices_black/ballerina_row0col3_black.png} &
    \hspace{0.01\linewidth}
    \includegraphics[width=0.095\linewidth,height=0.095\linewidth]{figs/comparisons/clipasso/black_man_128_strokes_seed_0_best.png} &
    \includegraphics[width=0.095\linewidth,height=0.095\linewidth]{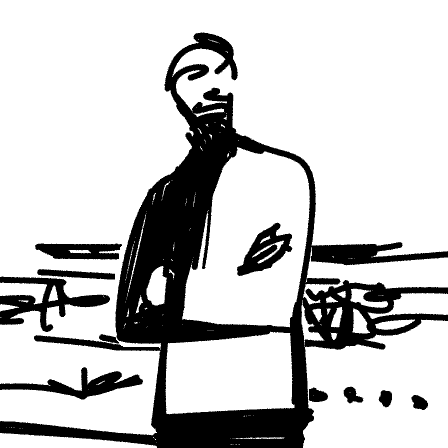} &
    \includegraphics[width=0.095\linewidth,height=0.095\linewidth]{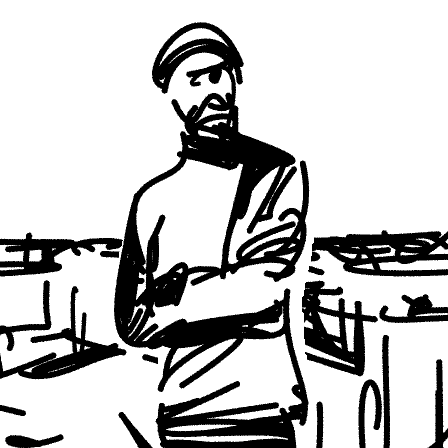} \\

    \includegraphics[width=0.095\linewidth,height=0.095\linewidth]{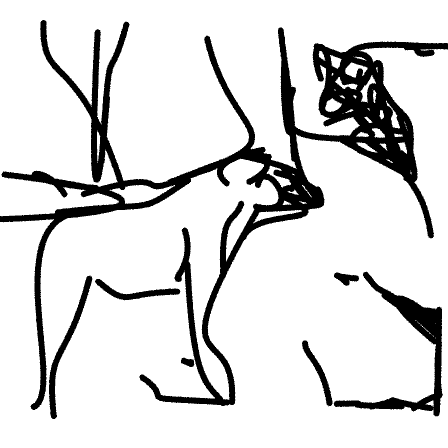} &
    \includegraphics[width=0.095\linewidth,height=0.095\linewidth]{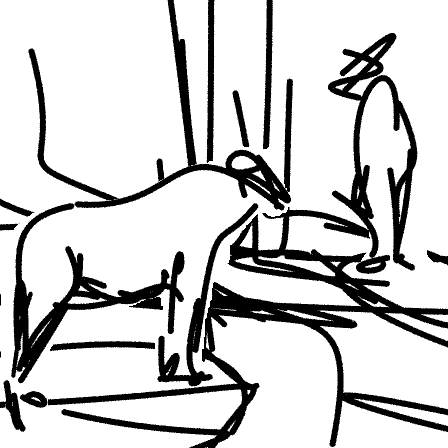} &
    \includegraphics[width=0.095\linewidth,height=0.095\linewidth]{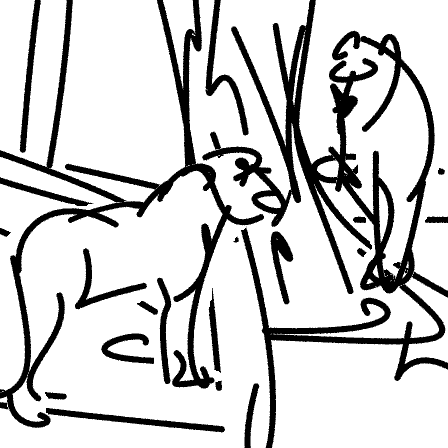} &
    \hspace{0.01\linewidth}
    \includegraphics[width=0.095\linewidth,height=0.095\linewidth]{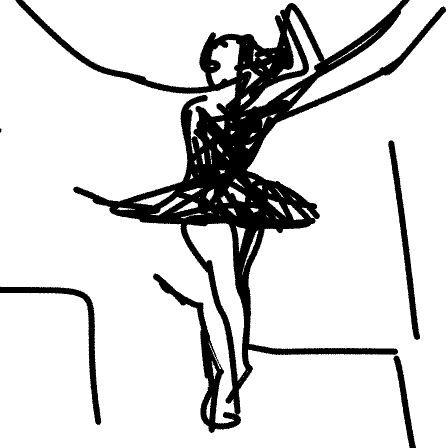} &
    \includegraphics[width=0.095\linewidth,height=0.095\linewidth]{figs/matrices_black/ballerina_row1col0_black.png} &
    \includegraphics[width=0.095\linewidth,height=0.095\linewidth]{figs/matrices_black/ballerina_row1col3_black.png} &
    \hspace{0.01\linewidth}
    \includegraphics[width=0.095\linewidth,height=0.095\linewidth]{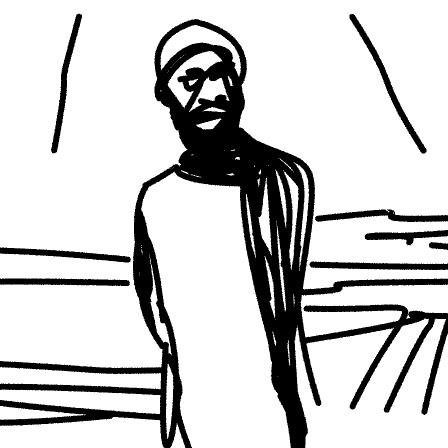} &
    \includegraphics[width=0.095\linewidth,height=0.095\linewidth]{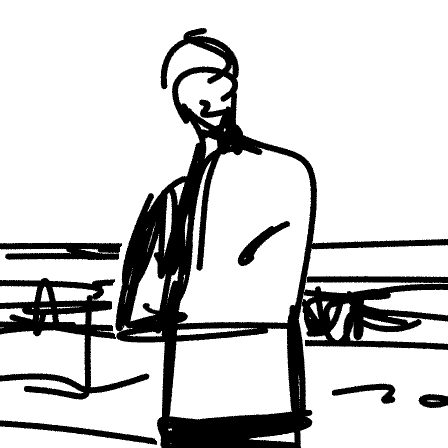} &
    \includegraphics[width=0.095\linewidth,height=0.095\linewidth]{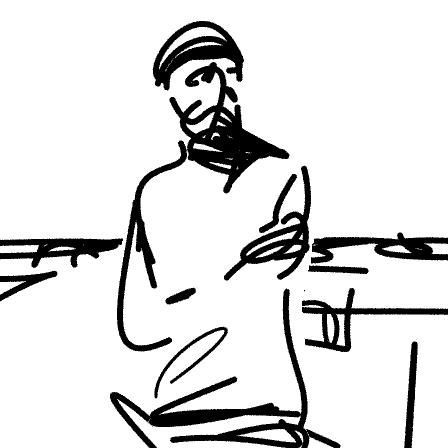} \\

    \includegraphics[width=0.095\linewidth,height=0.095\linewidth]{figs/comparisons/clipasso/lion_32_strokes_seed_2000_best.png} &
    \includegraphics[width=0.095\linewidth,height=0.095\linewidth]{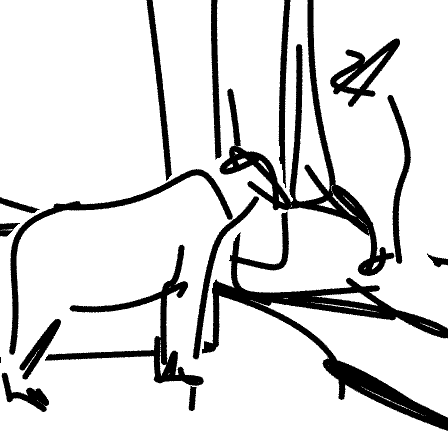} &
    \includegraphics[width=0.095\linewidth,height=0.095\linewidth]{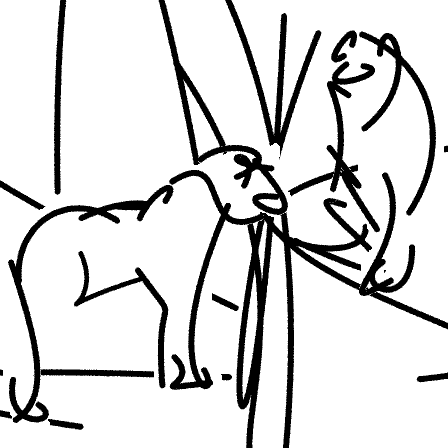} &
    \hspace{0.01\linewidth}
    \includegraphics[width=0.095\linewidth,height=0.095\linewidth]{figs/comparisons/clipasso/ballerina_32_strokes_seed_2000_best.png} &
    \includegraphics[width=0.095\linewidth,height=0.095\linewidth]{figs/matrices_black/ballerina_row2col0_black.png} &
    \includegraphics[width=0.095\linewidth,height=0.095\linewidth]{figs/matrices_black/ballerina_row2col3_black.png} &
    \hspace{0.01\linewidth}
    \includegraphics[width=0.095\linewidth,height=0.095\linewidth]{figs/comparisons/clipasso/black_man_32_strokes_seed_0_best.png} &
    \includegraphics[width=0.095\linewidth,height=0.095\linewidth]{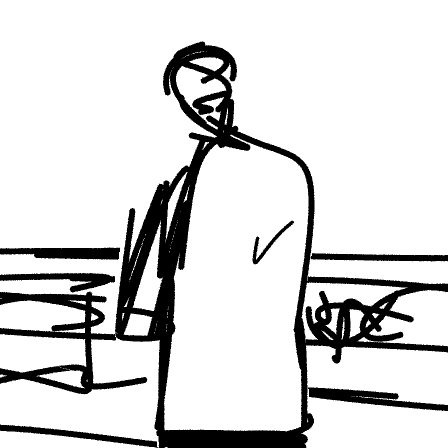} &
    \includegraphics[width=0.095\linewidth,height=0.095\linewidth]{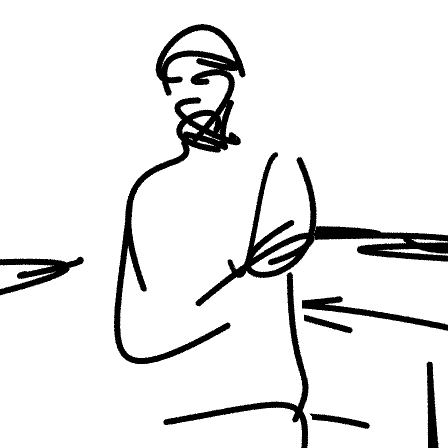} \\

    \includegraphics[width=0.095\linewidth,height=0.095\linewidth]{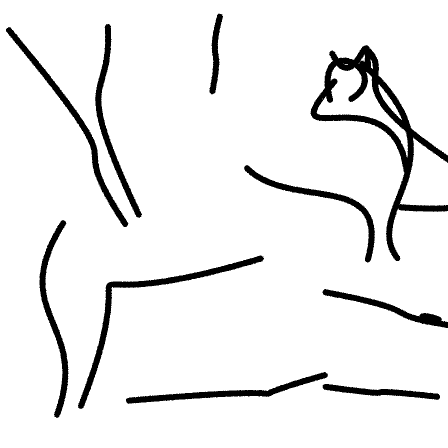} &
    \includegraphics[width=0.095\linewidth,height=0.095\linewidth]{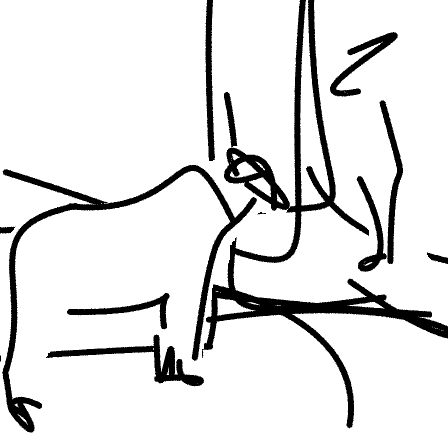} &
    \includegraphics[width=0.095\linewidth,height=0.095\linewidth]{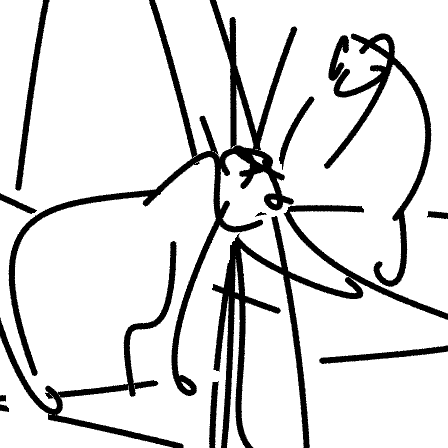} &
    \hspace{0.01\linewidth}
    \includegraphics[width=0.095\linewidth,height=0.095\linewidth]{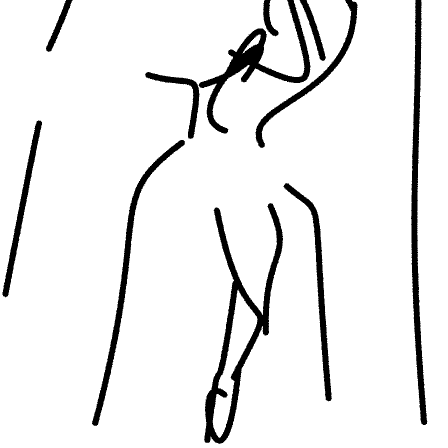} &
    \includegraphics[width=0.095\linewidth,height=0.095\linewidth]{figs/matrices_black/ballerina_row3col0_black.png} &
    \includegraphics[width=0.095\linewidth,height=0.095\linewidth]{figs/matrices_black/ballerina_row3col3_black.png} &
    \hspace{0.01\linewidth}
    \includegraphics[width=0.095\linewidth,height=0.095\linewidth]{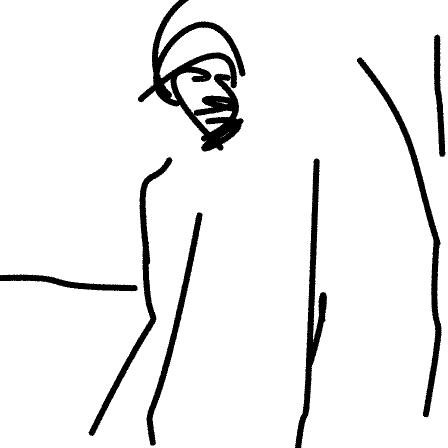} &
    \includegraphics[width=0.095\linewidth,height=0.095\linewidth]{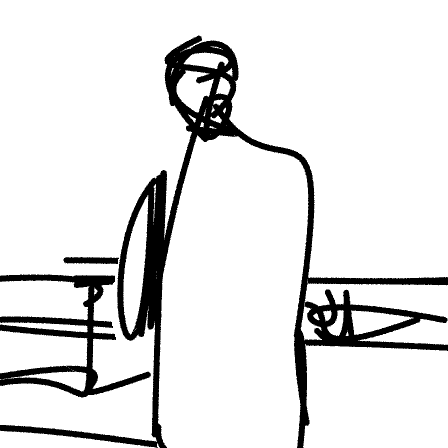} &
    \includegraphics[width=0.095\linewidth,height=0.095\linewidth]{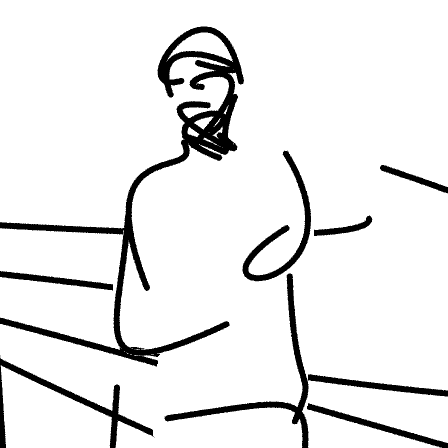} \\

    CLIPasso &
    \begin{tabular}{c} Ours \\ (Layer 2) \end{tabular} &
    
    \begin{tabular}{c} Ours \\ (Layer 11) \end{tabular} &
    \hspace{0.01\linewidth}
    CLIPasso &
    \begin{tabular}{c} Ours \\ (Layer 2) \end{tabular} &
    \begin{tabular}{c} Ours \\ (Layer 11) \end{tabular} &
    CLIPasso &
    \begin{tabular}{c} Ours \\ (Layer 2) \end{tabular} &
    \begin{tabular}{c} Ours \\ (Layer 11) \end{tabular}

    \end{tabular}
    
    \end{tabular}
    
    \begin{tabular}{c}
    
    \begin{tabular}{c c c c c c}
    
    \\ \\

    \includegraphics[width=0.095\linewidth,height=0.095\linewidth]{figs/inputs/bull.jpg} &
    \includegraphics[width=0.095\linewidth,height=0.095\linewidth]{figs/inputs/bull.jpg} &
    \includegraphics[width=0.095\linewidth,height=0.095\linewidth]{figs/inputs/bull.jpg} &
    \hspace{0.01\linewidth}
    \includegraphics[width=0.095\linewidth,height=0.095\linewidth]{figs/inputs/black_man.jpg} &
    \includegraphics[width=0.095\linewidth,height=0.095\linewidth]{figs/inputs/black_man.jpg} &
    \includegraphics[width=0.095\linewidth,height=0.095\linewidth]{figs/inputs/black_man.jpg} \\

    \includegraphics[width=0.095\linewidth,height=0.095\linewidth]{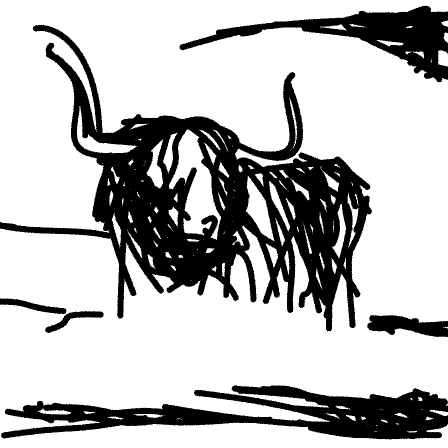} &
    \includegraphics[width=0.095\linewidth,height=0.095\linewidth]{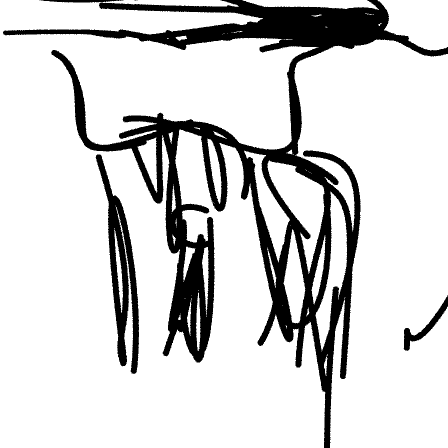} &
    \includegraphics[width=0.095\linewidth,height=0.095\linewidth]{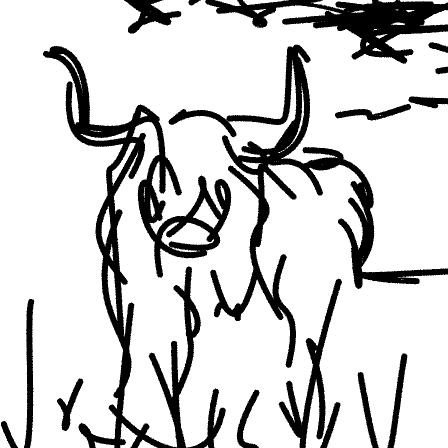} &
    \hspace{0.01\linewidth}
    \includegraphics[width=0.095\linewidth,height=0.095\linewidth]{figs/comparisons/clipasso/black_man_128_strokes_seed_0_best.png} &
    \includegraphics[width=0.095\linewidth,height=0.095\linewidth]{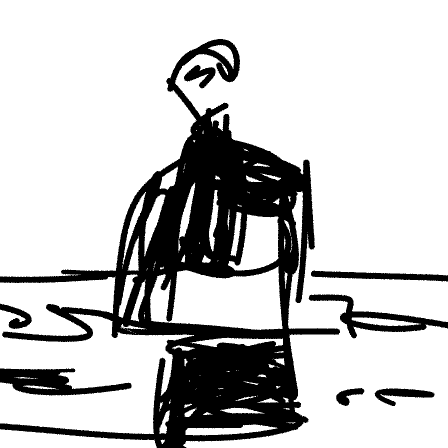} &
    \includegraphics[width=0.095\linewidth,height=0.095\linewidth]{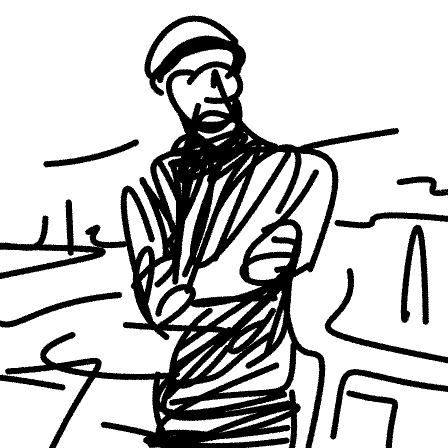} \\

    \includegraphics[width=0.095\linewidth,height=0.095\linewidth]{figs/comparisons/clipasso/objects/bull_64_strokes_seed_2000_best.png} &
    \includegraphics[width=0.095\linewidth,height=0.095\linewidth]{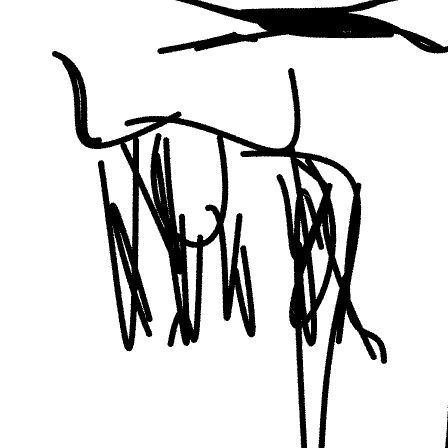} &
    \includegraphics[width=0.095\linewidth,height=0.095\linewidth]{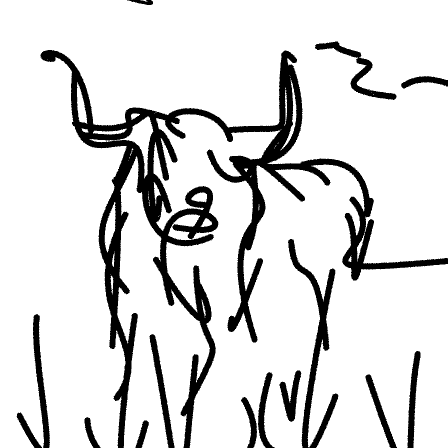} &
    \hspace{0.01\linewidth}
    \includegraphics[width=0.095\linewidth,height=0.095\linewidth]{figs/comparisons/clipasso/black_man_64_strokes_seed_1000_best.png} &
    \includegraphics[width=0.095\linewidth,height=0.095\linewidth]{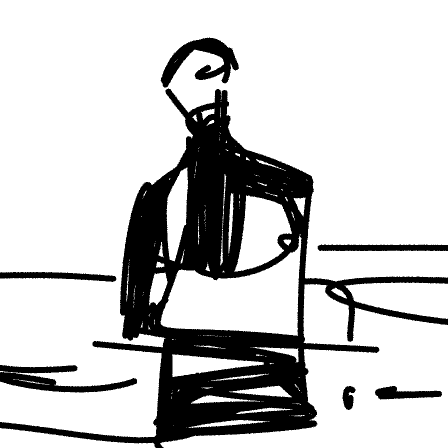} &
    \includegraphics[width=0.095\linewidth,height=0.095\linewidth]{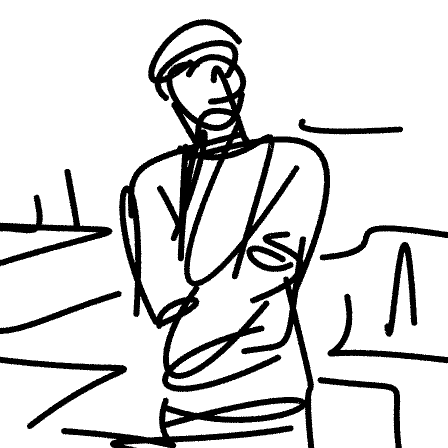} \\

    \includegraphics[width=0.095\linewidth,height=0.095\linewidth]{figs/comparisons/clipasso/objects/bull_32_strokes_seed_0_best.png} &
    \includegraphics[width=0.095\linewidth,height=0.095\linewidth]{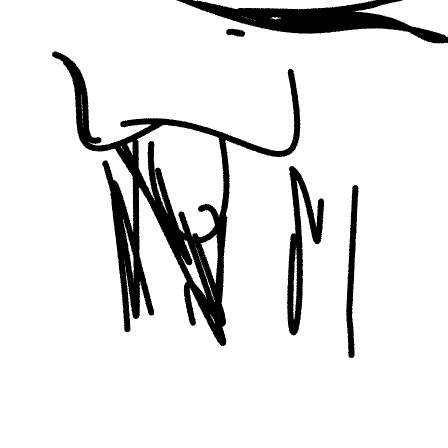} &
    \includegraphics[width=0.095\linewidth,height=0.095\linewidth]{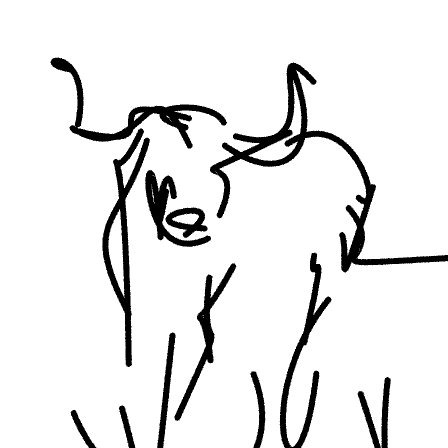} &
    \hspace{0.01\linewidth}
    \includegraphics[width=0.095\linewidth,height=0.095\linewidth]{figs/comparisons/clipasso/black_man_32_strokes_seed_0_best.png} &
    \includegraphics[width=0.095\linewidth,height=0.095\linewidth]{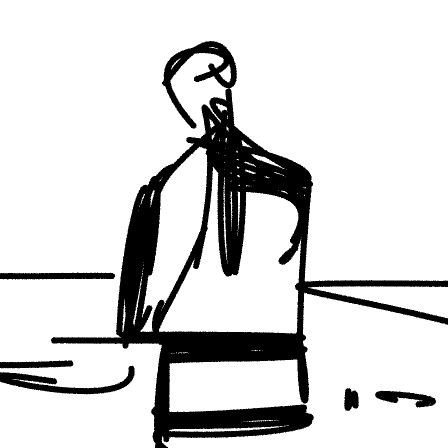} &
    \includegraphics[width=0.095\linewidth,height=0.095\linewidth]{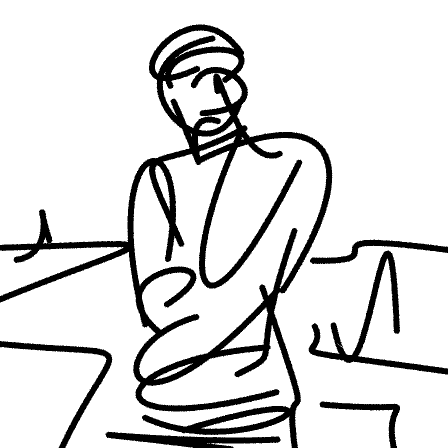} \\

    \includegraphics[width=0.095\linewidth,height=0.095\linewidth]{figs/comparisons/clipasso/objects/bull_16_strokes_seed_1000_best.png} &
    \includegraphics[width=0.095\linewidth,height=0.095\linewidth]{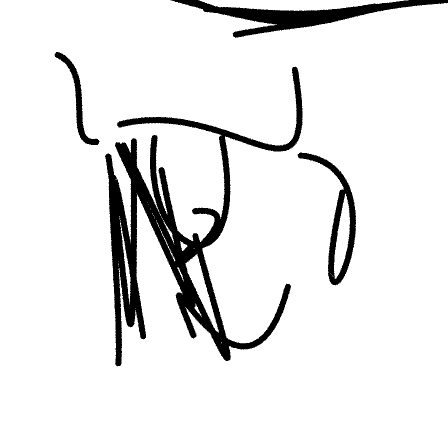} &
    \includegraphics[width=0.095\linewidth,height=0.095\linewidth]{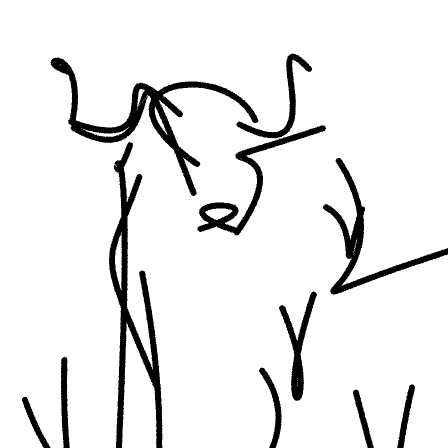} &
    \hspace{0.01\linewidth}
    \includegraphics[width=0.095\linewidth,height=0.095\linewidth]{figs/comparisons/clipasso/black_man_16_strokes_seed_1000_best.png} &
    \includegraphics[width=0.095\linewidth,height=0.095\linewidth]{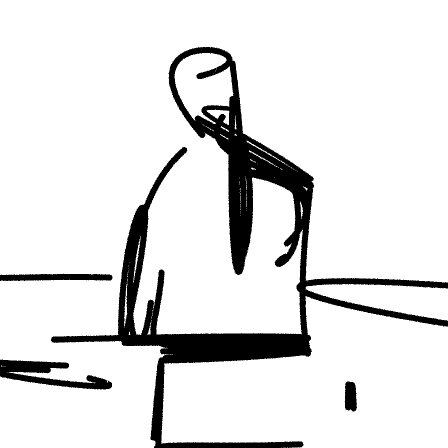} &
    \includegraphics[width=0.095\linewidth,height=0.095\linewidth]{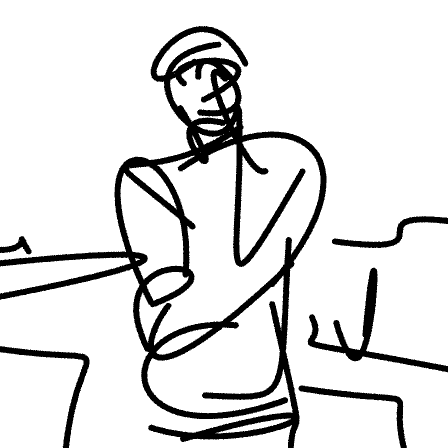} \\

    CLIPasso &
    \begin{tabular}{c} Ours \\ (Layer 2) \end{tabular} &
    
    \begin{tabular}{c} Ours \\ (Layer 11) \end{tabular} &
    \hspace{0.01\linewidth}
    CLIPasso &
    \begin{tabular}{c} Ours \\ (Layer 2) \end{tabular} &
    \begin{tabular}{c} Ours \\ (Layer 11) \end{tabular}
    
    \end{tabular}
    
    \end{tabular}
    
    \caption{Comparison to CLIPasso. For CLIPasso, we generate the sketches using 128, 64, 32, and 16 strokes. In the top set of results, we use our scene decomposition technique and apply our implicit simplification starting with 64 strokes for the foreground and background sketches. For the bottom set of results, we do not use the scene decomposition approach and start with simplification using $128$ strokes. We show our results for layers 2 and 11.}
    \label{fig:clipasso_comparisons_supp}
    
\end{figure*}